\documentclass[10pt,twocolumn,letterpaper]{article}

\usepackage{iccv}
\usepackage{times}
\usepackage{epsfig}
\usepackage{graphicx}
\usepackage{amsmath}
\usepackage{amssymb}

\usepackage[accsupp]{axessibility}

\usepackage{booktabs}
\usepackage{caption}
\usepackage{subcaption}
\usepackage{amsfonts}

\usepackage{algorithm}
\usepackage{algorithmic}
\usepackage{bbm}
\usepackage{color}

\usepackage{multirow}
\usepackage{multicol}
\usepackage{float}
\usepackage{pifont}
\newcommand{\cmark}{\ding{51}}%
\newcommand{\xmark}{\ding{55}}%
\usepackage{bm}
\usepackage{listings}

\usepackage[pagebackref=true,breaklinks=true,letterpaper=true,colorlinks,bookmarks=false]{hyperref}

\iccvfinalcopy 

\def\iccvPaperID{5366} 

\ificcvfinal\fi

\begin{document}

\title{Improving Sample Quality of Diffusion Models Using Self-Attention Guidance}

\author{
Susung Hong
\and
Gyuseong Lee
\and
Wooseok Jang
\and
Seungryong Kim
\and
Korea University, Seoul, Korea\\
{\tt\small \{susung1999, jpl358, jws1997, seungryong\_kim\}@korea.ac.kr}
}


\twocolumn[{
\maketitle
\ificcvfinal\thispagestyle{empty}\fi
\begin{figure}[H]
\centering
\hsize=\textwidth
\captionsetup[subfigure]{labelformat=empty}
\begin{subfigure}{.497\textwidth}
\centering
\lineskip=0pt
      \includegraphics[width=0.33\linewidth]{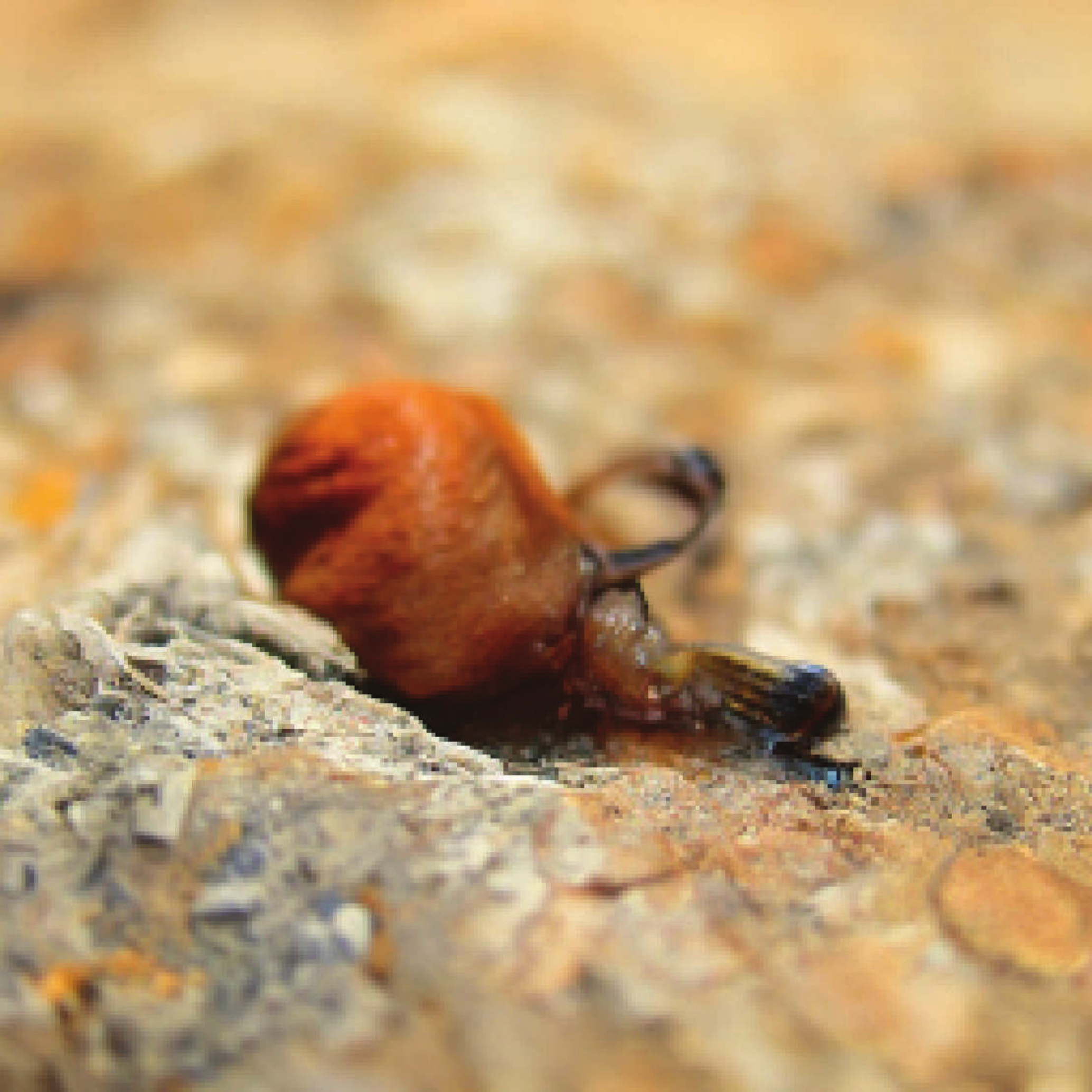}\hspace{-0.25em}
      \includegraphics[width=0.33\linewidth]{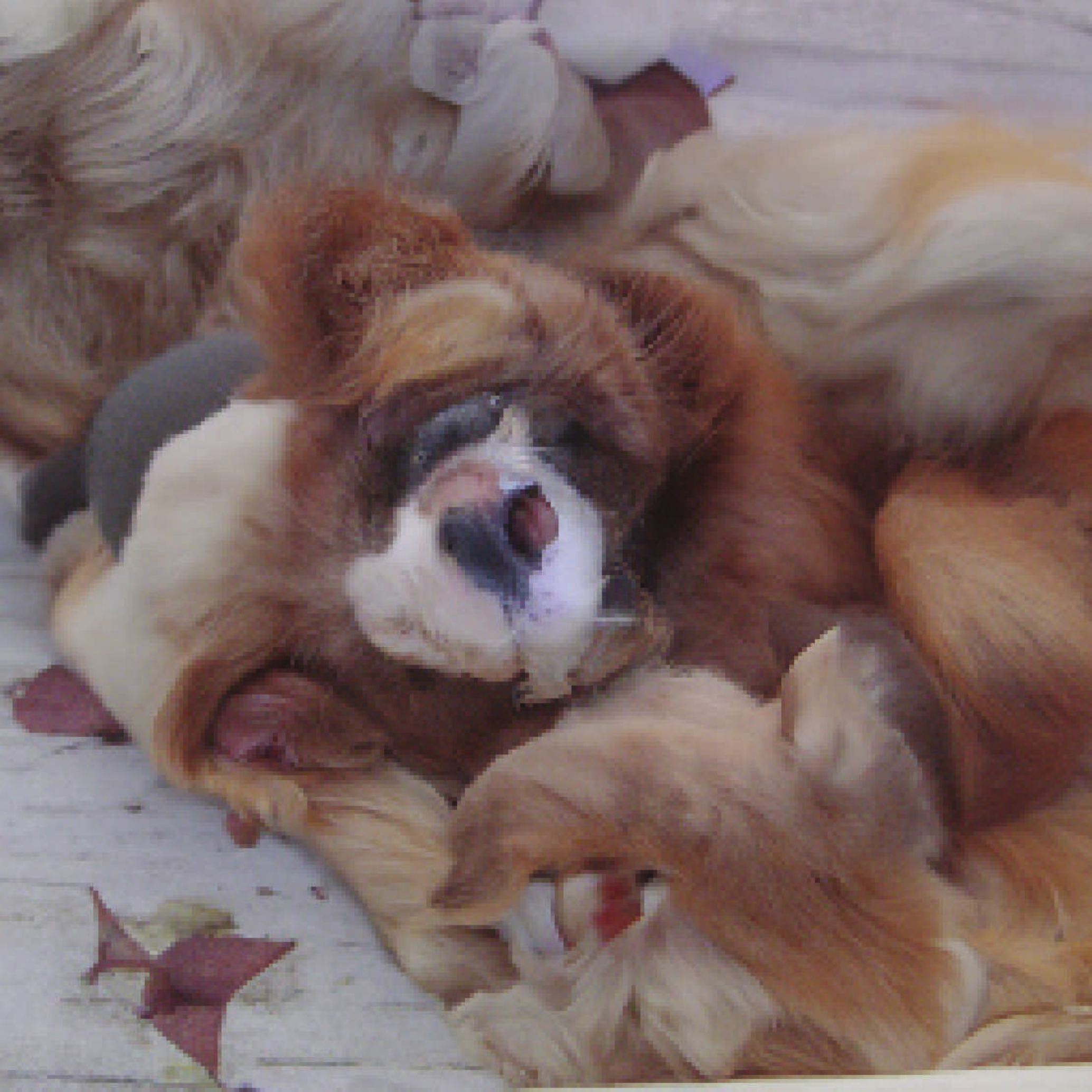}\hspace{-0.25em}
      \includegraphics[width=0.33\linewidth]{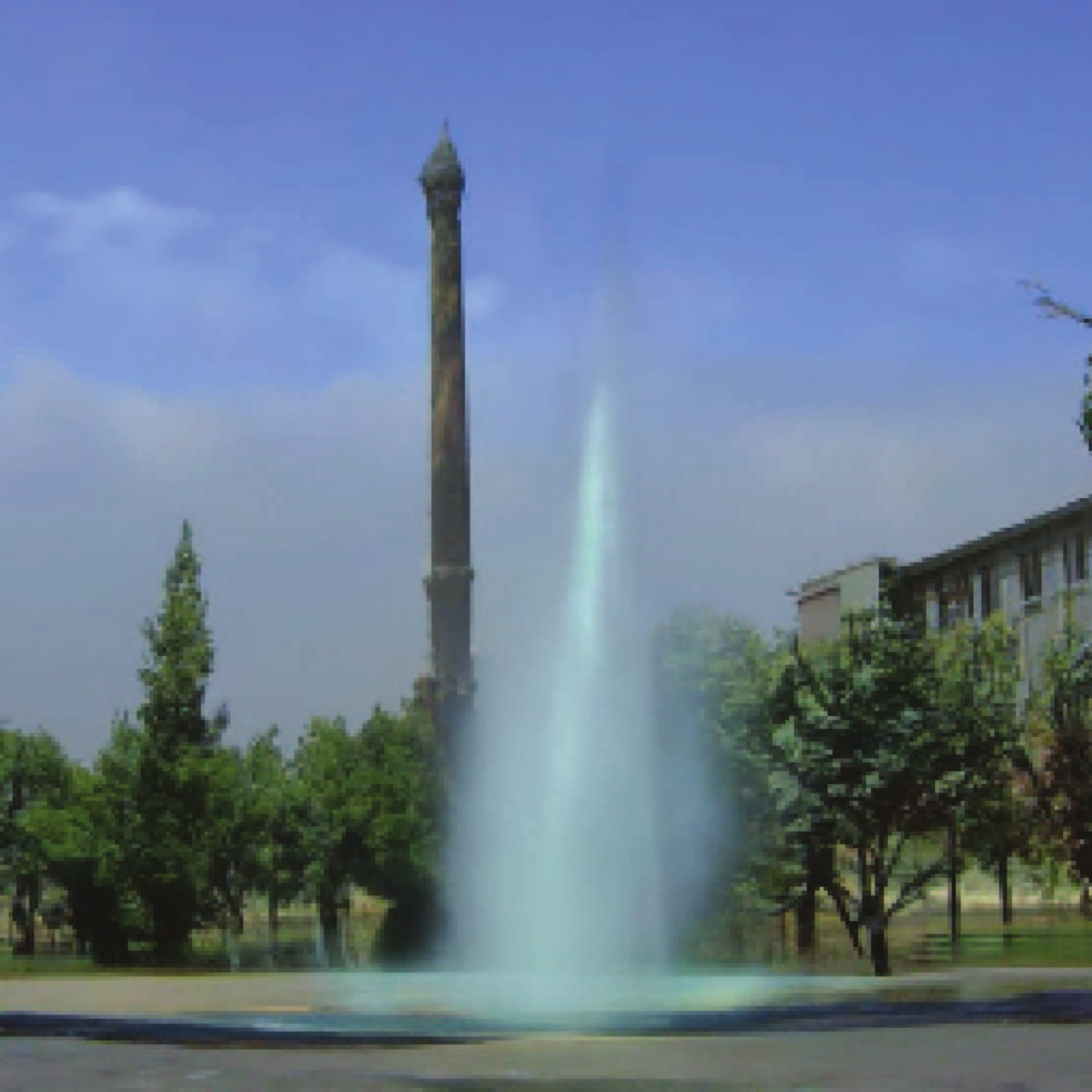}\\
      \includegraphics[width=0.33\linewidth]{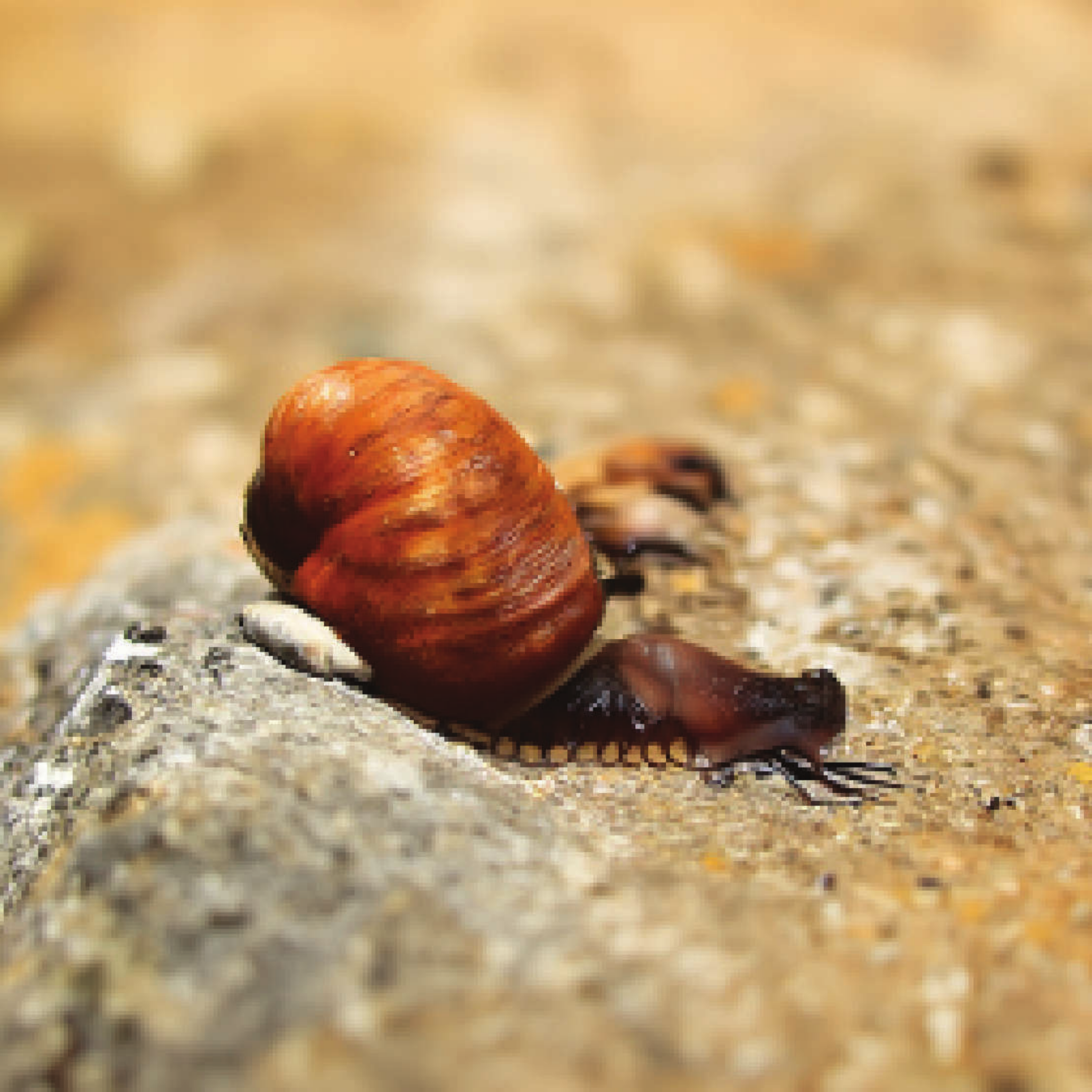}\hspace{-0.25em}
      \includegraphics[width=0.33\linewidth]{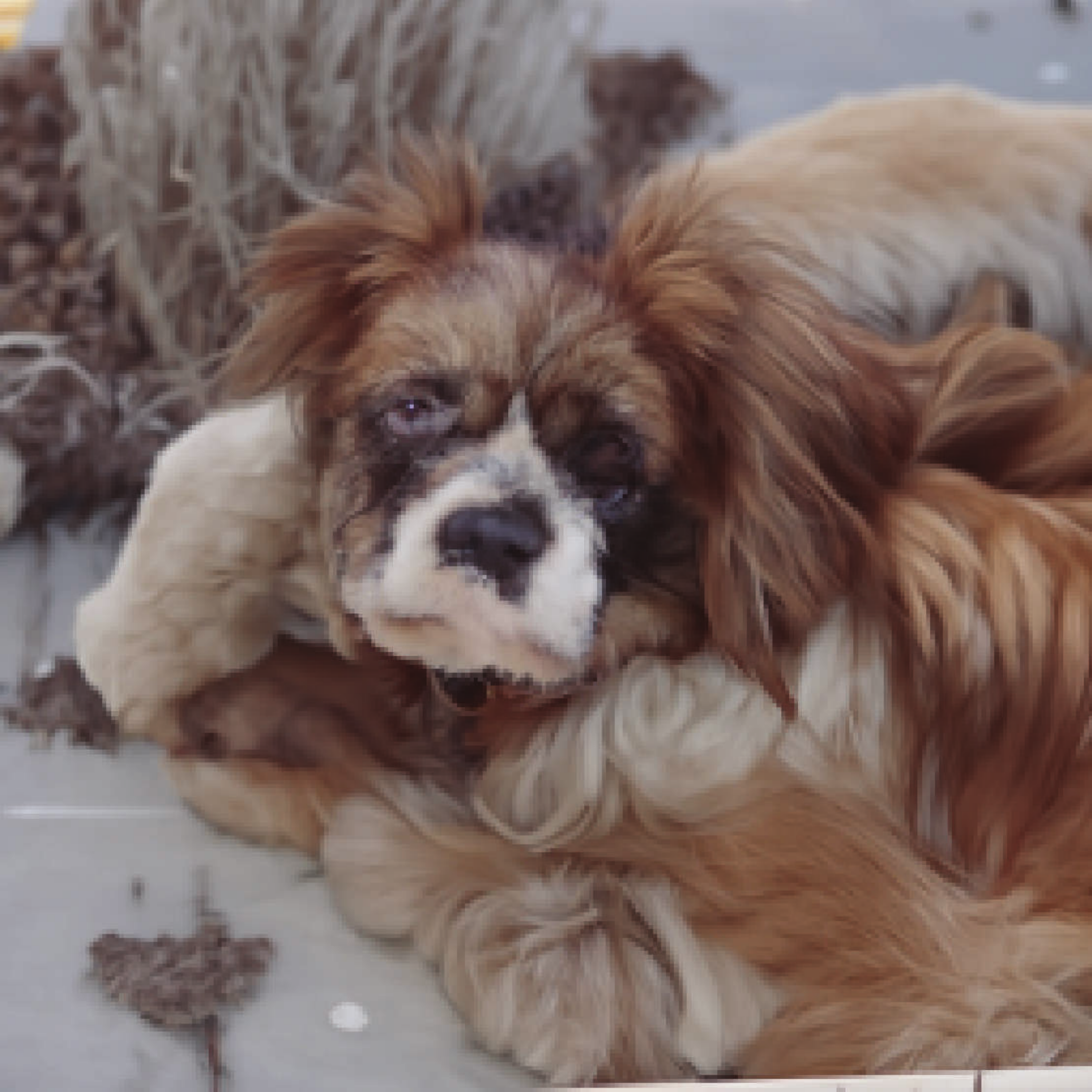}\hspace{-0.25em}
      \includegraphics[width=0.33\linewidth]{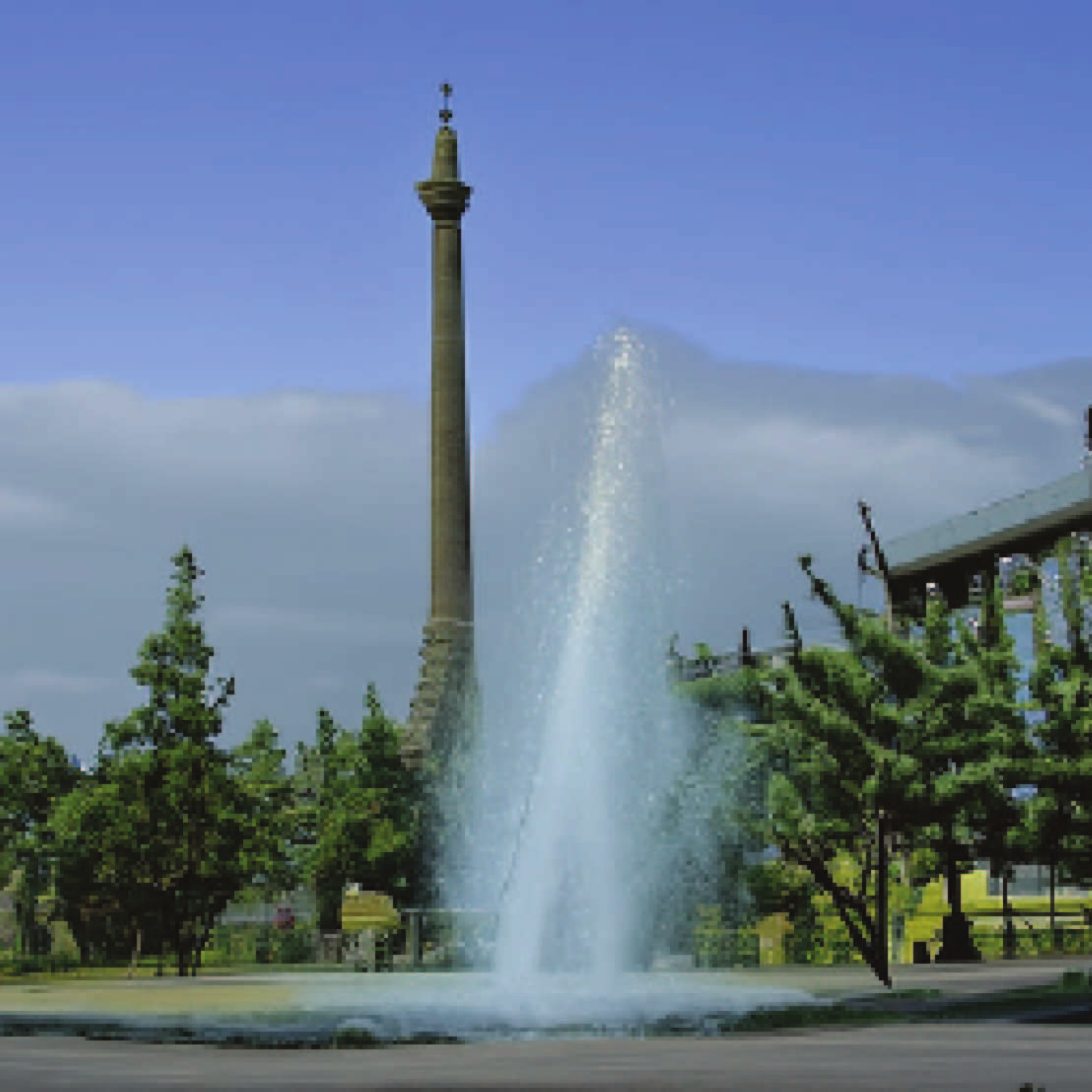}
\caption{(a) ADM~\cite{dhariwal2021diffusion} without (top) and with (bottom) SAG}
\end{subfigure}\hspace{-0.0em}
\begin{subfigure}{.497\textwidth}
\centering
\lineskip=0pt
      \includegraphics[width=0.33\linewidth]{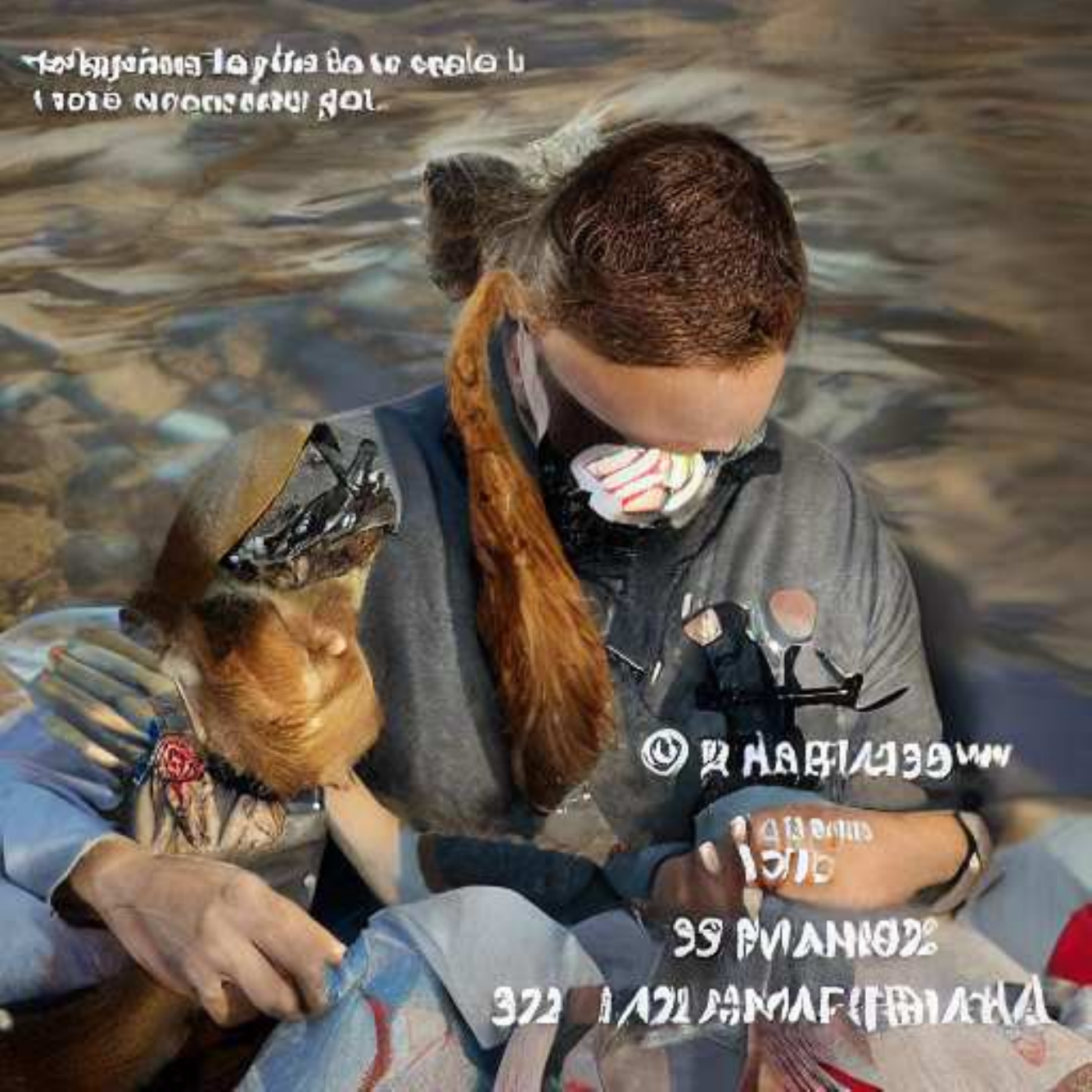}\hspace{-0.25em}
      \includegraphics[width=0.33\linewidth]{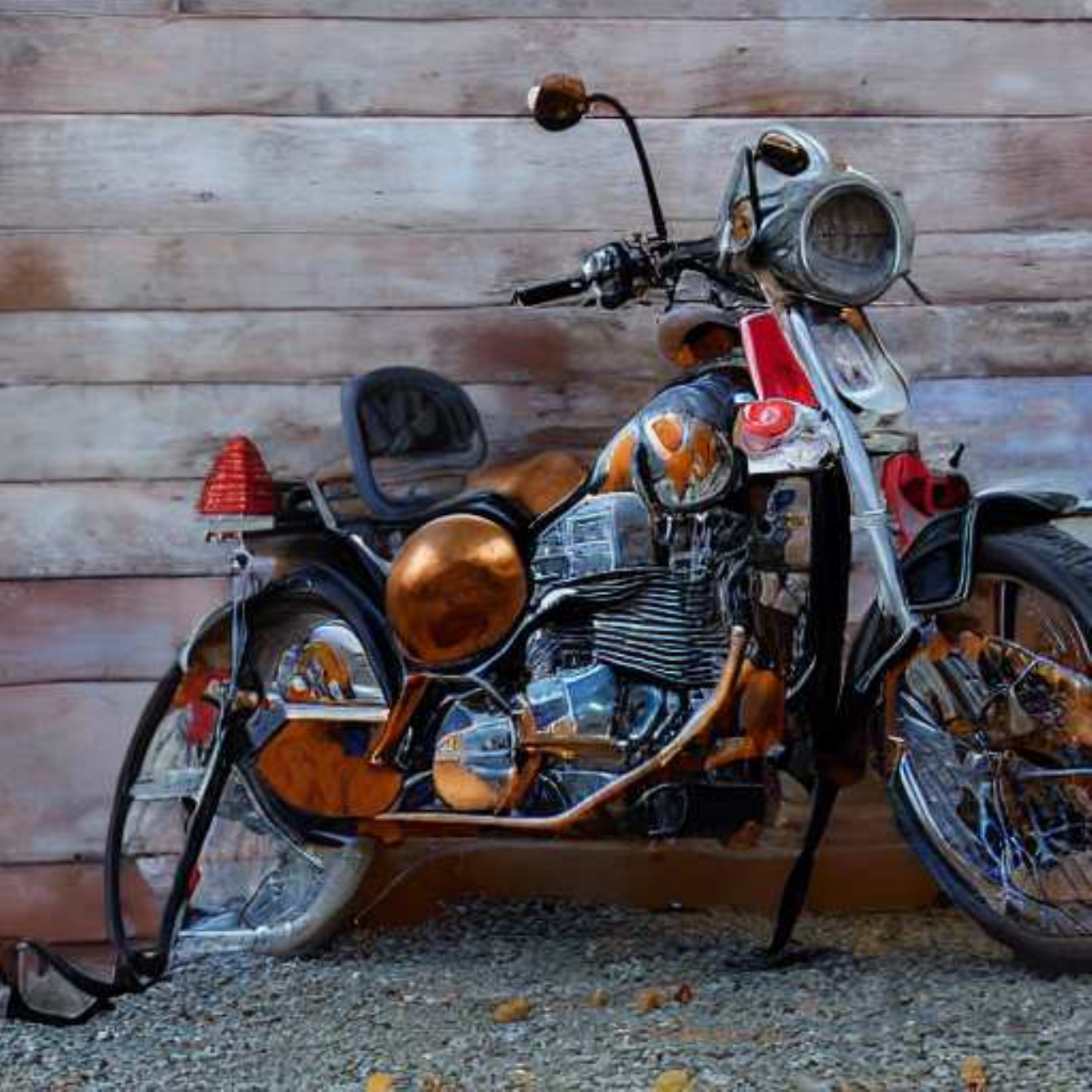}\hspace{-0.25em}
      \includegraphics[width=0.33\linewidth]{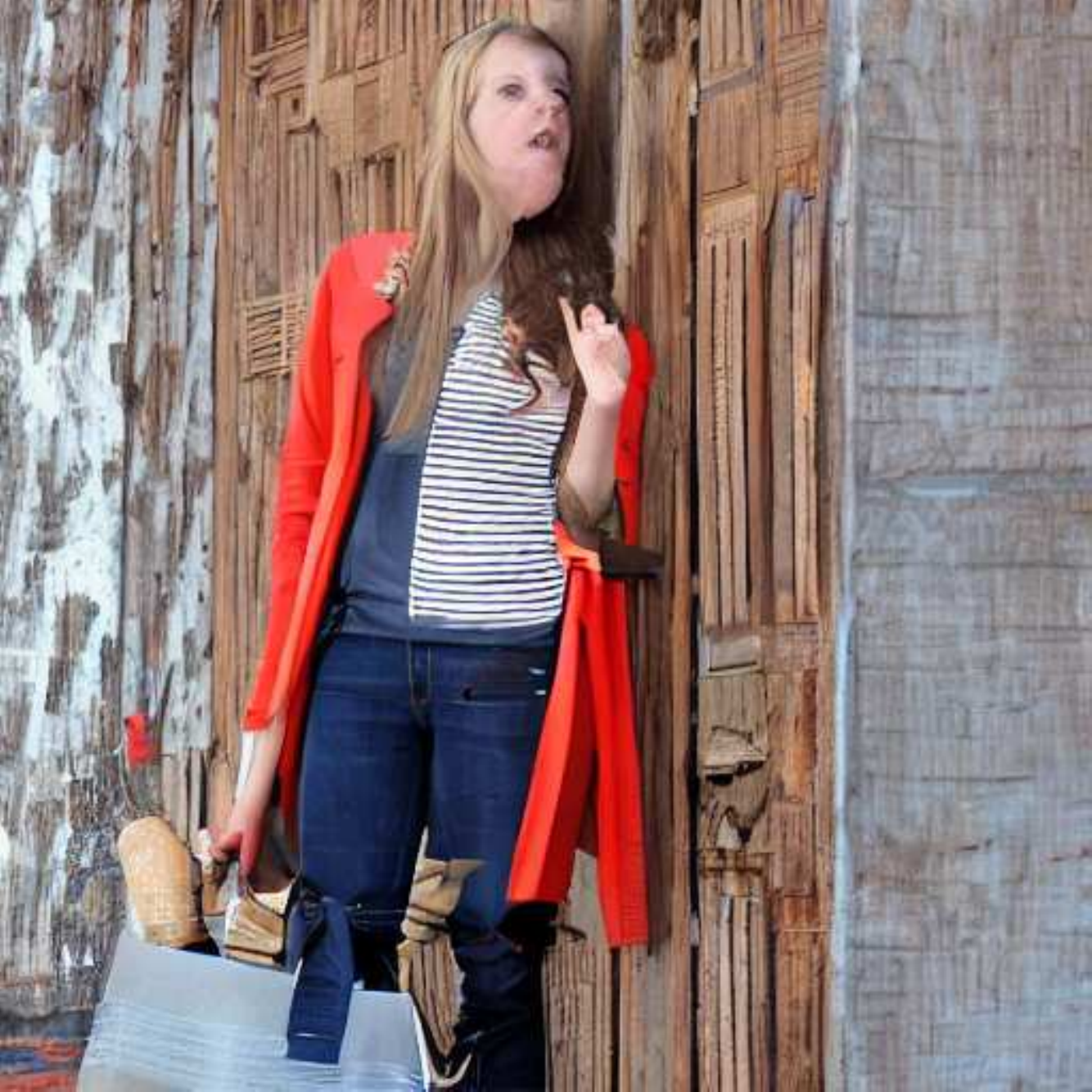} \\\vspace{-0.01pt}
      \includegraphics[width=0.33\linewidth]{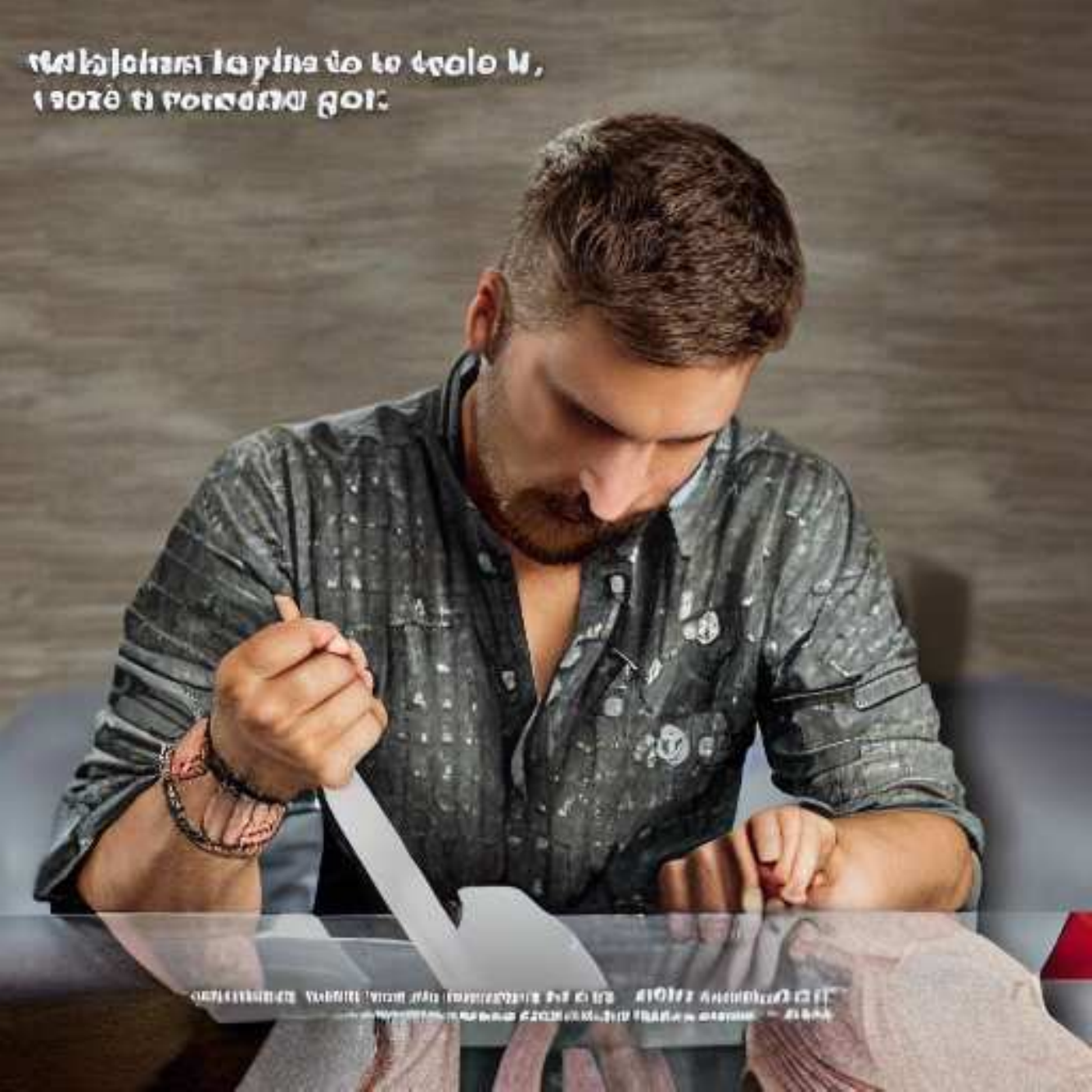}\hspace{-0.25em}
      \includegraphics[width=0.33\linewidth]{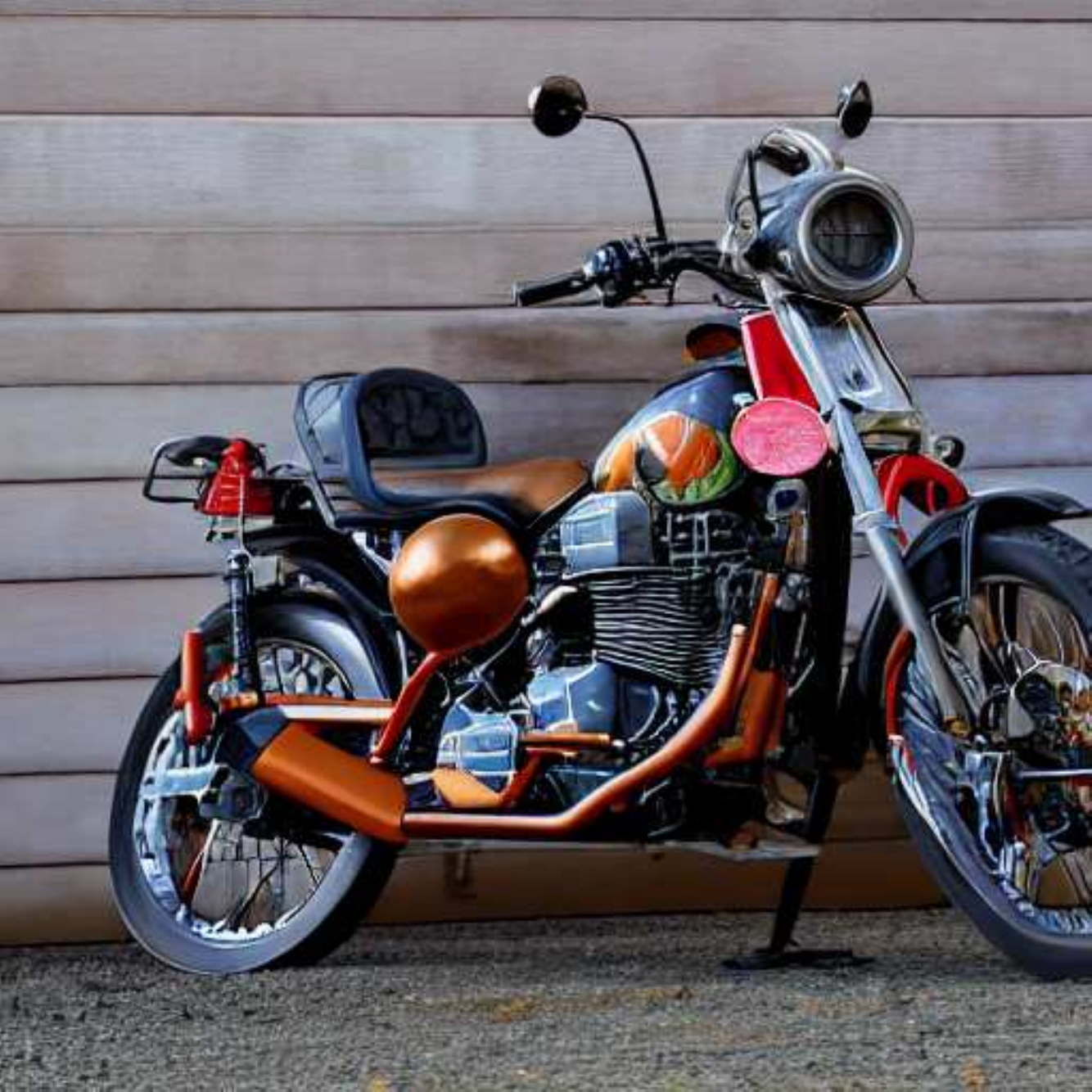}\hspace{-0.25em}
      \includegraphics[width=0.33\linewidth]{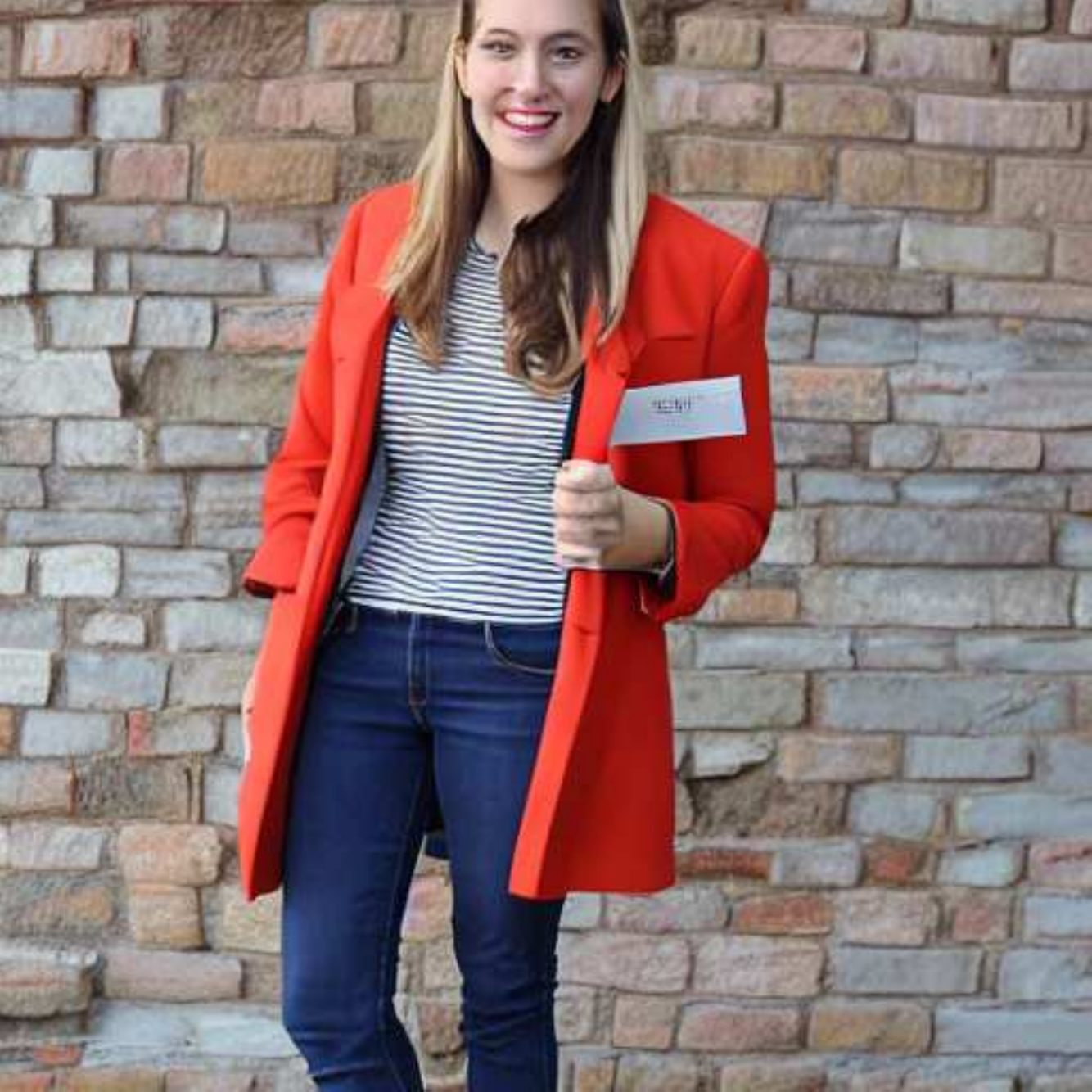}
\caption{(b) Stable Diffusion~\cite{rombach2022high} without (top) and with (bottom) SAG}
\end{subfigure}
\vspace{-20pt}
\caption{\textbf{Qualitative comparisons between unguided (top) and self-attention-guided (bottom) samples.} Unlike classifier guidance (CG)~\cite{dhariwal2021diffusion} or classifier-free guidance (CFG)~\cite{ho2021classifier}, self-attention guidance (SAG) does not necessarily require an external condition, \eg, a class label or text prompt, nor additional training, improving the details of the images generated by pre-trained diffusion models such as (a) unconditional ADM~\cite{dhariwal2021diffusion} and (b) Stable Diffusion~\cite{rombach2022high} with an empty prompt.}
\label{fig:branched}
\end{figure}
}]

\begin{abstract}
Denoising diffusion models (DDMs) have attracted attention for their exceptional generation quality and diversity. This success is largely attributed to the use of class- or text-conditional diffusion guidance methods, such as classifier and classifier-free guidance. In this paper, we present a more comprehensive perspective that goes beyond the traditional guidance methods. From this generalized perspective, we introduce novel condition- and training-free strategies to enhance the quality of generated images. As a simple solution, blur guidance improves the suitability of intermediate samples for their fine-scale information and structures, enabling diffusion models to generate higher quality samples with a moderate guidance scale. Improving upon this, Self-Attention Guidance (SAG) uses the intermediate self-attention maps of diffusion models to enhance their stability and efficacy. Specifically, SAG adversarially blurs only the regions that diffusion models attend to at each iteration and guides them accordingly. Our experimental results show that our SAG improves the performance of various diffusion models, including ADM, IDDPM, Stable Diffusion, and DiT. Moreover, combining SAG with conventional guidance methods leads to further improvement.
\end{abstract}

{\let\thefootnote\relax\footnote{The project page and code can be accessed at:\\\rightline{~
\urlstyle{rm}{\url{https://ku-cvlab.github.io/Self-Attention-Guidance/}}}}}
\section{Introduction}

Recently, denoising diffusion models (DDMs)~\cite{sohl2015deep, song2019generative, ho2020denoising, dhariwal2021diffusion, ho2022cascaded, rombach2022high}, which synthesize images from noise through an iterative denoising process, have been actively researched and attracted attention due to their exceptional performance in synthesizing high-quality and diverse images.

Behind this remarkable success lies the introduction of diffusion guidance methods~\cite{dhariwal2021diffusion,nichol2021glide,ho2021classifier}. Several studies have revealed that to improve the quality of image samples generated by diffusion models, guidance techniques using class labels~\cite{dhariwal2021diffusion, ho2021classifier} or captions~\cite{nichol2021glide} are essential. However, despite the significant improvement provided by these guidance methods, they are bounded within the limits of using external conditions. For example, classifier guidance (CG)~\cite{dhariwal2021diffusion} requires the training of an additional classifier, and classifier-free guidance (CFG)~\cite{ho2021classifier} adds complexity to the training process through label-dropping. In addition, both methods are limited by their need for hard-earned external conditions, which binds them to conditional settings.

In light of the limitations mentioned above, in this work, we present a more general formulation of diffusion guidance that can make use of information within the intermediate samples of diffusion models. This formulation detaches the necessary condition of traditional approaches~\cite{ho2021classifier,dhariwal2021diffusion,nichol2021glide}, \ie, the requirement for external information, from diffusion guidance, and facilitates a flexible and condition-free approach to guide diffusion models. This broadens the applicability of diffusion guidance to cases with or without external conditions.

Based on the generalized formulation and the intuition that any internal information within intermediate samples can also serve as guidance, we firstly propose blur guidance as a straightforward solution to improve sample quality. Blur guidance uses the eliminated information resulting from Gaussian blur to guide intermediate samples, exploiting the benign property of Gaussian blur that it naturally removes fine-scale details~\cite{hoogeboom2022blurring,lee2022progressive,rissanen2022generative}. While our results show that this method improves sample quality with a moderate guidance scale, it becomes problematic with a large guidance scale, since it may introduce structural ambiguity in entire regions, which makes it difficult to align the prediction of the degraded input with that of the original one.

\begin{figure}[t]
\centering
\begin{subfigure}{0.45\columnwidth}
  \centering
  \includegraphics[width=1\linewidth]{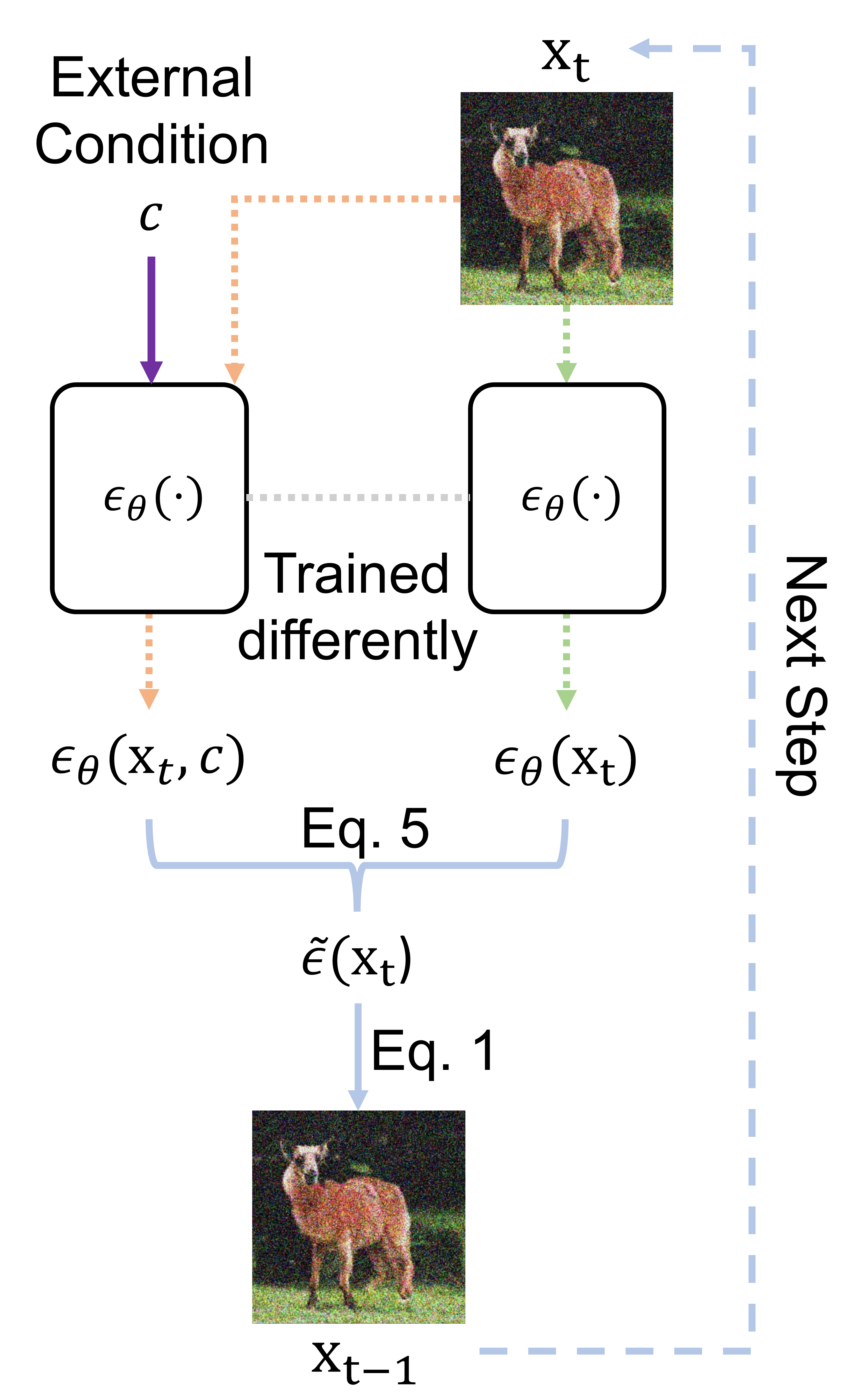}
  \caption{Classifier-free guidance}
\end{subfigure}\hfill
\begin{subfigure}{0.54\columnwidth}
  \centering
  \includegraphics[width=1\linewidth]{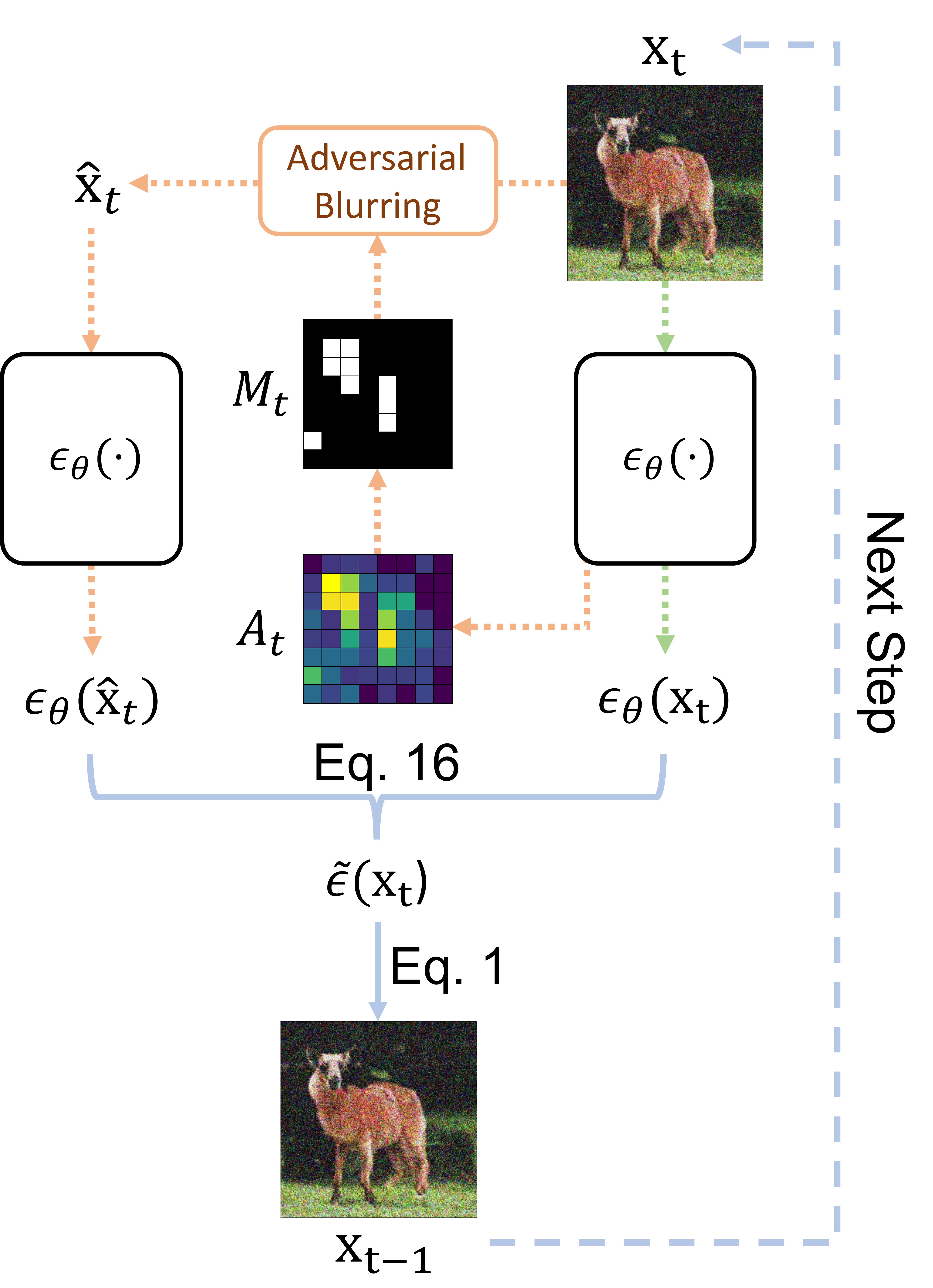}
  \caption{Self-attention guidance}
\end{subfigure}
\vspace{-5pt}
\caption{\textbf{Comparison of classifier-free guidance~\cite{ho2021classifier} and self-attention guidance (SAG).} Compared to classifier-free guidance that uses external class information, SAG extracts the internal information with the self-attention to guide the models, making it training- and condition-free.}\vspace{-10pt}
\label{fig:main_figure}
\end{figure}

To improve the effectiveness and stability of blur guidance with a larger guidance scale, we explore the self-attention mechanism of diffusion models. Generally, recent diffusion models~\cite{ho2020denoising,dhariwal2021diffusion,nichol2021improved,rombach2022high,ho2022cascaded,peebles2022scalable} are equipped with a self-attention module~\cite{vaswani2017attention,dosovitskiy2020image} within their architecture. Claiming that the self-attention is a key to capture salient information during generation process~\cite{jiang2021transgan,zhang2022styleswin,zhang2019self,hertz2022prompt}, we present Self-Attention Guidance (SAG), which adversarially blurs the region that contains salient information using the self-attention map of diffusion models and guides diffusion models with the residual information. Leveraging the attention maps during the reverse process of diffusion models, it can encouragingly boost the quality and reduce the artifacts through self-conditioning without requiring external information nor additional training, as shown in Fig.~\ref{fig:branched}. The pseudocode and pipeline are provided in Alg.~\ref{alg:sag} and Fig.~\ref{fig:main_figure}(b), respectively.

In experiments, we evaluate the effectiveness of the proposed approach by plugging it into various diffusion models including ADM~\cite{dhariwal2021diffusion}, IDDPM~\cite{nichol2021improved}, Stable Diffusion~\cite{rombach2022high}, and DiT~\cite{peebles2022scalable}, which demonstrates our method's broad applicability. We also show that in addition to the increased sample quality when using SAG alone, performance further improves when using it on top of existing guidance schemes, \ie, classifier~\cite{dhariwal2021diffusion} or classifier-free~\cite{ho2021classifier} guidance, demonstrating the orthogonality with the existing methods. Finally, we present ablation studies to validate our choices.

To sum up, our work has the following contributions:
\begin{itemize}

\item Generalizing conditional guidance methods~\cite{dhariwal2021diffusion,ho2021classifier,nichol2021glide} into a condition-free method that can be applied to any diffusion model without external conditions, expanding the applicability of guidance.
\vspace{-5pt}

\item Introducing novel guidance, dubbed Self-Attention Guidance (SAG), that uses the internal self-attention maps of diffusion models, improving sample quality without external conditions or additional fine-tuning.
\vspace{-5pt}

\item Demonstrating the orthogonality of SAG to existing conditional models and methods, enabling its flexible combination with others to achieve higher performance.
\vspace{-5pt}

\item Presenting extensive ablation studies to justify the design choices and demonstrate the effectiveness of the proposed method.

\end{itemize}

\section{Related Work}
\paragraph{Denoising diffusion models.}
Diffusion models~\cite{sohl2015deep}, which are closely related to score-based models~\cite{song2019generative, song2020score}, have attracted much attention owing to their superior sampling quality and diversity. As a pioneering work, DDPM~\cite{ho2020denoising} generates an image through an iterative process that progressively performs denoising to recover an image. Following this work, there have been several approaches to improve the sampling process, in terms of quality and speed~\cite{song2021denoising, nichol2021improved, rombach2022high, ho2022cascaded, dhariwal2021diffusion}. Notably, IDDPM~\cite{nichol2021improved} additionally predicts the variance of the reverse process of the diffusion model. DDIM~\cite{song2021denoising} accelerates the sampling speed by introducing the non-Markovian diffusion process. LDM~\cite{rombach2022high} reduces the computational cost by processing the diffusion process in the latent space.
\vspace{-10pt}

\renewcommand{\algorithmiccomment}[1]{\bgroup\quad\quad//~#1\egroup}
\begin{algorithm}[t]
\small
\caption{Self-Attention Guidance (SAG) Sampling}
\label{alg:sag}
\textbf{Functions}:\\
$\textrm{Model}(\mathbf{x}_t)$: a diffusion model that outputs the predicted noise $\epsilon_t$, variance $\Sigma_t$, and self-attention map $A_t$ given the input $\mathbf{x}_t$.\\
$\textrm{Gaussian-Blur}(\hat{\mathbf{x}}_0)$: a Gaussian blurring function.

\begin{algorithmic}[0] 
\STATE $\mathbf{x}_T \sim \mathcal{N}(0, \mathbf{I})$
\FOR{$t$ in $T, T-1, ..., 1$}
\STATE $\epsilon_t, \Sigma_t, A_t \leftarrow \textrm{Model}(\mathbf{x}_t)$
\STATE $M_t \leftarrow \mathbbm{1}(A_t > \psi)$
\STATE $\hat{\mathbf{x}}_0 \leftarrow (\mathbf{x}_t-\sqrt{1-\bar{\alpha}_t}\epsilon_t)/\sqrt{\bar{\alpha}_t}$\COMMENT{Eq.~\ref{eq:x0_pred}}
\STATE $\tilde{\mathbf{x}}_0 \leftarrow \textrm{Gaussian\text-Blur}(\hat{\mathbf{x}}_0)$
\STATE $\tilde{\mathbf{x}}_t \leftarrow \sqrt{\bar{\alpha}_t}\tilde{\mathbf{x}}_0 + \sqrt{1-\bar{\alpha}_t}\epsilon_t $
\STATE $\widehat{{\mathbf{x}}}_t \leftarrow (1-M_t)\odot {\mathbf{x}}_t+M_t\odot \tilde{\mathbf{x}}_t$\COMMENT{Eq.~\ref{eq:sel-blur}}
\STATE $\widehat{\epsilon}_t \leftarrow \textrm{Model}(\widehat{{\mathbf{x}}}_t)$
\STATE $\tilde\epsilon_t \leftarrow \widehat{\epsilon}_t + (1+s)(\epsilon_t - \widehat{\epsilon}_t)$\COMMENT{Eq.~\ref{eq:sag-ori}}
\STATE $\mathbf{x}_{t-1} \sim \mathcal{N}(\frac{1}{\sqrt{\bar{\alpha}_t}}(\mathbf{x}_t-\frac{1-\alpha_t}{\sqrt{1 - \bar{\alpha}_t}}\tilde{\epsilon}_t), \Sigma_t)$\COMMENT{Eq.~\ref{eq:xt-1_pred}}
\ENDFOR
\STATE \textbf{return} $\mathbf{x}_0$
\end{algorithmic} 
\end{algorithm}

\paragraph{Sampling guidance for diffusion models.}
Recent works have proposed diffusion guidance methods based on class labels to generate images with higher quality~\cite{dhariwal2021diffusion, ho2021classifier}. Classifier guidance (CG)~\cite{dhariwal2021diffusion} is an approach that uses a trained classifier that guides the reverse process toward a specific class distribution. As an alternative strategy without an additional classifier, Ho and Salimans~\cite{ho2021classifier} propose classifier-free guidance (CFG). Due to its simplicity of implementation and effectiveness, the guidance has been used in various high-quality diffusion models~\cite{ramesh2022hierarchical,rombach2022high,tang2022improved,wang2022pretraining,nichol2021glide,saharia2022photorealistic}. Adopting the concepts of the guidance methods above, Nichol \textit{et al.}~\cite{nichol2021glide} propose text-to-image generation with CLIP~\cite{radford2021learning} guidance and CFG. However, these approaches have limitations since they do not apply to unlabeled datasets and require additional training procedures~\cite{dhariwal2021diffusion, ho2021classifier}.
\vspace{-10pt}

\paragraph{Self-attention in generative models.}
A self-attention mechanism is the key ingredient of Transformer-based models~\cite{vaswani2017attention}. Notably, it has become a \textit{de facto} method in natural language processing tasks~\cite{vaswani2017attention} for its expressive power and capability to encode global context, which has inspired many works to incorporate this mechanism into computer vision~\cite{dosovitskiy2020image,jiang2021transgan,zhang2022styleswin,zhang2019self}. Among those, Jiang \textit{et al.}~\cite{jiang2021transgan} and Zhang \textit{et al.}~\cite{zhang2022styleswin,zhang2019self} attempt to bring self-attention into generative adversarial networks (GANs) for better image quality. Following this, diffusion models have also brought self-attention into their model architectures. DDPM~\cite{ho2020denoising} initiates this trend by introducing a self-attention layer at a coarse resolution of the U-Net~\cite{ronneberger2015u}. Inspired by this work, Dhariwal and Nichol~\cite{dhariwal2021diffusion} measure the boost performance according to the varying number of self-attention heads and resolutions. Concurrently, DiT~\cite{peebles2022scalable} even accomplishes high performance leveraging Transformer-based backbones.
\vspace{-10pt}

\paragraph{Internal representations of diffusion models.}
Motivated by the success of diffusion models in generation tasks, some works have tried to utilize the representations of diffusion models to do other tasks, such as semantic segmentation. Brempong \textit{et al.}~\cite{brempong2022denoising} show that the denoising pre-training boosts the performance on semantic segmentation, and Baranchuk \textit{et al.}~\cite{baranchuk2021label} propose a label-efficient strategy for semantic segmentation using the U-Net~\cite{ronneberger2015u} representations of diffusion models. While specific tasks such as text-driven manipulation using cross-attention has been researched concurrently~\cite{hertz2022prompt}, these are inherently different from improving and self-conditioning general diffusion models in a condition-free way leveraging the internal self-attention maps, which is mainly discussed in this paper.

\section{Preliminaries}
\label{sec:preliminaries}
\paragraph{Denoising diffusion probabilistic models.}
DDPM~\cite{ho2020denoising} is a model that recovers an image from white noise through an iterative denoising process. Formally, given an image $\mathbf{x}_0$ and a variance schedule $\beta_t$ at a timestep $t\in \{T, T-1, \ldots, 1\}$, we can obtain $\mathbf{x}_t$ through the forward process which is defined as a Markovian process. Similarly, given a trained diffusion model parameterized by $\epsilon_\theta(\mathbf{x}_t, t)$ and $\Sigma_\theta(\mathbf{x}_t, t)$, we can define the reverse process. In this case, we set $\Sigma_\theta(\mathbf{x}_t, t)$ to $\sigma_t^2 = \beta_t$~\cite{ho2020denoising} although it is possible to predict the variance~\cite{nichol2021improved, dhariwal2021diffusion}. Specifically, given $\mathbf{x}_T \sim \mathcal{N}(0, \mathbf{I})$ and $\Sigma_\theta(\mathbf{x}_t, t)$, DDPM samples $\mathbf{x}_{T-1}, \mathbf{x}_{T-2}, \ldots, \mathbf{x}_{0}$ by computing:
\begin{equation}
\mathbf{x}_{t-1} = \frac{1}{\sqrt{\bar{\alpha}_t}}(\mathbf{x}_t-\frac{\beta_t}{\sqrt{1-\bar{\alpha}_t}}\epsilon_{\theta}(\mathbf{x}_t, t))+\sigma_t\mathbf{z},
\label{eq:xt-1_pred}
\end{equation}
where $\alpha_t = 1 - \beta_t$, $\bar{\alpha}_t=\prod_{i=1}^{t}\alpha_i$, $\mathbf{z} \sim \mathcal{N}(0, \mathbf{I})$, and $\epsilon_\theta$ denotes a neural network parameterized by $\theta$. Note that for simplicity, we define $\epsilon_{\theta}(\mathbf{x}_t) := \epsilon_{\theta}(\mathbf{x}_t, t)$ for the rest of the paper. Using the reparameterization trick, we can obtain $\hat{\mathbf{x}}_0$,  an intermediate reconstruction of $\mathbf{x}_0$ at a timestep $t$, using the following equation:
\begin{equation}
\hat{\mathbf{x}}_0 = (\mathbf{x}_t-\sqrt{1-\bar{\alpha}_t}\epsilon_{\theta}(\mathbf{x}_t, t))/\sqrt{\bar{\alpha}_t}.
\label{eq:x0_pred}\vspace{-5pt}
\end{equation}

\paragraph{Classifier guidance and classifier-free guidance.}
To bring the capability of GANs' trading diversity for fidelity to diffusion models, Dhariwal and Nichol~\cite{dhariwal2021diffusion} propose the classifier guidance that uses an additional classifier $p(c|\mathbf x_t)$, where $c$ is a class label. The guidance can be formulated as the following with a guidance scale $s>0$:
\begin{equation}
    \tilde{\epsilon}(\mathbf x_t,c) = \epsilon_\theta(\mathbf x_t,c)-s\sigma_t\nabla_{\mathbf x_t}\log p(c|\mathbf x_t),
\label{eq:classifier-guidance}
\end{equation}
where $\epsilon_\theta(\mathbf x_t,c)$ is a conditional diffusion model, and $\tilde{\epsilon}(\mathbf x_t,c)$ is the guided output by the classifier. On the other hand, Ho and Salimans~\cite{ho2021classifier} present a classifier-free guidance strategy that achieves the similar effect as classifier guidance without the use of an additional classifier:
\begin{align}
\tilde{\epsilon}(\mathbf x_t,c)&={\epsilon_\theta}(\mathbf x_t,c)+s ({\epsilon_\theta}(\mathbf x_t,c)-\epsilon_\theta(\mathbf x_t))\\&=\epsilon_\theta(\mathbf x_t)+(1+s)(\epsilon_\theta(\mathbf x_t,c)-\epsilon_\theta(\mathbf x_t)).
\label{eq:classifier-free}
\end{align}
However, this method still demands hard-earned labels and confines the application to conditional diffusion models that use external conditions such as class or text~\cite{nichol2021glide,rombach2022high,saharia2022photorealistic} conditions. Moreover, it requires additional training detail that occasionally zero-outs the class embedding in the training phase~\cite{ho2021classifier}, thus imposing extra complexity.
\vspace{-10pt}

\paragraph{Self-attention in diffusion models.}
\label{sec:attn}
Several works of diffusion models use the U-Net structure~\cite{ronneberger2015u} with self-attention~\cite{vaswani2017attention} at one or some of the intermediate layers~\cite{ho2020denoising, dhariwal2021diffusion}. Moreover, very recently, diffusion models using Transformers~\cite{vaswani2017attention} as the backbone has also been proposed~\cite{peebles2022scalable}. Specifically, for the height $H$ and width $W$, given any feature map $X_t\in\mathbb{R}^{(HW) \times C}$ at a timestep $t$, the $N$-head self-attention is defined as:
\begin{equation}
Q^{h}_t = X_tW^{h}_Q,\quad K^{h}_t = X_tW^{h}_K,
\end{equation}
\begin{equation}
A^{h}_t = \textrm{softmax}(Q^{h}_t (K^{h}_t)^T/\sqrt{d}),
\label{eq:attn-map}
\end{equation}
where $W^{h}_Q, W^{h}_K \in \mathbb{R}^{C \times d}$ for $h=0,1,...,N-1$. Each $A^{h}_t$ is then right multiplied by $V^{h}_t = X_tW^{h}_V$, where $W^{h}_V \in \mathbb{R}^{C \times d}$.

\section{Generalizing Diffusion Guidance}
\label{sec:generalizing}

Although classifier guidance and classifier-free guidance have largely contributed to the conditional generation of diffusion models~\cite{dhariwal2021diffusion, ho2021classifier,nichol2021glide}, they depend on external inputs. In this work, we broaden our perspective by extending them to handle both cases: with or without external inputs. We also show how CFG~\cite{ho2021classifier} can be integrated into our framework at the end of this section.

At a given timestep $t$, the entire input for a diffusion model comprises a generalized condition represented as $\mathbf{h}_t$, and a perturbed sample $\bar{\mathbf{x}}_t$ that lacks $\mathbf{h}_t$. More specifically, the condition $\mathbf{h}_t$ can encompass internal information within $\mathbf{x}_t$, an external condition, or both. With this definition, the resulting guidance is formulated through the utilization of an imaginary regressor, $p_{\textrm{im}}(\mathbf{h}_t|\bar{\mathbf x}_t)$, which is assumed to predict $\mathbf{h}_t$ given $\bar{\mathbf{x}}_t$. Modifying guidance proposed in prior works~\cite{song2020score, dhariwal2021diffusion}, we present:
\begin{equation}
    \tilde{\epsilon}(\bar{\mathbf x}_t, \mathbf{h}_t) = \epsilon_\theta(\bar{\mathbf x}_t, \mathbf{h}_t)-s\sigma_t\nabla_{\bar{\mathbf x}_t}\log p_{\textrm{im}}(\mathbf{h}_t|\bar{\mathbf x}_t),
\label{eq:general-guidance}
\end{equation}
where we slightly abuse the notation for ${\epsilon_\theta}(\bar{\mathbf x}_t,\mathbf{h}_t)$ since we assume that the inputs are simply aggregated to match the original whole input. Intuitively, the gradient of the regressor, $\nabla_{\bar{\mathbf x}_t}\log p_{\textrm{im}}(\mathbf{h}_t|\bar{\mathbf x}_t)$, guides generated samples to be more suitable with that information.

With Bayes' rule, $p_{\textrm{im}}(\mathbf{h}|\bar{\mathbf x}_t)\propto p(\bar{\mathbf{x}}_t|\mathbf{h})/p(\bar{\mathbf{x}}_t)$, and the score of an imaginary regressor $p_{\textrm{im}}(\mathbf{h}_t|\bar{\mathbf x}_t)$ is derived:
\begin{equation}
    \nabla_{\bar{\mathbf x}_t}\log p_{\textrm{im}}(\mathbf{h}_t|\bar{\mathbf x}_t) = {-\frac{1}{\sigma_t} }(\epsilon^*(\bar{\mathbf x}_t,\mathbf{h}_t)-\epsilon^*(\bar{\mathbf x}_t)),
\end{equation}
where $\epsilon^*$ denotes the true score of the regressor. Eventually, this term is plugged into Eq.~\ref{eq:general-guidance} and produces:
\begin{align}
    \tilde{\epsilon}(\bar{\mathbf x}_t, \mathbf{h}_t) &= \epsilon_\theta(\bar{\mathbf x}_t, \mathbf{h}_t)-s\sigma_t\nabla_{\bar{\mathbf x}_t}\log p_{\textrm{im}}(\mathbf{h}_t|\bar{\mathbf x}_t)\\
    &={\epsilon_\theta}(\bar{\mathbf x}_t,\mathbf{h}_t)+s ({\epsilon_\theta}(\bar{\mathbf x}_t,\mathbf{h}_t)-\epsilon_\theta(\bar{\mathbf x}_t))
\label{eq:gg-prev}\\
    &={\epsilon_\theta}(\bar{\mathbf x}_t)+(1+s) ({\epsilon_\theta}(\bar{\mathbf x}_t,\mathbf{h}_t)-\epsilon_\theta(\bar{\mathbf x}_t)).
\label{eq:gg}
\end{align}
Note that Eq.~\ref{eq:gg-prev} induces a constraint that $\bar{\mathbf x}_t$ be in-manifold that the diffusion model $\epsilon_\theta$ defines. Also note that CFG~\cite{ho2021classifier} is a special case of Eq.~\ref{eq:gg} where $\bar{\mathbf x}_t = {\mathbf x}_t$, ${\mathbf h}_t = c$, and the imaginary regressor $p_{\textrm{im}}(\mathbf{h}_t|\bar{\mathbf x}_t)$ is reduced into the implicit classifier in \cite{ho2021classifier}.

Benefiting from this formulation, we can also define diffusion guidance on unconditional models, which have a sole noised image $\mathbf x_t$ as an input and no external label~\cite{ho2021classifier,dhariwal2021diffusion}, by making it self-conditional on visual information within the intermediate samples of the reverse process. In this light, we present comprehensive discussions on how to find appropriate ${\mathbf h}_t$ for unconditional models and according $\bar{\mathbf x}_t$, and subsequently propose guidance in Section~\ref{sec:method}.

\begin{figure}[t]
\captionsetup[subfigure]{justification=centering}

\begin{subfigure}{.12\textwidth}
  \centering
  \includegraphics[width=1.0\linewidth]{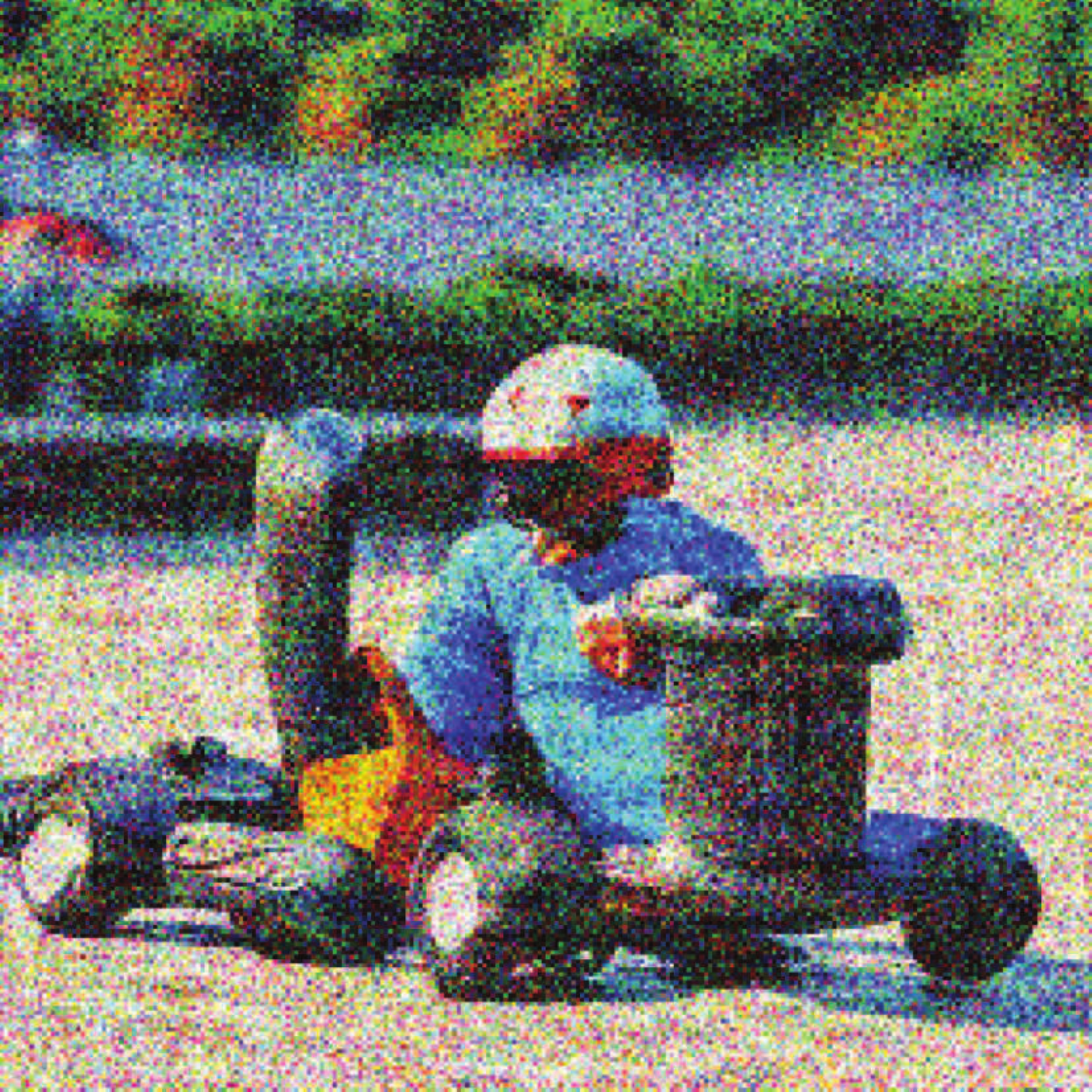}
\end{subfigure}\hspace{-0.25em}
\begin{subfigure}{.12\textwidth}
  \centering
  \includegraphics[width=1.0\linewidth]{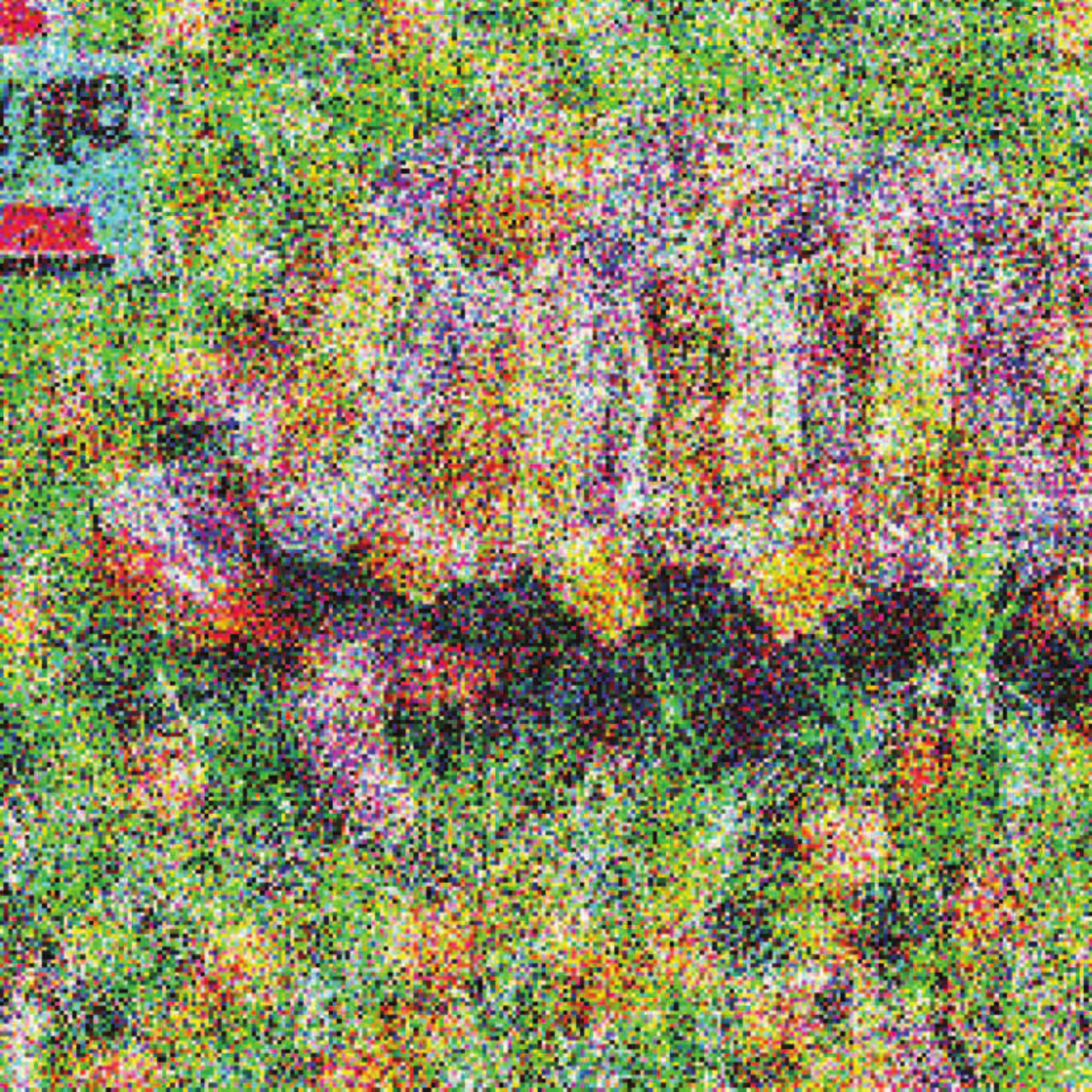}
\end{subfigure}\hspace{-0.25em}
\begin{subfigure}{.12\textwidth}
  \centering
  \includegraphics[width=1.0\linewidth]{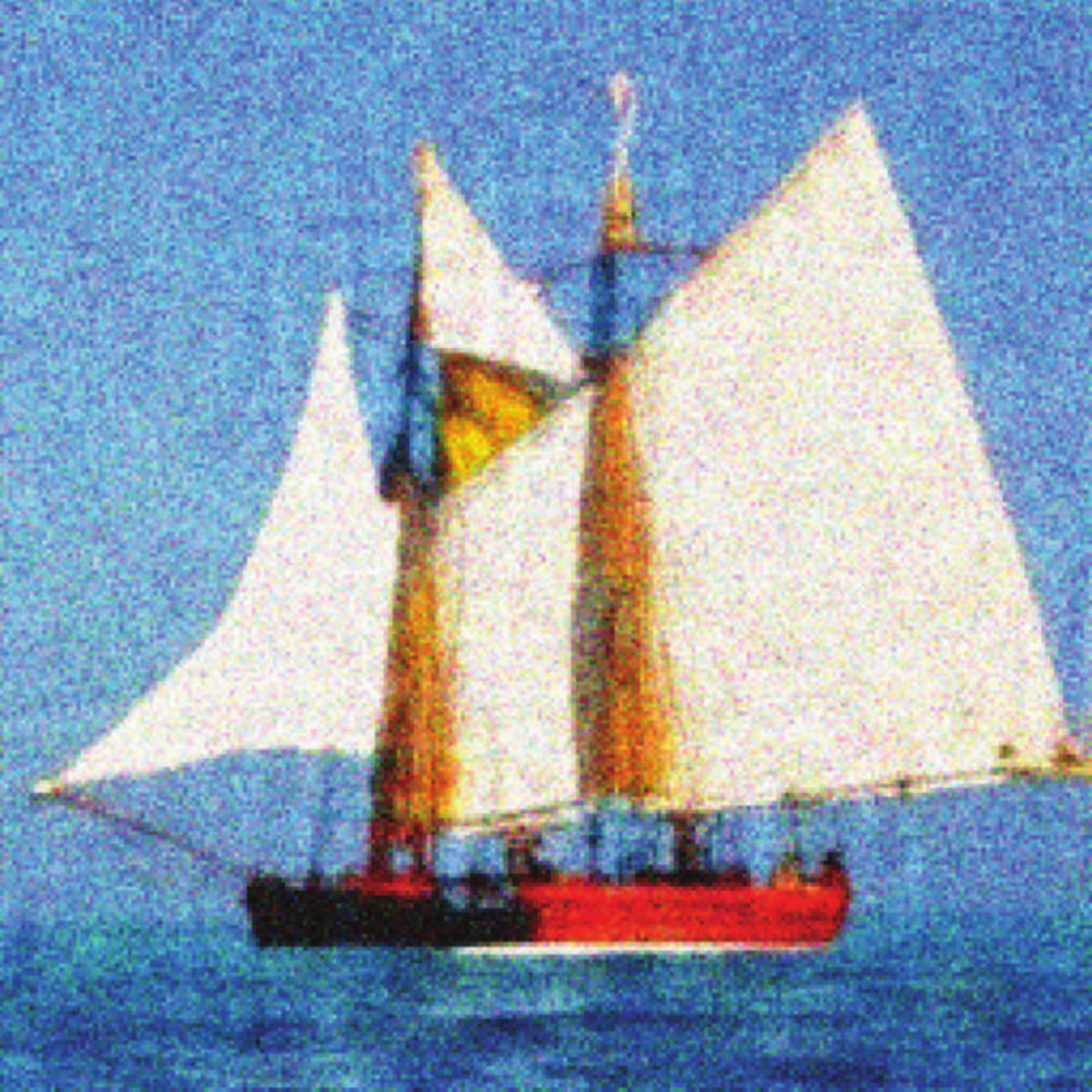}
\end{subfigure}\hspace{-0.25em}
\begin{subfigure}{.12\textwidth}
  \centering
  \includegraphics[width=1.0\linewidth]{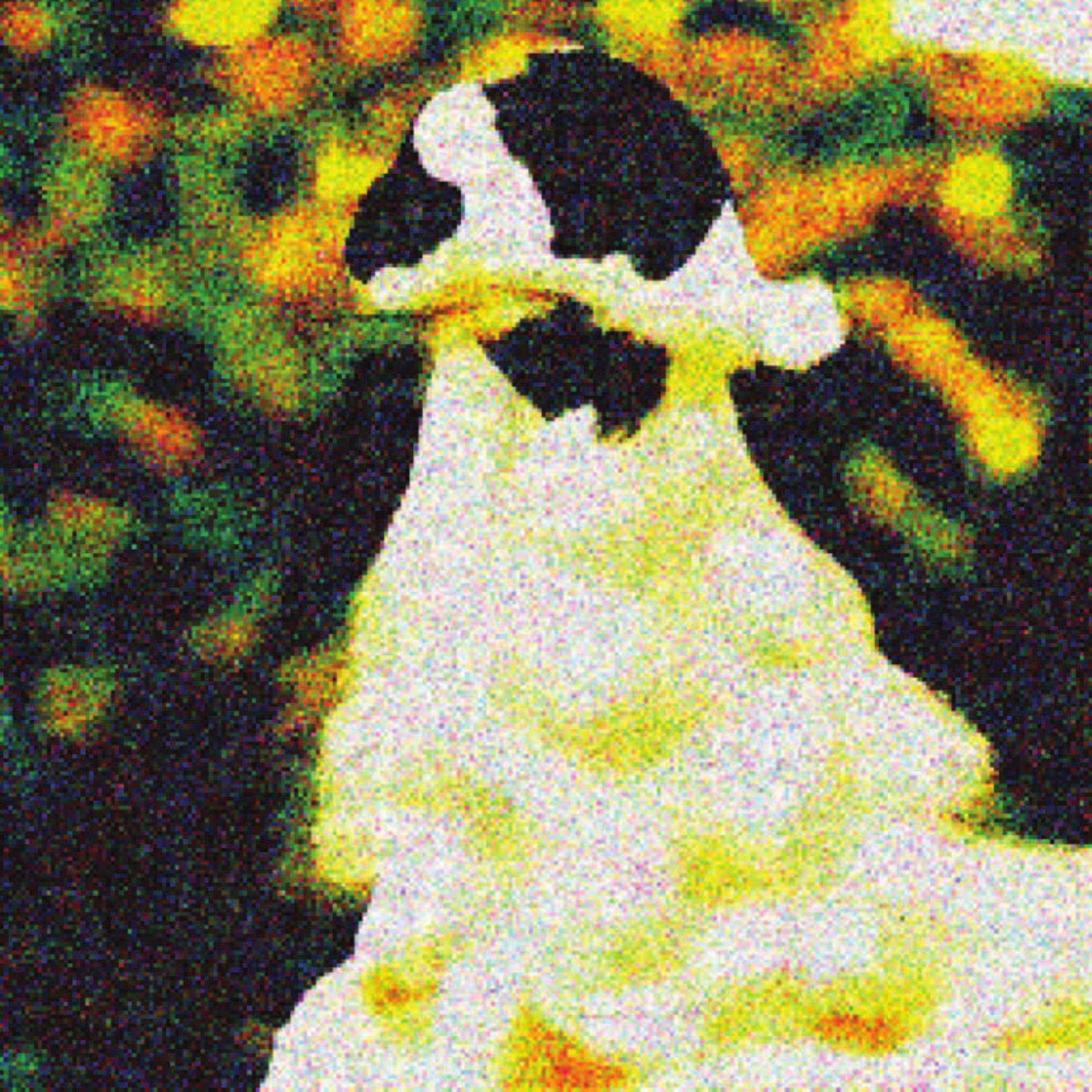}
\end{subfigure}\vspace{-0.9pt}\\
\begin{subfigure}{.12\textwidth}
  \centering
  \includegraphics[width=1.0\linewidth]{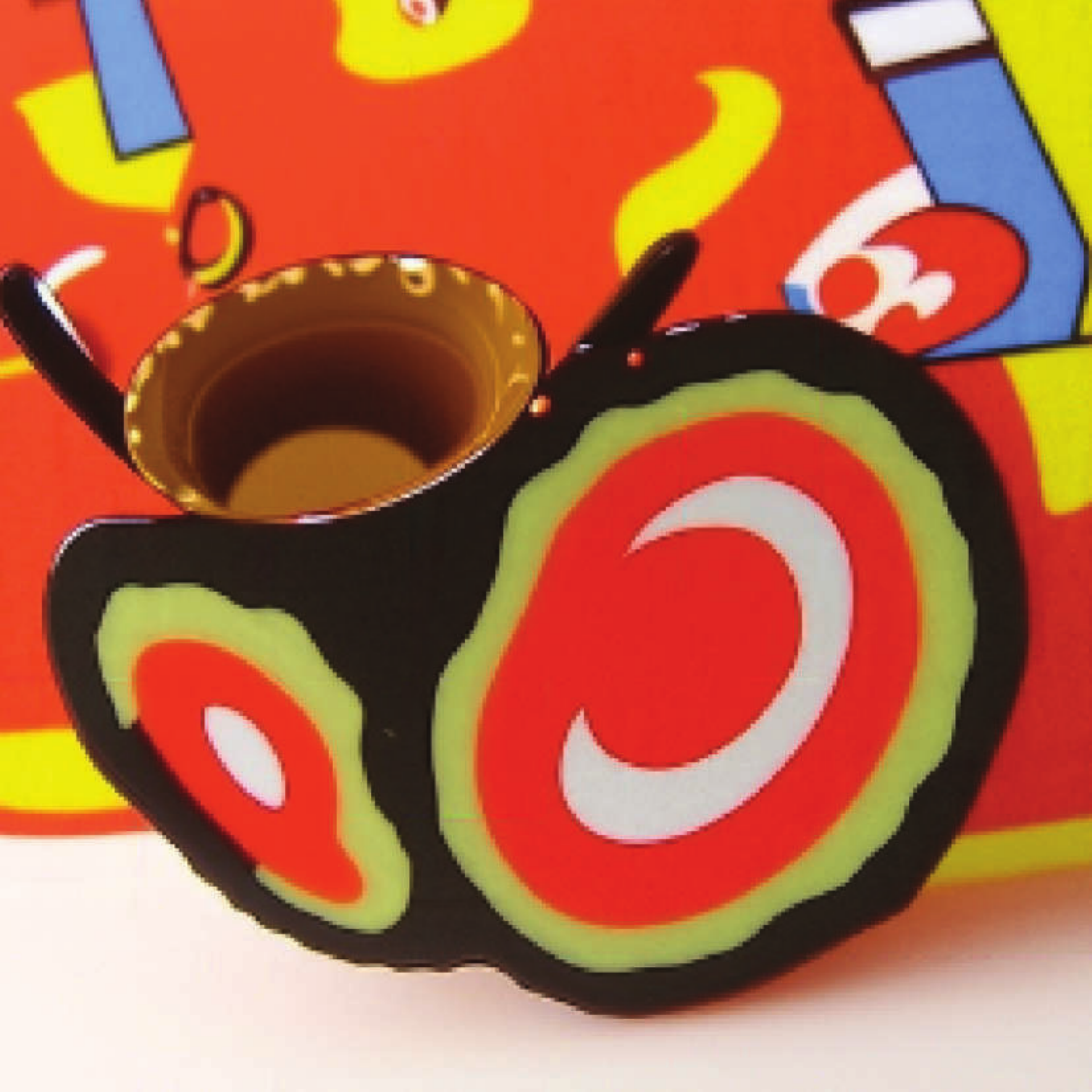}
\end{subfigure}\hspace{-0.25em}
\begin{subfigure}{.12\textwidth}
  \centering
  \includegraphics[width=1.0\linewidth]{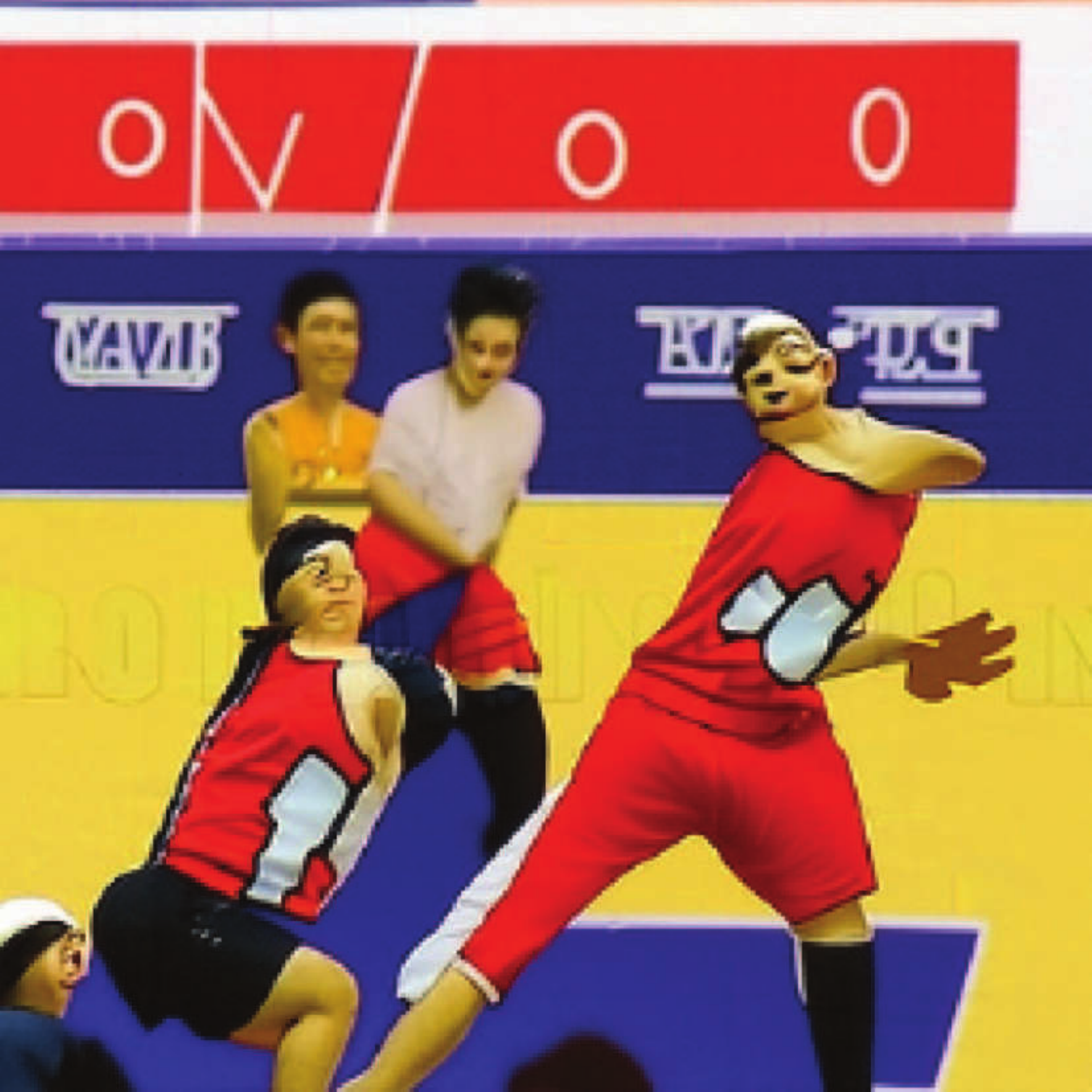}
\end{subfigure}\hspace{-0.25em}
\begin{subfigure}{.12\textwidth}
  \centering
  \includegraphics[width=1.0\linewidth]{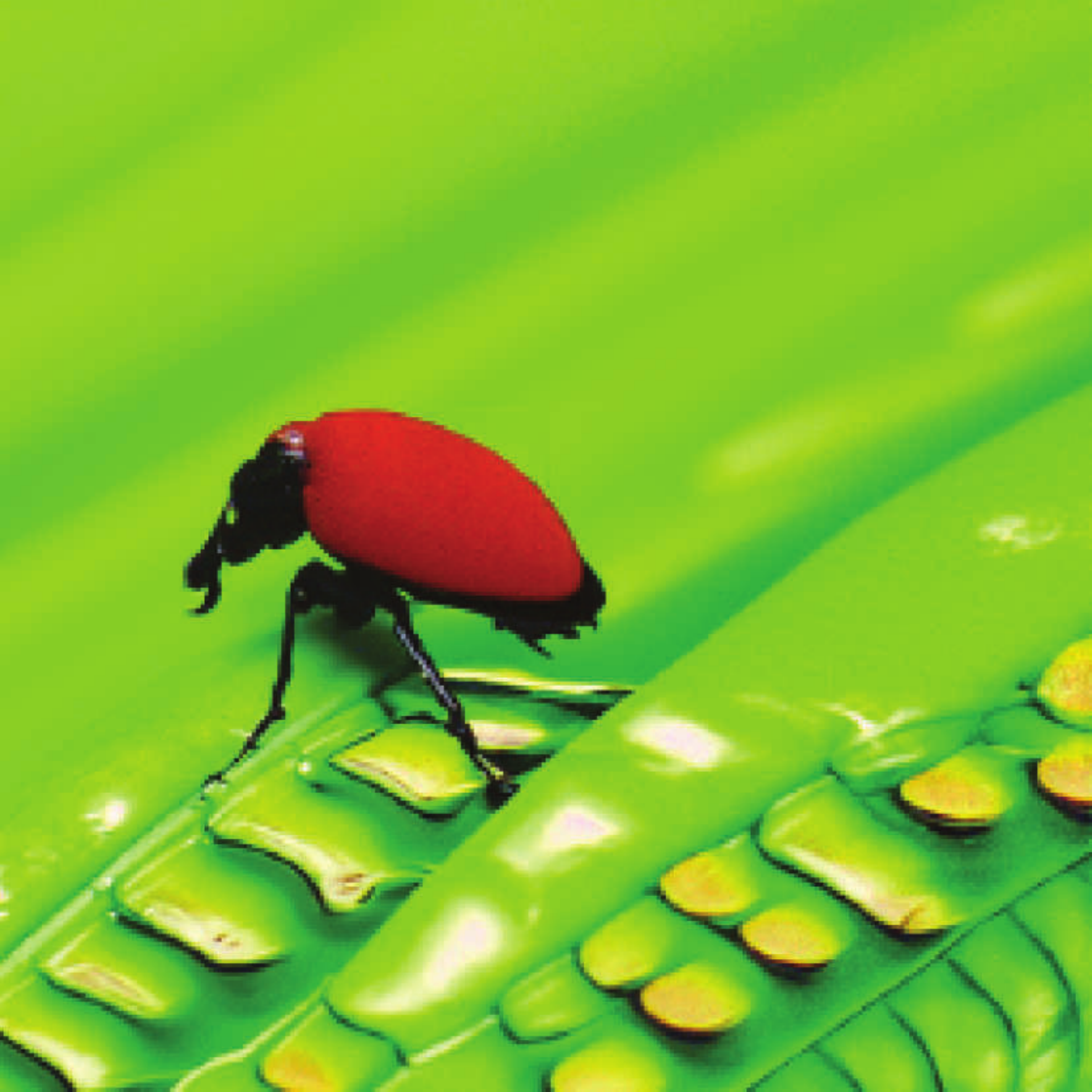}
\end{subfigure}\hspace{-0.25em}
\begin{subfigure}{.12\textwidth}
  \centering
  \includegraphics[width=1.0\linewidth]{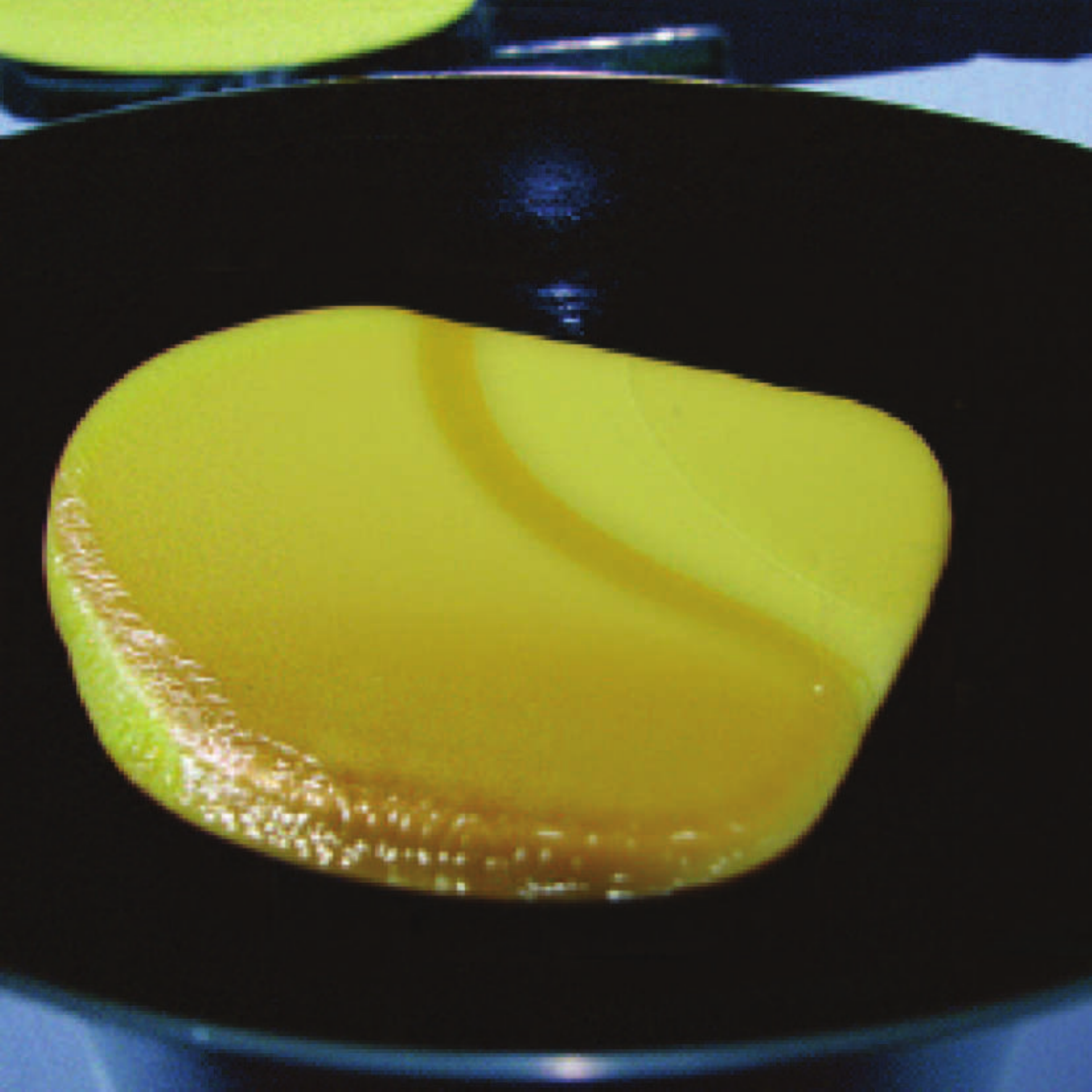}
\end{subfigure}
\vspace{-15pt}
\caption{\textbf{Comparison of blur guidance with self-attention guidance (SAG) under a large guidance scale.} Given an extreme guidance scale ($s=5.0$), blur guidance generates relatively noisy images (top) compared to those generated with SAG (bottom).}
\label{fig:unstab}\vspace{-10pt}
\end{figure}

\begin{figure}[t]
\captionsetup[subfigure]{justification=centering}

\begin{subfigure}{.12\textwidth}
  \centering
  \includegraphics[width=1.0\linewidth]{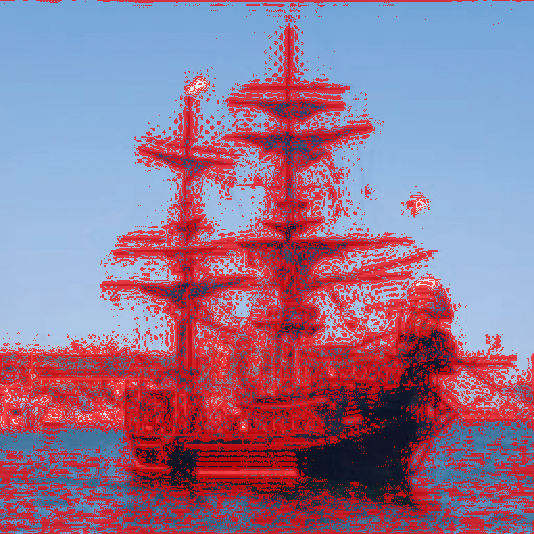}
\end{subfigure}\hspace{-0.25em}
\begin{subfigure}{.12\textwidth}
  \centering
  \includegraphics[width=1.0\linewidth]{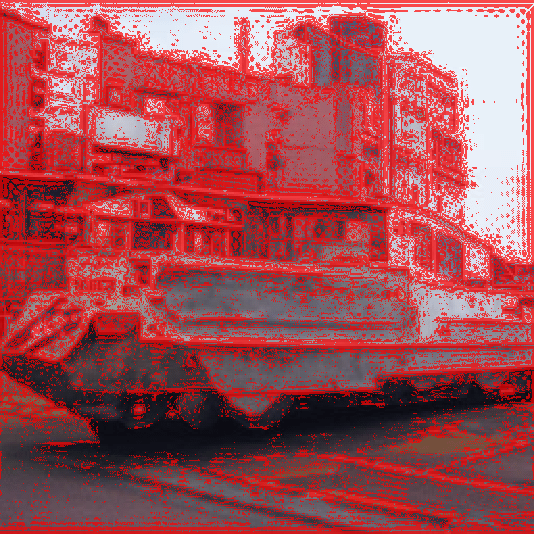}
\end{subfigure}\hspace{-0.25em}
\begin{subfigure}{.12\textwidth}
  \centering
  \includegraphics[width=1.0\linewidth]{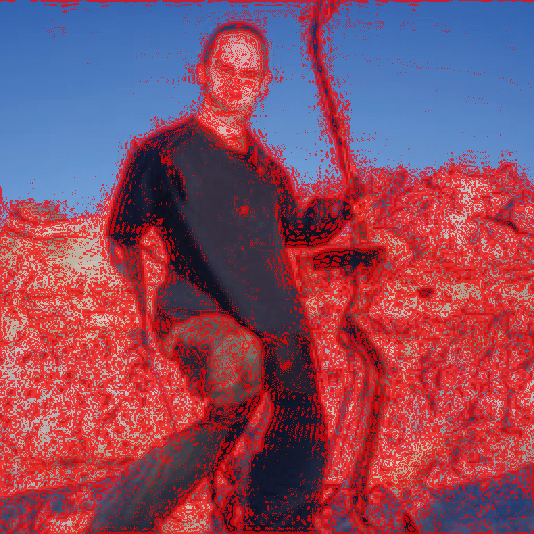}
\end{subfigure}\hspace{-0.25em}
\begin{subfigure}{.12\textwidth}
  \centering
  \includegraphics[width=1.0\linewidth]{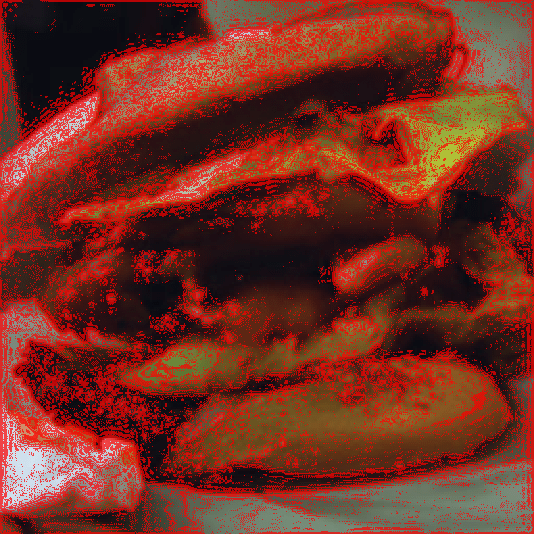}
\end{subfigure}\vspace{-0.9pt}\\
\begin{subfigure}{.12\textwidth}
  \centering
  \includegraphics[width=1.0\linewidth]{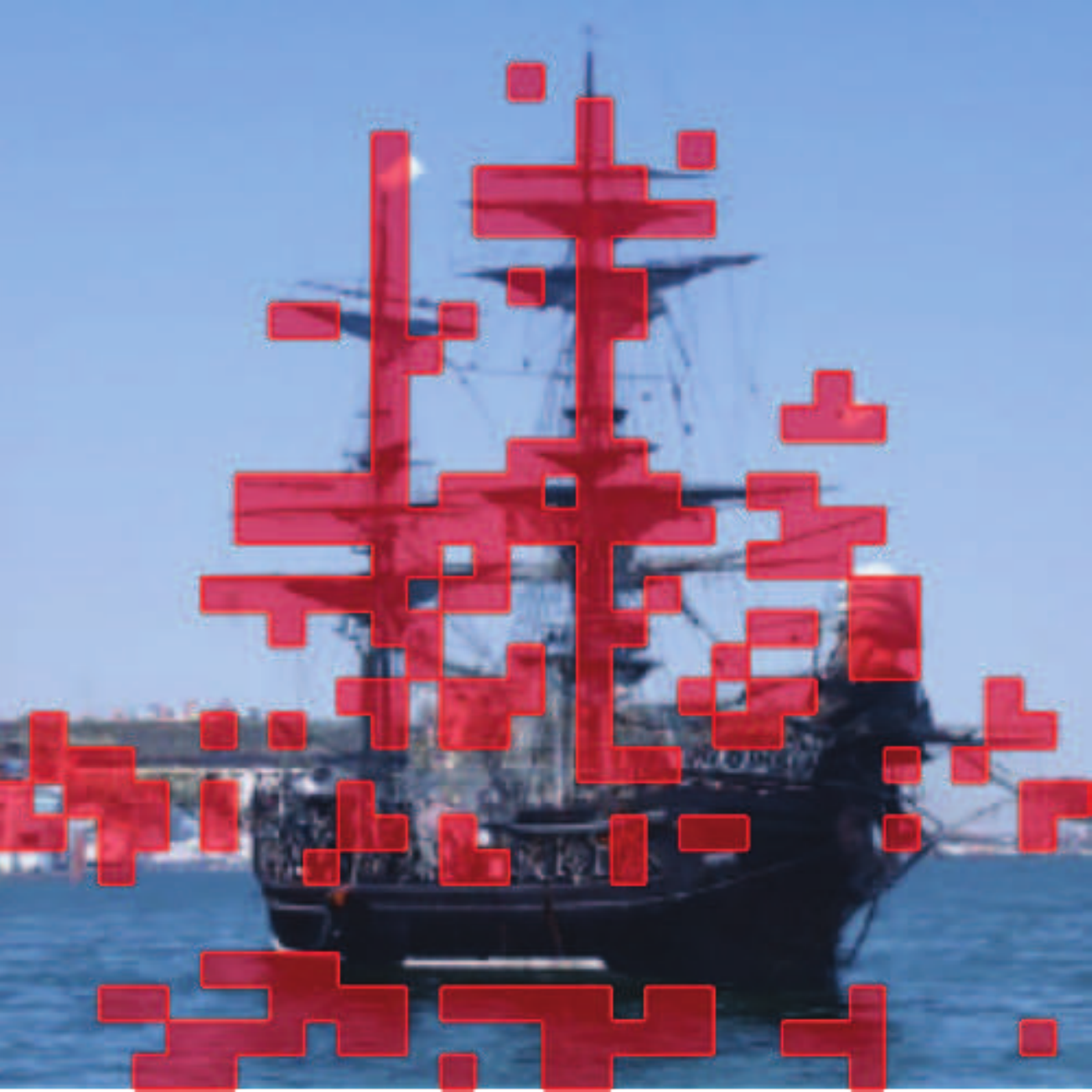}
\end{subfigure}\hspace{-0.25em}
\begin{subfigure}{.12\textwidth}
  \centering
  \includegraphics[width=1.0\linewidth]{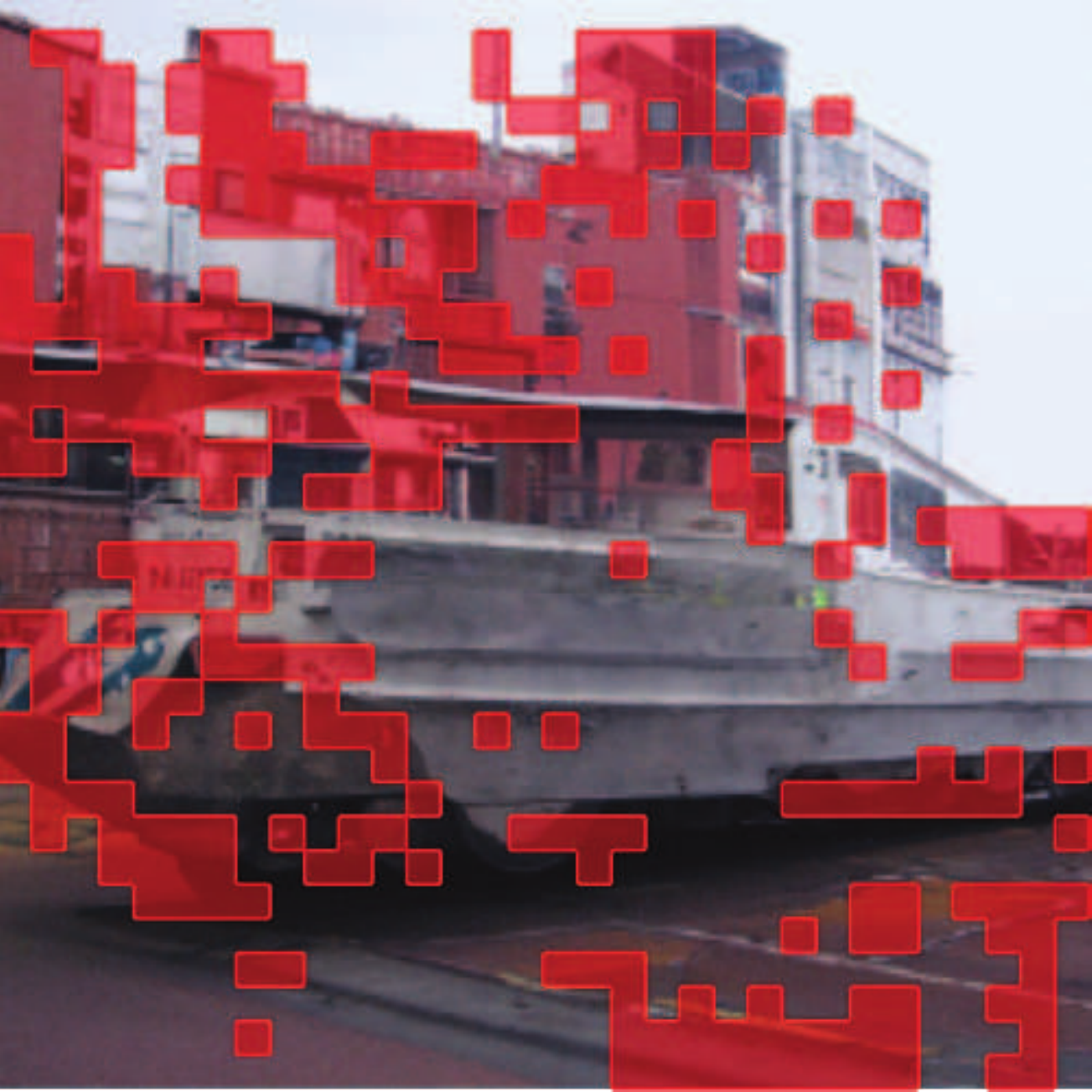}
\end{subfigure}\hspace{-0.25em}
\begin{subfigure}{.12\textwidth}
  \centering
  \includegraphics[width=1.0\linewidth]{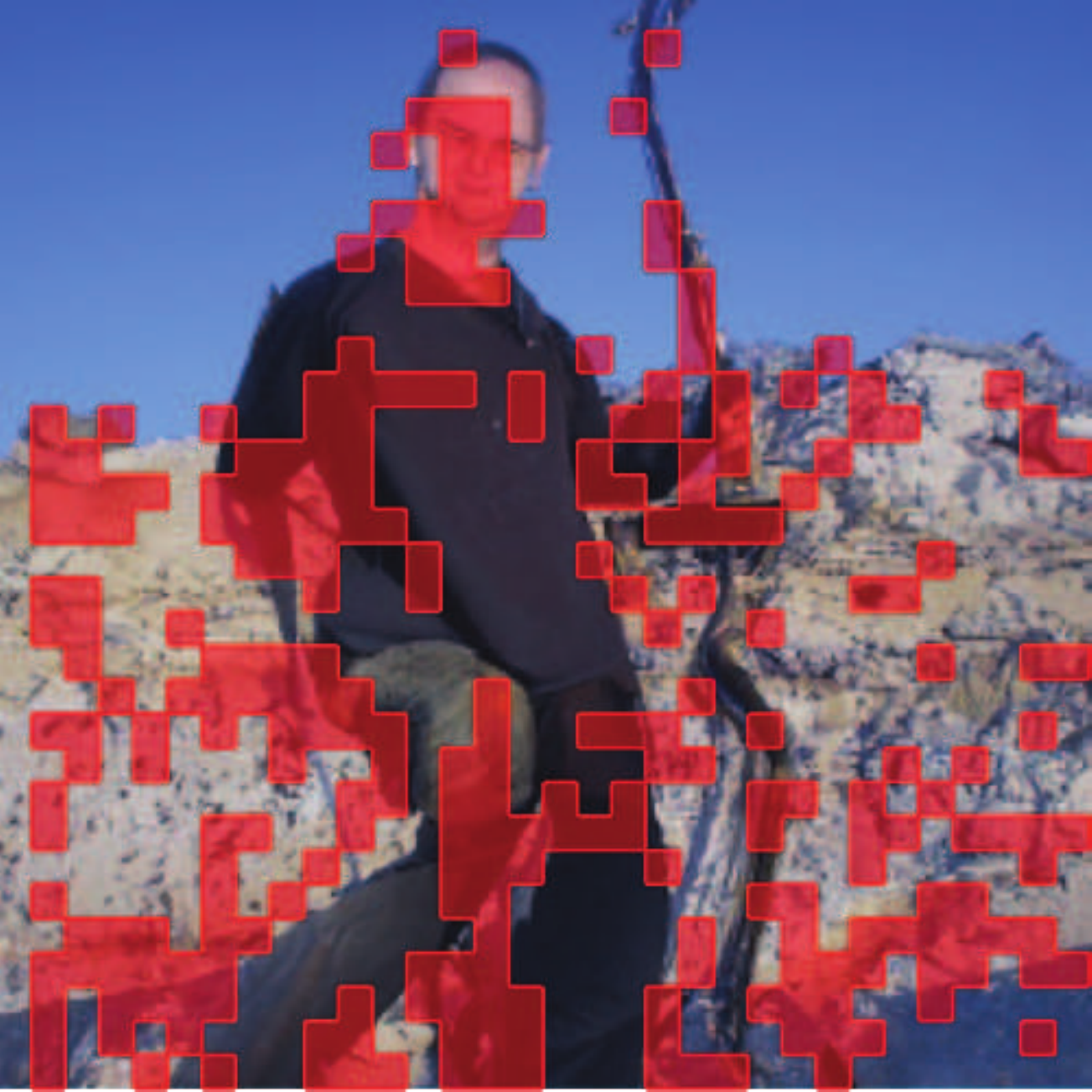}
\end{subfigure}\hspace{-0.25em}
\begin{subfigure}{.12\textwidth}
  \centering
  \includegraphics[width=1.0\linewidth]{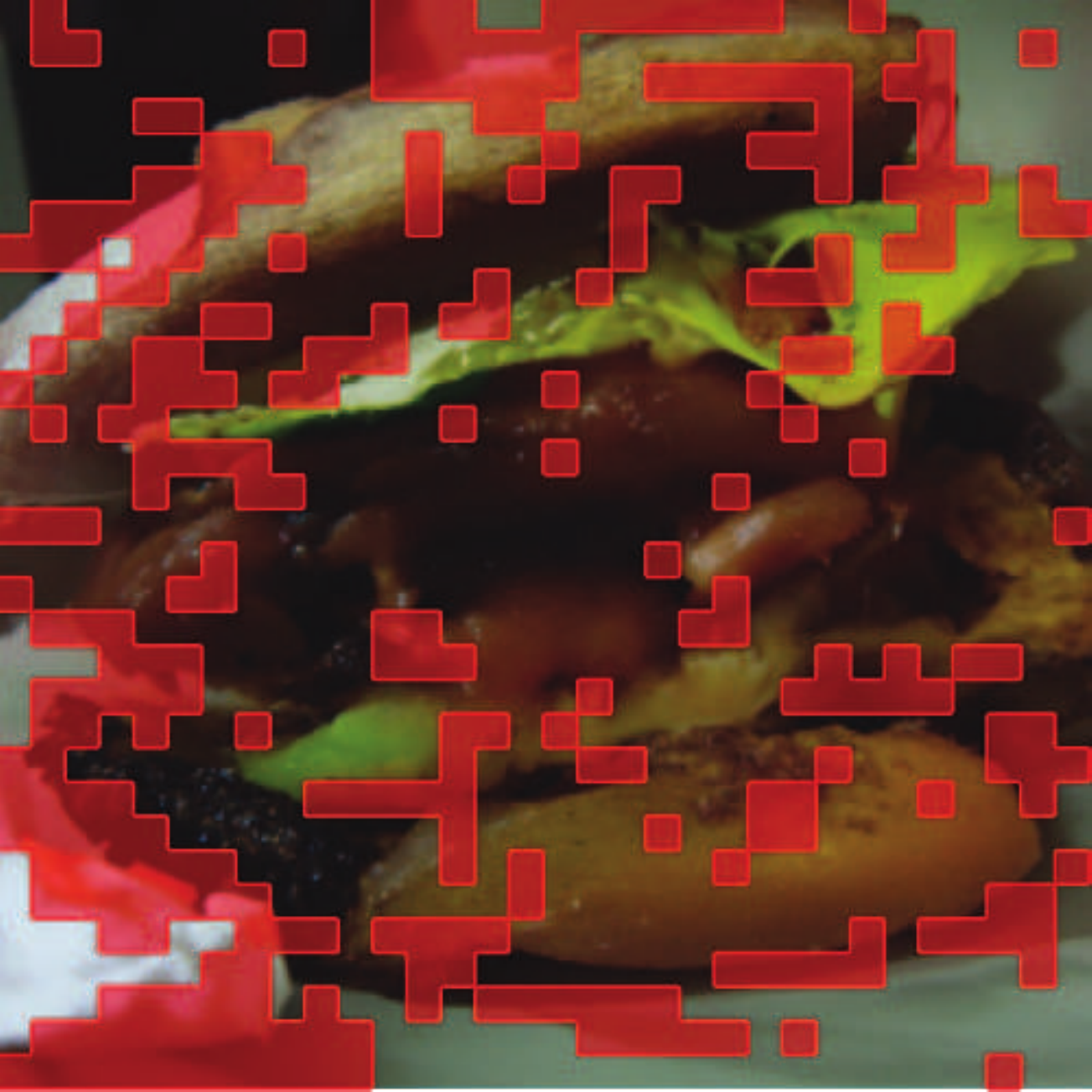}
\end{subfigure}
\vspace{-15pt}
\caption{\textbf{High-frequency masks (top) and the self-attention masks (bottom) of the finally generated images.} Note that the frequency masks are calculated after the generation process, while the self-attention masks are accumulated during the entire reverse process.}
\label{fig:attn-vis}\vspace{-10pt}
\end{figure}

\begin{table*}[t]
\small
\centering
\begin{tabular}{c|c|c|c|ccccc}
\noalign{\smallskip}\noalign{\smallskip}\toprule
Dataset & Input & \# of steps & SAG & FID ($\downarrow$) & sFID ($\downarrow$) & IS ($\uparrow$) & Precision ($\uparrow$) & Recall ($\uparrow$)\\
\midrule
\multirow{2}{*}{ImageNet 256$\times$256} & \multirow{2}{*}{Uncond.} & \multirow{2}{*}{250} &\xmark& 26.21 & 6.35 & 39.70 & 0.61 & \textbf{0.63} \\
&&&\cmark& \textbf{20.08} & \textbf{5.77} & \textbf{45.56} & \textbf{0.68} & 0.59 \\
\midrule
\multirow{2}{*}{ImageNet 256$\times$256} & \multirow{2}{*}{Cond.} & \multirow{2}{*}{250} &\xmark& 10.94 & 6.02 & 100.98 & 0.69 & \textbf{0.63} \\
&&&\cmark& \textbf{9.41} & \textbf{5.28} & \textbf{104.79} & \textbf{0.70} & 0.62 \\
\midrule
\multirow{2}{*}{LSUN Cat 256$\times$256} & \multirow{2}{*}{Uncond.} & \multirow{2}{*}{250} &\xmark& 7.03 & 8.24 & - & \textbf{0.60} & \textbf{0.53} \\
&&&\cmark& \textbf{6.87} & \textbf{8.21} & - & \textbf{0.60} & 0.50 \\
\midrule
\multirow{2}{*}{LSUN Horse 256$\times$256} & \multirow{2}{*}{Uncond.} & \multirow{2}{*}{250} &\xmark& 3.45 & 7.55 & - & \textbf{0.68} & \textbf{0.56} \\
&&&\cmark& \textbf{3.43} & \textbf{7.51} & - & \textbf{0.68} & 0.55 \\
\bottomrule
\end{tabular}
\vspace{-5pt}
\caption{\textbf{50K results of self-attention guidance on ADM~\cite{dhariwal2021diffusion} pre-trained on 256$\times$256 images.} The best values are in bold.} \vspace{-10pt}
\label{tab:main}
\end{table*}

\section{Utilizing the Self-Attention Map to Improve Sample Quality}
\label{sec:method}

The derivation presented in Section~\ref{sec:generalizing} implies that by extracting salient information ${\mathbf h}_t$ contained in ${\mathbf x}_t$, it is possible to provide guidance to the reverse process of diffusion models. Inspired by this implication, we propose an innovative guidance technique, Self-Attention Guidance (SAG), which effectively capture the salient information for reverse process while mitigating the risk of out-of-distribution issues of $\bar{\mathbf x}_t$ in pre-trained diffusion models. We first explain blur guidance, which is a primitive form of SAG, in Section~\ref{sec:blur}, and then we introduce SAG in Section~\ref{sec:sag}.

\subsection{Blur Guidance for Diffusion Models}\label{sec:blur}
Gaussian blur is a linear filtering technique that involves convolving an input signal $\hat{\mathbf{x}}_0$ with a Gaussian filter $G_\sigma$ to produce an output $\tilde{\mathbf{x}}_0$. Formally, $\tilde{\mathbf{x}}_0 = \hat{\mathbf{x}}_0 \ast G_\sigma$, where $\ast$ represents the convolution operator. As the standard deviation $\sigma$ increases, Gaussian blur reduces the fine-scale details within the input signals and smooths them towards constant~\cite{rissanen2022generative}, resulting in locally indistinguishable ones.

It is evident that there is an information imbalance between $\tilde{\mathbf{x}}_0$ and $\hat{\mathbf{x}}_0$, where $\hat{\mathbf{x}}_0$ contains more fine-scale information. Based on this fundamental insight, we introduce a specialized version of Eq.~\ref{eq:gg}, which we refer to as blur guidance in this paper. In essence, blur guidance intentionally excludes the information from intermediate reconstructions (Eq.~\ref{eq:x0_pred}) during the diffusion process, using this information to guide our predictions towards enhancing the relevance of the images to the information. In detail, blur guidance makes the original prediction deviate more from the prediction of the blurred input. Moreover, we note that Gaussian blur has a benign property in that it prevents the resulting signals from deviating significantly from the original manifold with a moderate $\sigma$, \ie, blurring occurs naturally in images~\cite{hoogeboom2022blurring,lee2022progressive,rissanen2022generative}, which makes Gaussian blur particularly suitable for its application to pre-trained diffusion models. These models generally include latent diffusion models~\cite{rombach2022high}, given that the spatial latents also contains low-level information such as local structures~\cite{esser2021taming,ommer2007learning}.

\begin{table}[t]
\centering
\small
\begin{tabular}{c|c|c|c|c}
\noalign{\smallskip}\noalign{\smallskip}\toprule
Schedule & Objective & Input & SAG & FID ($\downarrow$) \\
\midrule
\multirow{2}{*}{cosine} & \multirow{2}{*}{$L_{\textrm{hybrid}}$} & \multirow{2}{*}{Uncond.} & \xmark & 19.2 \\
 &&& \cmark & \textbf{18.0} \\
\bottomrule
\end{tabular}
\vspace{-5pt}
\caption{\textbf{A 50K result of self-attention guidance on IDDPM~\cite{nichol2021improved} pre-trained on ImageNet 64$\times$64.}} \vspace{-10pt}
\label{tab:iddpm}
\end{table}

\begin{figure}[t]
\center
\includegraphics[width=0.9\linewidth]{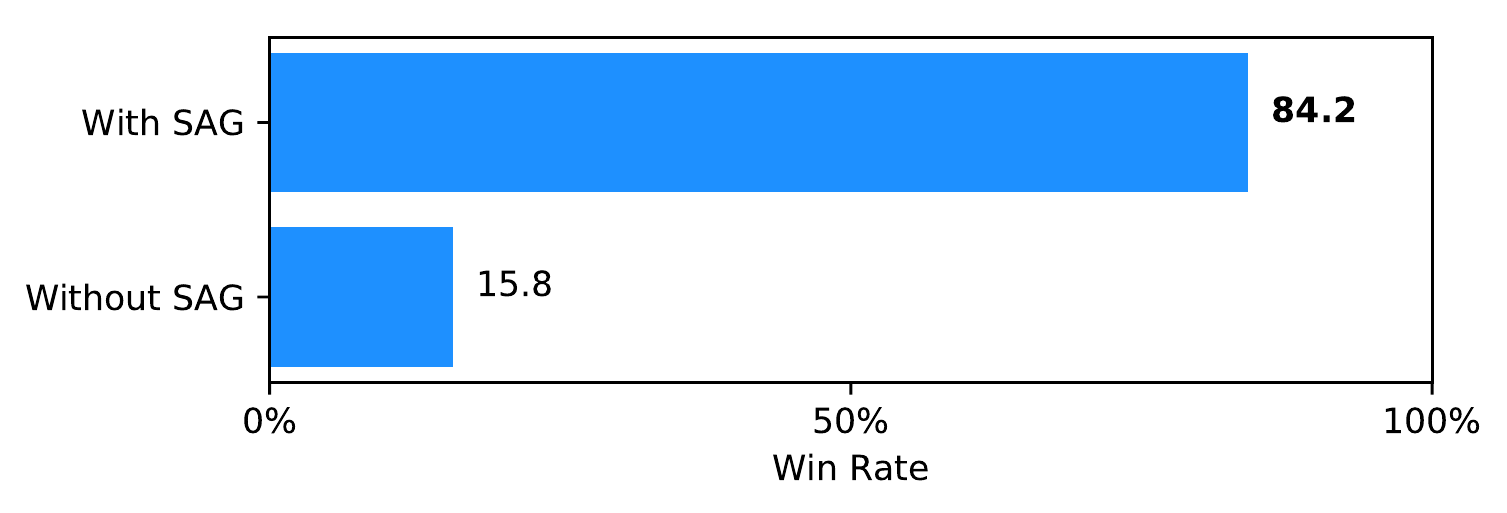}
\vspace{-5pt}
\caption{\textbf{A result of human evaluation on self-attention guidance with pairs sampled from Stable Diffusion~\cite{rombach2022high}.}}
\vspace{-10pt}
\label{fig:stable}
\end{figure}

To be specific, we first blur $\hat{\mathbf{x}}_0$ at Eq.~\ref{eq:x0_pred} with a Gaussian filter $G_\sigma$. Subsequently, we diffuse it again with the noise $\epsilon_{\theta}(\mathbf{x}_t)$ to produce $\tilde{\mathbf{x}}_t$. It is important to note that by doing this, we bypass the side effect of blur that reduces Gaussian noise, making the guidance depend on the intermediate content rather than the random noise. For brevity and to incorporate diffusion models in latent space~\cite{rombach2022high}, we let $\mathbf{x}_t$ represent either noised images or the spatial latents~\cite{esser2021taming,rombach2022high}.

The blur guidance is then incorporated into Eq.~\ref{eq:gg} by setting $\bar{\mathbf x}_t=\tilde{\mathbf x}_t$ and ${\mathbf h}_t={\mathbf{x}}_t - {\tilde{\mathbf{x}}_t}$. In practice, the joint input $(\tilde{\mathbf x}_t,\mathbf{h}_t)$ is simply computed as the summation $\mathbf{x}_t=\tilde{\mathbf x}_t+\mathbf{h}_t$. The term ${\mathbf{x}}_t - {\tilde{\mathbf{x}}_t}$ retains the information present before the blurring process, thus guiding the diffusion process to be more appropriate to the removed salient information in the original input. Our results, as shown in Table~\ref{tab:abl_masking} ``Global", demonstrate the effect of blur guidance in improving the baseline in terms of the quality metrics.

Despite its benefit with moderate guidance scales, the application of blur guidance on existing models with large guidance scales ($s>5.0$) produces noisy results, as shown in the top row of Fig.~\ref{fig:unstab}. We assume that this is because global blur introduces structural ambiguity across entire regions. This makes it difficult to align the prediction of the degraded input with that of the original, contributing to the noisy outcome accumulated over $t$. This issue highlights the need for a more adaptive approach that can capture finer and more relevant information during the reverse process than the global blurring.

\begin{figure}[t]
\begin{subfigure}{1.0\columnwidth}
    \includegraphics[width=1.0\linewidth]{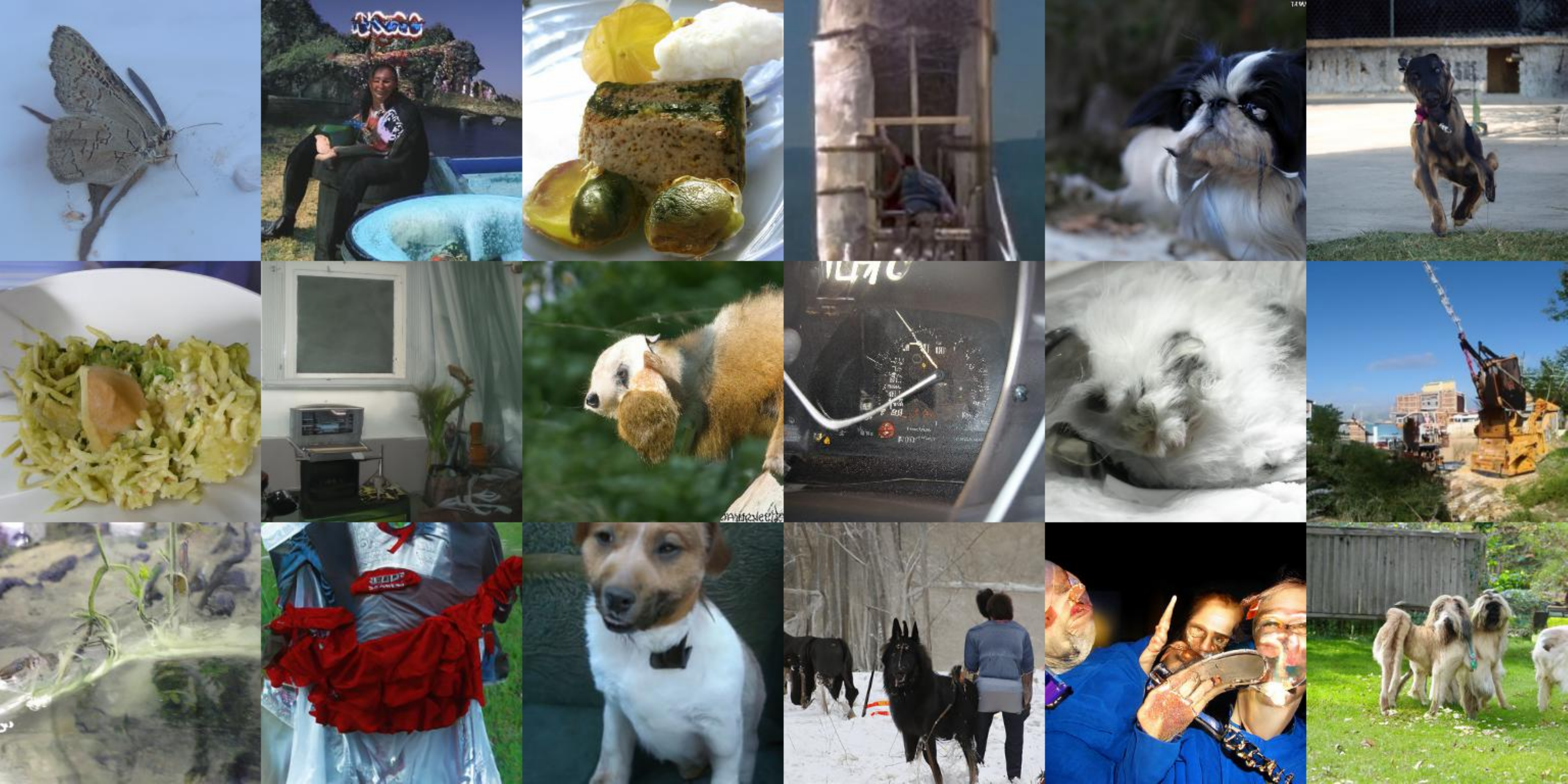}
\vspace{-15pt}
    \caption{Results from ADM~\cite{dhariwal2021diffusion}.}
\end{subfigure}
\begin{subfigure}{1.0\columnwidth}
  \includegraphics[width=1.0\linewidth]{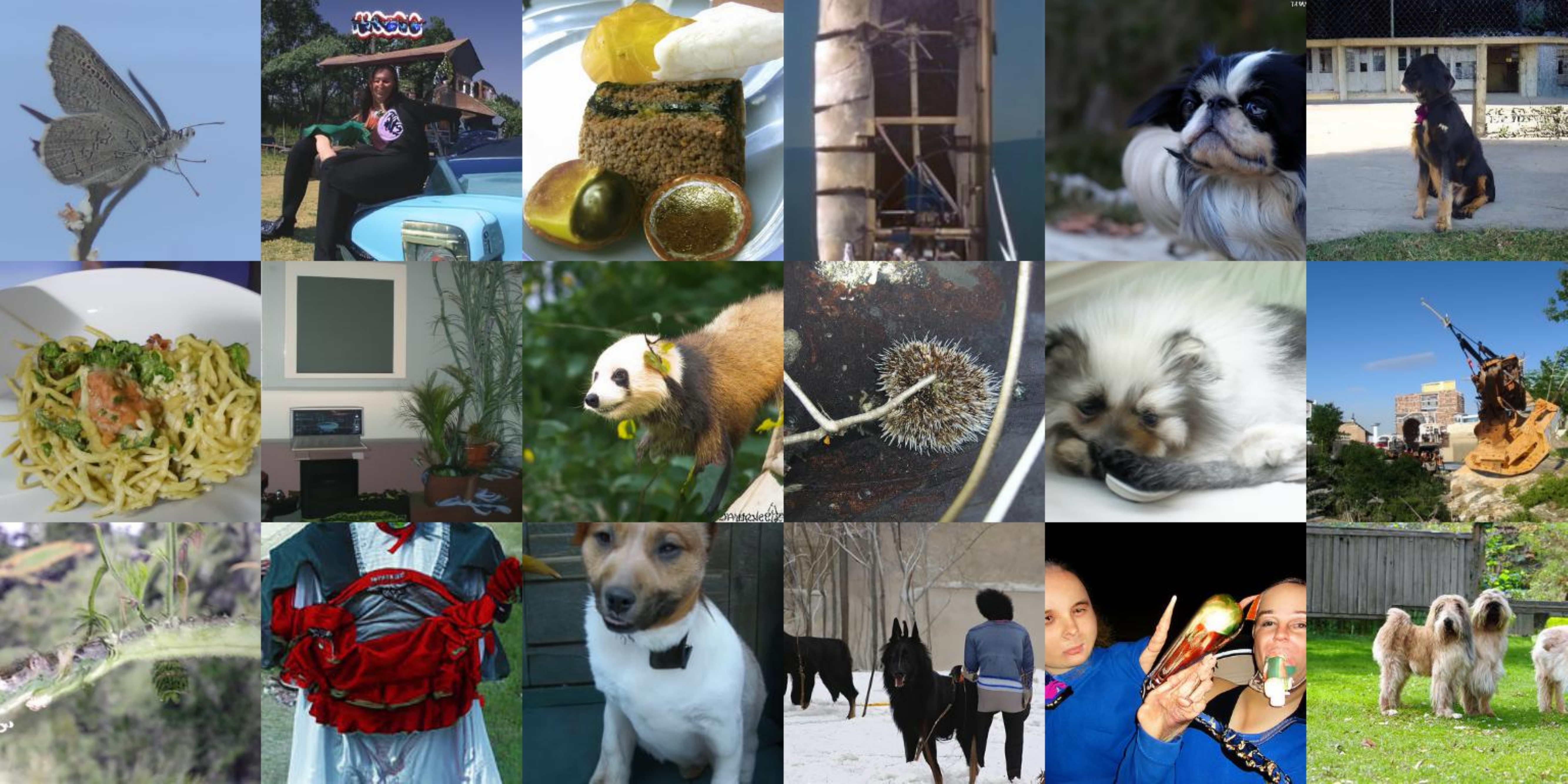}
\vspace{-15pt}
  \caption{Results from ADM~\cite{dhariwal2021diffusion} with our method (SAG).}
\end{subfigure}
\vspace{-20pt}
\caption{\textbf{Uncurated samples from ADM~\cite{dhariwal2021diffusion} without and with our method (SAG).} Both results are sampled from unconditional ADM pre-trained on ImageNet 256$\times$256~\cite{deng2009imagenet}, and share the same random seed. The samples guided by SAG typically show fewer artifacts, benefiting from the self-conditioning of the internal conditions.}
\label{fig:unsel_compare}
\vspace{-10pt}
\end{figure}

\begin{figure}[t]
\begin{subfigure}{1.0\columnwidth}
    \includegraphics[width=1.0\linewidth]{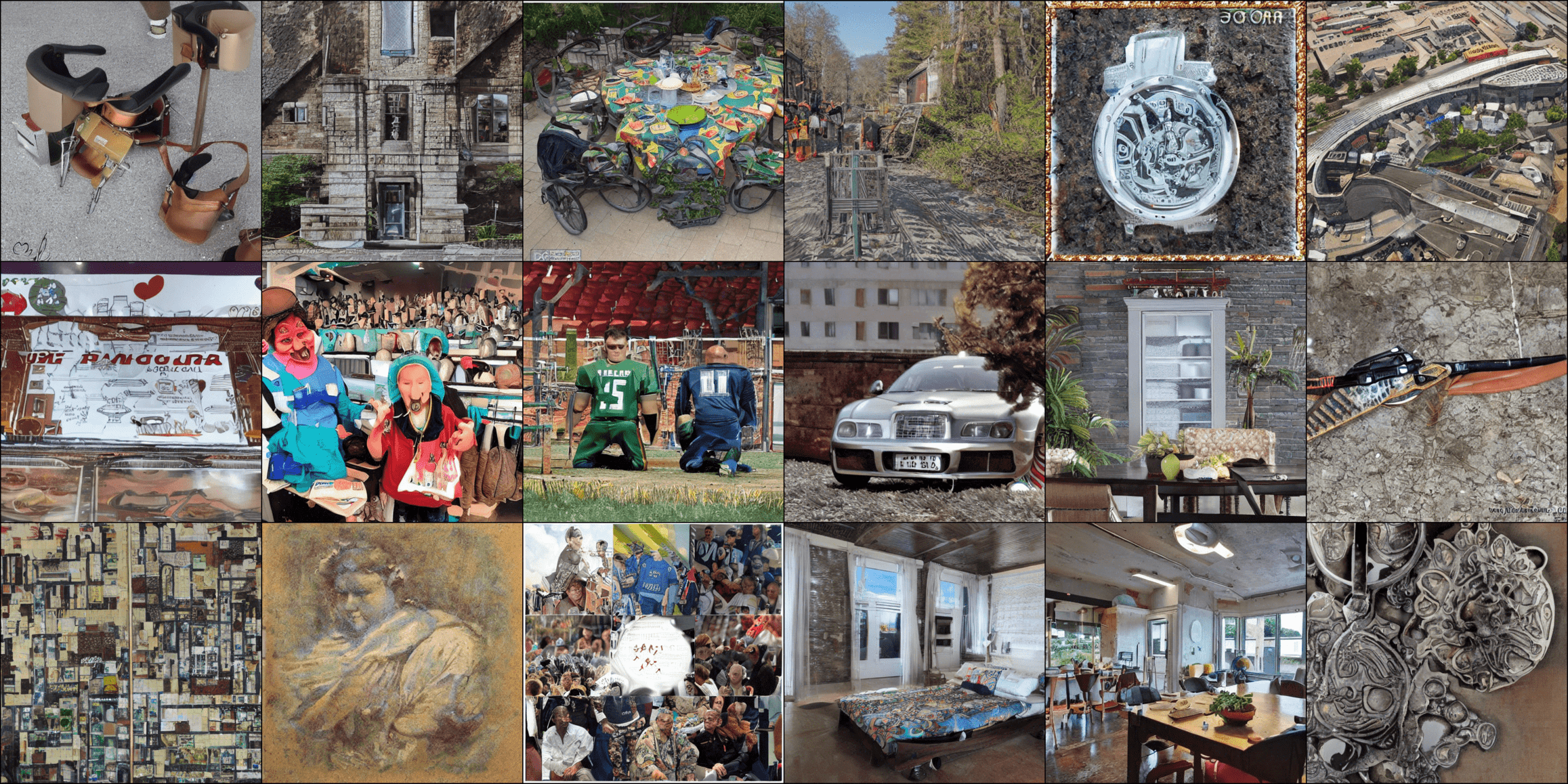}
\vspace{-15pt}
    \caption{Results from Stable Diffusion~\cite{rombach2022high}.}
\end{subfigure}
\begin{subfigure}{1.0\columnwidth}
  \includegraphics[width=1.0\linewidth]{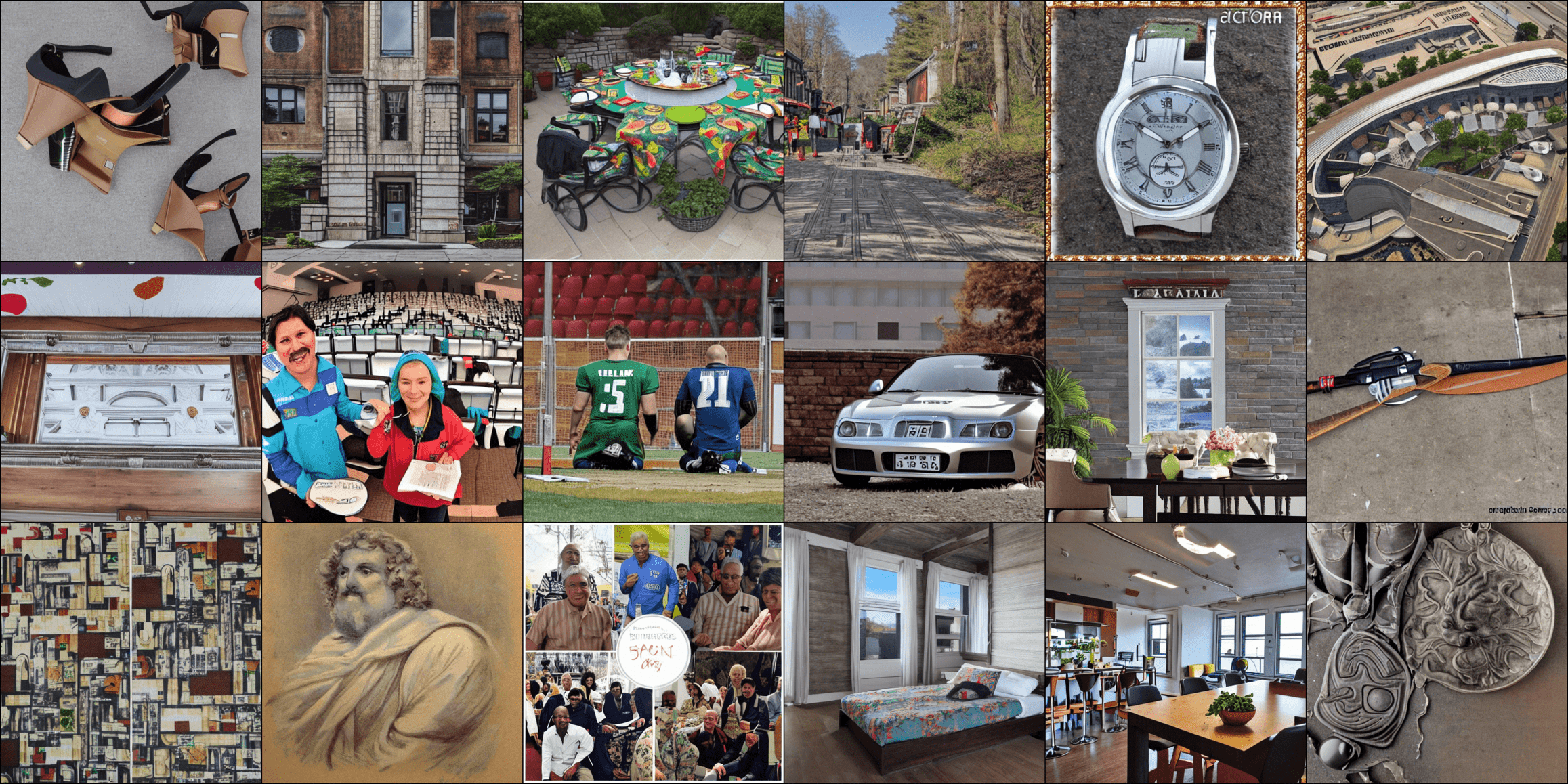}
\vspace{-15pt}
  \caption{Results from Stable Diffusion~\cite{rombach2022high} with our method (SAG).}
\end{subfigure}
\vspace{-20pt}
\caption{\textbf{Uncurated samples from Stable Diffusion~\cite{rombach2022high} without and with our method (SAG).} Both results are sampled from Stable Diffusion, and share the same random seed. The prompt is set to an empty prompt with a single space (`` '').}
\label{fig:unsel_sd}
\vspace{-10pt}
\end{figure}

\subsection{Self-Attention Guidance for Diffusion Models}\label{sec:sag}

The self-attention mechanism~\cite{dosovitskiy2020image,vaswani2017attention} has been shown to be a key component of diffusion models~\cite{dhariwal2021diffusion,ho2020denoising}. This mechanism, which is implemented in the backbones of diffusion models, allows the model to attend to salient parts of the input during the generative process~\cite{jiang2021transgan,zhang2022styleswin,zhang2019self,hertz2022prompt}. A particular example of the information capture is illustrated in Fig.~\ref{fig:attn-vis}, which shows that the region of the self-attention masks from ADM~\cite{dhariwal2021diffusion} overlaps with the high-frequency details that diffusion models ought to elaborate on and that are one of essential factors of image generation~\cite{chen2021ssd,xu2022dhg,yang2022wavegan} and human perception~\cite{de2013image}. See the appendix for more examples and analyses.

Building upon this intuition, we propose SAG, which leverages the self-attention maps of diffusion models. In essence, we adversarially blur self-attended patches of ${\mathbf x}_t$, \ie, conceal the information of patches that diffusion models attend to. We then use the concealed information to guide diffusion models. In addition, it can be shown that $\bar{\mathbf{x}}_t$ of self-attention guidance contains intact regions of ${\mathbf{x}}_t$, which means that it does not cause the structural ambiguity of the inputs and thus mitigates the problem of global blur.

To obtain the aggregated self-attention map from Eq.~\ref{eq:attn-map}, we conduct global average pooling (GAP) to aggregate the stacked self-attention maps $A^{S}_t \in \mathbb{R}^{N\times(HW)\times(HW)}$ to the dimension $ \mathbb{R}^{HW}$, followed by reshaping to $\mathbb{R}^{H \times W}$ and subsequent nearest-neighbor upsampling to match the resolution of $\mathbf{x}_t$:
\begin{equation}
{A}_t = \textrm{Upsample}(\textrm{Reshape}(\textrm{GAP}(A^{S}_t))).
\label{eq:real-attn-map}
\end{equation}
Generalizing blur guidance, given a masking threshold $\psi$, which is practically set to the mean value of $A_t$, SAG blurs only the masked patches of $\mathbf{x}_t$ according to the self-attention map and is formulated as follows:
\begin{align}
&M_t = \mathbbm{1}(A_t > \psi),\\
&{\widehat{\mathbf{x}}}_t=(1-M_t)\odot {\mathbf{x}}_t+M_t\odot \tilde{\mathbf{x}}_t,~\label{eq:sel-blur}\\
&\tilde\epsilon(\mathbf{x}_t) = \epsilon_\theta(\widehat{{\mathbf{x}}}_t) + (1+s) (\epsilon_\theta(\mathbf{x}_t) - \epsilon_\theta({\widehat{\mathbf{x}}}_t)),~\label{eq:sag-ori}
\end{align}
where $\odot$ denotes the Hadamard product and $\tilde{\mathbf{x}}_t$ is obtained in the same manner as that in Sec.~\ref{sec:blur}. Note that Eq.~\ref{eq:sag-ori} is also a special case of Eq.~\ref{eq:gg} where ${\mathbf{h}}_t = M_t\odot {\mathbf{x}}_t - M_t\odot \tilde{\mathbf{x}}_t$, $\bar{\mathbf x}_t={\widehat{\mathbf x}}_t$, and the joint input undergoes the simple summation as in Sec.~\ref{sec:blur}. Unlike blur guidance, ${\widehat{\mathbf{x}}}_t$ explicitly contains intact patches of ${\mathbf{x}}_t$, preventing the output $\epsilon_\theta(\widehat{{\mathbf{x}}}_t)$ from deviating too far from the original with even a large scale (Fig.~\ref{fig:unstab}) as well as effectively concealing the information critical for the reverse process in an adversarial manner.

\begin{figure}[t]
\captionsetup[subfigure]{labelformat=empty}
\captionsetup[subfigure]{justification=centering}
\begin{subfigure}{.12\textwidth}
  \centering
  \includegraphics[width=1.0\linewidth]{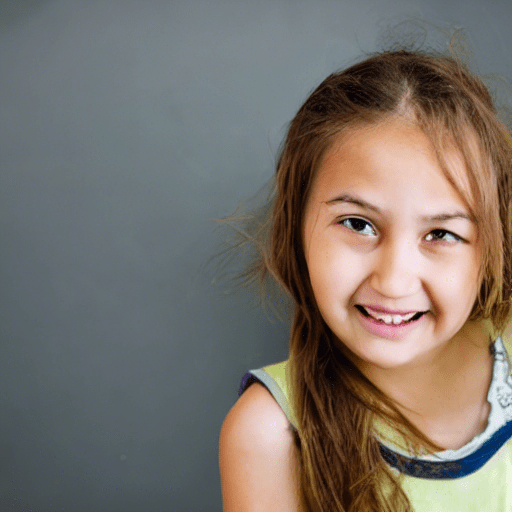}
\end{subfigure}\hspace{-0.25em}
\begin{subfigure}{.12\textwidth}
  \centering
  \includegraphics[width=1.0\linewidth]{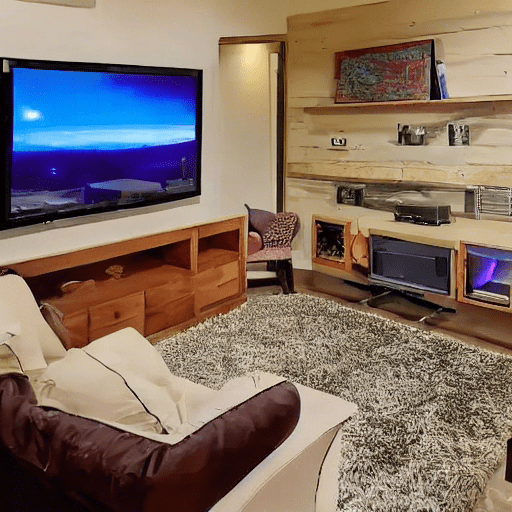}
\end{subfigure}\hspace{-0.25em}
\begin{subfigure}{.12\textwidth}
  \centering
  \includegraphics[width=1.0\linewidth]{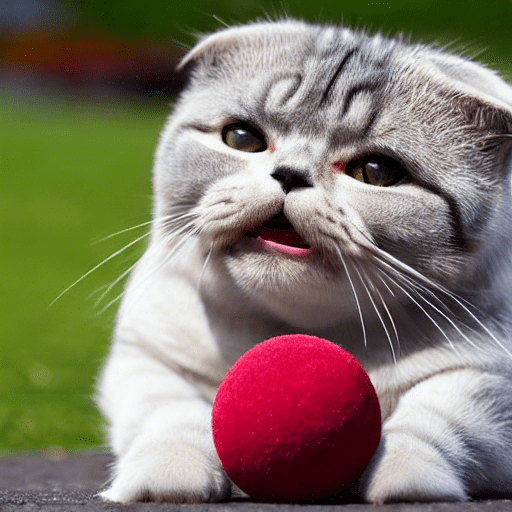}
\end{subfigure}\hspace{-0.25em}
\begin{subfigure}{.12\textwidth}
  \centering
  \includegraphics[width=1.0\linewidth]{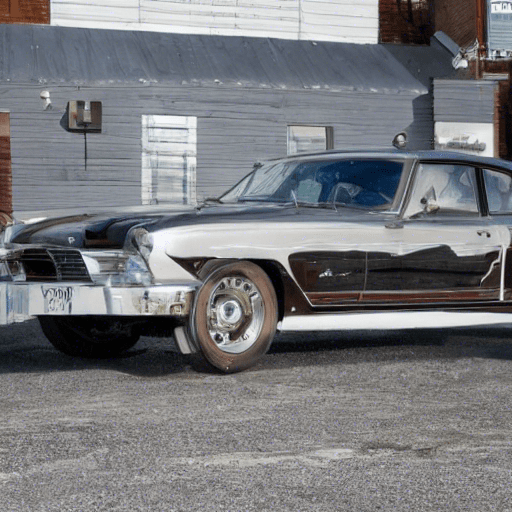}
\end{subfigure}\vspace{-0.9pt}\\
\begin{subfigure}{.12\textwidth}
  \centering
  \includegraphics[width=1.0\linewidth]{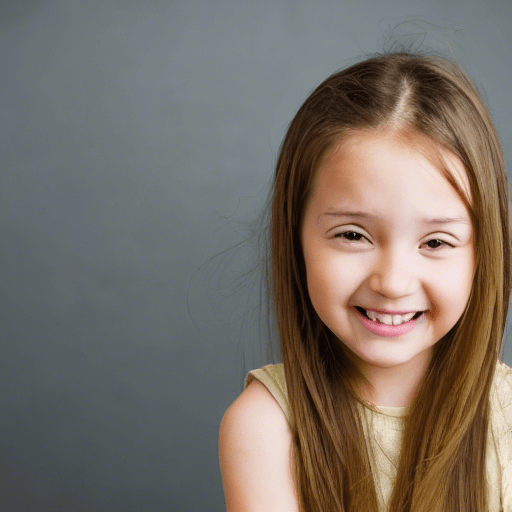}
  \caption{``A girl\\ showing a\\ smiling face."}
\end{subfigure}\hspace{-0.25em}
\begin{subfigure}{.12\textwidth}
  \centering
  \includegraphics[width=1.0\linewidth]{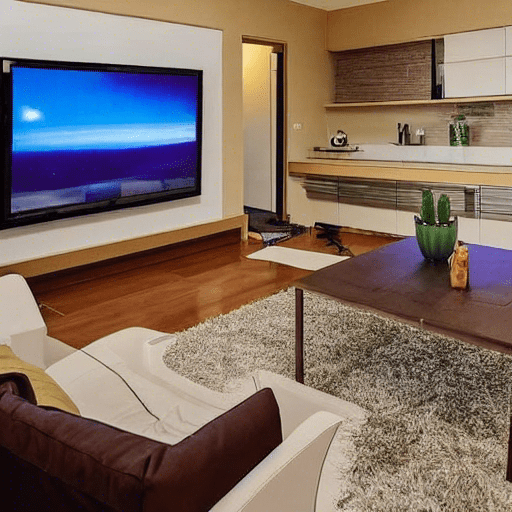}
  \caption{``A living area with a television and a table."}
\end{subfigure}\hspace{-0.25em}
\begin{subfigure}{.12\textwidth}
  \centering
  \includegraphics[width=1.0\linewidth]{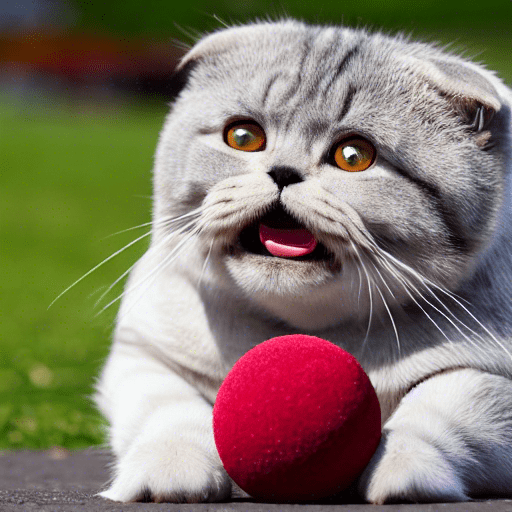}
  \caption{``A Scottish Fold playing with a ball."}
\end{subfigure}\hspace{-0.25em}
\begin{subfigure}{.12\textwidth}
  \centering
  \includegraphics[width=1.0\linewidth]{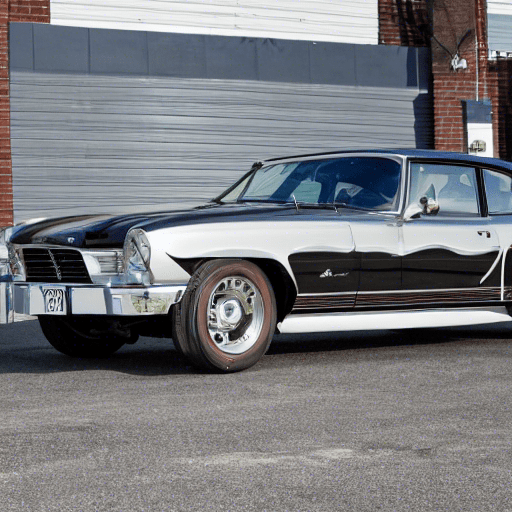}
  \caption{`` "\linebreak\linebreak}
\end{subfigure}
\vspace{-20pt}
\caption{\textbf{Text-to-image results of Stable Diffusion~\cite{rombach2022high}, where the top row is sampled with only CFG and the bottom row with CFG and SAG.} SAG helps the model generate a high-quality image that is more self-conditioned and has fewer artifacts even with an empty prompt (4th column), exhibiting independence from external information.}
\label{fig:stable-t2i}\vspace{-10pt}
\end{figure}

\section{Experiments}
\subsection{Experimental Settings}

For the experiments, we use two servers with 8 NVIDIA GeForce RTX 3090 GPUs each to sample from. We build upon the pre-trained models of ADM~\cite{dhariwal2021diffusion}, IDDPM~\cite{nichol2021improved}, Stable Diffusion~\cite{rombach2022high}, and DiT~\cite{peebles2022scalable}. We take all the weights for our experiments from their publicly available repositories, and use the same evaluation metrics as \cite{dhariwal2021diffusion}, including FID~\cite{heusel2017gans}, sFID~\cite{nash2021generating}, IS~\cite{salimans2016improved}, and Improved Precision and Recall~\cite{kynkaanniemi2019improved}.

\subsection{Experimental Results}

\paragraph{Unconditional generation with SAG.}
We show the effectiveness of SAG on the unconditional models, which demonstrates our condition-free property that CG and CFG do not possess. We use unconditionally pre-trained ADM~\cite{dhariwal2021diffusion} and IDDPM~\cite{nichol2021improved} for this experiment, and evaluate 50k samples for the metrics.

We evaluate pre-trained ADM~\cite{dhariwal2021diffusion} on ImageNet~\cite{deng2009imagenet} 256$\times$256, LSUN Cat~\cite{yu2015lsun}, and LSUN Horse~\cite{yu2015lsun}. As shown in Table~\ref{tab:main}, we observe that SAG consistently improves the FID, sFID and IS of unconditional, while it lowers the recall. As explained in recent studies~\cite{dhariwal2021diffusion,ho2021classifier}, we suspect for the lower recall that there also exists a trade-off relationship between sample fidelity and diversity. Nevertheless, the qualitative improvement is made due to the self-conditioning of our method, as we can see the comparison of unselected samples in Fig.~\ref{fig:unsel_compare}.

Subsequently, we include the results of the unconditional model of IDDPM~\cite{nichol2021improved} equipped with the proposed method, which is trained on ImageNet at resolution 64$\times$64. The result is in Table~\ref{tab:iddpm}, which also shows an improvement in terms of FID by applying SAG.
\vspace{-10pt}

\begin{table}[t]
\footnotesize
\centering
\begin{tabular}{cc|cccc}
\noalign{\smallskip}\noalign{\smallskip}\toprule
CG~\cite{dhariwal2021diffusion} & SAG & FID ($\downarrow$) & sFID ($\downarrow$) & Precision ($\uparrow$) & Recall ($\uparrow$) \\
\midrule
\xmark & \xmark & 5.91 & 5.09 & 0.70 & \textbf{0.65} \\
\cmark & \xmark & 2.97 & 5.09 & 0.78 & 0.59 \\
\xmark & \cmark & 5.11 & \textbf{4.09} & 0.72 & \textbf{0.65} \\
\cmark & \cmark & \textbf{2.58} & 4.35 & \textbf{0.79} & 0.59 \\
\bottomrule
\end{tabular}
\vspace{-5pt}
\caption{\textbf{Compatibility of SAG with CG~\cite{dhariwal2021diffusion}.} The results are from ADM trained on ImageNet 128$\times$128.} \vspace{-10pt}
\label{tab:abl_compat}
\end{table}

\begin{table}[t]
\centering
\footnotesize
\begin{tabular}{c|cc|c}
\noalign{\smallskip}\noalign{\smallskip}\toprule
Model & CFG~\cite{ho2021classifier} & SAG & FID ($\downarrow$) \\
\midrule
\multirow{2}{*}{DiT-XL/2~\cite{peebles2022scalable}} & \cmark & \xmark & 2.27 \\
                                                & \cmark & \cmark & \textbf{2.16} \\
\bottomrule
\end{tabular}
\vspace{-5pt}
\caption{\textbf{Compatibility of SAG with CFG~\cite{ho2021classifier}.} The results are from DiT-XL/2 trained on ImageNet 256$\times$256.} \vspace{-10pt}
\label{tab:dit_cfg}
\end{table}

\begin{table}[t]
\footnotesize
\centering
\begin{tabular}{c|cc}
\noalign{\smallskip}\noalign{\smallskip}\toprule
Masking strategy & FID ($\downarrow$) & IS ($\uparrow$) \\
\midrule
Baseline & 5.98 & 141.72 \\
\midrule
Global (blur guidance in Sec.~\ref{sec:blur}) & 5.82 & 143.15 \\
High-frequency & 5.74 & 148.87 \\
Random & 5.68 & 148.99 \\
Square & 5.68 & 146.50 \\
Self-attention (SAG in Sec.~\ref{sec:sag}) & \textbf{5.47} & \textbf{151.12} \\
DINO~\cite{caron2021emerging}-attention & 5.63 & 146.18 \\
\bottomrule
\end{tabular}
\vspace{-5pt}
\caption{\textbf{Ablation study of the masking strategy.} The results are from ADM trained on ImageNet 128$\times$128.} \vspace{-10pt}
\label{tab:abl_masking}
\end{table}

\paragraph{Conditional generation with SAG.}
While our method is effective on unconditional models, Eq.~\ref{eq:gg} implies the condition-agnosticity, meaning that SAG can also be applied to conditional models. To evaluate SAG on conditional models, we perform an experiment on ADM~\cite{dhariwal2021diffusion} that is conditionally trained on ImageNet 256x256. The results are presented in Table~\ref{tab:main}, which demonstrates a similar effect on conditional models as on unconditional ones.
\vspace{-10pt}

\paragraph{Stable Diffusion with SAG.}

We compare our results with Stable Diffusion~\cite{rombach2022high} using human evaluation (see the appendix for the protocol) on 500 pairs of images with and without SAG. We use empty prompt for Stable Diffusion and the same random seed for each pair. The results show samples with SAG are more visually favorable or realistic to human. See Fig.~\ref{fig:stable} and Fig.~\ref{fig:unsel_sd}.

In addition, we also broaden the range to text-to-image (T2I) generation by utilizing Stable Diffusion and fusing CFG with SAG, although SAG is not intended for the specific T2I task. Notably, in Fig.~\ref{fig:stable-t2i}, the image samples generated from the model with SAG show higher quality and fewer artifacts due to the self-conditioning effect of SAG. Interestingly, even with an empty prompt (Fig.~\ref{fig:unsel_sd} and Fig.~\ref{fig:stable-t2i} 4th column), we observe an obviously improved quality. This corroborates the independence of SAG with an external condition.

\begin{figure}[t]
\begin{center}
\begin{subfigure}{.23\textwidth}
  \centering
  \includegraphics[width=1.0\linewidth]{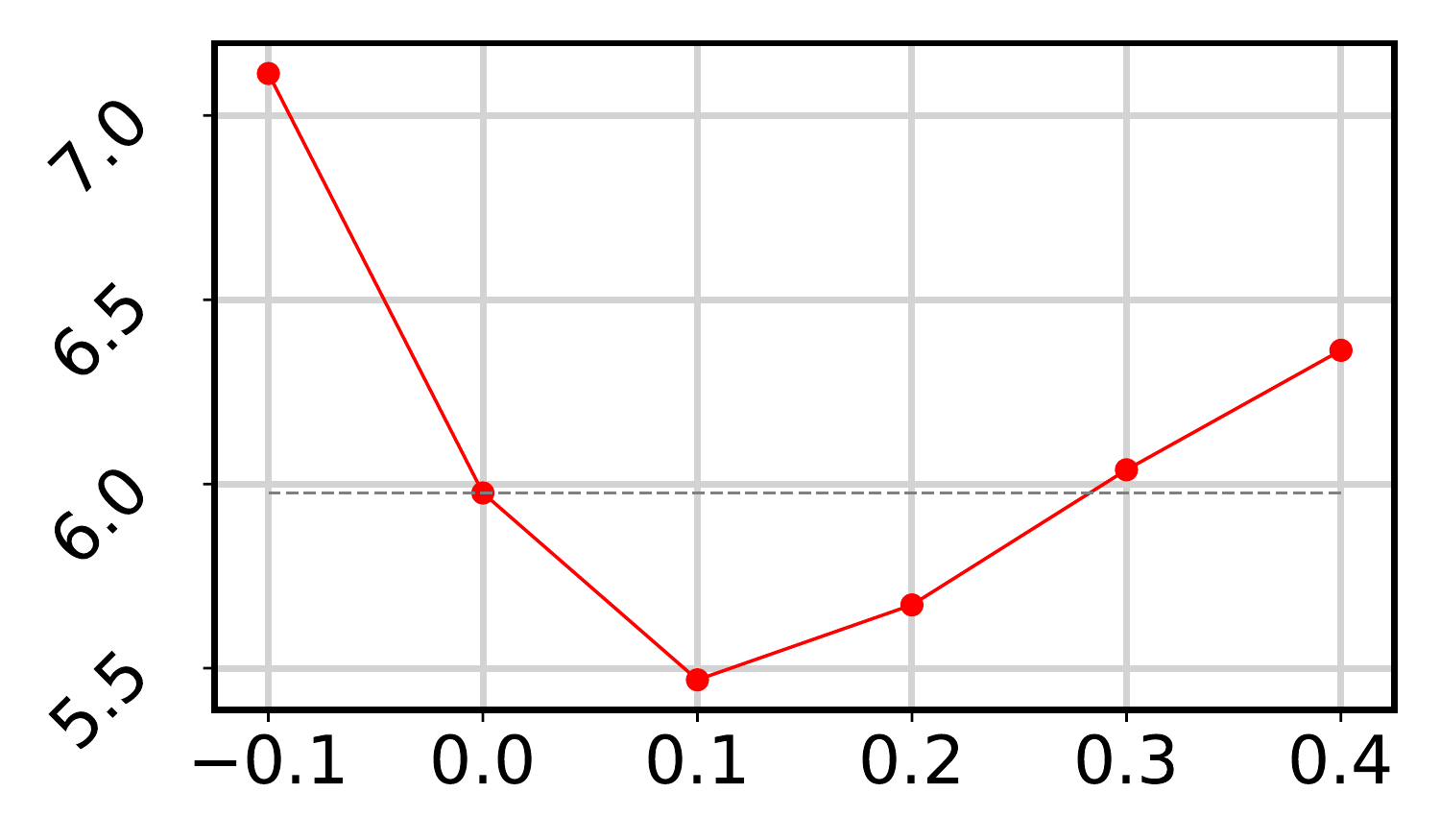}
  \caption{FID}
\end{subfigure}
\begin{subfigure}{.23\textwidth}
  \centering
  \includegraphics[width=1.0\linewidth]{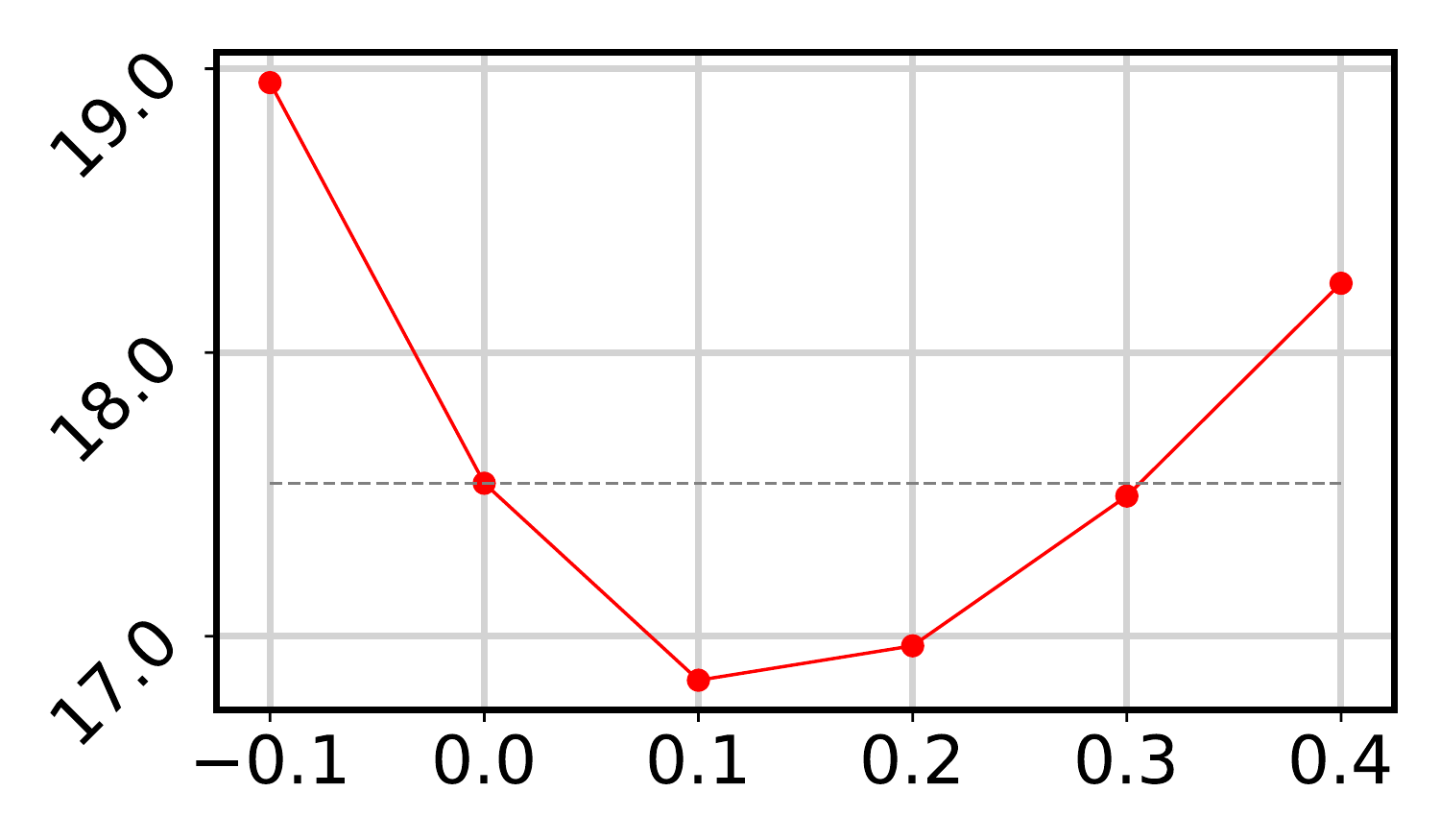}
  \caption{sFID}
\end{subfigure}\\
\begin{subfigure}{.23\textwidth}
  \centering
  \includegraphics[width=1.0\linewidth]{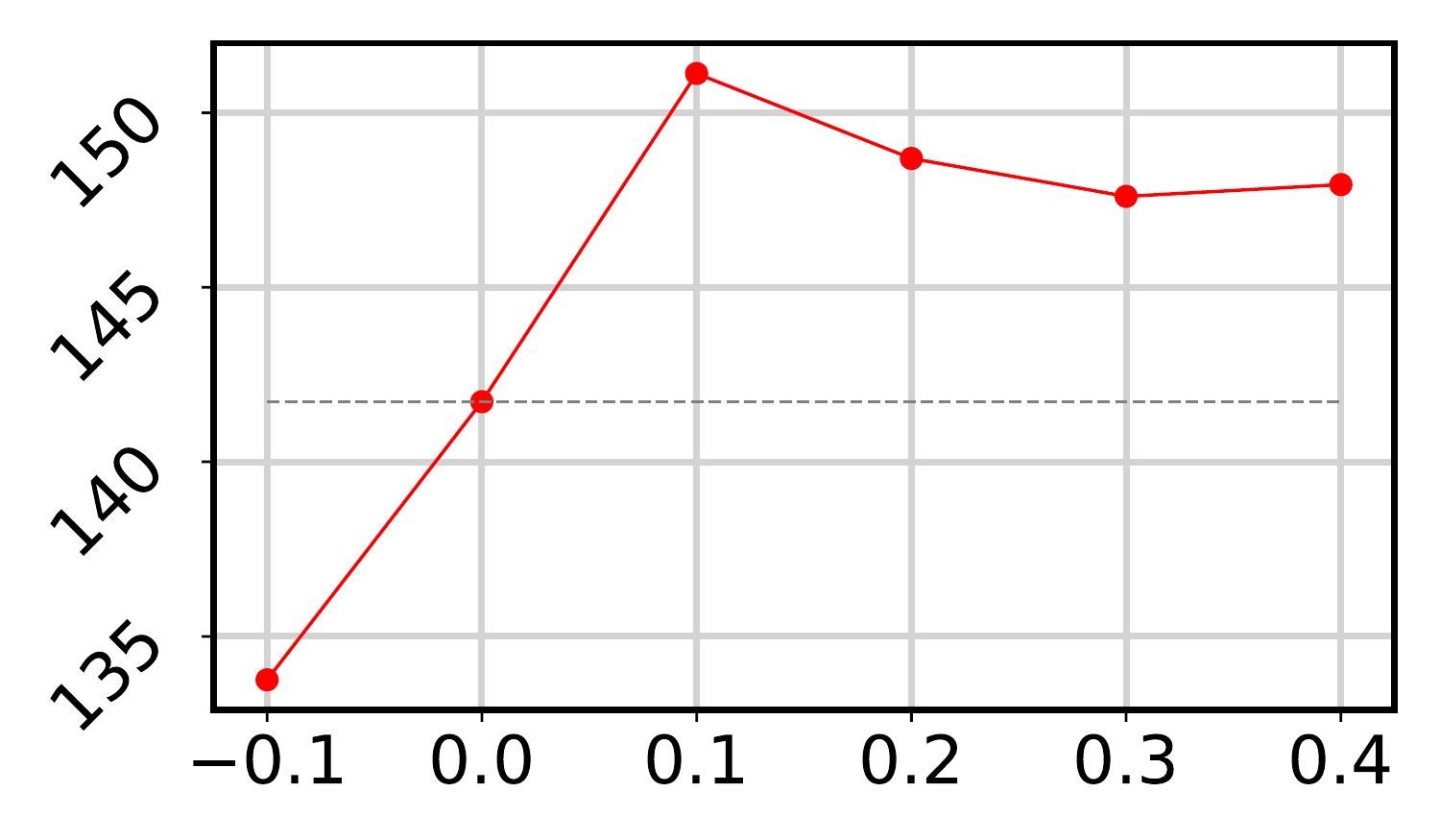}
  \caption{Inception Score}
\end{subfigure}
\begin{subfigure}{.23\textwidth}
  \centering
  \includegraphics[width=1.0\linewidth]{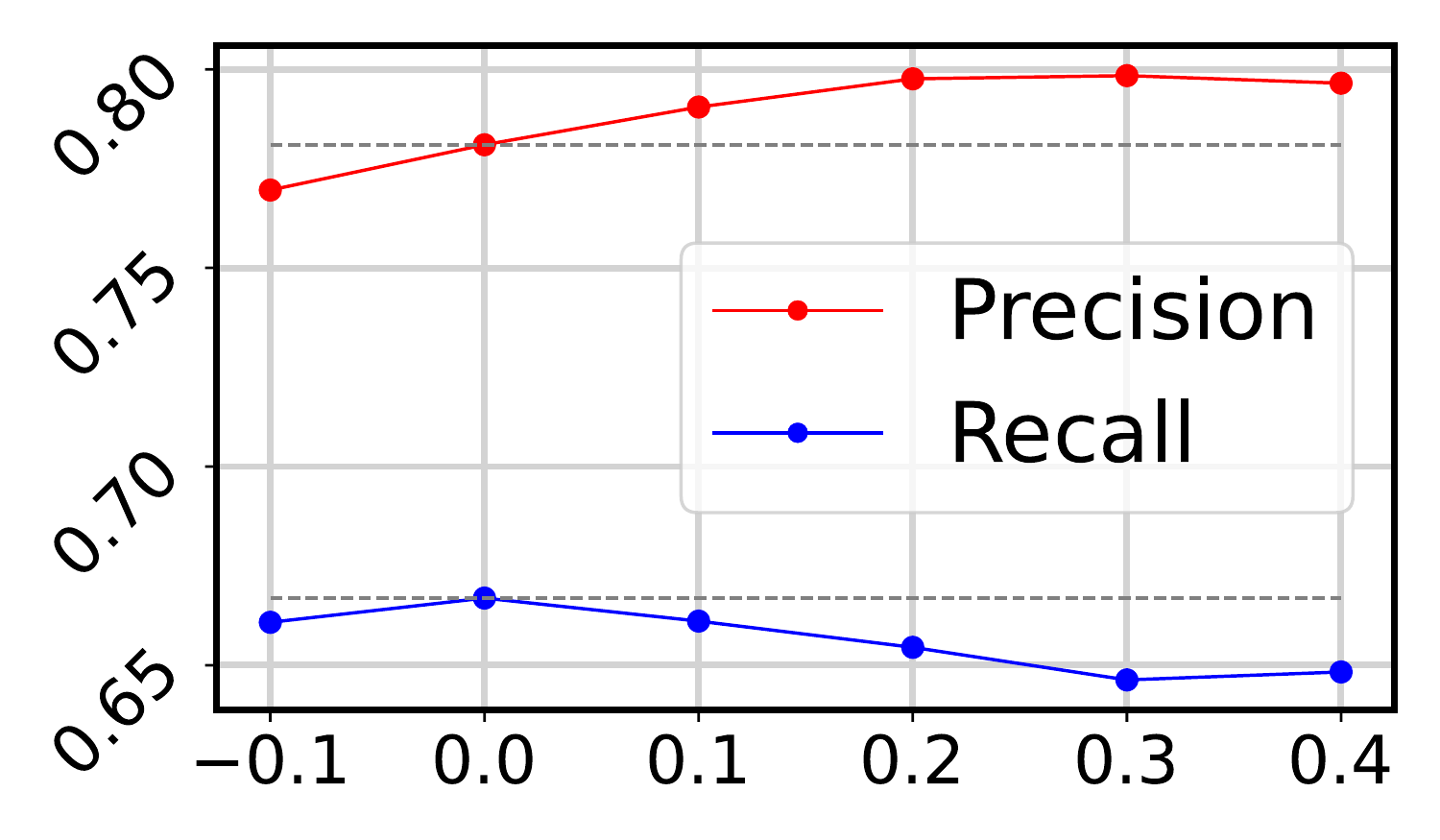}
  \caption{Precision \& Recall}
\end{subfigure}
\end{center}
\vspace{-15pt}
\caption{\textbf{Ablation study of the guidance scale.} The x-axis is guidance scale, and the dotted line denotes the performance of the baseline, \ie, the model without SAG. The results are from ADM trained on ImageNet 128$\times$128.} \vspace{-10pt}
\label{fig:abl_guide_scale}
\end{figure}

\subsection{Ablation Studies and Analyses}
\label{sec:abl}

\paragraph{Orthogonality with CG and CFG.}
Although designing SAG for unguided models, we can combine it with CG~\cite{dhariwal2021diffusion} that utilizes external conditions to further improve the performance. To this end, we test four cases to use the guidance, with or without CG and SAG. The metrics are evaluated on 50k samples generated by the ImageNet 128$\times$128 model~\cite{dhariwal2021diffusion}. As shown in Table~\ref{tab:abl_compat}, we observe additional improvements in FID and precision when using both of them, yet in terms of sFID only giving SAG is the best. This implies that SAG have an orthogonal component with and can be used simultaneously with traditional guidance.

Moreover, CFG~\cite{ho2021classifier} is another method of providing class-conditional guidance. However, it requires diffusion models to be trained in a specific manner. Therefore, we use DiT-XL/2~\cite{peebles2022scalable}, a Transformer~\cite{vaswani2017attention}-based model which has self-attention layers as well. The 50k results are presented in Table~\ref{tab:dit_cfg}. They show that samples guided by CFG also benefit from the self-conditioning effect of SAG. Note that the combined effect of SAG and CFG is also corroborated by text-to-image samples in Fig.~\ref{fig:stable-t2i}.
\vspace{-10pt}

\paragraph{Masking strategy.}
We test various masking strategies to verify the effectiveness of our self-attention masking with 10k samples on ADM~\cite{dhariwal2021diffusion}. Those strategies replace the masking function of SAG at each timestep. For a fair comparison, we mask 40\% of the pixels of the image for the other masking schemes, which is the equivalent portion of the masked area when the threshold of the self-attention masking is $1.0$. The results are in Table~\ref{tab:abl_masking}. We find that the self-attention masking strategy outperforms other masking strategies. Notably, applying global masking, \ie blur guidance, shows the worst performance among the schemes, which validates the motivation for SAG. In addition, we applied the high-frequency mask using FFT on $\hat{\mathbf{x}}_0$, as well as the self-attention mask of DINO~\cite{caron2021emerging}. However, these methods demonstrated worse performance than ours in terms of FID and IS metrics. Therefore, this result indicates that the self-attention masking is a sufficiently effective method.
\vspace{-10pt}

\begin{table}[t]
\centering
\resizebox{1.0\linewidth}{!}{
\begin{tabular}{c|c|cccc|c}
\toprule
\multirow{2}{*}{$\sigma$} & Baseline & \multirow{2}{*}{$\sigma=1$} & \multirow{2}{*}{$\sigma=3$} & \multirow{2}{*}{$\sigma=9$} & \multirow{2}{*}{$\sigma=27$} & Avg. pixel\\
& ($\sigma\rightarrow 0$) & & & & & ($\sigma\rightarrow \infty$)\\
\midrule
FID ($\downarrow$)& 5.98 & 5.58 & \textbf{5.47} & 5.70 & 5.80 & 5.84\\
IS  ($\uparrow$) & 141.72 & 145.85 & \textbf{151.12} & 148.70 & 147.83 & 147.52\\
\bottomrule
\end{tabular}}
\vspace{-5pt}
\caption{\textbf{Ablation study of the sigma ($\sigma$) of Gaussian blur.} The results are from ADM trained on ImageNet 128$\times$128.} \vspace{-10pt}
\label{tab:abl_sigma}
\end{table}

\paragraph{Guidance scale.}
We also evaluate the performance changes as the guidance scale changes with 10k samples on ADM~\cite{dhariwal2021diffusion}. As shown in Fig.~\ref{fig:abl_guide_scale}, we test the scales of $-0.1$, $0.1$, $0.2$, $0.3$, and $0.4$ to ADM and obtain the best FID, sFID, and Inception Score at the guidance scale $s=0.1$. The precision metric shows the best results when the guidance scale is $s=0.3$. We also find out that applying self-attention guidance with a negative scale ($s=-0.1$) or a scale that is too large ($s\ge{}0.4$) harms the sample quality.
\vspace{-10pt}

\paragraph{Gaussian blur.}
We examine the effect of changes on $\sigma$ using 10k samples, testing for $\sigma \in \{1, 3, 9, 27\}$ and the extreme cases. As $\sigma\rightarrow\infty$, the filter gradually blurs the signal content, reducing every pixel to the average value. Conversely, if $\sigma\rightarrow0$, the signal remains unchanged. The results are in Table~\ref{tab:abl_sigma}. SAG is robust against linear changes in $\sigma$, while there still exists an optimal $\sigma$ that yields the best performance. Note that the impact also depends on the input resolution; for instance, a higher input resolution generally requires a larger $\sigma$.

\begin{table}[t]
\footnotesize
\centering
\begin{tabular}{c|c|c|c}
\noalign{\smallskip}\noalign{\smallskip}\toprule
& No guidance & SAG & CFG~\cite{ho2021classifier} \\
\midrule
GPU memory & 12,167MB & 12,209MB & 12,218MB \\
Run-time & 108.27s & 186.60s & 190.27s \\
\bottomrule
\end{tabular}
\vspace{-5pt}
\caption{\textbf{Computational cost.}} \vspace{-15pt}
\label{tab:complexity}
\end{table}

\paragraph{Computational cost.}

We report the computational cost of SAG and CFG~\cite{ho2021classifier} in Table~\ref{tab:complexity}. The memory and time consumption of SAG is almost the same as CFG, which indicates that the overhead due to the operations in SAG (\eg, blurring and masking) is negligible. However, due to the additional step, the cost is high compared to no guidance.

\section{Conclusion}

We present a novel and general formulation of guidance that utilizes internal information within diffusion models for synthesizing high-quality images. Our method, self-attention guidance, is condition- and training-free, and can be applied to various diffusion models, such as ADM, IDDPM, Stable Diffusion, and DiT, improving their quality and reducing the artifacts via self-conditioning. The results of our experiments demonstrate the effectiveness of our proposed method and the orthogonality of self-attention guidance to existing guidance methods. With the findings and the generalization of guidance, we believe that our work opens new avenues for further research in the field of denoising diffusion models and their guidance.

\section*{Acknowledgements}
This research was supported by the MSIT, Korea (IITP-2022-2020-0-01819, ICT Creative Consilience program), and National Research Foundation of Korea (NRF-2021R1C1C1006897). This research was also supported by Samsung Mobile eXperience Business department.

{\small
\bibliographystyle{ieee_fullname}
\bibliography{egbib}
}

\onecolumn

\begin{center}
\Large
\textbf{Appendix}
\end{center}

In this document, we provide additional details of DDPM~\cite{dhariwal2021diffusion}, implementation details of our method, more analyses and results, and the human evaluation protocol. We also discuss the limitations and future work at the end.

\appendix

\section{Denoising Diffusion Probabilistic Models}

DDPM~\cite{ho2020denoising} is a generative model that generates an image from white noise with iterative denoising steps. Given an image $\mathbf{x}_0$ and a variance schedule $\beta_t$ for an arbitrary timestep $t \in \{1, 2, \ldots, T\}$, the forward process of DDPM is defined as a Markov process of the form:
\begin{equation}
q(\mathbf {x}_{t+1}|\mathbf{x}_t) = \mathcal {N}(\mathbf{x}_{t+1};\sqrt{1-\beta_t}\mathbf{x}_t, \beta_t\mathbf{I}).
\end{equation}
Note that we can directly get $\mathbf{x}_t$ from $\mathbf{x}_0$ in the closed form:
\begin{equation}
q(\mathbf{x}_t|\mathbf{x}_0) = \mathcal{N}(\mathbf{x}_t;\sqrt{\bar{\alpha}_t}\mathbf{x}_0, (1 - \bar{\alpha}_t)\mathbf{I}),
\end{equation}
where $\alpha_t = 1 - \beta_t$, and $\bar{\alpha}_t=\prod_{i=1}^{t}\alpha_i$. Similarly, the reverse process is defined as:
\begin{equation}
p_{\theta}(\mathbf {x}_{t-1}|\mathbf{x}_t) = \mathcal {N}(\mathbf{x}_{t-1};\mu_{\theta}(\mathbf{x}_t, t), \Sigma_{\theta}(\mathbf{x}_t, t)\mathbf{I}),
\end{equation}
where $\mu_{\theta}$ and $\Sigma_{\theta}$ denote neural networks with parameter $\theta$.

For the training phase, with $\Sigma_{\theta}$ fixed to a constant $\sigma_t^2=\beta_t$ as in DDPM, $p_\theta(\mathbf{x}_{t-1}|\mathbf{x}_t)$ is compared with the following forward posterior:
\begin{equation}
q(\mathbf{x}_{t-1}|\mathbf{x}_0, \mathbf{x}_t) = \mathcal{N}(\mathbf{x}_{t-1};\tilde{\mu}_{t}(\mathbf{x}_0, \mathbf{x}_t), \tilde{\beta}_t\mathbf{I}),
\end{equation}
where $\tilde{\mu}_t=\frac{\sqrt{\bar{\alpha}_{t-1}}\beta_t}{1-\bar{\alpha}_t}\mathbf{x}_0+\frac{\sqrt{\alpha_t}(1-\bar{\alpha}_{t-1})}{1-\bar{\alpha}_t}\mathbf{x}_t$, and $\tilde{\beta}_t=\frac{1-\bar{\alpha}_{t-1}}{1-\bar{\alpha}_t}\beta_t$. However, instead of directly comparing $\mu_{\theta}$ to $\tilde{\mu}_{t}$, Ho \textit{et al.}~\cite{ho2020denoising} discover that it is beneficial to optimize $\epsilon_\theta$ with the following simplified objective after reparameterization:
\begin{equation}
\mathbf{x}_t = \sqrt{\bar{\alpha}_t}\mathbf{x}_0+\sqrt{1-\bar{\alpha}_t}\boldsymbol{\epsilon},\quad\textrm{where}\quad \boldsymbol{\epsilon} \sim \mathcal{N}(0, \mathbf{I}),
\label{eq:sup-reparam}
\end{equation}
\begin{equation}
L_\mathrm{simple} = \mathbb{E}_{\mathbf{x}_0, t, \boldsymbol{\epsilon}}[||\boldsymbol{\epsilon} - \epsilon_{\theta}(\sqrt{\bar{\alpha}_t}\mathbf{x}_0+\sqrt{1-\bar{\alpha}_t}\boldsymbol{\epsilon}, t)||^2].
\end{equation}

For sampling $\mathbf{x}_{t-1} \sim p_\theta(\mathbf{x}_{t-1}|\mathbf{x}_t)$, we can compute the following from $\mathbf{x}_T$ to $\mathbf{x}_0$:
\begin{equation}
\mathbf{x}_{t-1} = \frac{1}{\sqrt{\bar{\alpha}_t}}(\mathbf{x}_t-\frac{\beta_t}{\sqrt{1-\bar{\alpha}_t}}\epsilon_{\theta}(\mathbf{x}_t, t))+\sigma_t\mathbf{z},
\end{equation}
where $\mathbf{z} \sim \mathcal{N}(0, \mathbf{I})$. Rewriting Eq.~\ref{eq:sup-reparam}, we can get $\hat{\mathbf{x}}_0$ which is a prediction of $\mathbf{x}_0$ at each timestep with the following formula:
\begin{equation}
\hat{\mathbf{x}}_0 = (\mathbf{x}_t-\sqrt{1-\bar{\alpha}_t}\epsilon_{\theta}(\mathbf{x}_t, t))/\sqrt{\bar{\alpha}_t}.
\end{equation}

\newpage

\section{Additional Implementation Details}

\subsection{Environmental setting}
For the experiments, we use two servers of 8 NVIDIA GeForce RTX 3090 GPUs each to sample from the pre-trained models of ADM~\cite{dhariwal2021diffusion}, IDDPM~\cite{nichol2021improved}, Stable Diffusion v1.4~\cite{rombach2022high}, and DiT~\cite{peebles2022scalable}. We build upon the PyTorch~\cite{paszke2019pytorch} implementation of these models, taking all the weights for our experiments from their publicly available repository.

\subsection{Selective blurring}
In practice, we efficiently implement selective blurring in Sec.~{5.2}. At the first step, we blur the intermediate reconstruction $\hat{\mathbf{x}}_0$ of $\mathbf{x}_t$~\cite{ho2020denoising}. Then, we apply masks $1-M_t$ and $M_t$ on $\hat{\mathbf{x}}_0$ and the blurred version of $\hat{\mathbf{x}}_0$, respectively. Finally, we aggregate the output and then noise it again with the predicted noise $\epsilon_\theta(\mathbf{x}_t)$ that we use for computing $\hat{\mathbf{x}}_0$ above. This process ends up producing the same $\widehat{\mathbf{x}}_t$ as Eq.~{15} in the main paper.

\subsection{Combination of SAG and CFG}
Na\"ively, in order to combine SAG with CFG~\cite{ho2021classifier} in Stable Diffusion~\cite{rombach2022high} and DiT~\cite{peebles2022scalable}, we have to compute SAG through the conditional and unconditional models, which requires us four feedforward steps. In practice, the guided prediction of noise can be efficiently calculated as follows:
\begin{equation}
\label{eq:double-guidance}
\tilde\epsilon(\mathbf{x}_t) = \epsilon_\theta(\mathbf{x}_t, c) + s_{\textrm{c}} (\epsilon_\theta(\mathbf{x}_t, c) - \epsilon_\theta(\mathbf{x}_t)) + s_{\textrm{s}} (\epsilon_\theta(\mathbf{x}_t) - \epsilon_\theta(\bar{\mathbf{x}}_t)),
\end{equation}
where $s_{\textrm{c}}$ and $s_{\textrm{s}}$ denote the scales of CFG and SAG, respectively, and $c$ denotes a text prompt.

\subsection{Hyperparameter settings}
In Table~\ref{tab:hp}, we report our hyperparameter settings for our experiments. In the ablation studies in the main paper, we set the other parameters to the constants in Table~\ref{tab:hp}, while testing the ablated parameter. Note that $\sigma$ is dependent on the input resolution.

\begin{table*}[h]
\small
\centering
  \begin{tabular}{cc|ccc|c}
  \toprule
    \multicolumn{2}{c}{\multirow{3}{*}{Model}} &
      \multicolumn{3}{c|}{Self-attention} &
      \multicolumn{1}{c}{Gaussian-blur}\\
    &
      &\multicolumn{3}{c|}{parameter} &
      \multicolumn{1}{c}{parameter}\\
    & & Guidance scale & Threshold & Layer & $\sigma$ \\
      \midrule
    \multirow{10}{*}{ADM~\cite{dhariwal2021diffusion}} &ImageNet 256$\times$256 & \multirow{2}{*}{0.5, 0.8} & \multirow{2}{*}{1.0} & \multirow{2}{*}{Output 2} & \multirow{2}{*}{9} \\
    &(unconditional) & & & & \\
    \cmidrule{2-6}
    &ImageNet 256$\times$256 & \multirow{2}{*}{0.2} & \multirow{2}{*}{1.0} & \multirow{2}{*}{Output 2} & \multirow{2}{*}{9} \\
    &(conditional) & & & &\\
    \cmidrule{2-6}
    &LSUN Cat 256$\times$256 & 0.05 & 1.0 & Output 2 & 9\\
    \cmidrule{2-6}
    &LSUN Horse 256$\times$256 & 0.01 & 1.0 & Output 2 & 9\\
    \cmidrule{2-6}
    &ImageNet 128$\times$128 & 0.1 & 1.0 & Output 8 & 3\\
    \midrule
     \multirow{2}{*}{IDDPM~\cite{nichol2021improved}}&ImageNet 64$\times$64 & \multirow{2}{*}{0.05} & \multirow{2}{*}{1.0} & \multirow{2}{*}{Output 7} & \multirow{2}{*}{1} \\
    &(unconditional) & & & &\\
    \midrule
     \multicolumn{2}{c|}{Stable Diffusion~\cite{rombach2022high}} & 0.75, 1.0 & 1.0 & Middle & 1 \\
    \midrule
     \multicolumn{2}{c|}{DiT~\cite{peebles2022scalable}} & 0.005 & 1.0 & 13th block & 1 \\
    \bottomrule
  \end{tabular}
\caption{Hyperparameter settings.}
\label{tab:hp}
\end{table*}

\begin{figure*}[h]
\centering
\small
\begin{tabular}{lc}
Synthesized &
\begin{minipage}[c]{1.0\textwidth}
\begin{subfigure}{.10\textwidth}
  \centering
  \includegraphics[width=1.0\linewidth]{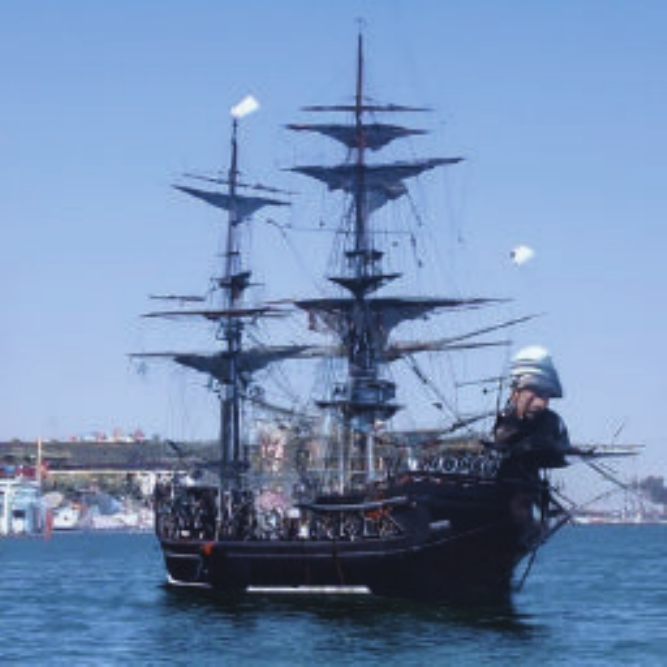}
\end{subfigure}
\begin{subfigure}{.10\textwidth}
  \centering
  \includegraphics[width=1.0\linewidth]{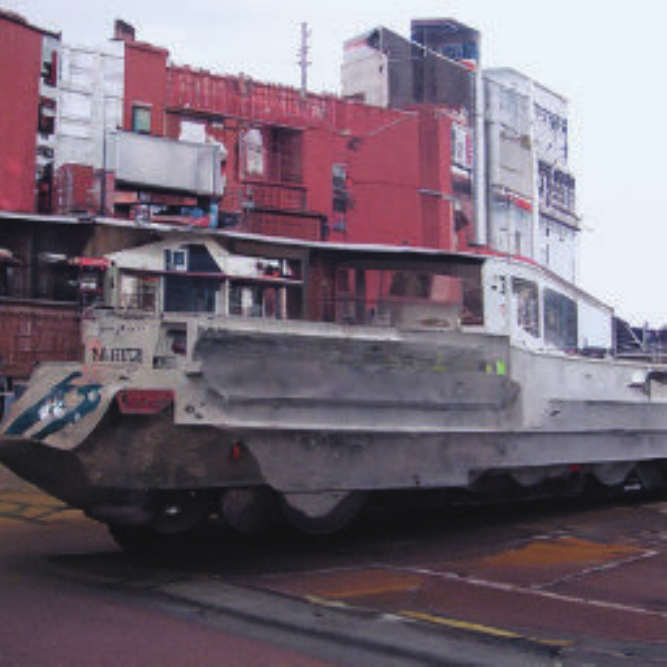}
\end{subfigure}
\begin{subfigure}{.10\textwidth}
  \centering
  \includegraphics[width=1.0\linewidth]{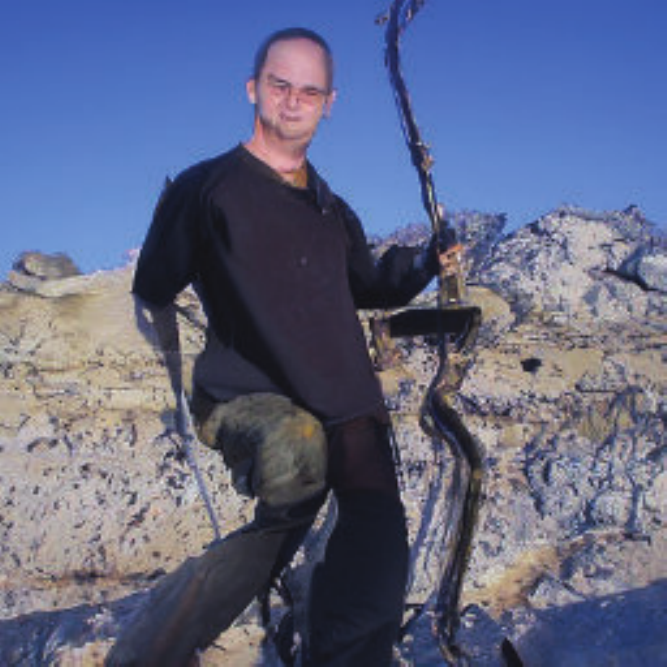}
\end{subfigure}
\begin{subfigure}{.10\textwidth}
  \centering
  \includegraphics[width=1.0\linewidth]{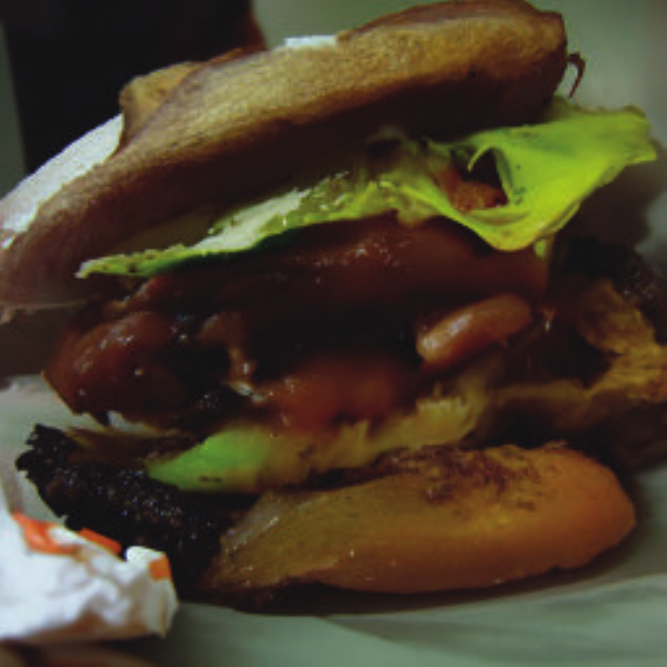}
\end{subfigure}
\begin{subfigure}{.10\textwidth}
  \centering
  \includegraphics[width=1.0\linewidth]{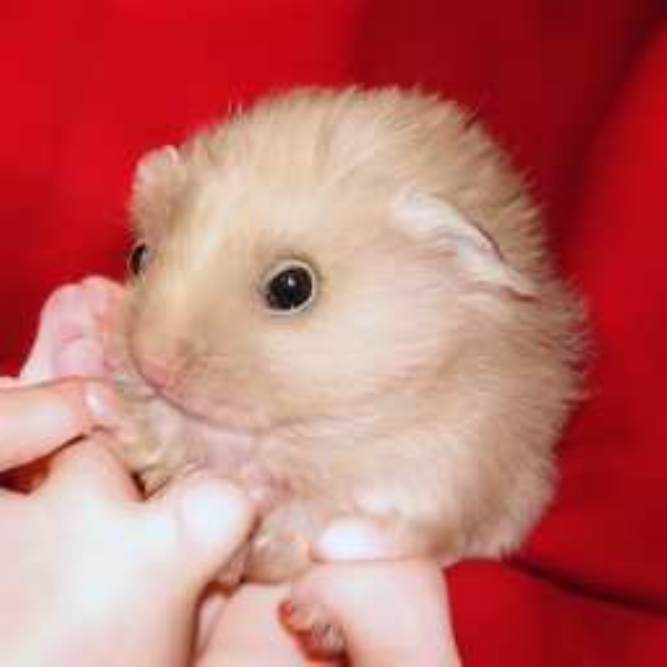}
\end{subfigure}
\begin{subfigure}{.10\textwidth}
  \centering
  \includegraphics[width=1.0\linewidth]{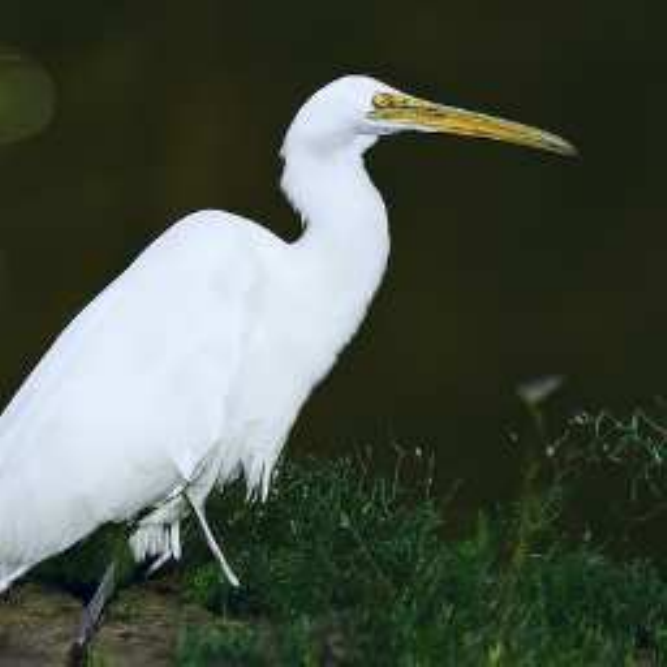}
\end{subfigure}
\begin{subfigure}{.10\textwidth}
  \centering
  \includegraphics[width=1.0\linewidth]{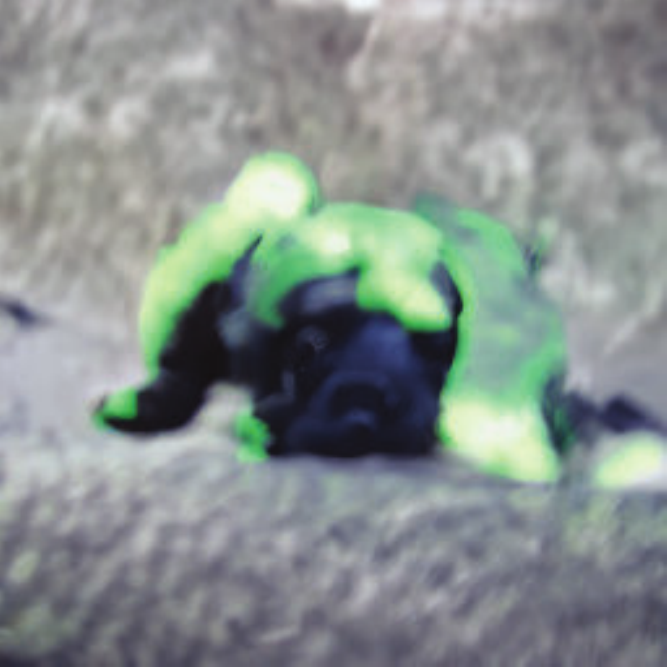}
\end{subfigure}
\begin{subfigure}{.10\textwidth}
  \centering
  \includegraphics[width=1.0\linewidth]{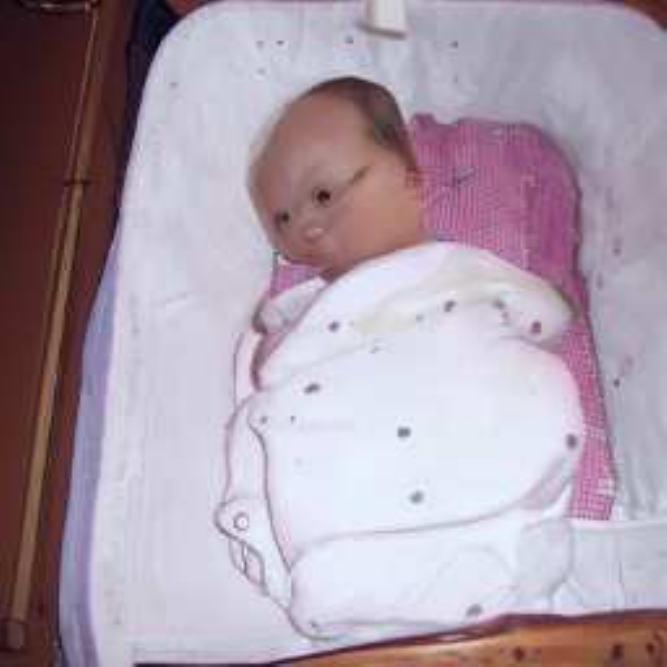}
\end{subfigure} 
\end{minipage}\\
(a) DINO &
\begin{minipage}[c]{1.0\textwidth}
\begin{subfigure}{.10\textwidth}
  \centering
  \includegraphics[width=1.0\linewidth]{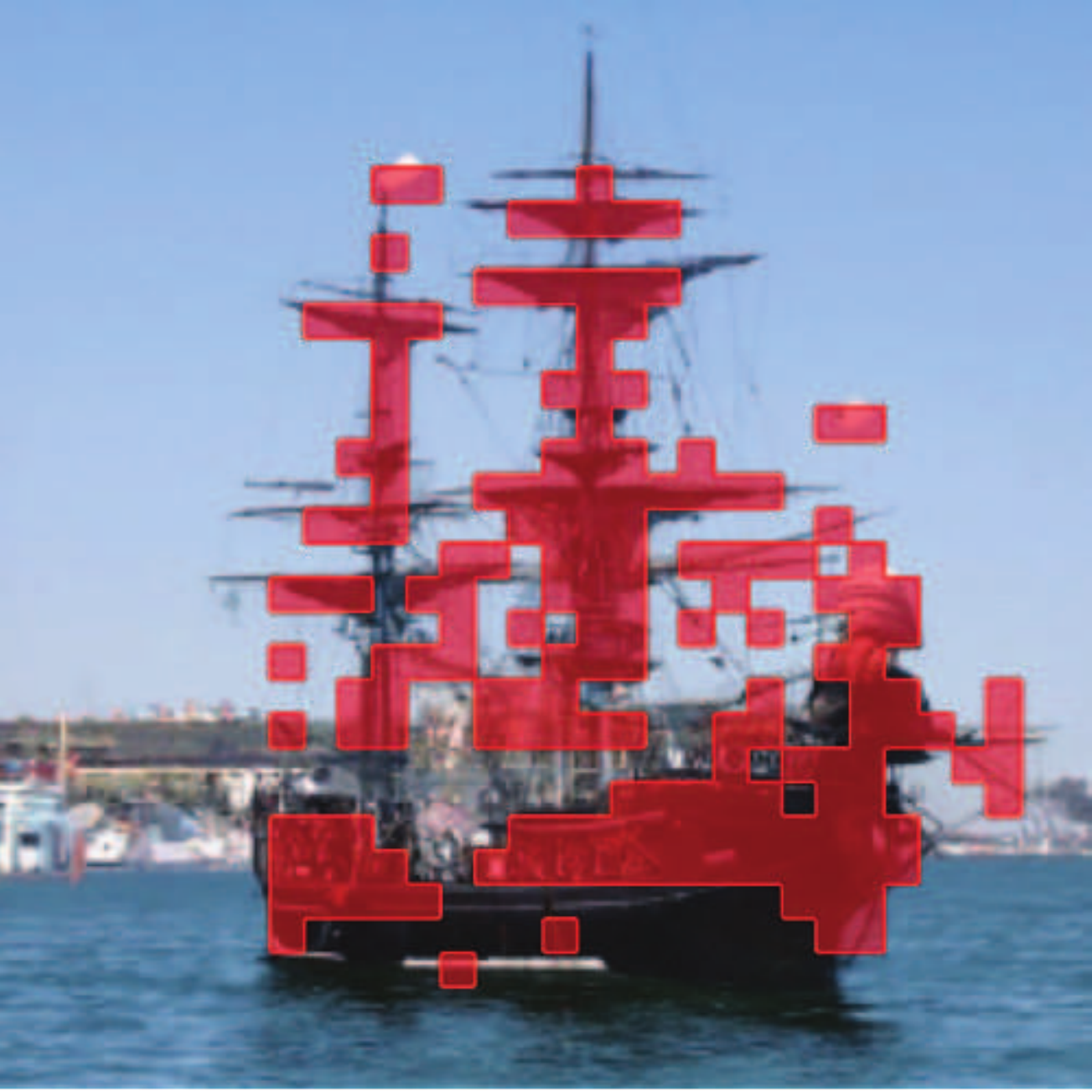}
\end{subfigure}
\begin{subfigure}{.10\textwidth}
  \centering
  \includegraphics[width=1.0\linewidth]{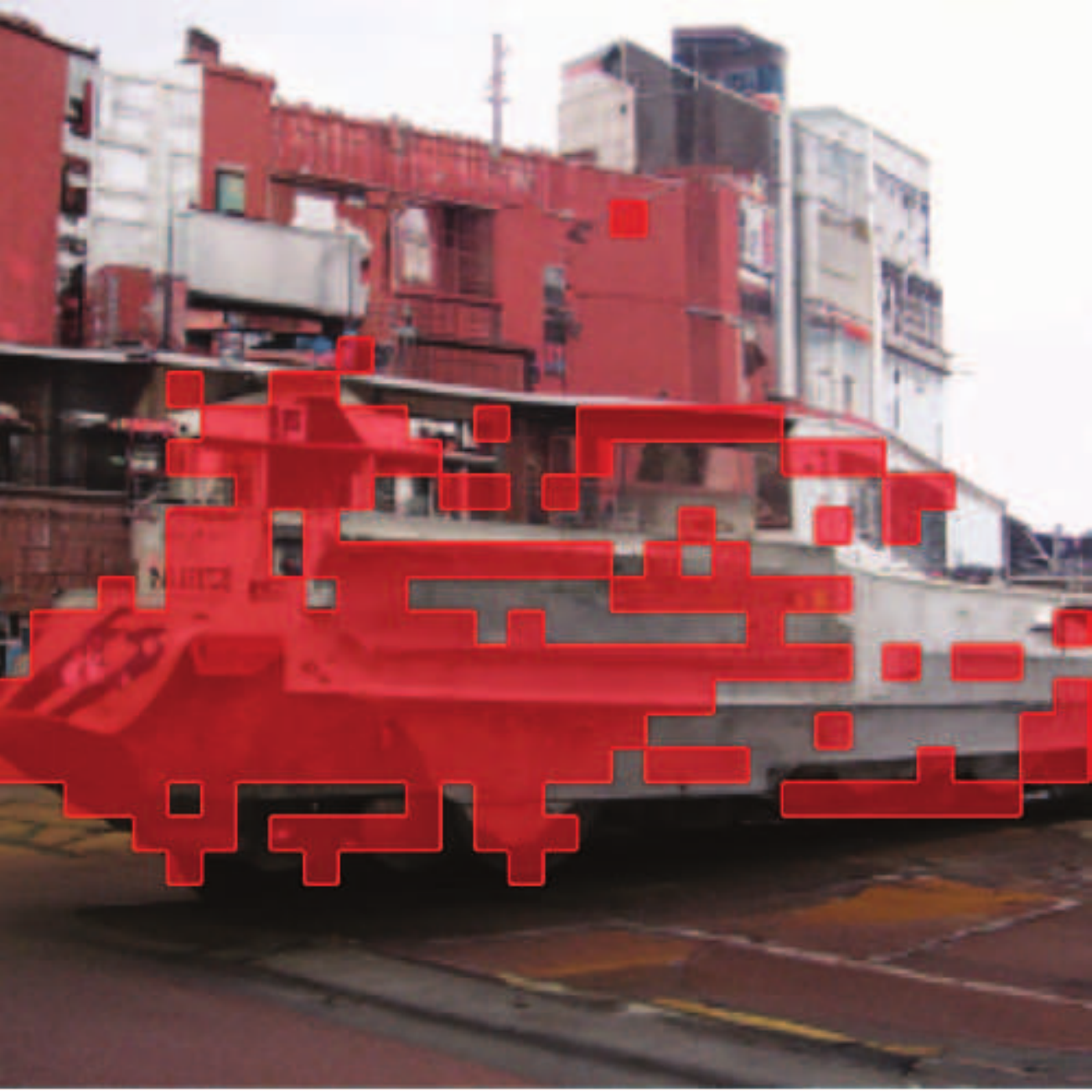}
\end{subfigure}
\begin{subfigure}{.10\textwidth}
  \centering
  \includegraphics[width=1.0\linewidth]{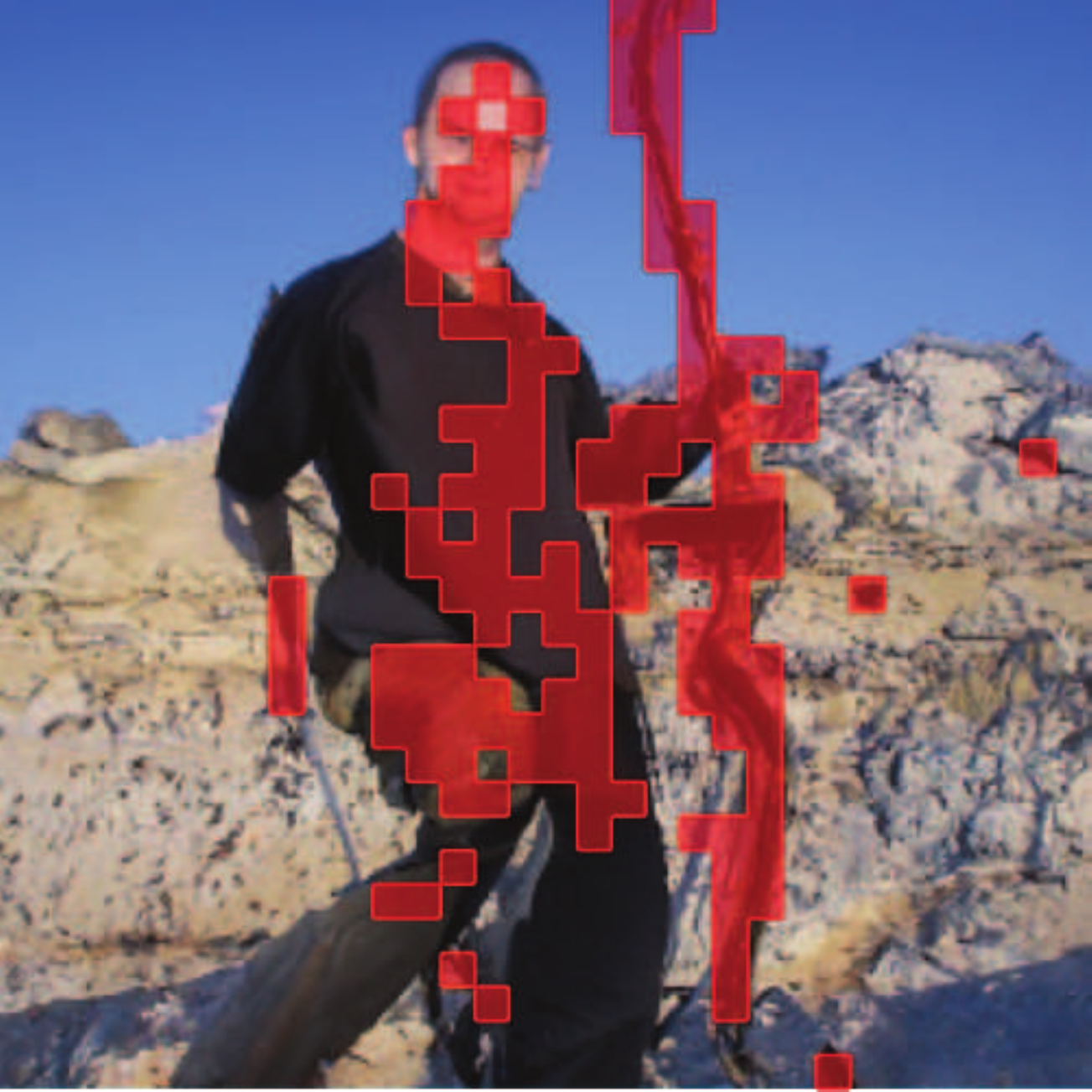}
\end{subfigure}
\begin{subfigure}{.10\textwidth}
  \centering
  \includegraphics[width=1.0\linewidth]{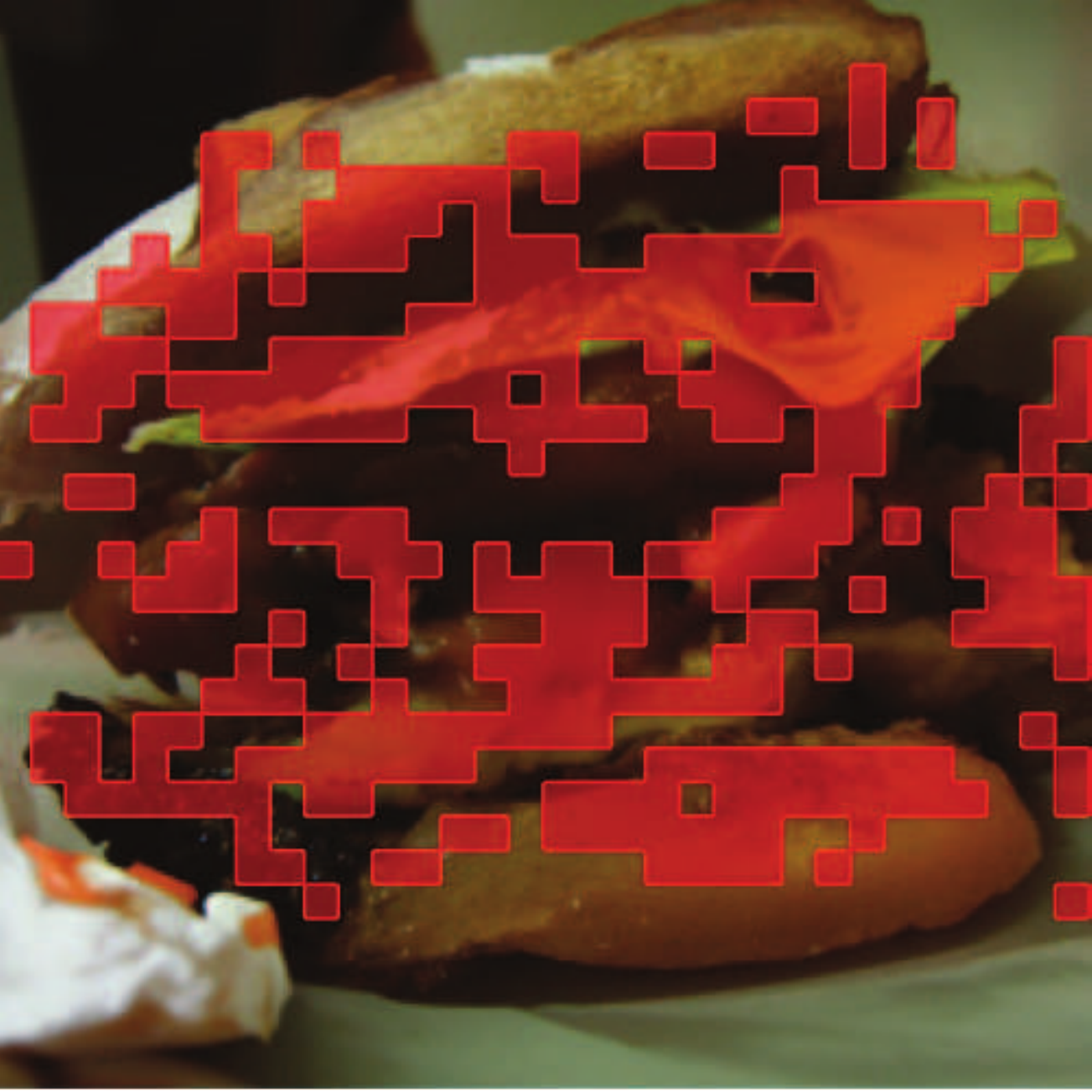}
\end{subfigure}
\begin{subfigure}{.10\textwidth}
  \centering
  \includegraphics[width=1.0\linewidth]{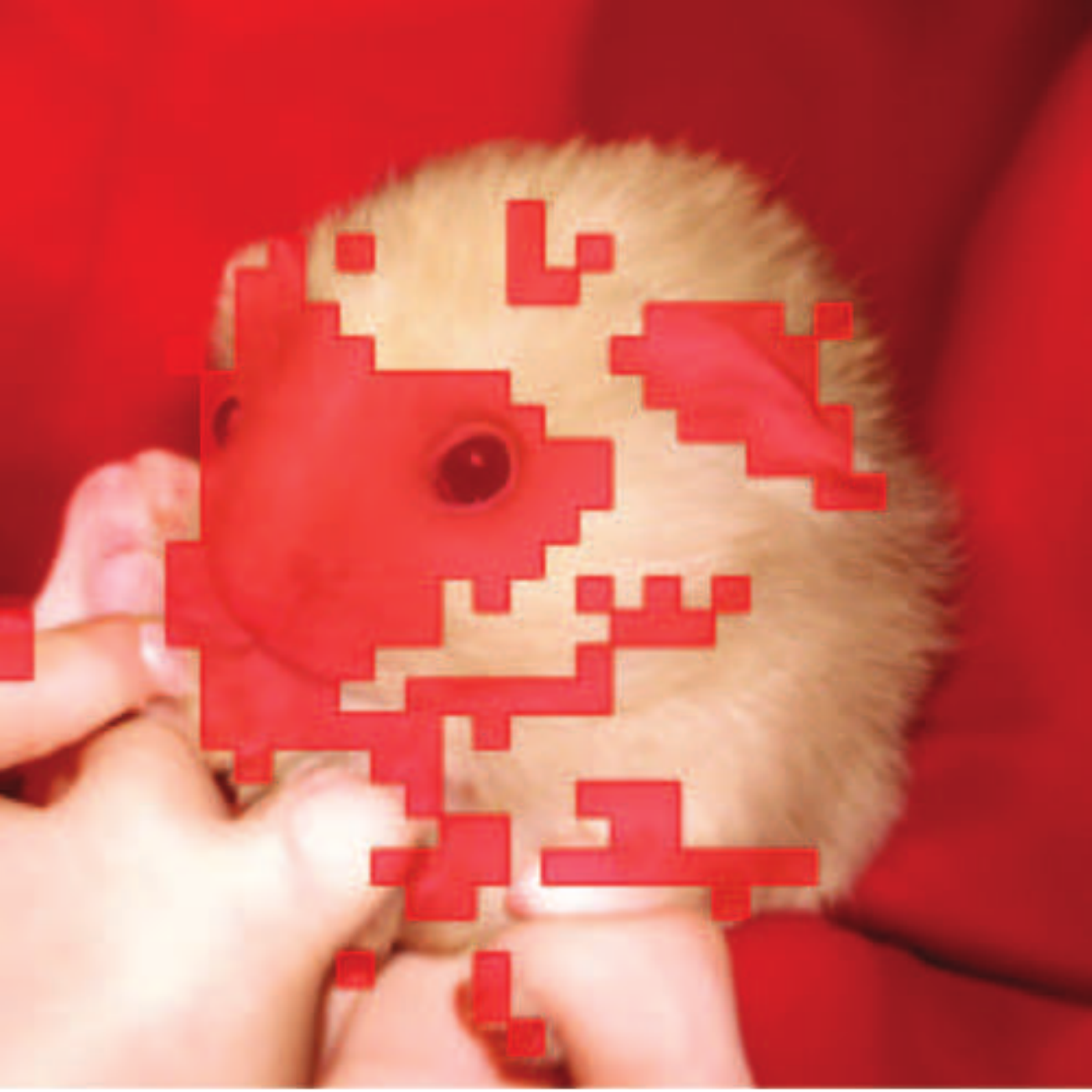}
\end{subfigure}
\begin{subfigure}{.10\textwidth}
  \centering
  \includegraphics[width=1.0\linewidth]{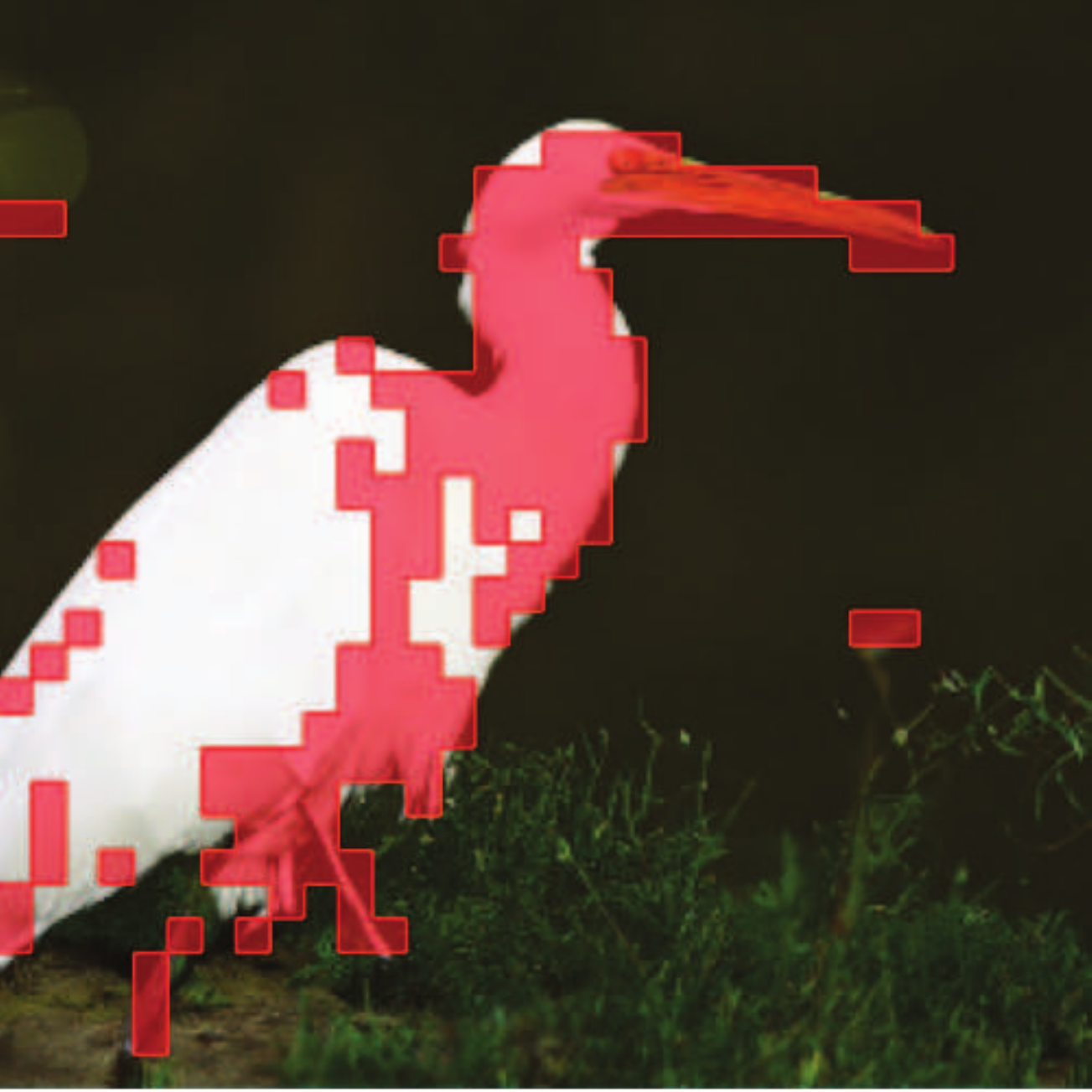}
\end{subfigure}
\begin{subfigure}{.10\textwidth}
  \centering
  \includegraphics[width=1.0\linewidth]{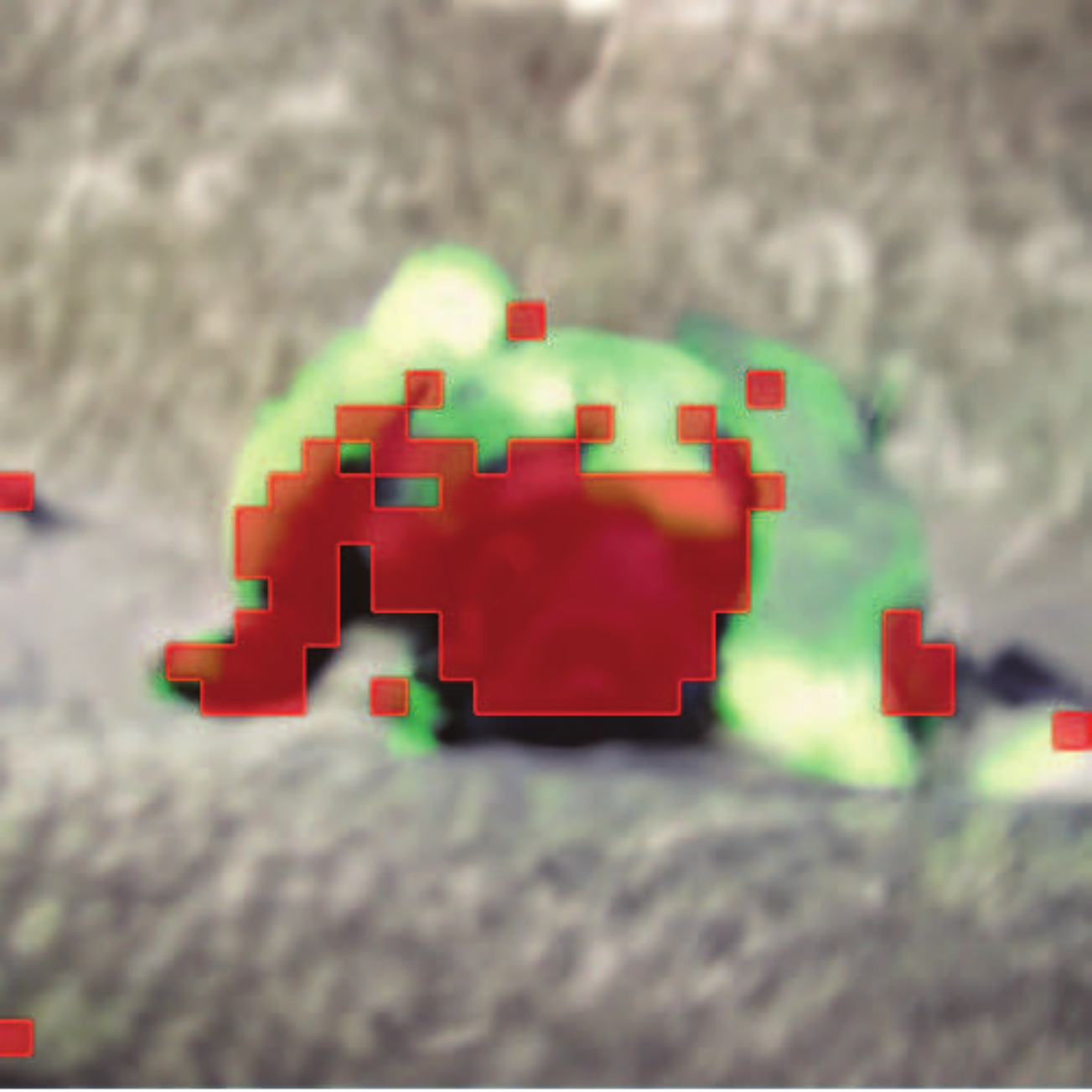}
\end{subfigure}
\begin{subfigure}{.10\textwidth}
  \centering
  \includegraphics[width=1.0\linewidth]{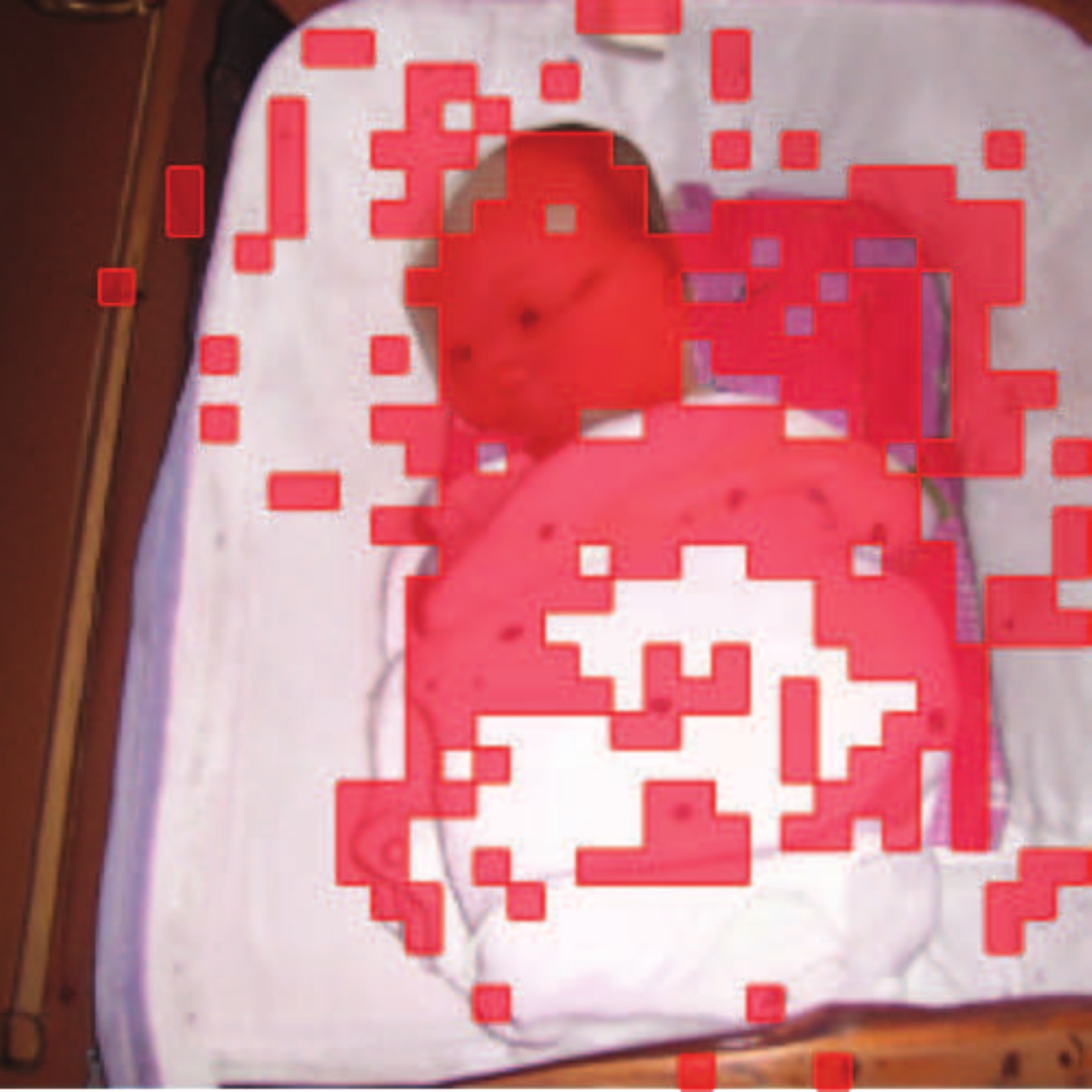}
\end{subfigure}
\end{minipage}\\
(b) ADM &
\begin{minipage}[c]{1.0\textwidth}
\begin{subfigure}{.10\textwidth}
  \centering
  \includegraphics[width=1.0\linewidth]{Figure/5_sa_cmyk.pdf}
\end{subfigure}
\begin{subfigure}{.10\textwidth}
  \centering
  \includegraphics[width=1.0\linewidth]{Figure/0_sa_cmyk.pdf}
\end{subfigure}
\begin{subfigure}{.10\textwidth}
  \centering
  \includegraphics[width=1.0\linewidth]{Figure/2_sa_cmyk.pdf}
\end{subfigure}
\begin{subfigure}{.10\textwidth}
  \centering
  \includegraphics[width=1.0\linewidth]{Figure/3_sa_cmyk.pdf}
\end{subfigure}
\begin{subfigure}{.10\textwidth}
  \centering
  \includegraphics[width=1.0\linewidth]{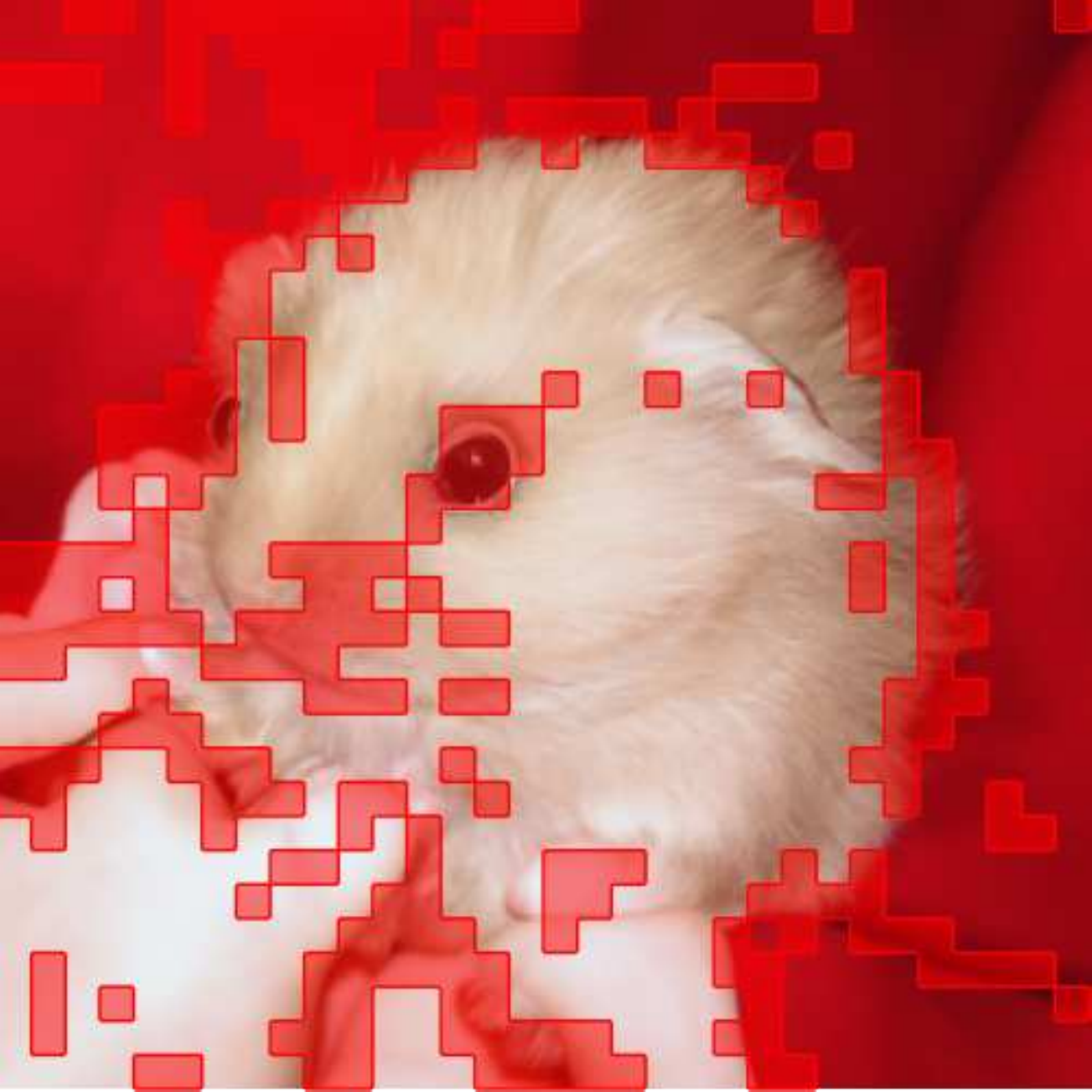}
\end{subfigure}
\begin{subfigure}{.10\textwidth}
  \centering
  \includegraphics[width=1.0\linewidth]{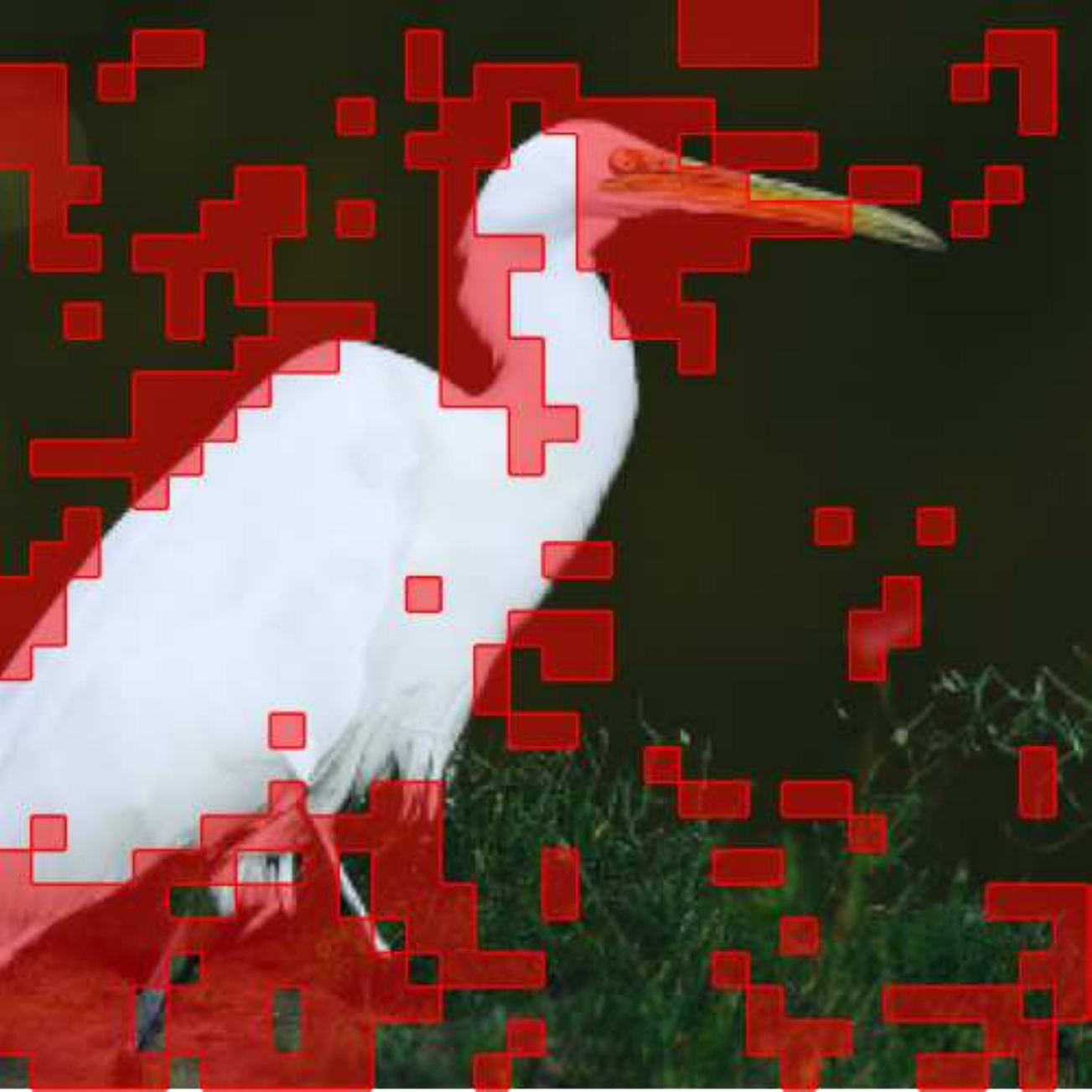}
\end{subfigure}
\begin{subfigure}{.10\textwidth}
  \centering
  \includegraphics[width=1.0\linewidth]{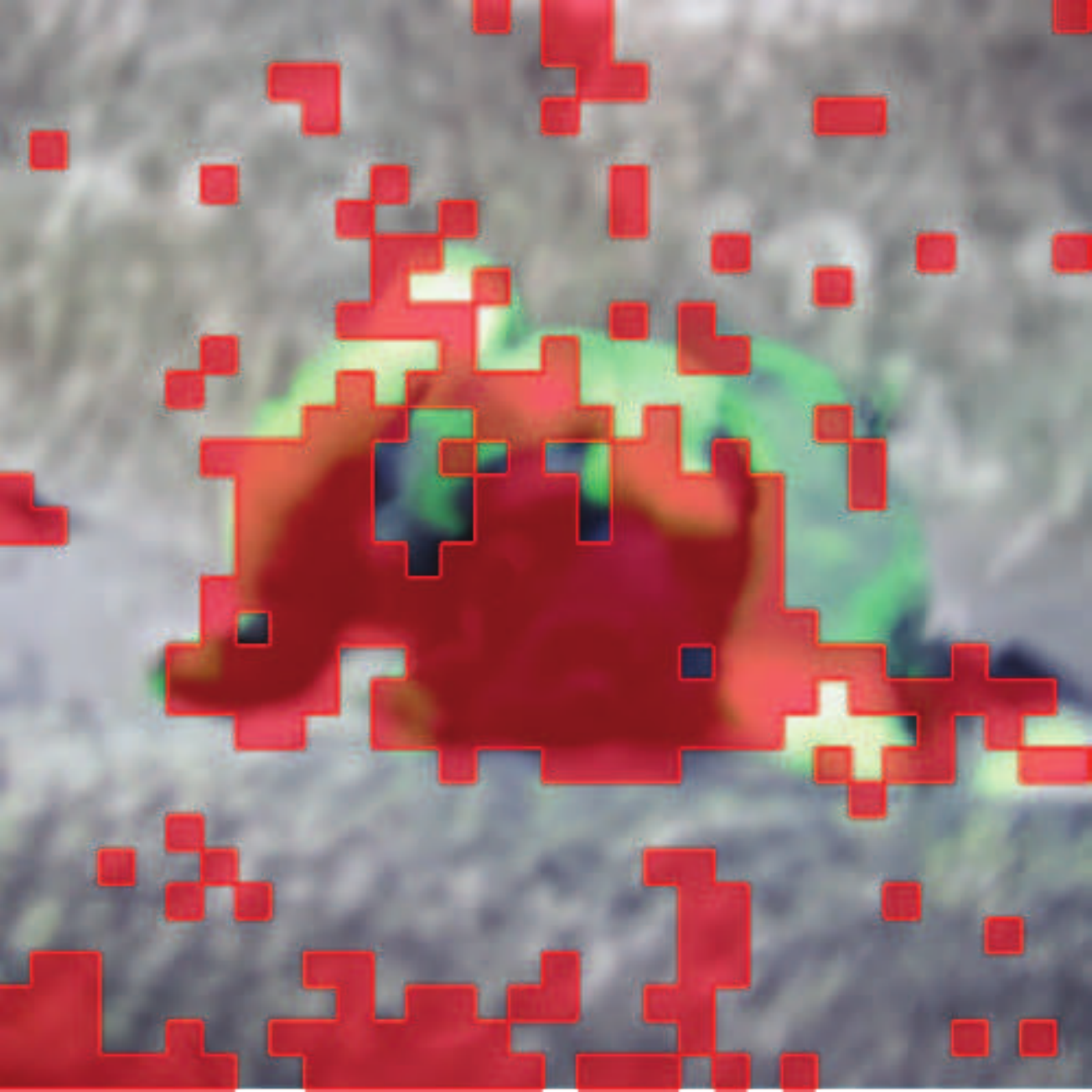}
\end{subfigure}
\begin{subfigure}{.10\textwidth}
  \centering
  \includegraphics[width=1.0\linewidth]{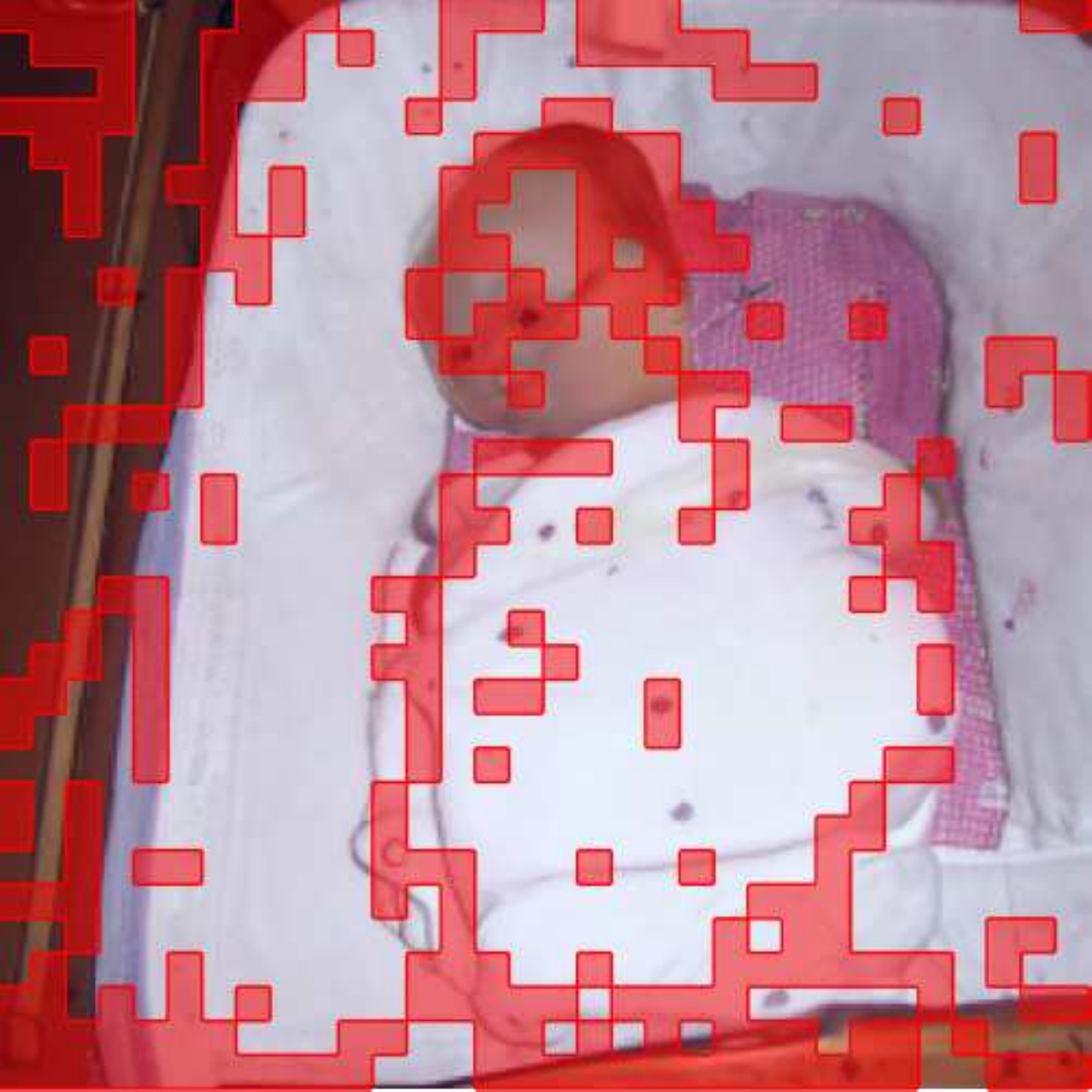}
\end{subfigure}
\end{minipage}\\
\end{tabular}
\caption{\textbf{Comparison between self-attention masks of DINO~\cite{caron2021emerging} and ADM~\cite{dhariwal2021diffusion}:} (a) the self-attention masks extracted from DINO~\cite{caron2021emerging}, (b) the self-attention masks extracted from ADM~\cite{dhariwal2021diffusion}.}
\label{fig:dino-vis}
\end{figure*}

\begin{figure}[t]
\begin{subfigure}{.255\textwidth}
  \centering  \captionsetup{justification=centering}
  \includegraphics[width=1.0\linewidth]{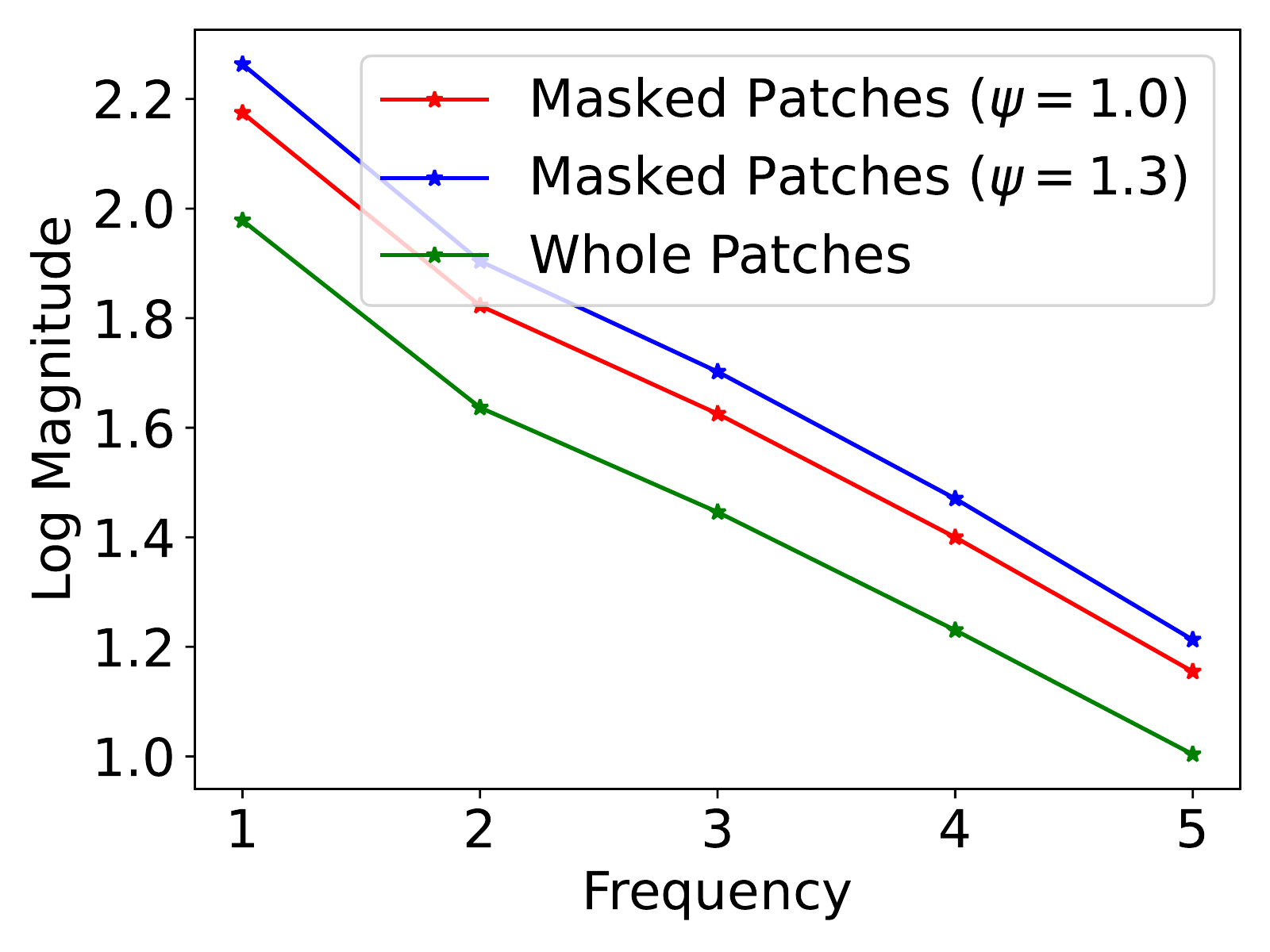}
  \caption{Frequency-magnitude plot of\\8$\times$8 patches}
\end{subfigure}\hfill
\begin{subfigure}{.255\textwidth}
  \centering \captionsetup{justification=centering}
  \includegraphics[width=1.0\linewidth]{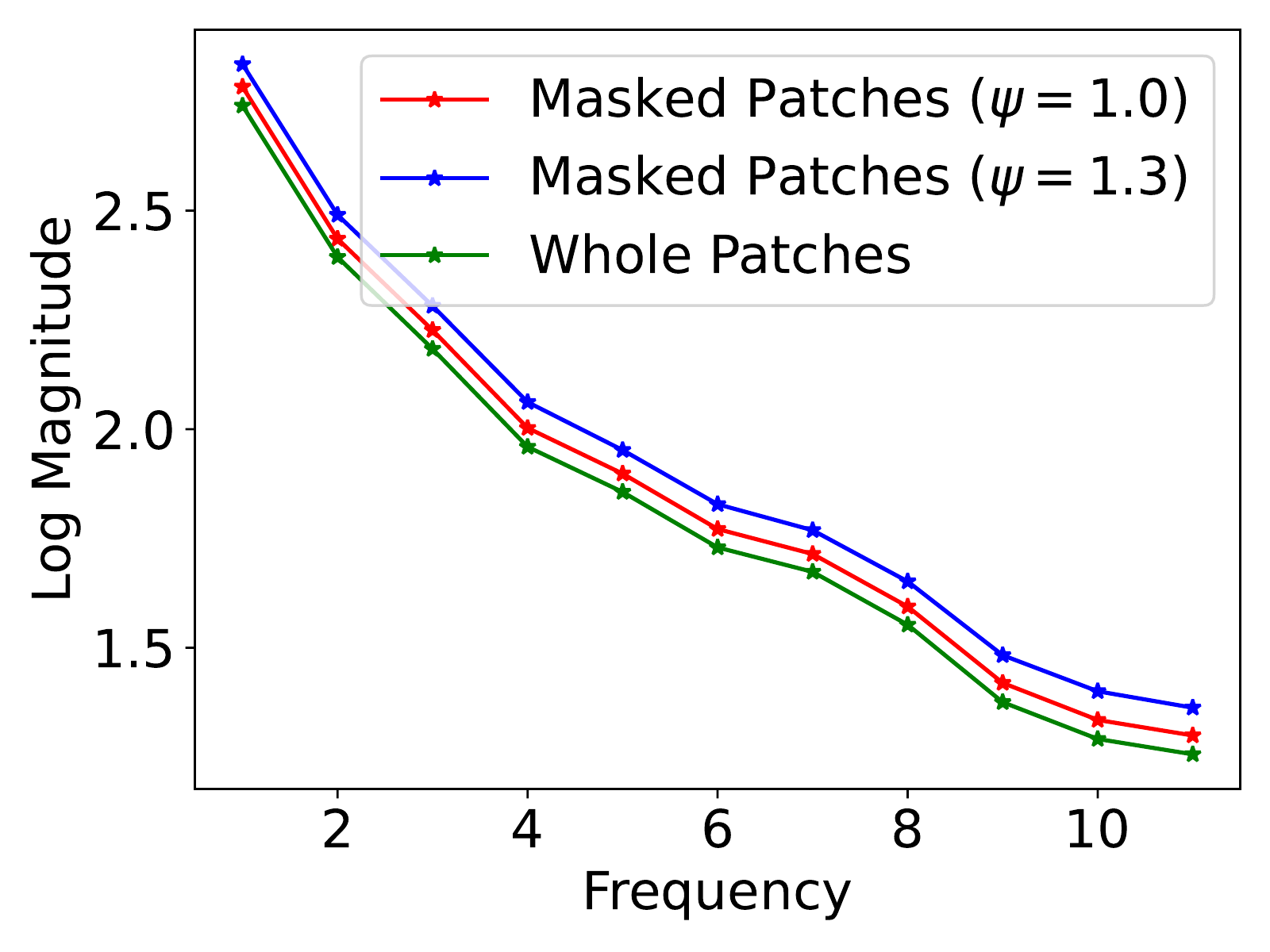}
  \caption{Frequency-magnitude plot of\\16$\times$16 patches}
\end{subfigure}\hfill
\begin{subfigure}{.255\textwidth}
  \centering \captionsetup{justification=centering}
  \includegraphics[width=1.0\linewidth]{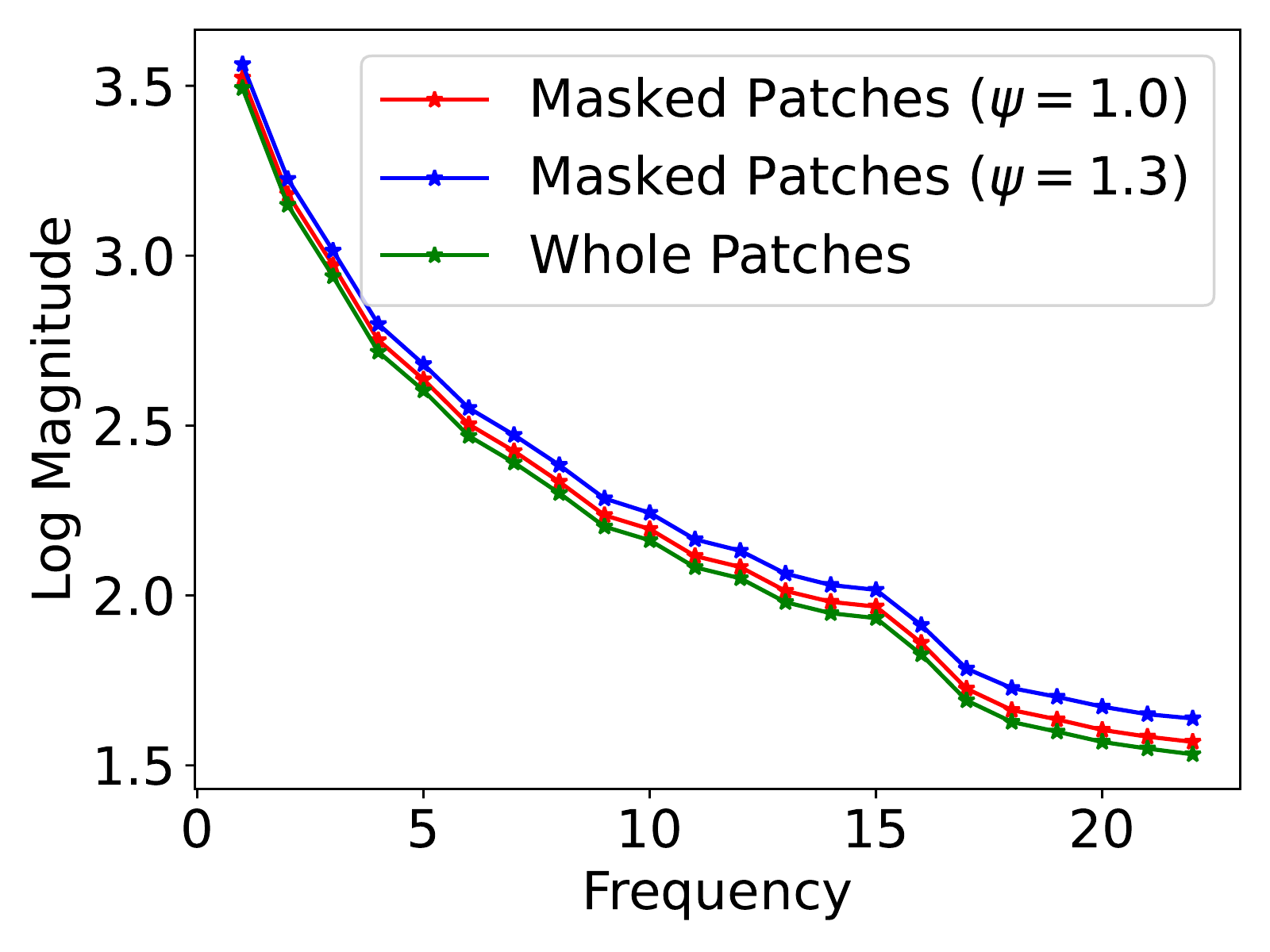}
  \caption{Frequency-magnitude plot of\\32$\times$32 patches}
\end{subfigure}
\begin{subfigure}{.21\textwidth}
  \centering \captionsetup{justification=centering}
    \includegraphics[width=1.0\linewidth]{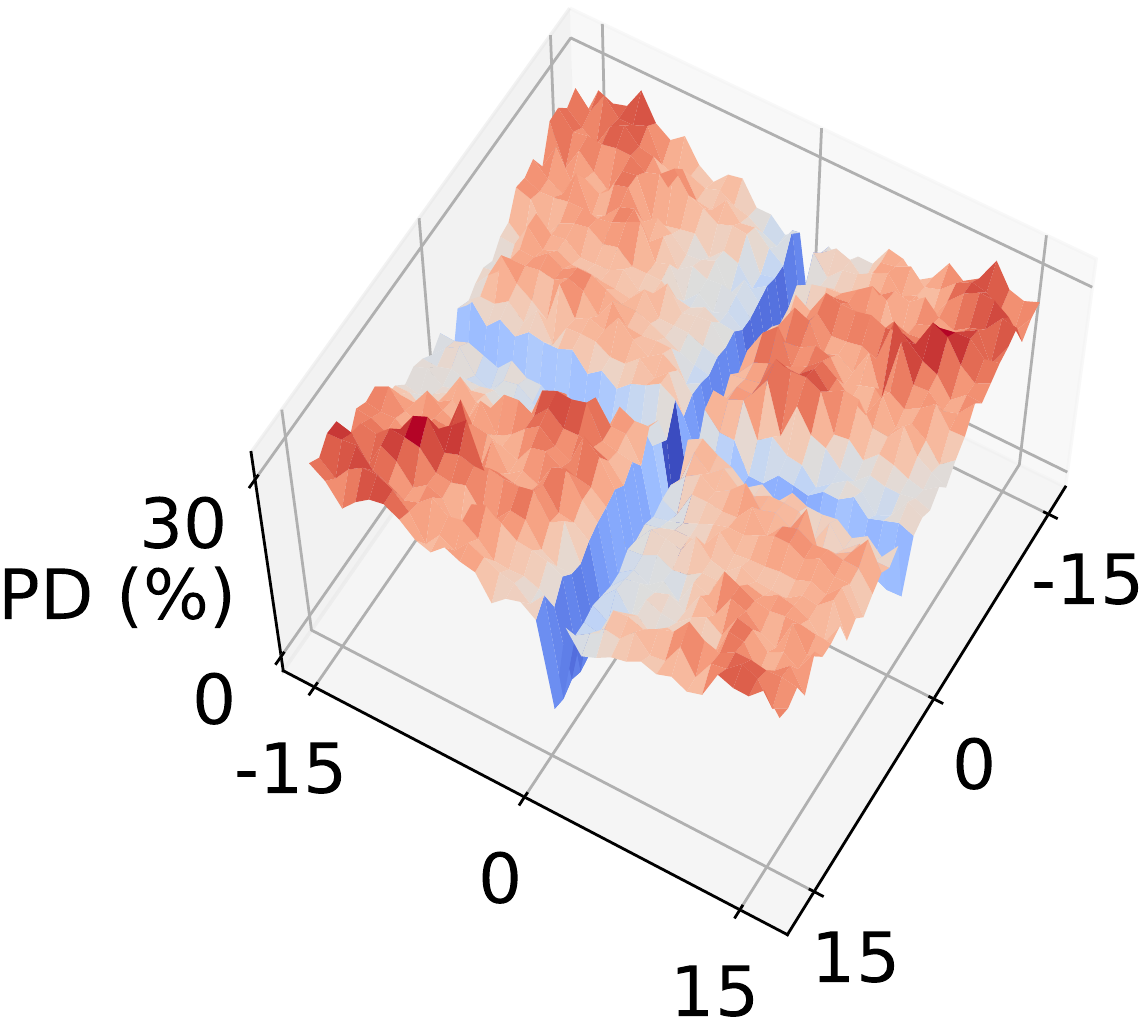}
  \caption{Percentage difference (PD) \\in frequency domain}
\end{subfigure}
\caption{\textbf{Frequency analysis of the self-attention masks:} (a), (b) and (c) show the frequency-magnitude graphs of 8$\times$8, 16$\times$16, and 32$\times$32 patches, respectively. $\psi$ denotes the masking threshold. (d) is a 3D visualization that shows the percentage difference of magnitude between masked and non-masked patches in the frequency domain regarding the 32$\times$32 patches.}
\label{fig:freq_vis}
\end{figure}

\newpage

\section{Additional Analyses and Results}

\subsection{Exploring the self-attention in diffusion models}
\begin{figure*}[p]
\centering
\begin{tabular}{lc}
\begin{subfigure}{.05\textwidth}
  \centering
  \includegraphics[width=1.0\linewidth]{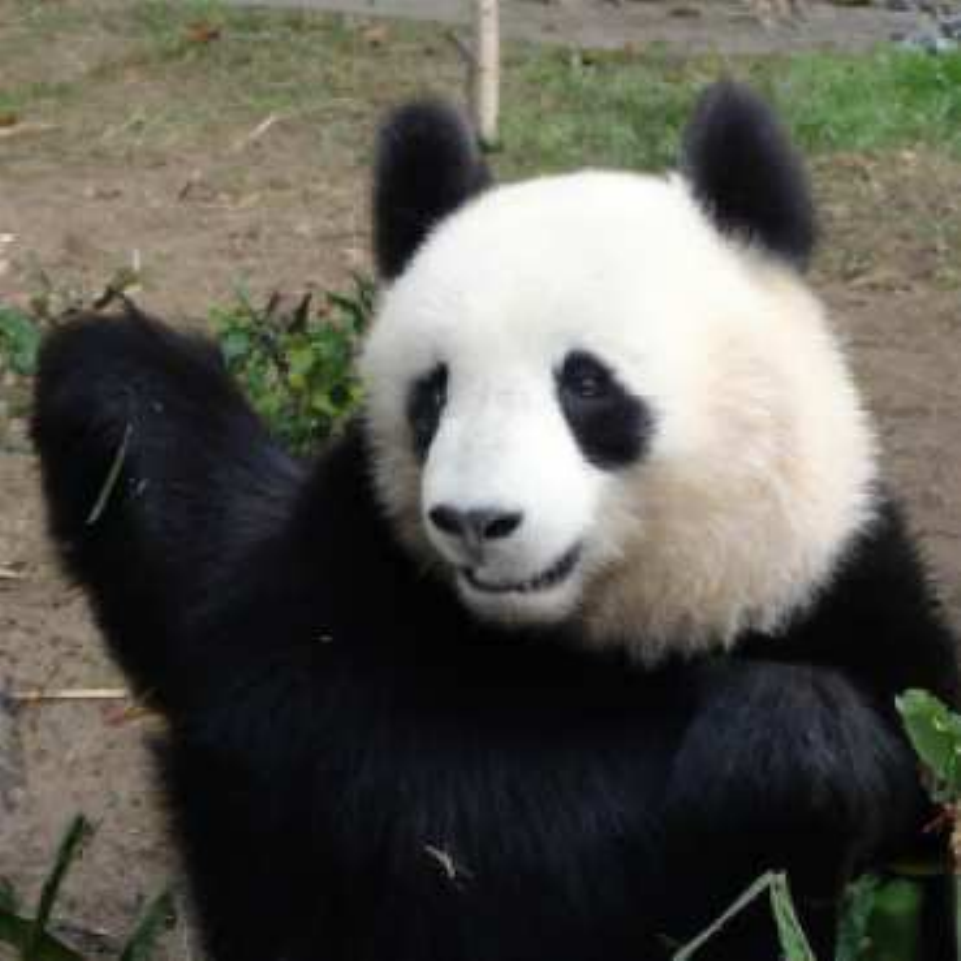}
\end{subfigure}
&
\begin{subfigure}{.05\textwidth}
  \centering
  \includegraphics[width=1.0\linewidth]{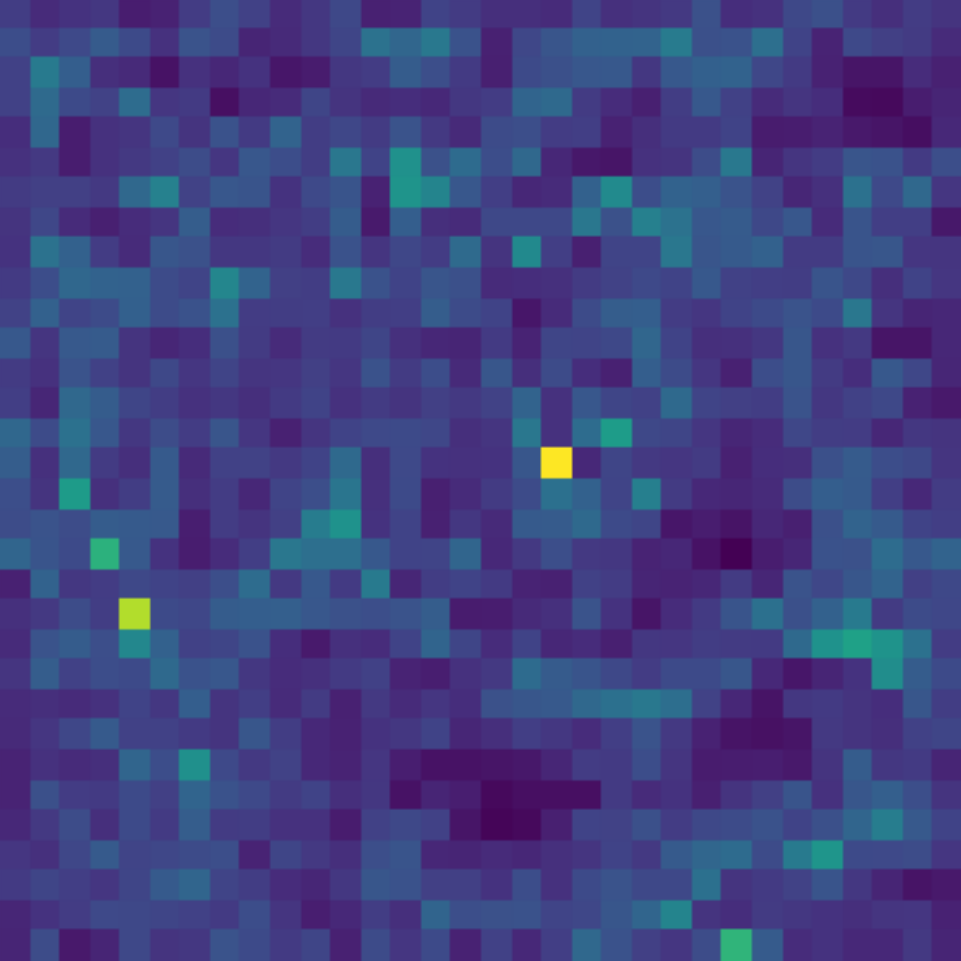}
\end{subfigure}
\begin{subfigure}{.05\textwidth}
  \centering
  \includegraphics[width=1.0\linewidth]{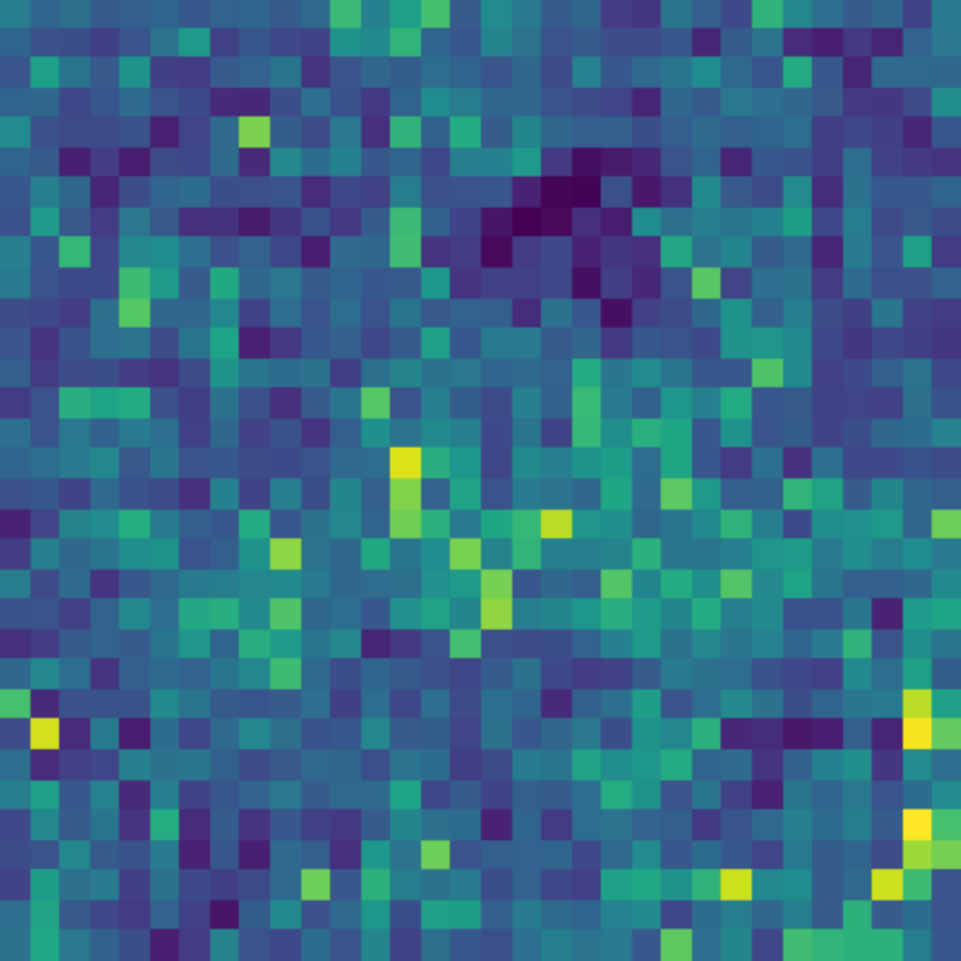}
\end{subfigure}
\begin{subfigure}{.05\textwidth}
  \centering
  \includegraphics[width=1.0\linewidth]{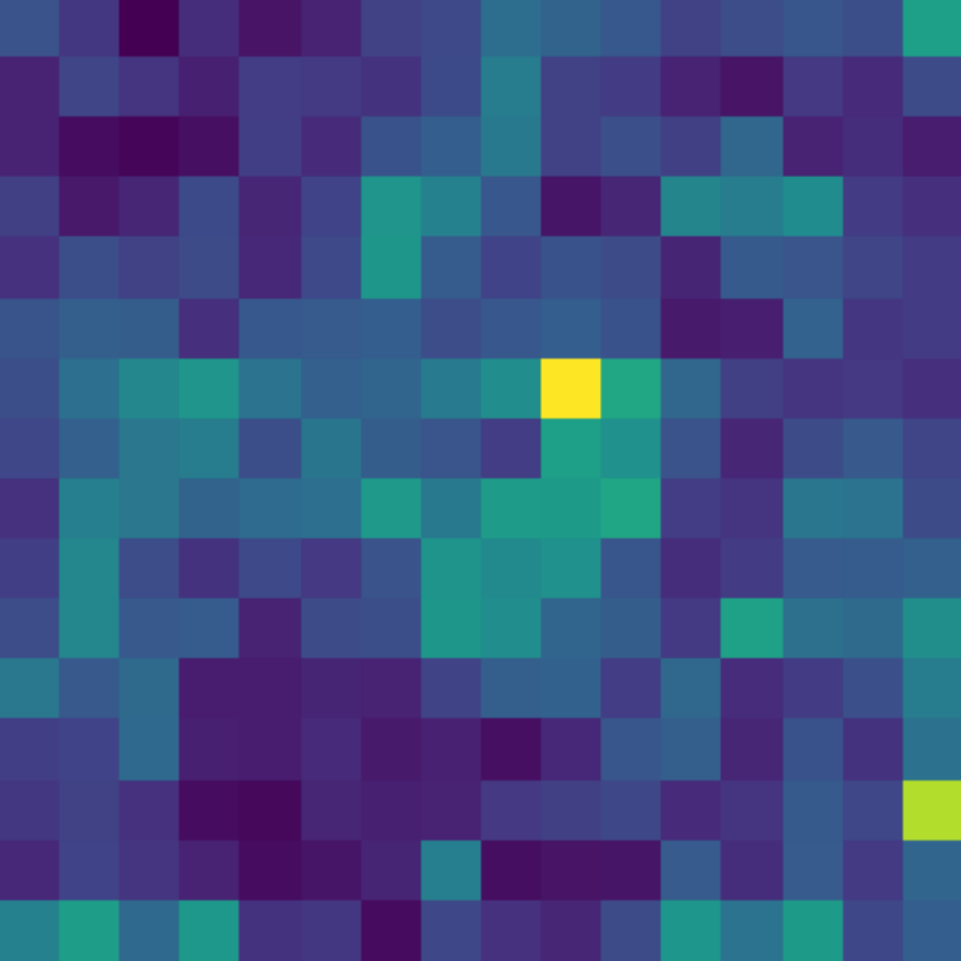}
\end{subfigure}
\begin{subfigure}{.05\textwidth}
  \centering
  \includegraphics[width=1.0\linewidth]{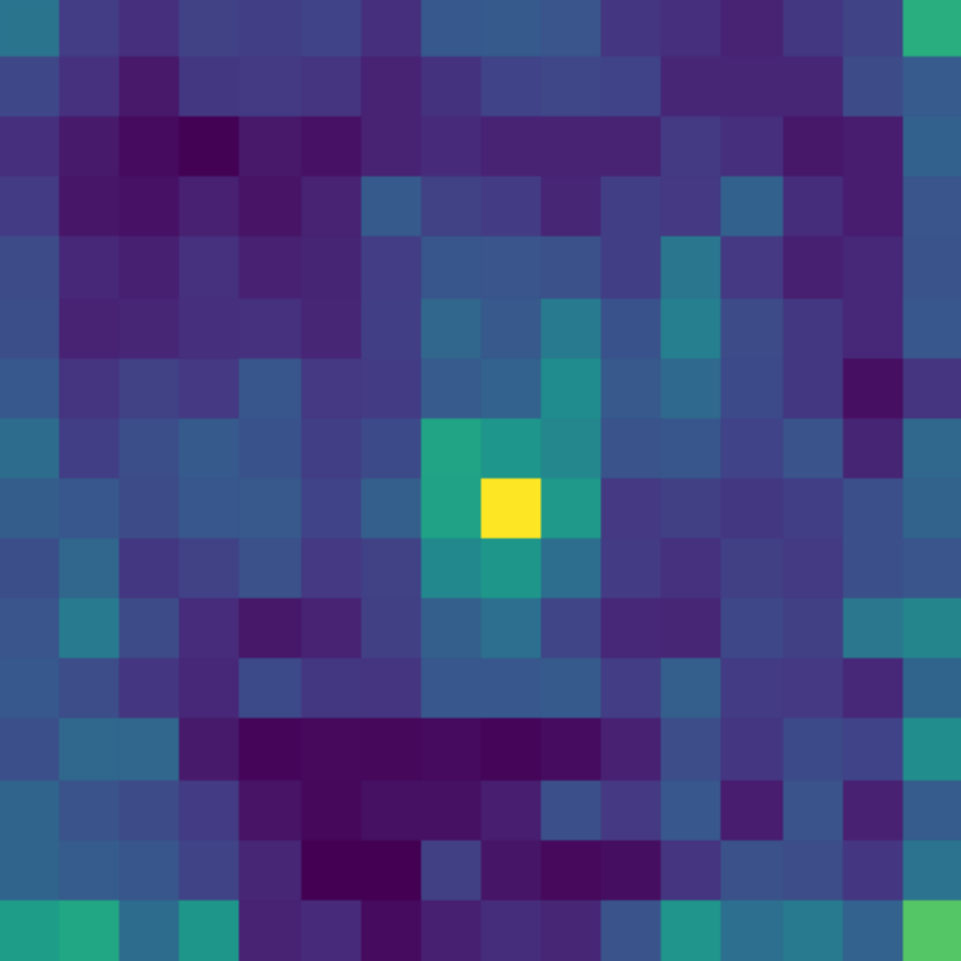}
\end{subfigure}
\begin{subfigure}{.05\textwidth}
  \centering
  \includegraphics[width=1.0\linewidth]{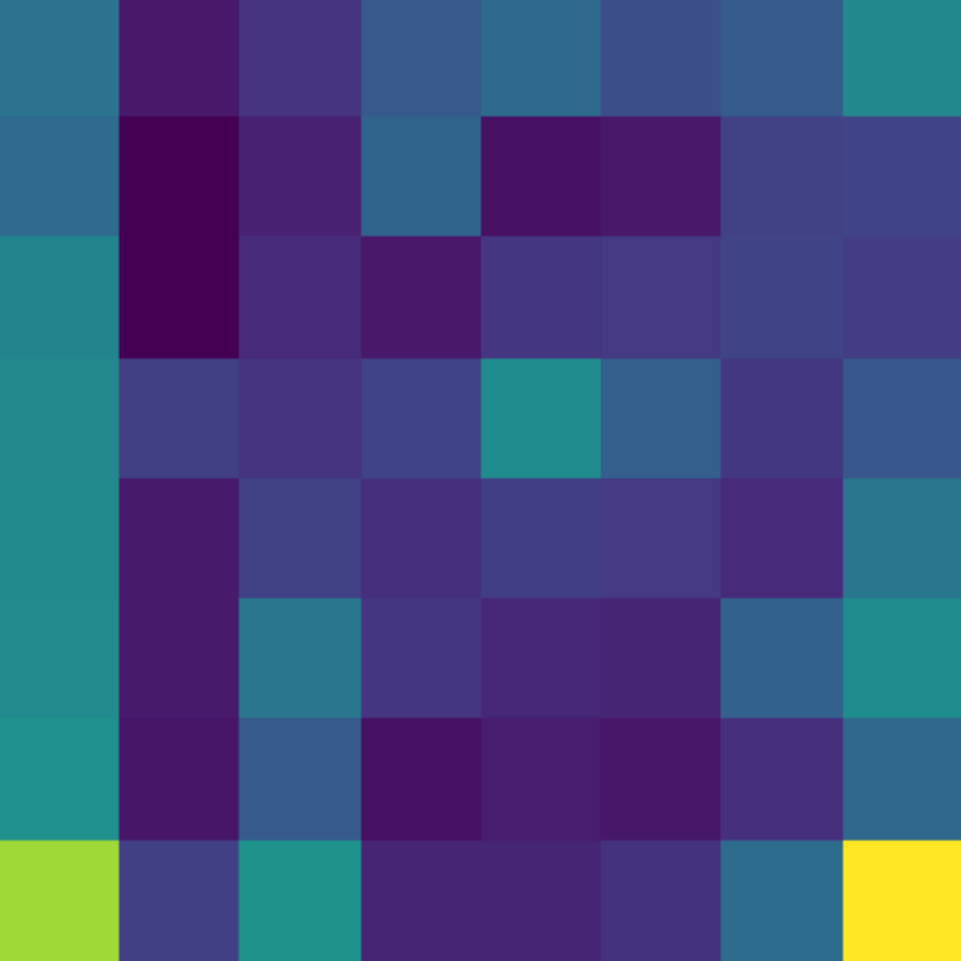}
\end{subfigure}
\begin{subfigure}{.05\textwidth}
  \centering
  \includegraphics[width=1.0\linewidth]{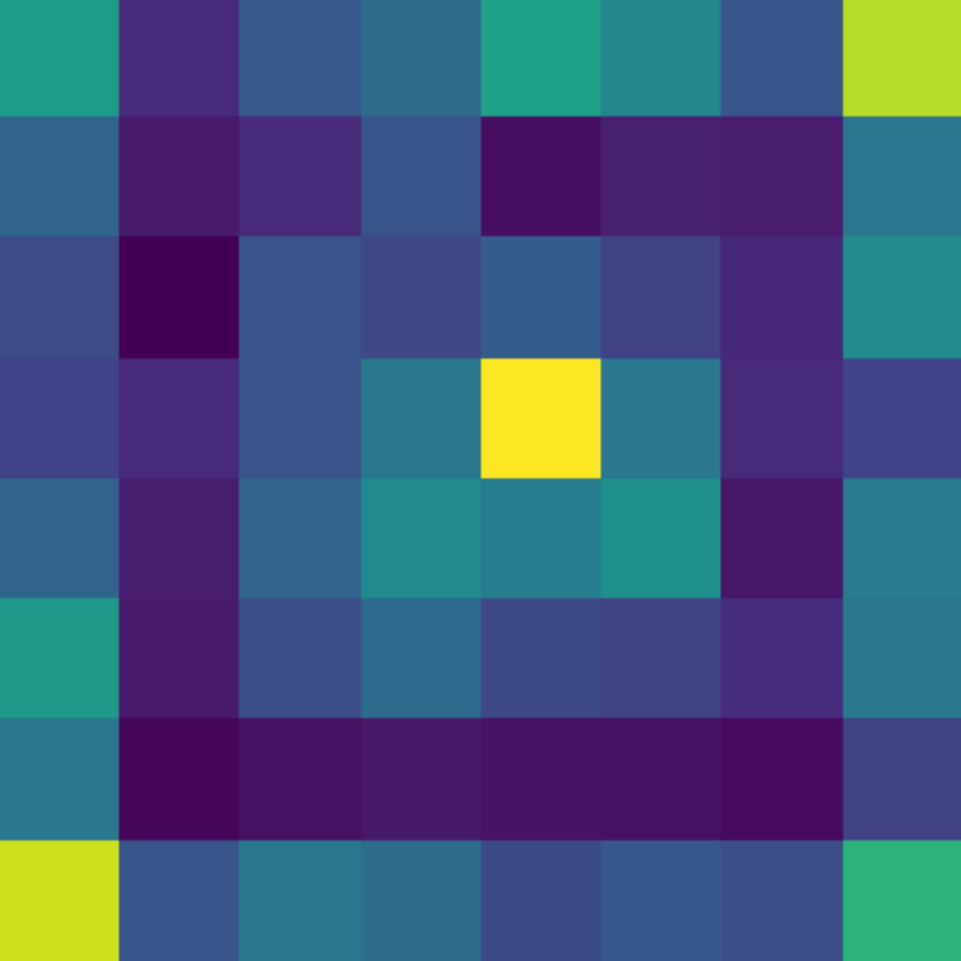}
\end{subfigure}
\begin{subfigure}{.05\textwidth}
  \centering
  \includegraphics[width=1.0\linewidth]{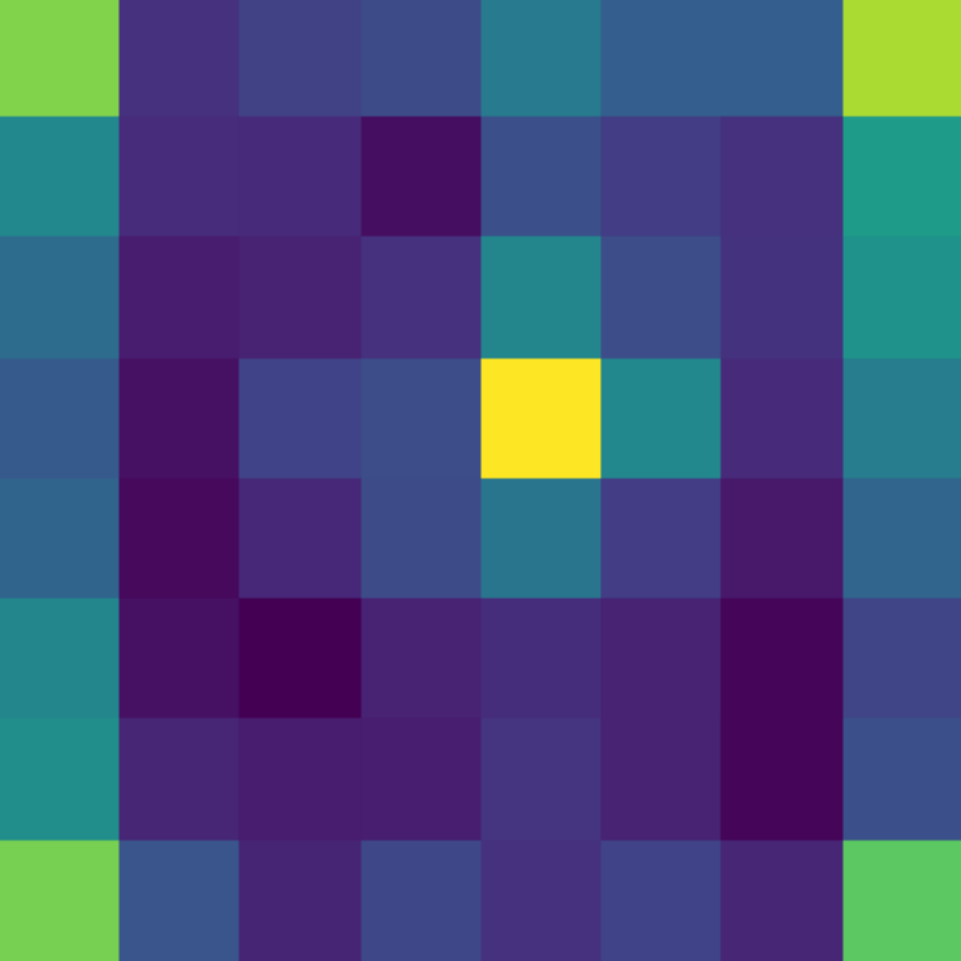}
\end{subfigure}
\begin{subfigure}{.05\textwidth}
  \centering
  \includegraphics[width=1.0\linewidth]{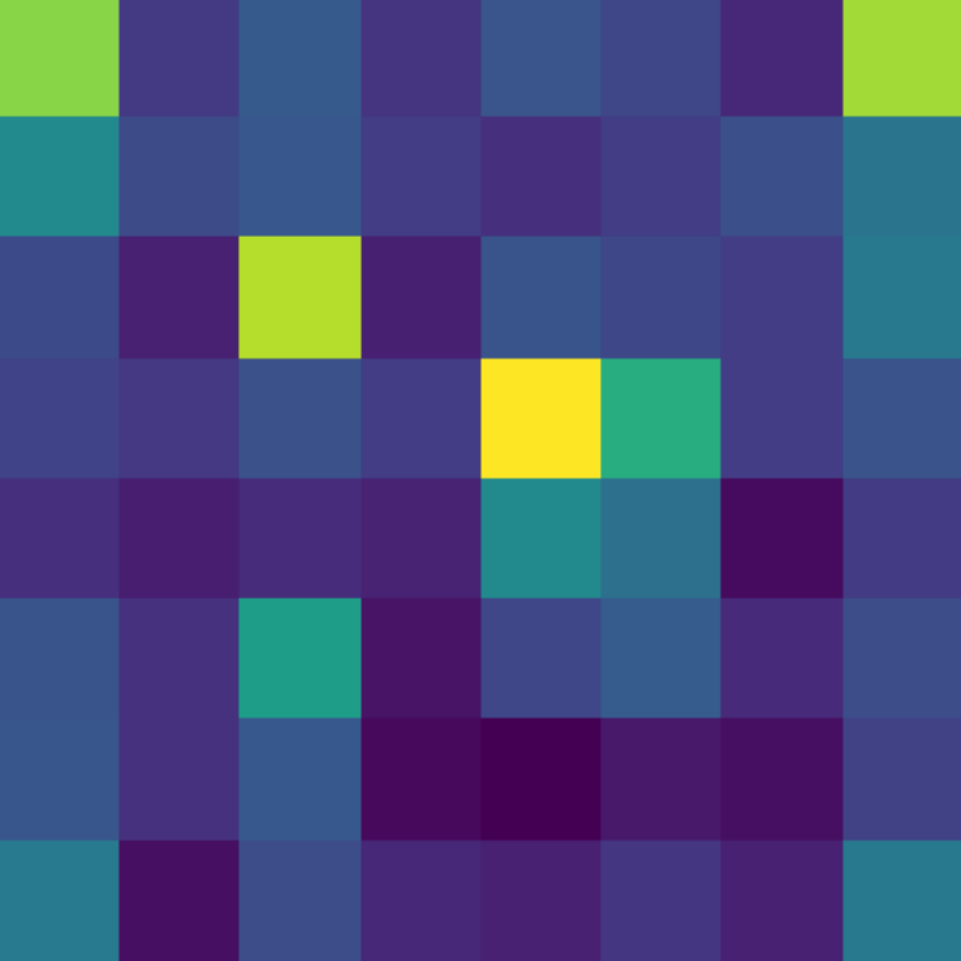}
\end{subfigure}
\begin{subfigure}{.05\textwidth}
  \centering
  \includegraphics[width=1.0\linewidth]{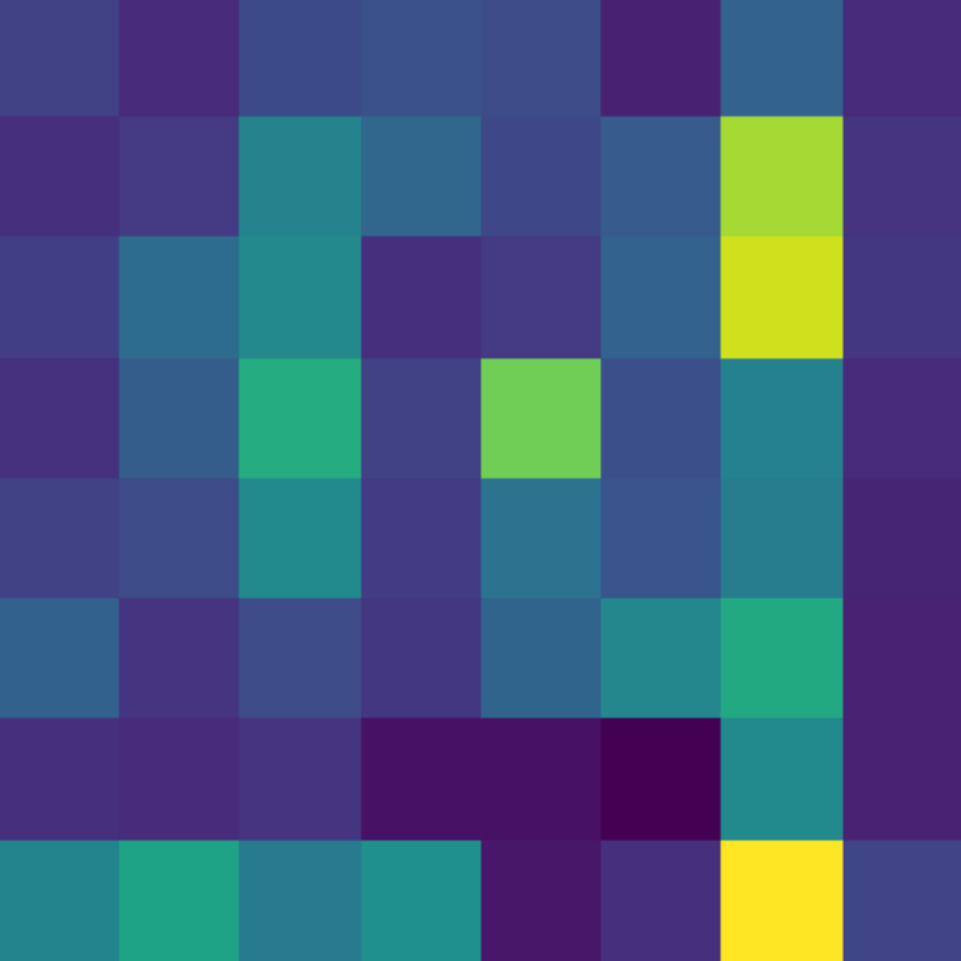}
\end{subfigure}
\begin{subfigure}{.05\textwidth}
  \centering
  \includegraphics[width=1.0\linewidth]{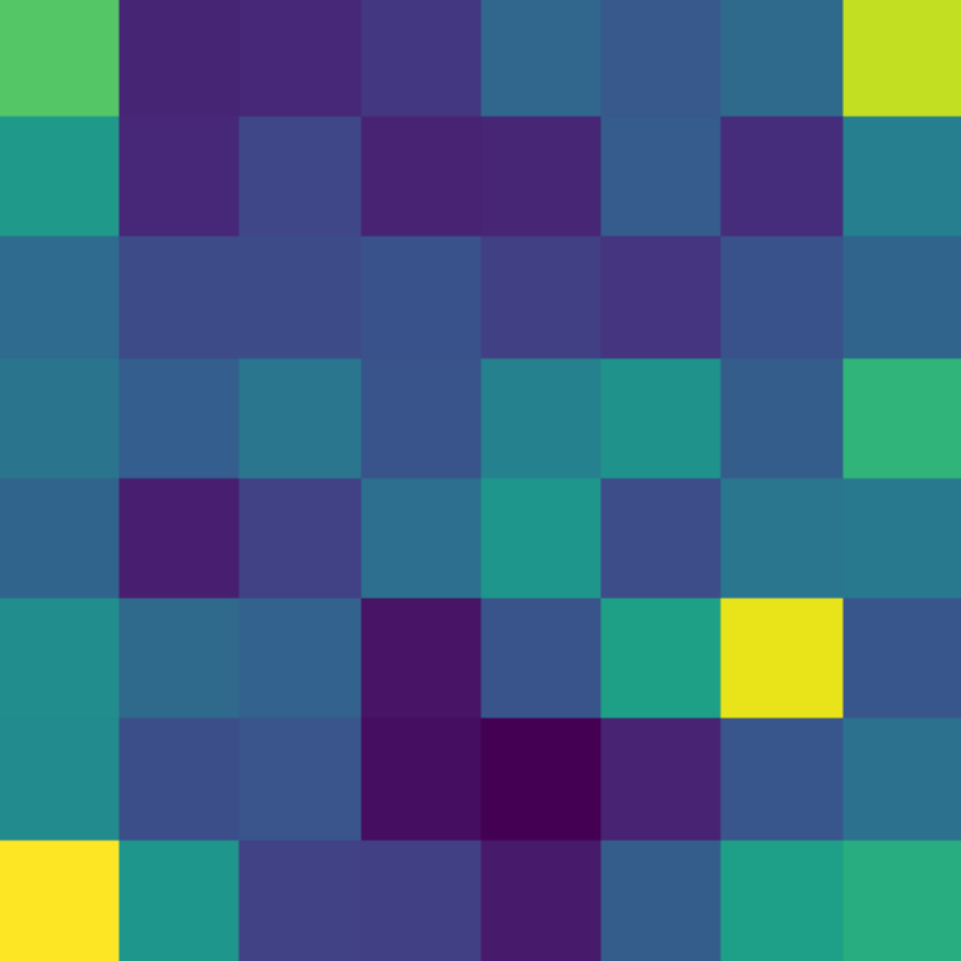}
\end{subfigure}
\begin{subfigure}{.05\textwidth}
  \centering
  \includegraphics[width=1.0\linewidth]{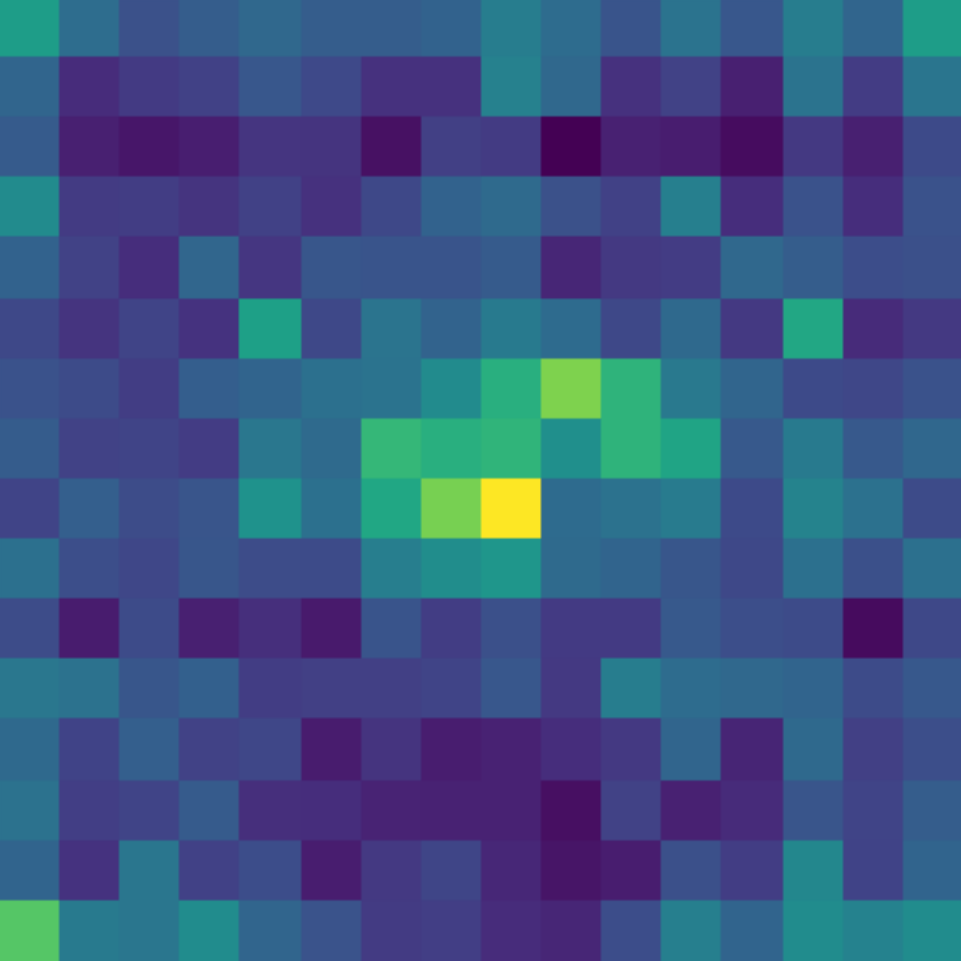}
\end{subfigure}
\begin{subfigure}{.05\textwidth}
  \centering
  \includegraphics[width=1.0\linewidth]{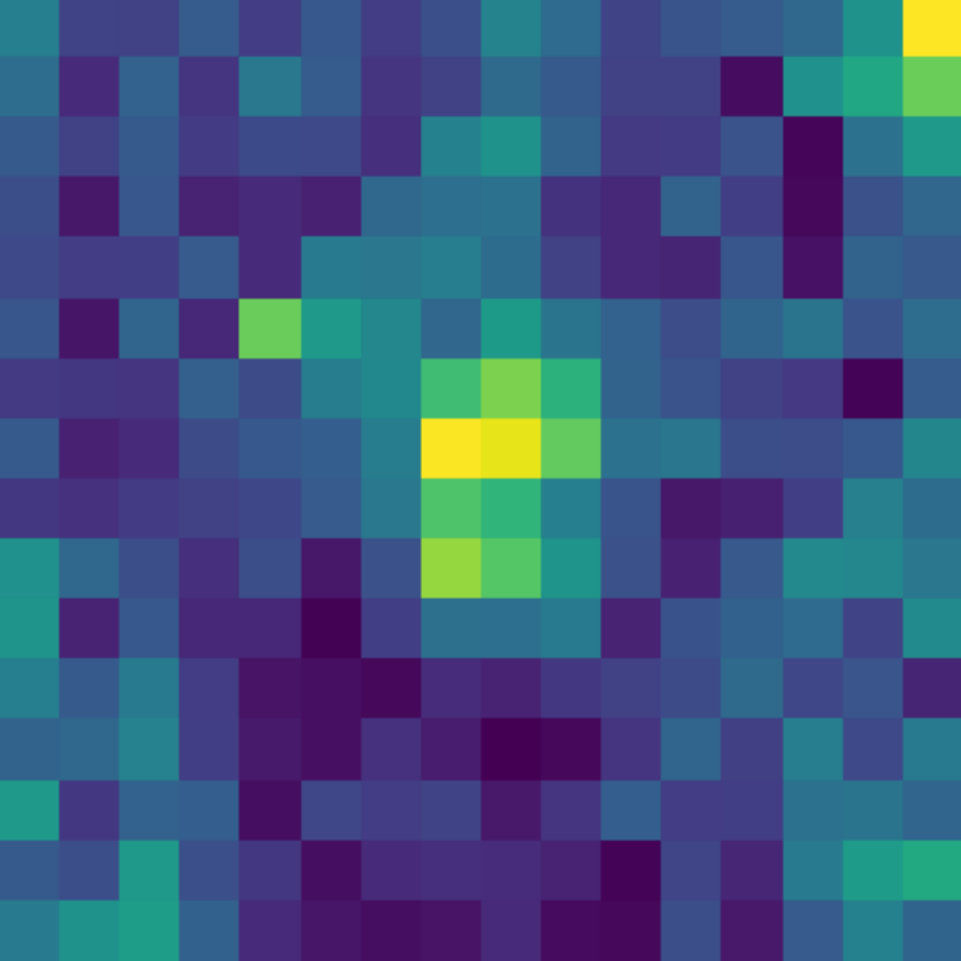}
\end{subfigure}
\begin{subfigure}{.05\textwidth}
  \centering
  \includegraphics[width=1.0\linewidth]{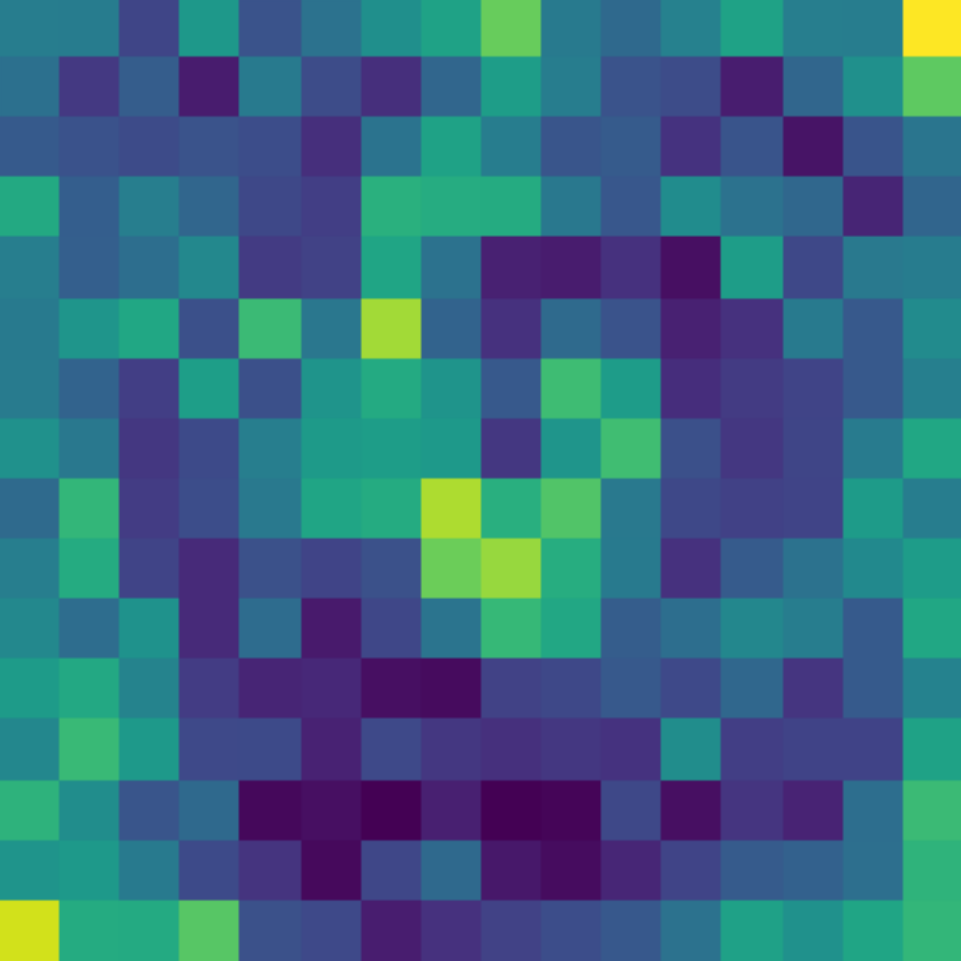}
\end{subfigure}
\begin{subfigure}{.05\textwidth}
  \centering
  \includegraphics[width=1.0\linewidth]{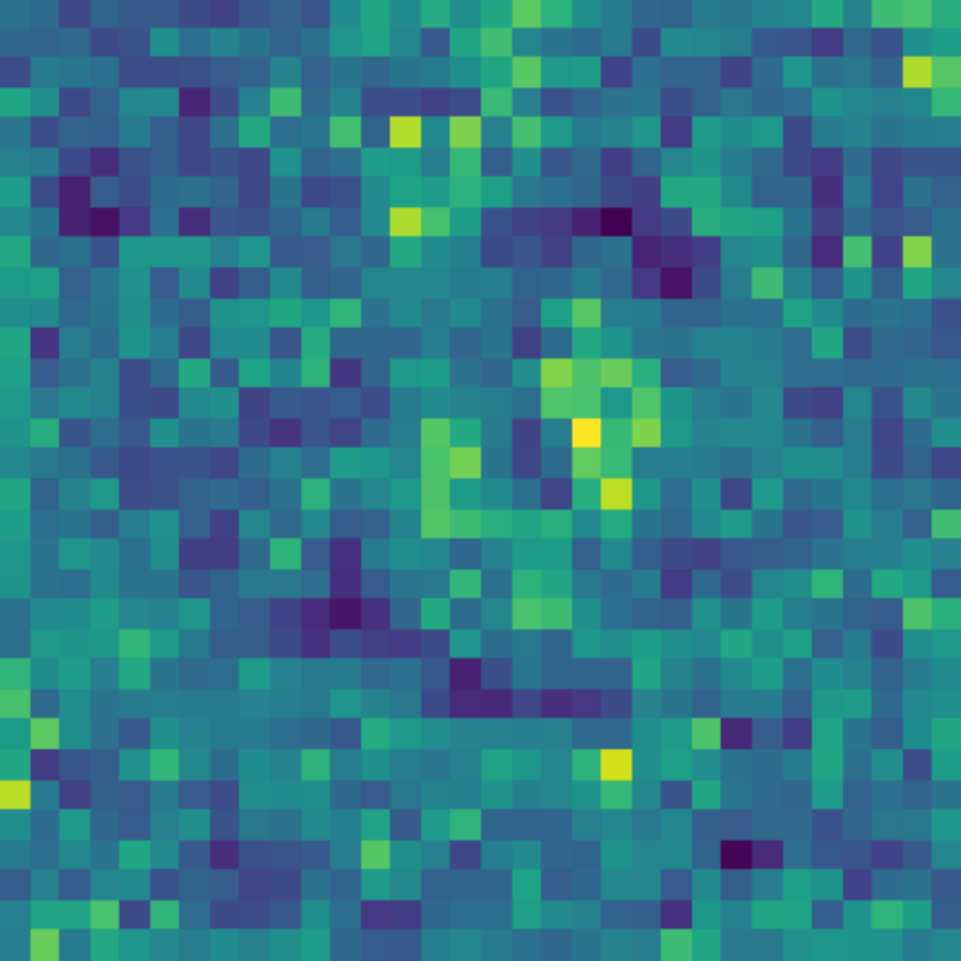}
\end{subfigure}
\begin{subfigure}{.05\textwidth}
  \centering
  \includegraphics[width=1.0\linewidth]{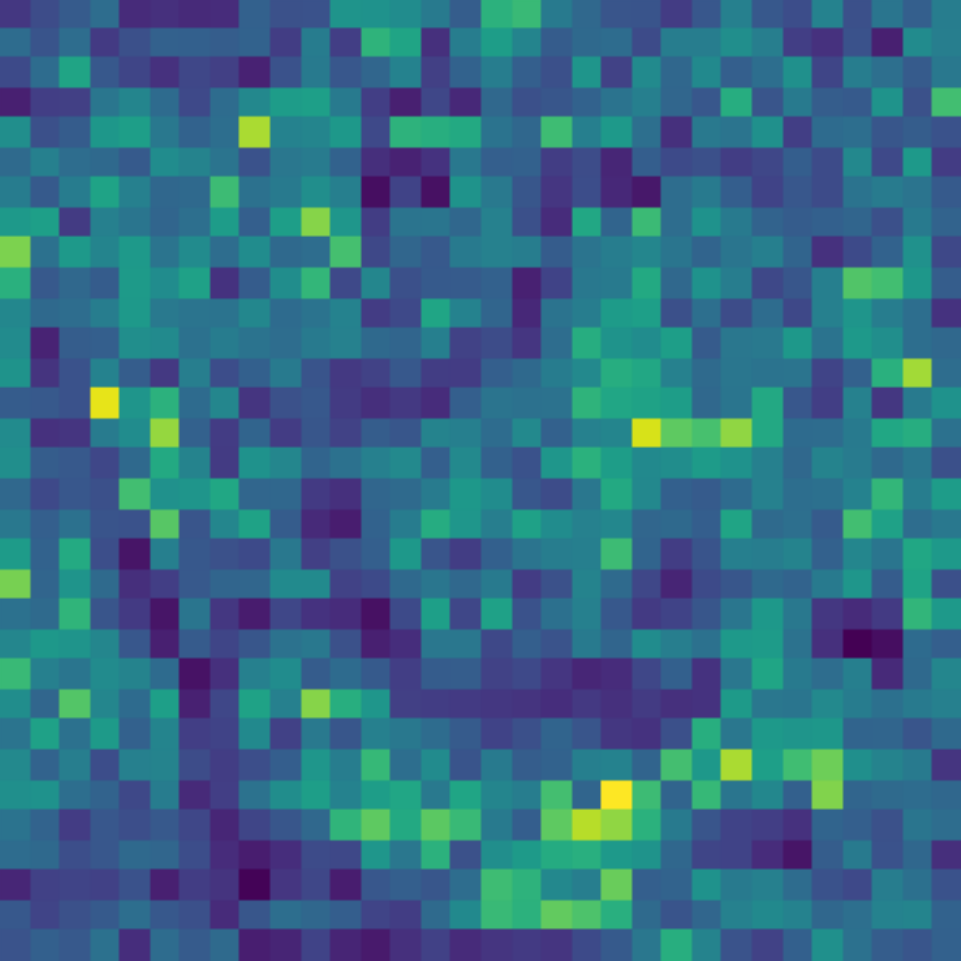}
\end{subfigure}
\begin{subfigure}{.05\textwidth}
  \centering
  \includegraphics[width=1.0\linewidth]{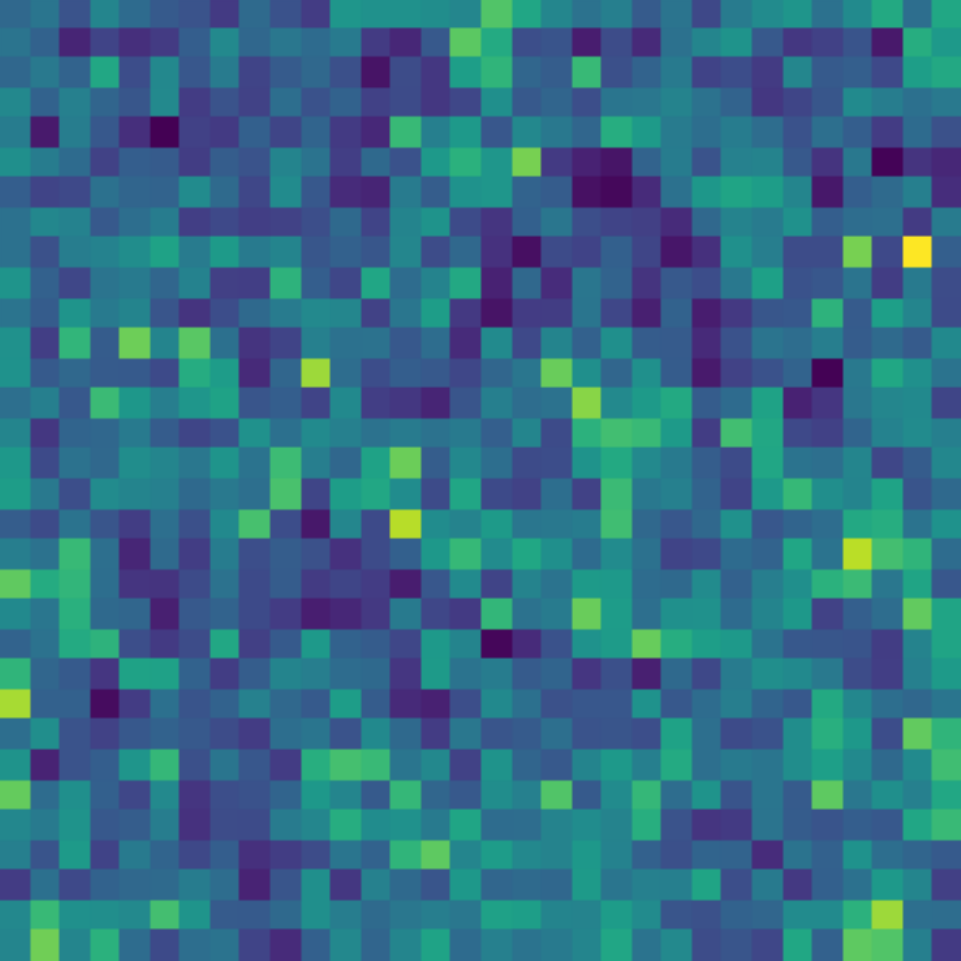}
\end{subfigure}
\\
&
\begin{subfigure}{.05\textwidth}
  \centering
  \includegraphics[width=1.0\linewidth]{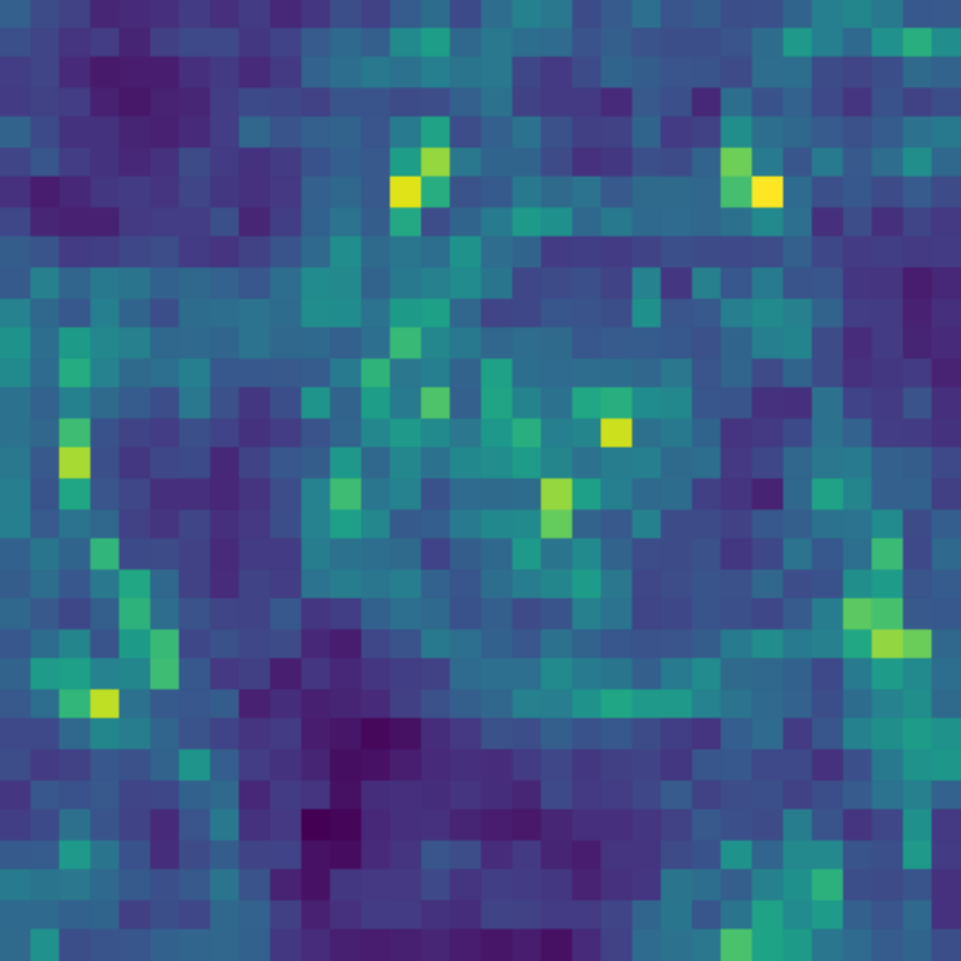}
\end{subfigure}
\begin{subfigure}{.05\textwidth}
  \centering
  \includegraphics[width=1.0\linewidth]{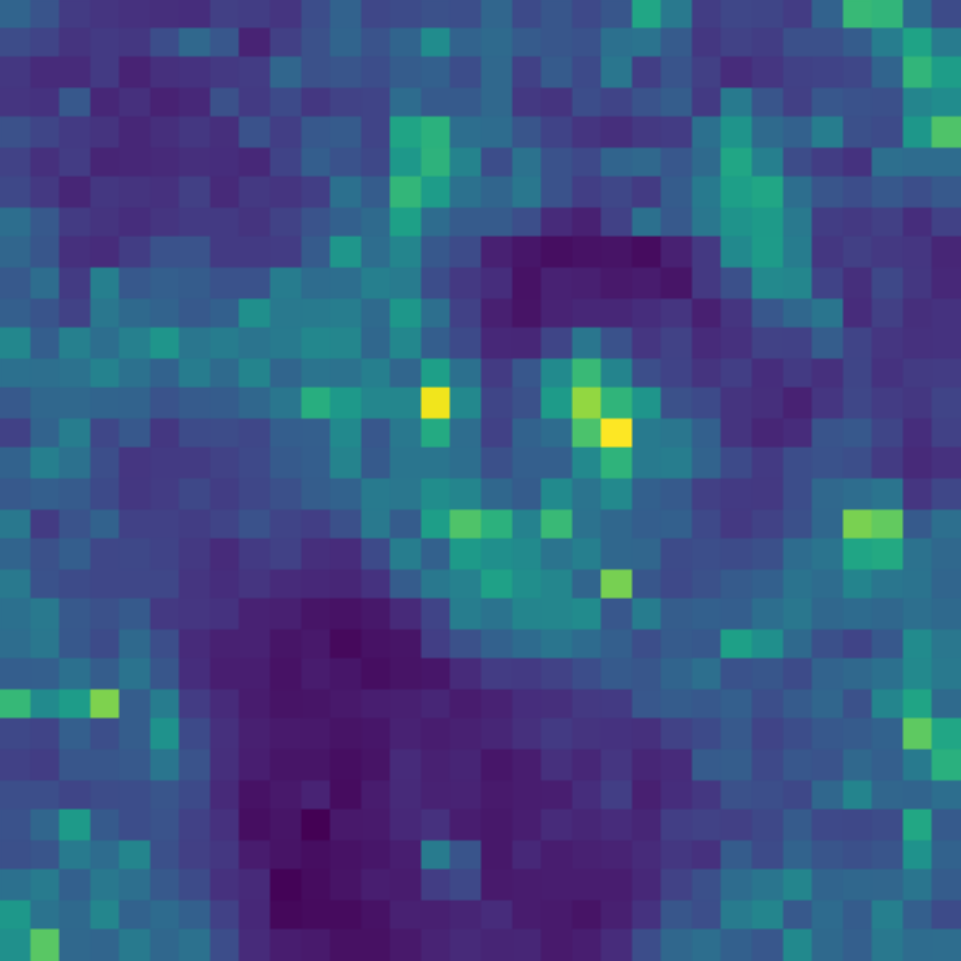}
\end{subfigure}
\begin{subfigure}{.05\textwidth}
  \centering
  \includegraphics[width=1.0\linewidth]{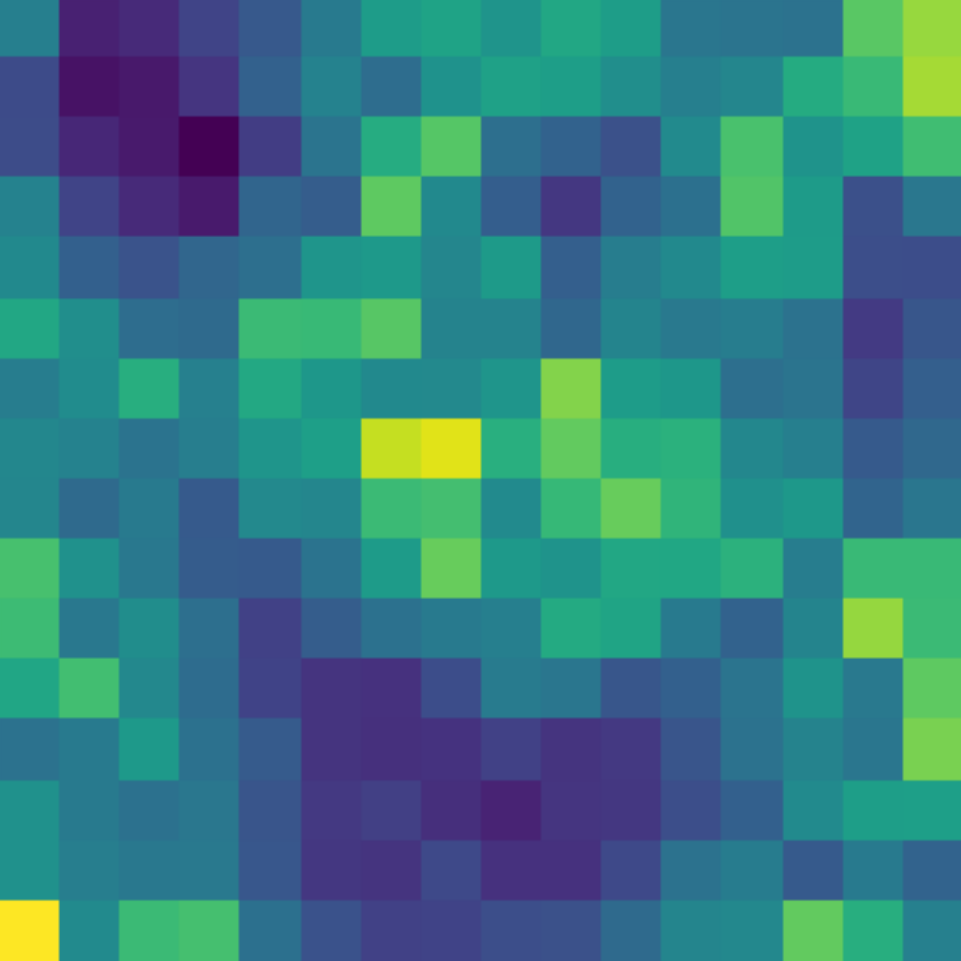}
\end{subfigure}
\begin{subfigure}{.05\textwidth}
  \centering
  \includegraphics[width=1.0\linewidth]{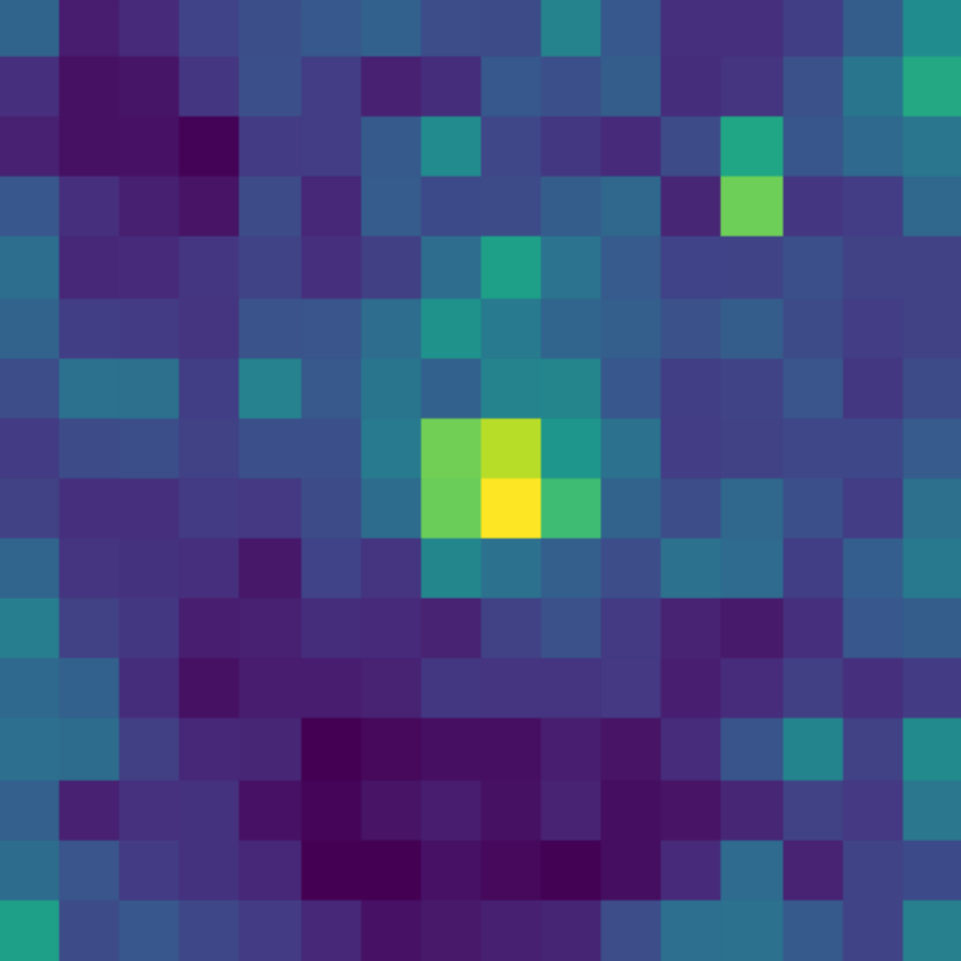}
\end{subfigure}
\begin{subfigure}{.05\textwidth}
  \centering
  \includegraphics[width=1.0\linewidth]{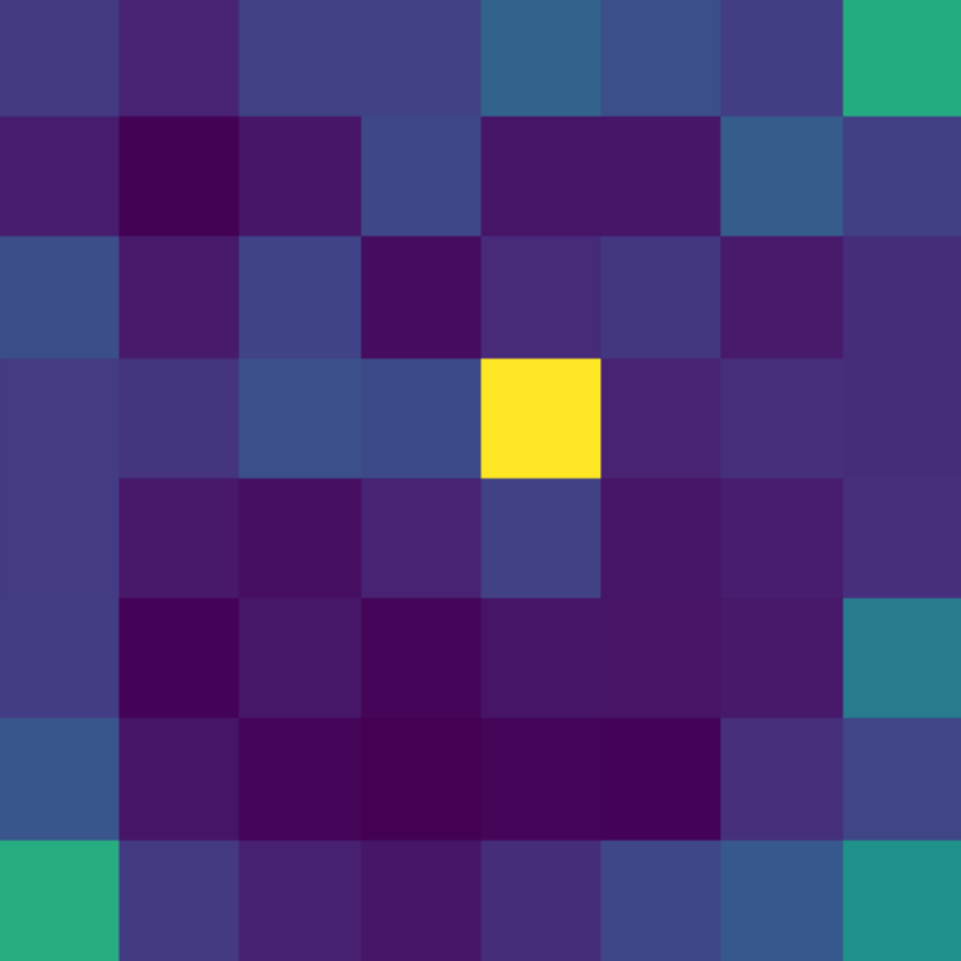}
\end{subfigure}
\begin{subfigure}{.05\textwidth}
  \centering
  \includegraphics[width=1.0\linewidth]{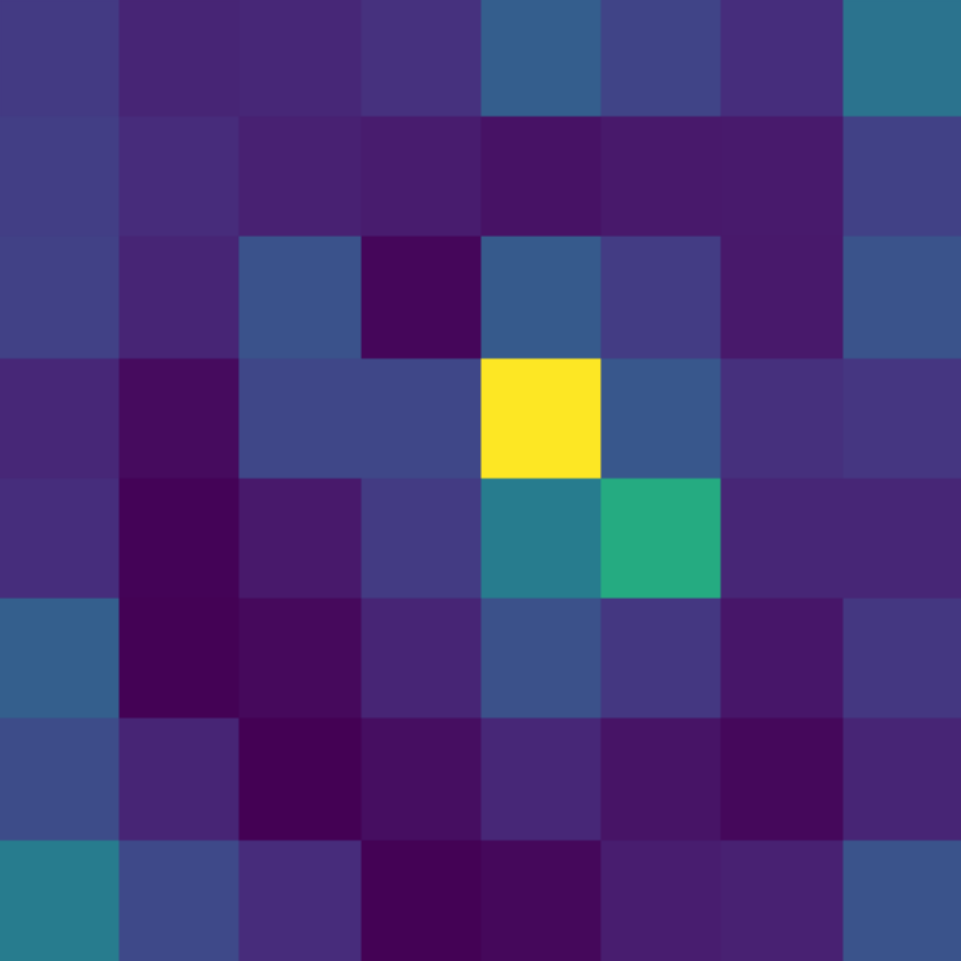}
\end{subfigure}
\begin{subfigure}{.05\textwidth}
  \centering
  \includegraphics[width=1.0\linewidth]{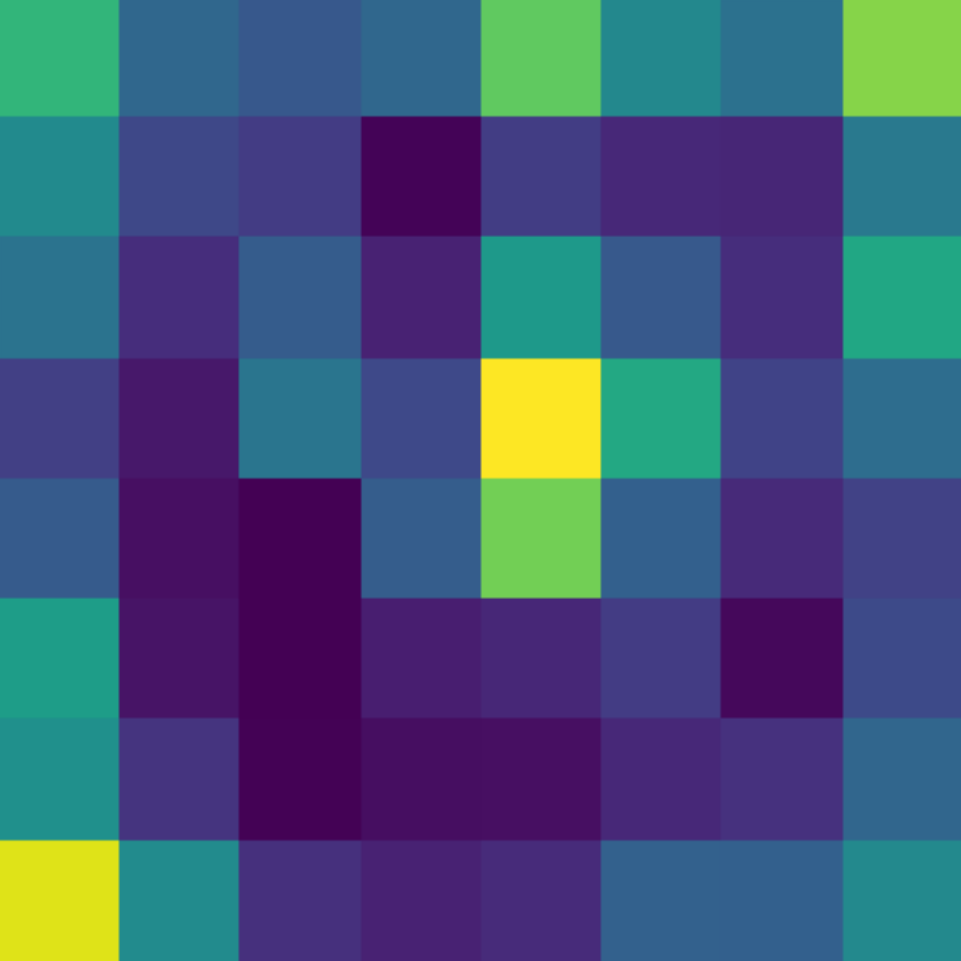}
\end{subfigure}
\begin{subfigure}{.05\textwidth}
  \centering
  \includegraphics[width=1.0\linewidth]{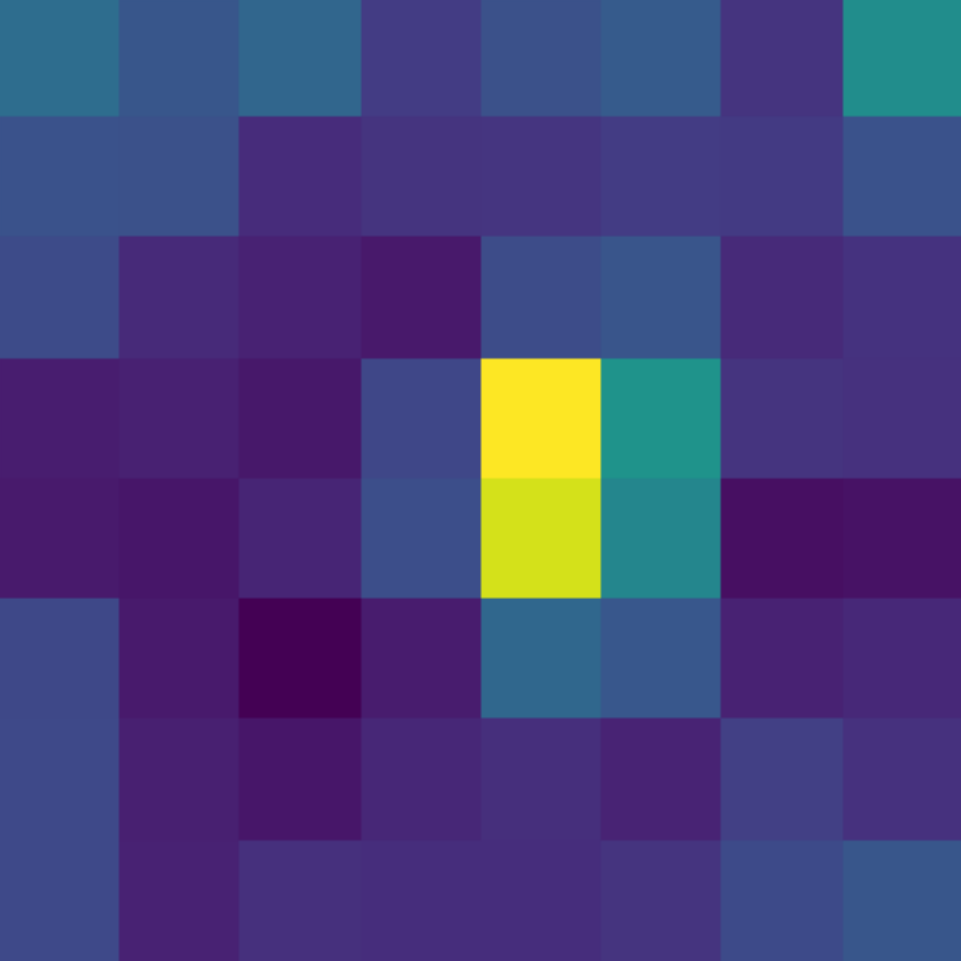}
\end{subfigure}
\begin{subfigure}{.05\textwidth}
  \centering
  \includegraphics[width=1.0\linewidth]{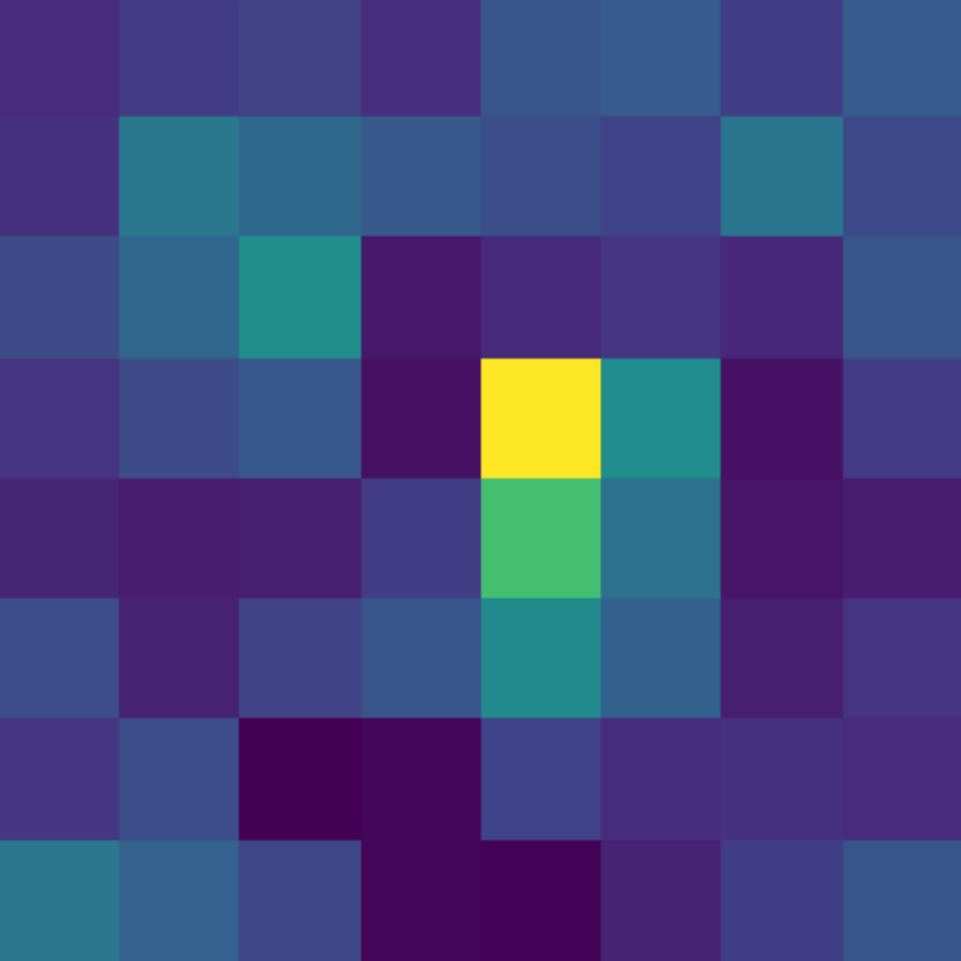}
\end{subfigure}
\begin{subfigure}{.05\textwidth}
  \centering
  \includegraphics[width=1.0\linewidth]{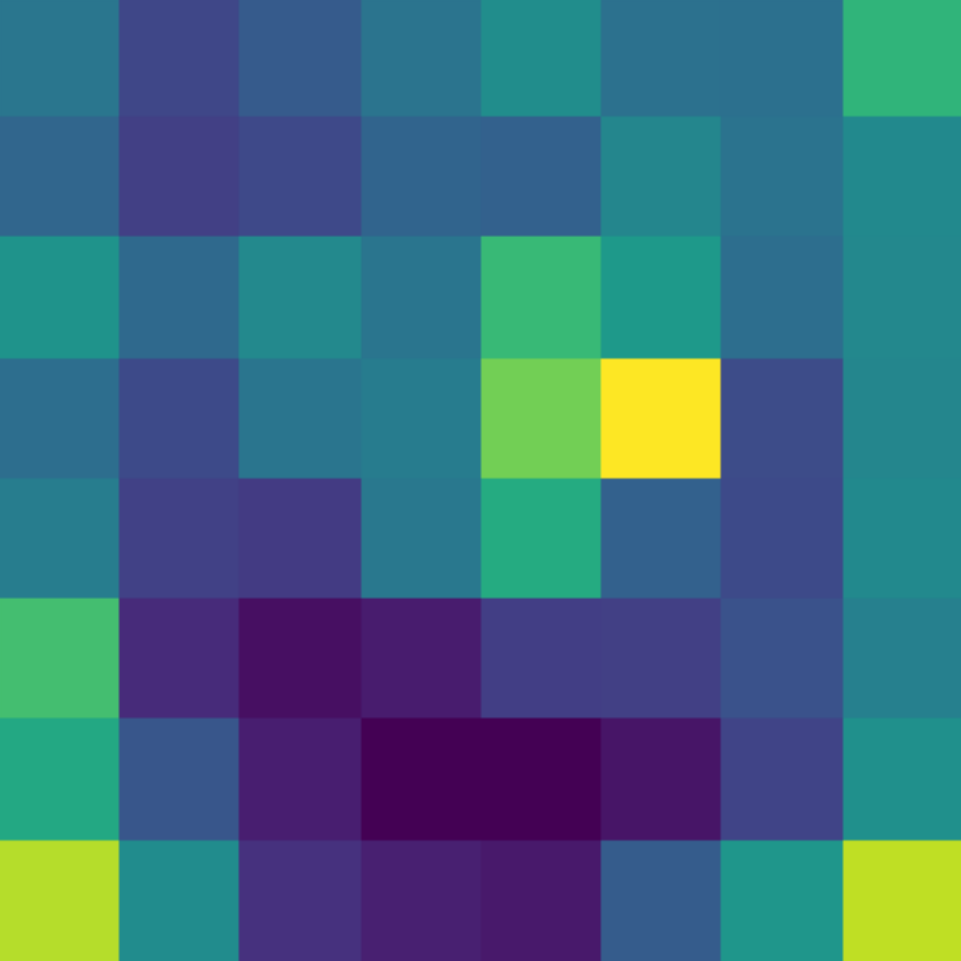}
\end{subfigure}
\begin{subfigure}{.05\textwidth}
  \centering
  \includegraphics[width=1.0\linewidth]{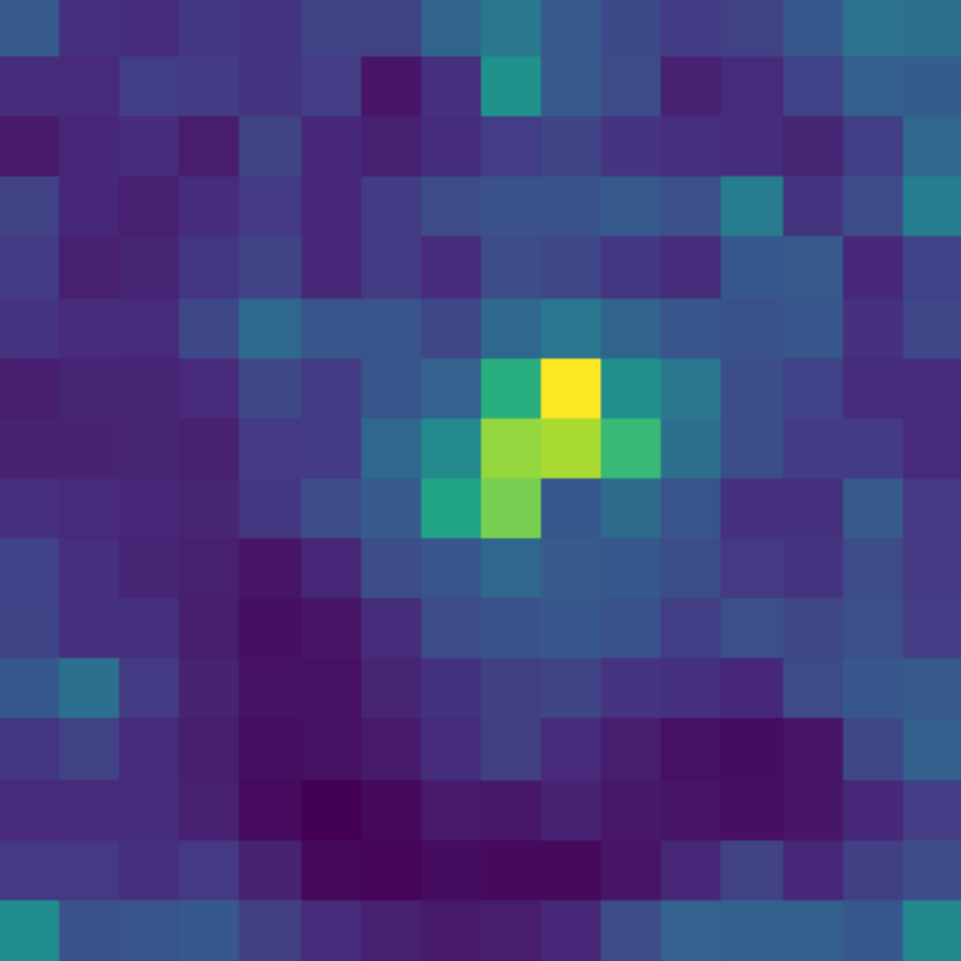}
\end{subfigure}
\begin{subfigure}{.05\textwidth}
  \centering
  \includegraphics[width=1.0\linewidth]{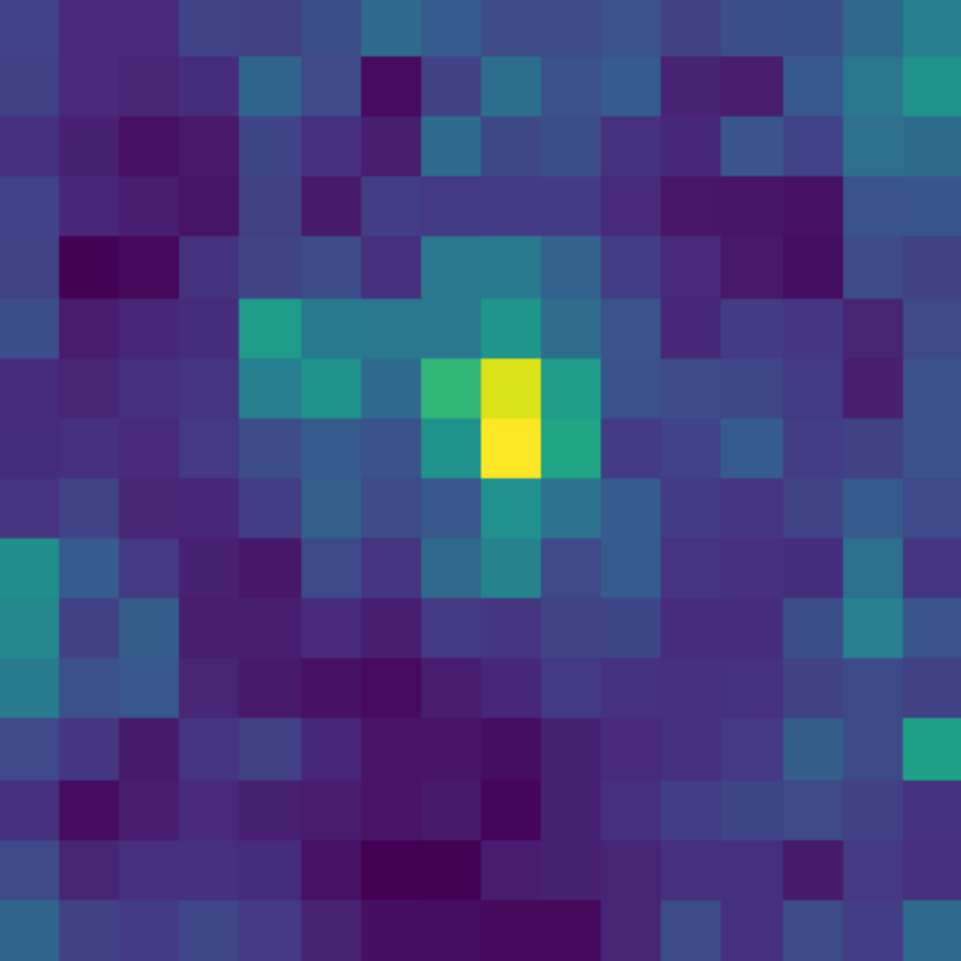}
\end{subfigure}
\begin{subfigure}{.05\textwidth}
  \centering
  \includegraphics[width=1.0\linewidth]{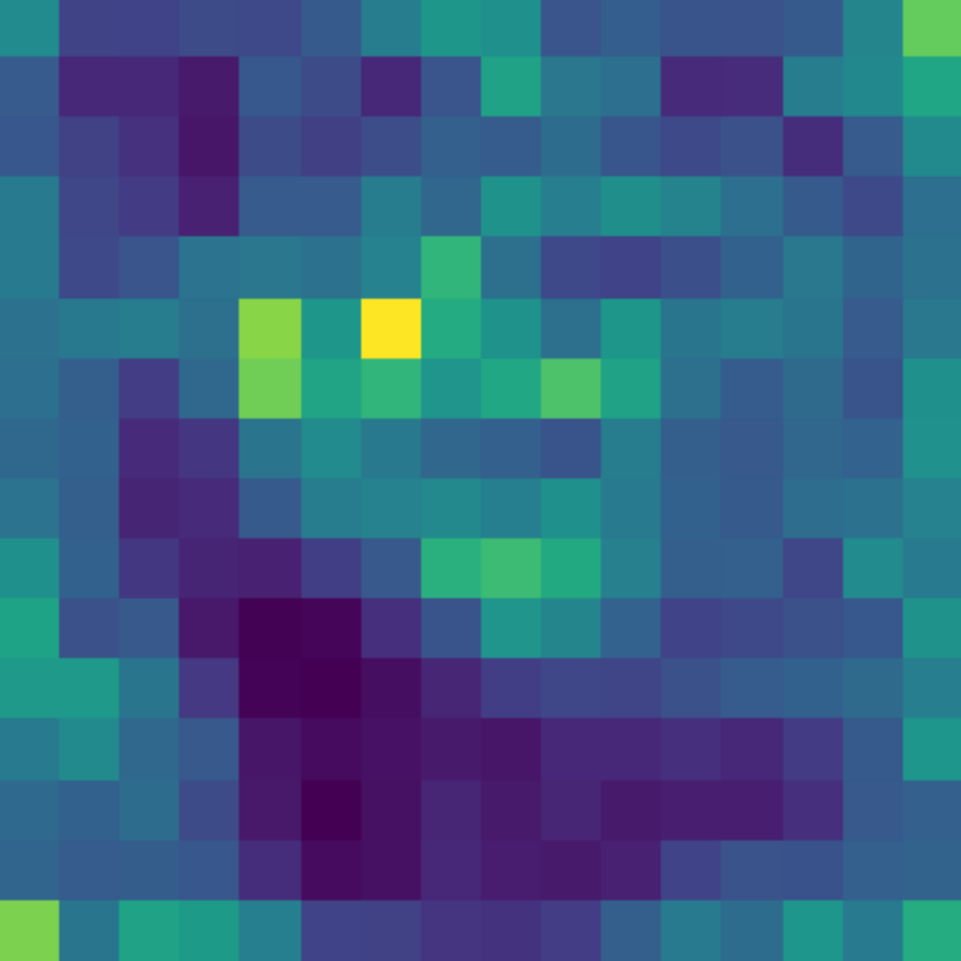}
\end{subfigure}
\begin{subfigure}{.05\textwidth}
  \centering
  \includegraphics[width=1.0\linewidth]{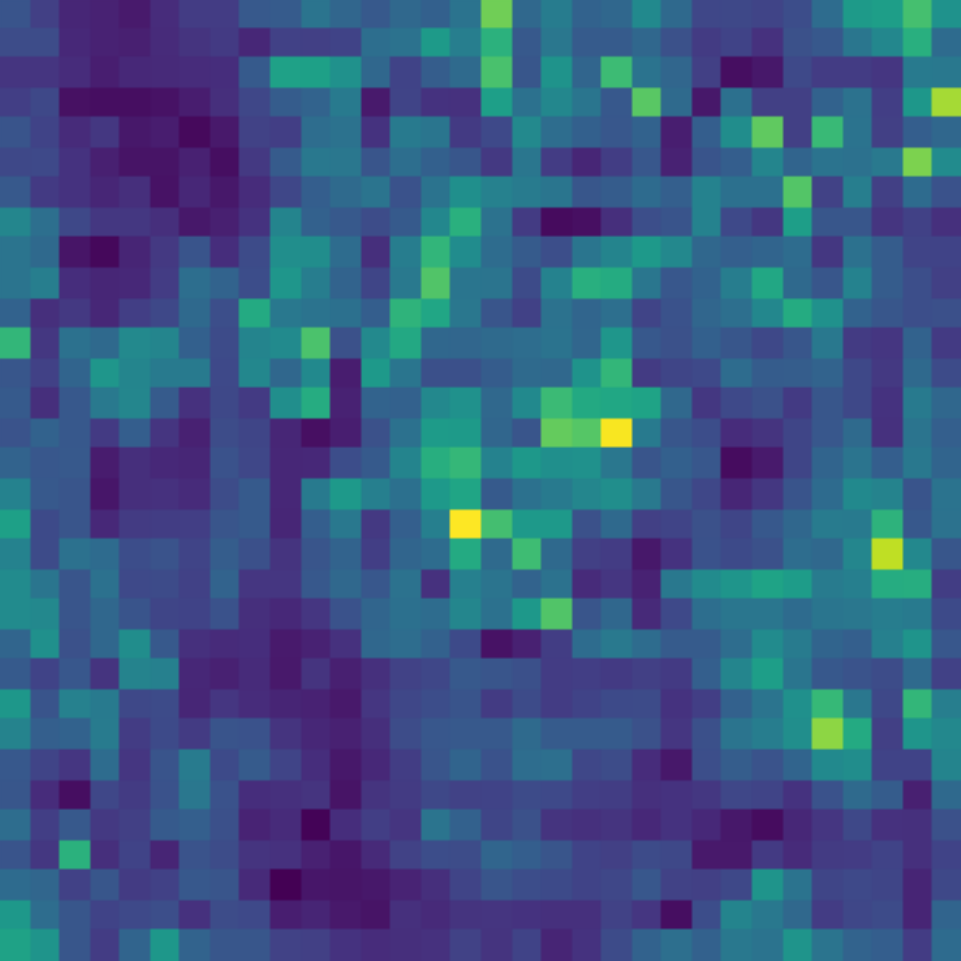}
\end{subfigure}
\begin{subfigure}{.05\textwidth}
  \centering
  \includegraphics[width=1.0\linewidth]{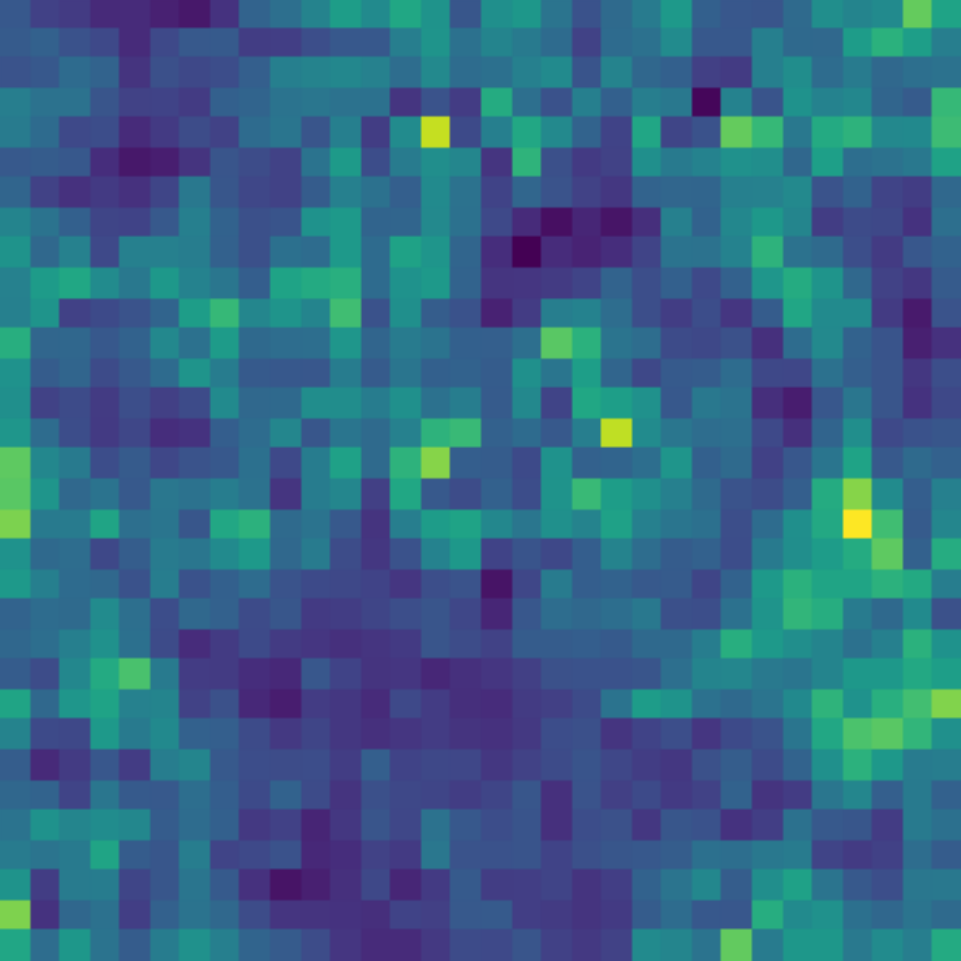}
\end{subfigure}
\begin{subfigure}{.05\textwidth}
  \centering
  \includegraphics[width=1.0\linewidth]{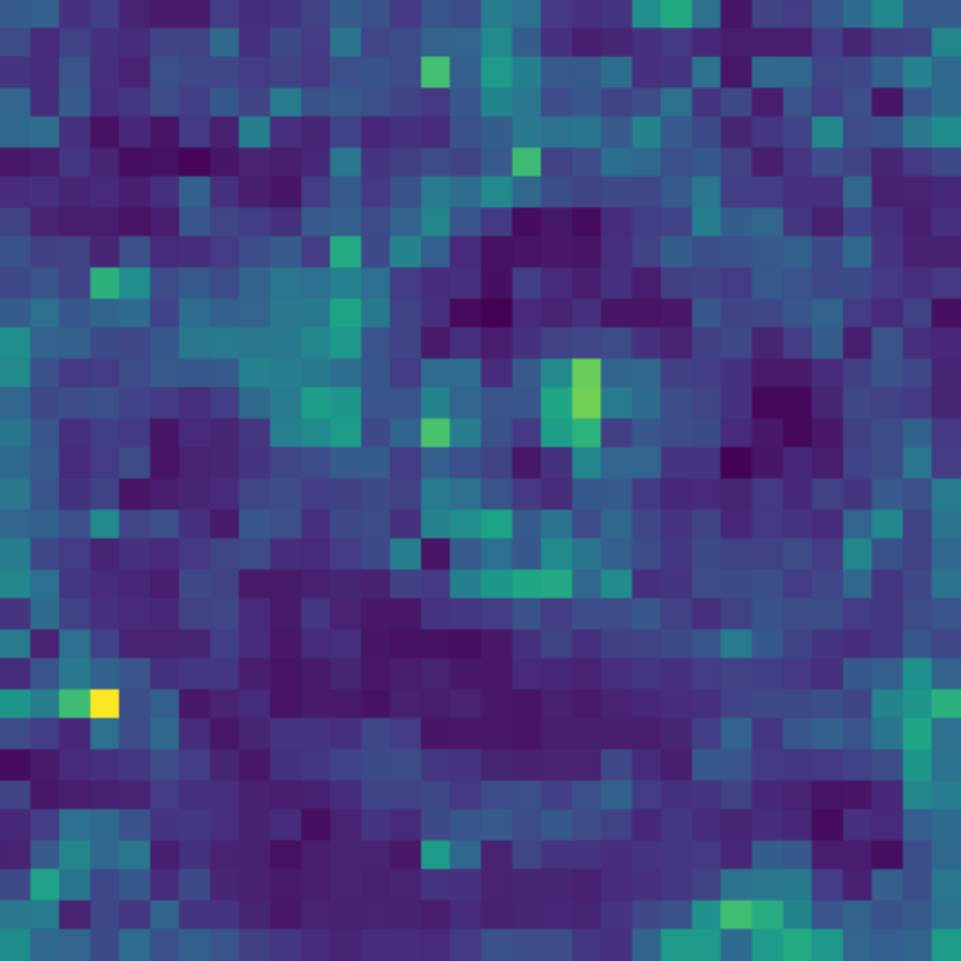}
\end{subfigure}
\\
&
\begin{subfigure}{.05\textwidth}
  \centering
  \includegraphics[width=1.0\linewidth]{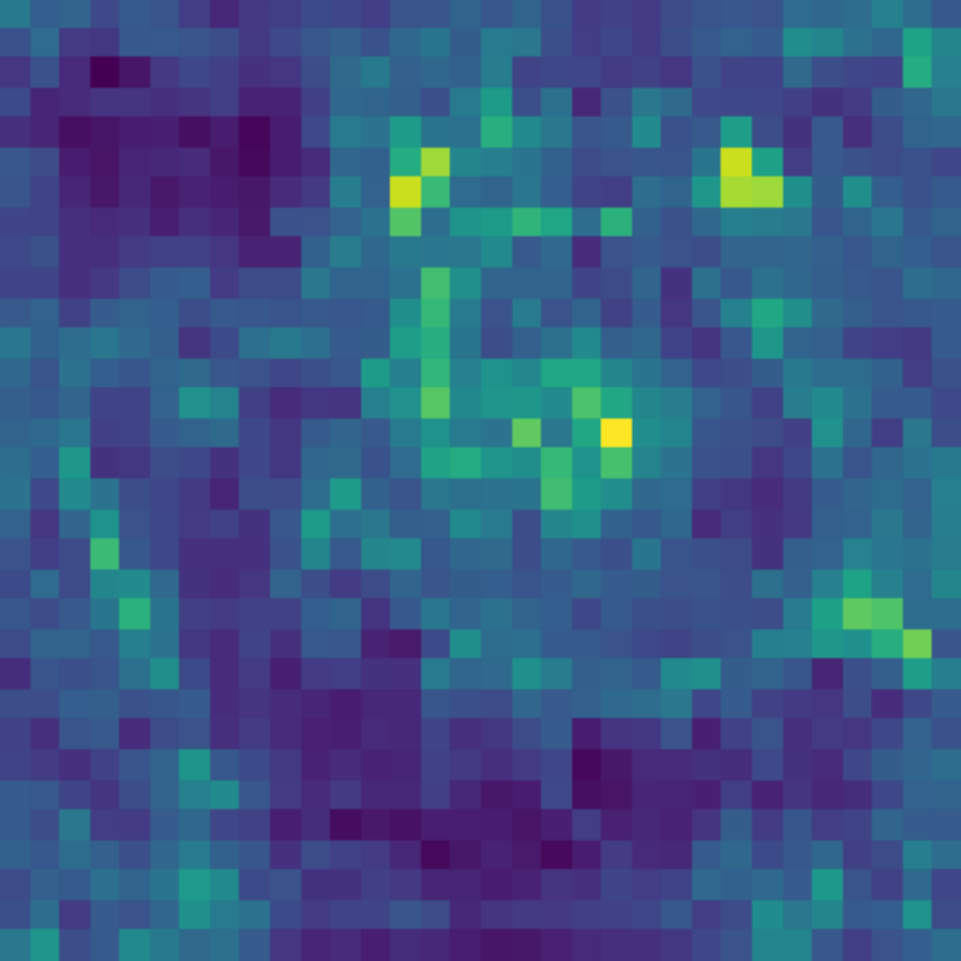}
\end{subfigure}
\begin{subfigure}{.05\textwidth}
  \centering
  \includegraphics[width=1.0\linewidth]{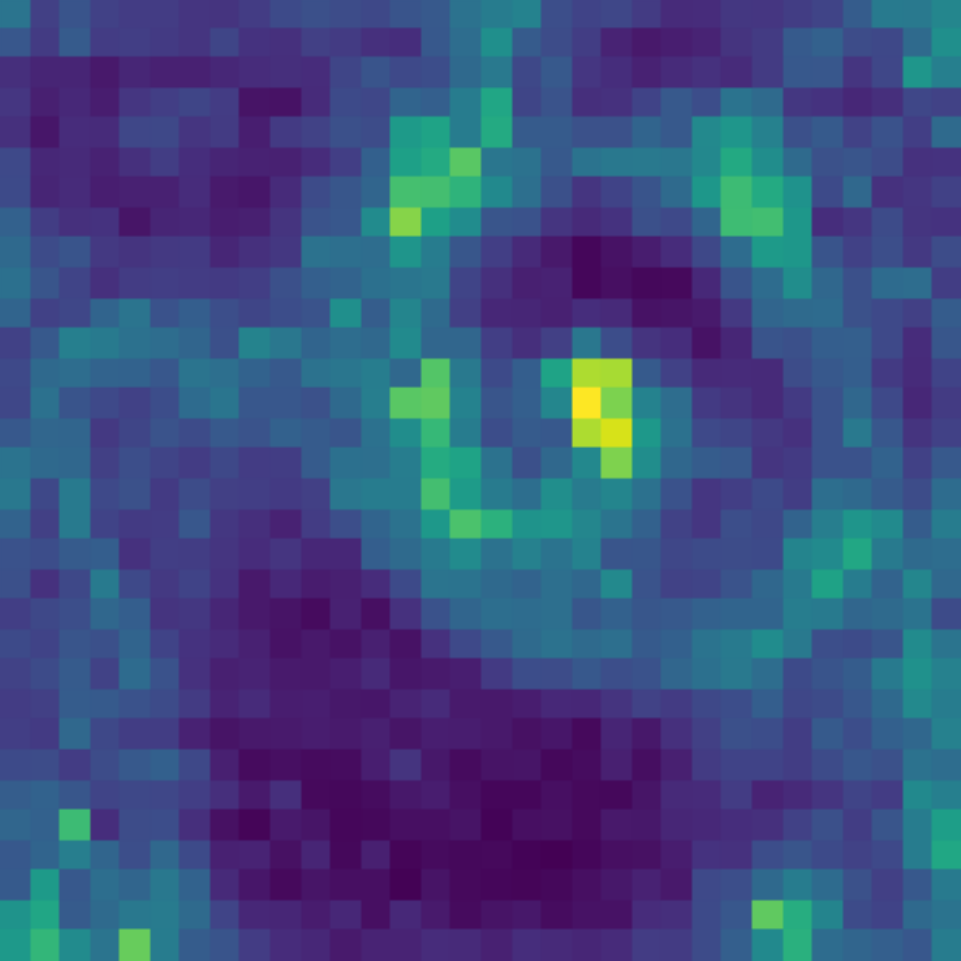}
\end{subfigure}
\begin{subfigure}{.05\textwidth}
  \centering
  \includegraphics[width=1.0\linewidth]{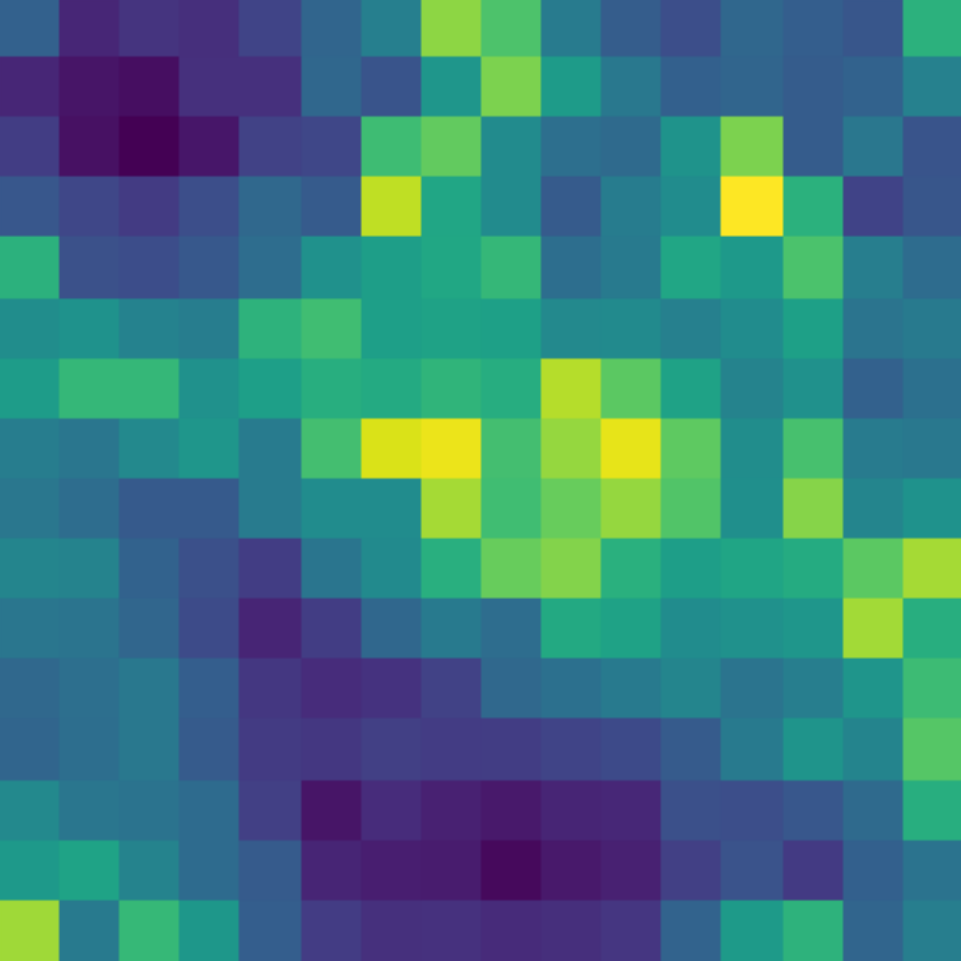}
\end{subfigure}
\begin{subfigure}{.05\textwidth}
  \centering
  \includegraphics[width=1.0\linewidth]{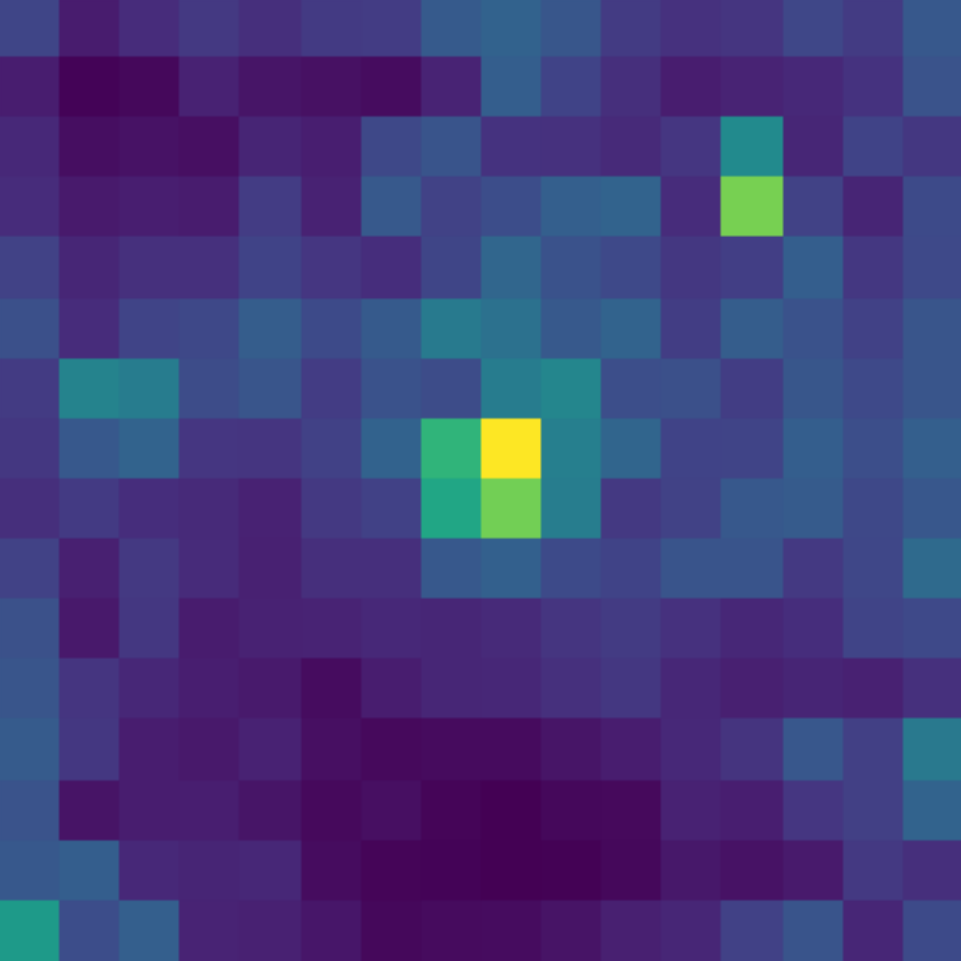}
\end{subfigure}
\begin{subfigure}{.05\textwidth}
  \centering
  \includegraphics[width=1.0\linewidth]{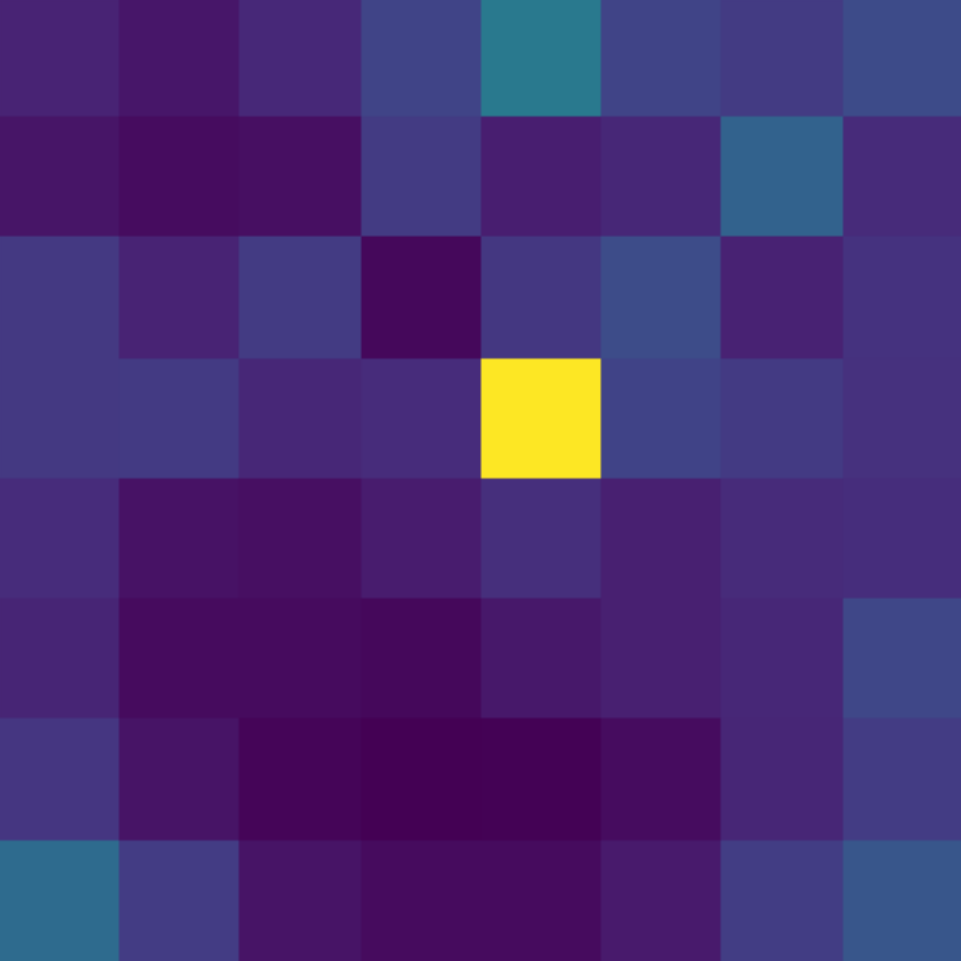}
\end{subfigure}
\begin{subfigure}{.05\textwidth}
  \centering
  \includegraphics[width=1.0\linewidth]{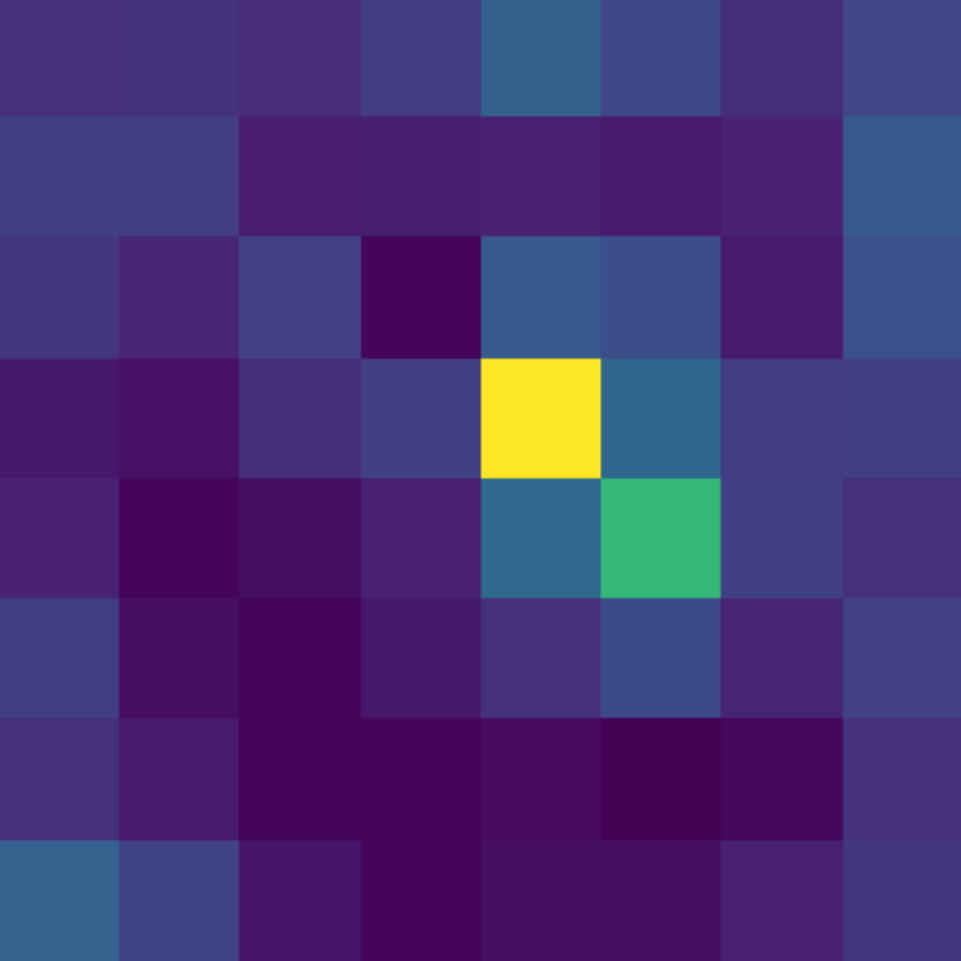}
\end{subfigure}
\begin{subfigure}{.05\textwidth}
  \centering
  \includegraphics[width=1.0\linewidth]{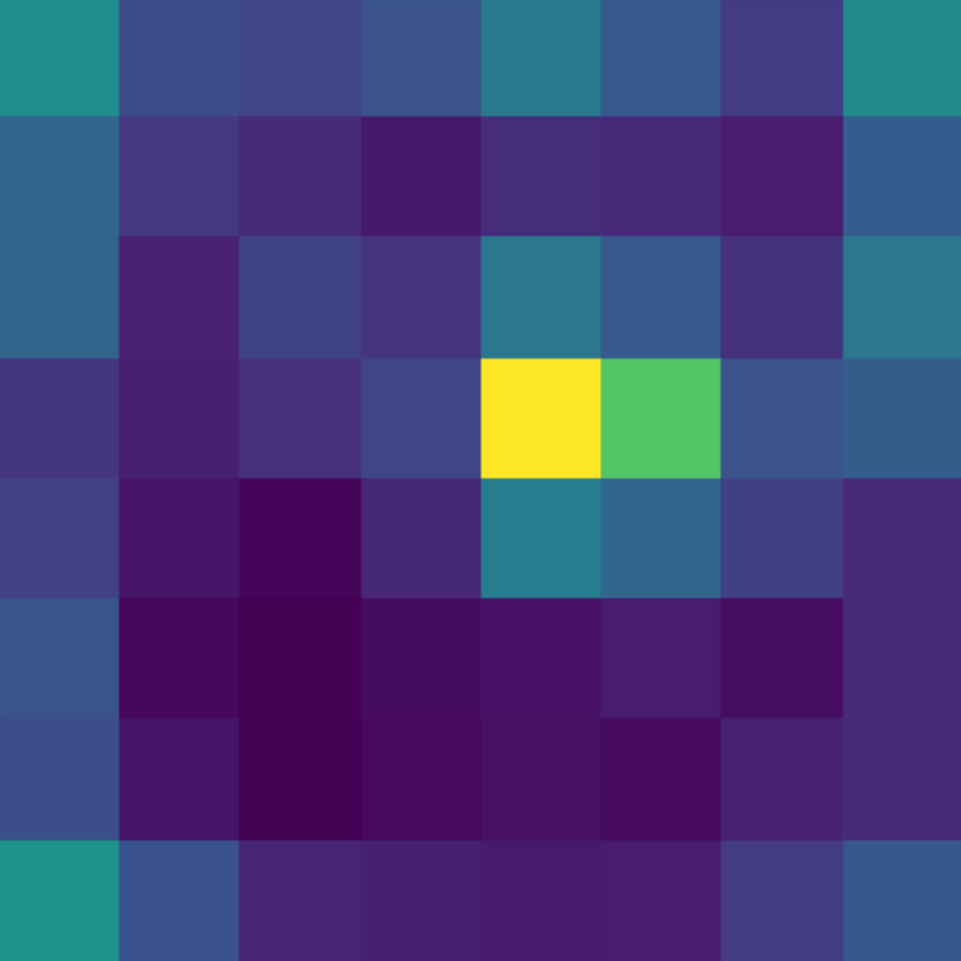}
\end{subfigure}
\begin{subfigure}{.05\textwidth}
  \centering
  \includegraphics[width=1.0\linewidth]{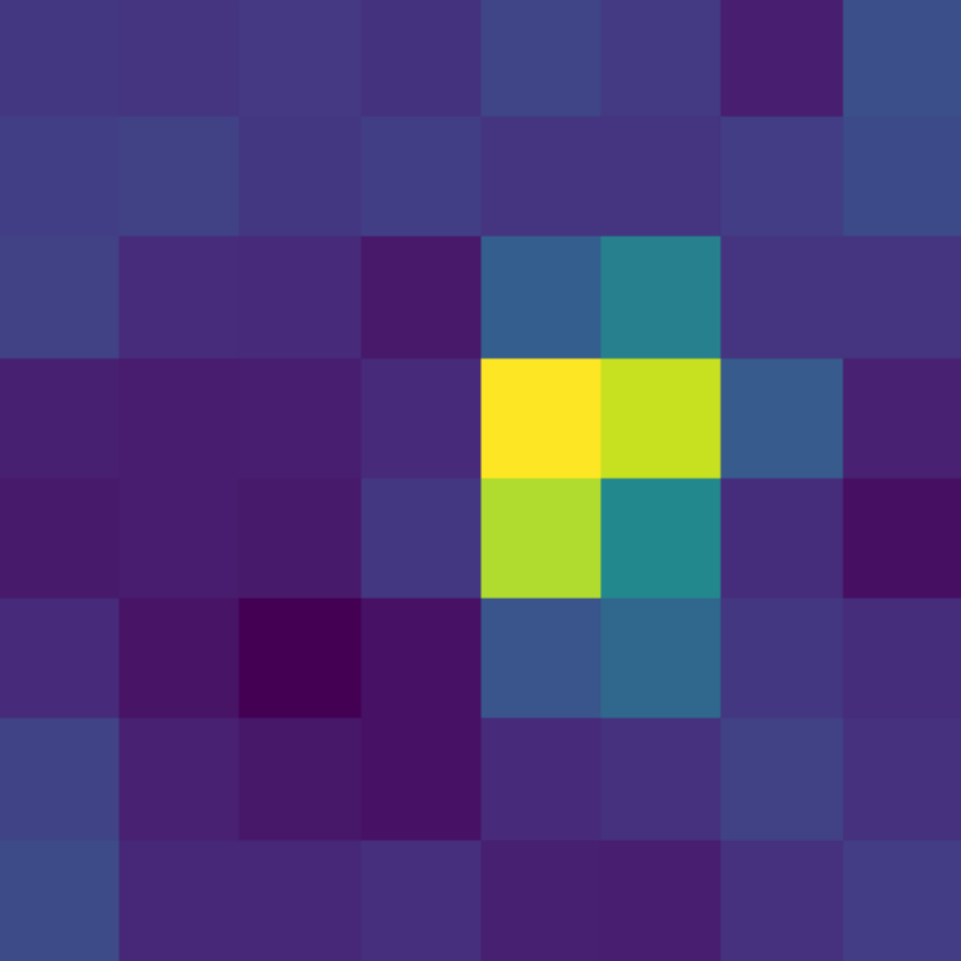}
\end{subfigure}
\begin{subfigure}{.05\textwidth}
  \centering
  \includegraphics[width=1.0\linewidth]{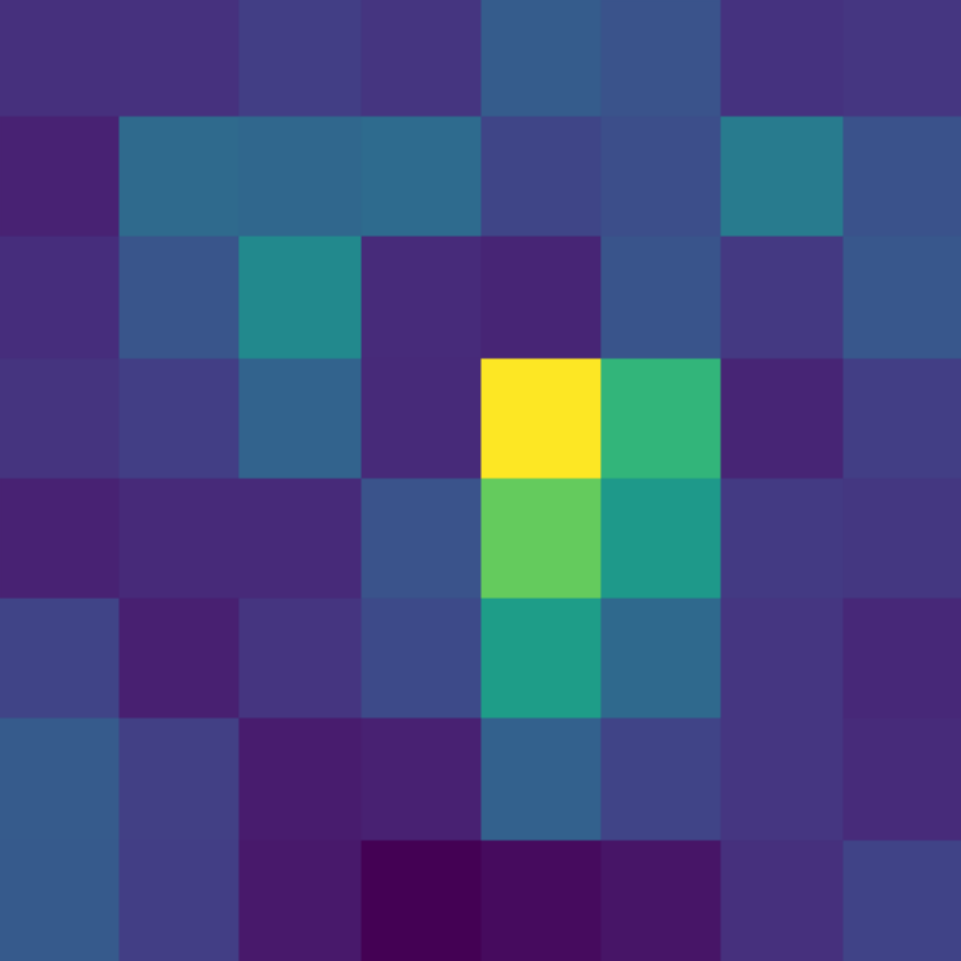}
\end{subfigure}
\begin{subfigure}{.05\textwidth}
  \centering
  \includegraphics[width=1.0\linewidth]{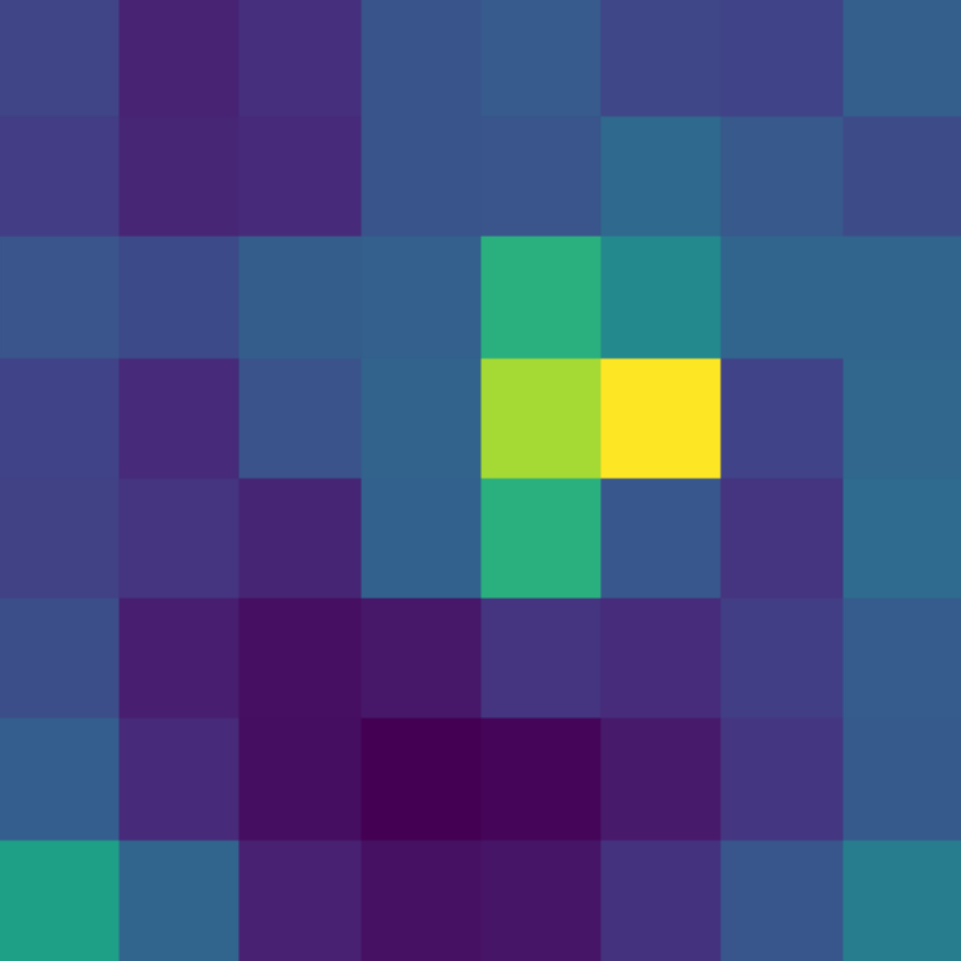}
\end{subfigure}
\begin{subfigure}{.05\textwidth}
  \centering
  \includegraphics[width=1.0\linewidth]{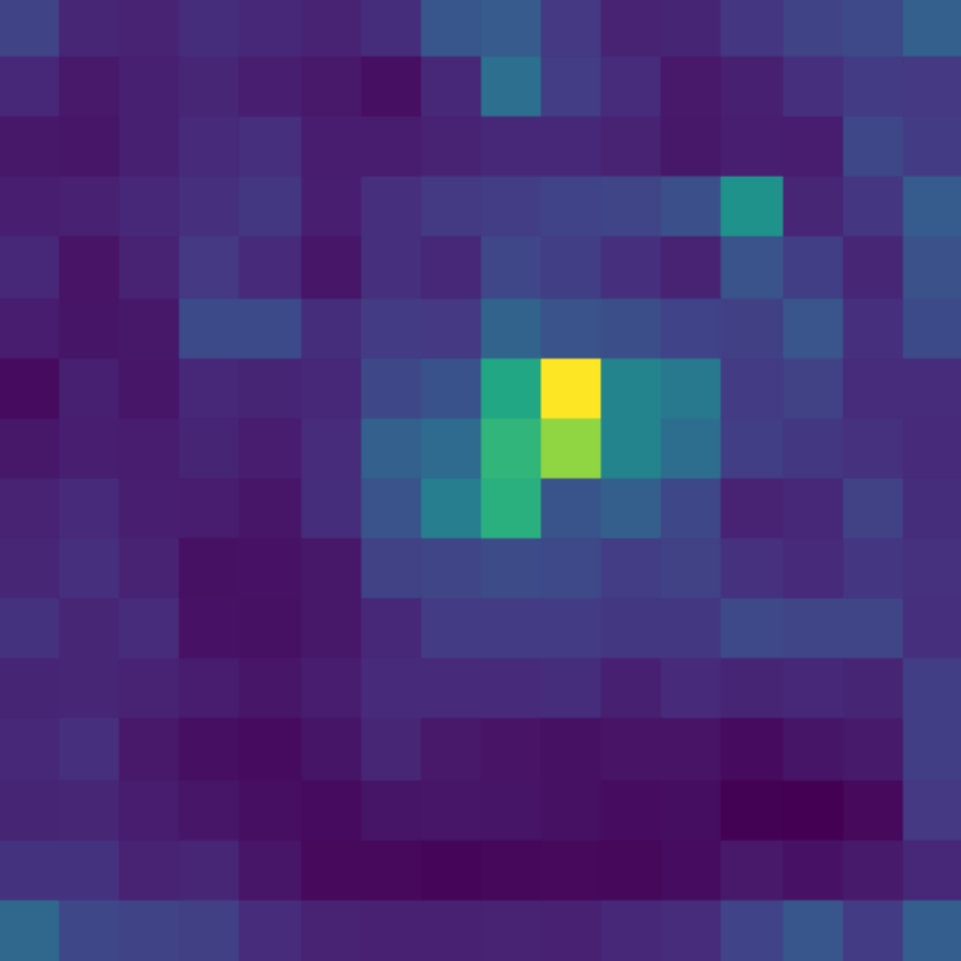}
\end{subfigure}
\begin{subfigure}{.05\textwidth}
  \centering
  \includegraphics[width=1.0\linewidth]{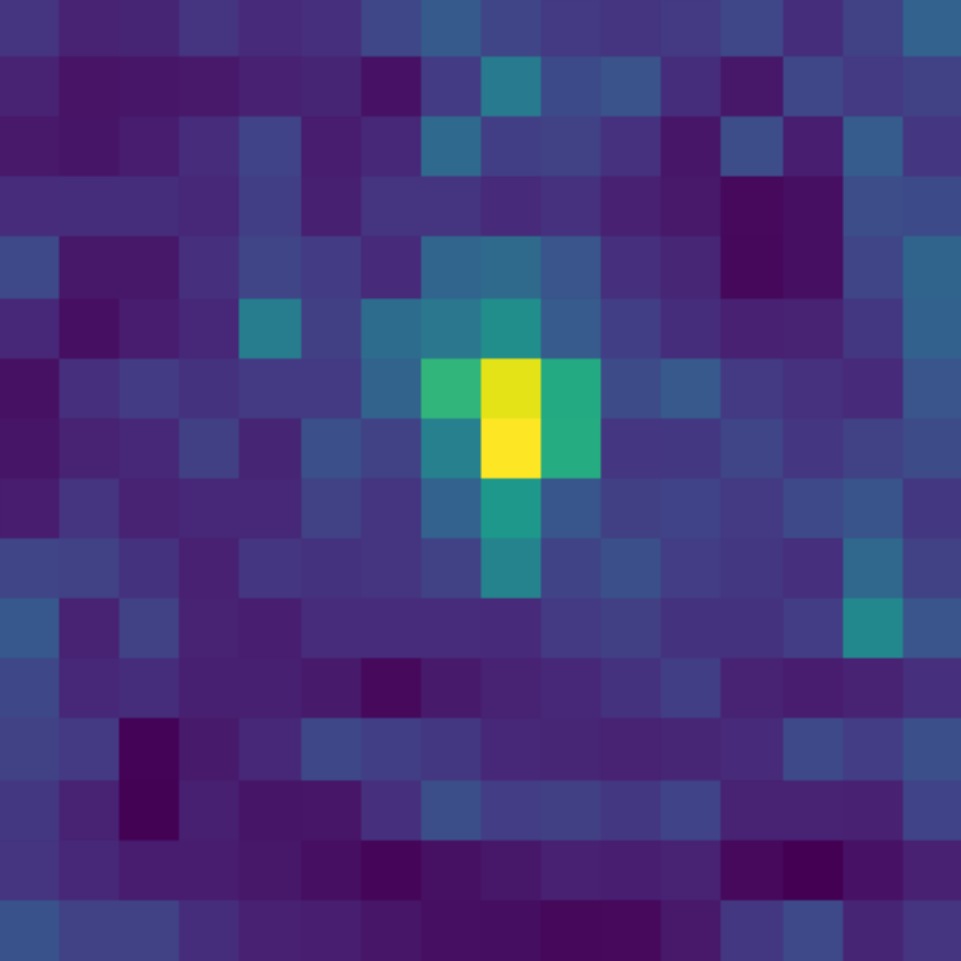}
\end{subfigure}
\begin{subfigure}{.05\textwidth}
  \centering
  \includegraphics[width=1.0\linewidth]{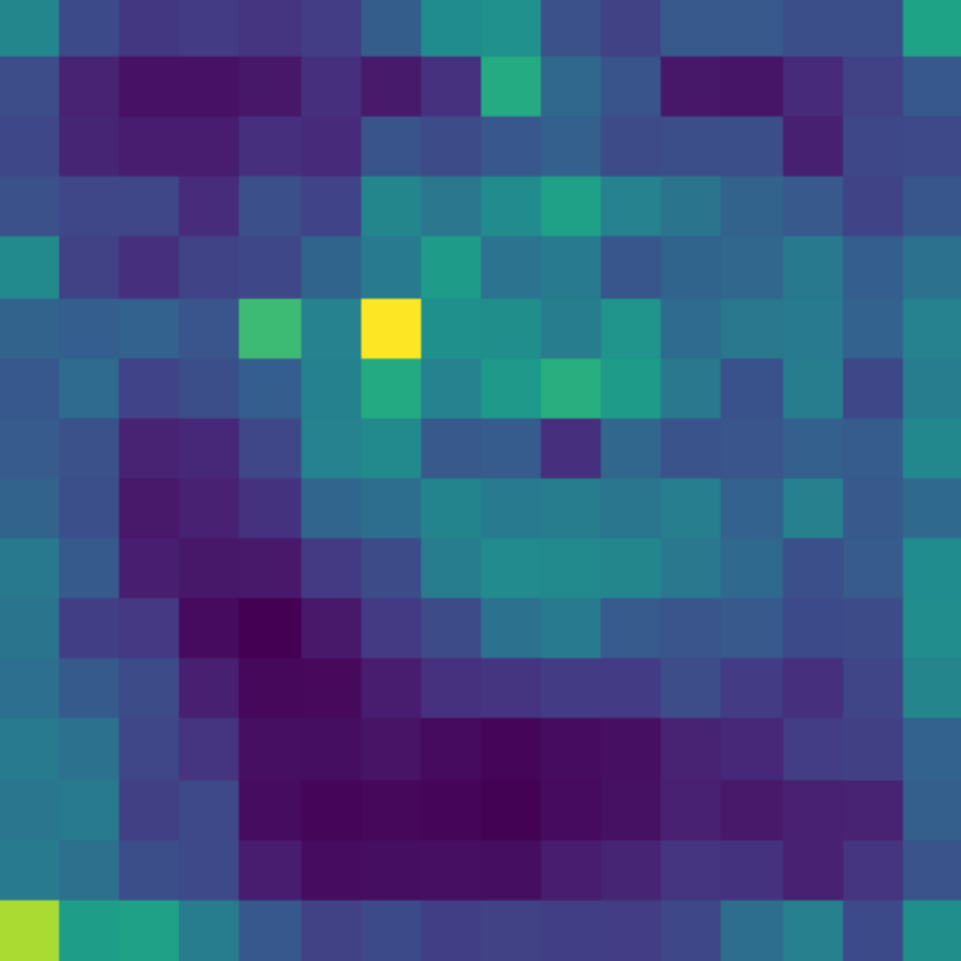}
\end{subfigure}
\begin{subfigure}{.05\textwidth}
  \centering
  \includegraphics[width=1.0\linewidth]{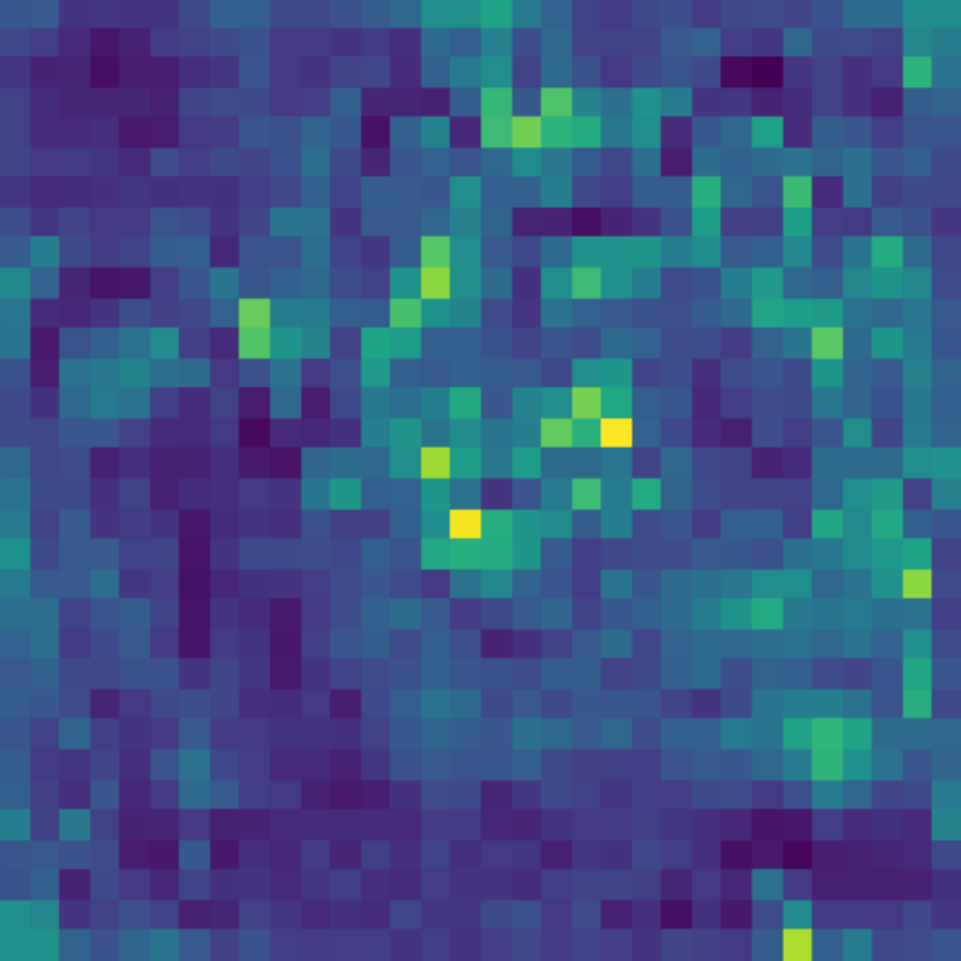}
\end{subfigure}
\begin{subfigure}{.05\textwidth}
  \centering
  \includegraphics[width=1.0\linewidth]{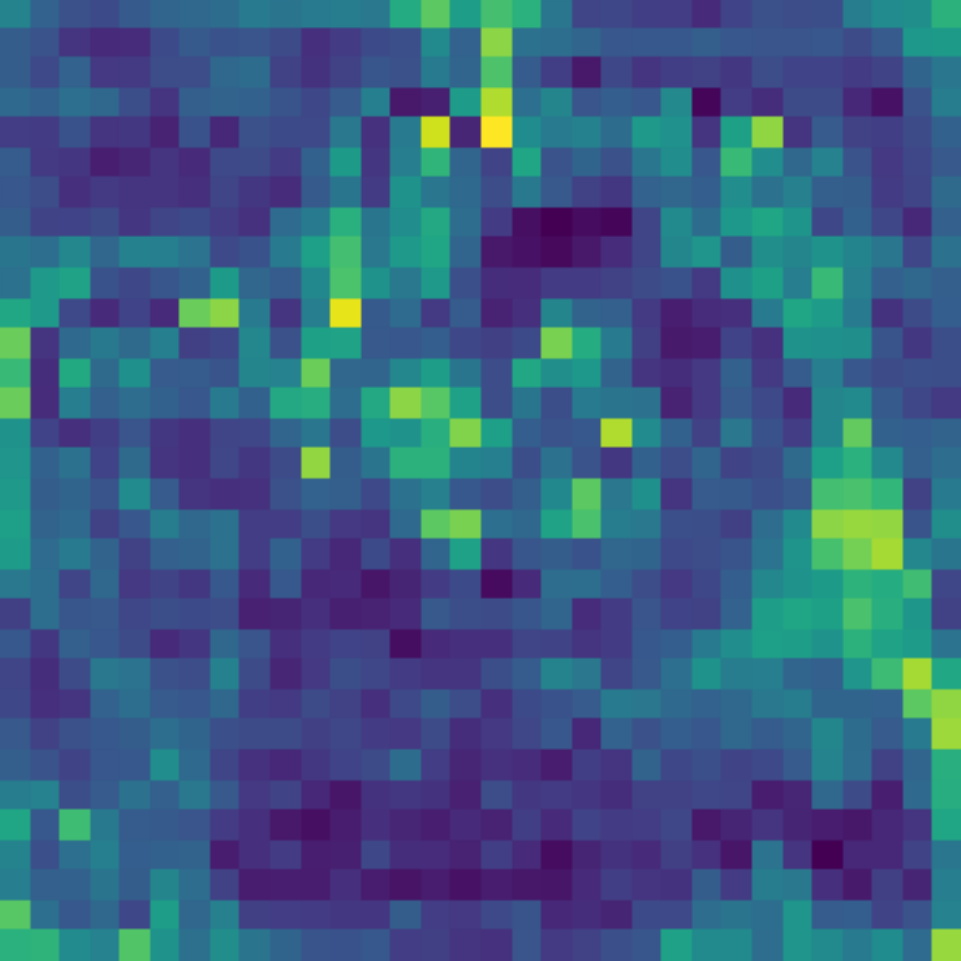}
\end{subfigure}
\begin{subfigure}{.05\textwidth}
  \centering
  \includegraphics[width=1.0\linewidth]{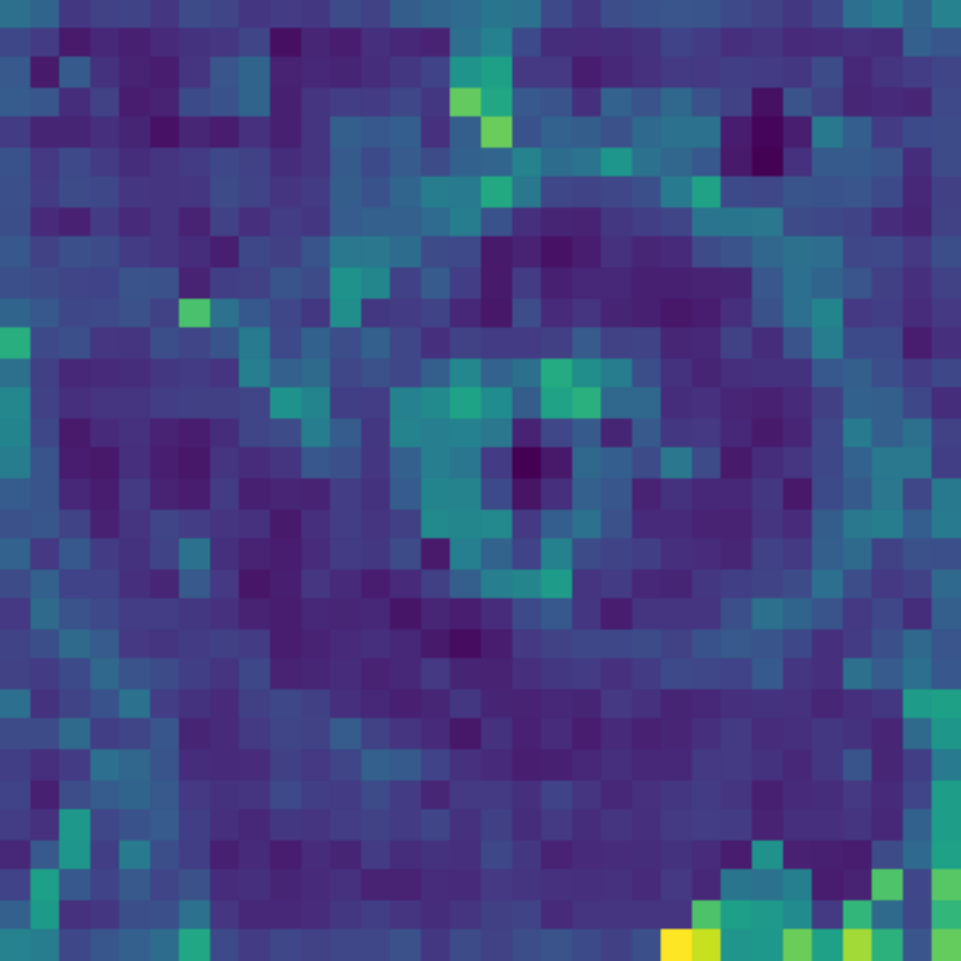}
\end{subfigure}
\\
&
\begin{subfigure}{.05\textwidth}
  \centering
  \includegraphics[width=1.0\linewidth]{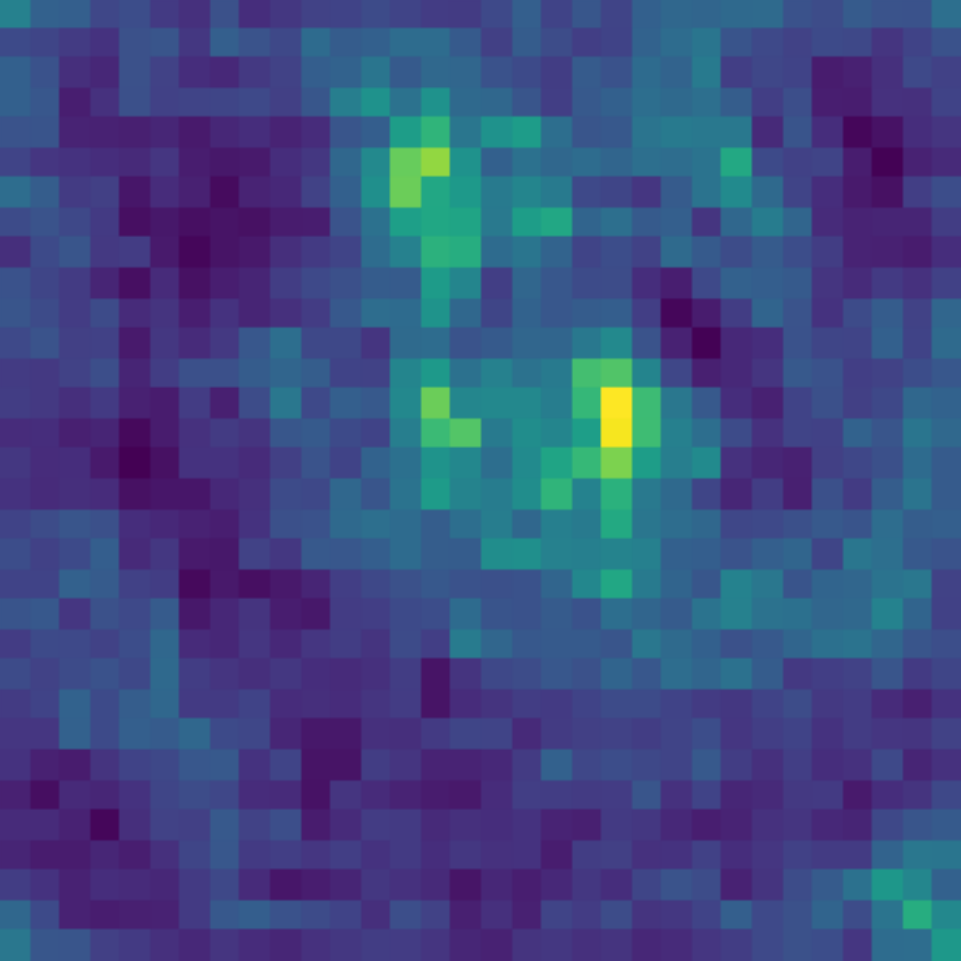}
\end{subfigure}
\begin{subfigure}{.05\textwidth}
  \centering
  \includegraphics[width=1.0\linewidth]{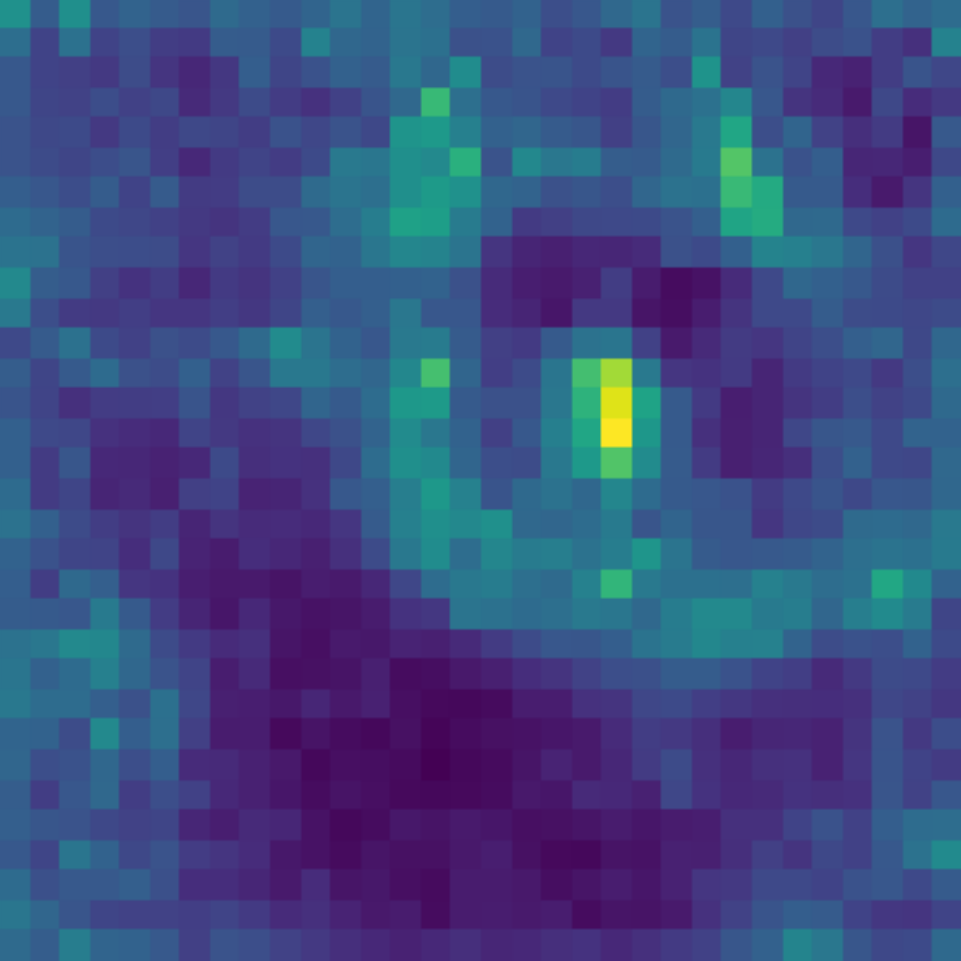}
\end{subfigure}
\begin{subfigure}{.05\textwidth}
  \centering
  \includegraphics[width=1.0\linewidth]{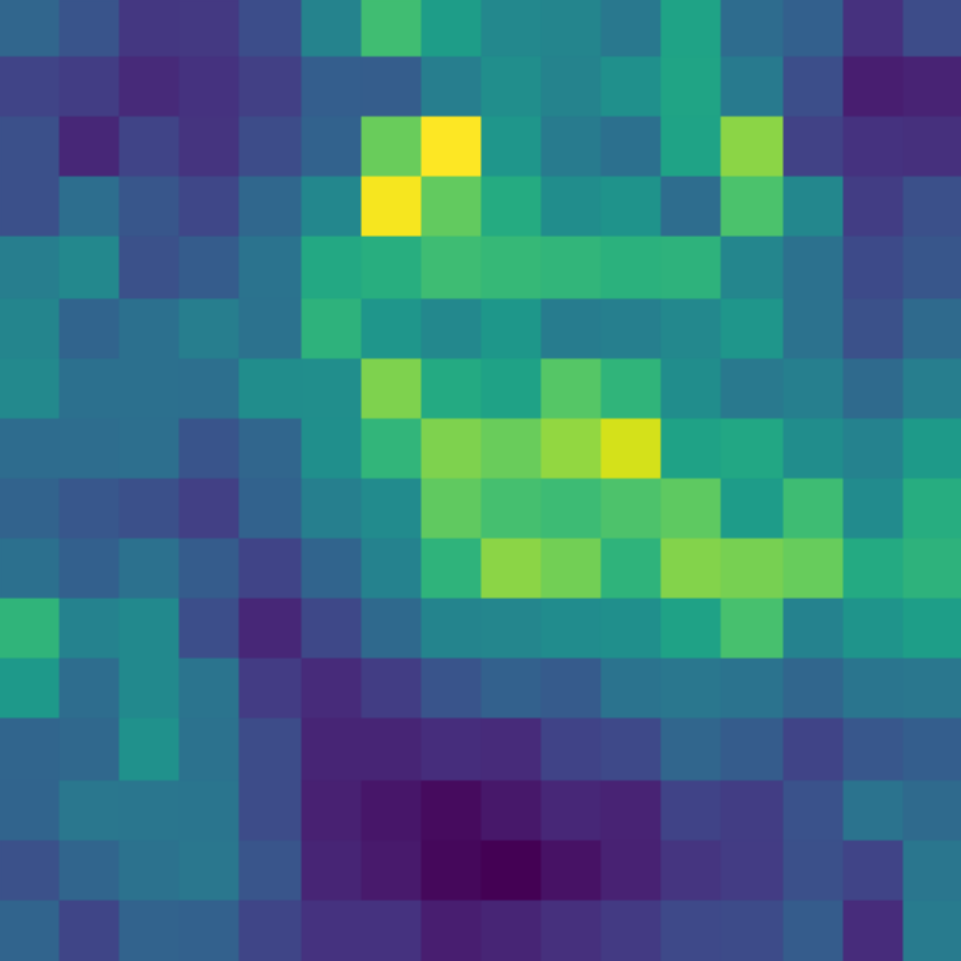}
\end{subfigure}
\begin{subfigure}{.05\textwidth}
  \centering
  \includegraphics[width=1.0\linewidth]{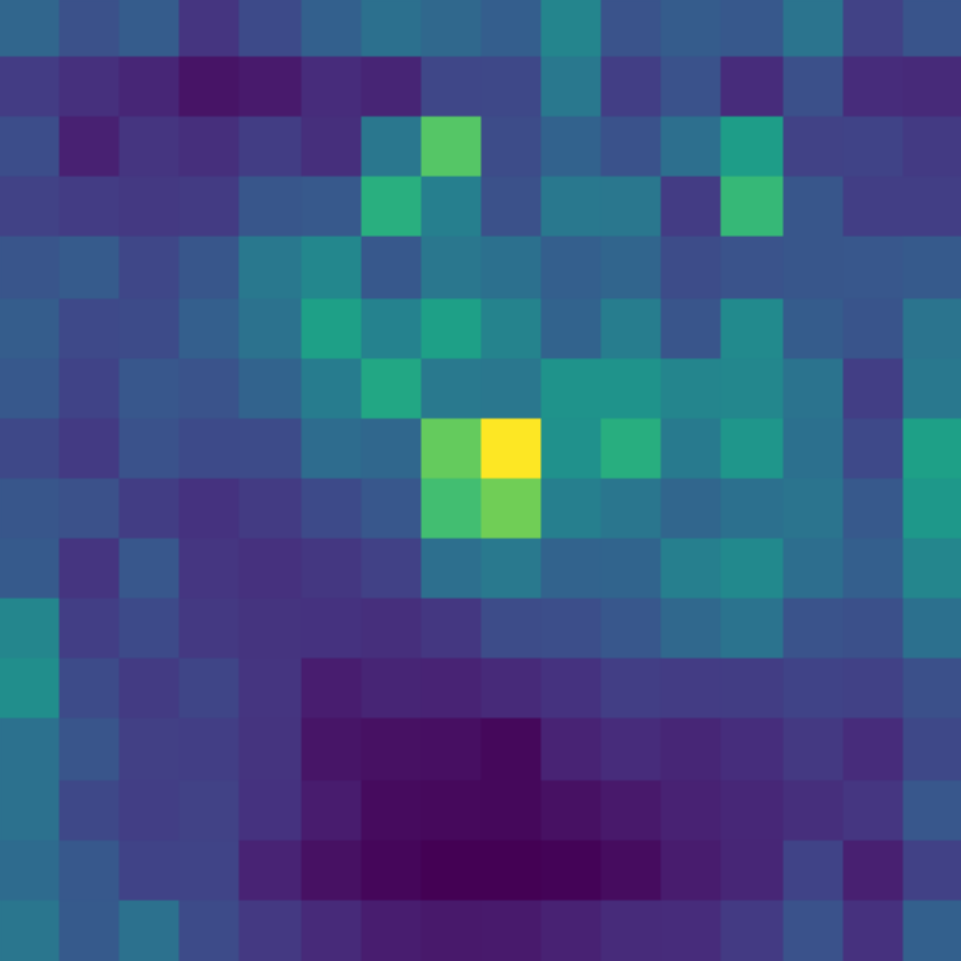}
\end{subfigure}
\begin{subfigure}{.05\textwidth}
  \centering
  \includegraphics[width=1.0\linewidth]{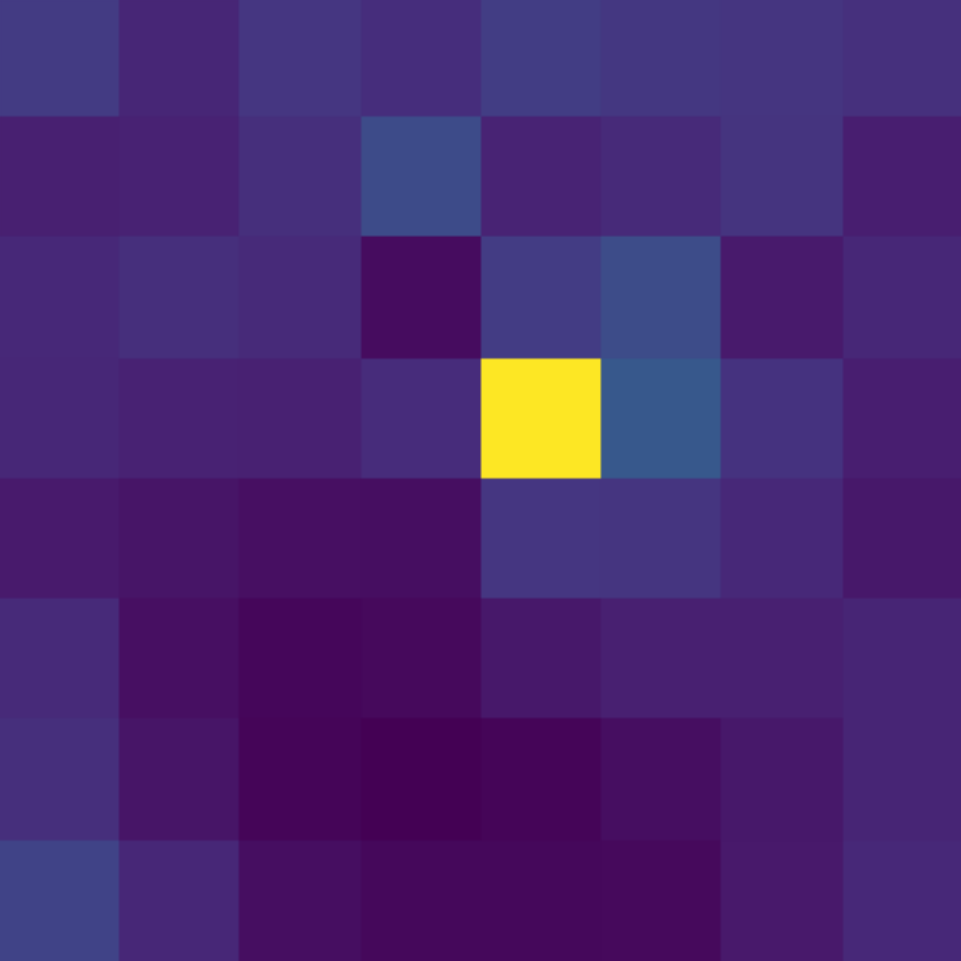}
\end{subfigure}
\begin{subfigure}{.05\textwidth}
  \centering
  \includegraphics[width=1.0\linewidth]{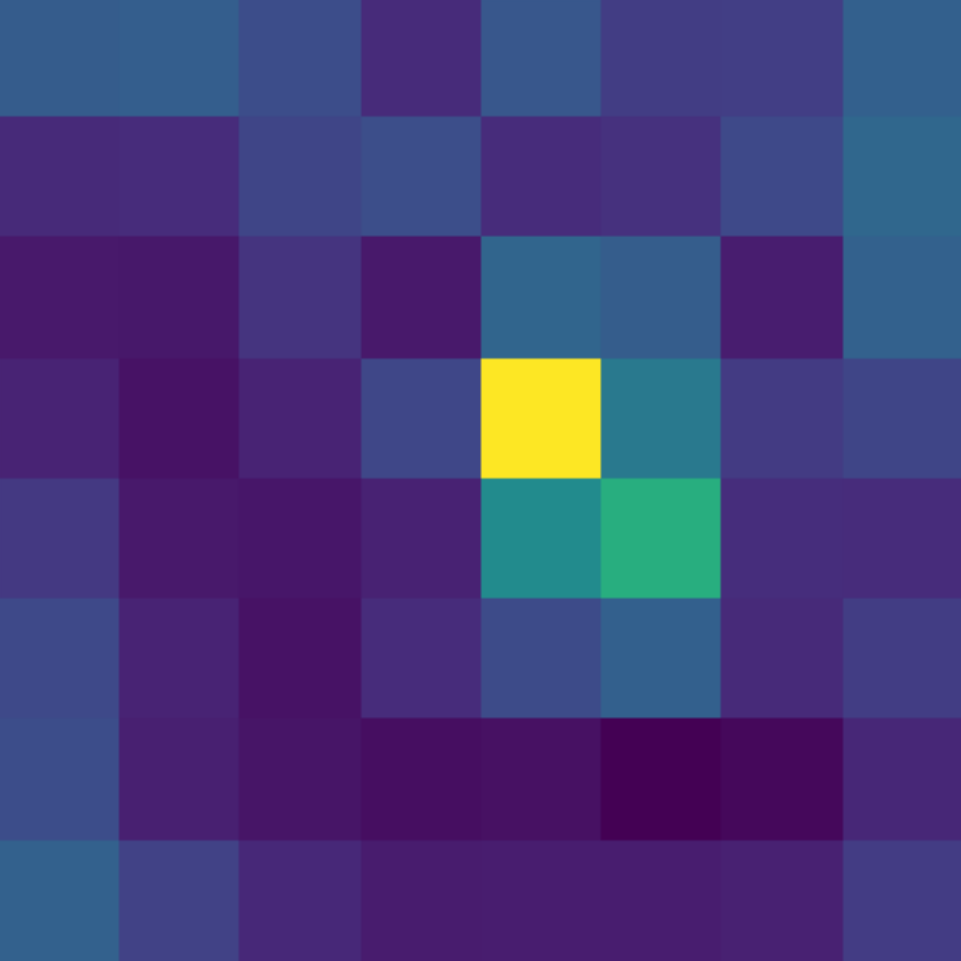}
\end{subfigure}
\begin{subfigure}{.05\textwidth}
  \centering
  \includegraphics[width=1.0\linewidth]{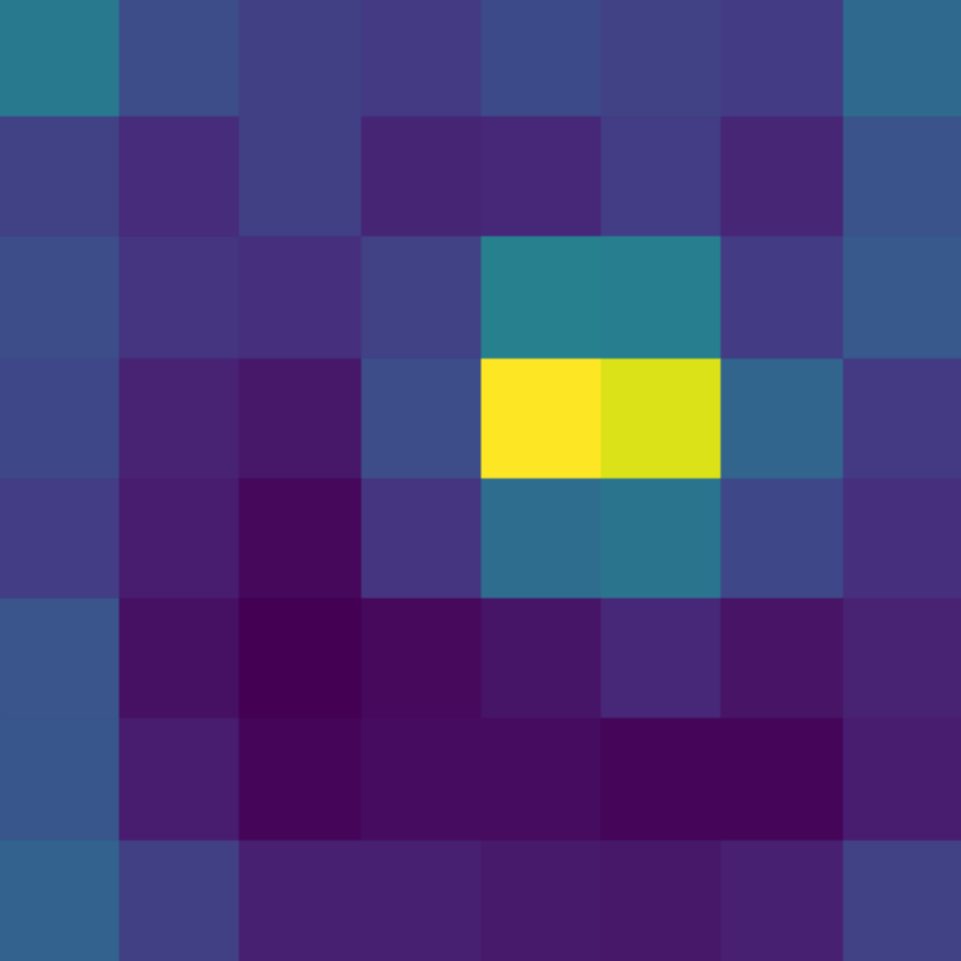}
\end{subfigure}
\begin{subfigure}{.05\textwidth}
  \centering
  \includegraphics[width=1.0\linewidth]{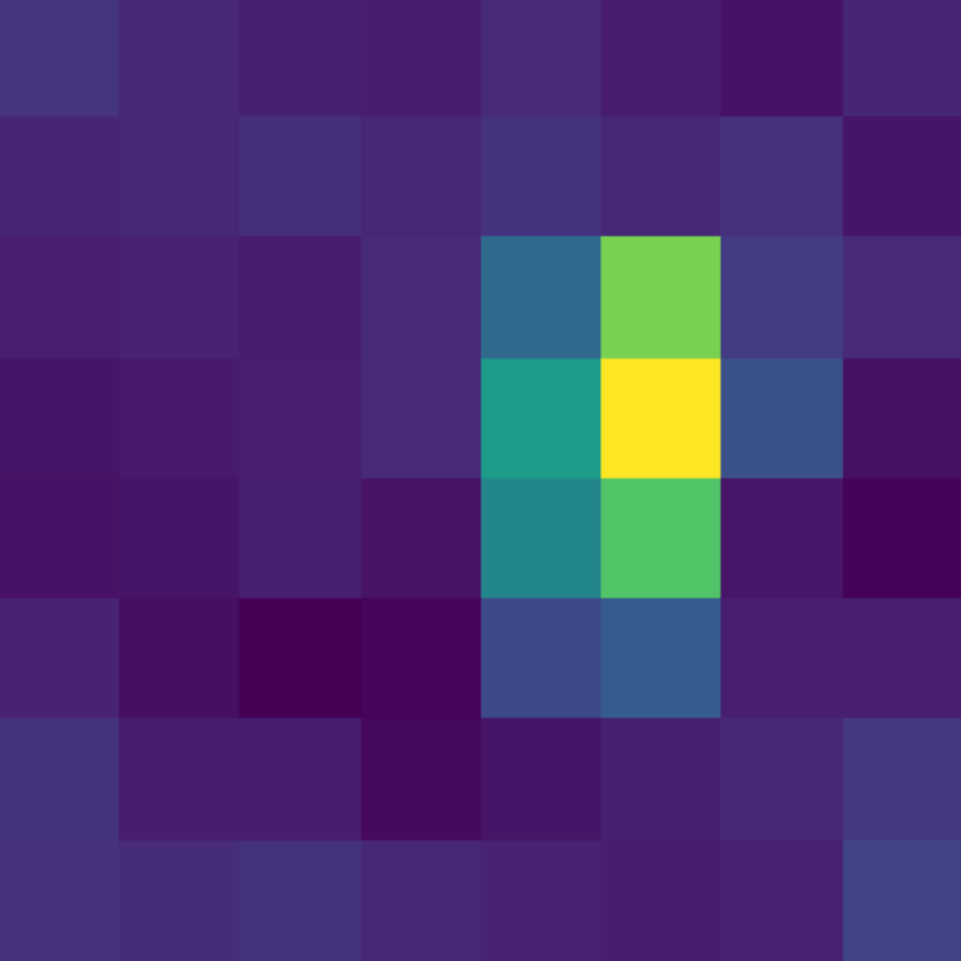}
\end{subfigure}
\begin{subfigure}{.05\textwidth}
  \centering
  \includegraphics[width=1.0\linewidth]{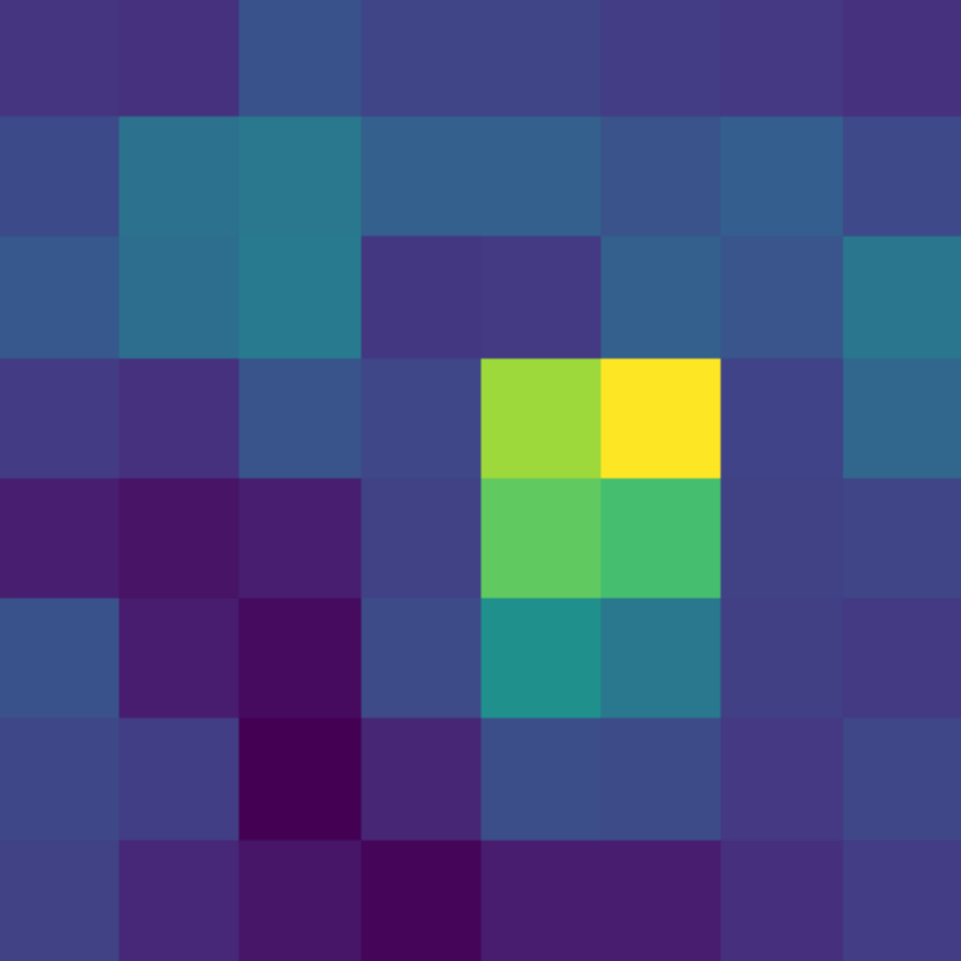}
\end{subfigure}
\begin{subfigure}{.05\textwidth}
  \centering
  \includegraphics[width=1.0\linewidth]{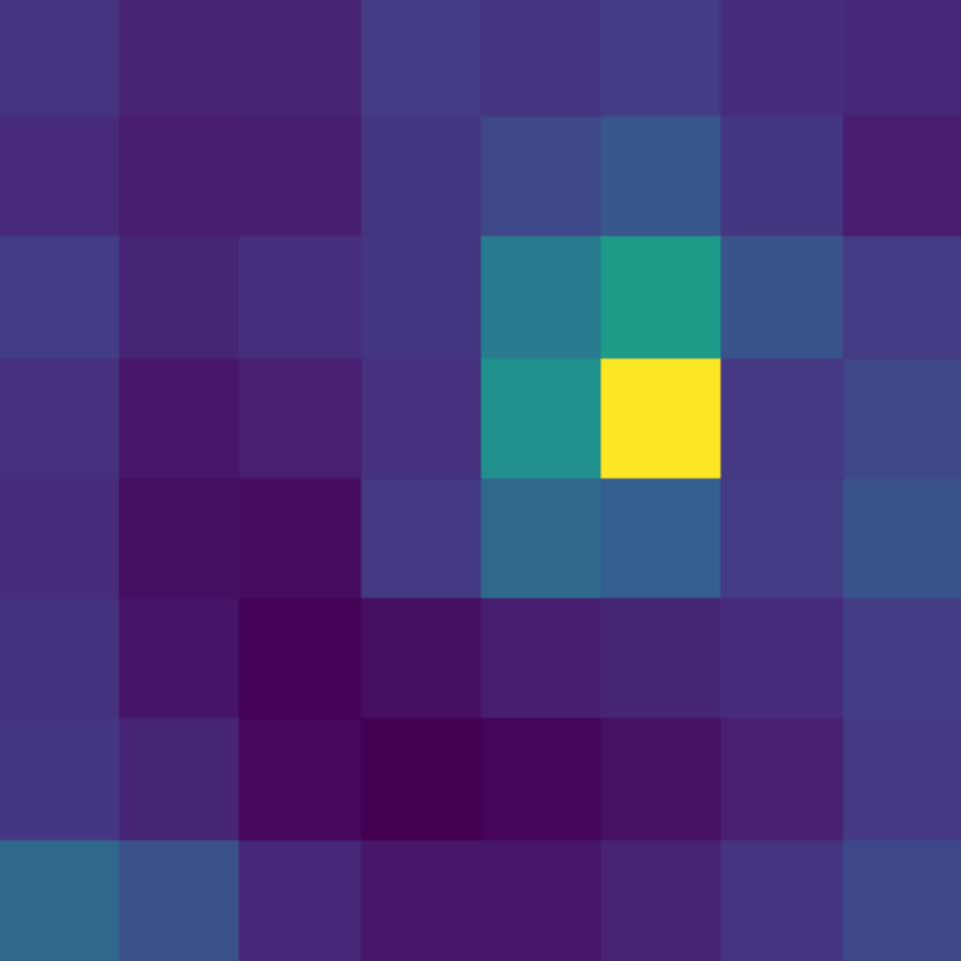}
\end{subfigure}
\begin{subfigure}{.05\textwidth}
  \centering
  \includegraphics[width=1.0\linewidth]{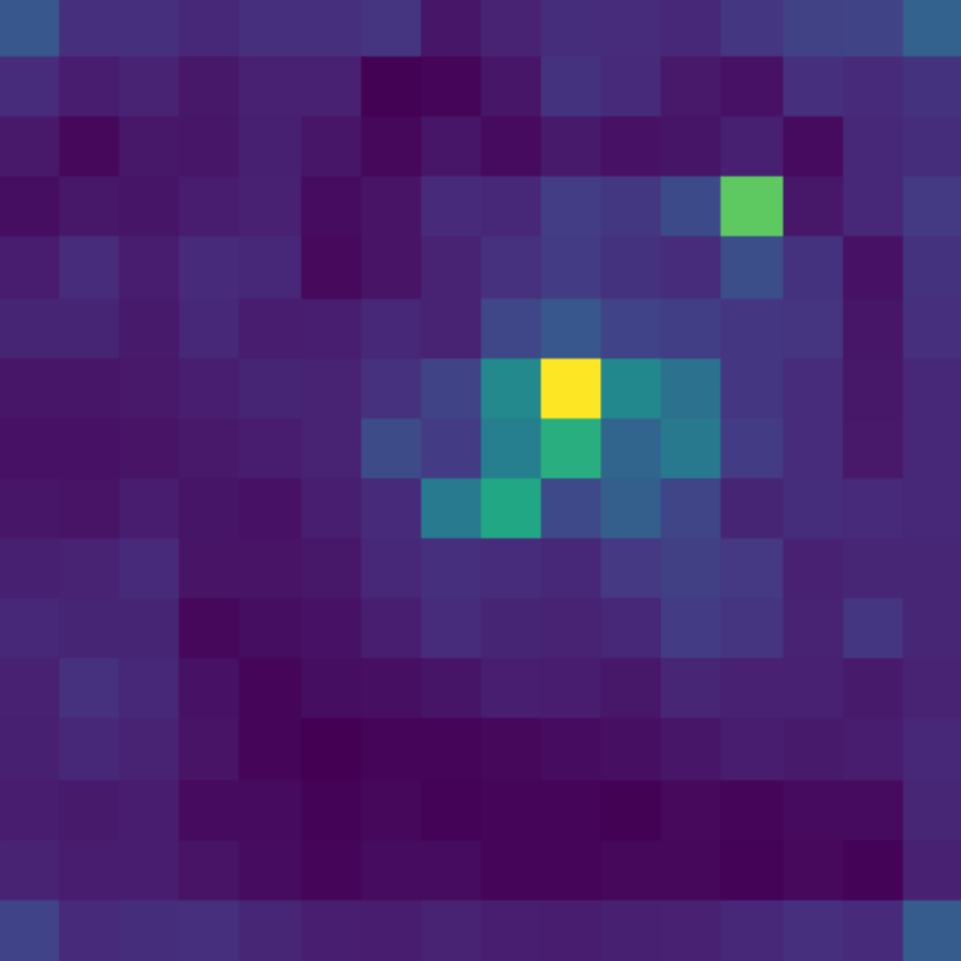}
\end{subfigure}
\begin{subfigure}{.05\textwidth}
  \centering
  \includegraphics[width=1.0\linewidth]{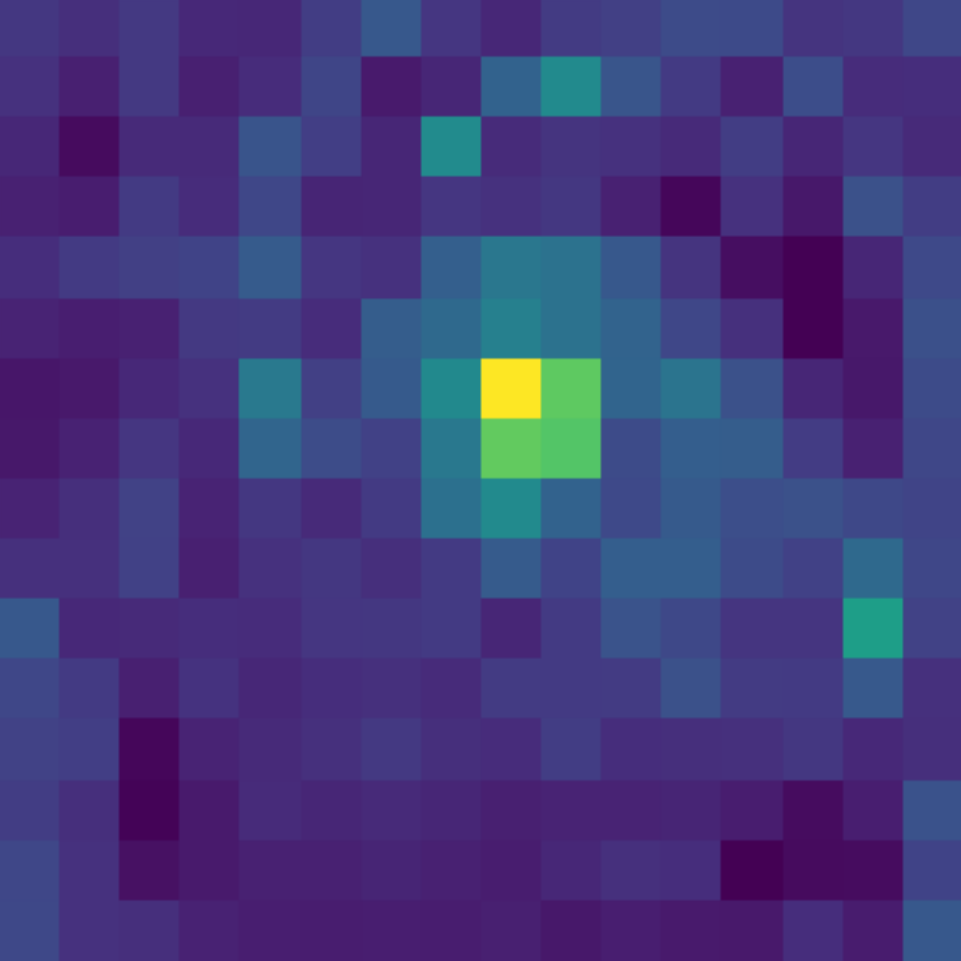}
\end{subfigure}
\begin{subfigure}{.05\textwidth}
  \centering
  \includegraphics[width=1.0\linewidth]{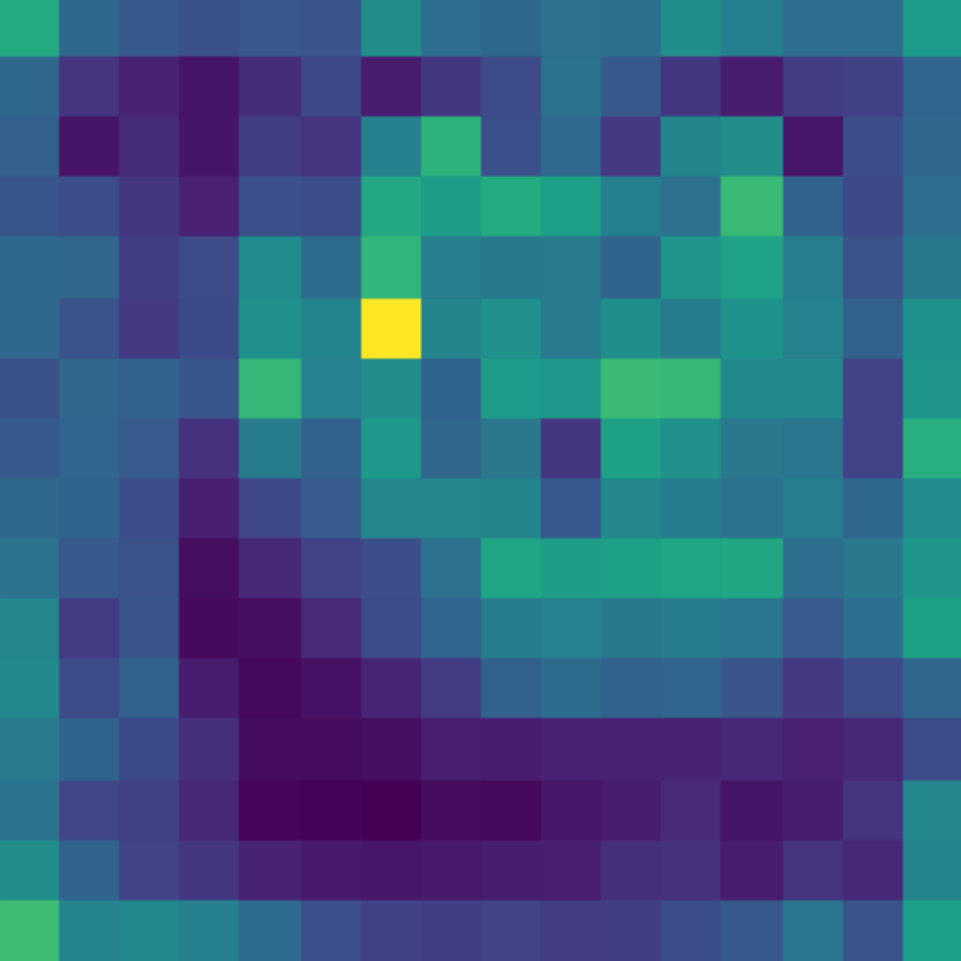}
\end{subfigure}
\begin{subfigure}{.05\textwidth}
  \centering
  \includegraphics[width=1.0\linewidth]{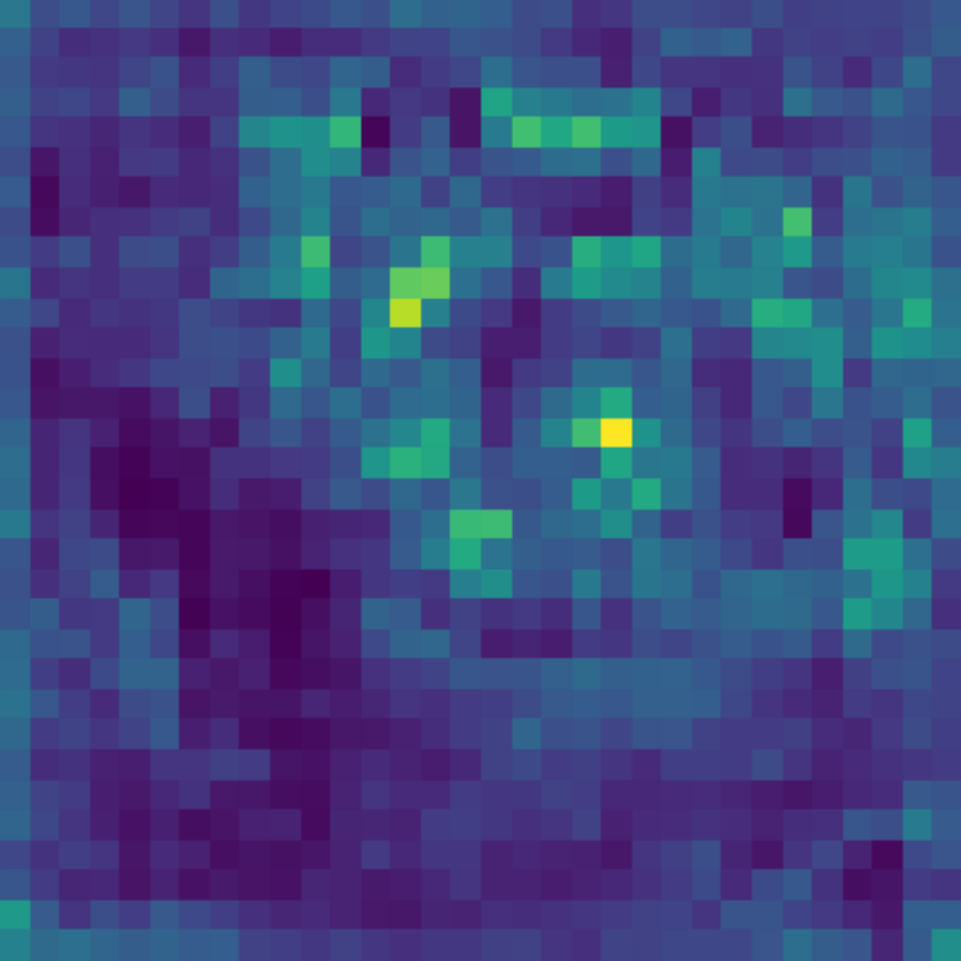}
\end{subfigure}
\begin{subfigure}{.05\textwidth}
  \centering
  \includegraphics[width=1.0\linewidth]{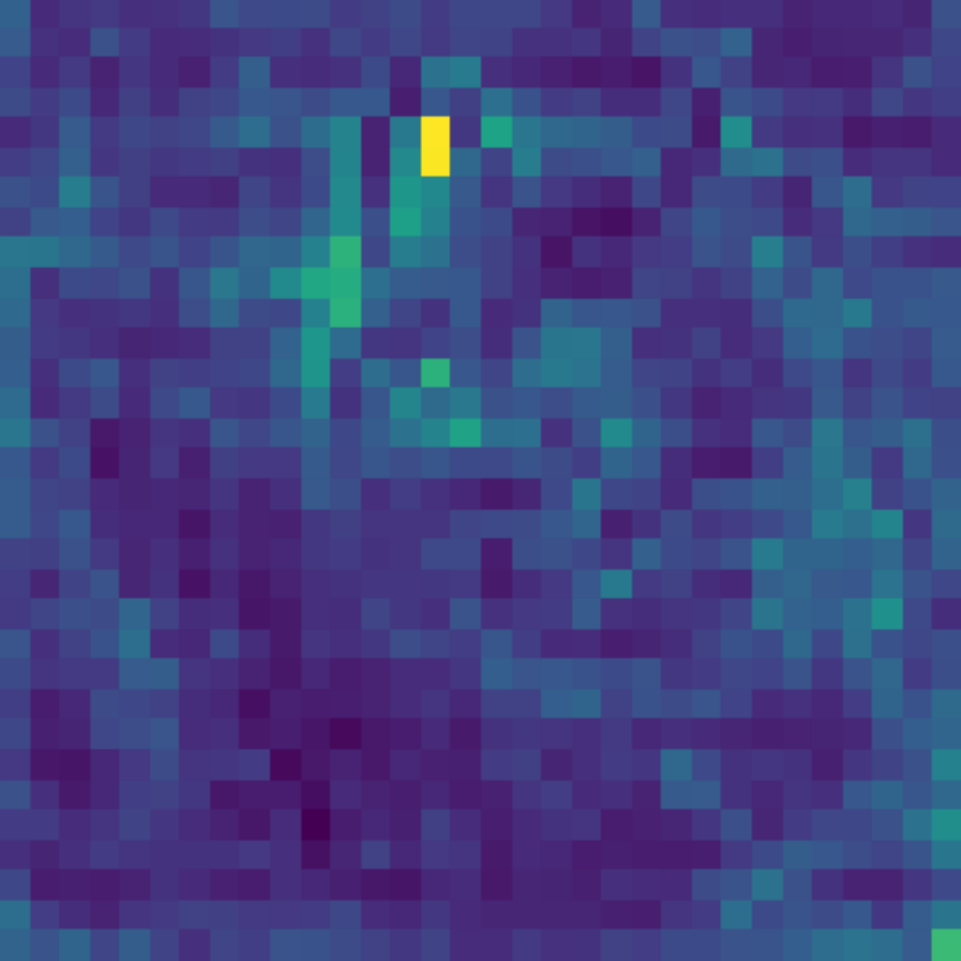}
\end{subfigure}
\begin{subfigure}{.05\textwidth}
  \centering
  \includegraphics[width=1.0\linewidth]{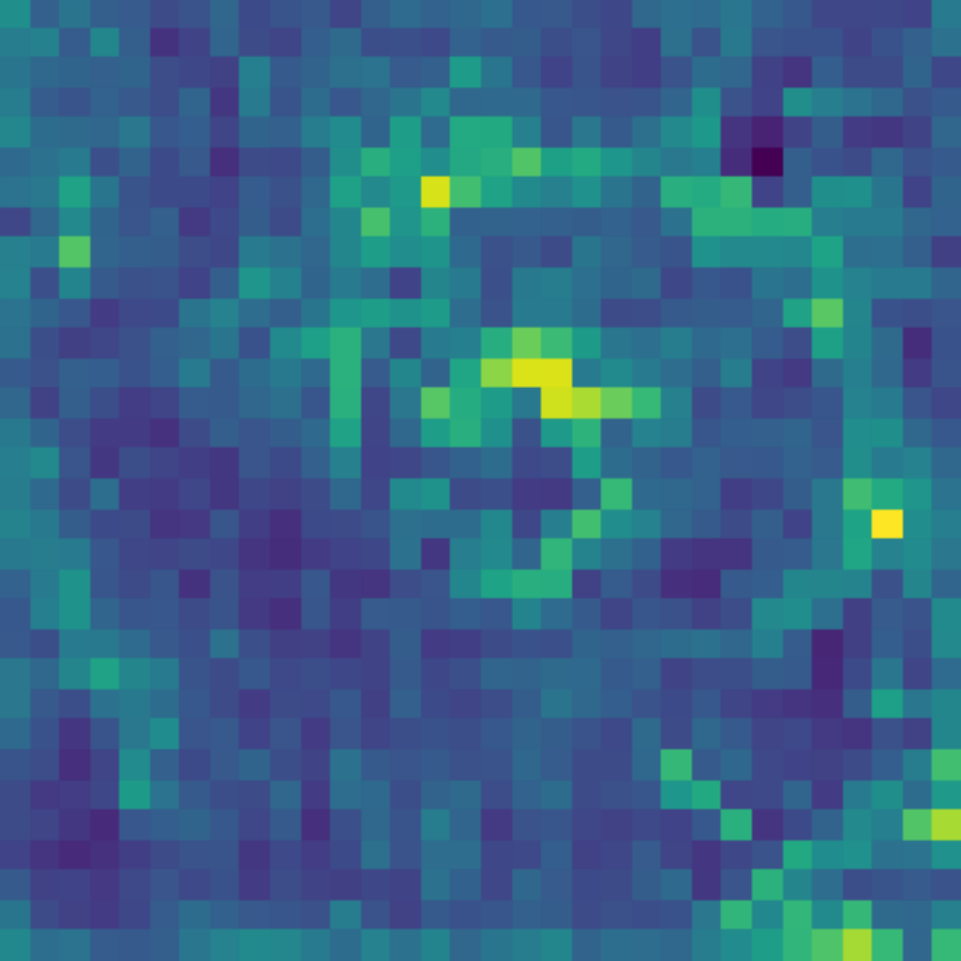}
\end{subfigure}
\\
&
\begin{subfigure}{.05\textwidth}
  \centering
  \includegraphics[width=1.0\linewidth]{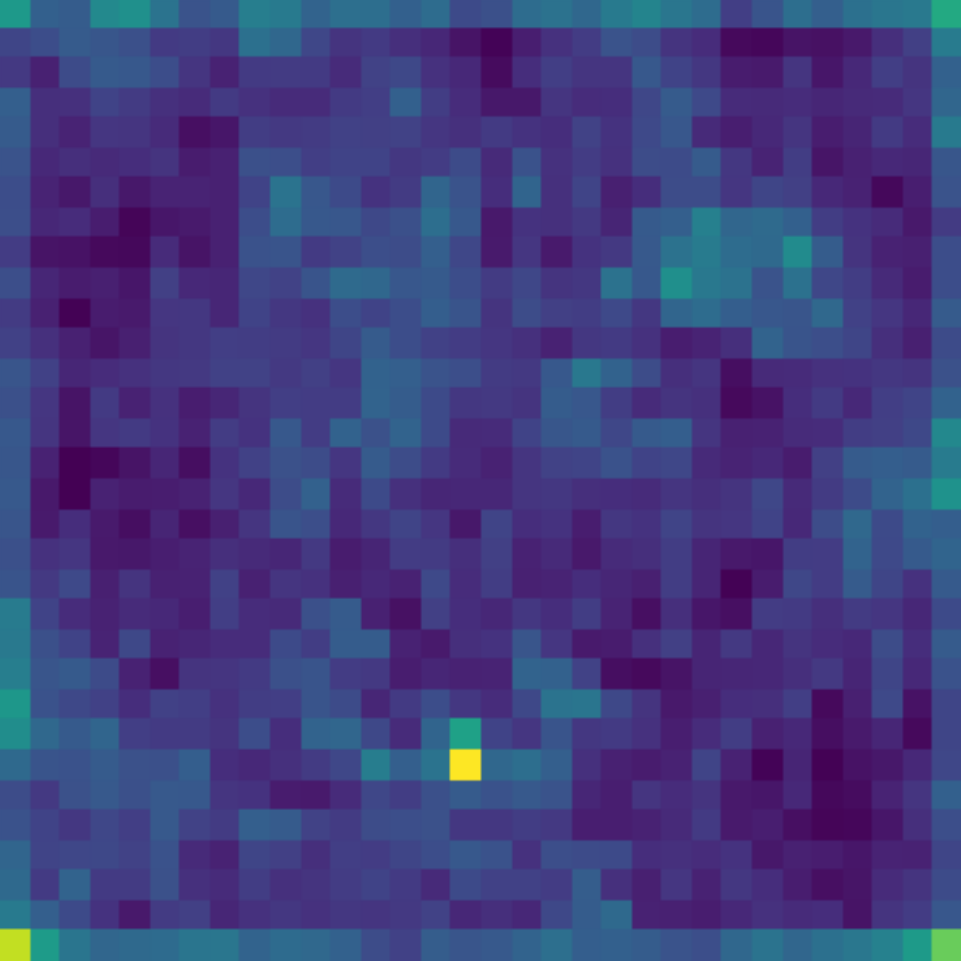}
\end{subfigure}
\begin{subfigure}{.05\textwidth}
  \centering
  \includegraphics[width=1.0\linewidth]{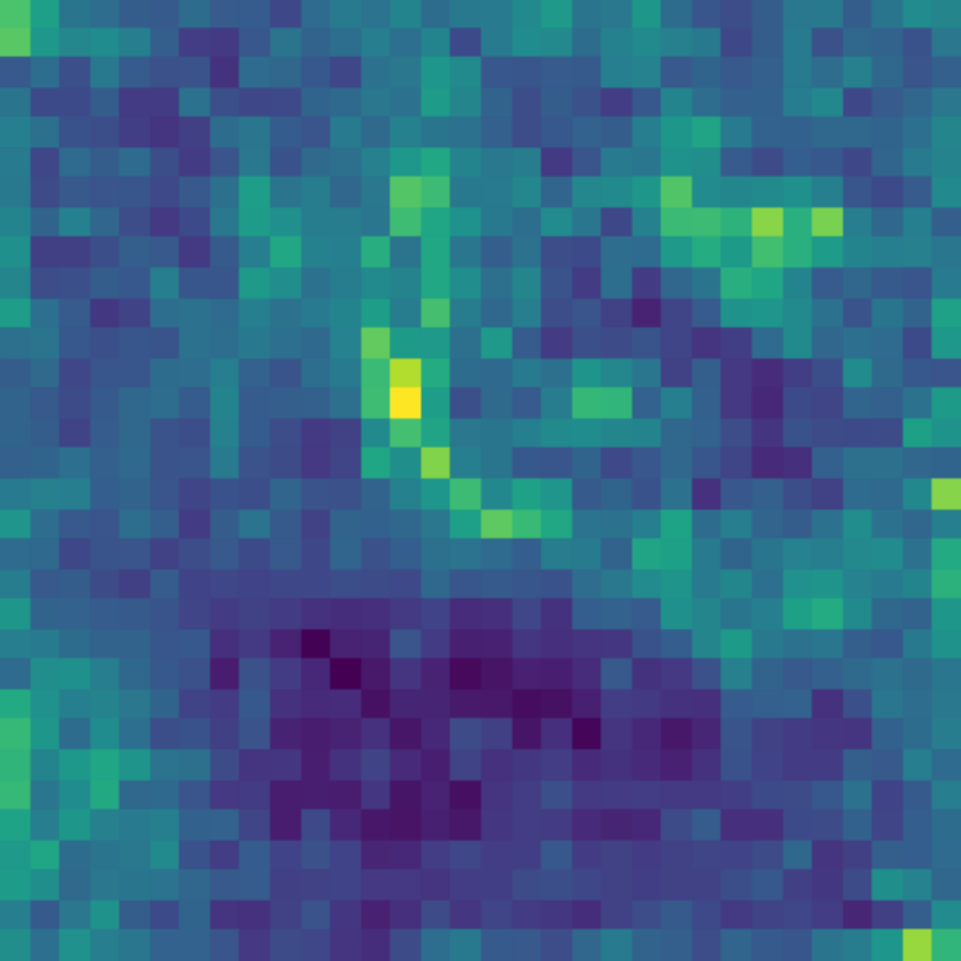}
\end{subfigure}
\begin{subfigure}{.05\textwidth}
  \centering
  \includegraphics[width=1.0\linewidth]{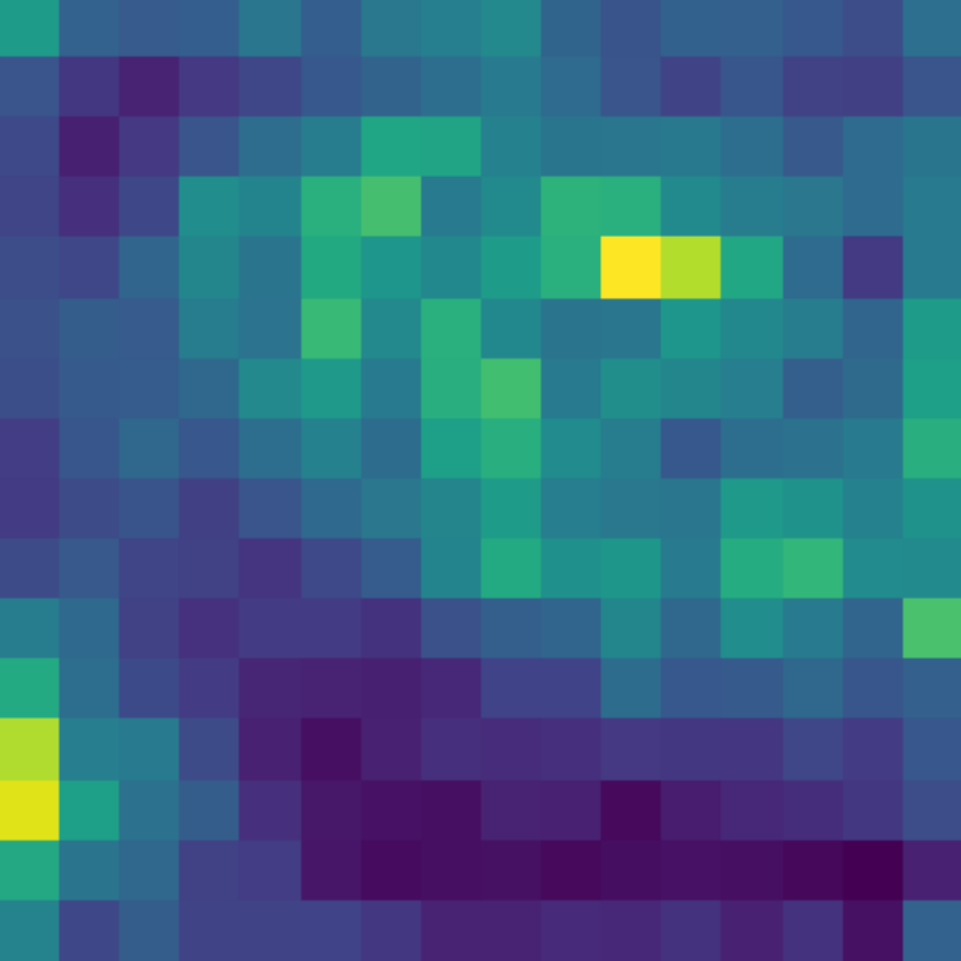}
\end{subfigure}
\begin{subfigure}{.05\textwidth}
  \centering
  \includegraphics[width=1.0\linewidth]{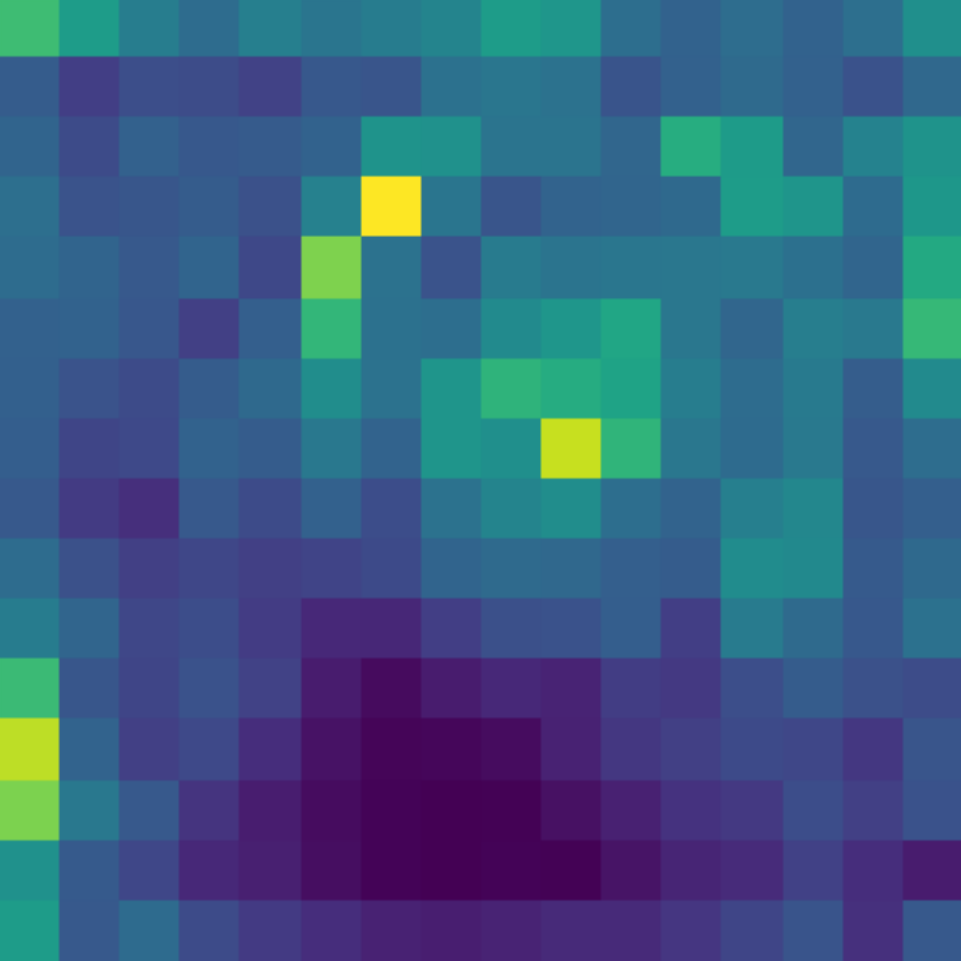}
\end{subfigure}
\begin{subfigure}{.05\textwidth}
  \centering
  \includegraphics[width=1.0\linewidth]{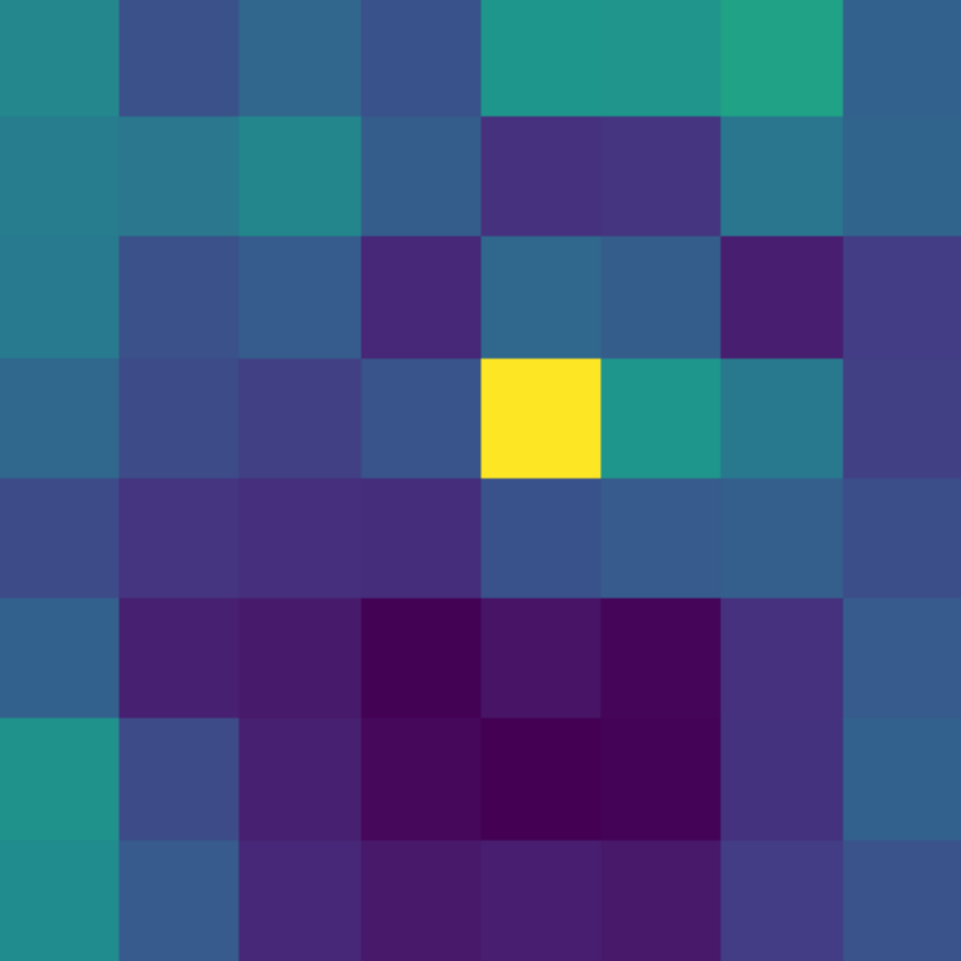}
\end{subfigure}
\begin{subfigure}{.05\textwidth}
  \centering
  \includegraphics[width=1.0\linewidth]{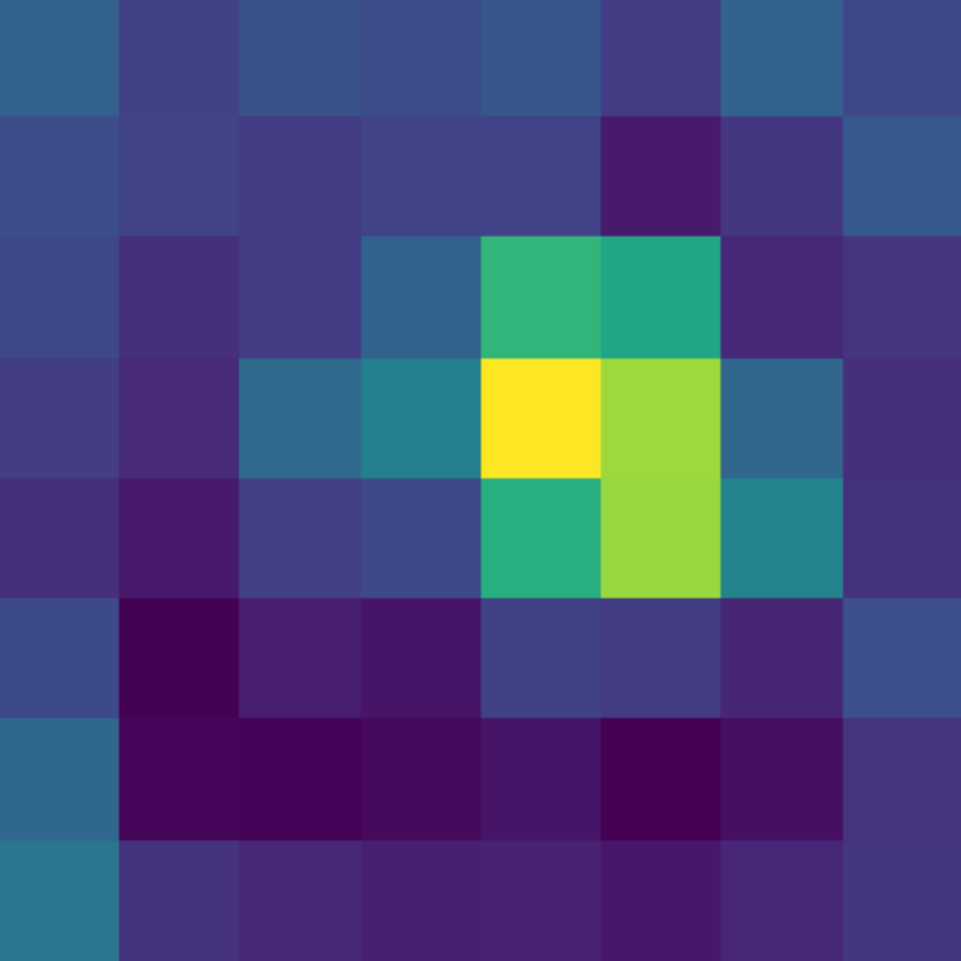}
\end{subfigure}
\begin{subfigure}{.05\textwidth}
  \centering
  \includegraphics[width=1.0\linewidth]{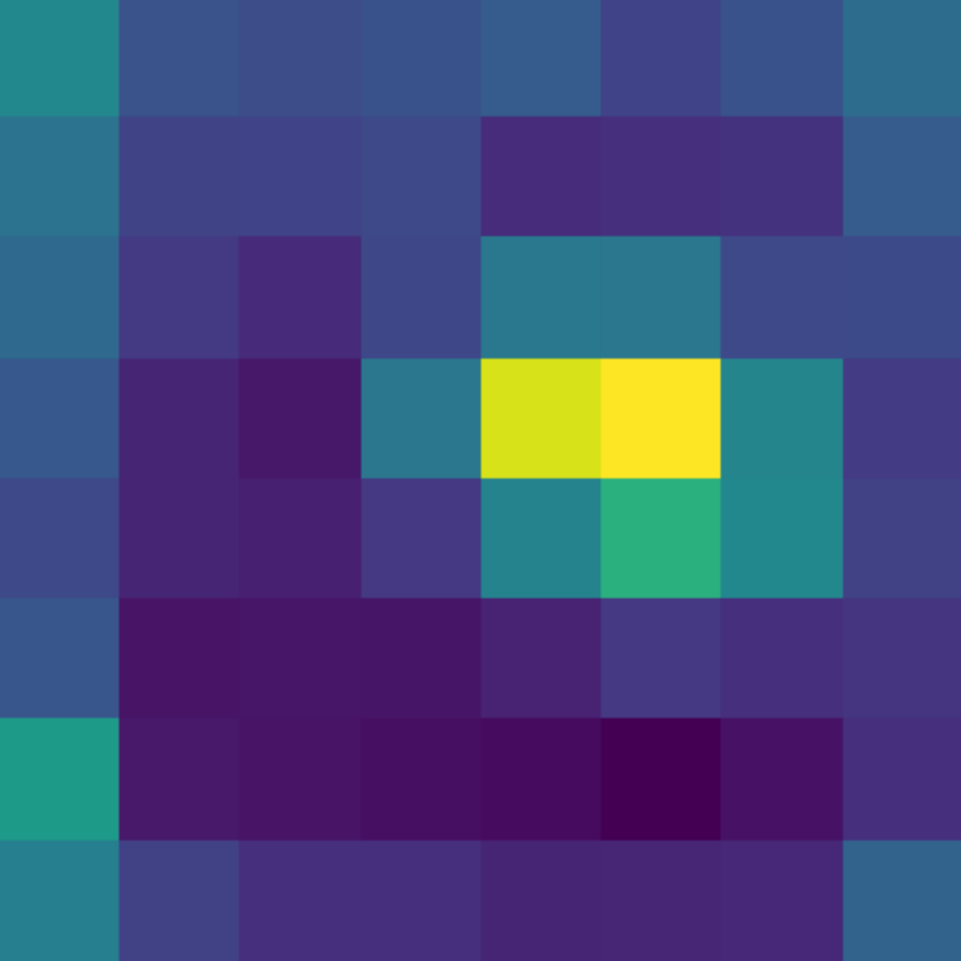}
\end{subfigure}
\begin{subfigure}{.05\textwidth}
  \centering
  \includegraphics[width=1.0\linewidth]{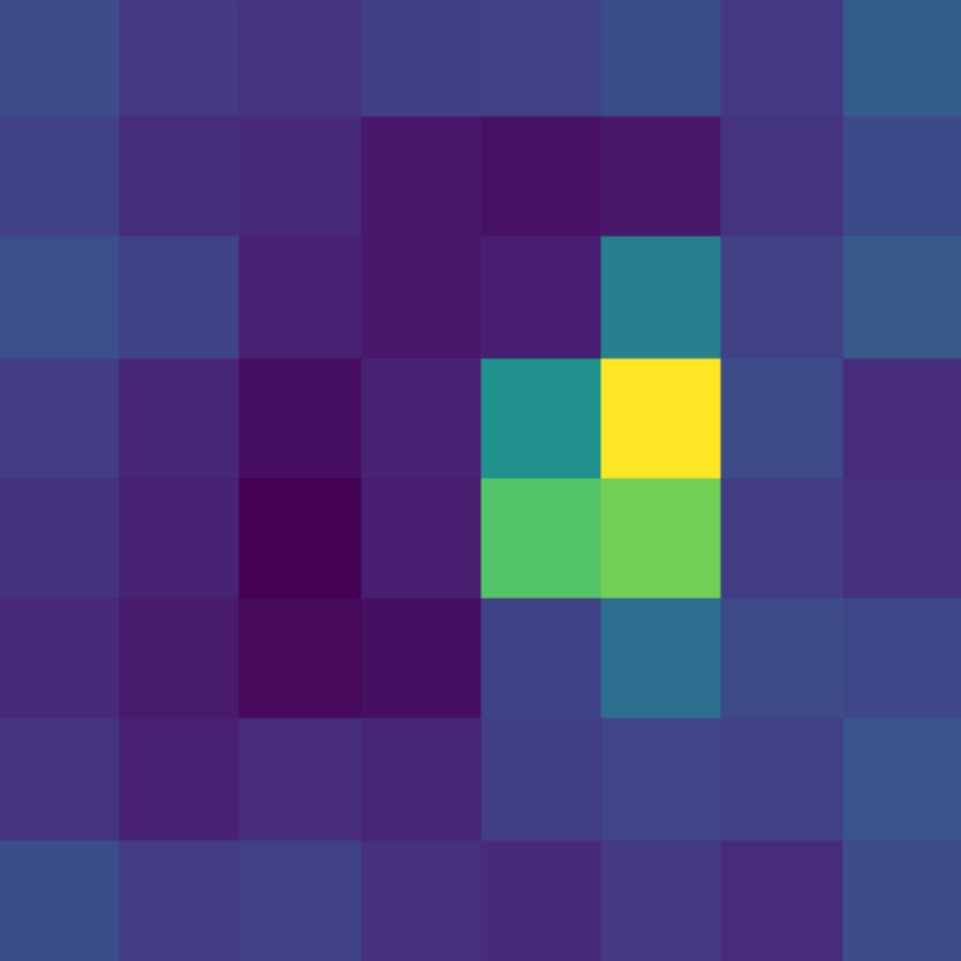}
\end{subfigure}
\begin{subfigure}{.05\textwidth}
  \centering
  \includegraphics[width=1.0\linewidth]{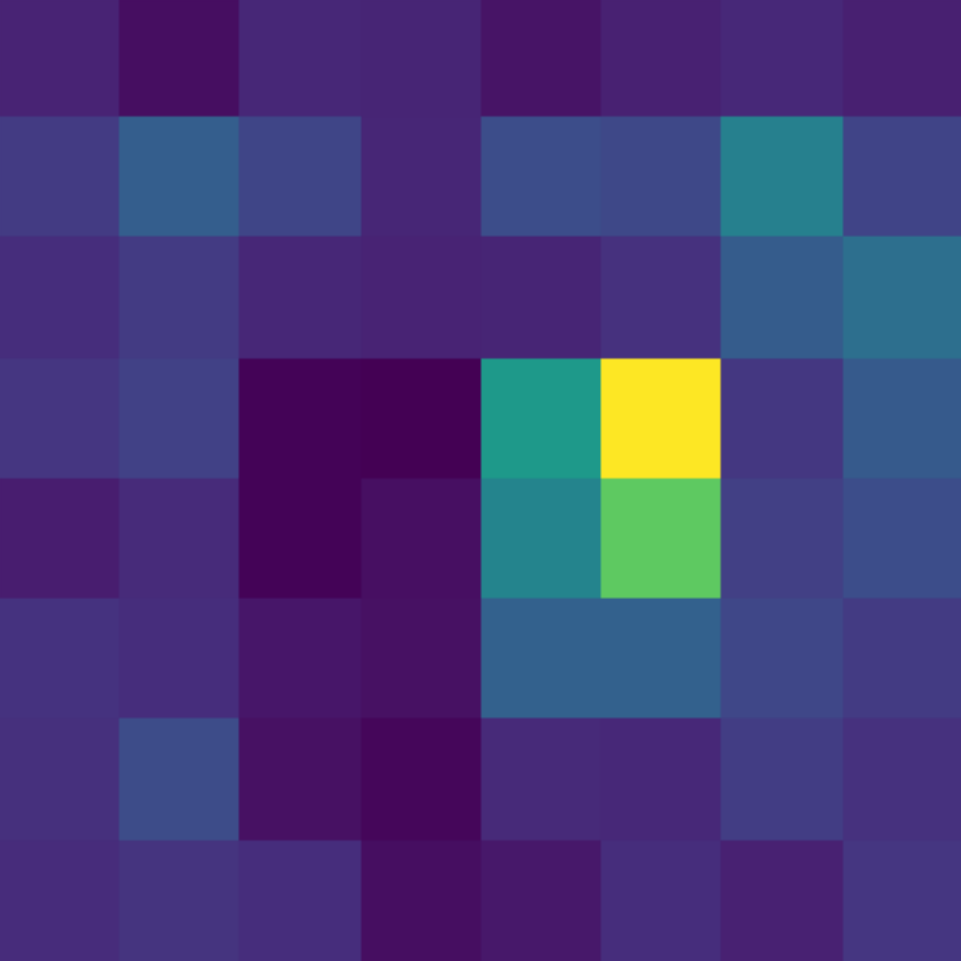}
\end{subfigure}
\begin{subfigure}{.05\textwidth}
  \centering
  \includegraphics[width=1.0\linewidth]{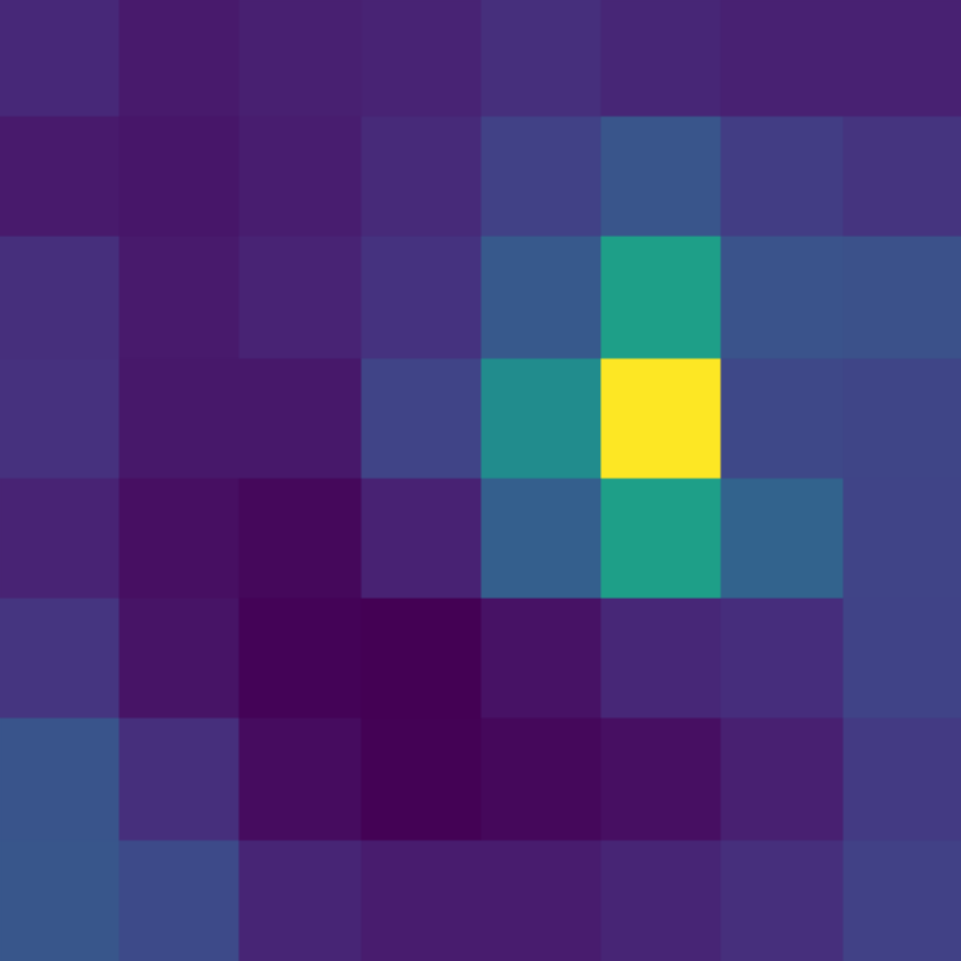}
\end{subfigure}
\begin{subfigure}{.05\textwidth}
  \centering
  \includegraphics[width=1.0\linewidth]{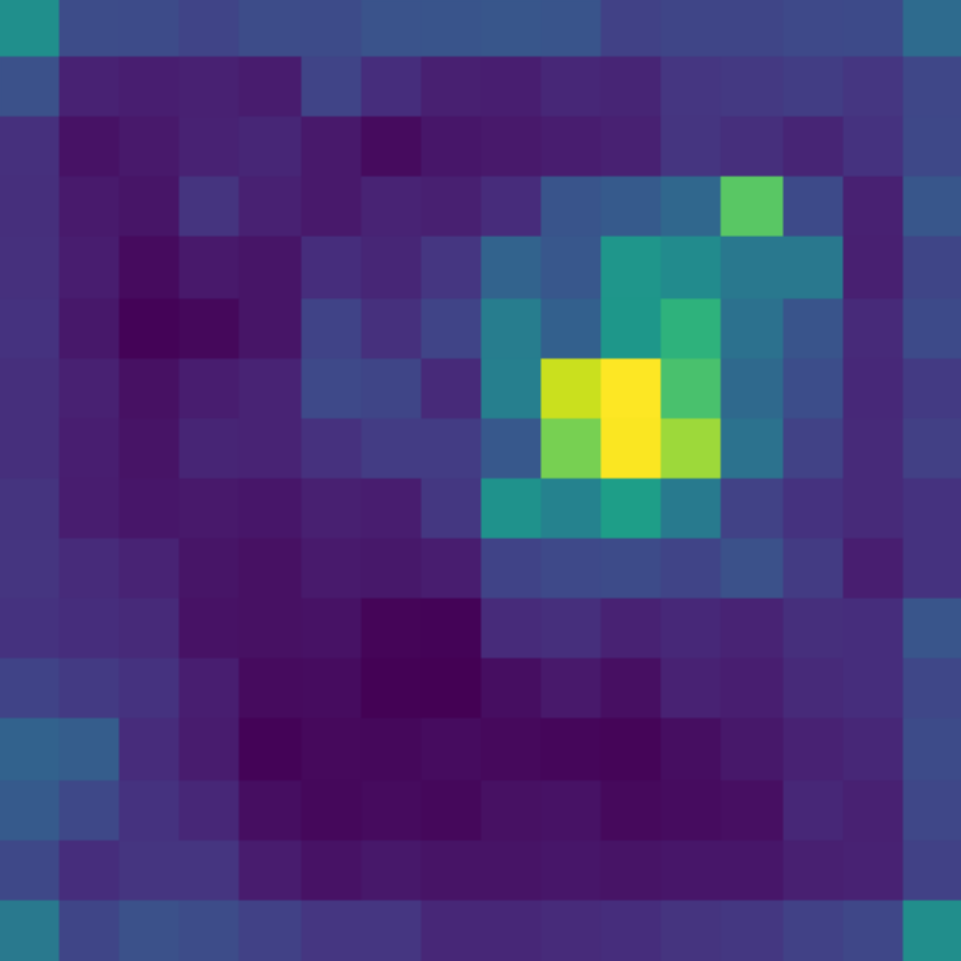}
\end{subfigure}
\begin{subfigure}{.05\textwidth}
  \centering
  \includegraphics[width=1.0\linewidth]{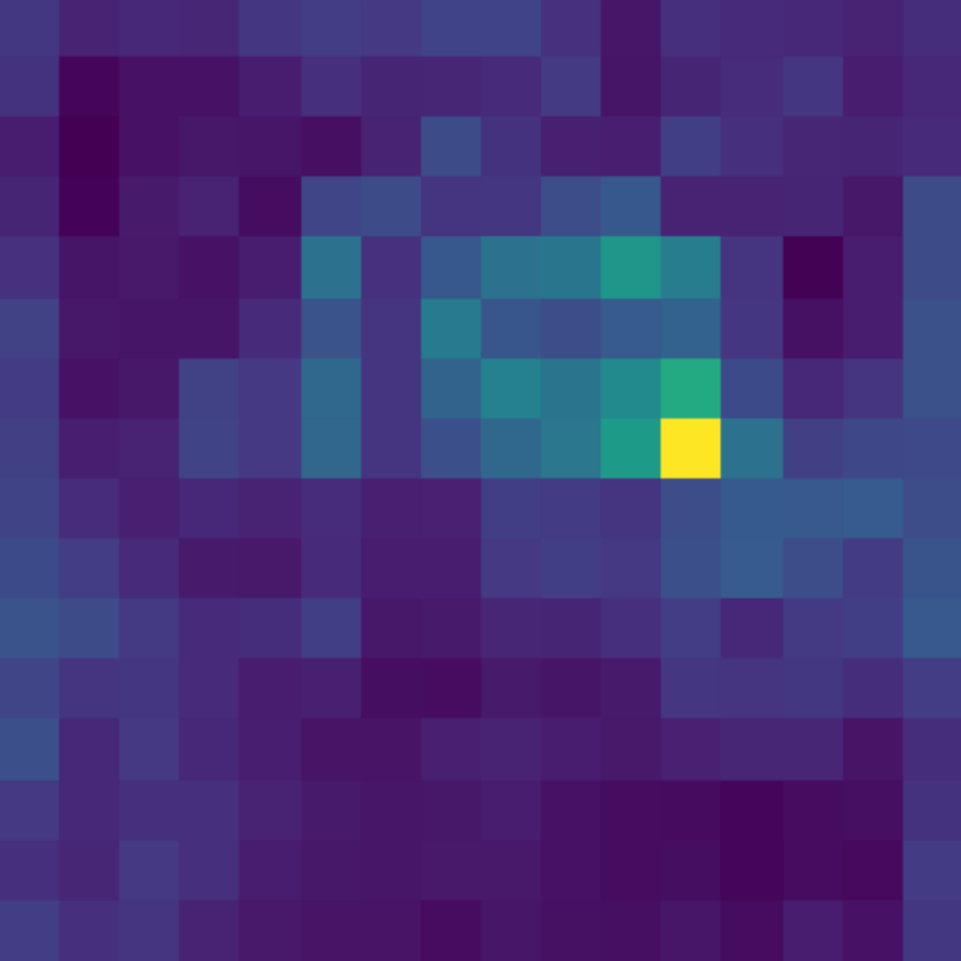}
\end{subfigure}
\begin{subfigure}{.05\textwidth}
  \centering
  \includegraphics[width=1.0\linewidth]{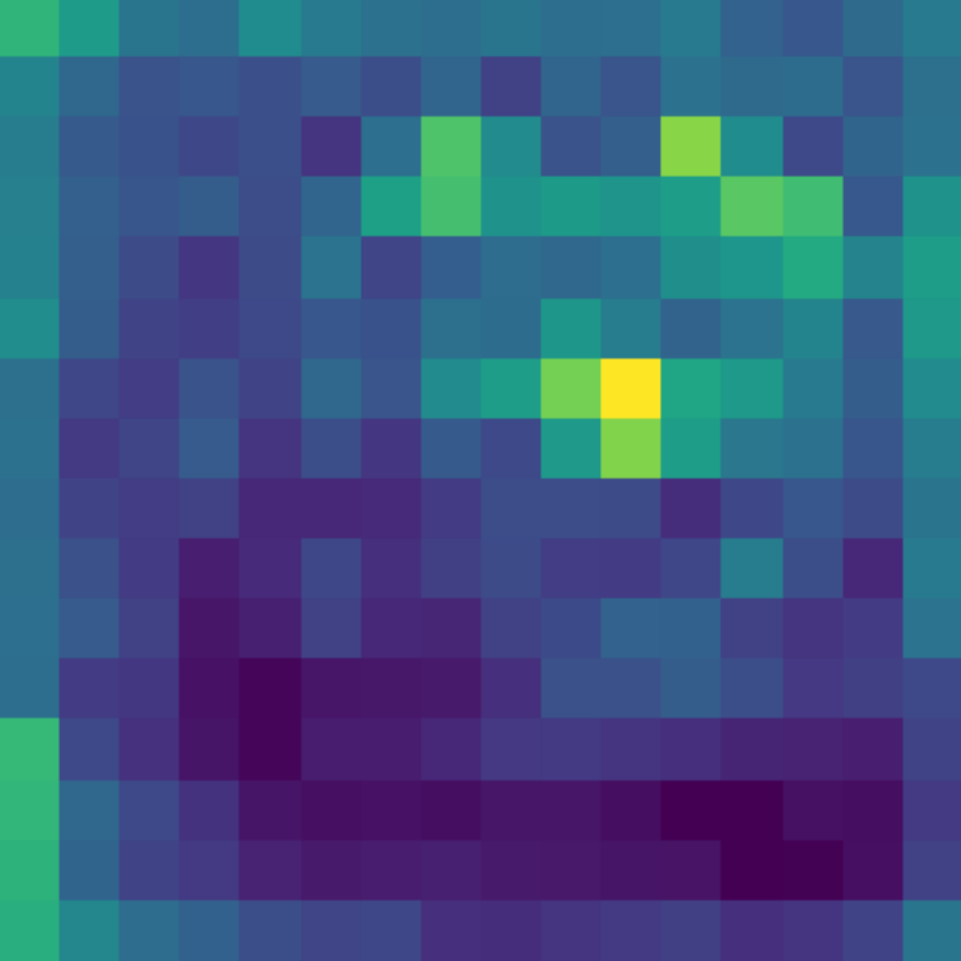}
\end{subfigure}
\begin{subfigure}{.05\textwidth}
  \centering
  \includegraphics[width=1.0\linewidth]{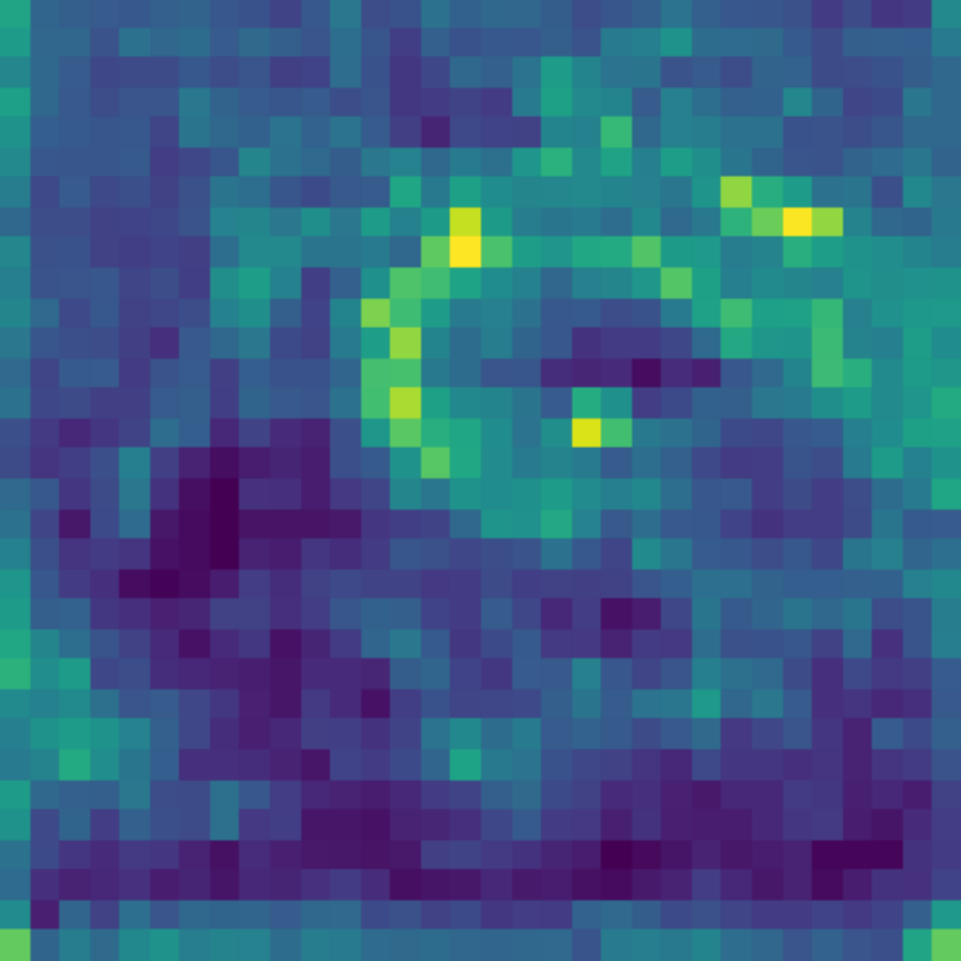}
\end{subfigure}
\begin{subfigure}{.05\textwidth}
  \centering
  \includegraphics[width=1.0\linewidth]{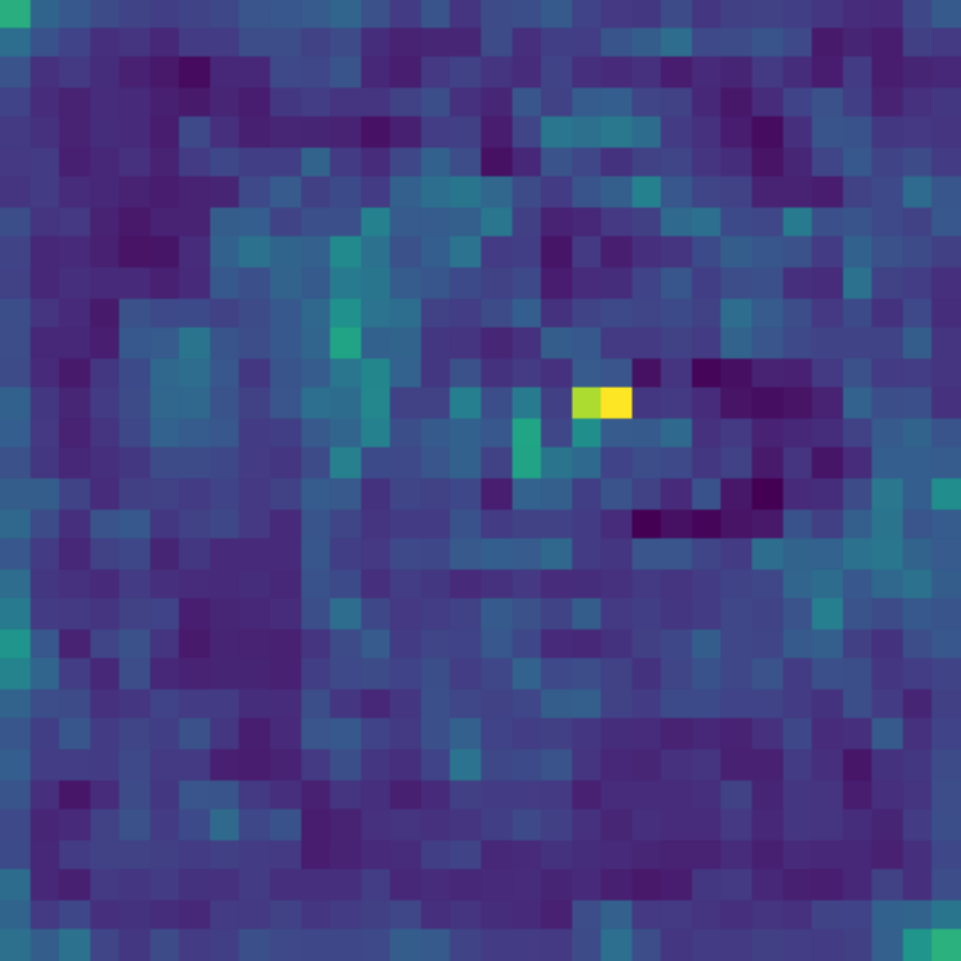}
\end{subfigure}
\begin{subfigure}{.05\textwidth}
  \centering
  \includegraphics[width=1.0\linewidth]{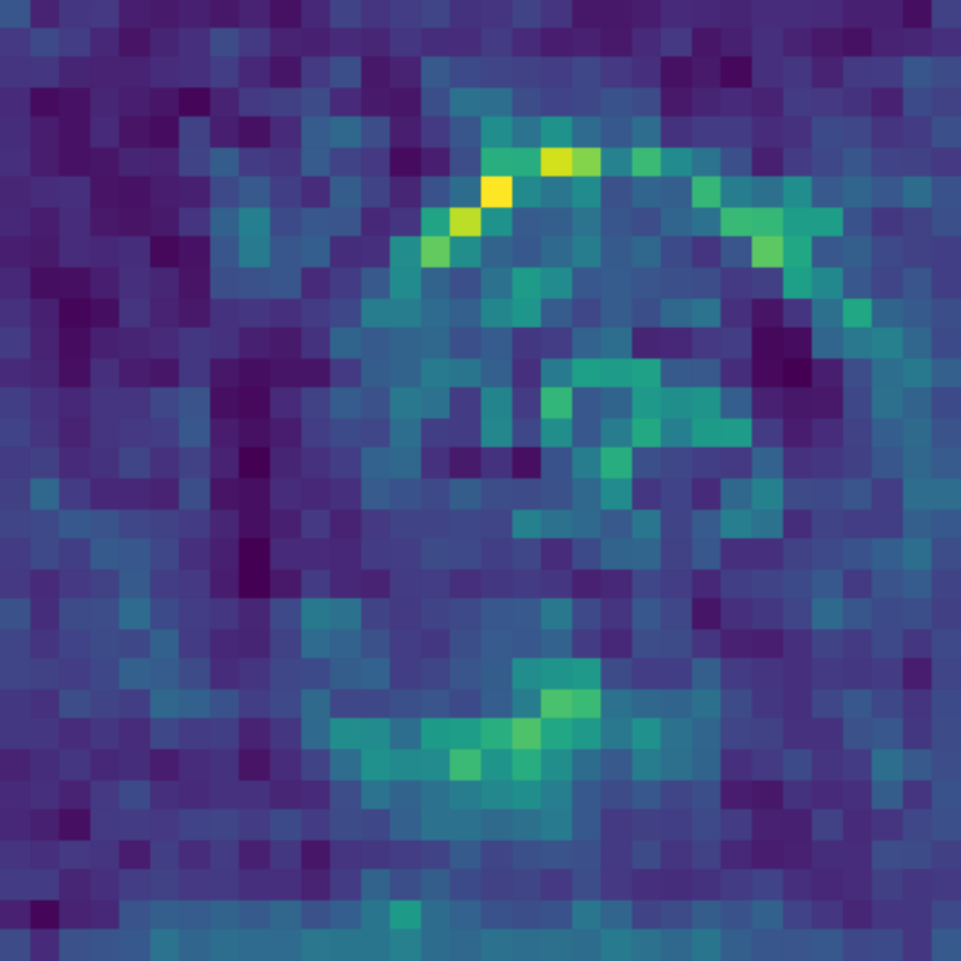}
\end{subfigure}
\\
&
\begin{subfigure}{.05\textwidth}
  \centering
  \includegraphics[width=1.0\linewidth]{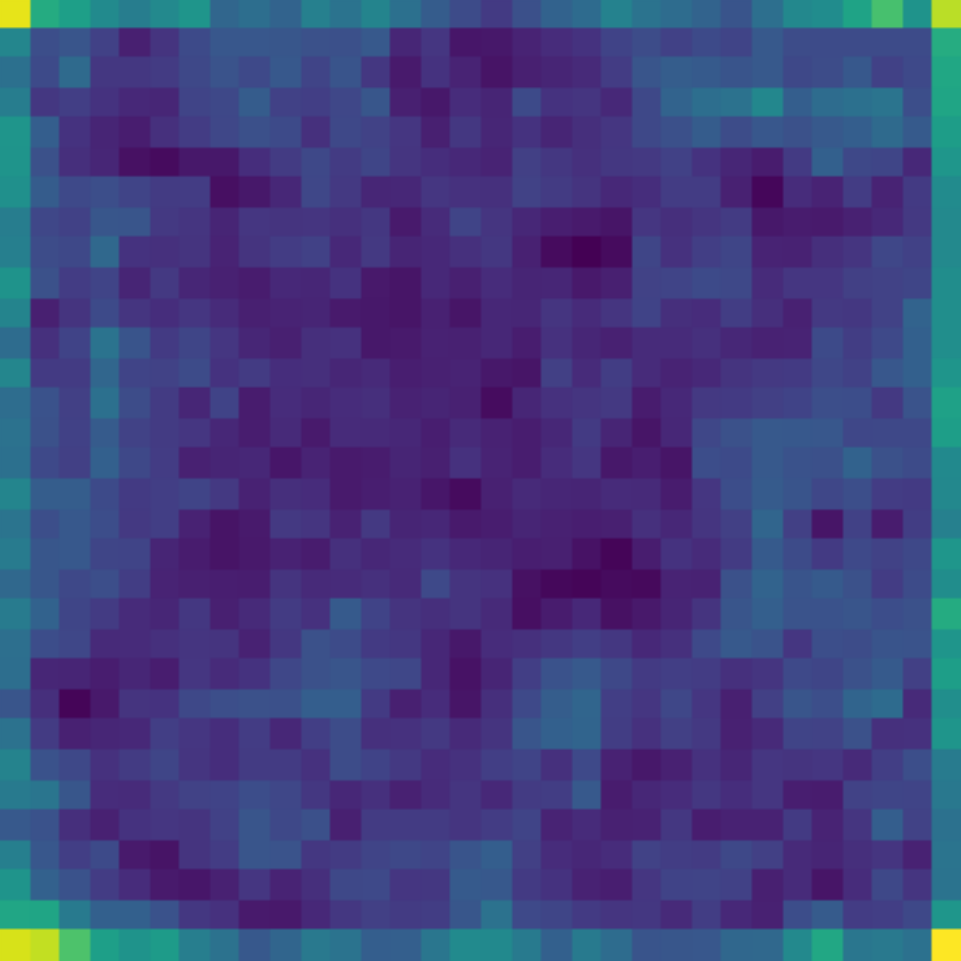}
\end{subfigure}
\begin{subfigure}{.05\textwidth}
  \centering
  \includegraphics[width=1.0\linewidth]{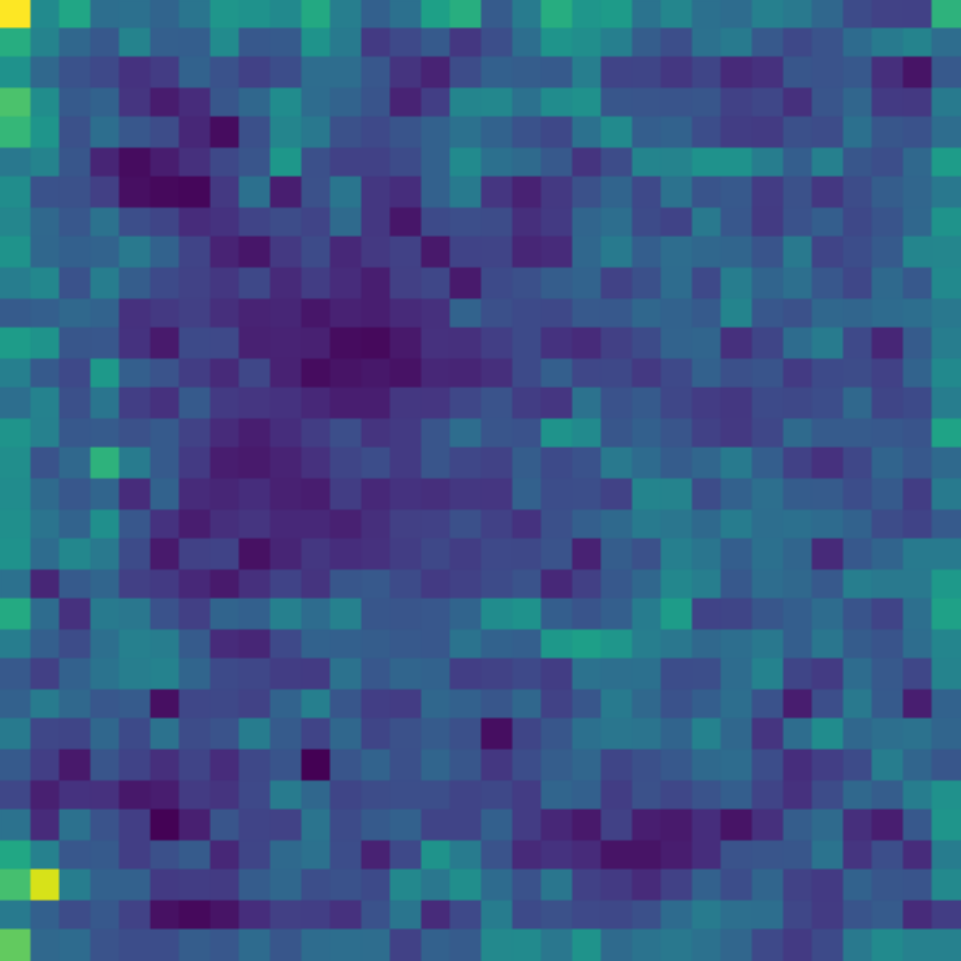}
\end{subfigure}
\begin{subfigure}{.05\textwidth}
  \centering
  \includegraphics[width=1.0\linewidth]{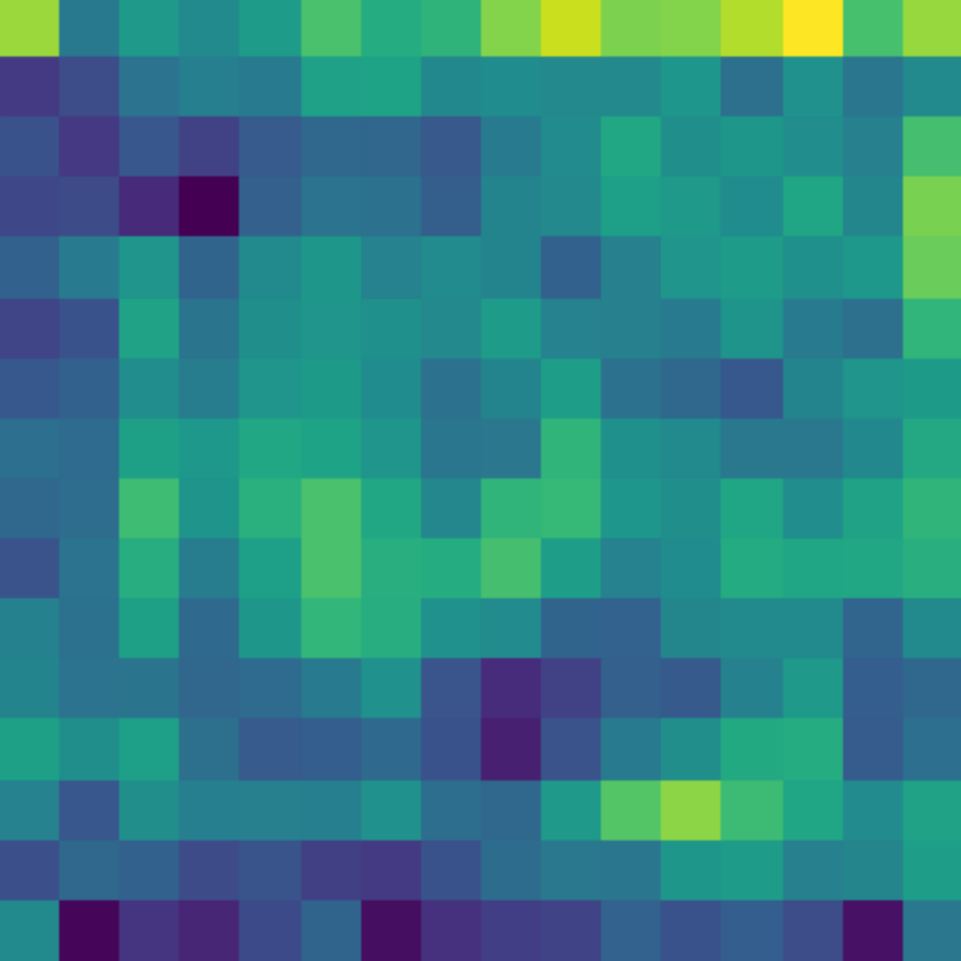}
\end{subfigure}
\begin{subfigure}{.05\textwidth}
  \centering
  \includegraphics[width=1.0\linewidth]{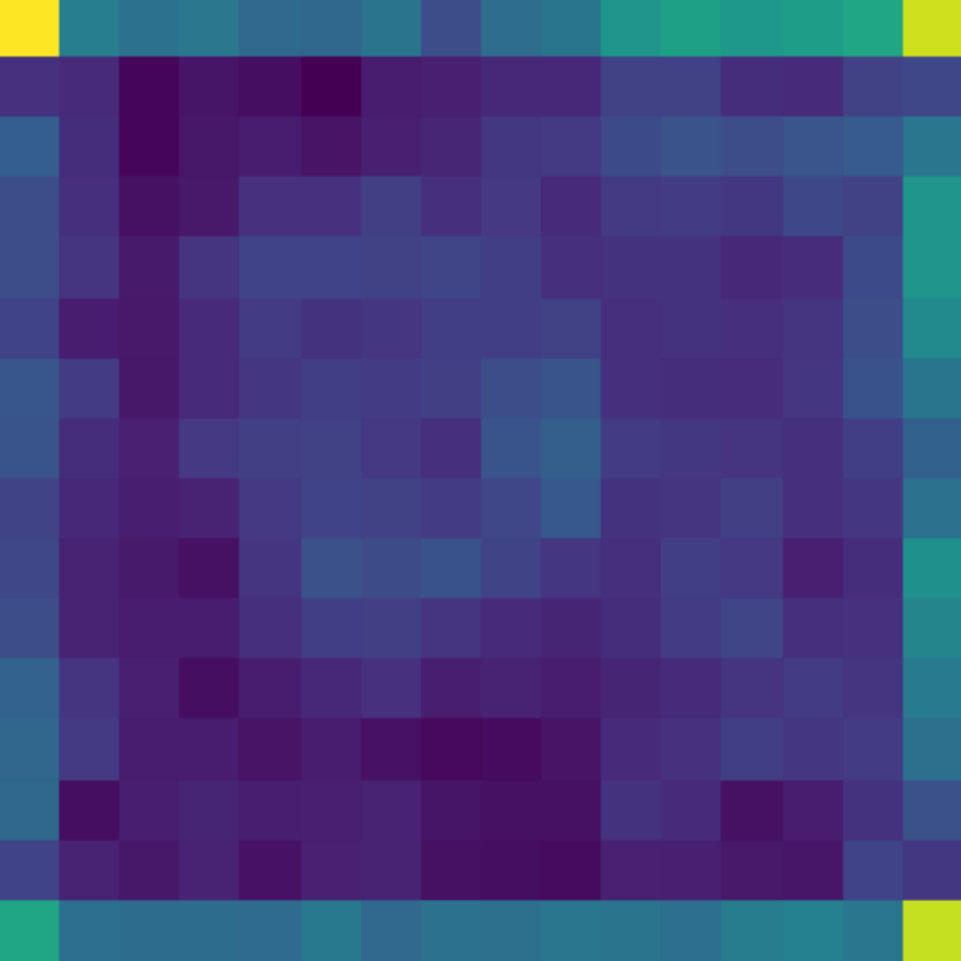}
\end{subfigure}
\begin{subfigure}{.05\textwidth}
  \centering
  \includegraphics[width=1.0\linewidth]{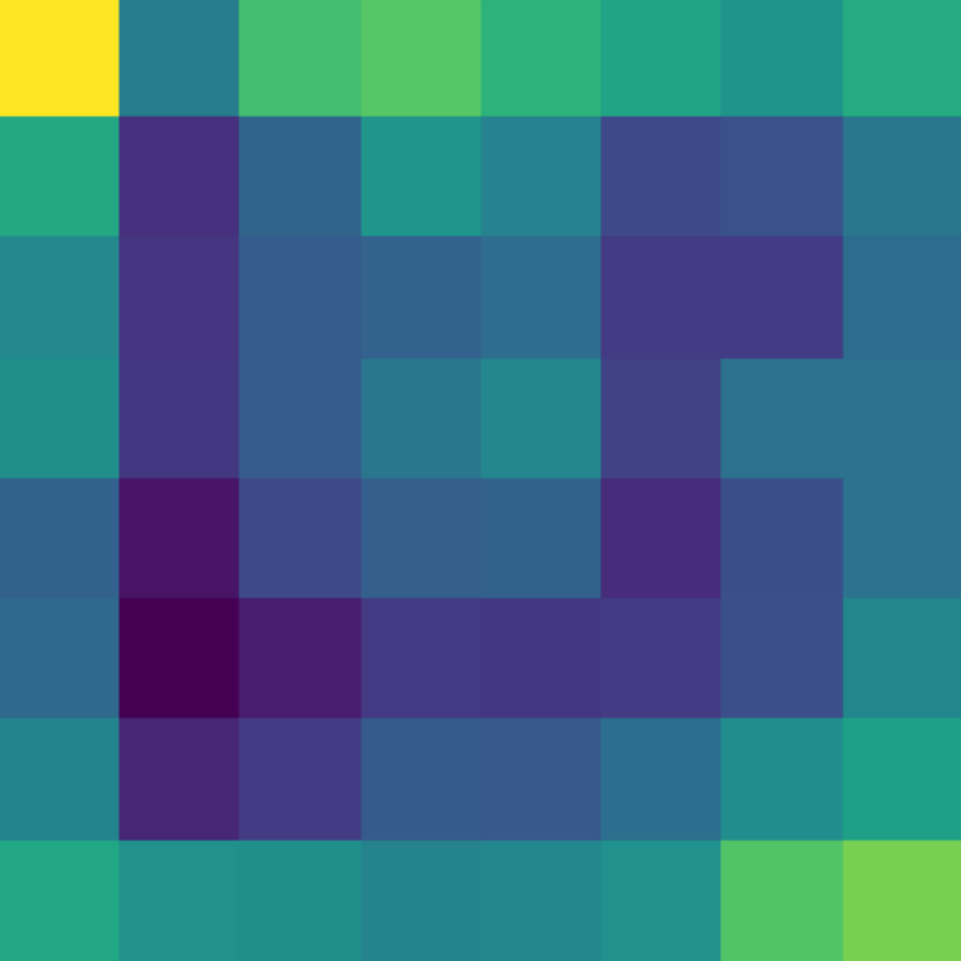}
\end{subfigure}
\begin{subfigure}{.05\textwidth}
  \centering
  \includegraphics[width=1.0\linewidth]{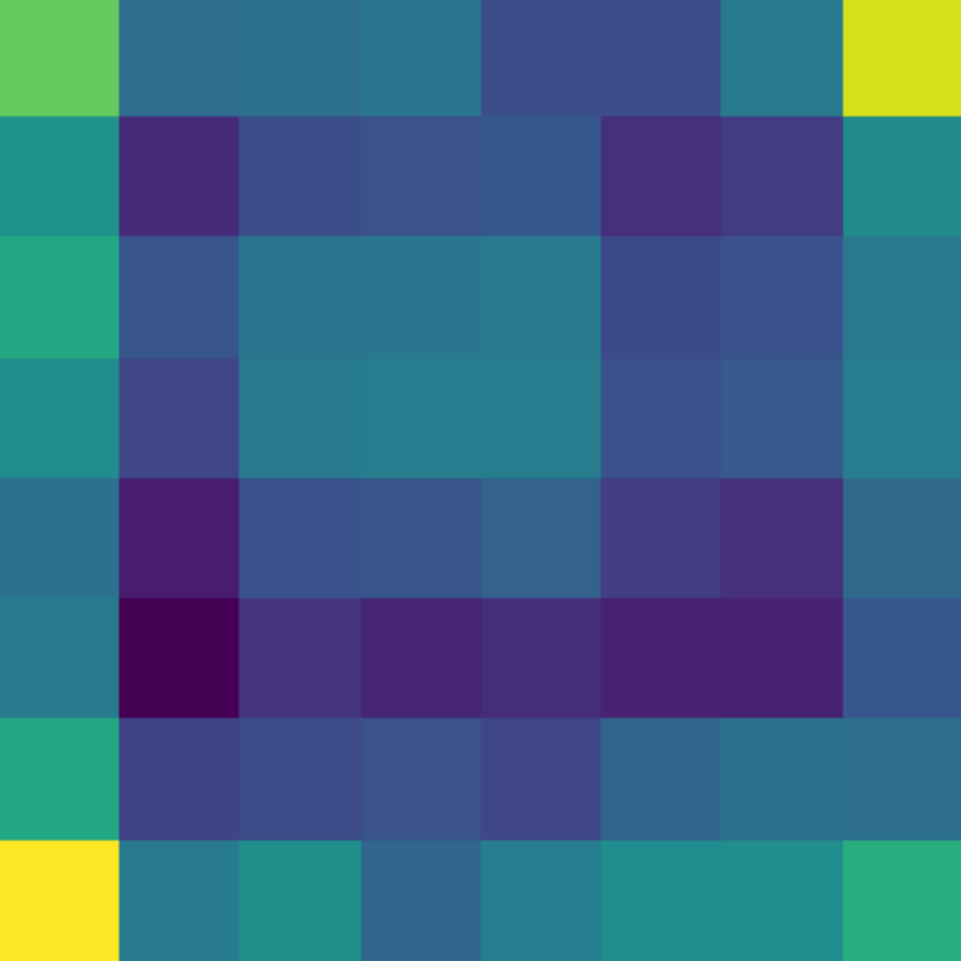}
\end{subfigure}
\begin{subfigure}{.05\textwidth}
  \centering
  \includegraphics[width=1.0\linewidth]{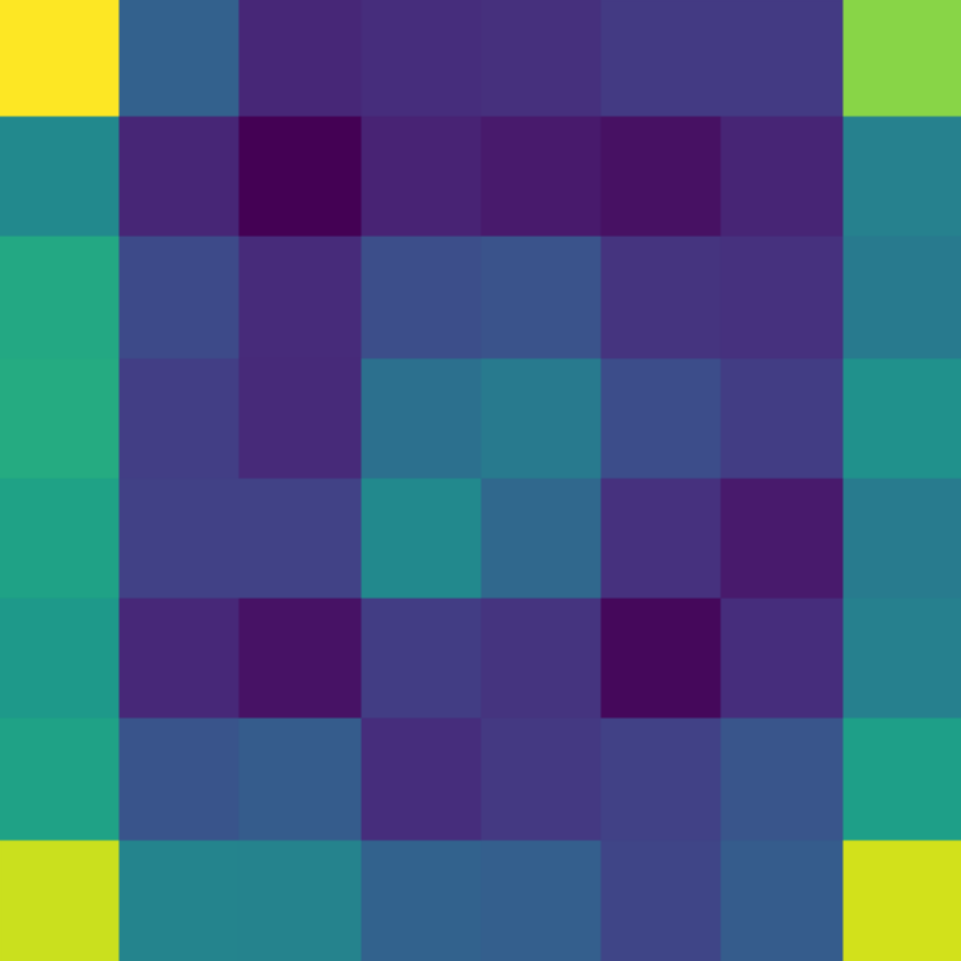}
\end{subfigure}
\begin{subfigure}{.05\textwidth}
  \centering
  \includegraphics[width=1.0\linewidth]{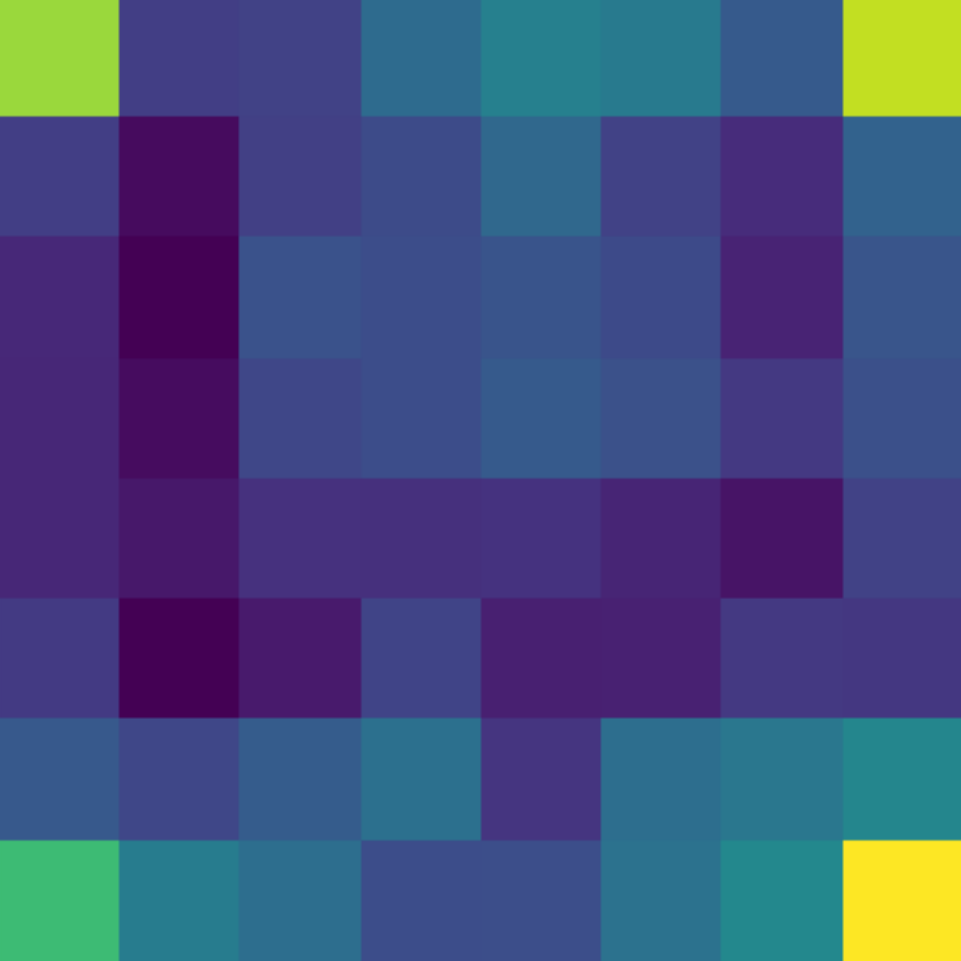}
\end{subfigure}
\begin{subfigure}{.05\textwidth}
  \centering
  \includegraphics[width=1.0\linewidth]{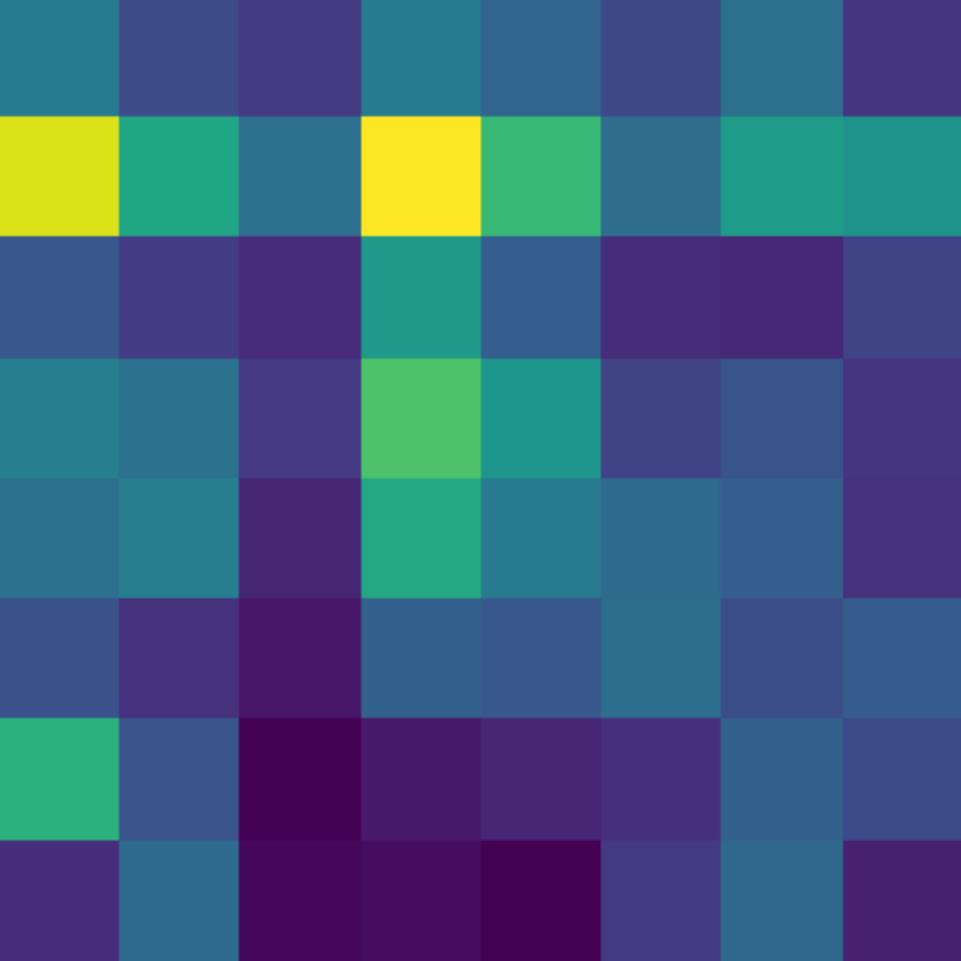}
\end{subfigure}
\begin{subfigure}{.05\textwidth}
  \centering
  \includegraphics[width=1.0\linewidth]{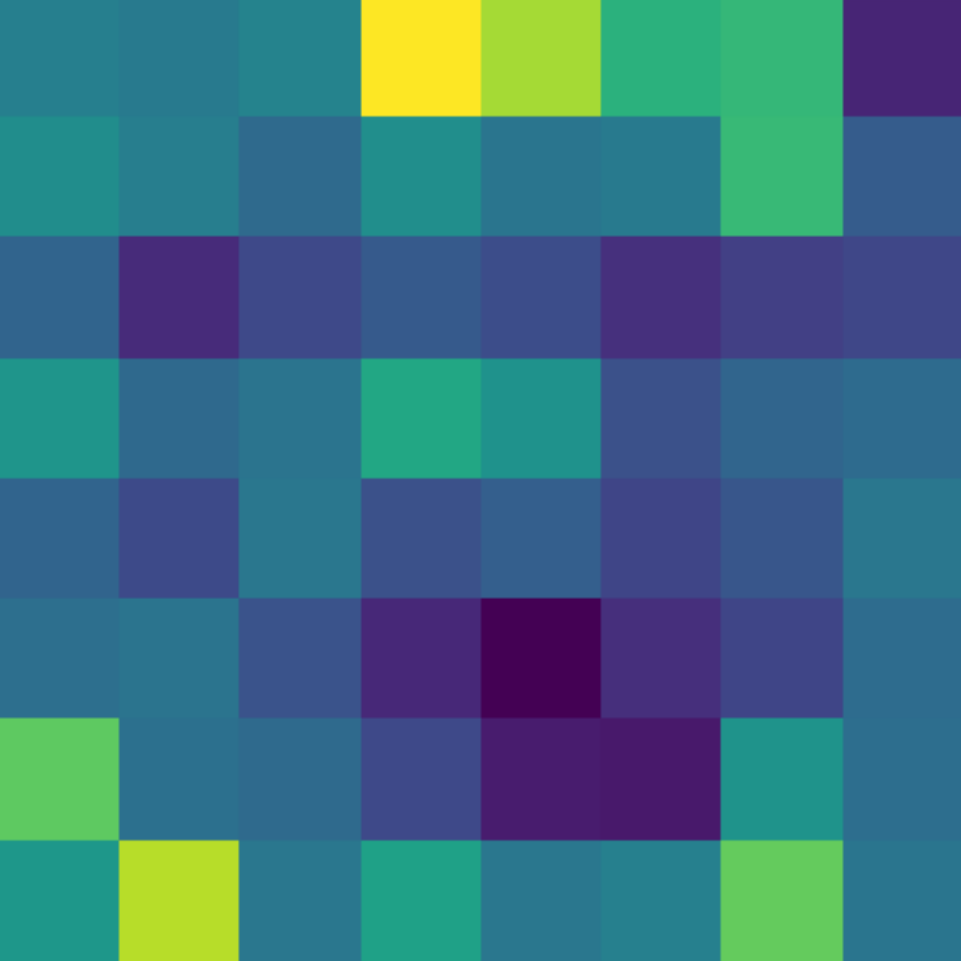}
\end{subfigure}
\begin{subfigure}{.05\textwidth}
  \centering
  \includegraphics[width=1.0\linewidth]{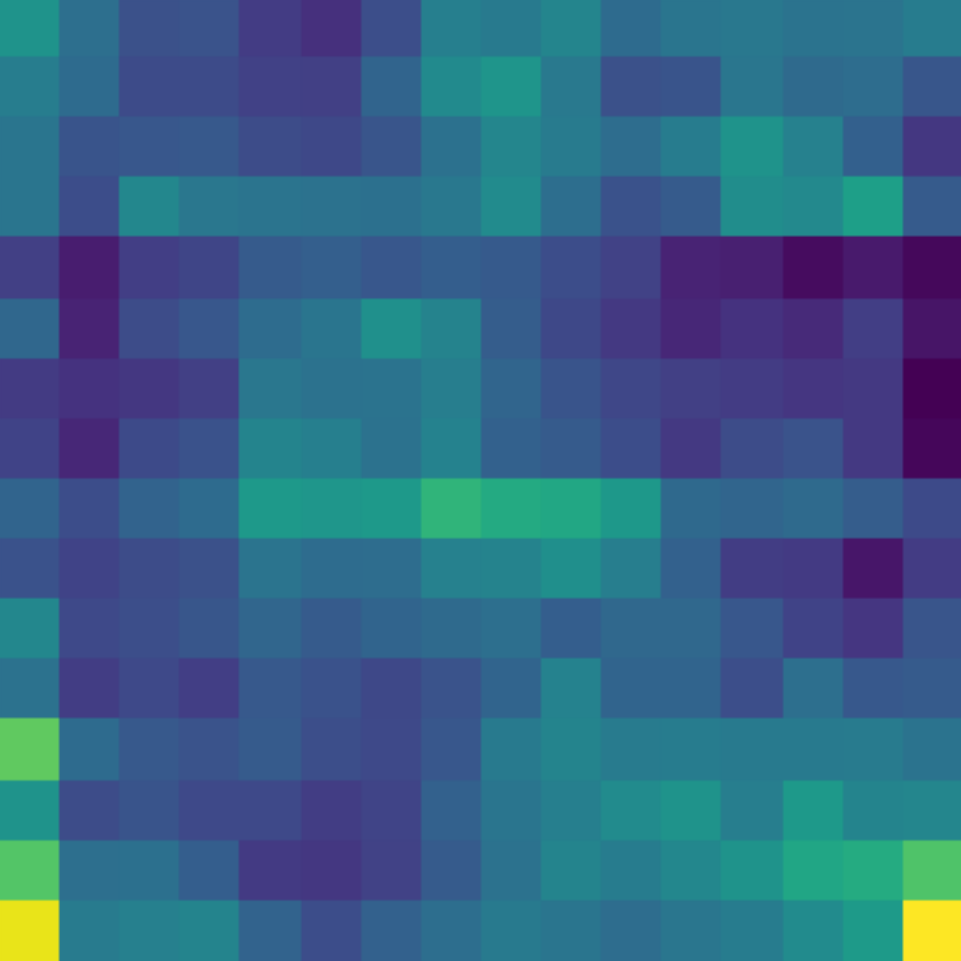}
\end{subfigure}
\begin{subfigure}{.05\textwidth}
  \centering
  \includegraphics[width=1.0\linewidth]{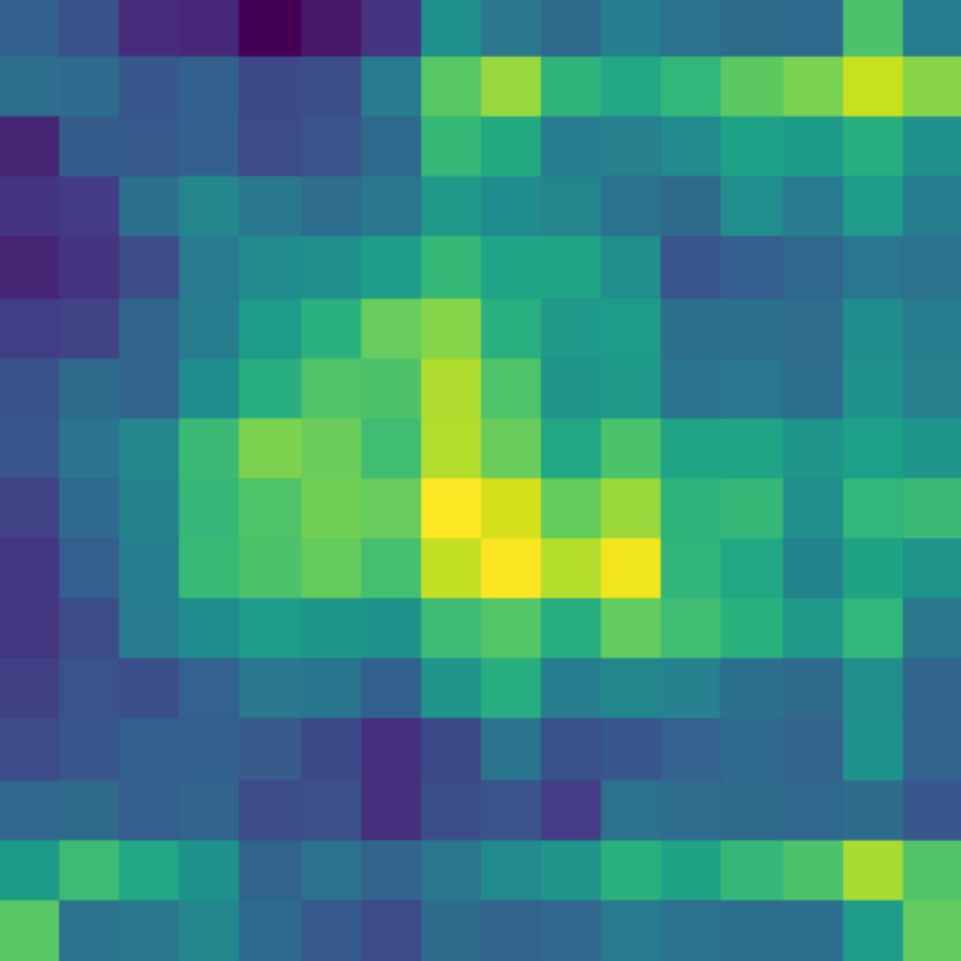}
\end{subfigure}
\begin{subfigure}{.05\textwidth}
  \centering
  \includegraphics[width=1.0\linewidth]{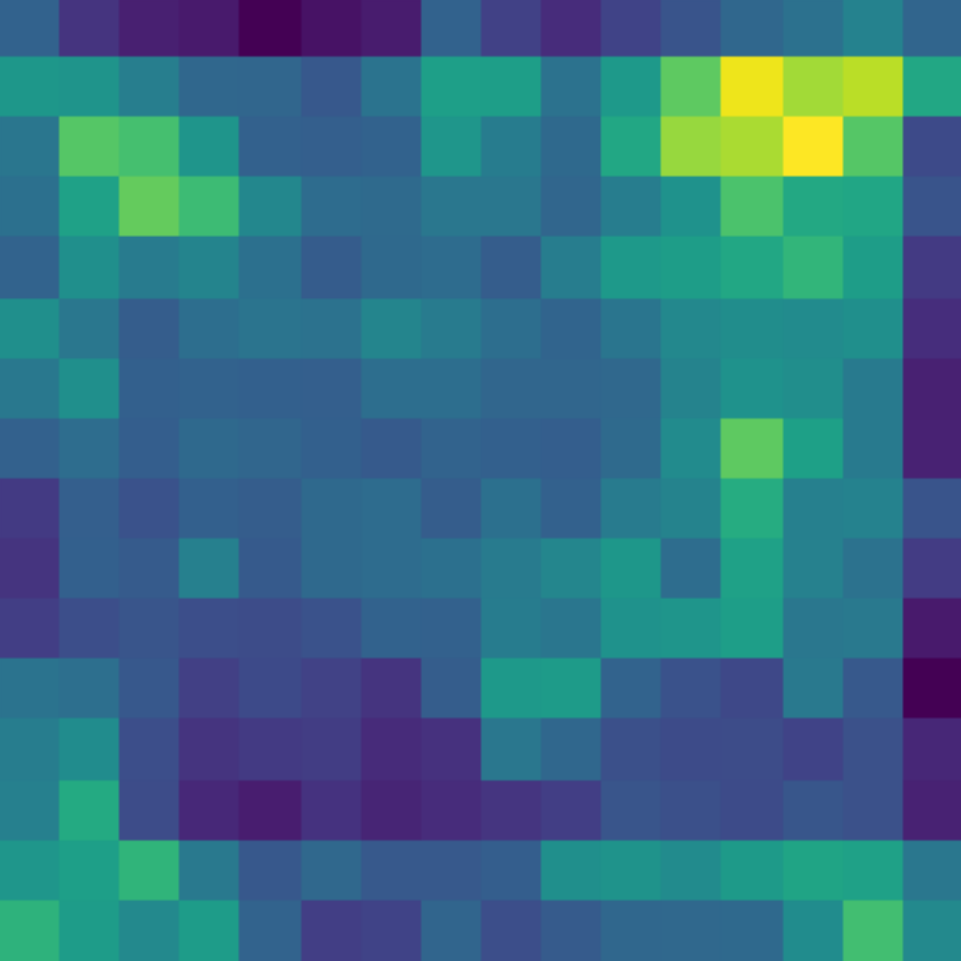}
\end{subfigure}
\begin{subfigure}{.05\textwidth}
  \centering
  \includegraphics[width=1.0\linewidth]{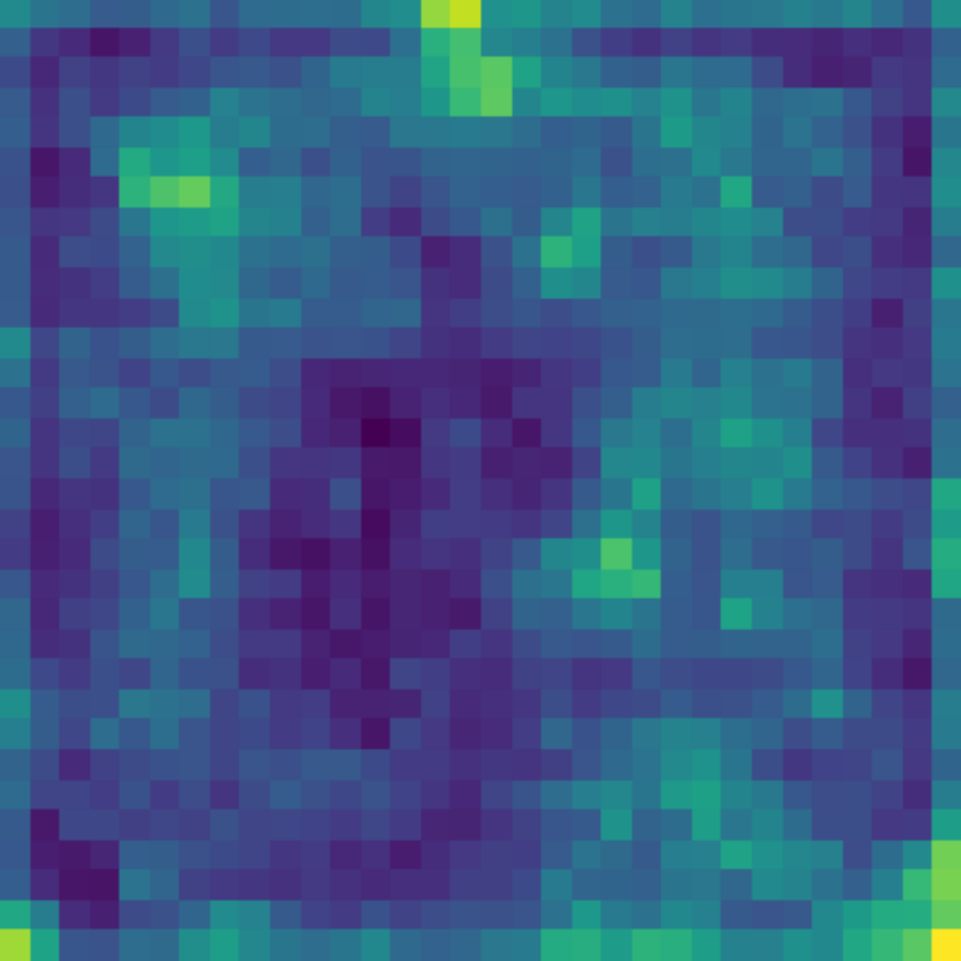}
\end{subfigure}
\begin{subfigure}{.05\textwidth}
  \centering
  \includegraphics[width=1.0\linewidth]{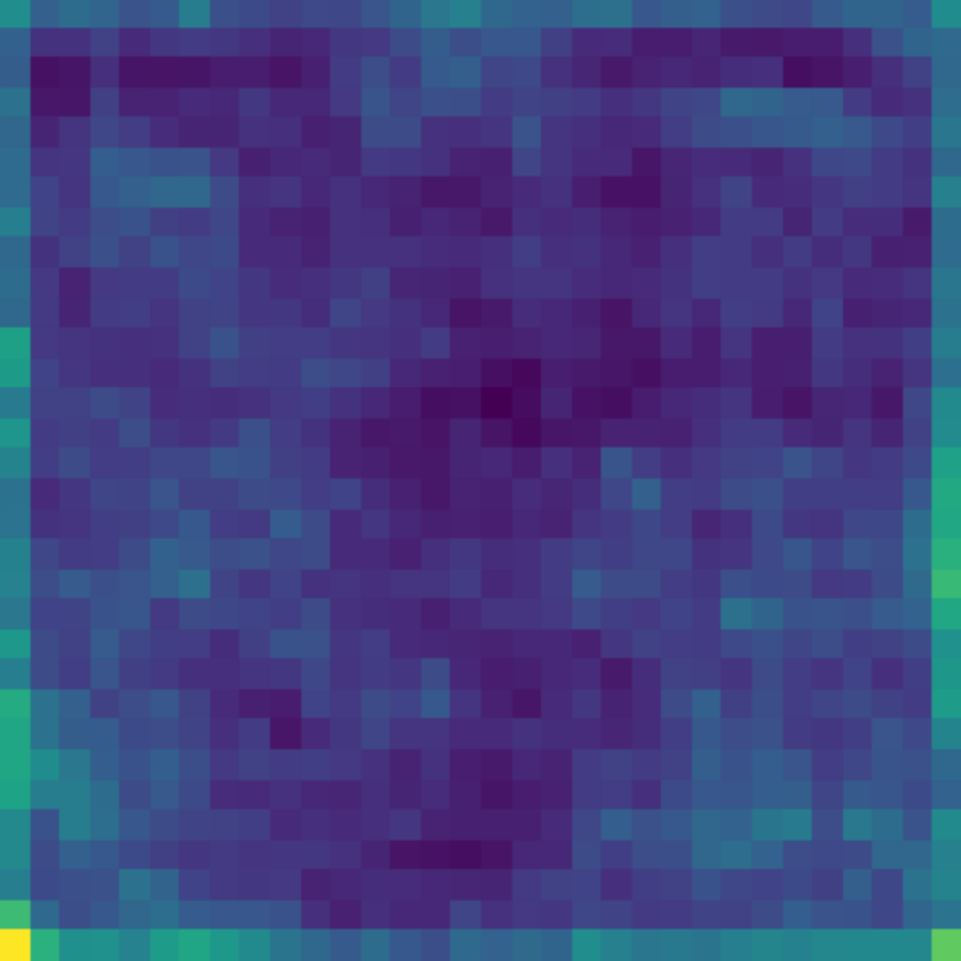}
\end{subfigure}
\begin{subfigure}{.05\textwidth}
  \centering
  \includegraphics[width=1.0\linewidth]{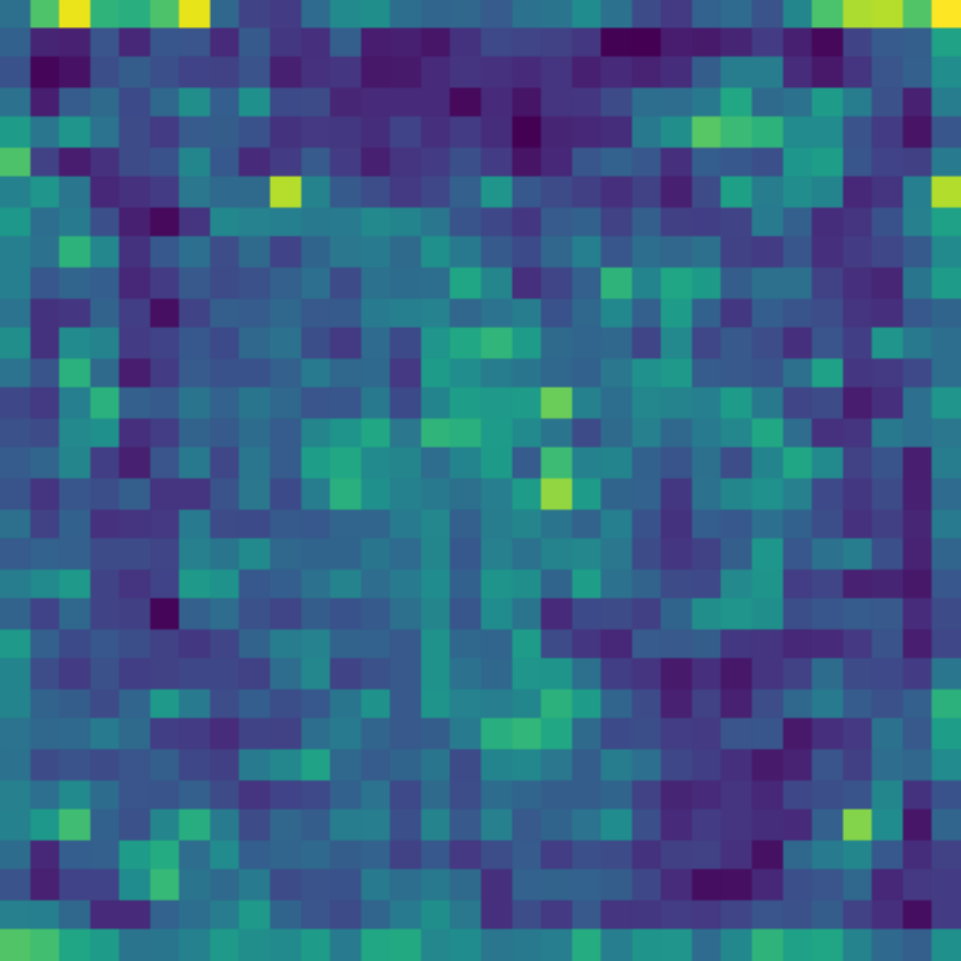}
\end{subfigure}\vspace{10pt}
\\

\begin{subfigure}{.05\textwidth}
  \centering
  \includegraphics[width=1.0\linewidth]{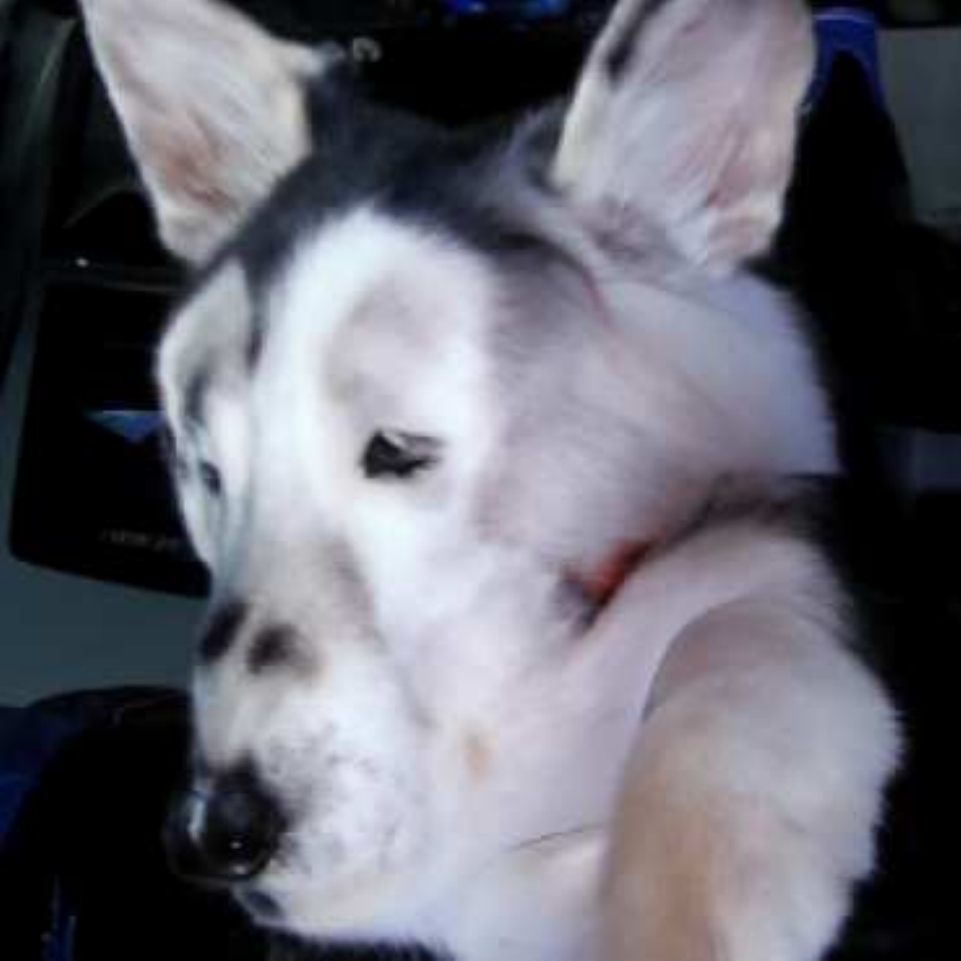}
\end{subfigure}
&
\begin{subfigure}{.05\textwidth}
  \centering
  \includegraphics[width=1.0\linewidth]{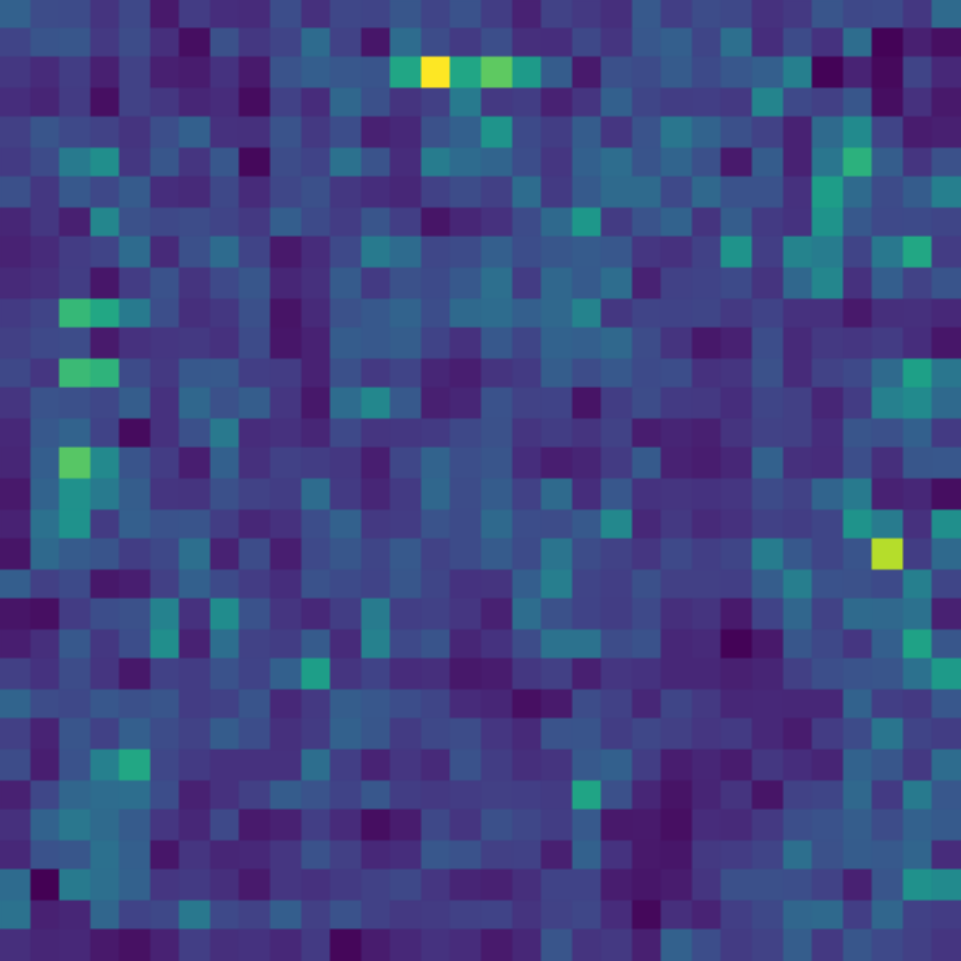}
\end{subfigure}
\begin{subfigure}{.05\textwidth}
  \centering
  \includegraphics[width=1.0\linewidth]{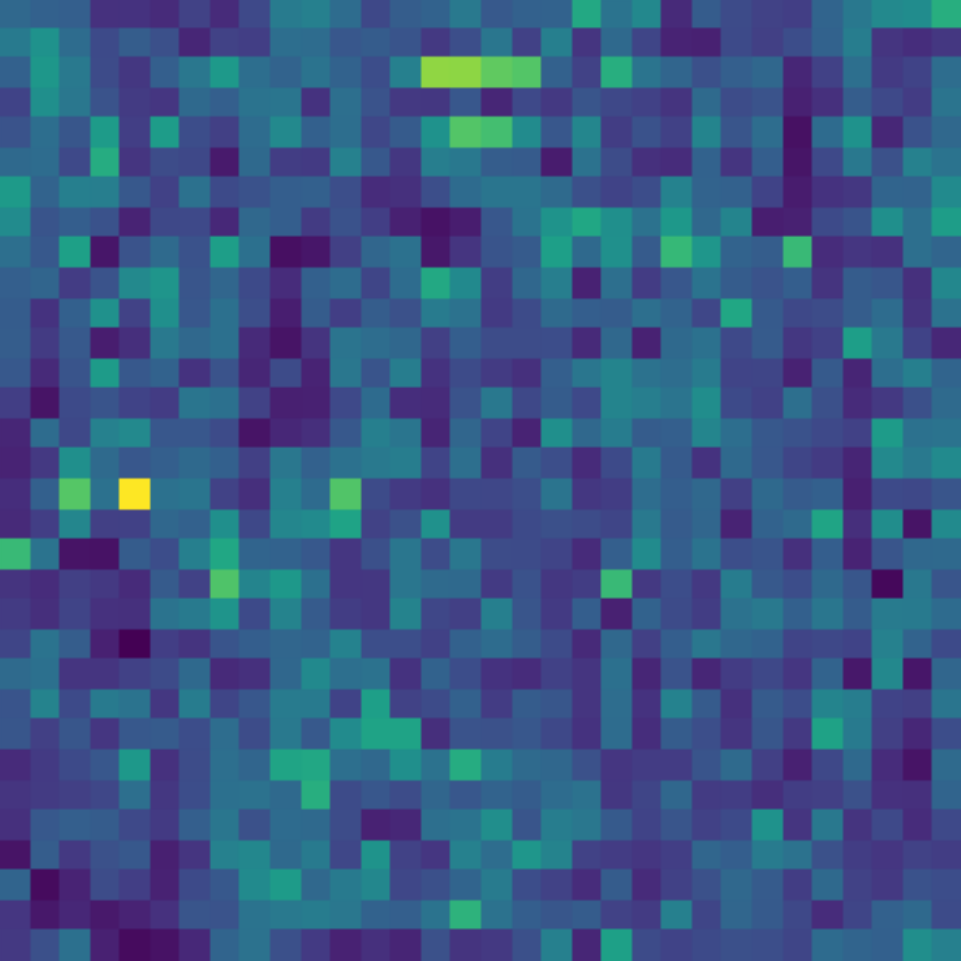}
\end{subfigure}
\begin{subfigure}{.05\textwidth}
  \centering
  \includegraphics[width=1.0\linewidth]{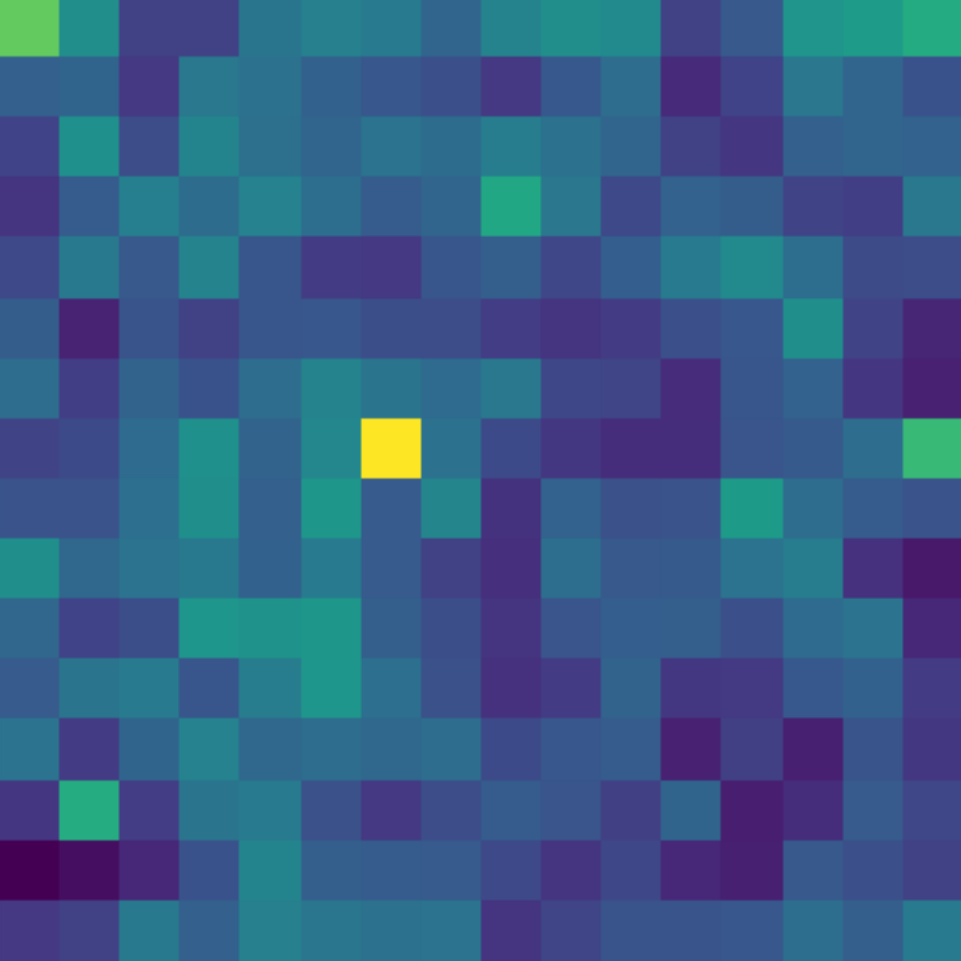}
\end{subfigure}
\begin{subfigure}{.05\textwidth}
  \centering
  \includegraphics[width=1.0\linewidth]{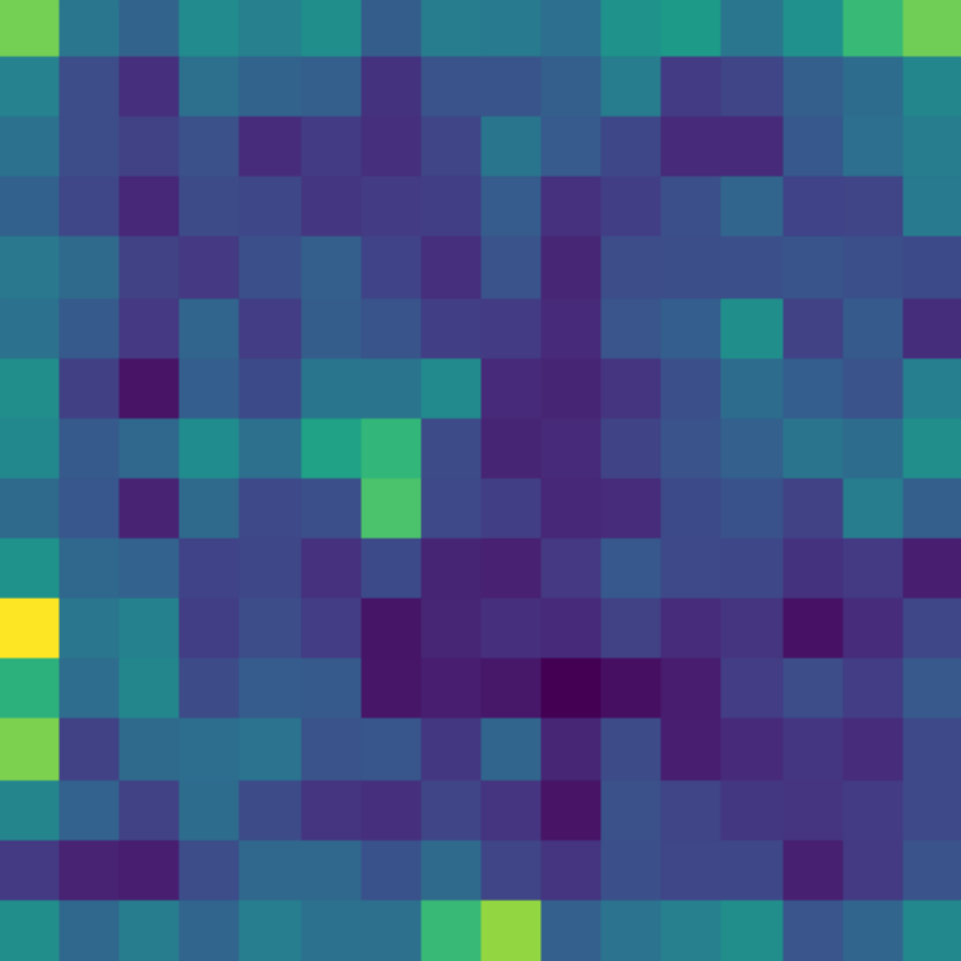}
\end{subfigure}
\begin{subfigure}{.05\textwidth}
  \centering
  \includegraphics[width=1.0\linewidth]{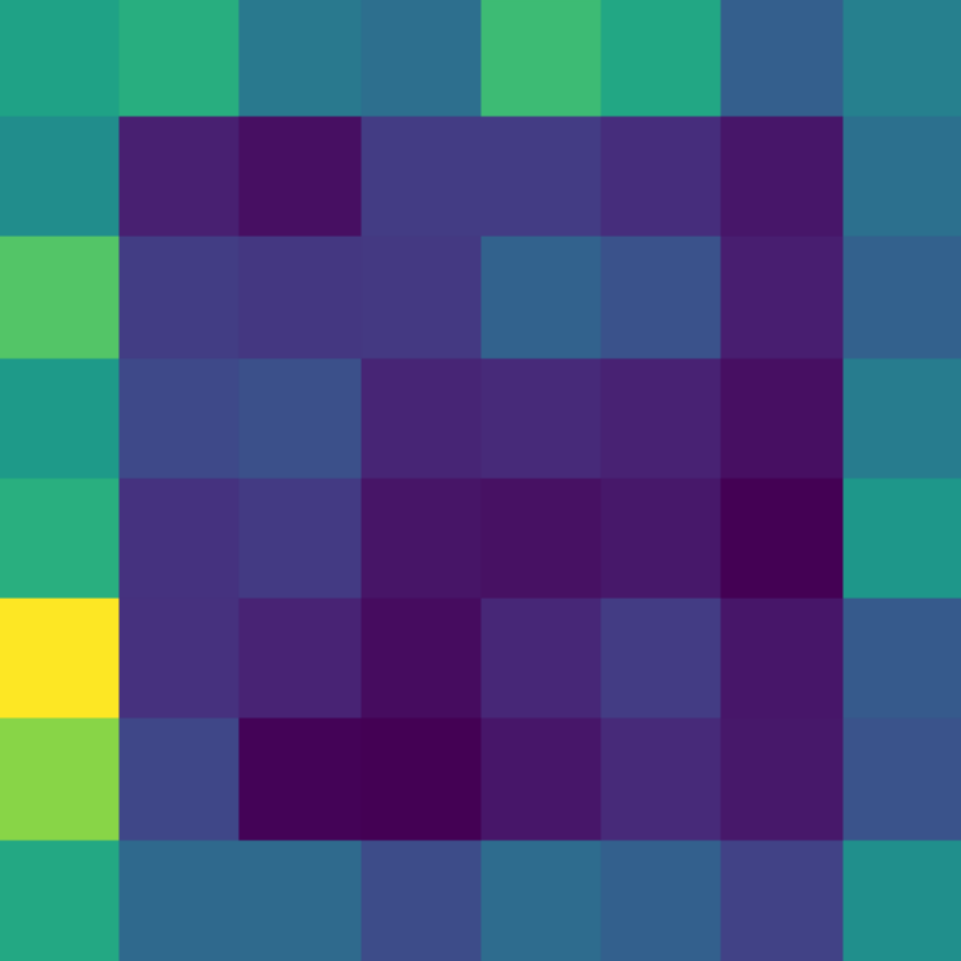}
\end{subfigure}
\begin{subfigure}{.05\textwidth}
  \centering
  \includegraphics[width=1.0\linewidth]{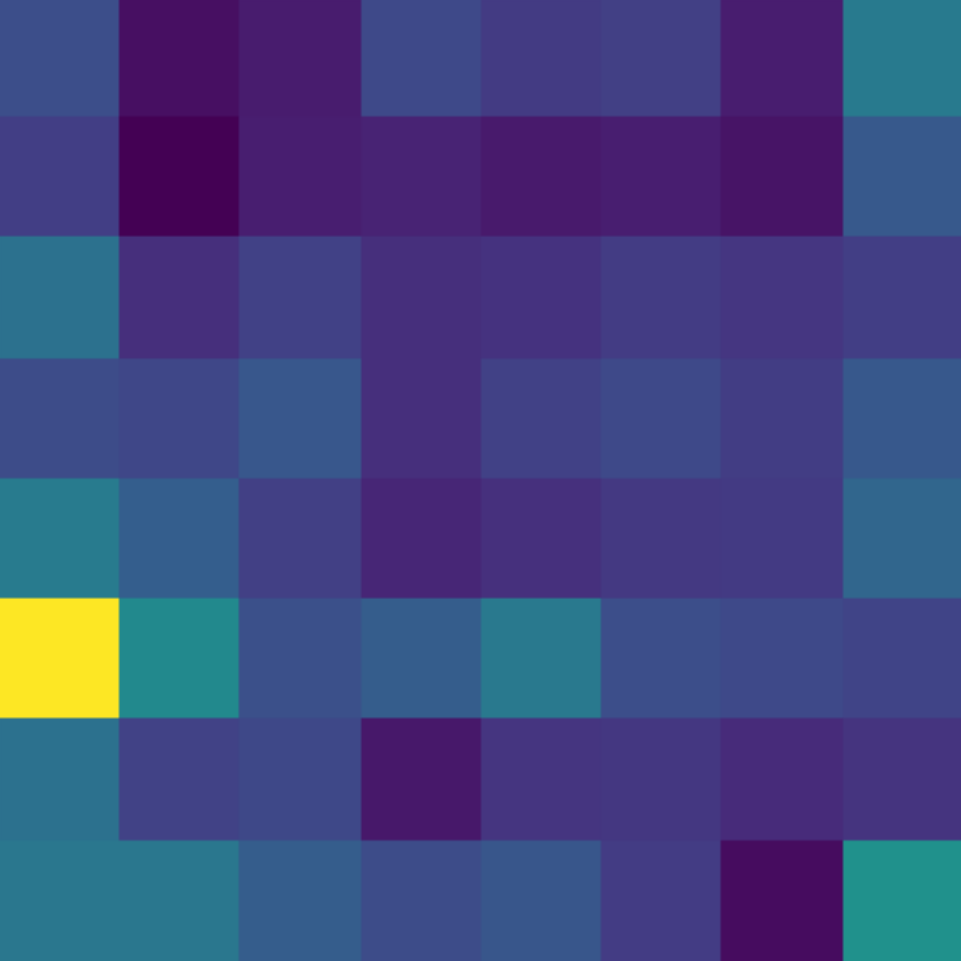}
\end{subfigure}
\begin{subfigure}{.05\textwidth}
  \centering
  \includegraphics[width=1.0\linewidth]{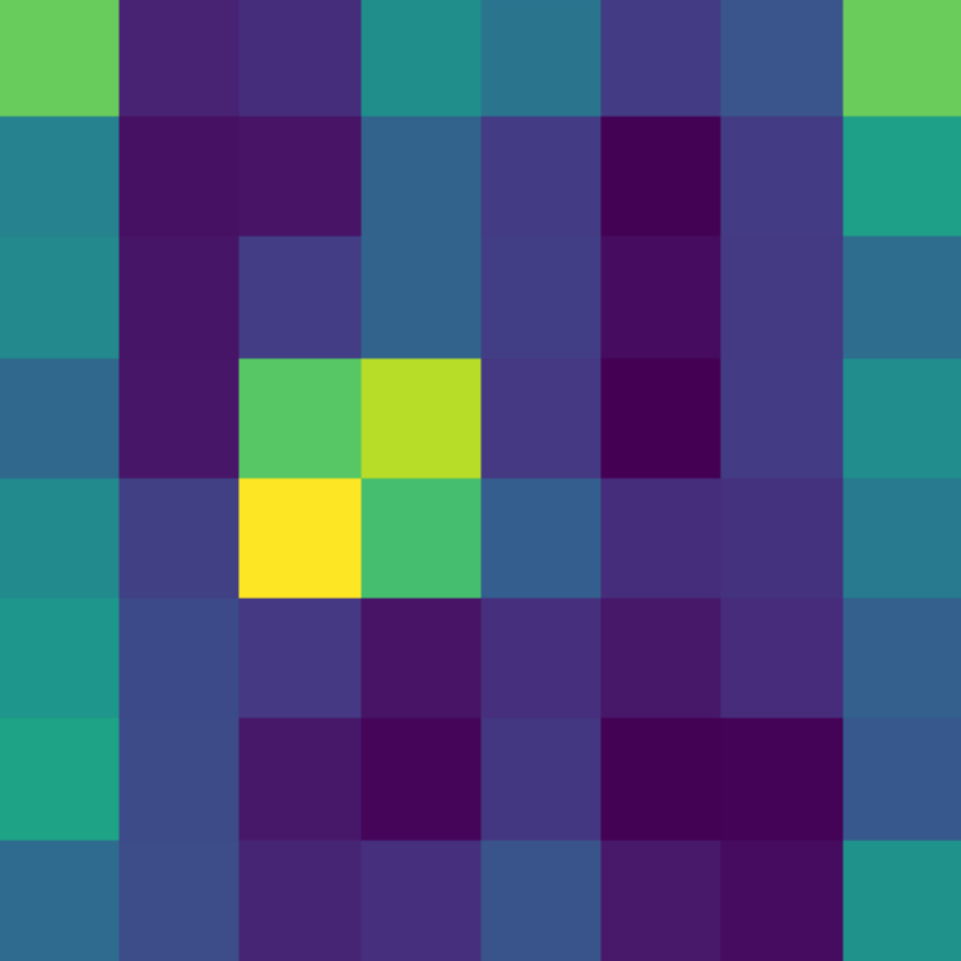}
\end{subfigure}
\begin{subfigure}{.05\textwidth}
  \centering
  \includegraphics[width=1.0\linewidth]{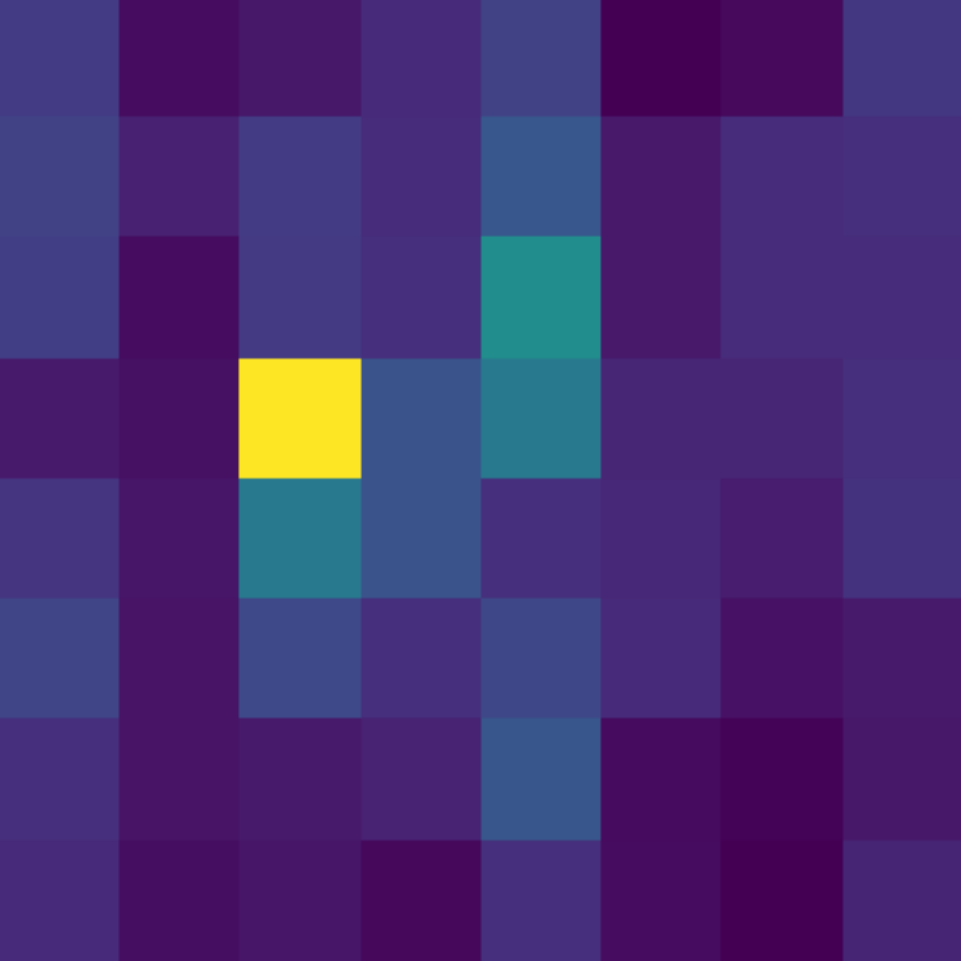}
\end{subfigure}
\begin{subfigure}{.05\textwidth}
  \centering
  \includegraphics[width=1.0\linewidth]{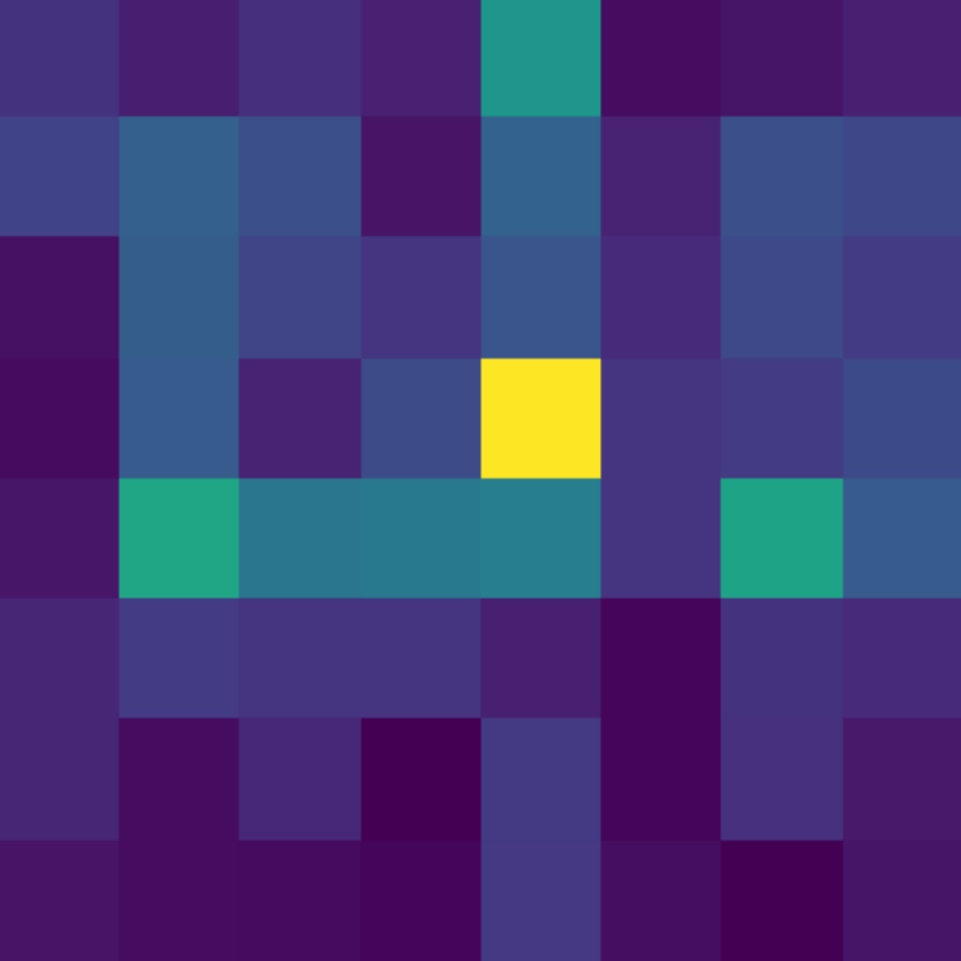}
\end{subfigure}
\begin{subfigure}{.05\textwidth}
  \centering
  \includegraphics[width=1.0\linewidth]{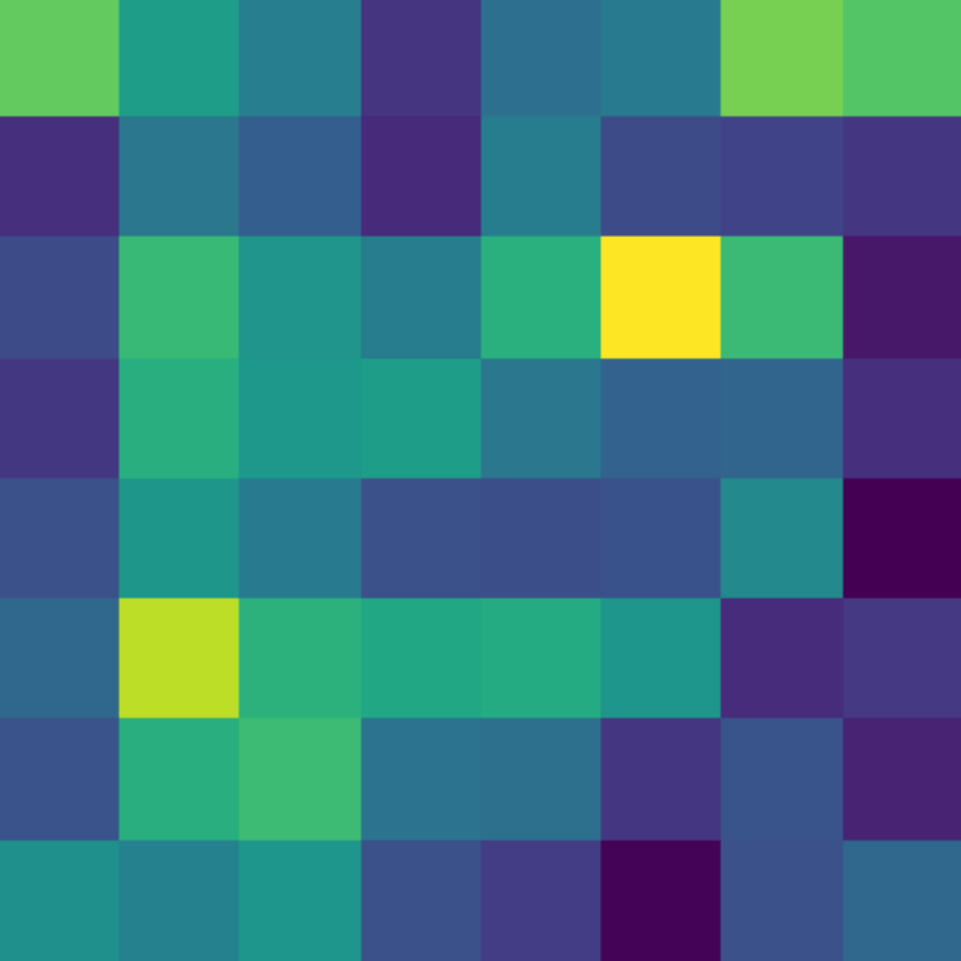}
\end{subfigure}
\begin{subfigure}{.05\textwidth}
  \centering
  \includegraphics[width=1.0\linewidth]{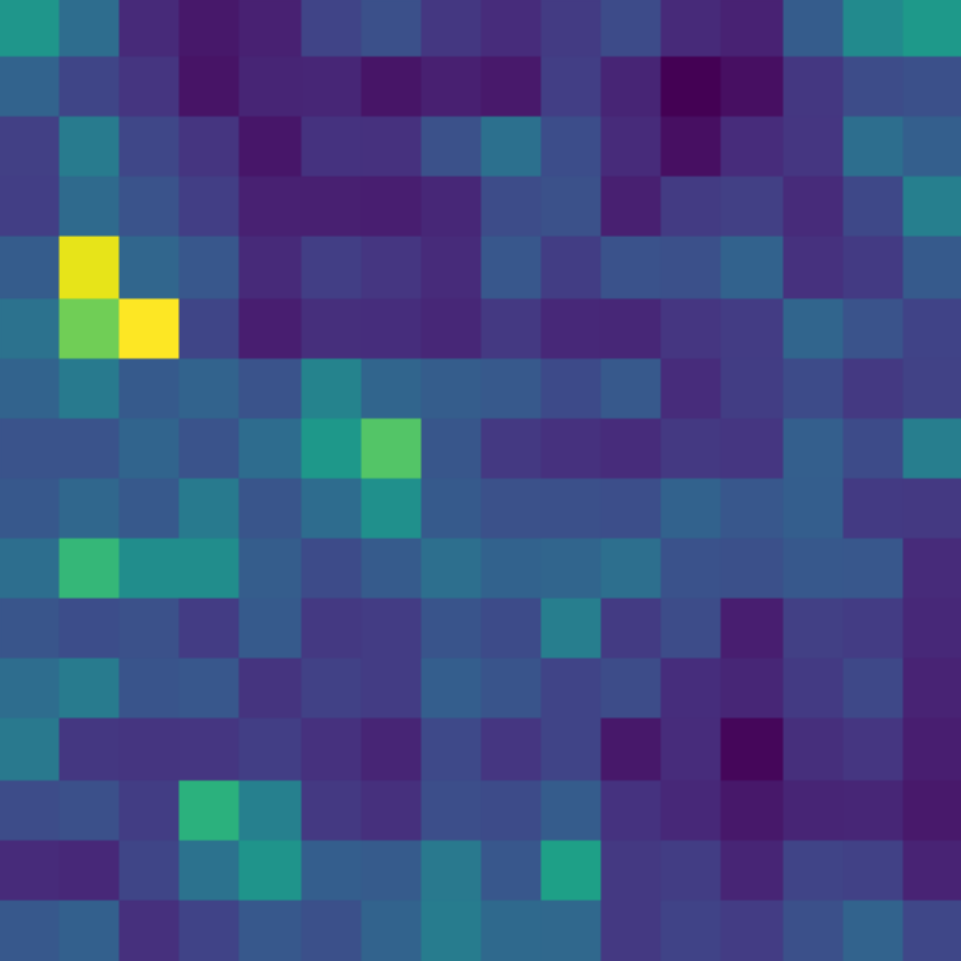}
\end{subfigure}
\begin{subfigure}{.05\textwidth}
  \centering
  \includegraphics[width=1.0\linewidth]{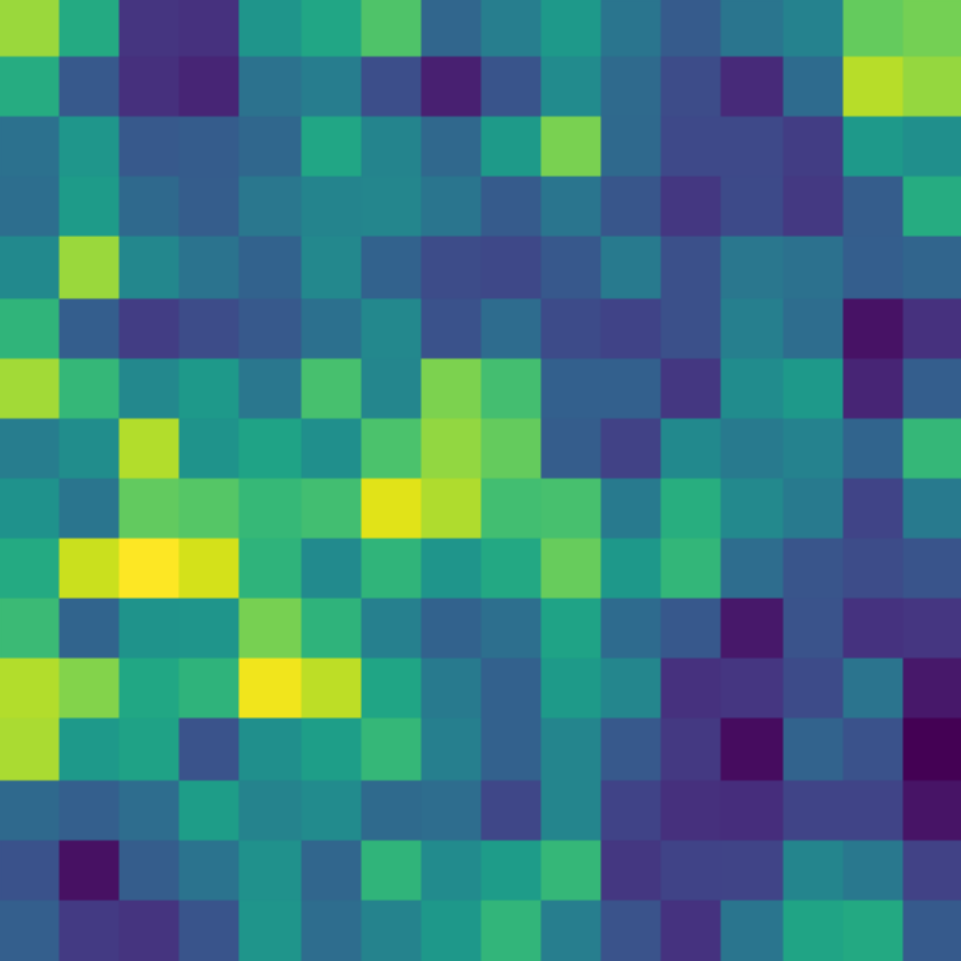}
\end{subfigure}
\begin{subfigure}{.05\textwidth}
  \centering
  \includegraphics[width=1.0\linewidth]{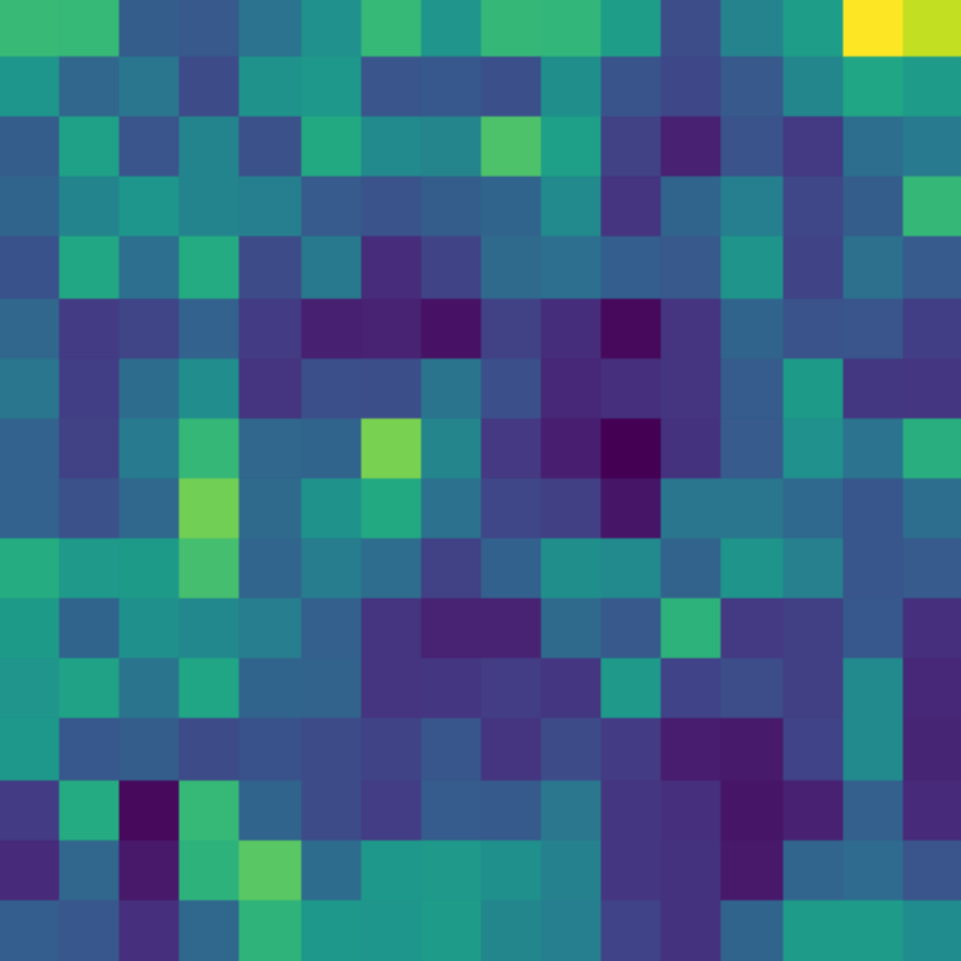}
\end{subfigure}
\begin{subfigure}{.05\textwidth}
  \centering
  \includegraphics[width=1.0\linewidth]{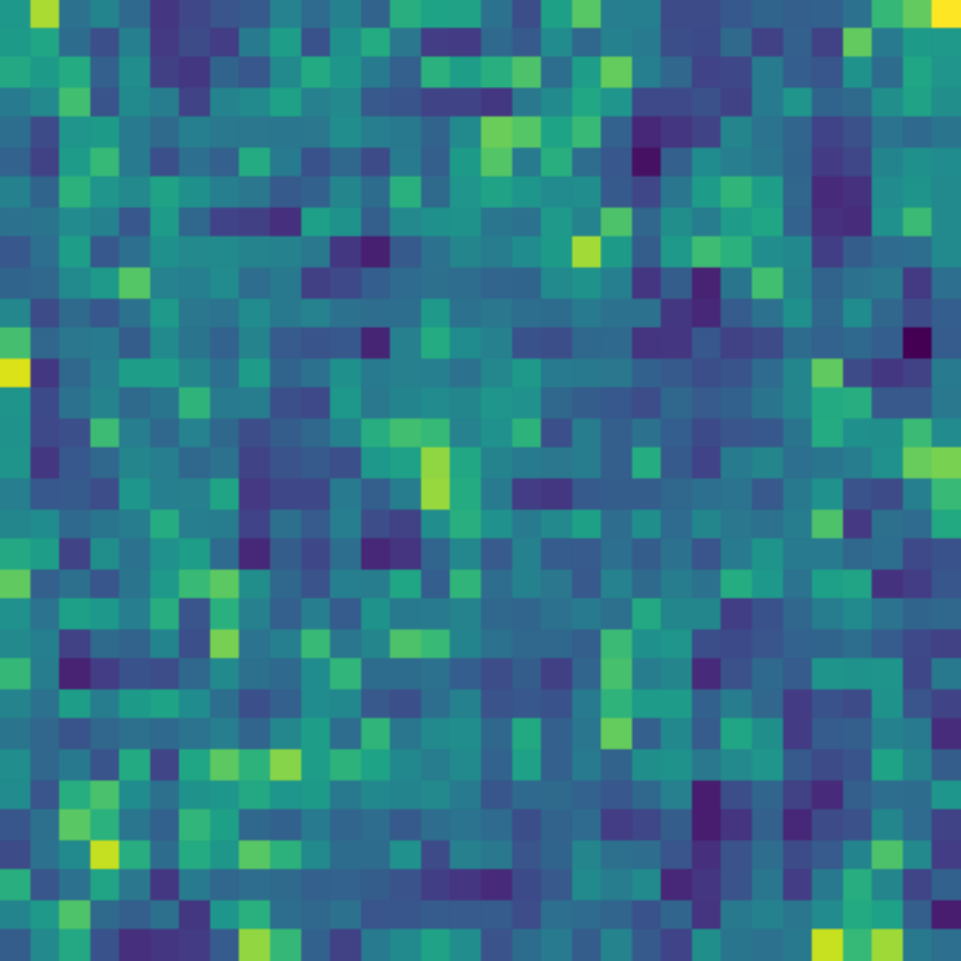}
\end{subfigure}
\begin{subfigure}{.05\textwidth}
  \centering
  \includegraphics[width=1.0\linewidth]{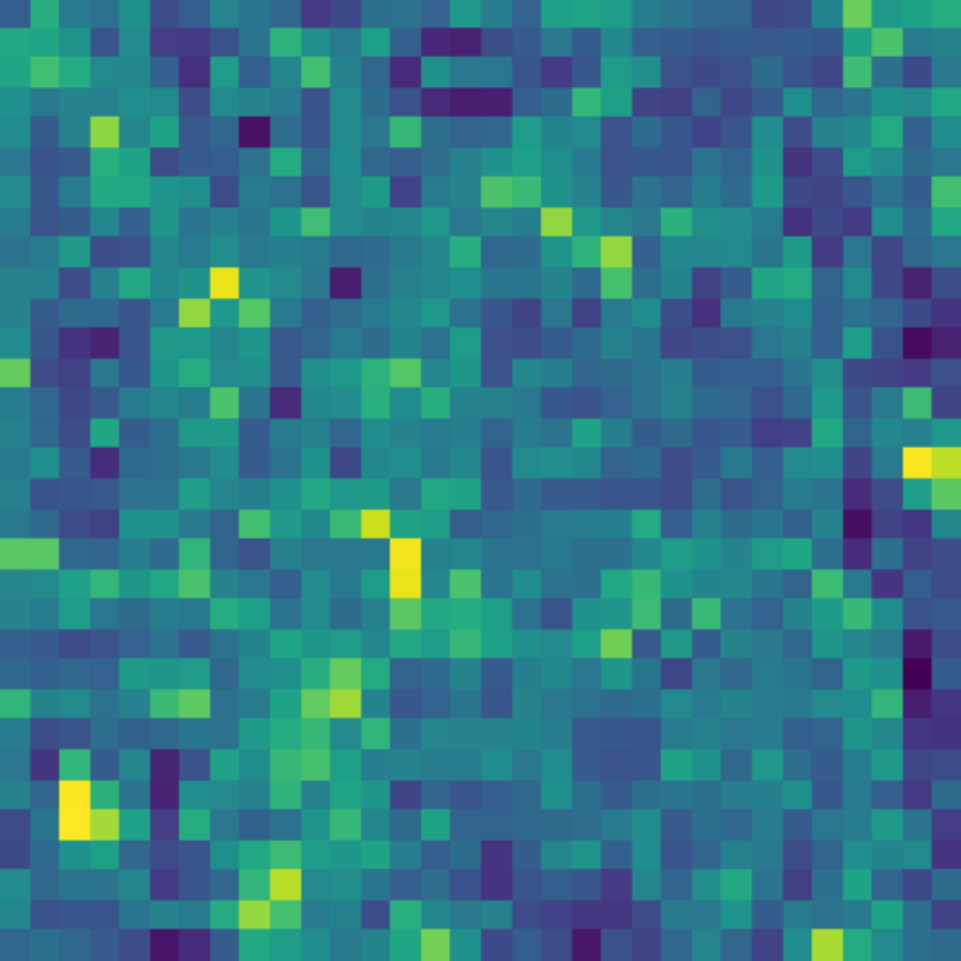}
\end{subfigure}
\begin{subfigure}{.05\textwidth}
  \centering
  \includegraphics[width=1.0\linewidth]{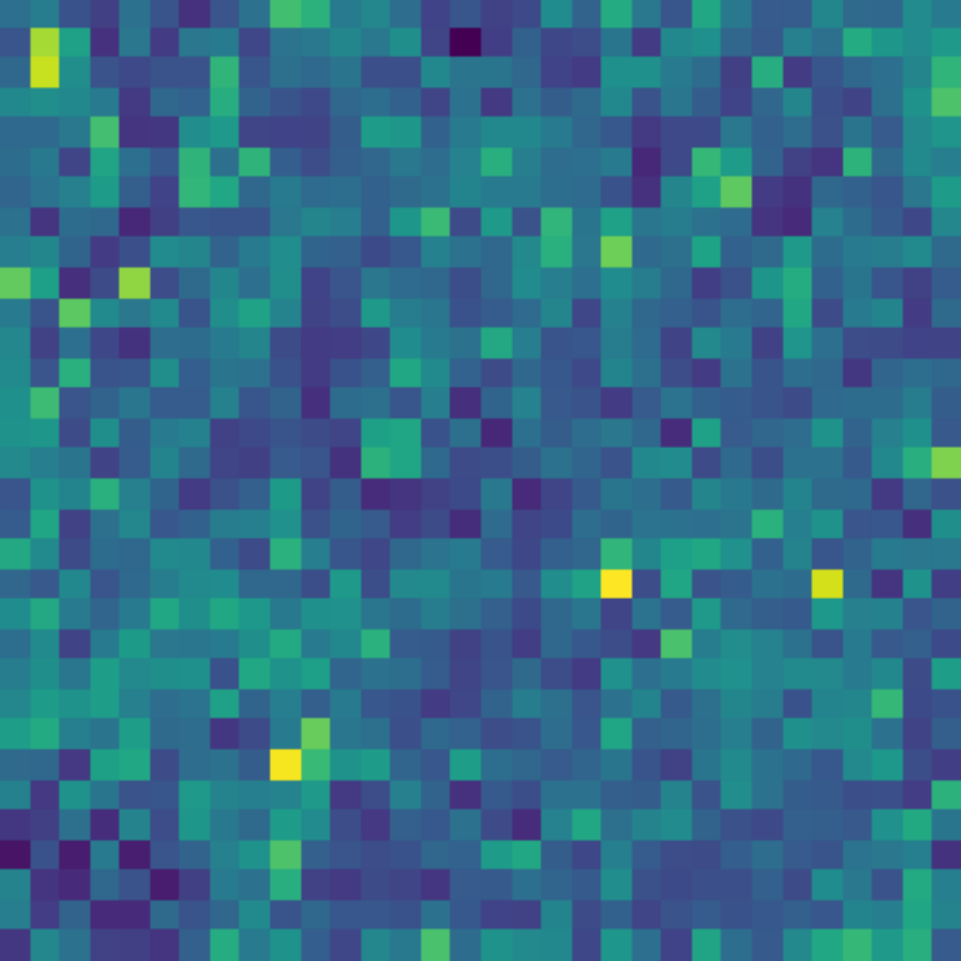}
\end{subfigure}
\\
&
\begin{subfigure}{.05\textwidth}
  \centering
  \includegraphics[width=1.0\linewidth]{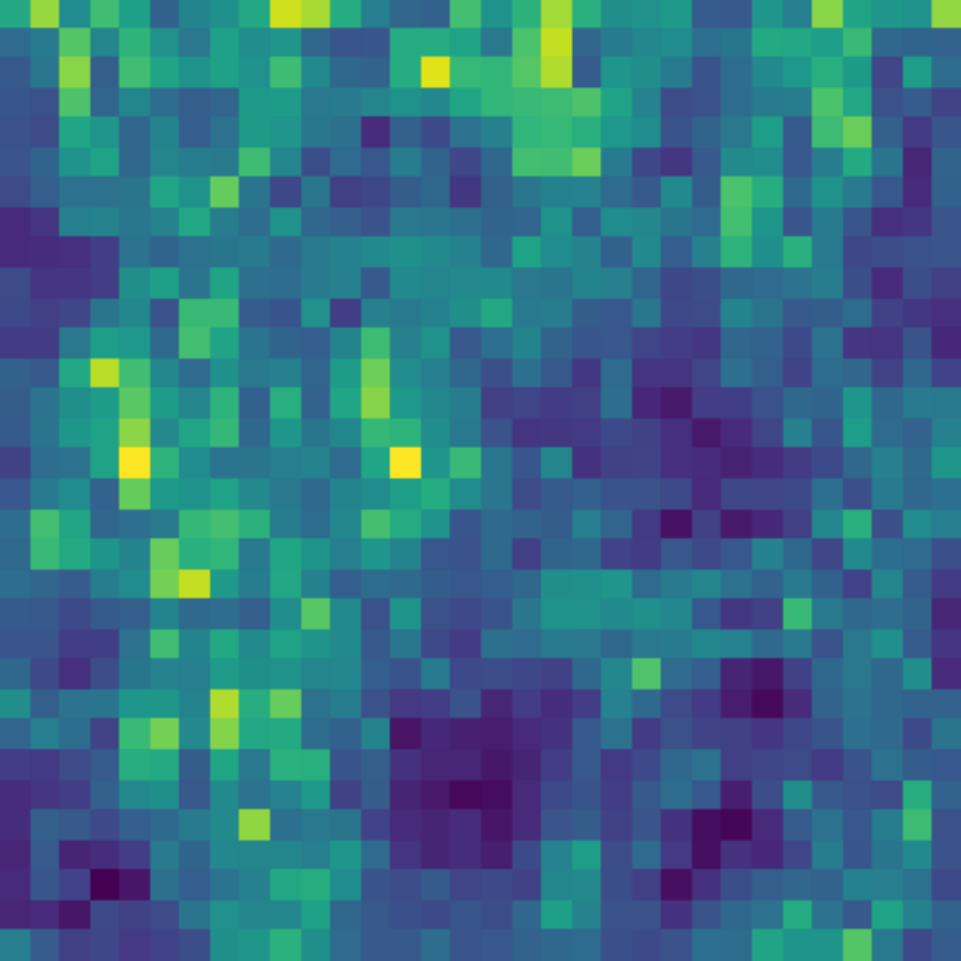}
\end{subfigure}
\begin{subfigure}{.05\textwidth}
  \centering
  \includegraphics[width=1.0\linewidth]{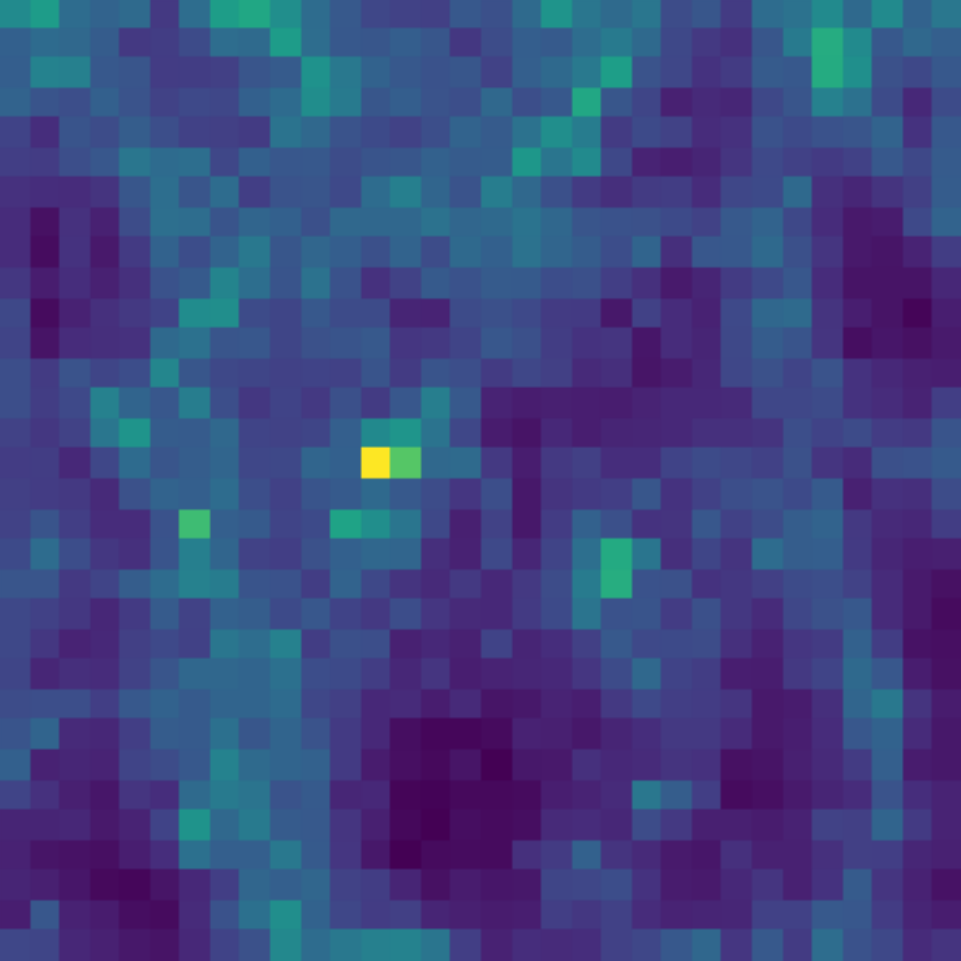}
\end{subfigure}
\begin{subfigure}{.05\textwidth}
  \centering
  \includegraphics[width=1.0\linewidth]{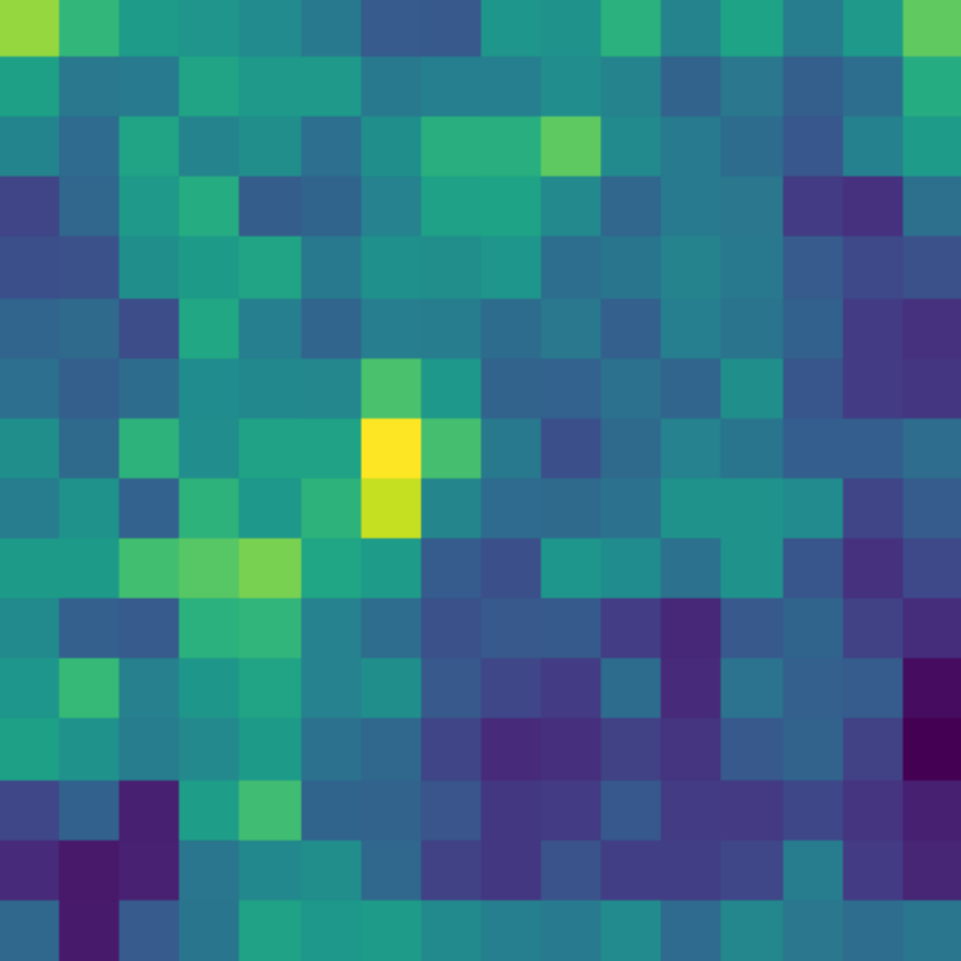}
\end{subfigure}
\begin{subfigure}{.05\textwidth}
  \centering
  \includegraphics[width=1.0\linewidth]{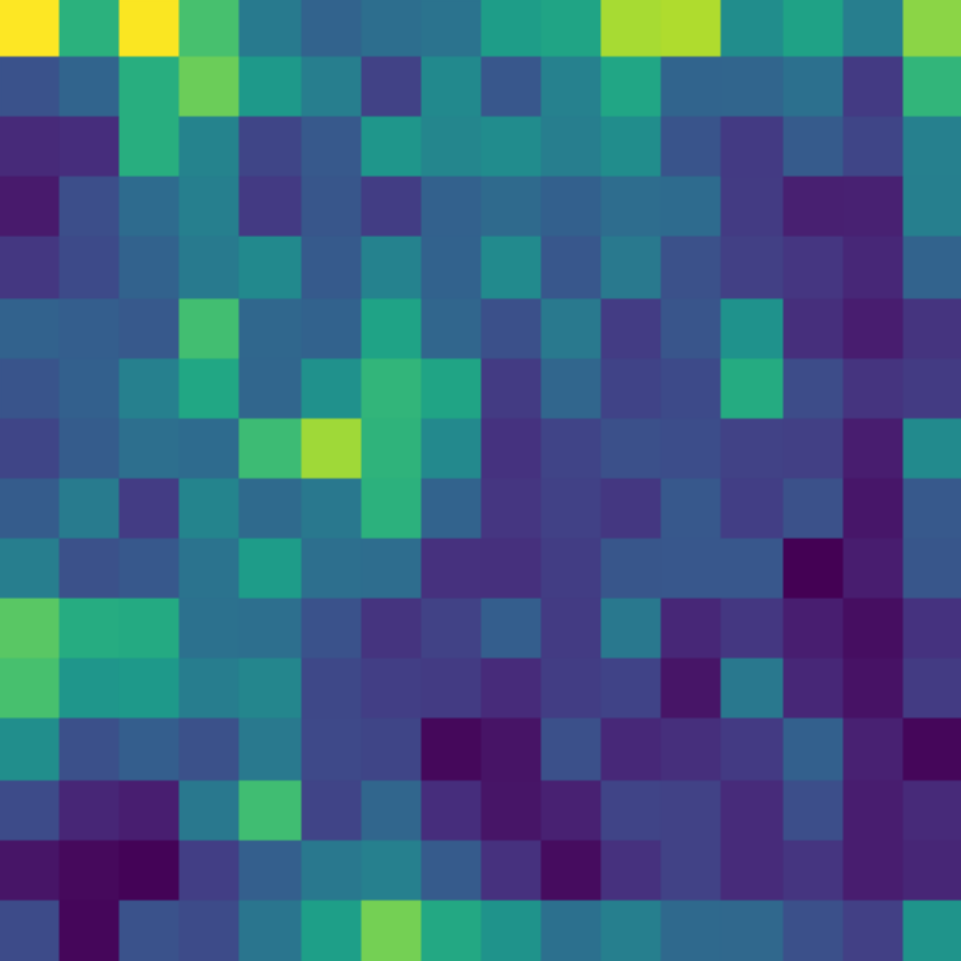}
\end{subfigure}
\begin{subfigure}{.05\textwidth}
  \centering
  \includegraphics[width=1.0\linewidth]{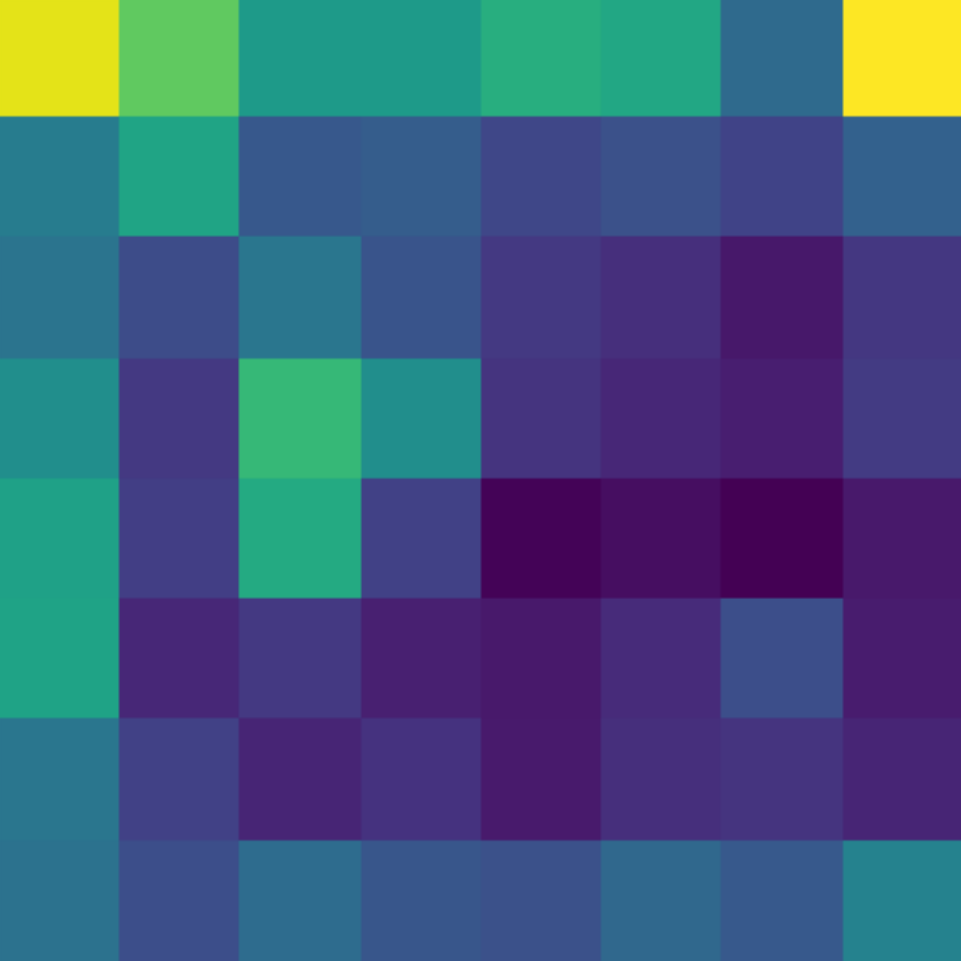}
\end{subfigure}
\begin{subfigure}{.05\textwidth}
  \centering
  \includegraphics[width=1.0\linewidth]{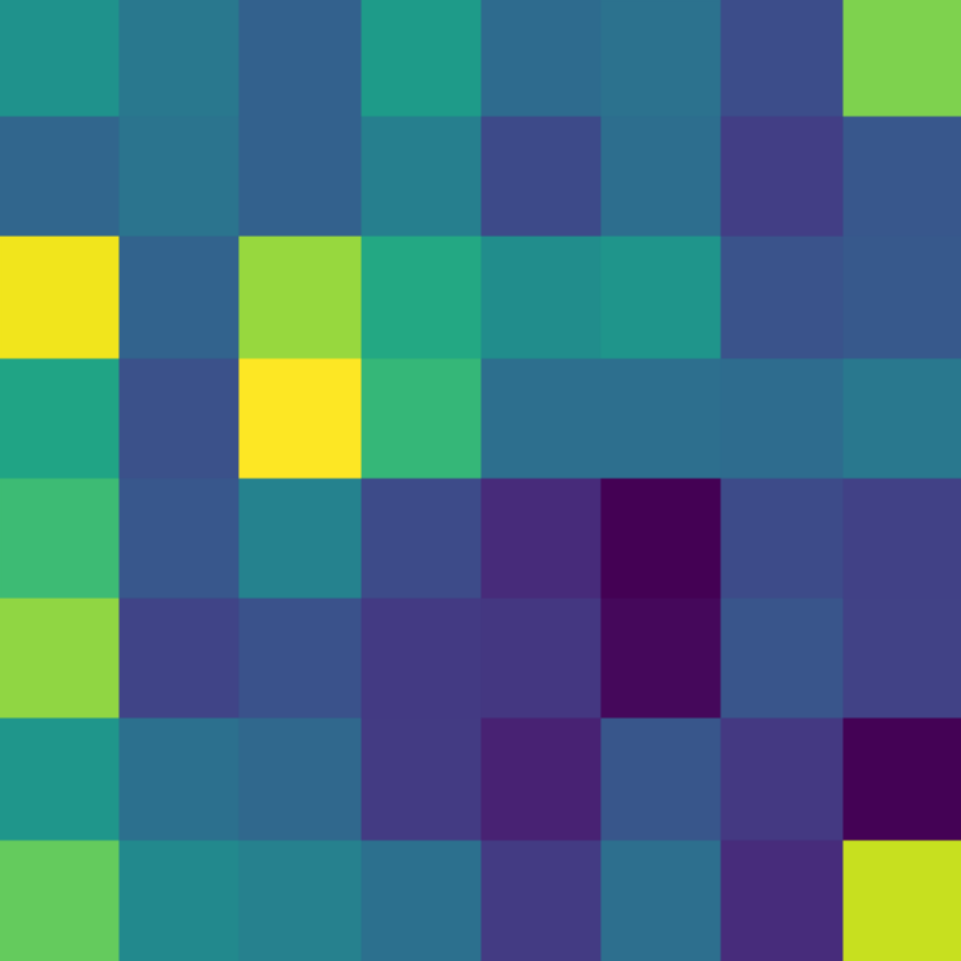}
\end{subfigure}
\begin{subfigure}{.05\textwidth}
  \centering
  \includegraphics[width=1.0\linewidth]{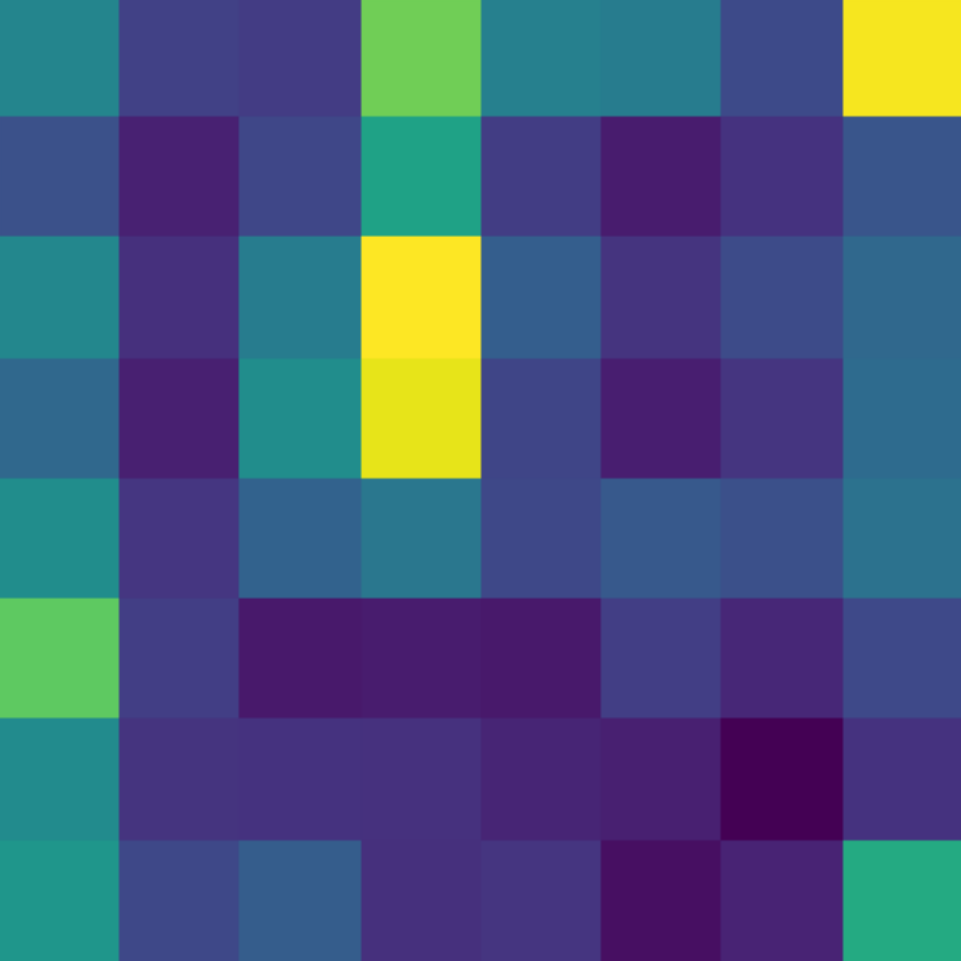}
\end{subfigure}
\begin{subfigure}{.05\textwidth}
  \centering
  \includegraphics[width=1.0\linewidth]{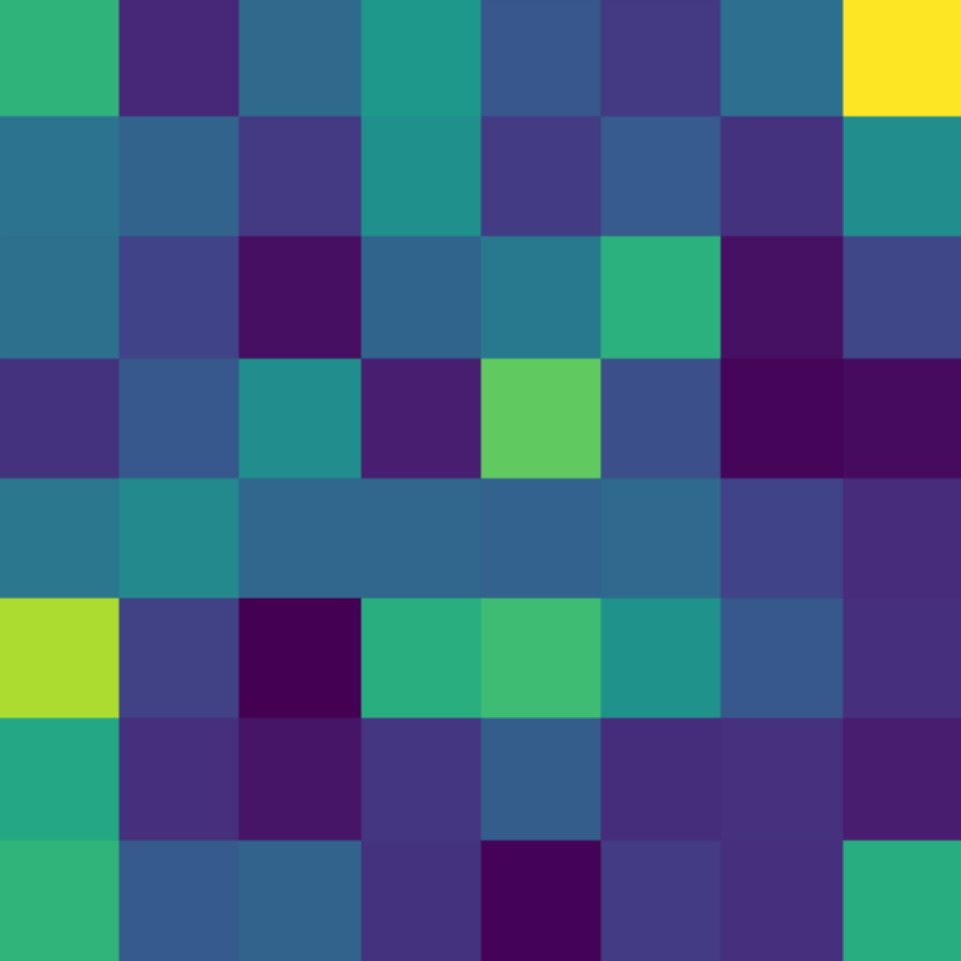}
\end{subfigure}
\begin{subfigure}{.05\textwidth}
  \centering
  \includegraphics[width=1.0\linewidth]{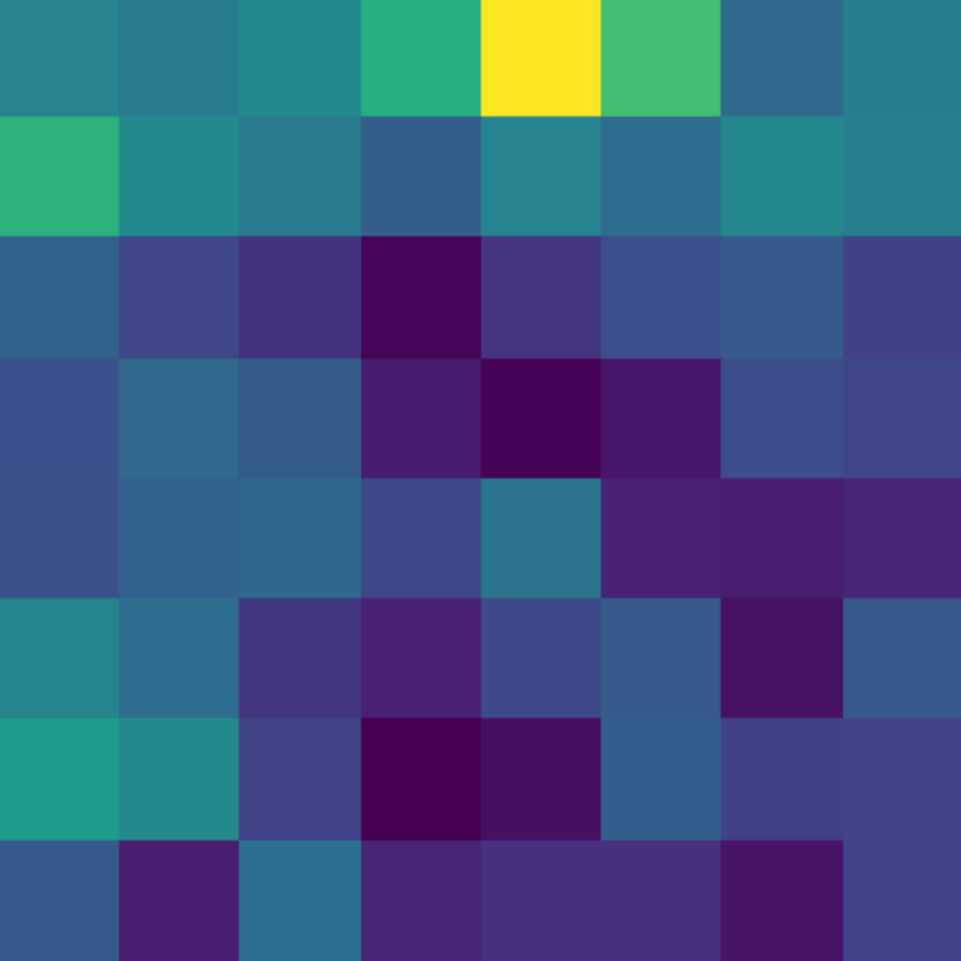}
\end{subfigure}
\begin{subfigure}{.05\textwidth}
  \centering
  \includegraphics[width=1.0\linewidth]{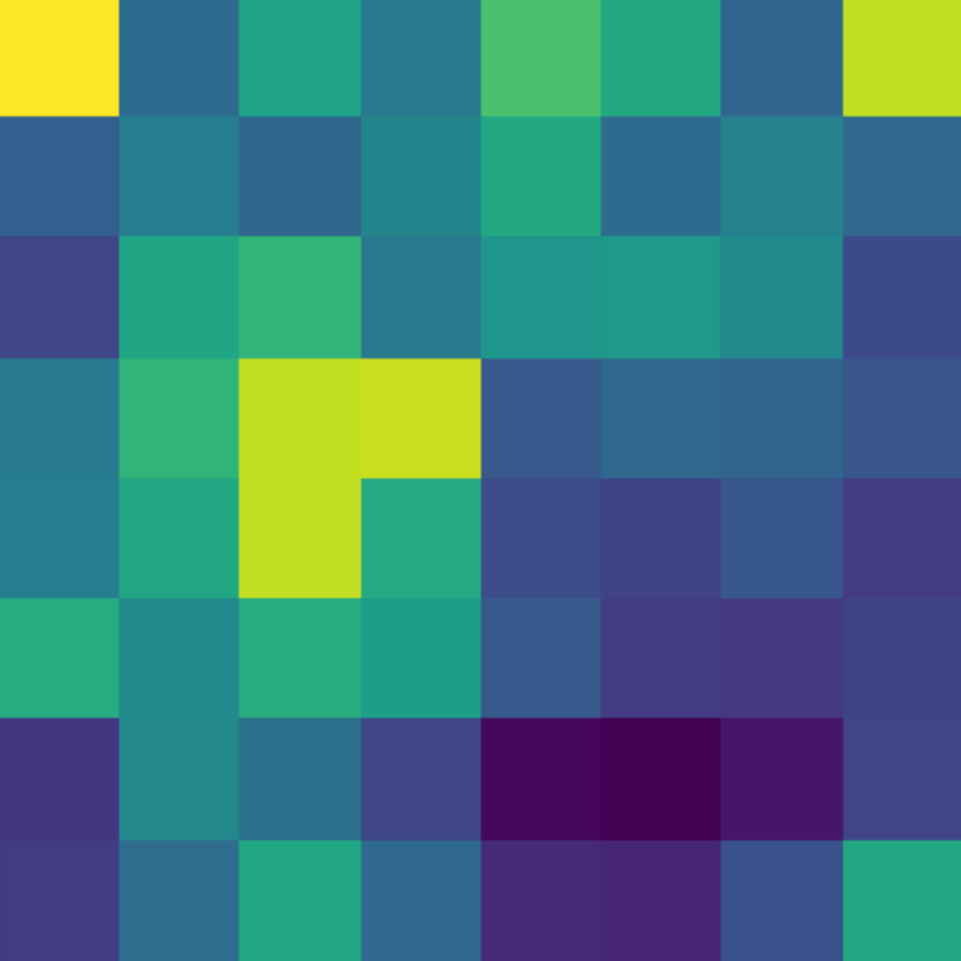}
\end{subfigure}
\begin{subfigure}{.05\textwidth}
  \centering
  \includegraphics[width=1.0\linewidth]{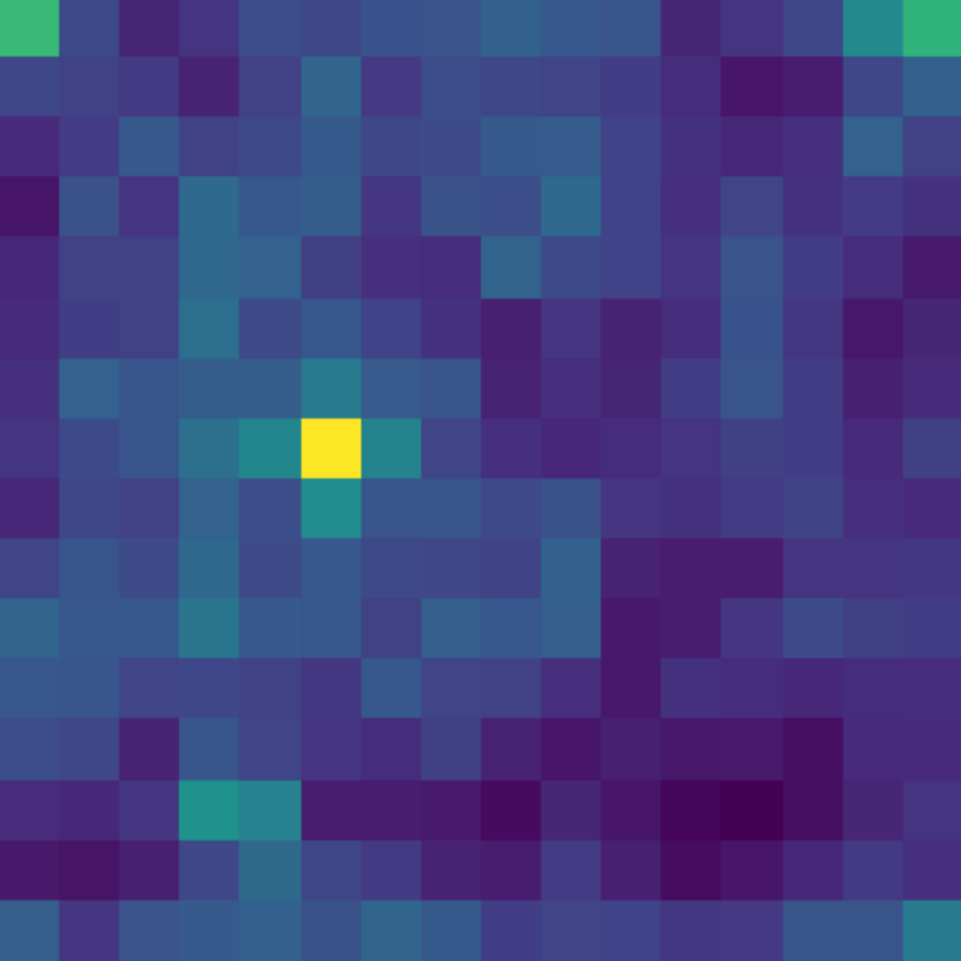}
\end{subfigure}
\begin{subfigure}{.05\textwidth}
  \centering
  \includegraphics[width=1.0\linewidth]{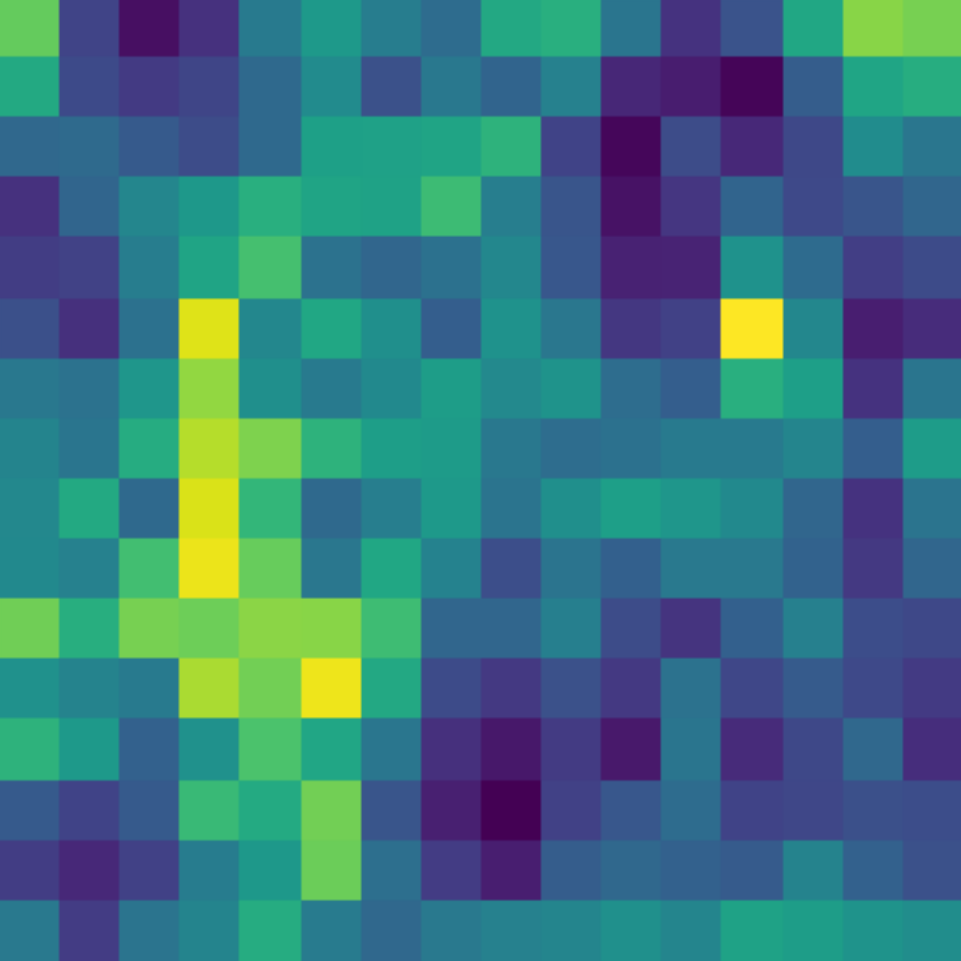}
\end{subfigure}
\begin{subfigure}{.05\textwidth}
  \centering
  \includegraphics[width=1.0\linewidth]{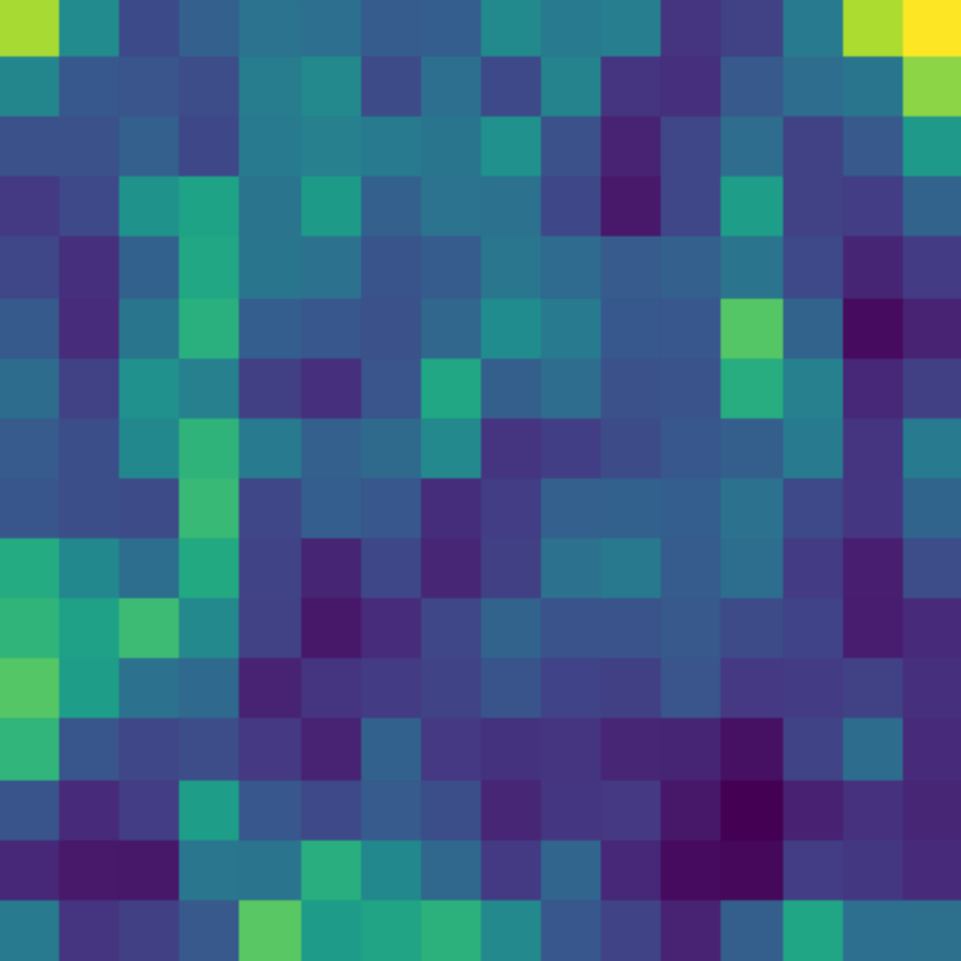}
\end{subfigure}
\begin{subfigure}{.05\textwidth}
  \centering
  \includegraphics[width=1.0\linewidth]{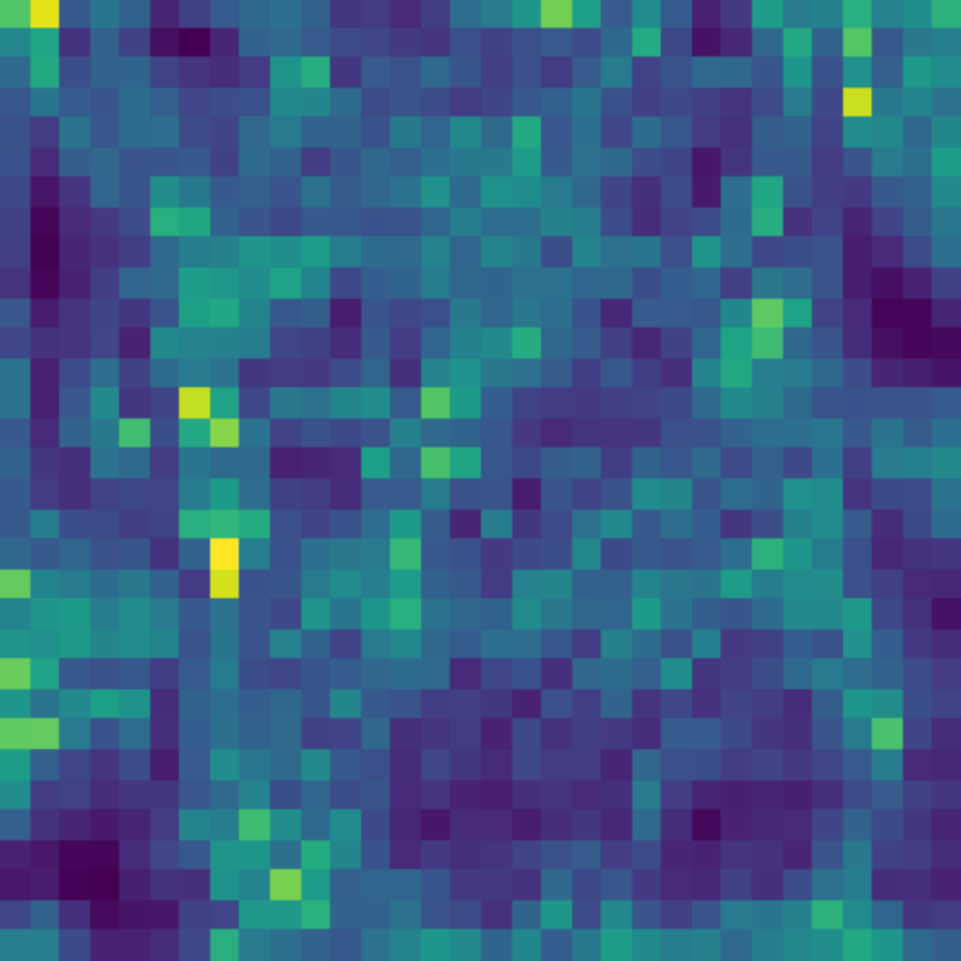}
\end{subfigure}
\begin{subfigure}{.05\textwidth}
  \centering
  \includegraphics[width=1.0\linewidth]{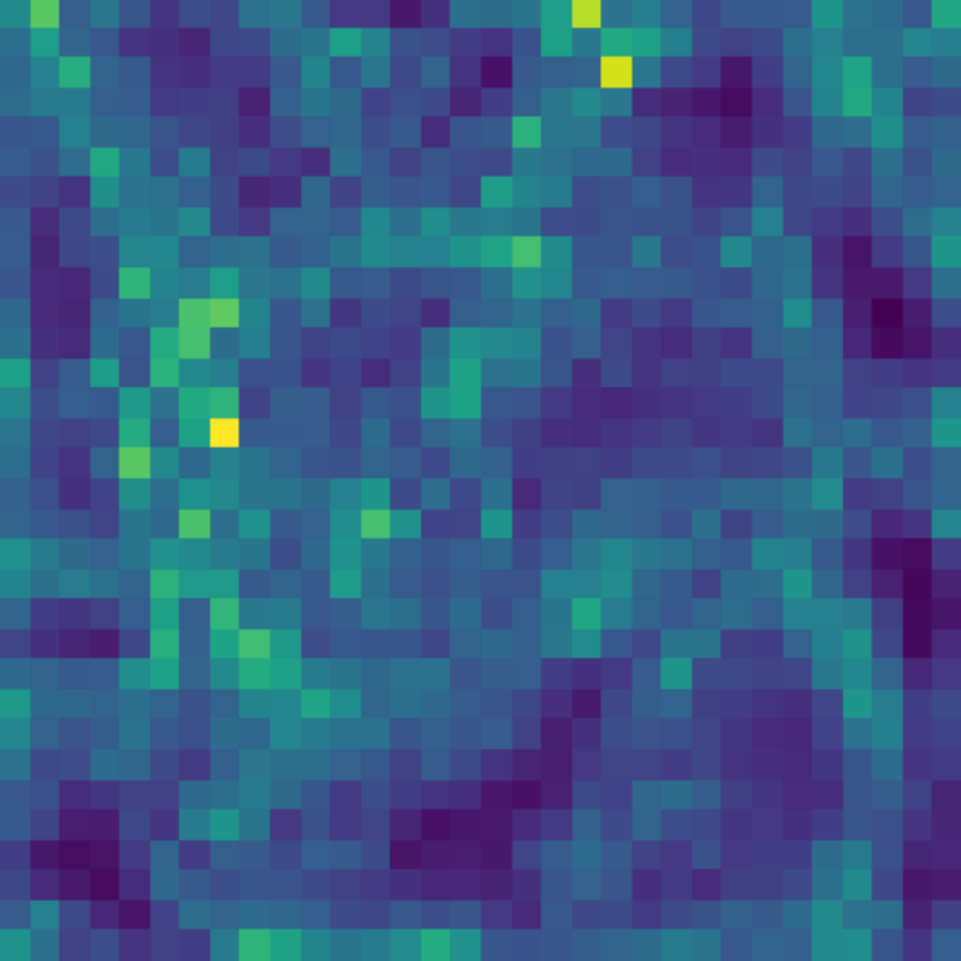}
\end{subfigure}
\begin{subfigure}{.05\textwidth}
  \centering
  \includegraphics[width=1.0\linewidth]{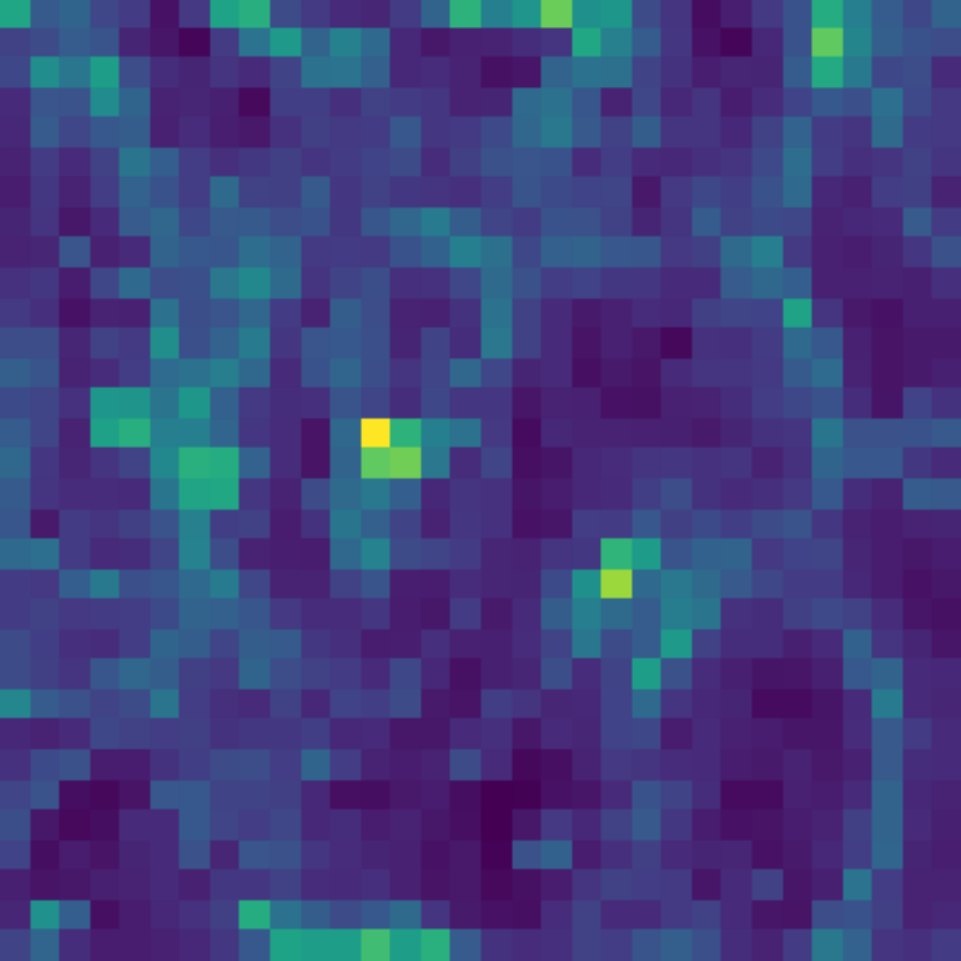}
\end{subfigure}
\\
&
\begin{subfigure}{.05\textwidth}
  \centering
  \includegraphics[width=1.0\linewidth]{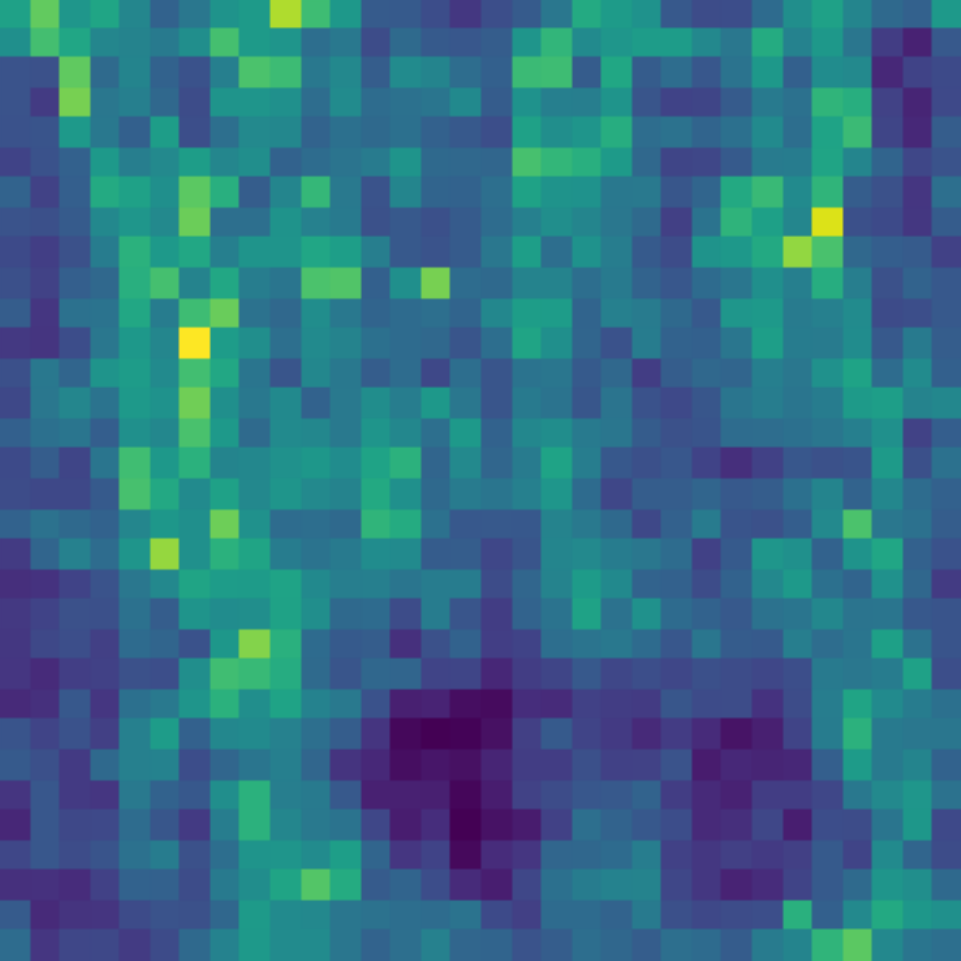}
\end{subfigure}
\begin{subfigure}{.05\textwidth}
  \centering
  \includegraphics[width=1.0\linewidth]{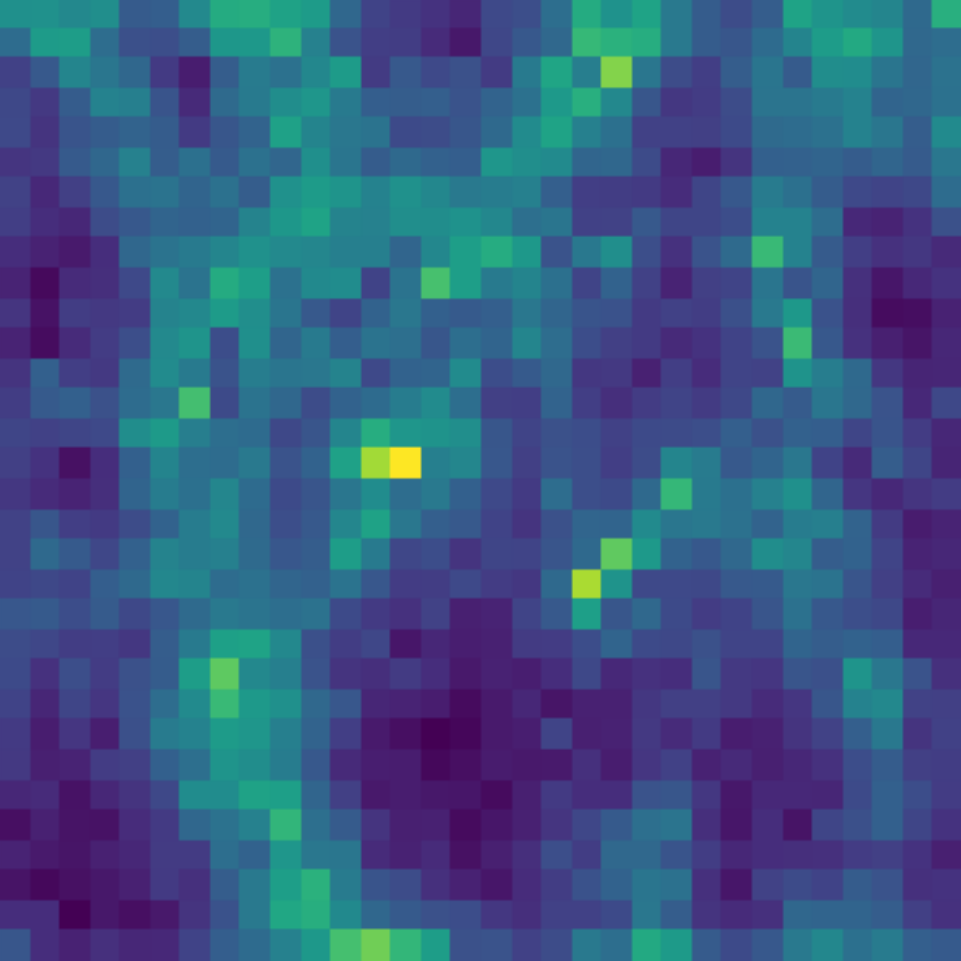}
\end{subfigure}
\begin{subfigure}{.05\textwidth}
  \centering
  \includegraphics[width=1.0\linewidth]{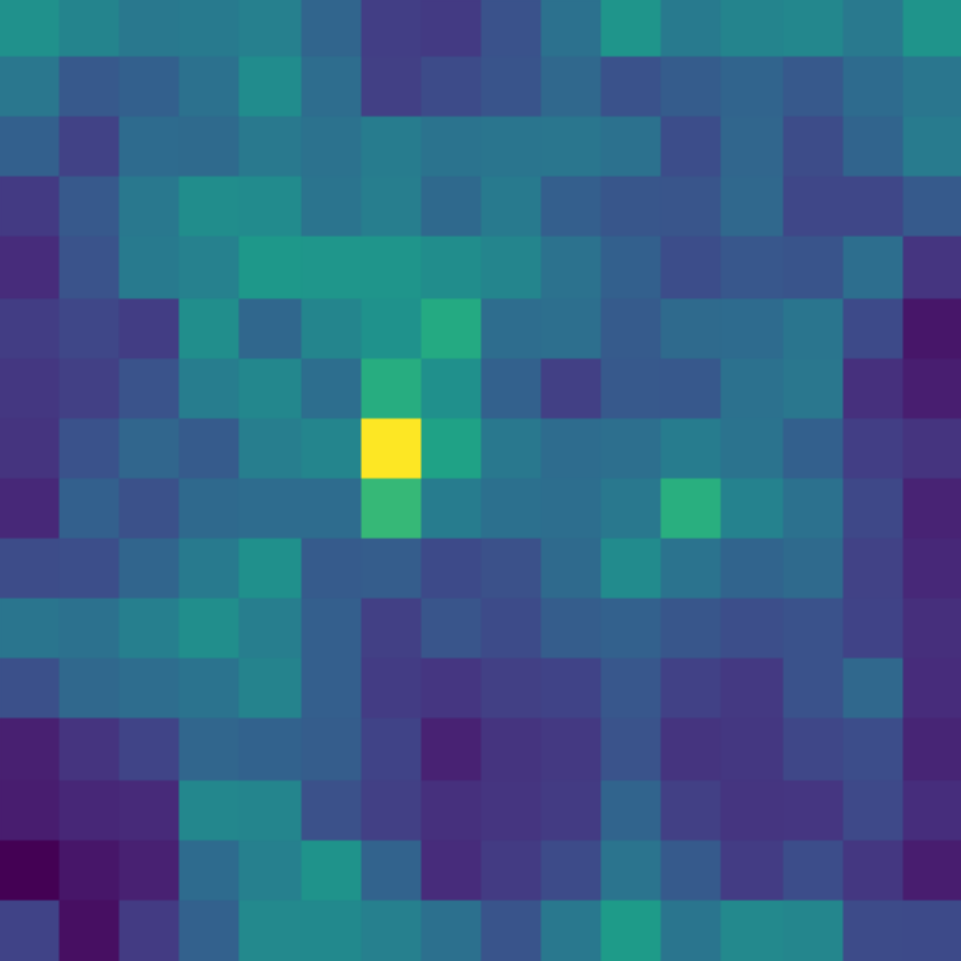}
\end{subfigure}
\begin{subfigure}{.05\textwidth}
  \centering
  \includegraphics[width=1.0\linewidth]{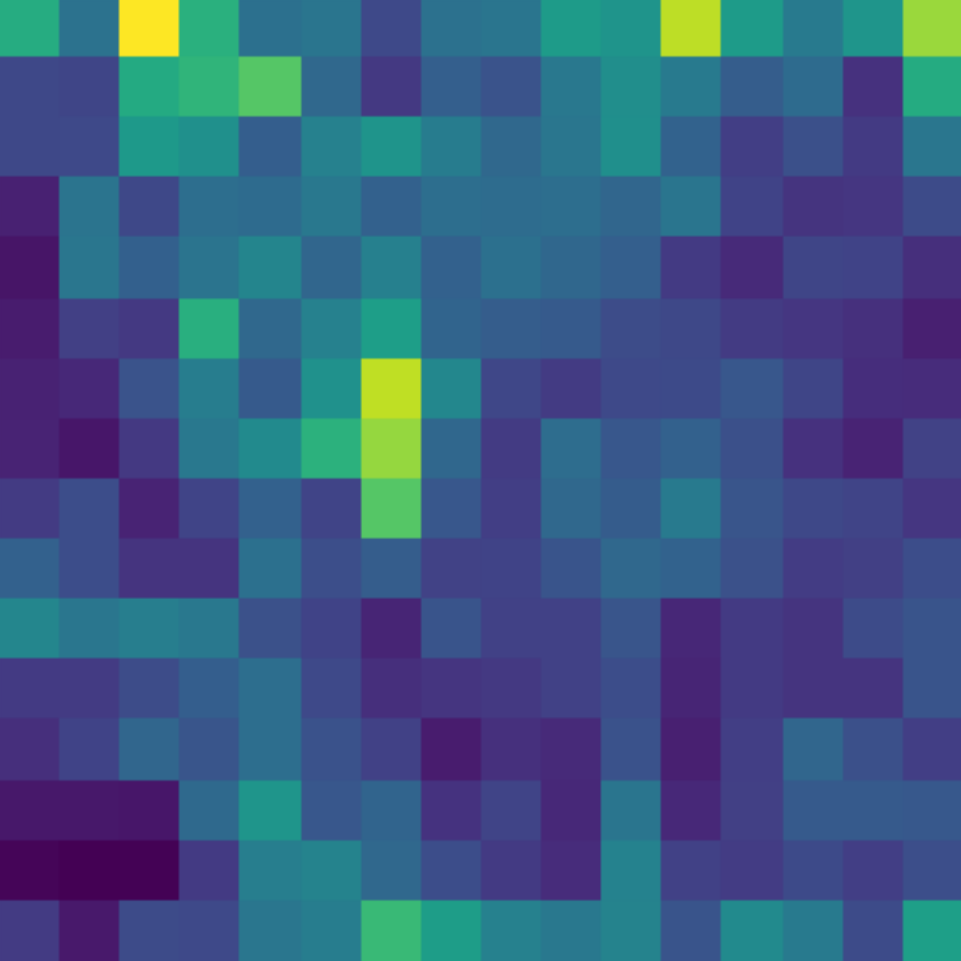}
\end{subfigure}
\begin{subfigure}{.05\textwidth}
  \centering
  \includegraphics[width=1.0\linewidth]{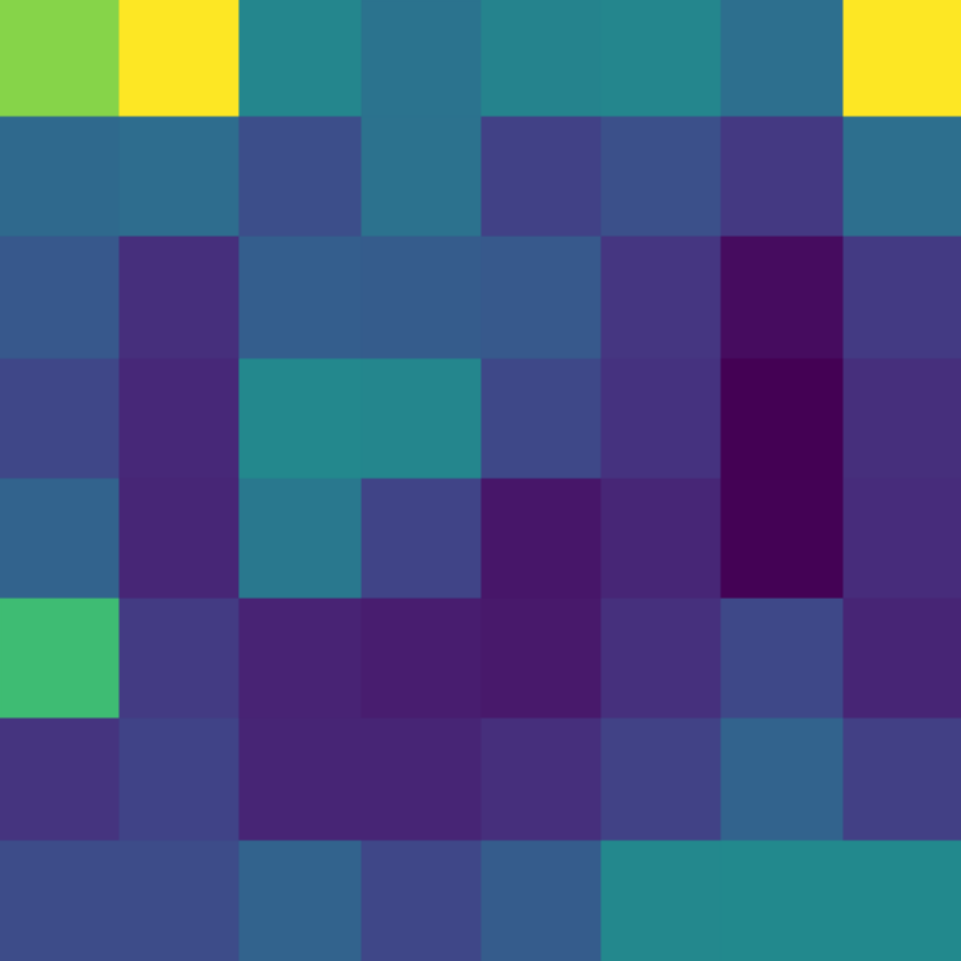}
\end{subfigure}
\begin{subfigure}{.05\textwidth}
  \centering
  \includegraphics[width=1.0\linewidth]{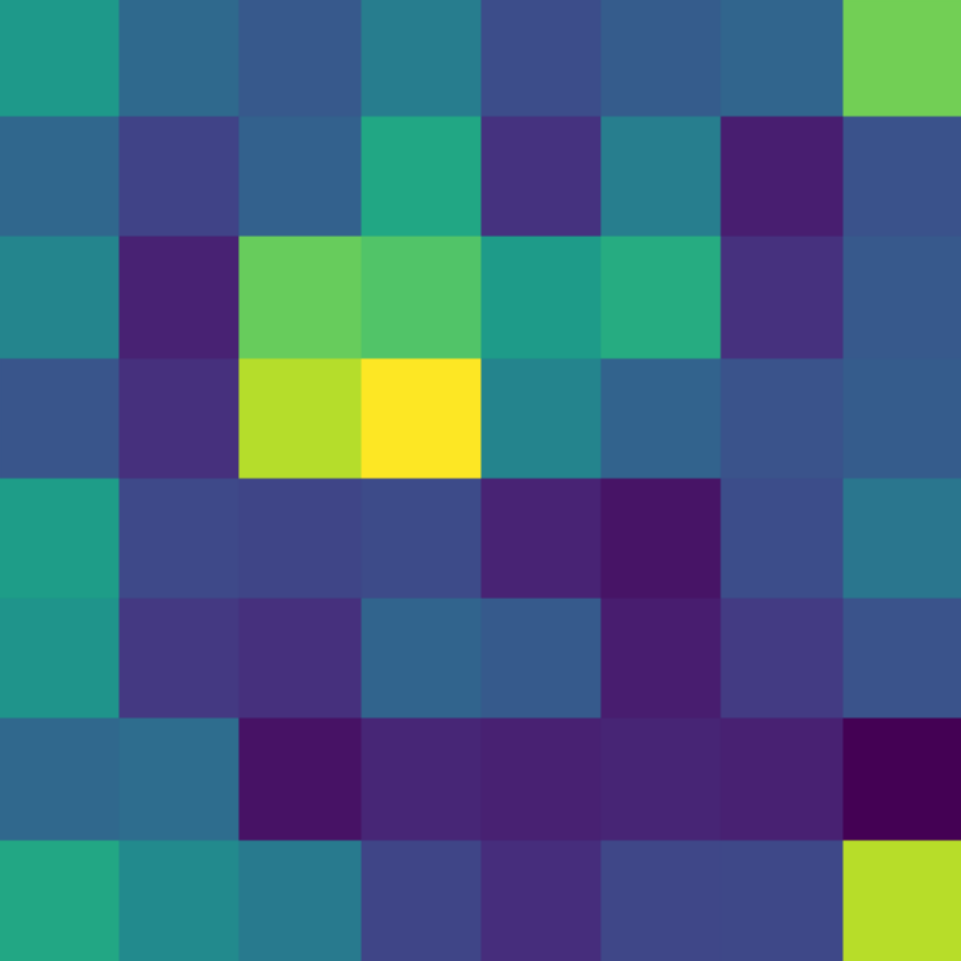}
\end{subfigure}
\begin{subfigure}{.05\textwidth}
  \centering
  \includegraphics[width=1.0\linewidth]{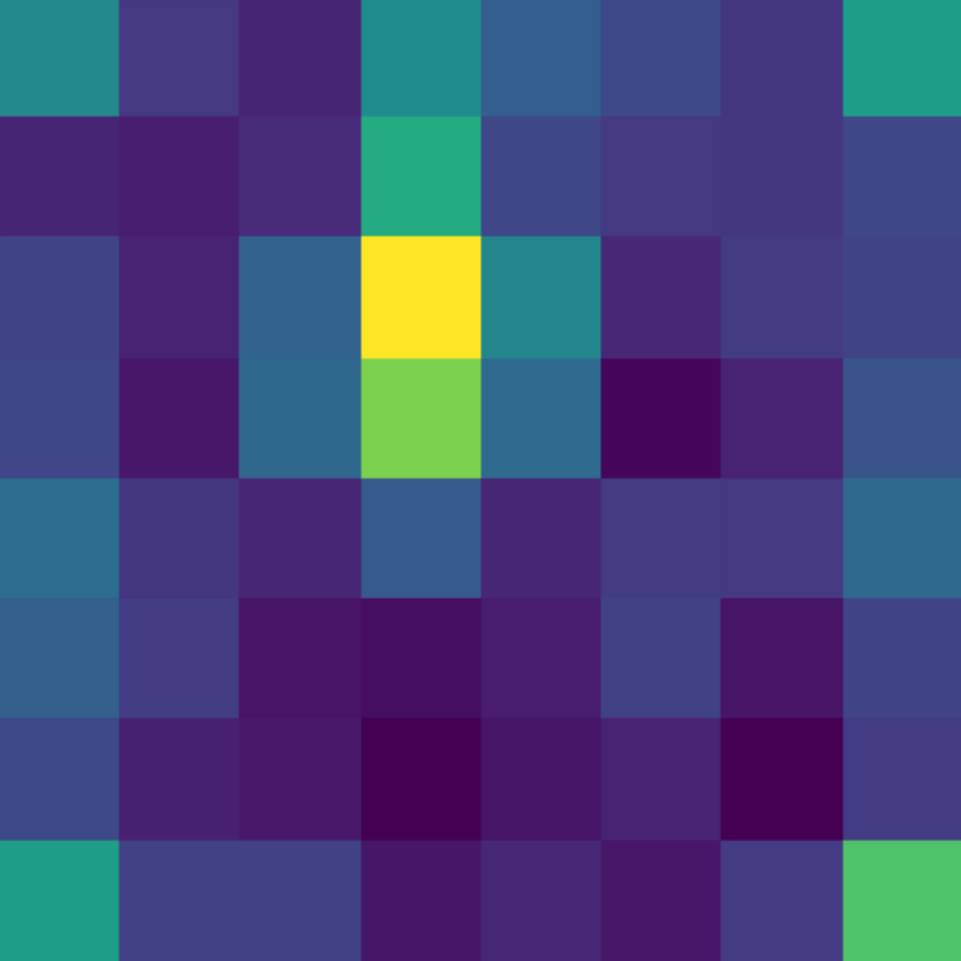}
\end{subfigure}
\begin{subfigure}{.05\textwidth}
  \centering
  \includegraphics[width=1.0\linewidth]{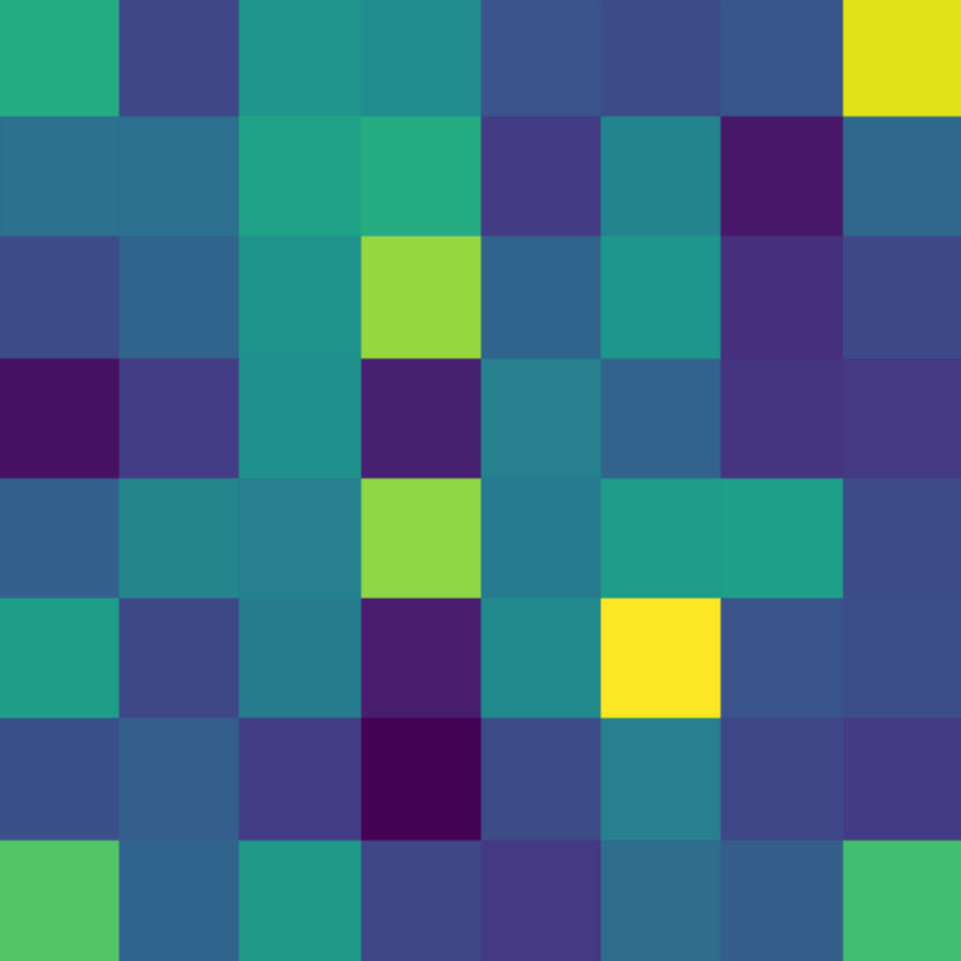}
\end{subfigure}
\begin{subfigure}{.05\textwidth}
  \centering
  \includegraphics[width=1.0\linewidth]{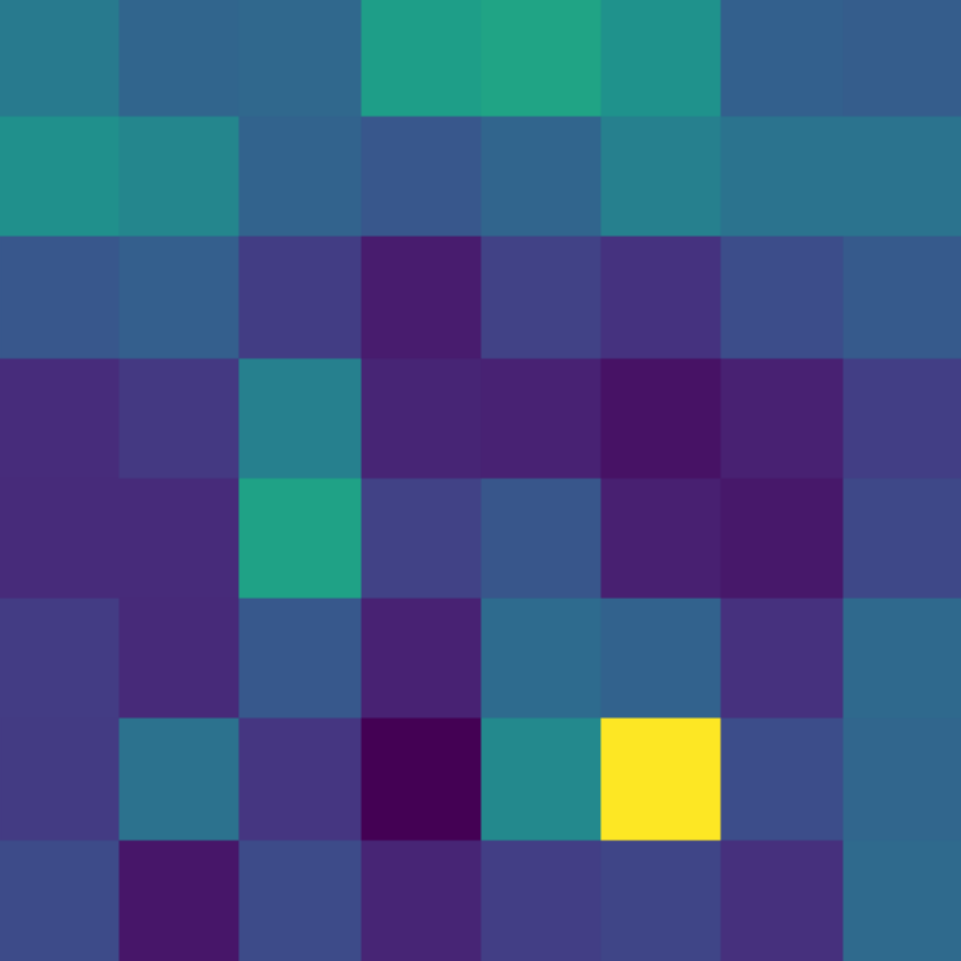}
\end{subfigure}
\begin{subfigure}{.05\textwidth}
  \centering
  \includegraphics[width=1.0\linewidth]{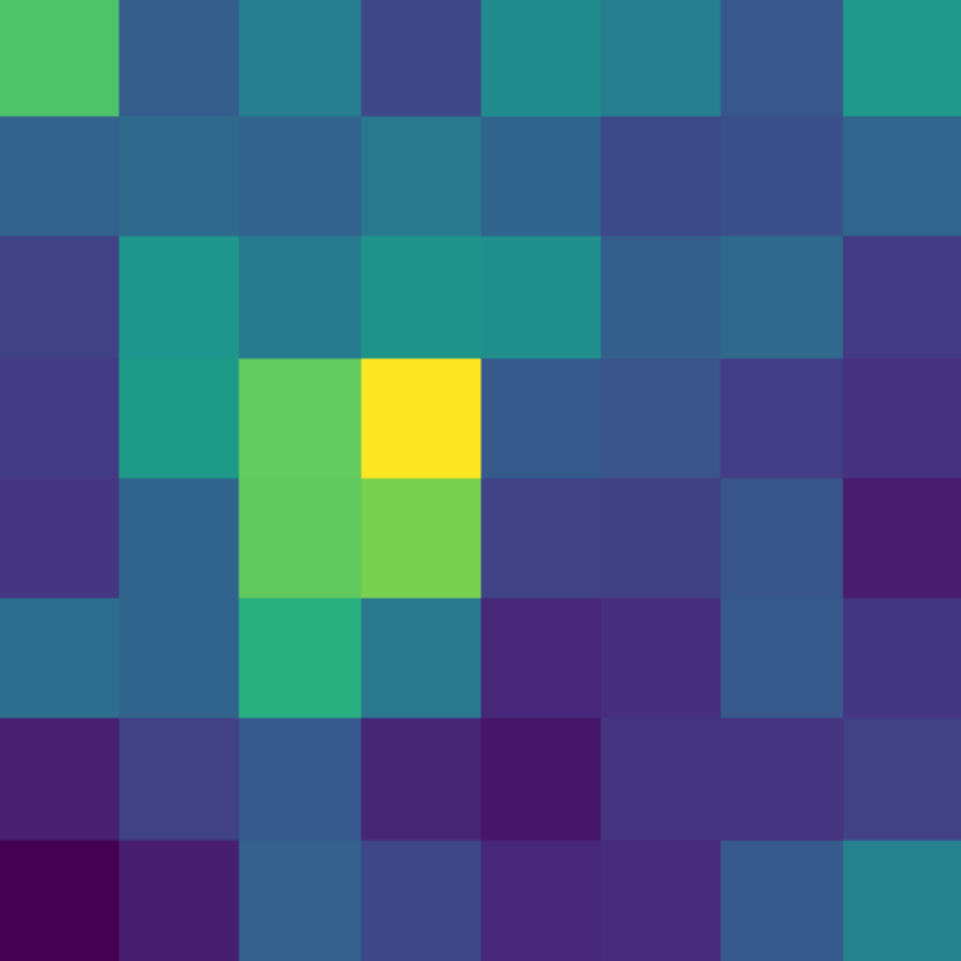}
\end{subfigure}
\begin{subfigure}{.05\textwidth}
  \centering
  \includegraphics[width=1.0\linewidth]{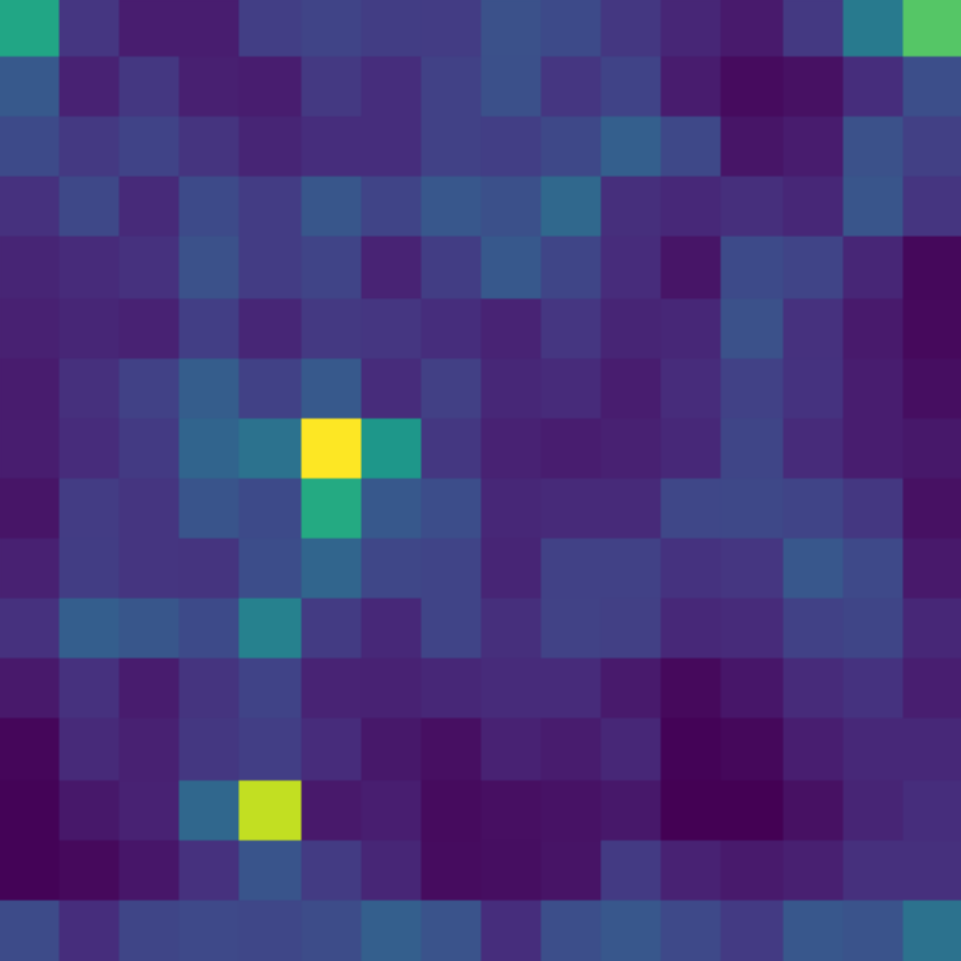}
\end{subfigure}
\begin{subfigure}{.05\textwidth}
  \centering
  \includegraphics[width=1.0\linewidth]{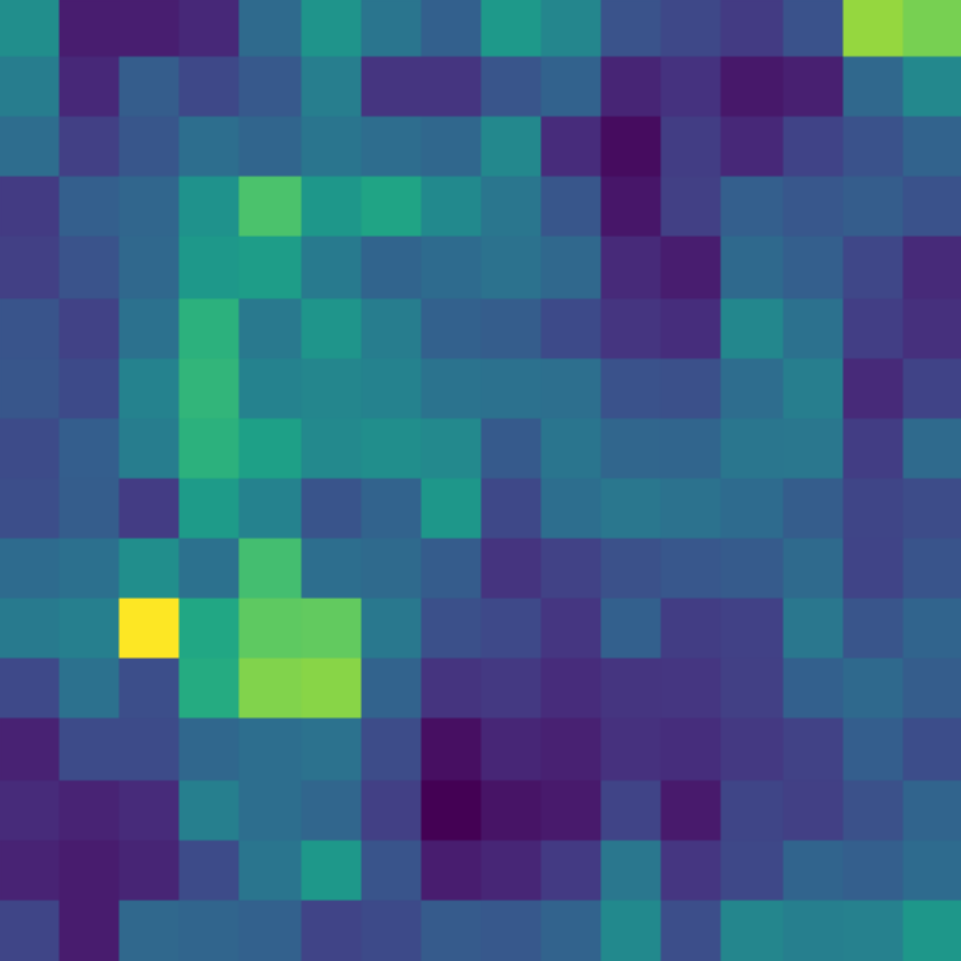}
\end{subfigure}
\begin{subfigure}{.05\textwidth}
  \centering
  \includegraphics[width=1.0\linewidth]{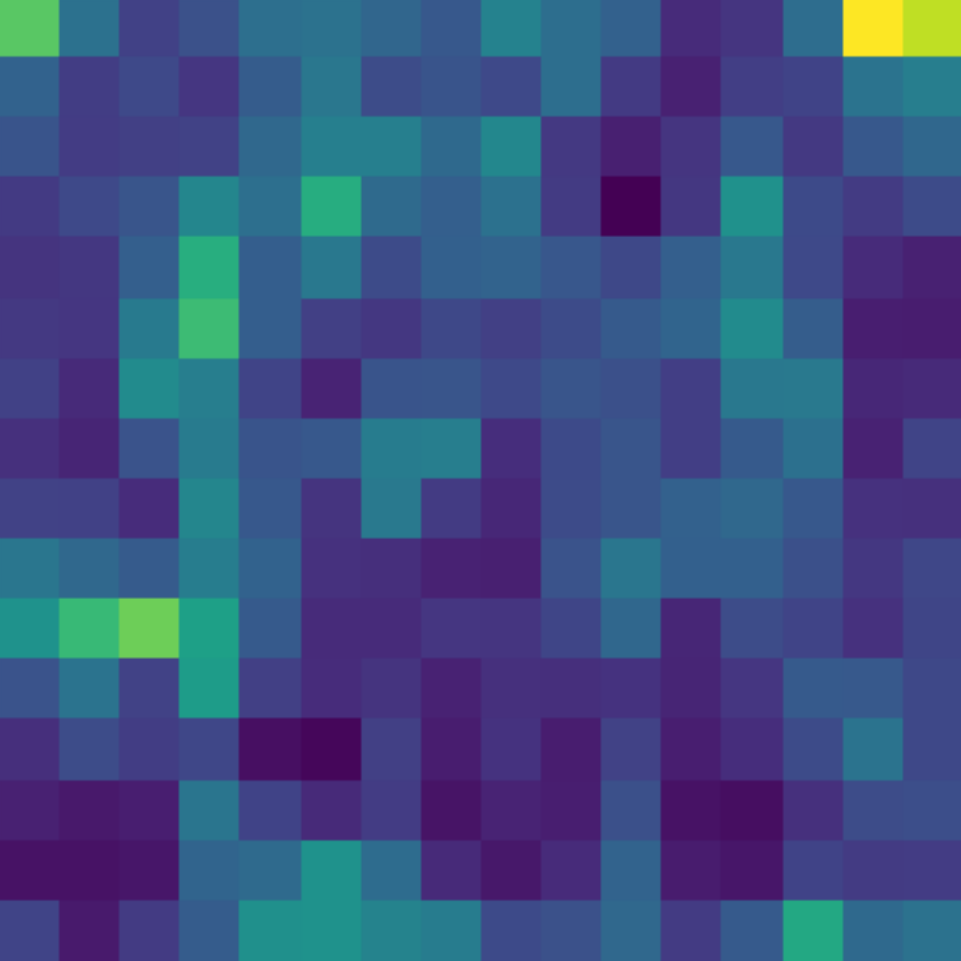}
\end{subfigure}
\begin{subfigure}{.05\textwidth}
  \centering
  \includegraphics[width=1.0\linewidth]{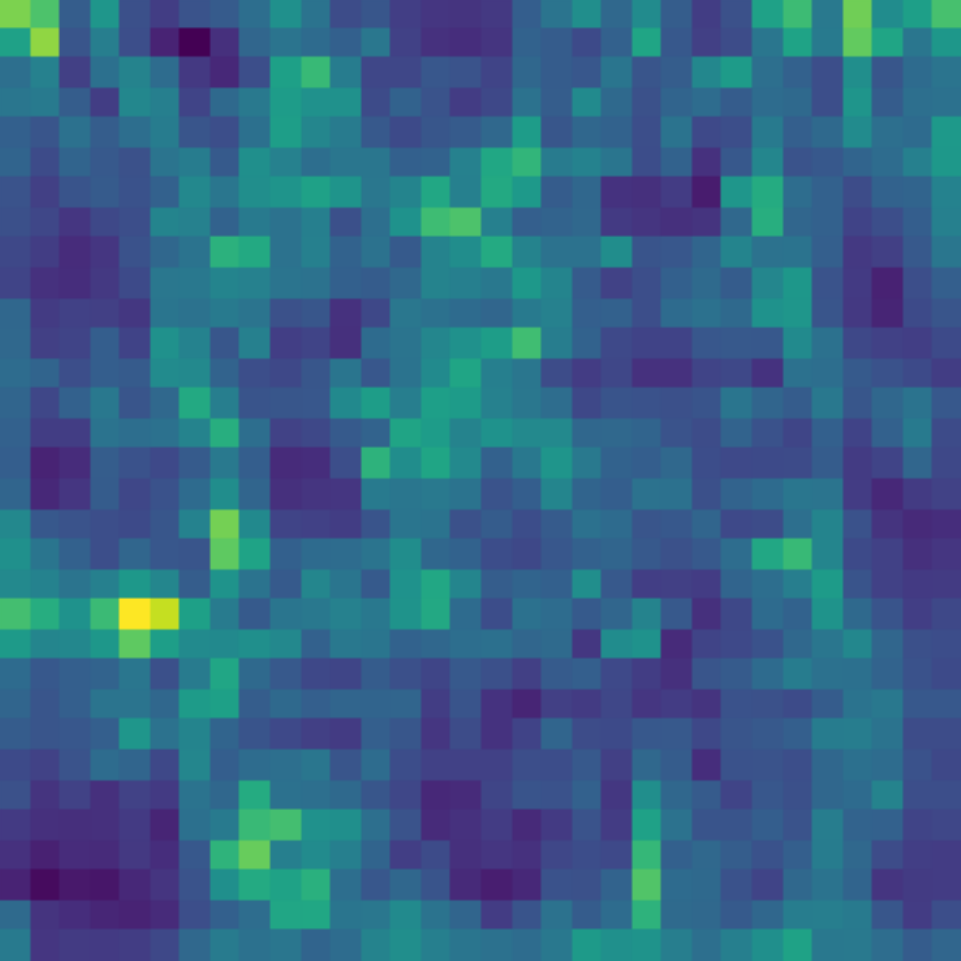}
\end{subfigure}
\begin{subfigure}{.05\textwidth}
  \centering
  \includegraphics[width=1.0\linewidth]{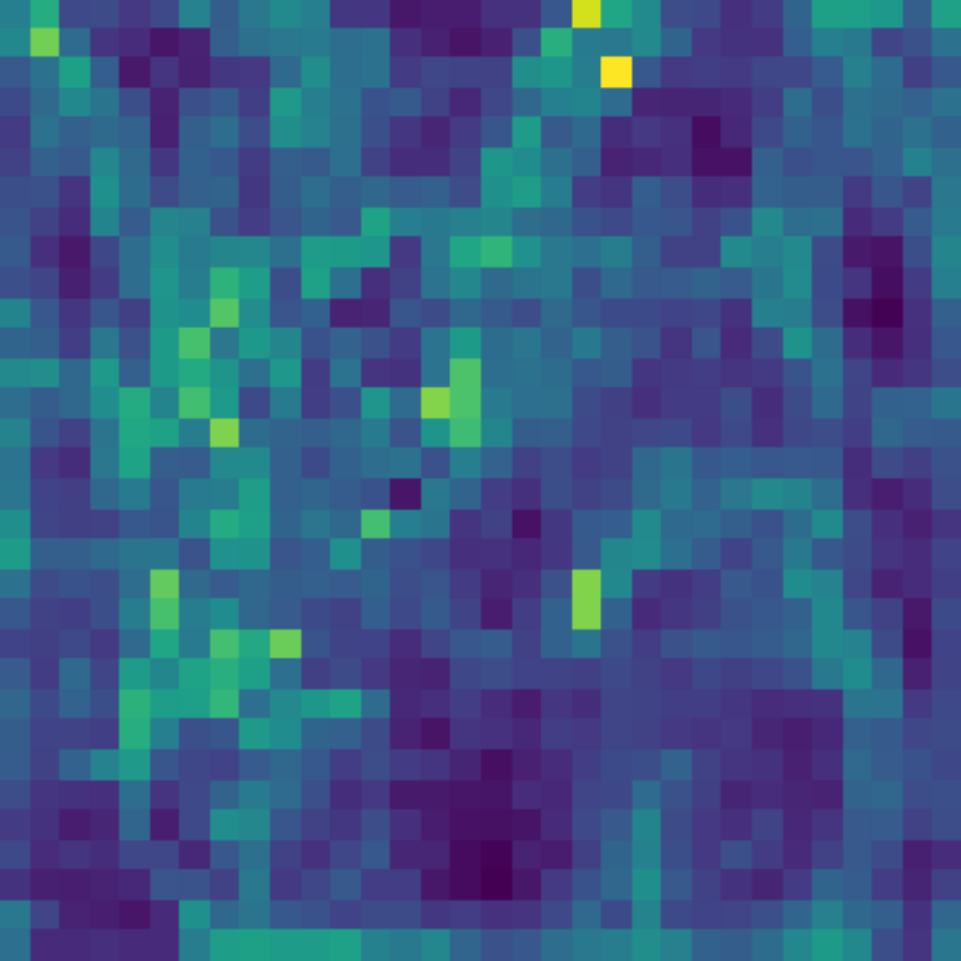}
\end{subfigure}
\begin{subfigure}{.05\textwidth}
  \centering
  \includegraphics[width=1.0\linewidth]{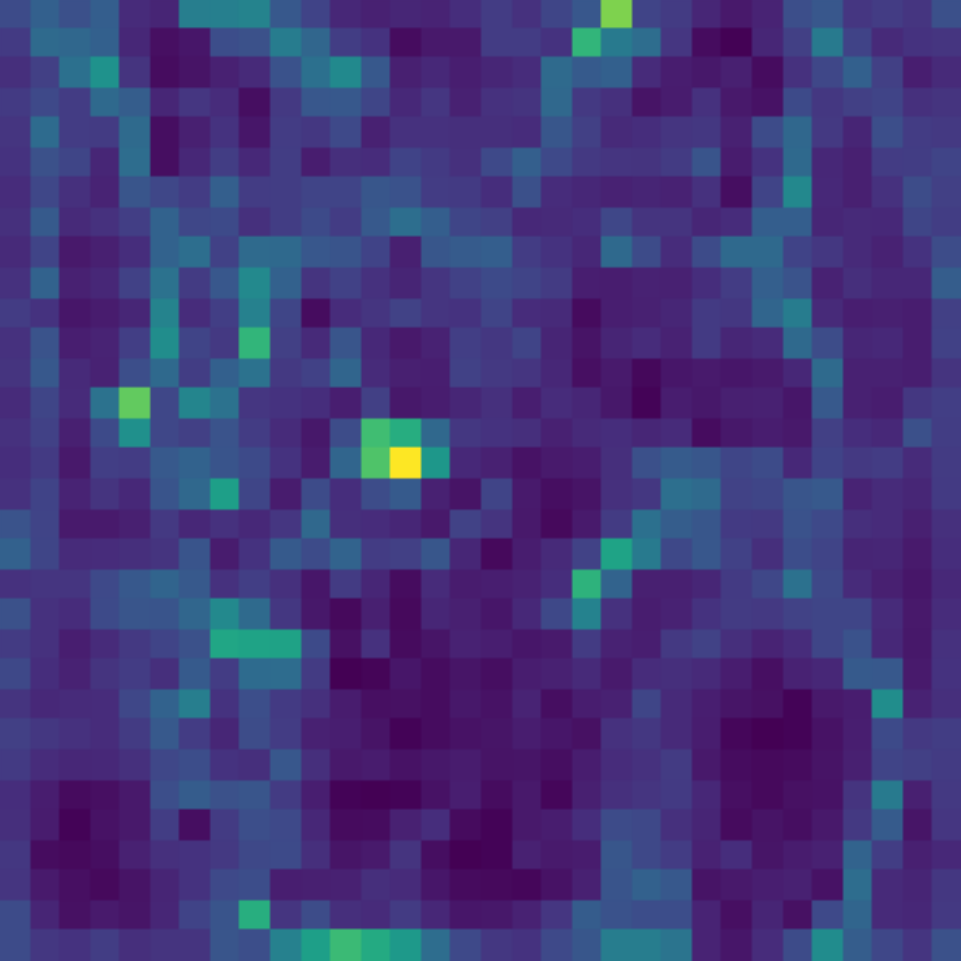}
\end{subfigure}
\\
&
\begin{subfigure}{.05\textwidth}
  \centering
  \includegraphics[width=1.0\linewidth]{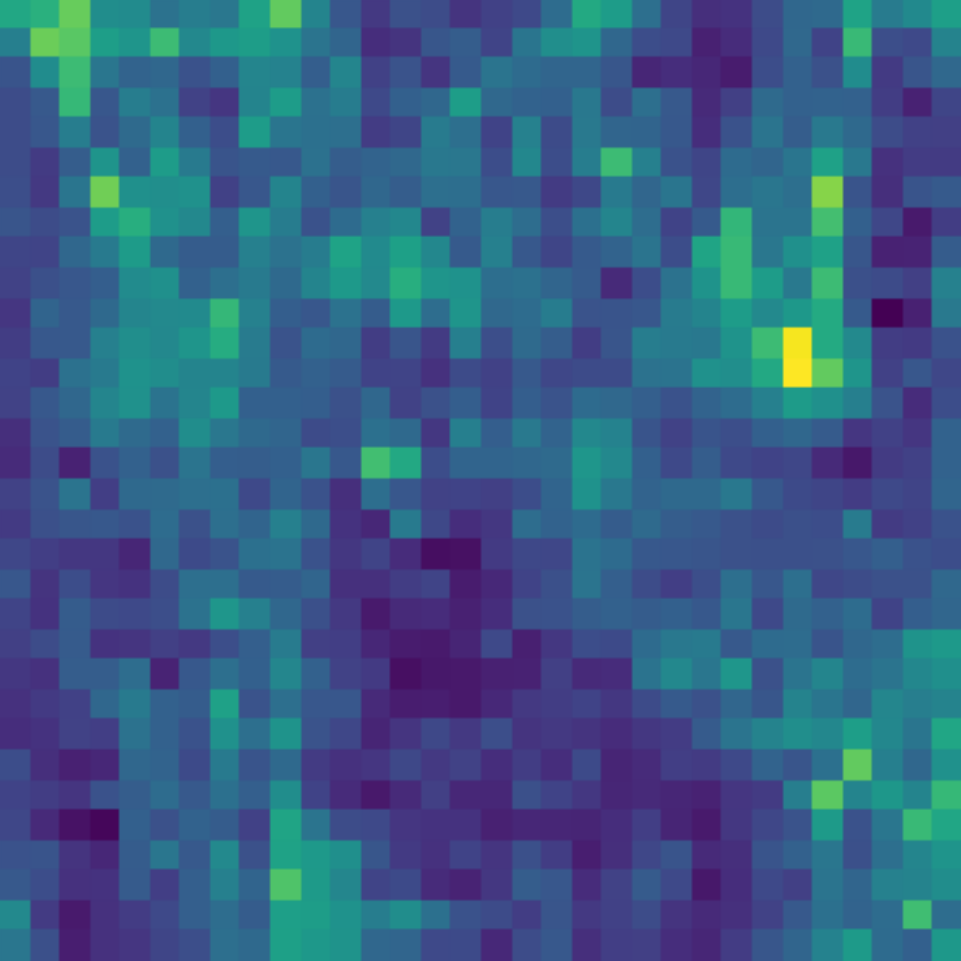}
\end{subfigure}
\begin{subfigure}{.05\textwidth}
  \centering
  \includegraphics[width=1.0\linewidth]{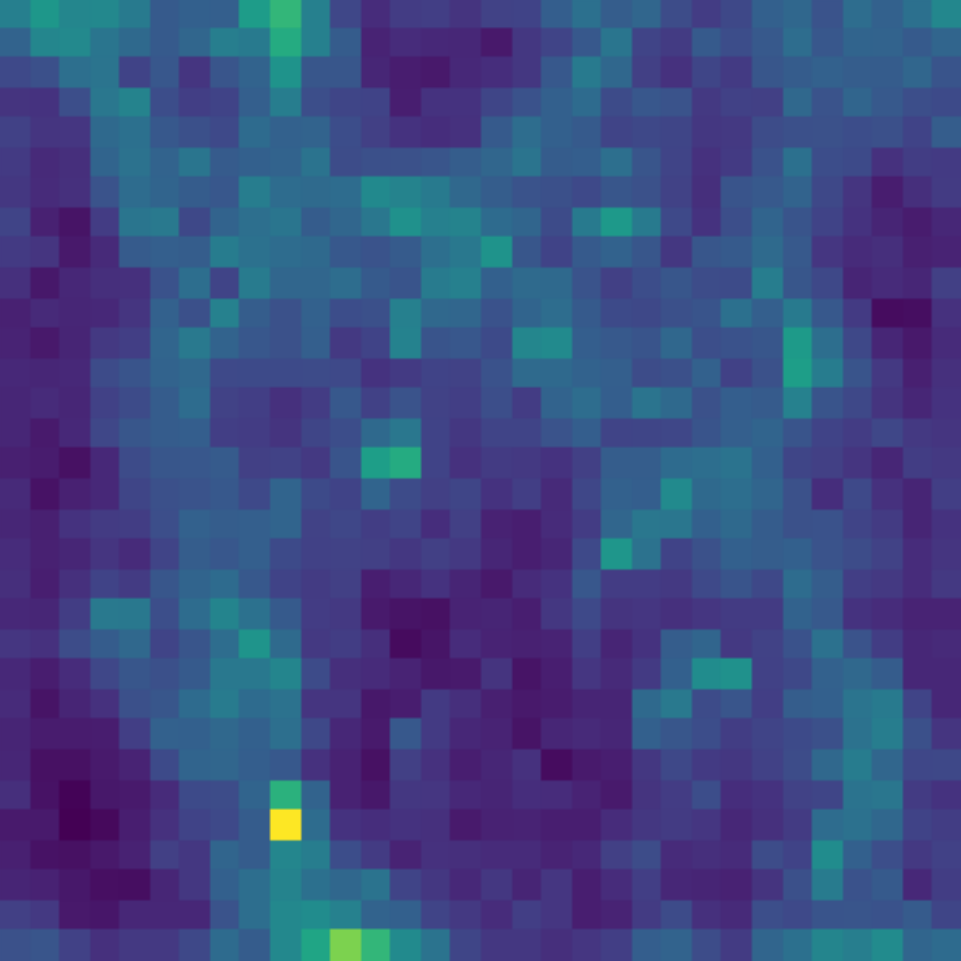}
\end{subfigure}
\begin{subfigure}{.05\textwidth}
  \centering
  \includegraphics[width=1.0\linewidth]{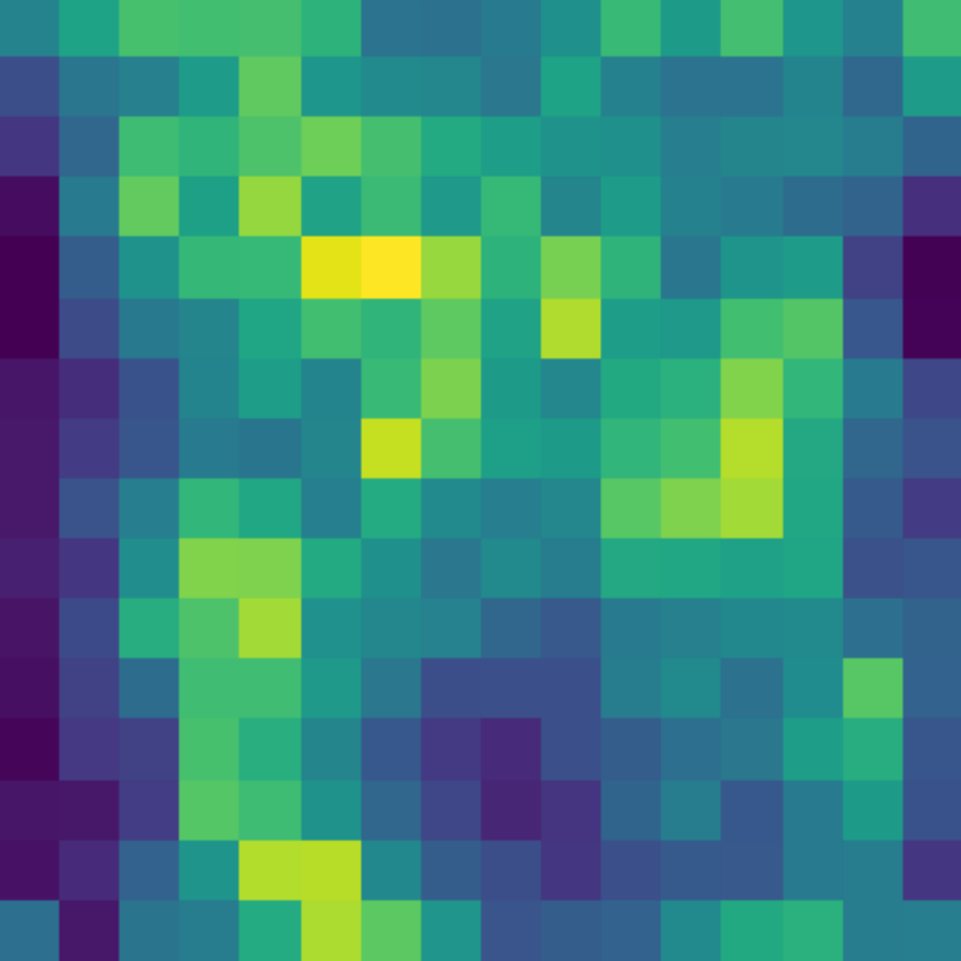}
\end{subfigure}
\begin{subfigure}{.05\textwidth}
  \centering
  \includegraphics[width=1.0\linewidth]{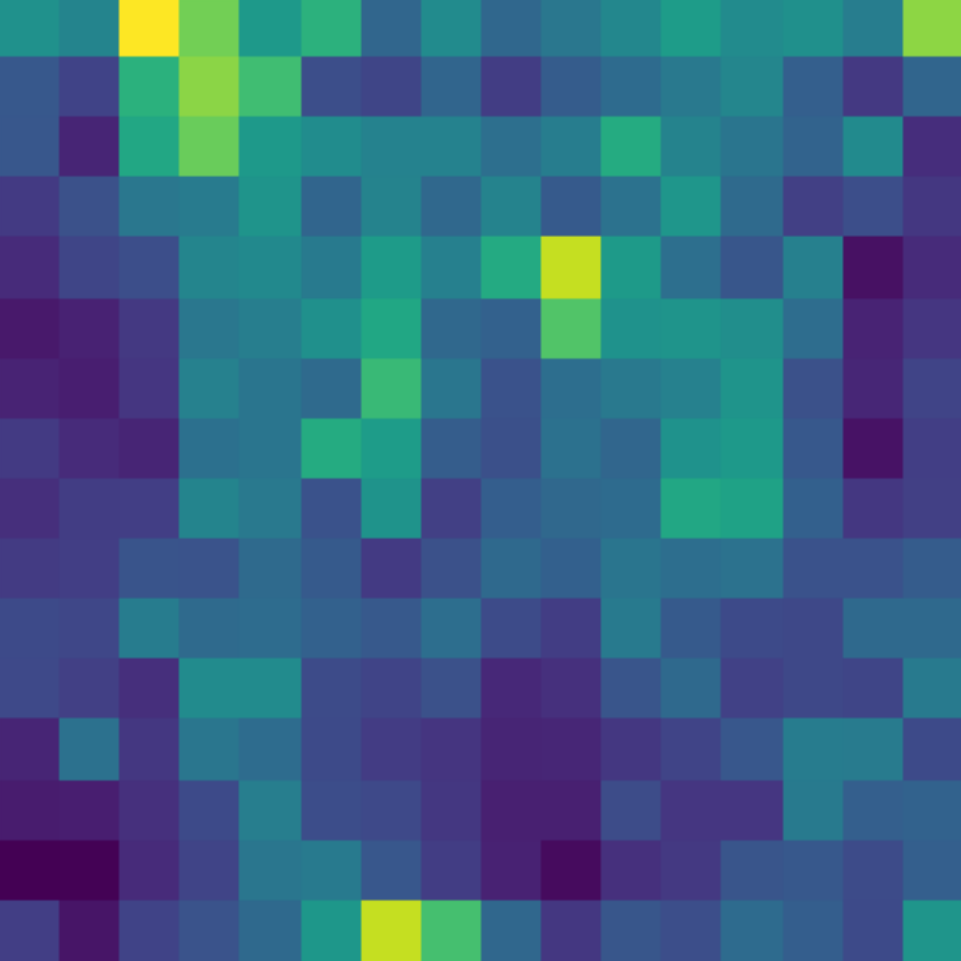}
\end{subfigure}
\begin{subfigure}{.05\textwidth}
  \centering
  \includegraphics[width=1.0\linewidth]{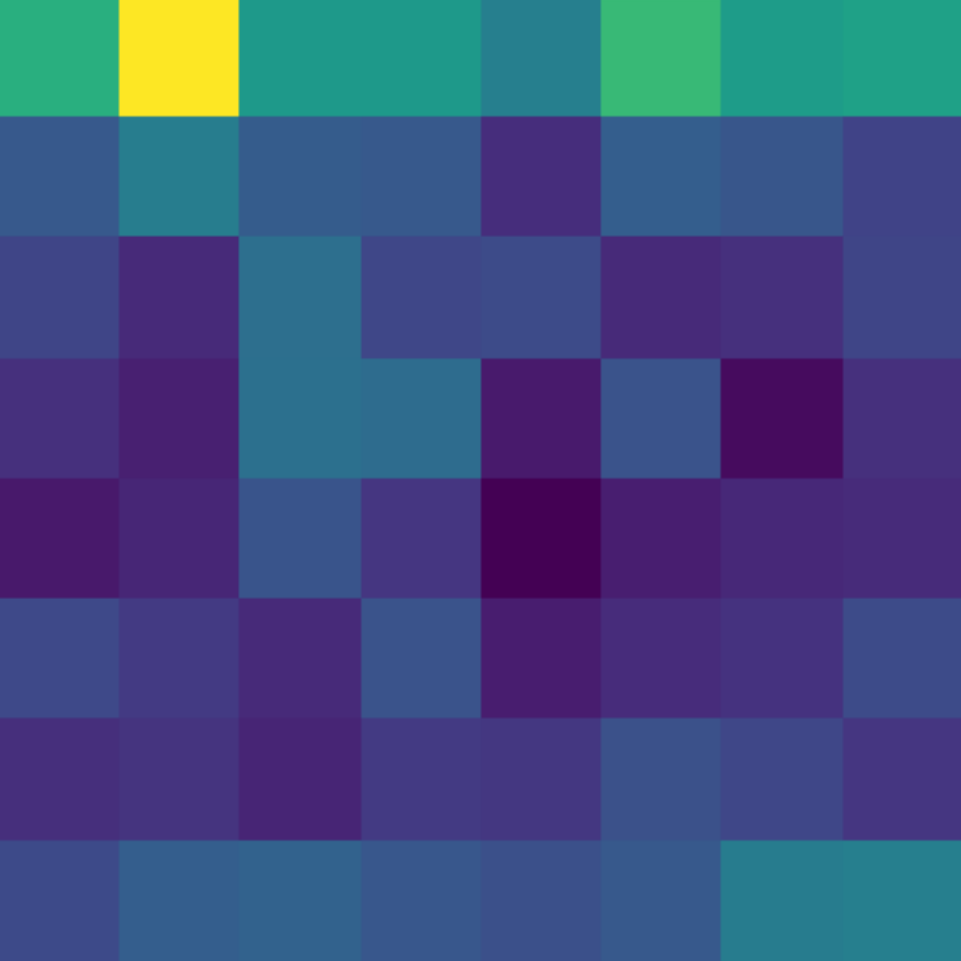}
\end{subfigure}
\begin{subfigure}{.05\textwidth}
  \centering
  \includegraphics[width=1.0\linewidth]{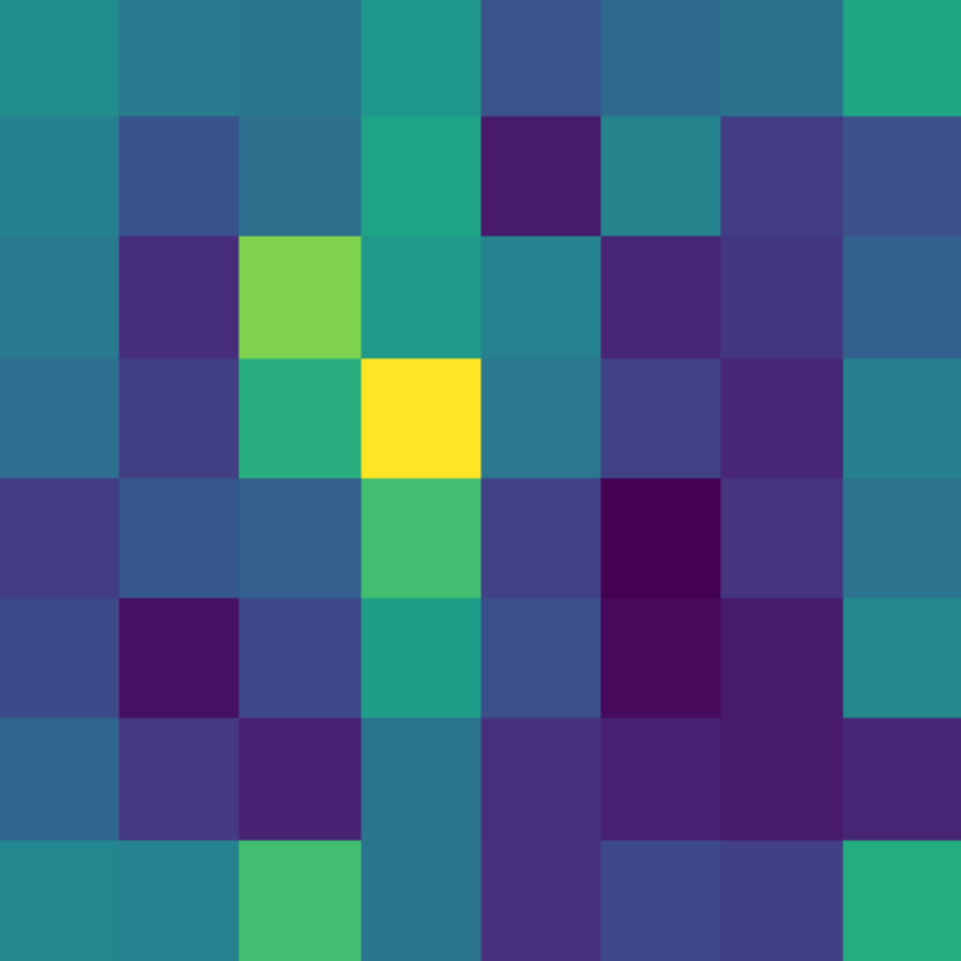}
\end{subfigure}
\begin{subfigure}{.05\textwidth}
  \centering
  \includegraphics[width=1.0\linewidth]{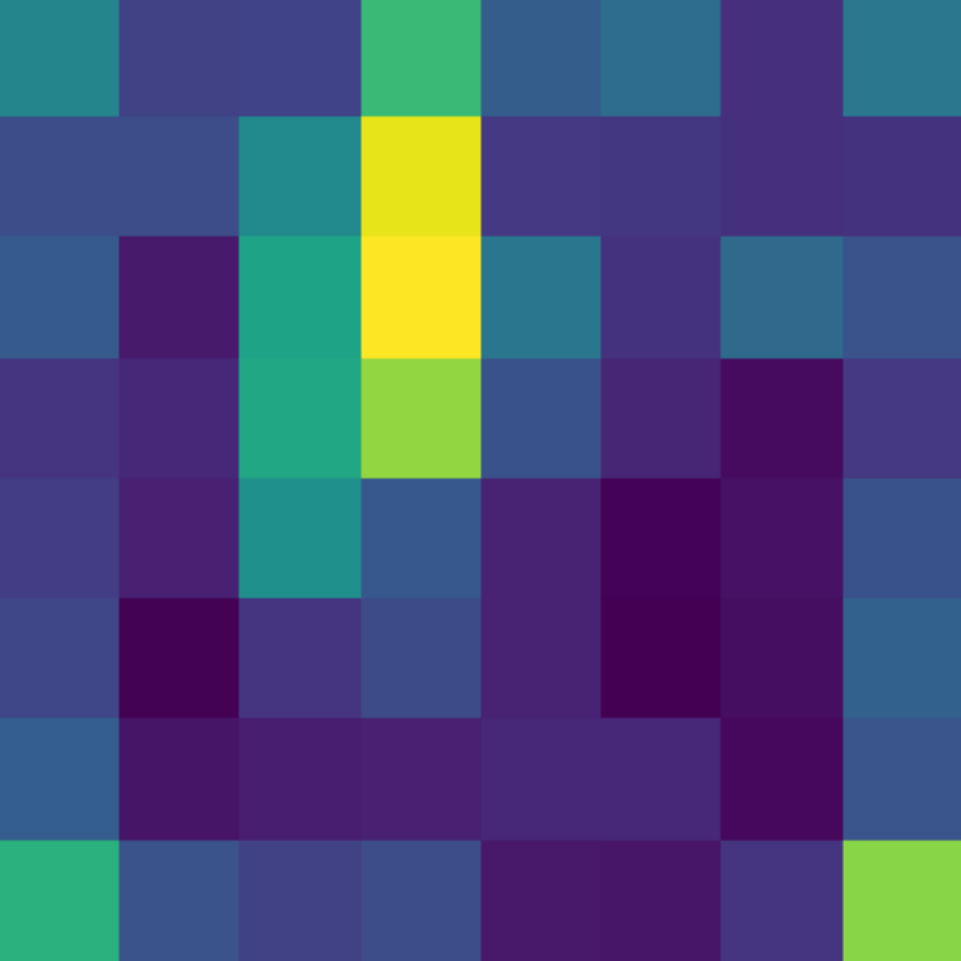}
\end{subfigure}
\begin{subfigure}{.05\textwidth}
  \centering
  \includegraphics[width=1.0\linewidth]{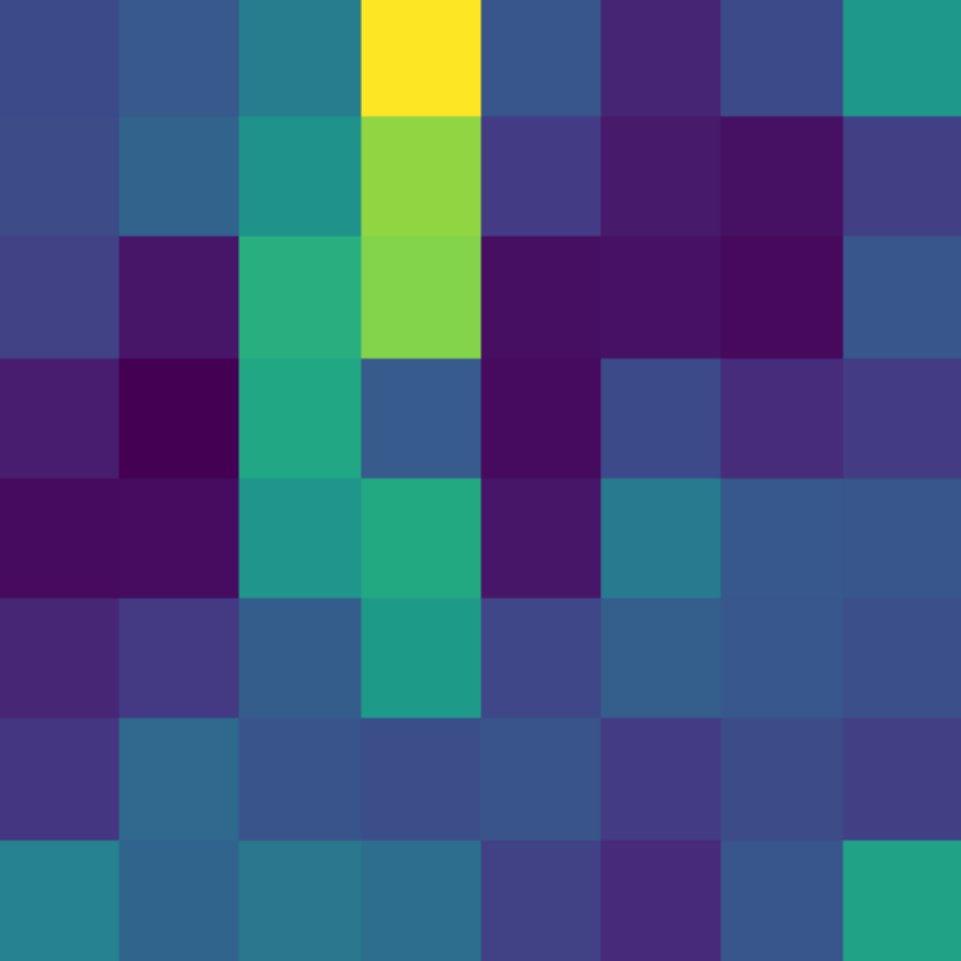}
\end{subfigure}
\begin{subfigure}{.05\textwidth}
  \centering
  \includegraphics[width=1.0\linewidth]{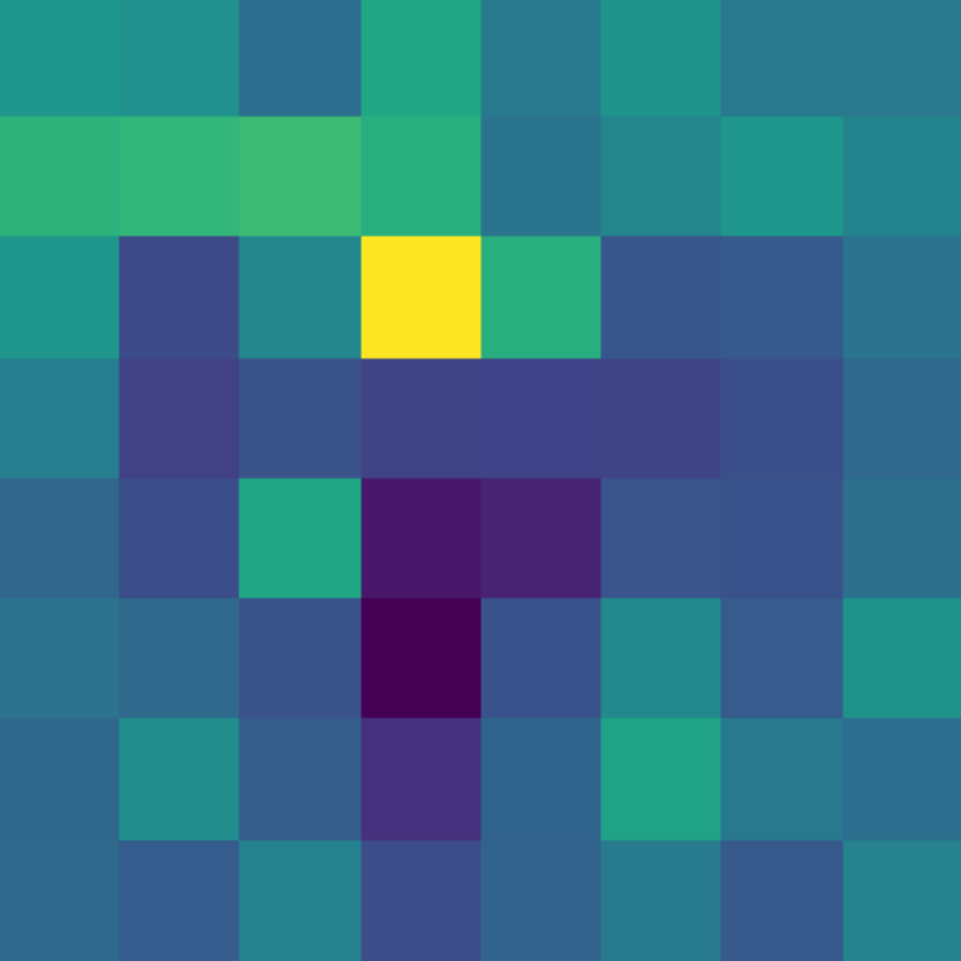}
\end{subfigure}
\begin{subfigure}{.05\textwidth}
  \centering
  \includegraphics[width=1.0\linewidth]{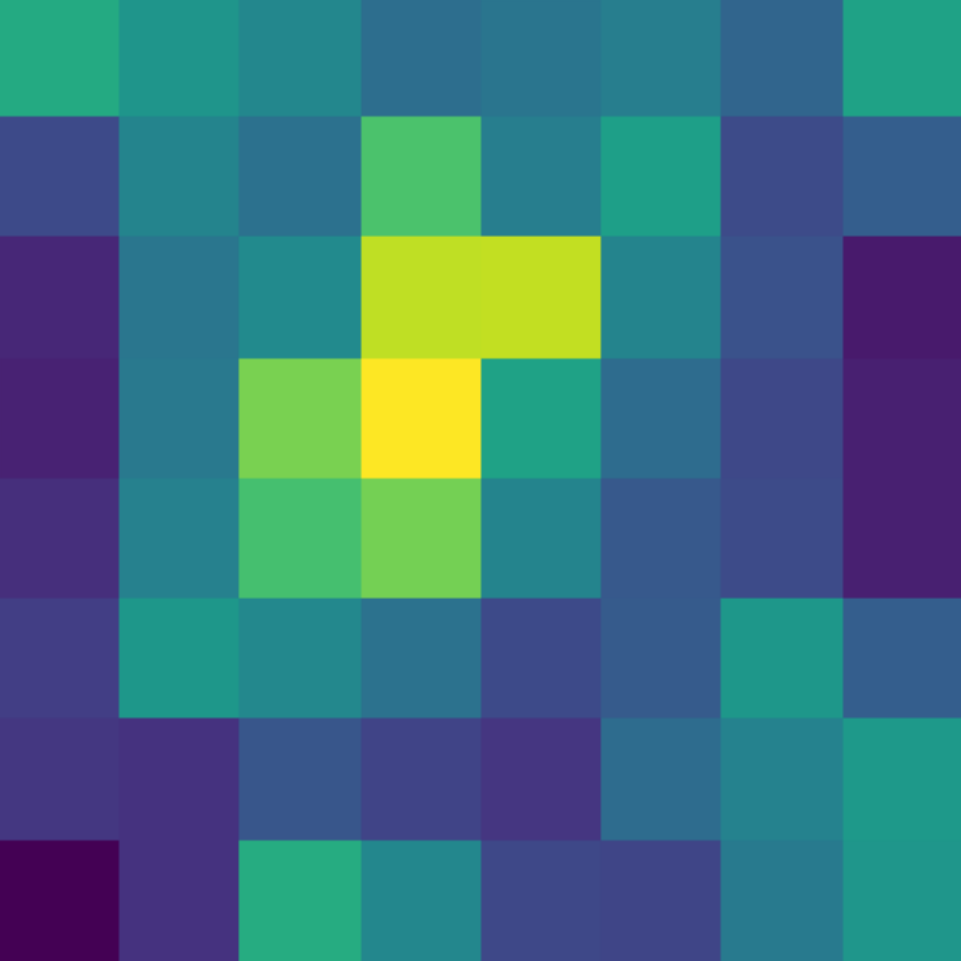}
\end{subfigure}
\begin{subfigure}{.05\textwidth}
  \centering
  \includegraphics[width=1.0\linewidth]{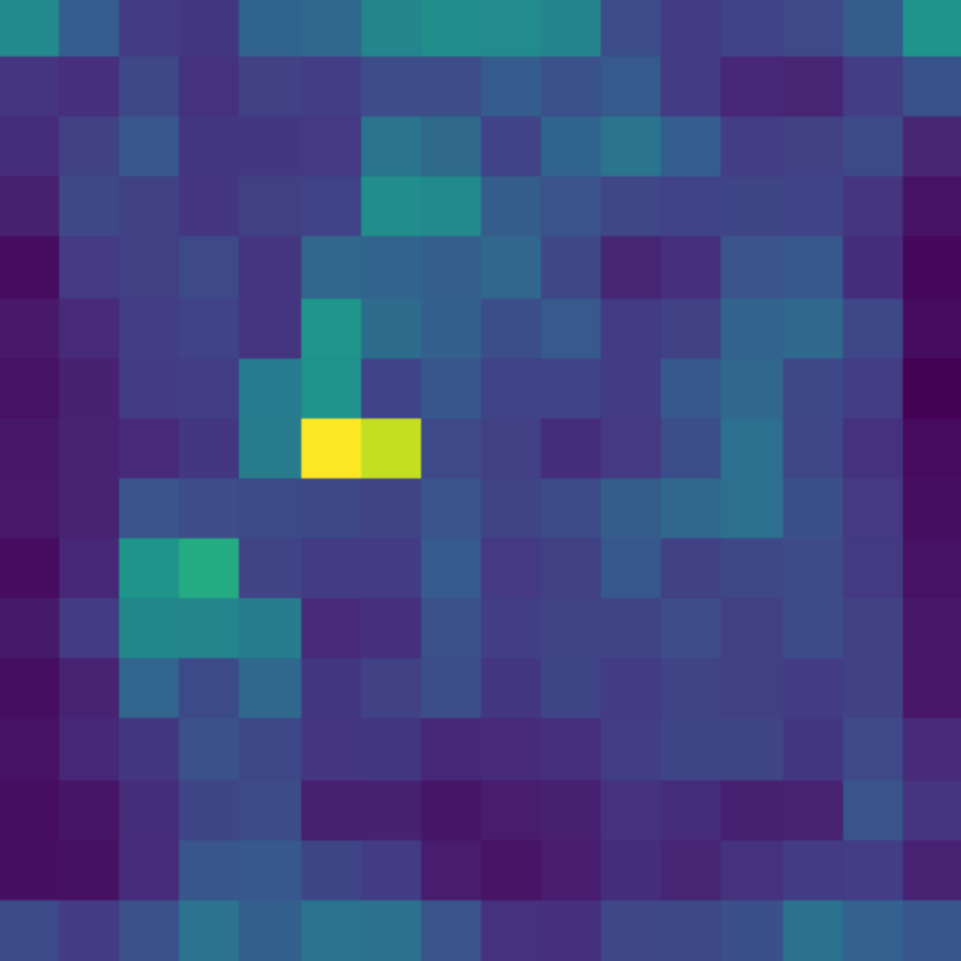}
\end{subfigure}
\begin{subfigure}{.05\textwidth}
  \centering
  \includegraphics[width=1.0\linewidth]{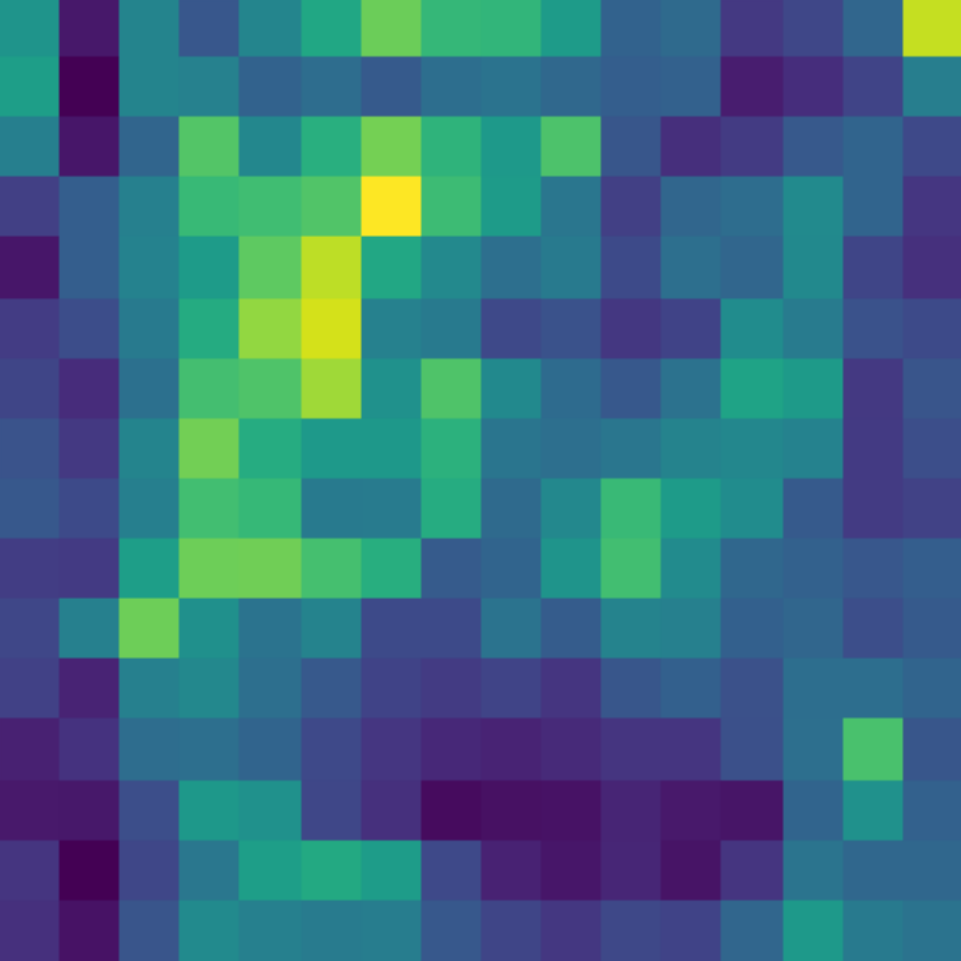}
\end{subfigure}
\begin{subfigure}{.05\textwidth}
  \centering
  \includegraphics[width=1.0\linewidth]{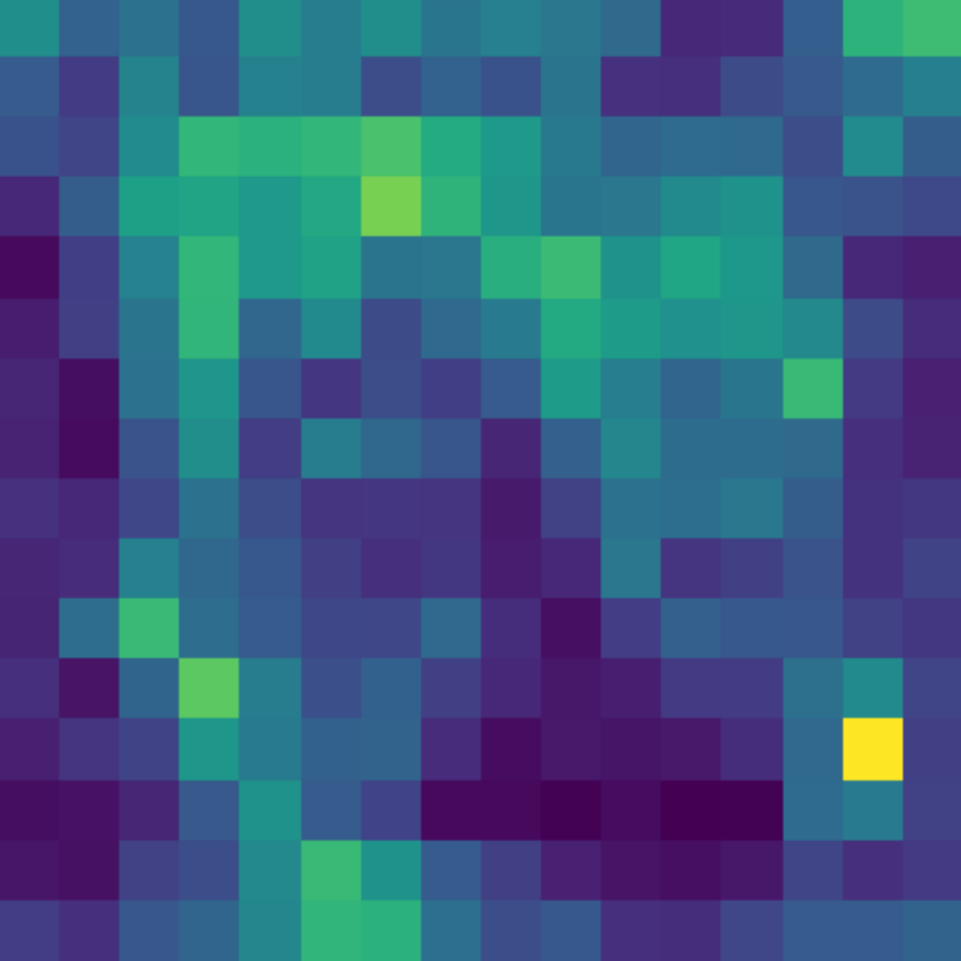}
\end{subfigure}
\begin{subfigure}{.05\textwidth}
  \centering
  \includegraphics[width=1.0\linewidth]{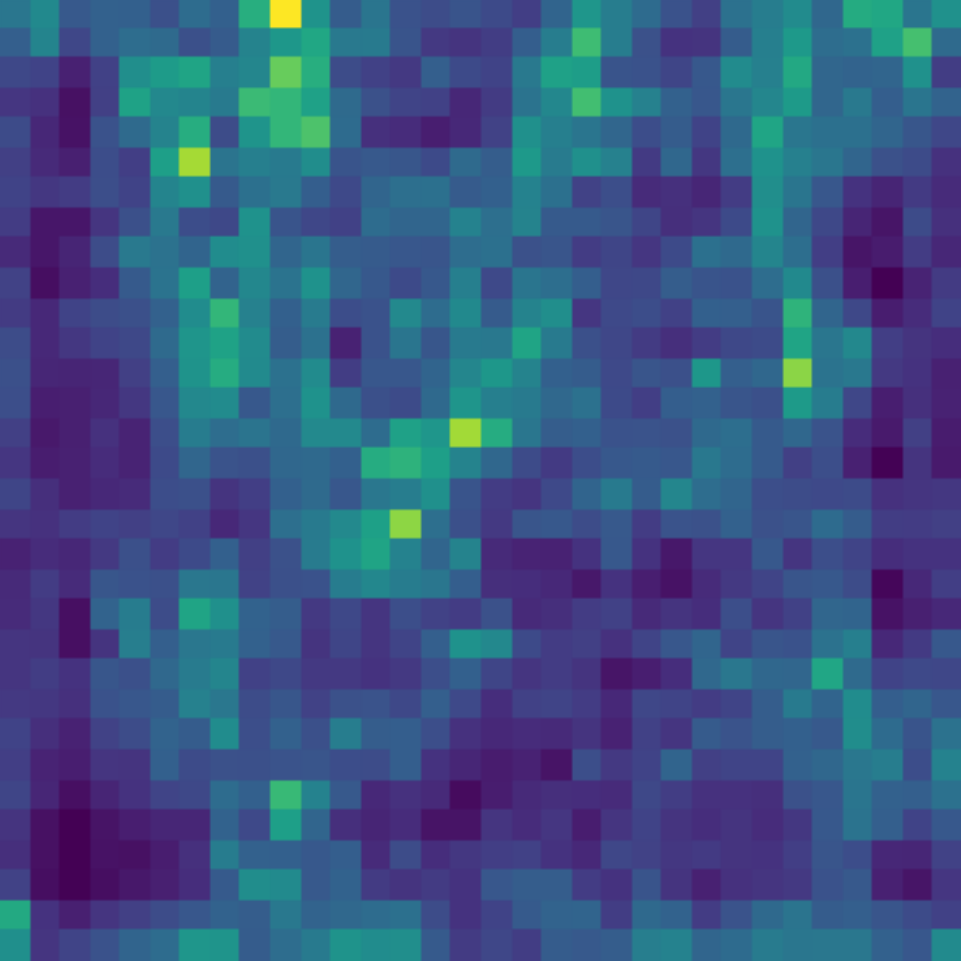}
\end{subfigure}
\begin{subfigure}{.05\textwidth}
  \centering
  \includegraphics[width=1.0\linewidth]{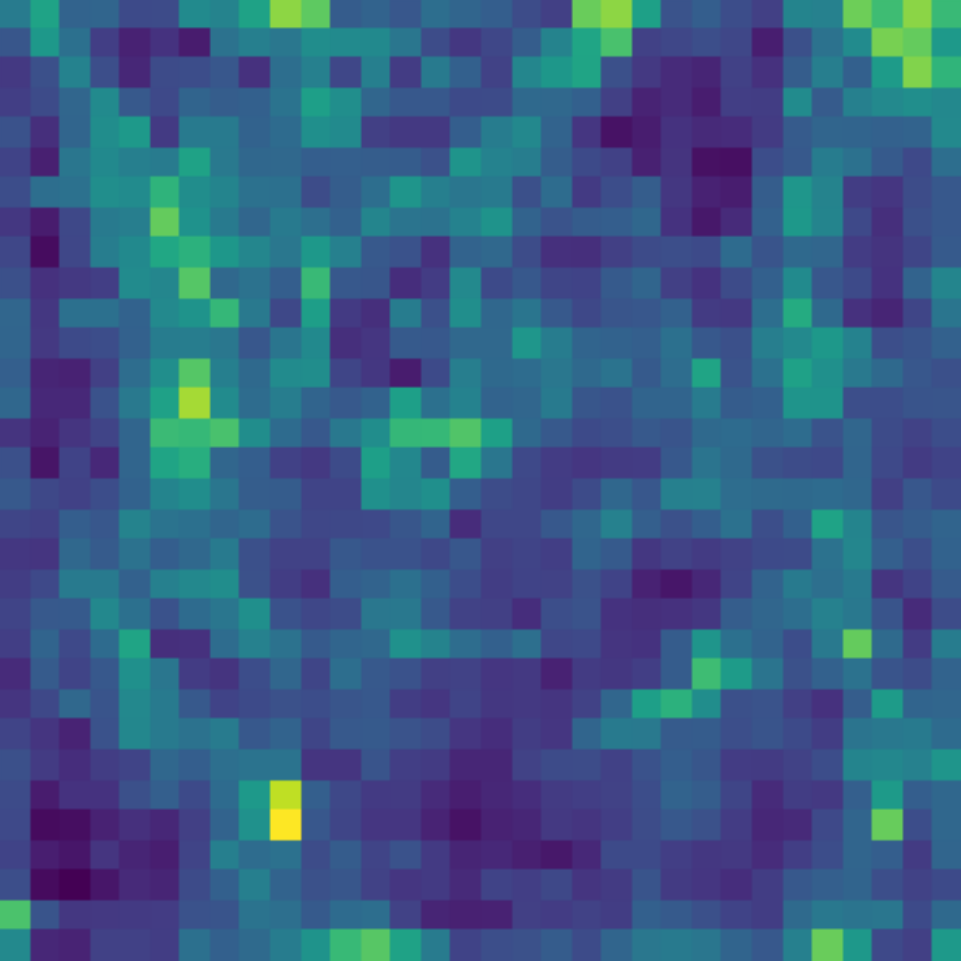}
\end{subfigure}
\begin{subfigure}{.05\textwidth}
  \centering
  \includegraphics[width=1.0\linewidth]{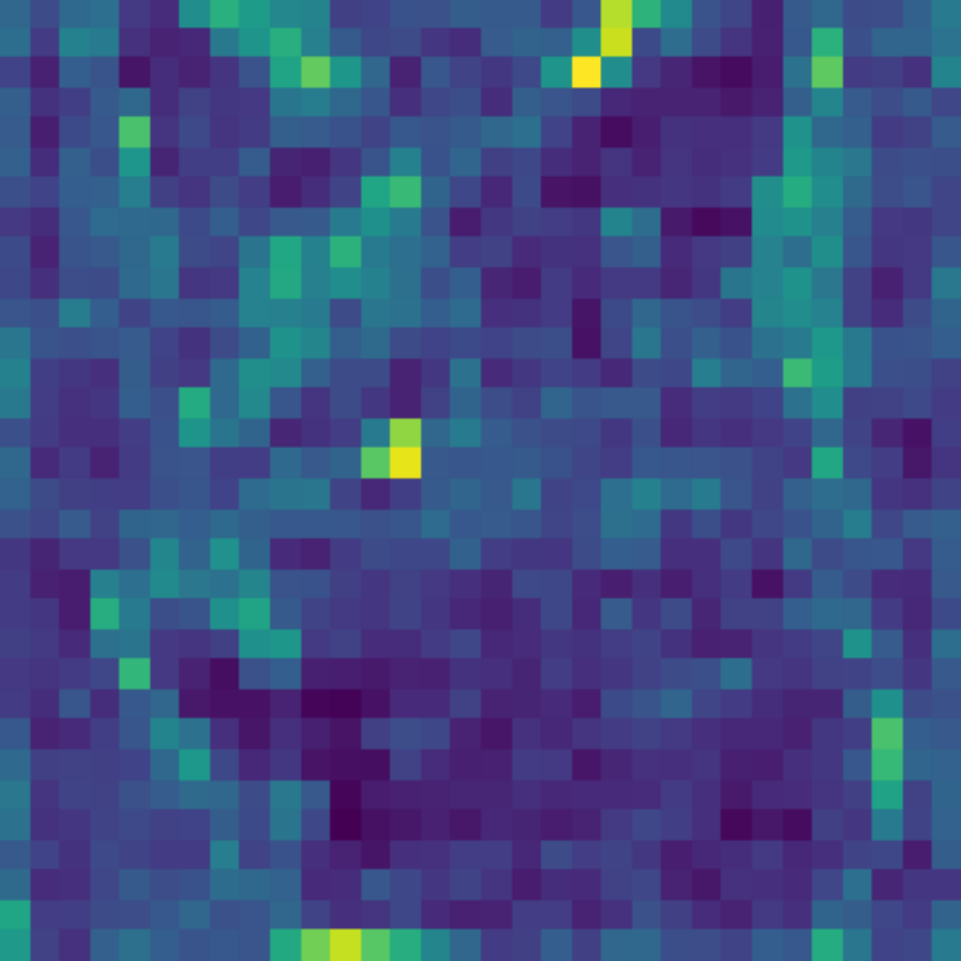}
\end{subfigure}
\\
&
\begin{subfigure}{.05\textwidth}
  \centering
  \includegraphics[width=1.0\linewidth]{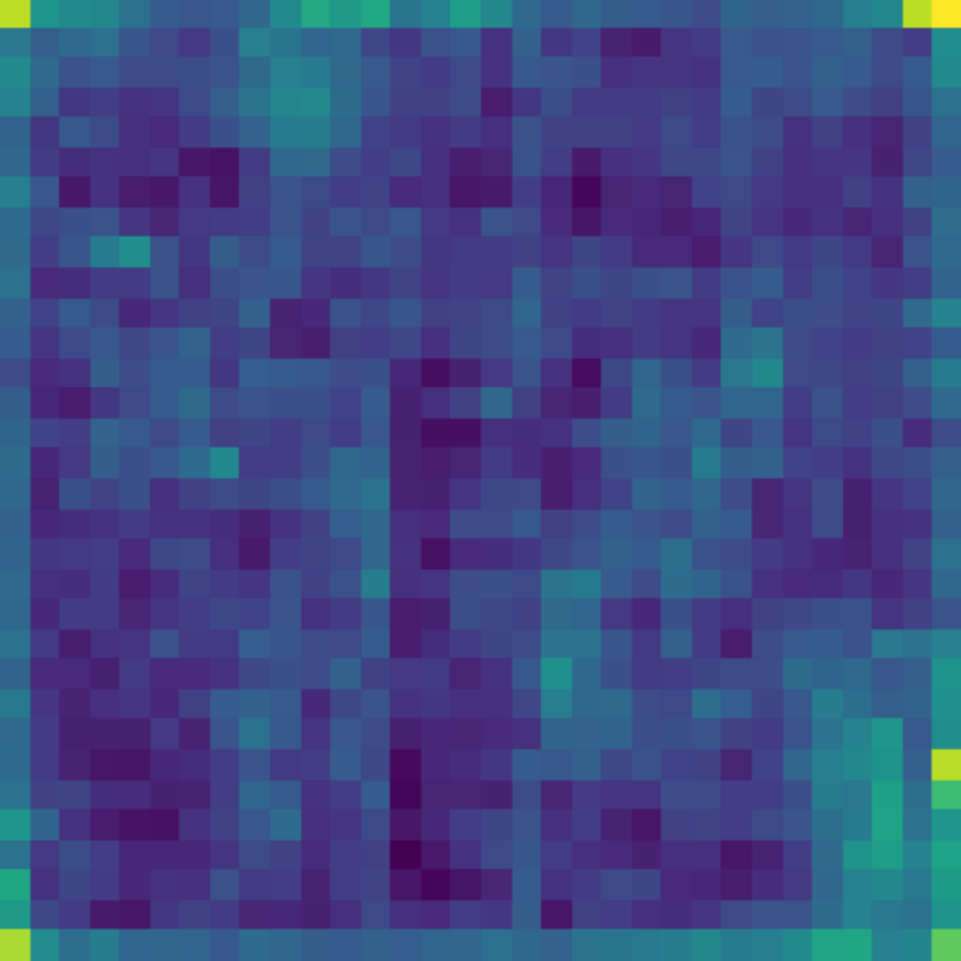}
\end{subfigure}
\begin{subfigure}{.05\textwidth}
  \centering
  \includegraphics[width=1.0\linewidth]{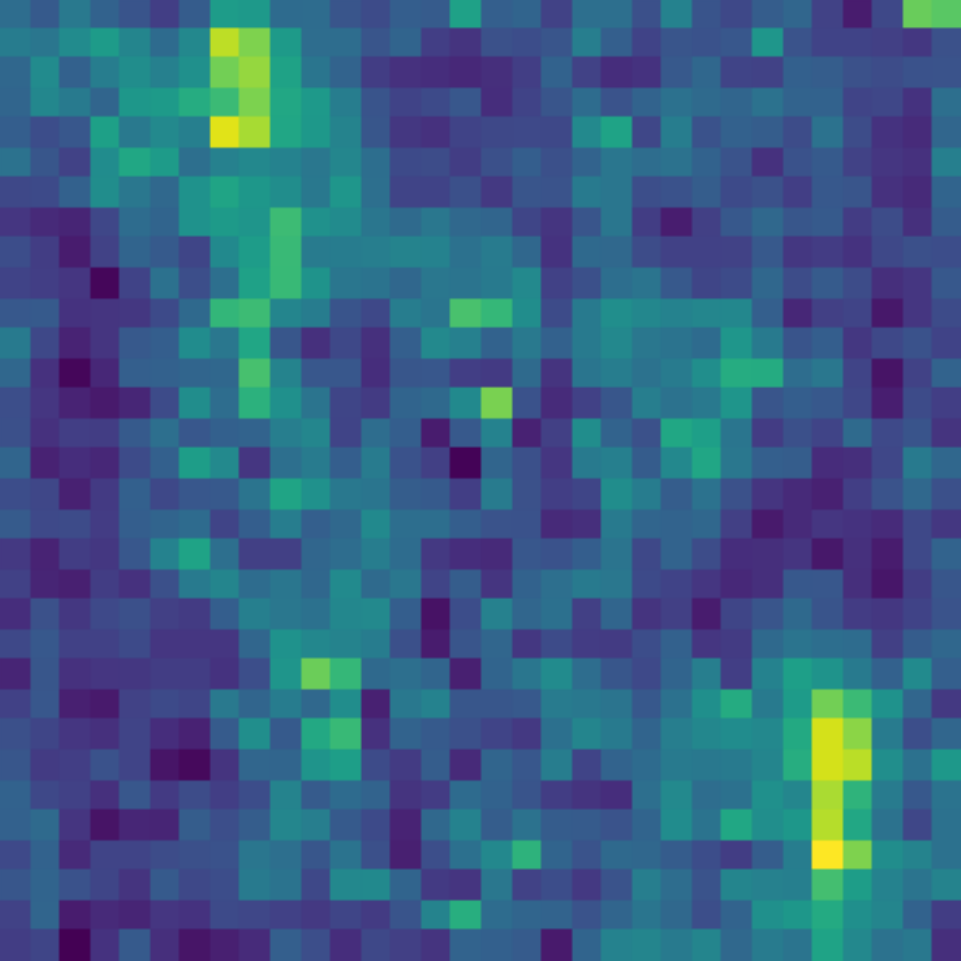}
\end{subfigure}
\begin{subfigure}{.05\textwidth}
  \centering
  \includegraphics[width=1.0\linewidth]{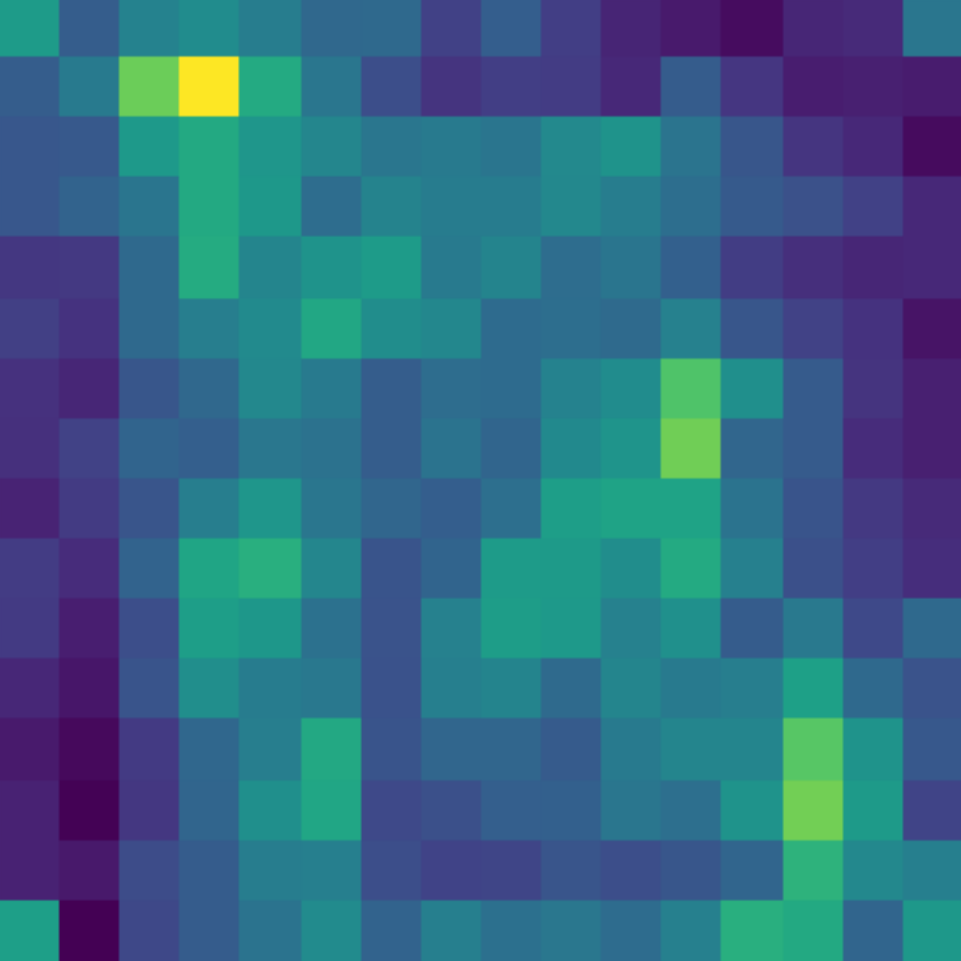}
\end{subfigure}
\begin{subfigure}{.05\textwidth}
  \centering
  \includegraphics[width=1.0\linewidth]{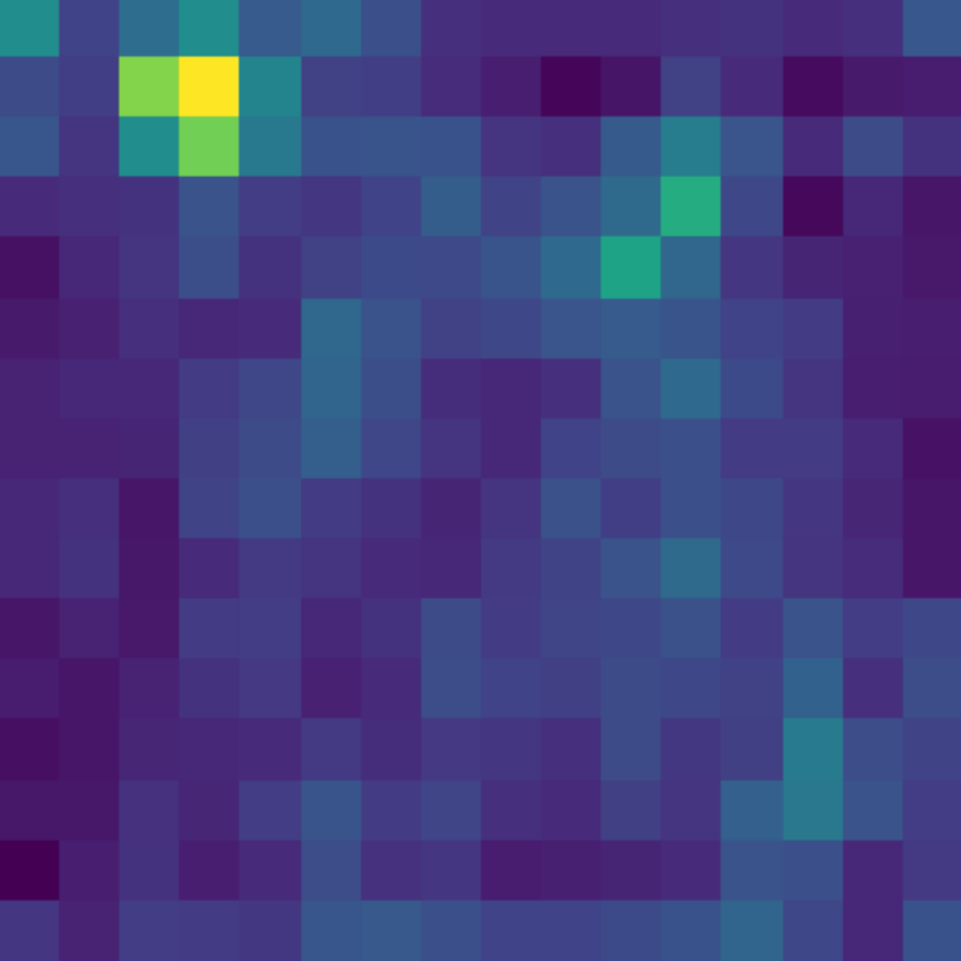}
\end{subfigure}
\begin{subfigure}{.05\textwidth}
  \centering
  \includegraphics[width=1.0\linewidth]{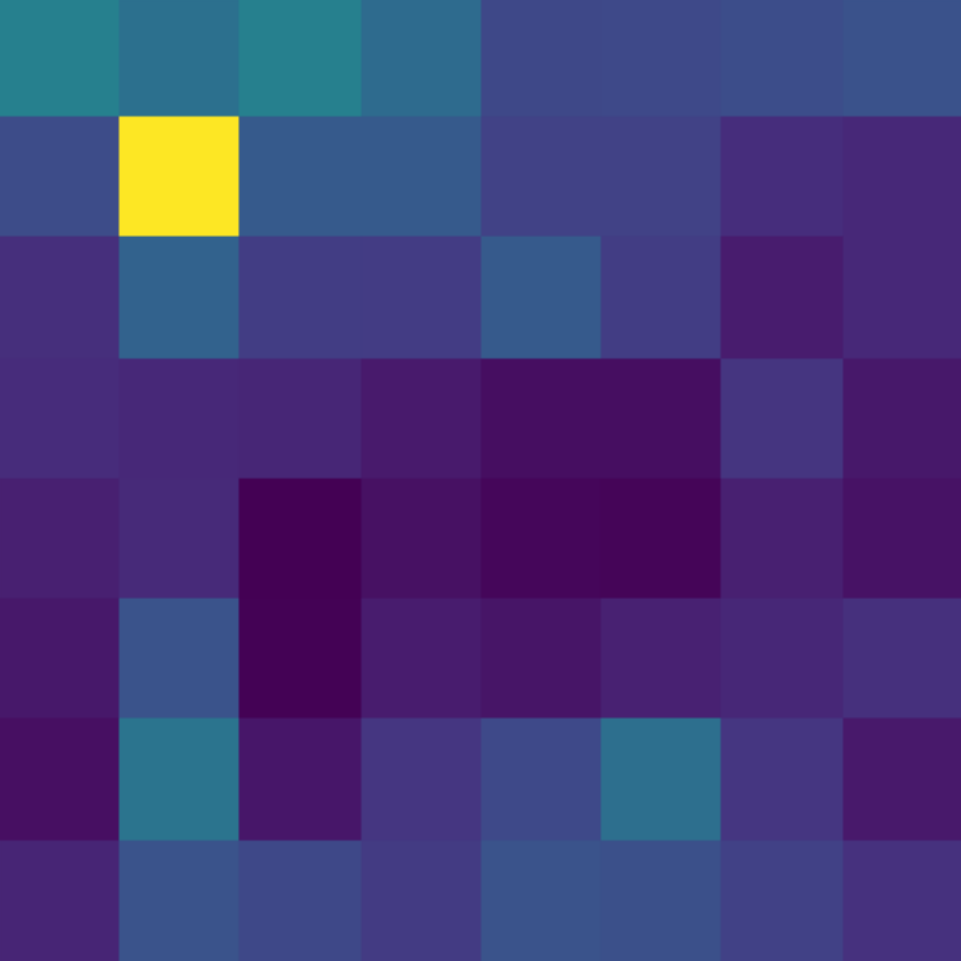}
\end{subfigure}
\begin{subfigure}{.05\textwidth}
  \centering
  \includegraphics[width=1.0\linewidth]{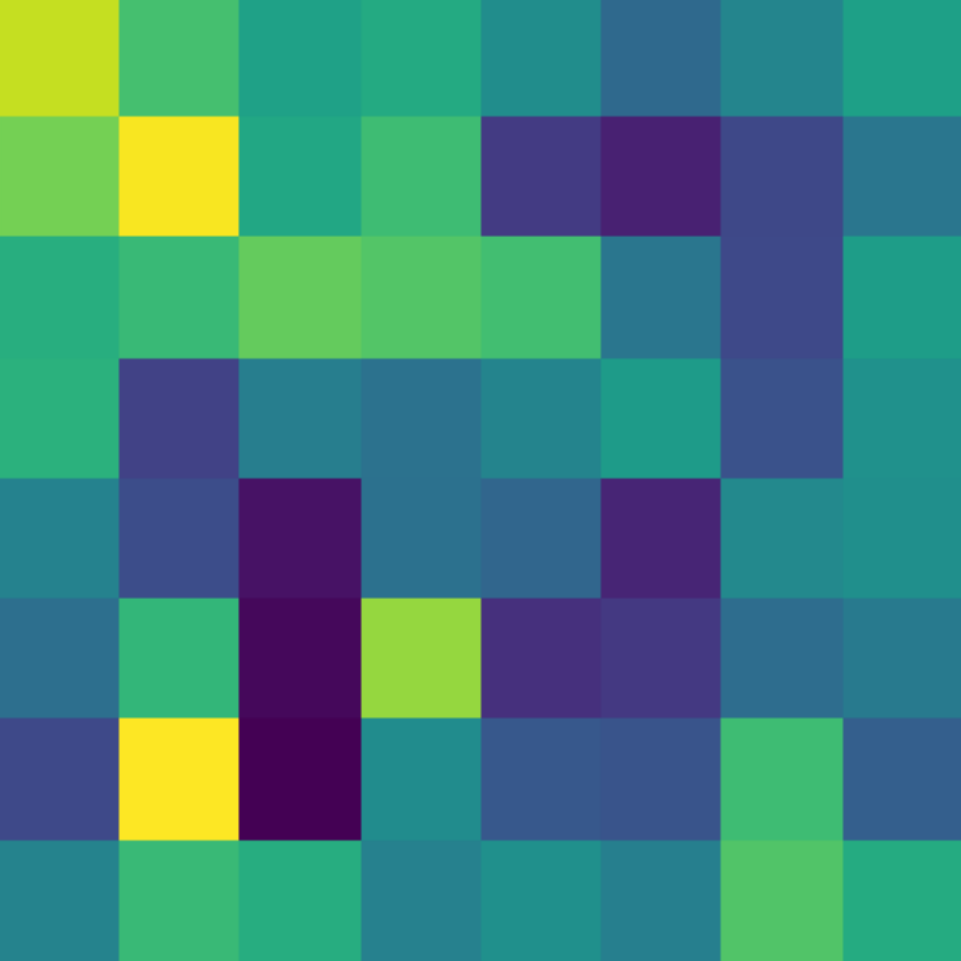}
\end{subfigure}
\begin{subfigure}{.05\textwidth}
  \centering
  \includegraphics[width=1.0\linewidth]{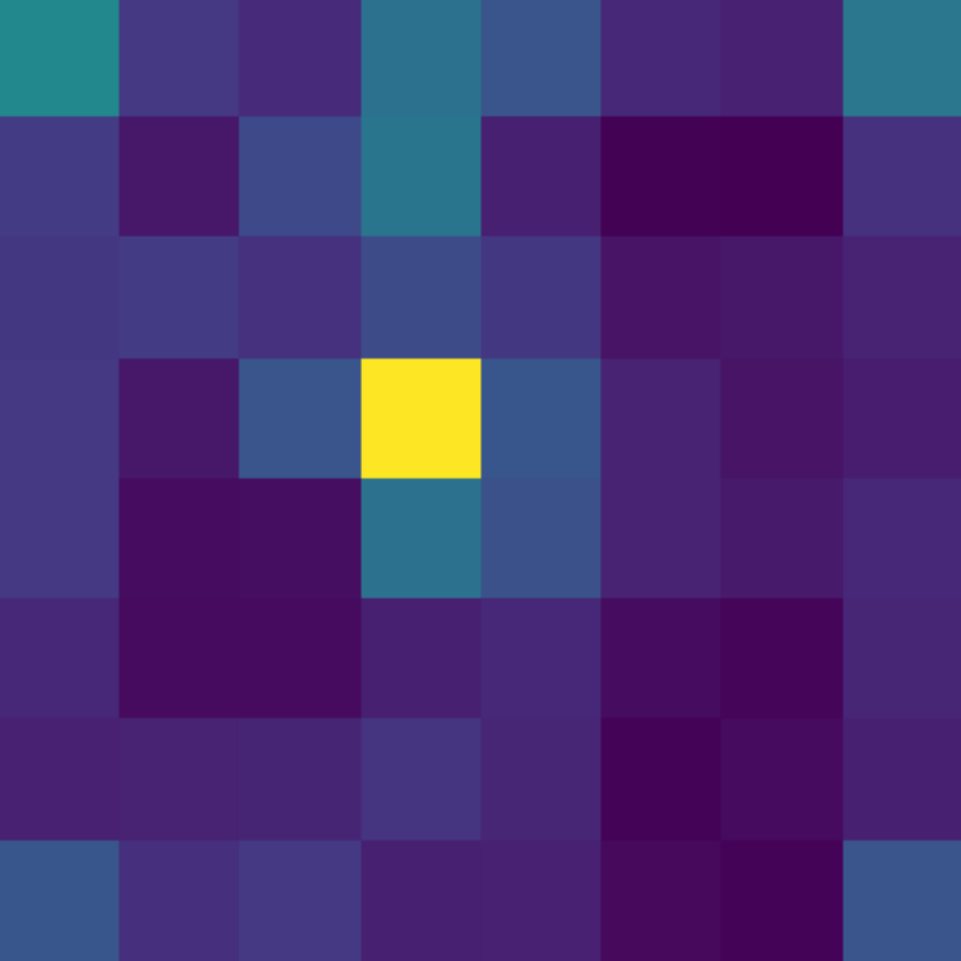}
\end{subfigure}
\begin{subfigure}{.05\textwidth}
  \centering
  \includegraphics[width=1.0\linewidth]{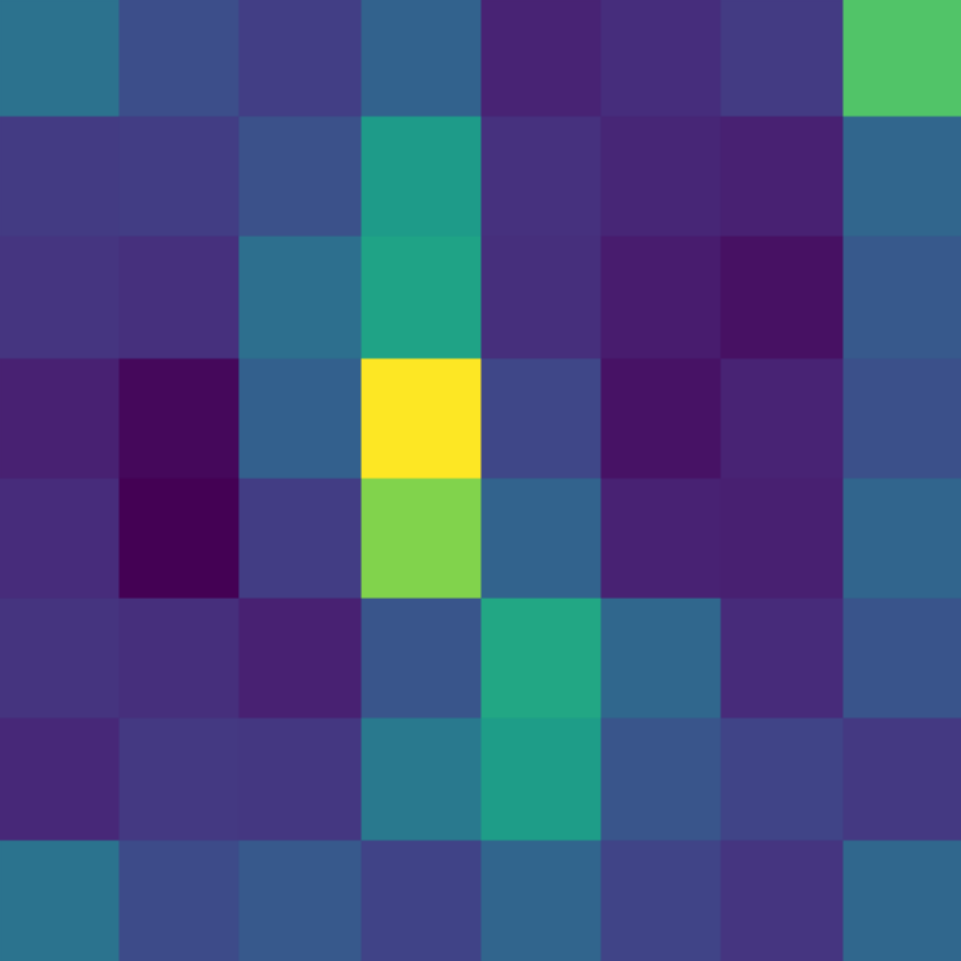}
\end{subfigure}
\begin{subfigure}{.05\textwidth}
  \centering
  \includegraphics[width=1.0\linewidth]{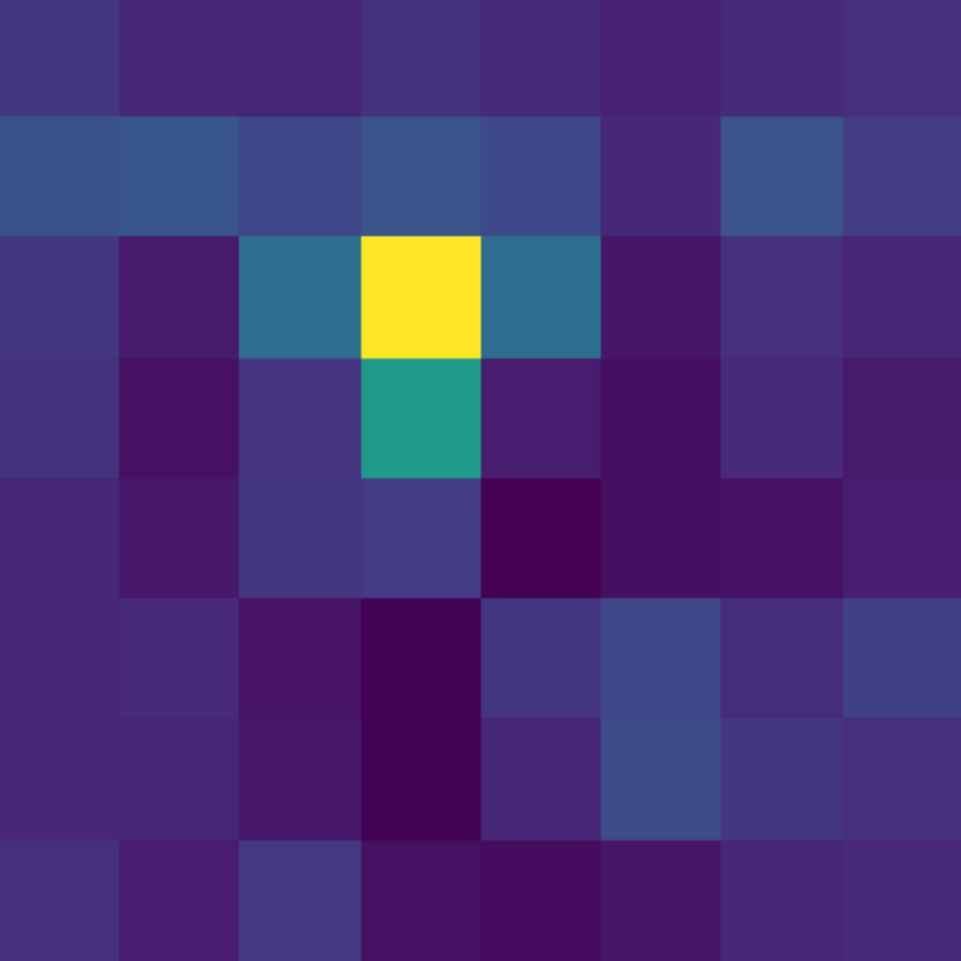}
\end{subfigure}
\begin{subfigure}{.05\textwidth}
  \centering
  \includegraphics[width=1.0\linewidth]{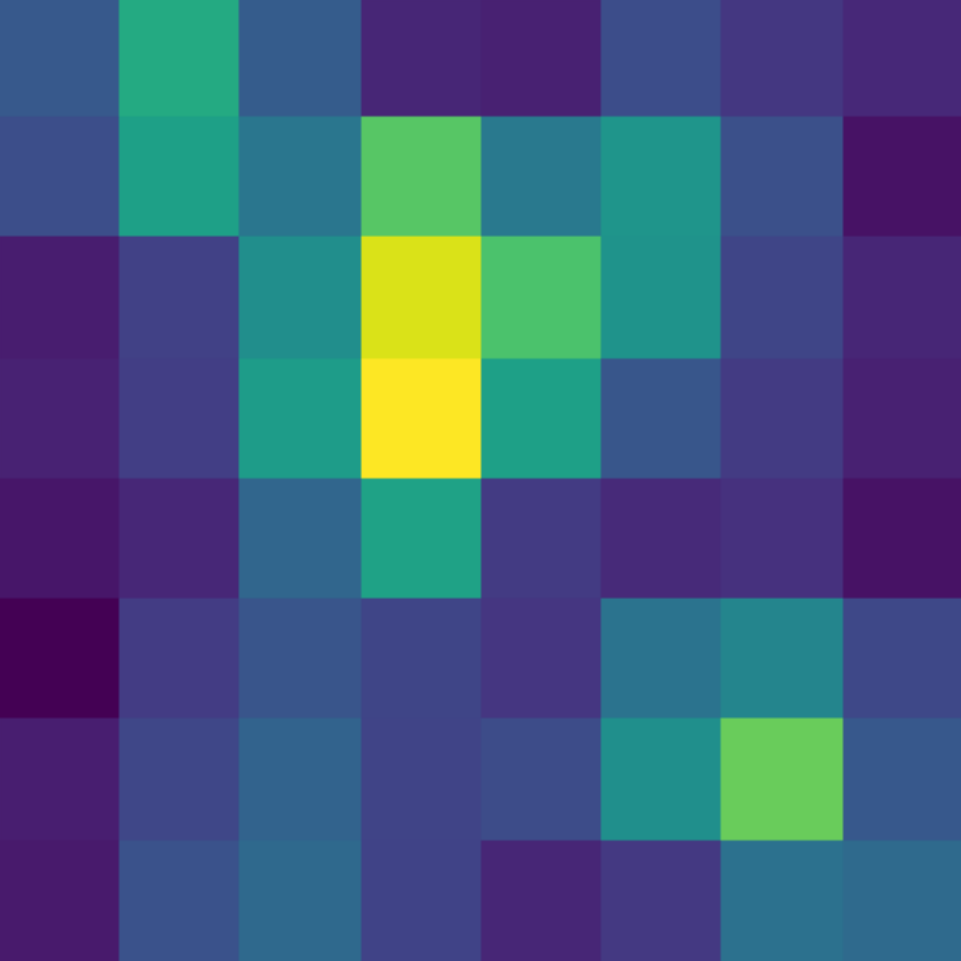}
\end{subfigure}
\begin{subfigure}{.05\textwidth}
  \centering
  \includegraphics[width=1.0\linewidth]{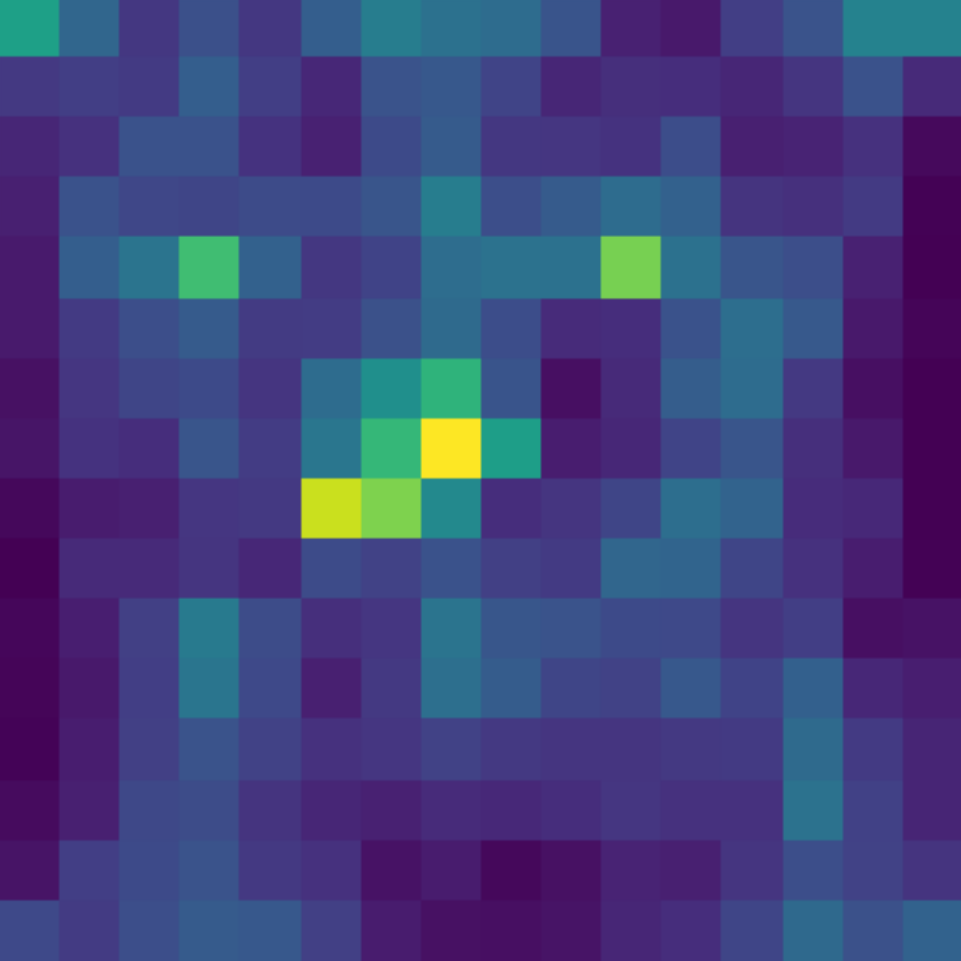}
\end{subfigure}
\begin{subfigure}{.05\textwidth}
  \centering
  \includegraphics[width=1.0\linewidth]{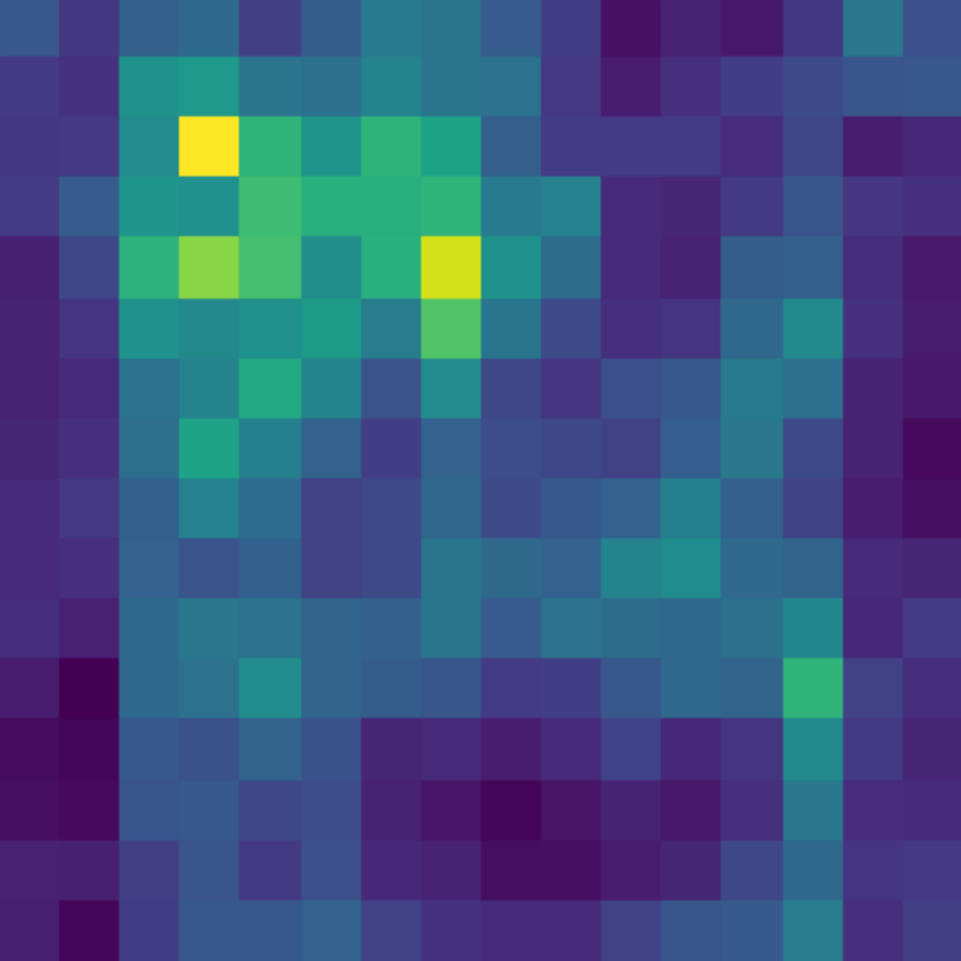}
\end{subfigure}
\begin{subfigure}{.05\textwidth}
  \centering
  \includegraphics[width=1.0\linewidth]{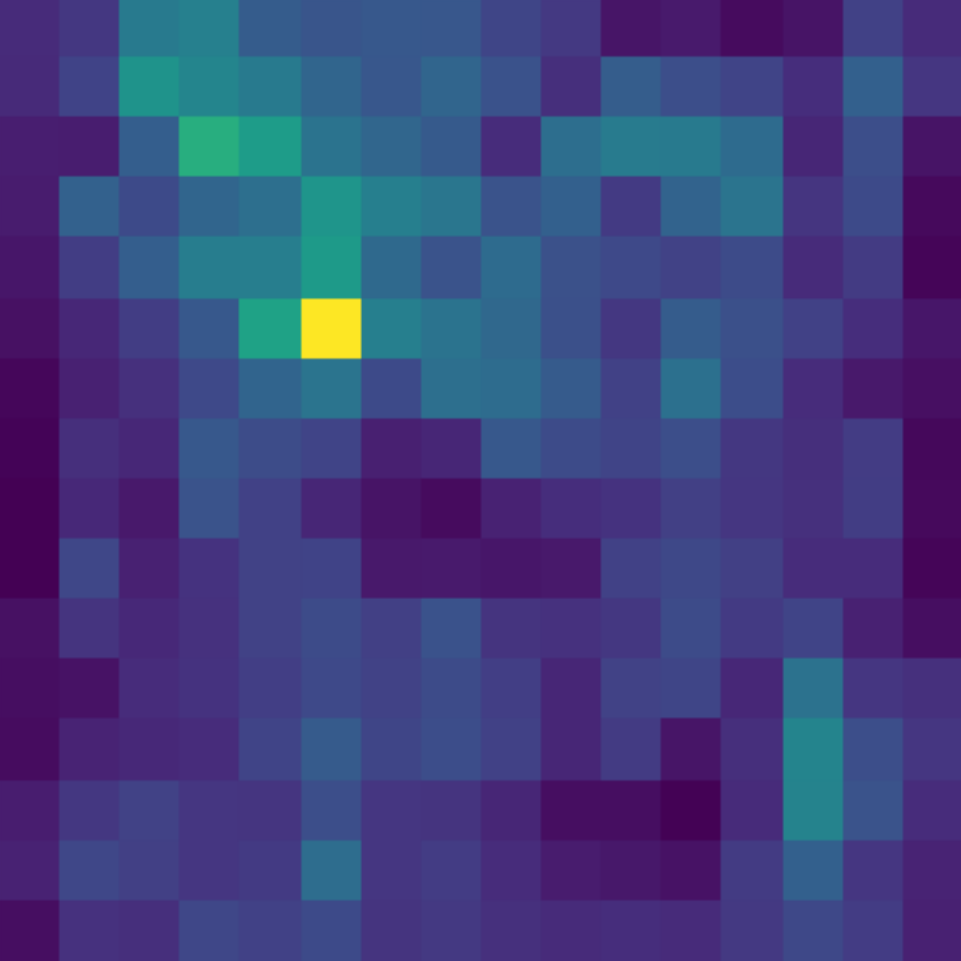}
\end{subfigure}
\begin{subfigure}{.05\textwidth}
  \centering
  \includegraphics[width=1.0\linewidth]{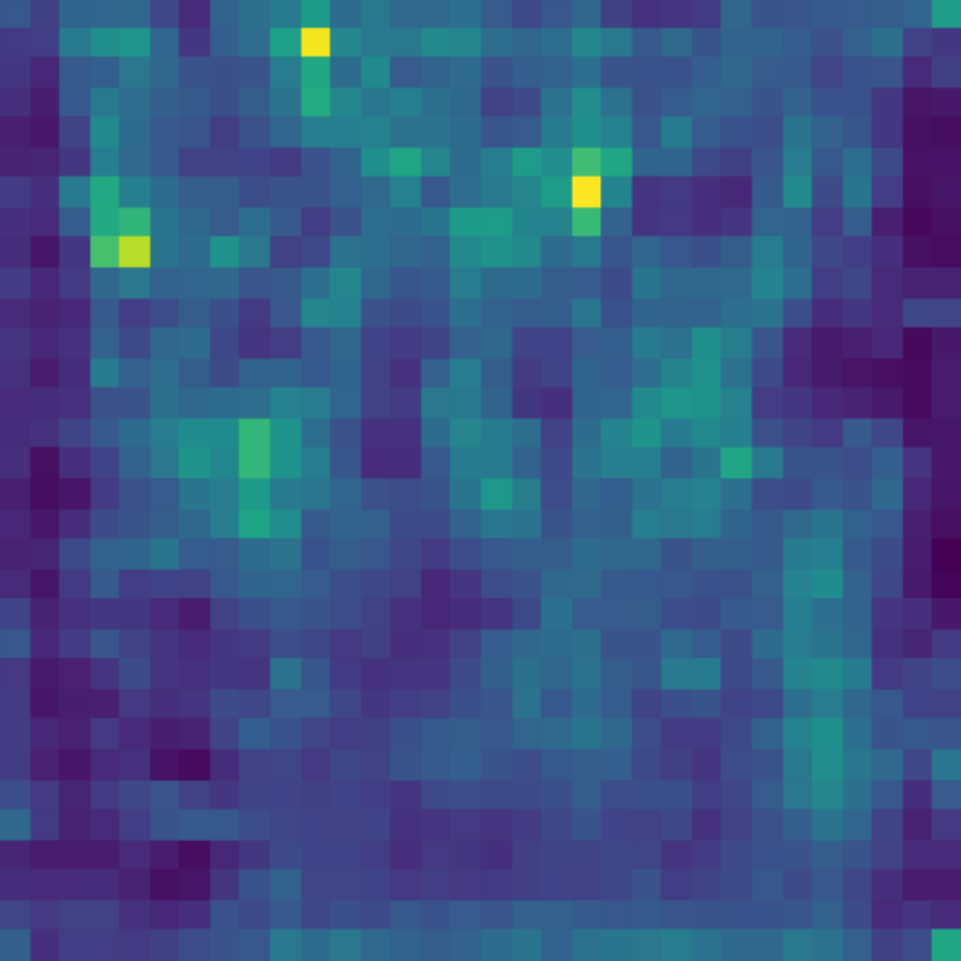}
\end{subfigure}
\begin{subfigure}{.05\textwidth}
  \centering
  \includegraphics[width=1.0\linewidth]{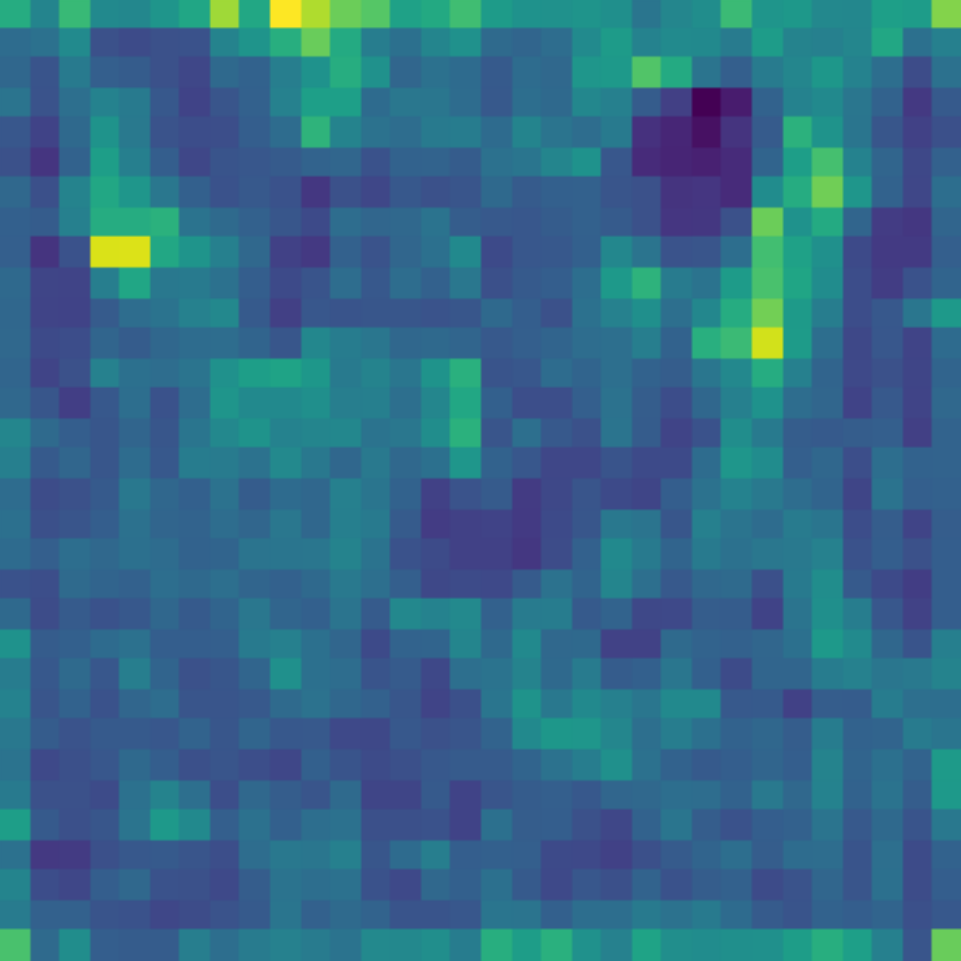}
\end{subfigure}
\begin{subfigure}{.05\textwidth}
  \centering
  \includegraphics[width=1.0\linewidth]{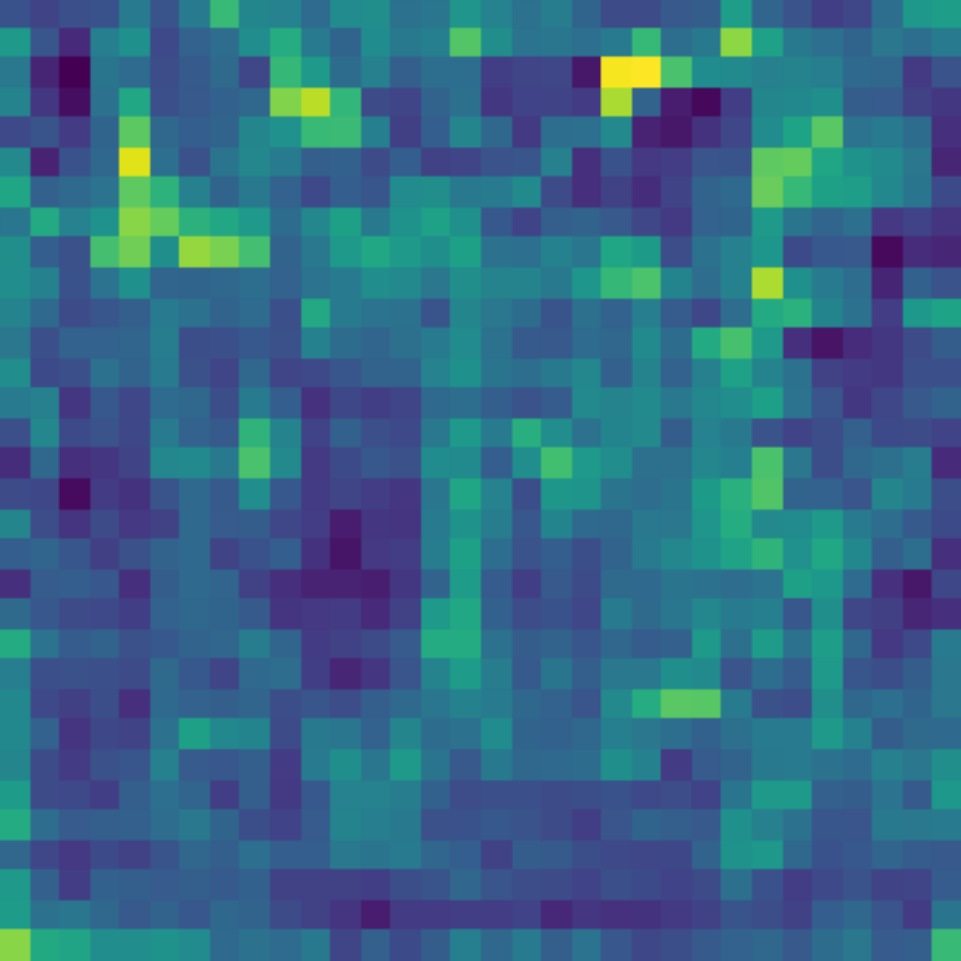}
\end{subfigure}
\\
&
\begin{subfigure}{.05\textwidth}
  \centering
  \includegraphics[width=1.0\linewidth]{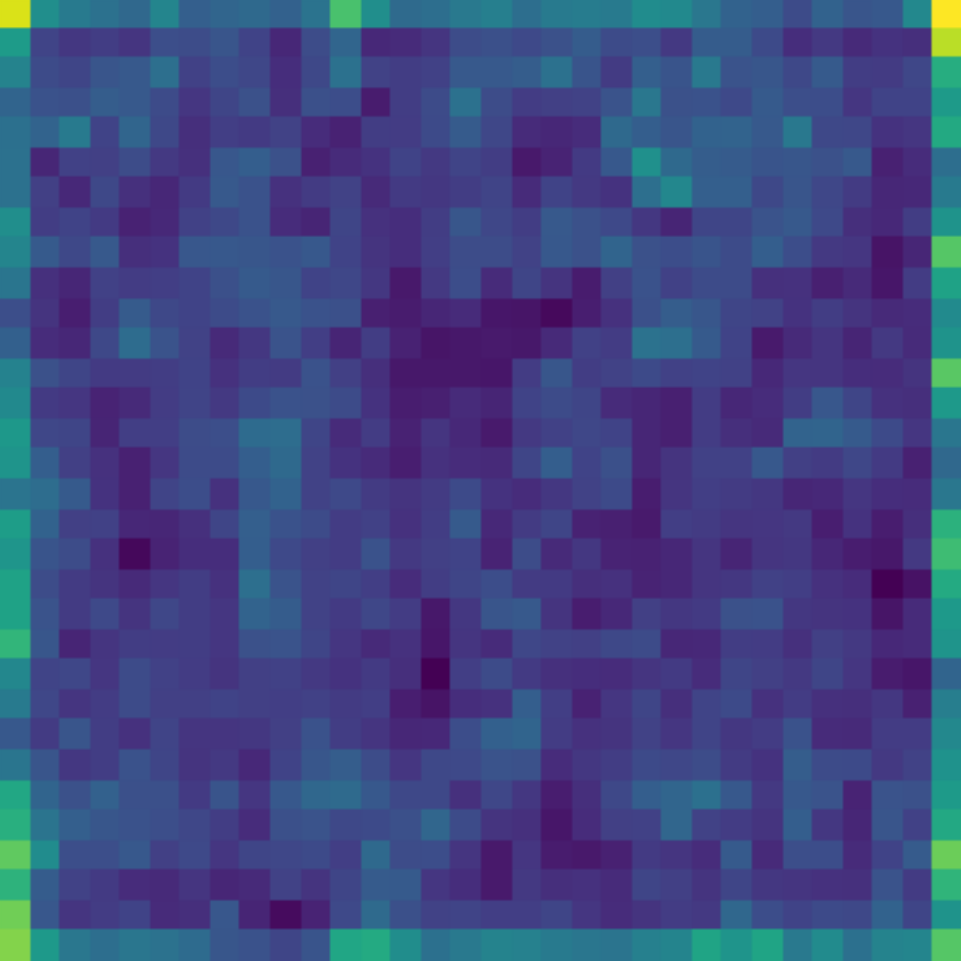}
\end{subfigure}
\begin{subfigure}{.05\textwidth}
  \centering
  \includegraphics[width=1.0\linewidth]{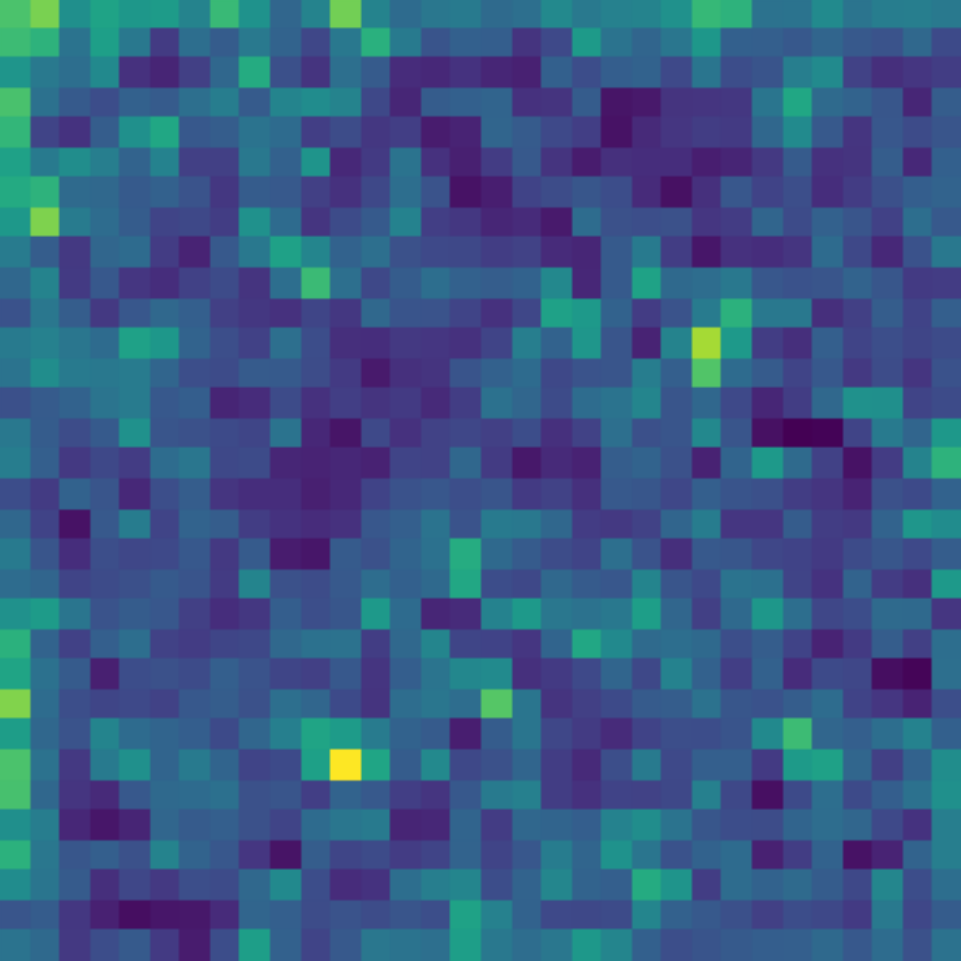}
\end{subfigure}
\begin{subfigure}{.05\textwidth}
  \centering
  \includegraphics[width=1.0\linewidth]{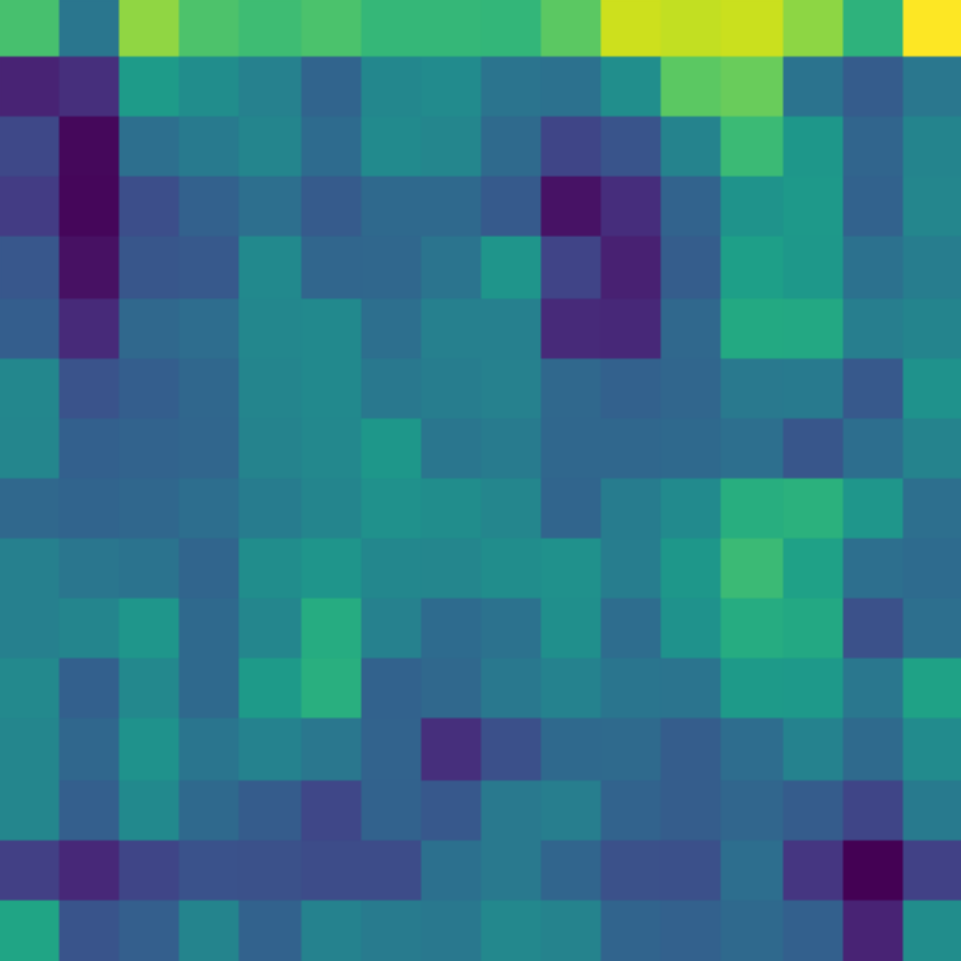}
\end{subfigure}
\begin{subfigure}{.05\textwidth}
  \centering
  \includegraphics[width=1.0\linewidth]{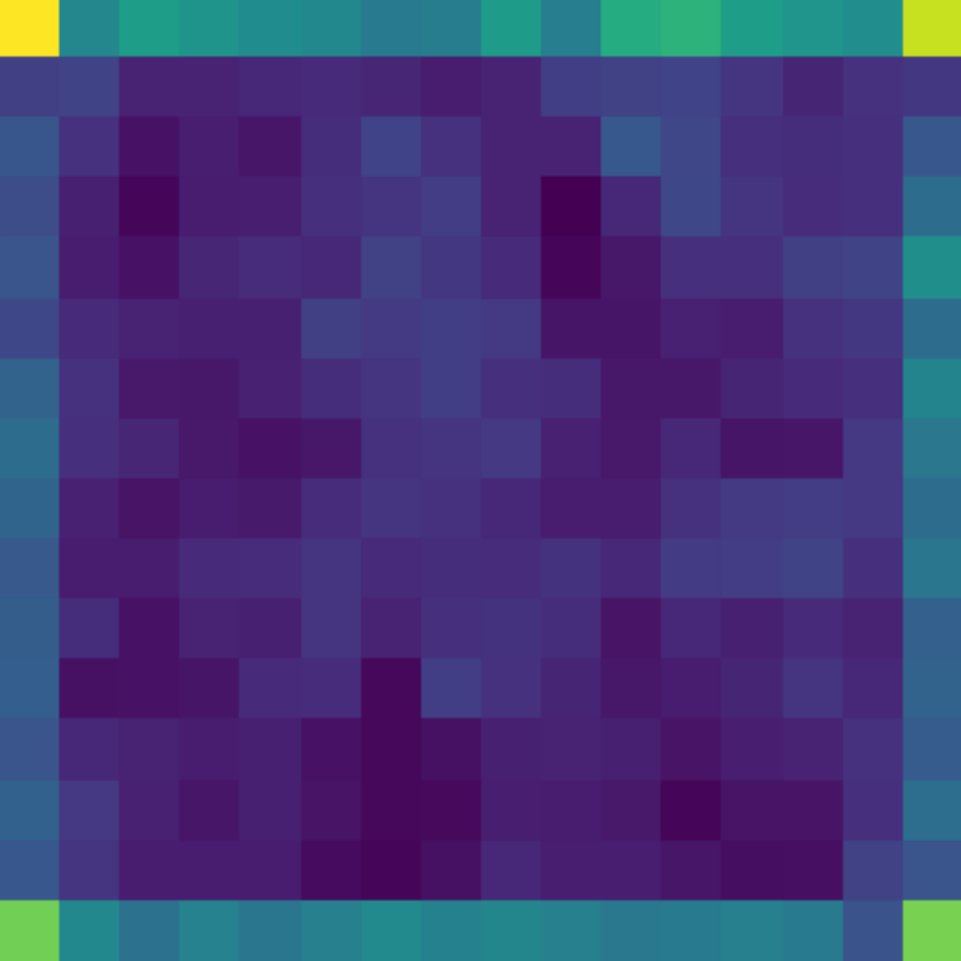}
\end{subfigure}
\begin{subfigure}{.05\textwidth}
  \centering
  \includegraphics[width=1.0\linewidth]{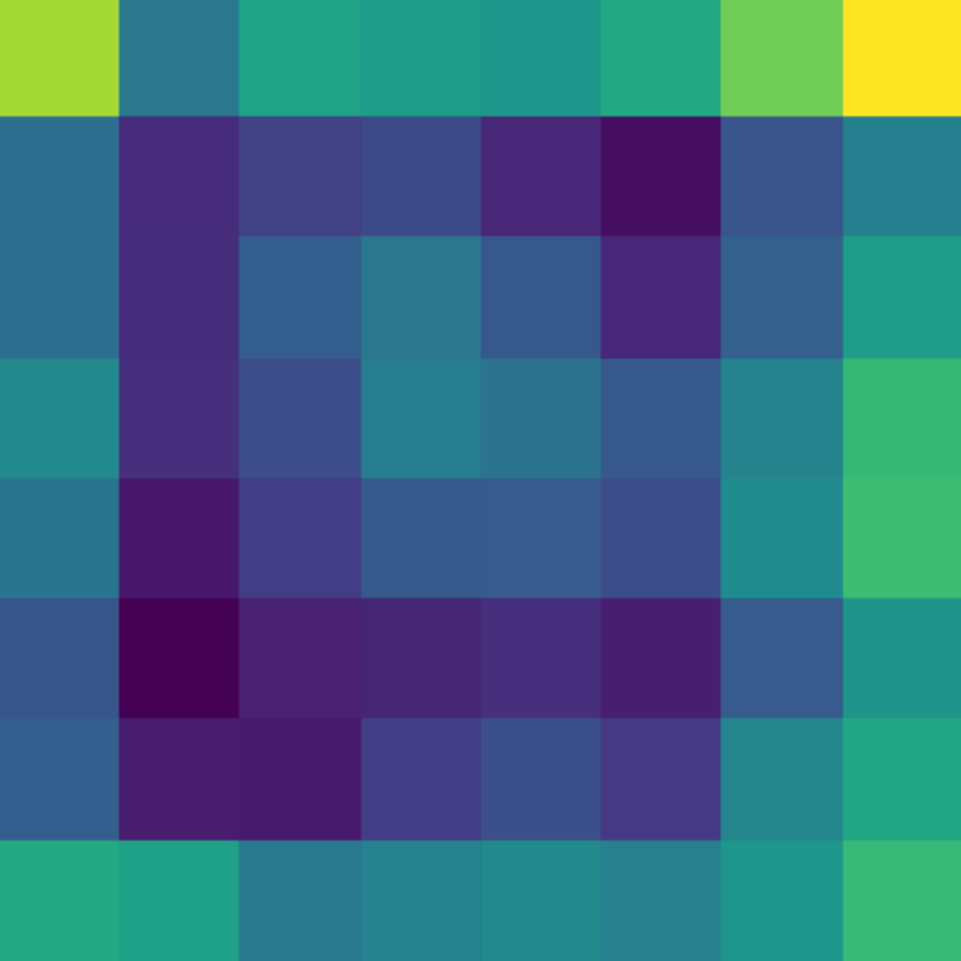}
\end{subfigure}
\begin{subfigure}{.05\textwidth}
  \centering
  \includegraphics[width=1.0\linewidth]{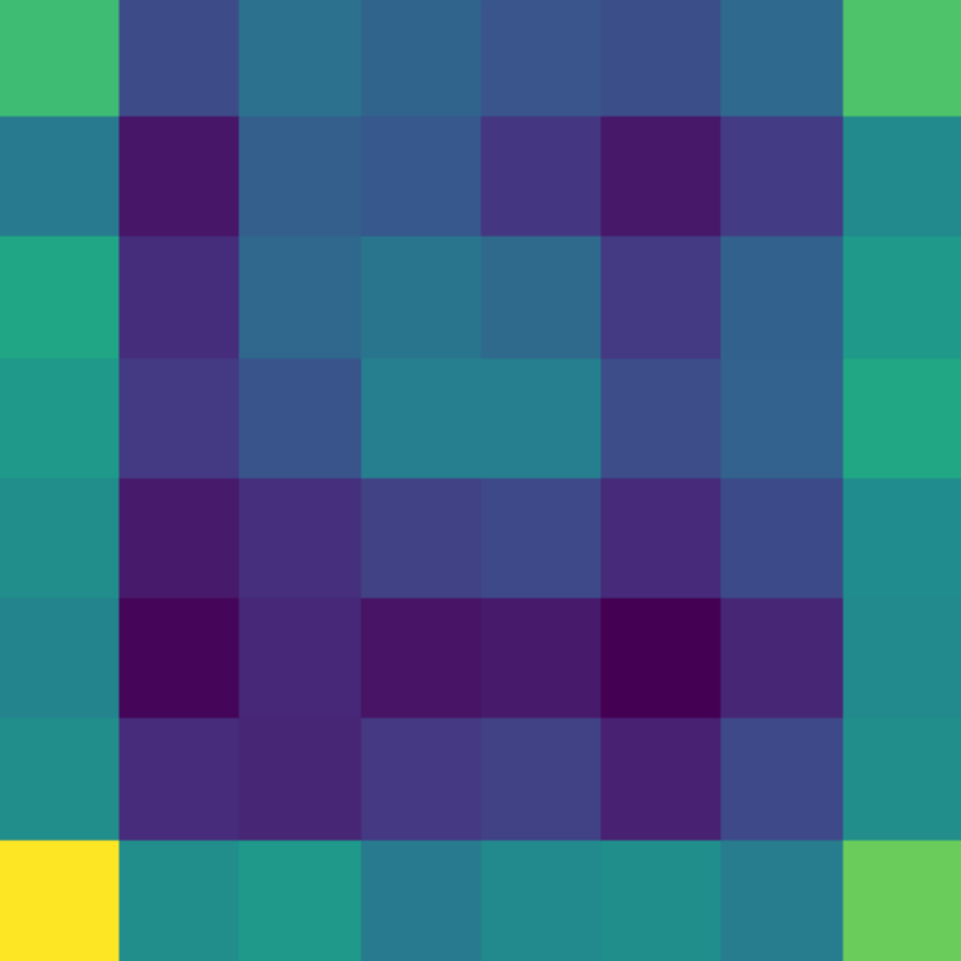}
\end{subfigure}
\begin{subfigure}{.05\textwidth}
  \centering
  \includegraphics[width=1.0\linewidth]{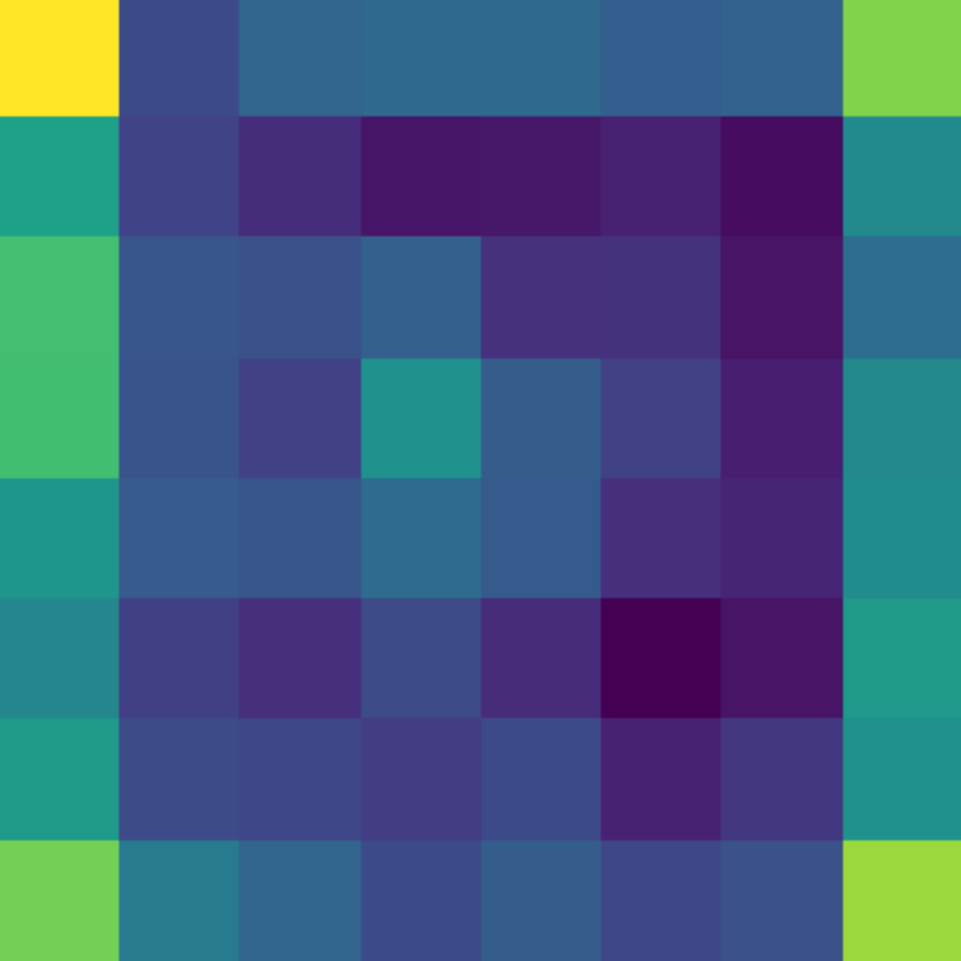}
\end{subfigure}
\begin{subfigure}{.05\textwidth}
  \centering
  \includegraphics[width=1.0\linewidth]{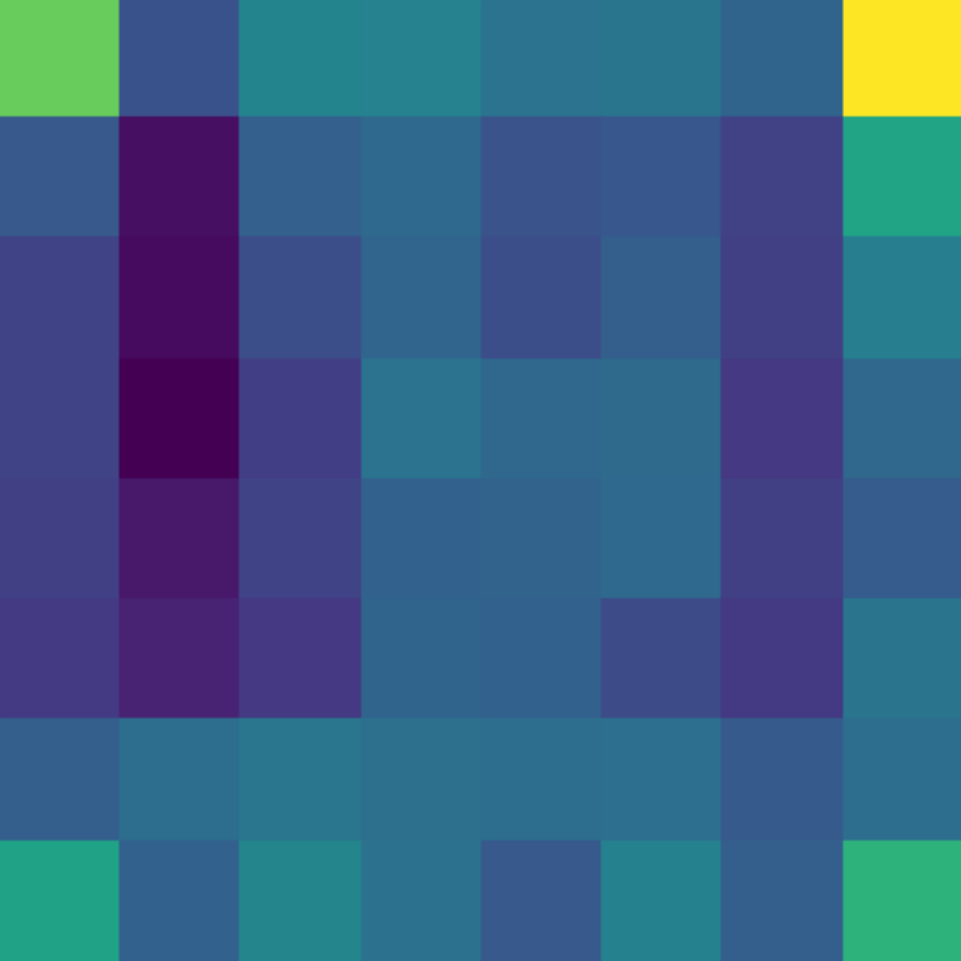}
\end{subfigure}
\begin{subfigure}{.05\textwidth}
  \centering
  \includegraphics[width=1.0\linewidth]{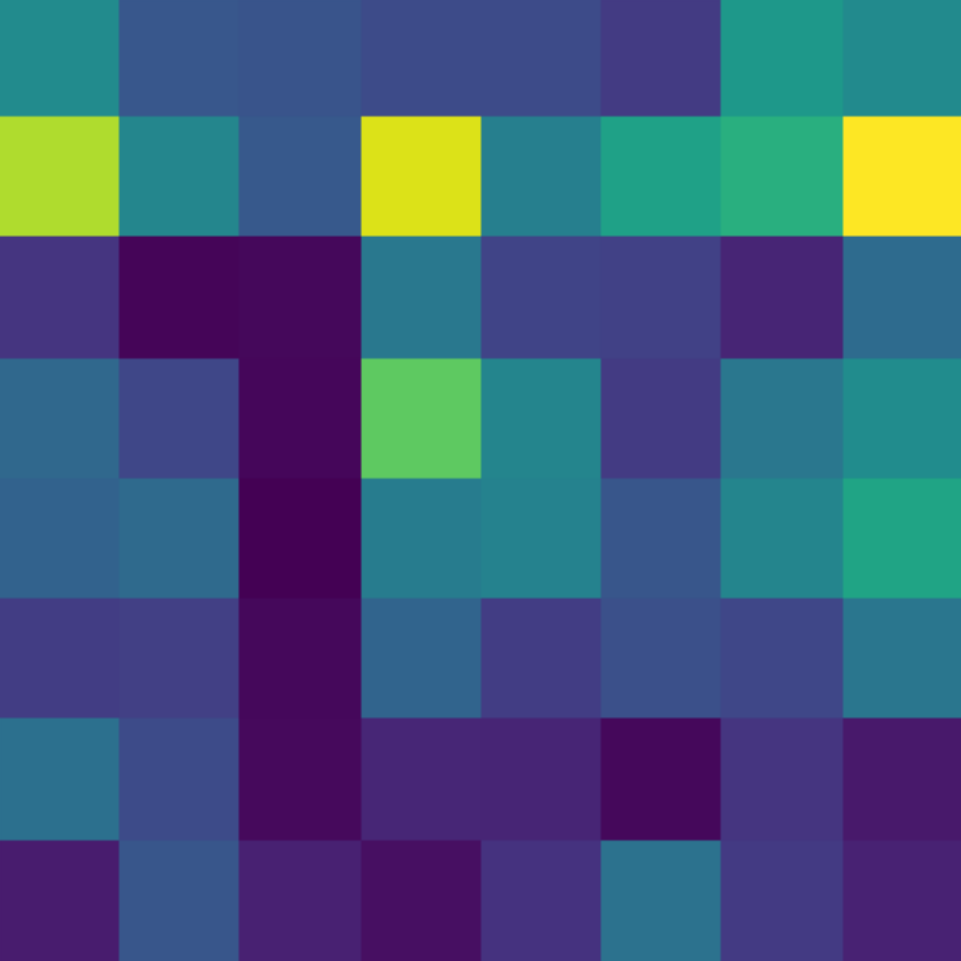}
\end{subfigure}
\begin{subfigure}{.05\textwidth}
  \centering
  \includegraphics[width=1.0\linewidth]{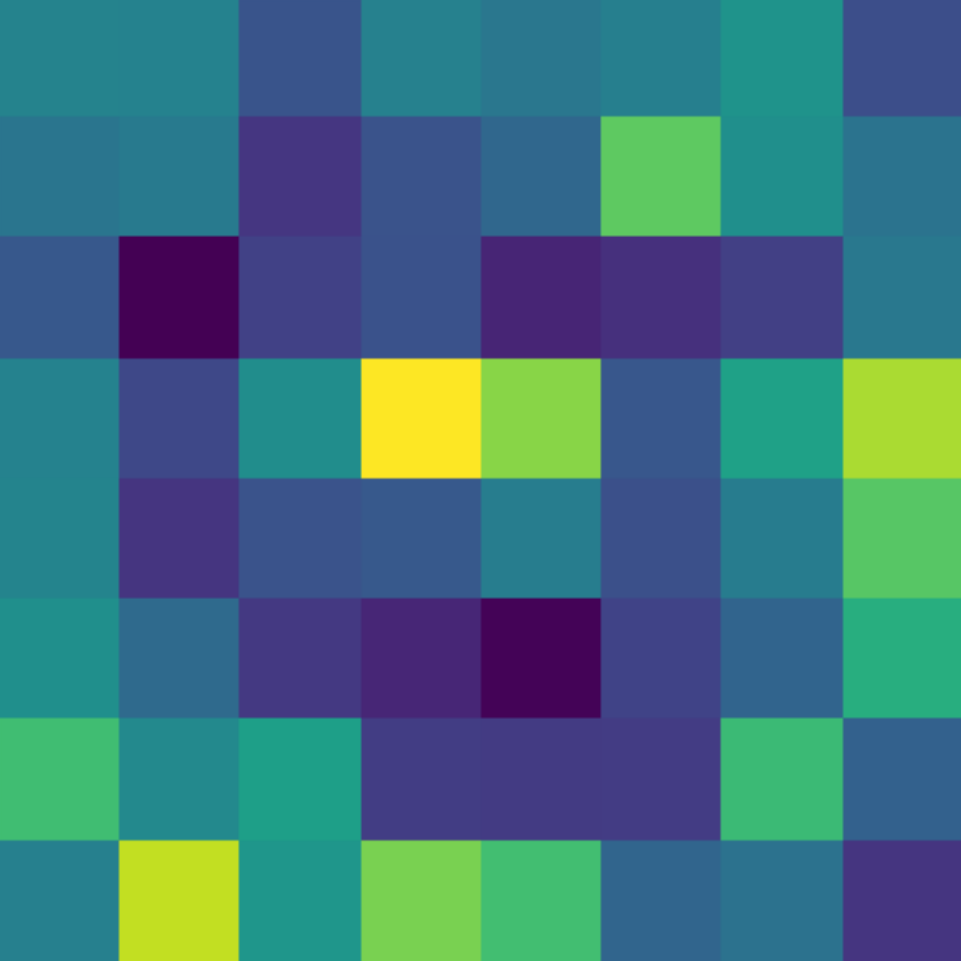}
\end{subfigure}
\begin{subfigure}{.05\textwidth}
  \centering
  \includegraphics[width=1.0\linewidth]{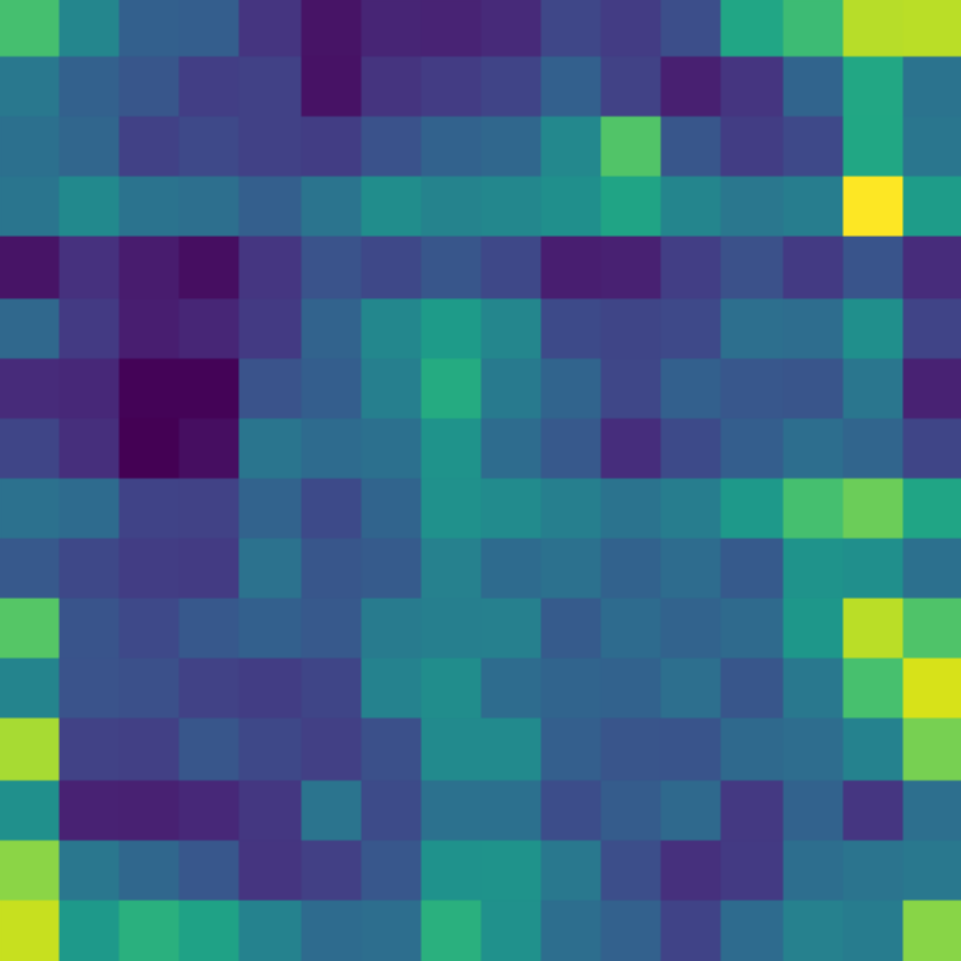}
\end{subfigure}
\begin{subfigure}{.05\textwidth}
  \centering
  \includegraphics[width=1.0\linewidth]{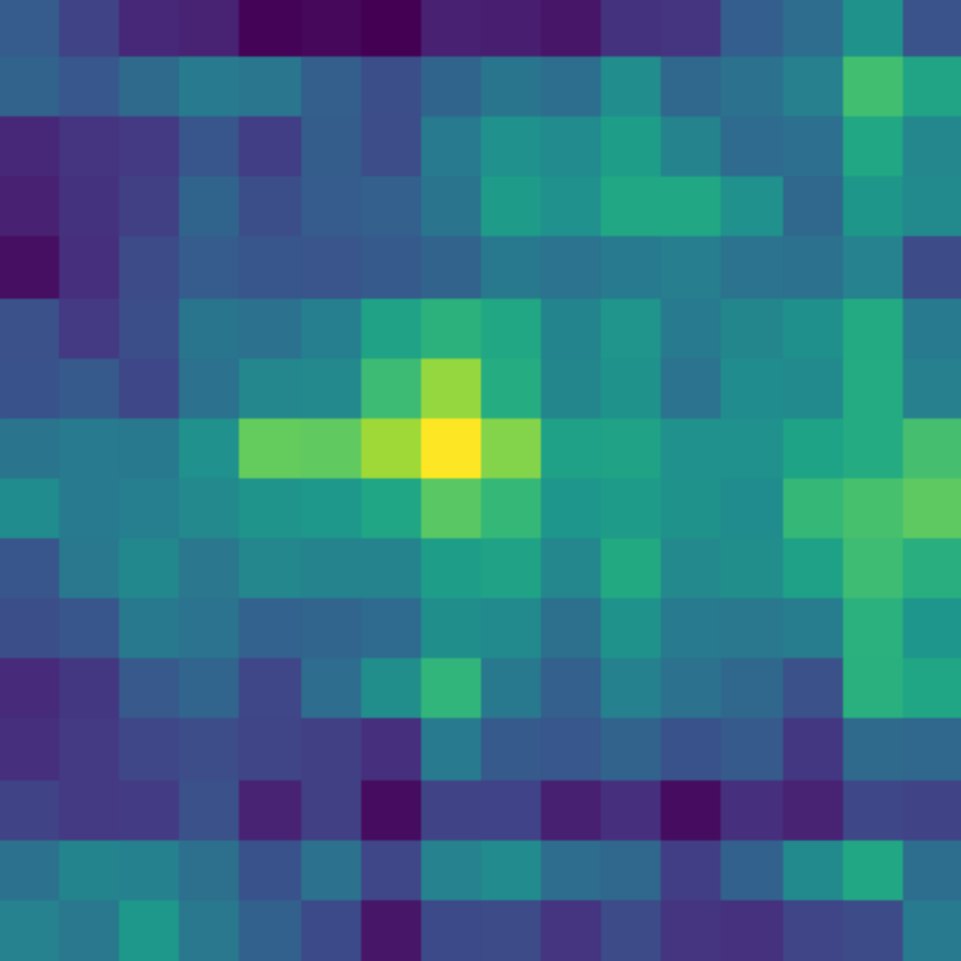}
\end{subfigure}
\begin{subfigure}{.05\textwidth}
  \centering
  \includegraphics[width=1.0\linewidth]{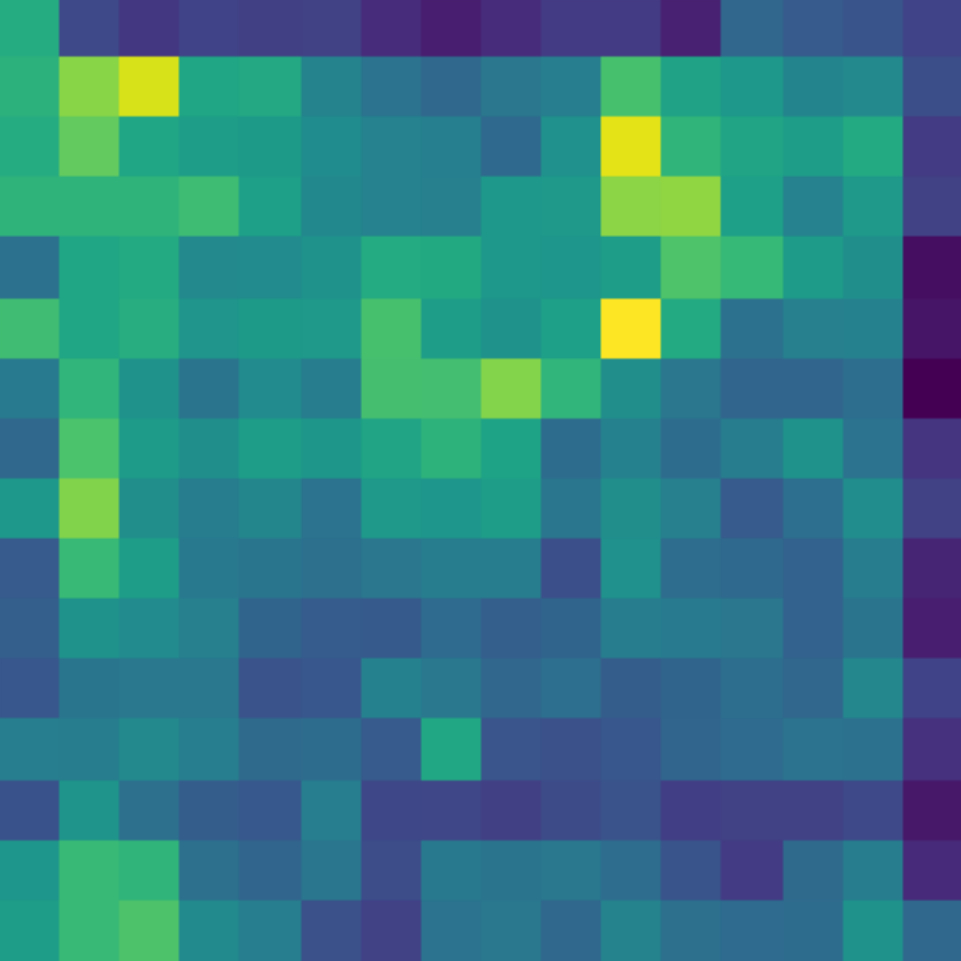}
\end{subfigure}
\begin{subfigure}{.05\textwidth}
  \centering
  \includegraphics[width=1.0\linewidth]{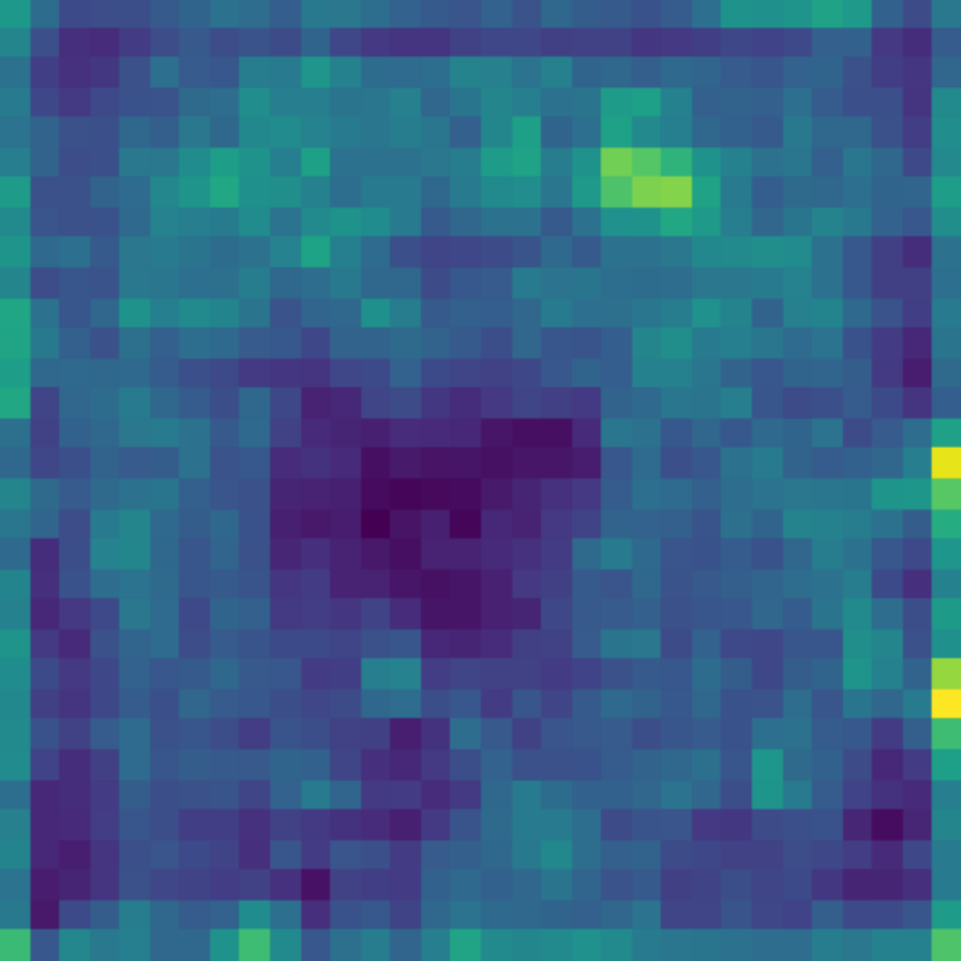}
\end{subfigure}
\begin{subfigure}{.05\textwidth}
  \centering
  \includegraphics[width=1.0\linewidth]{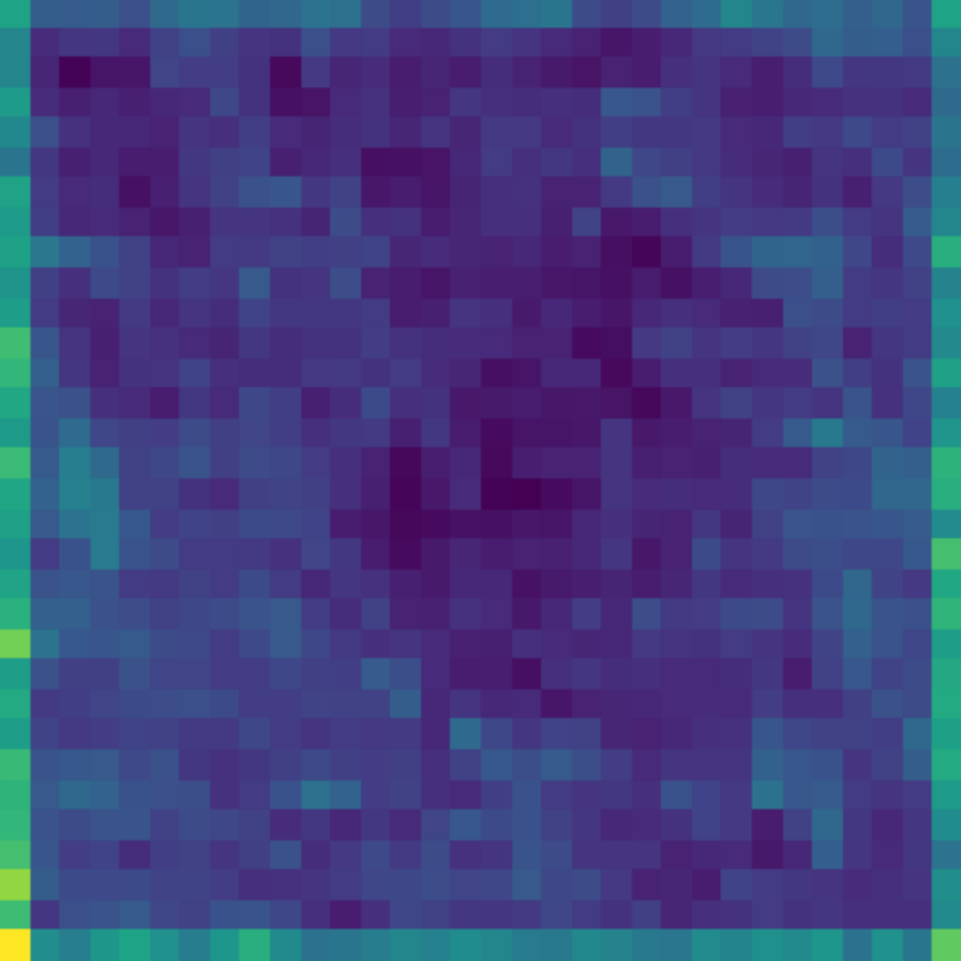}
\end{subfigure}
\begin{subfigure}{.05\textwidth}
  \centering
  \includegraphics[width=1.0\linewidth]{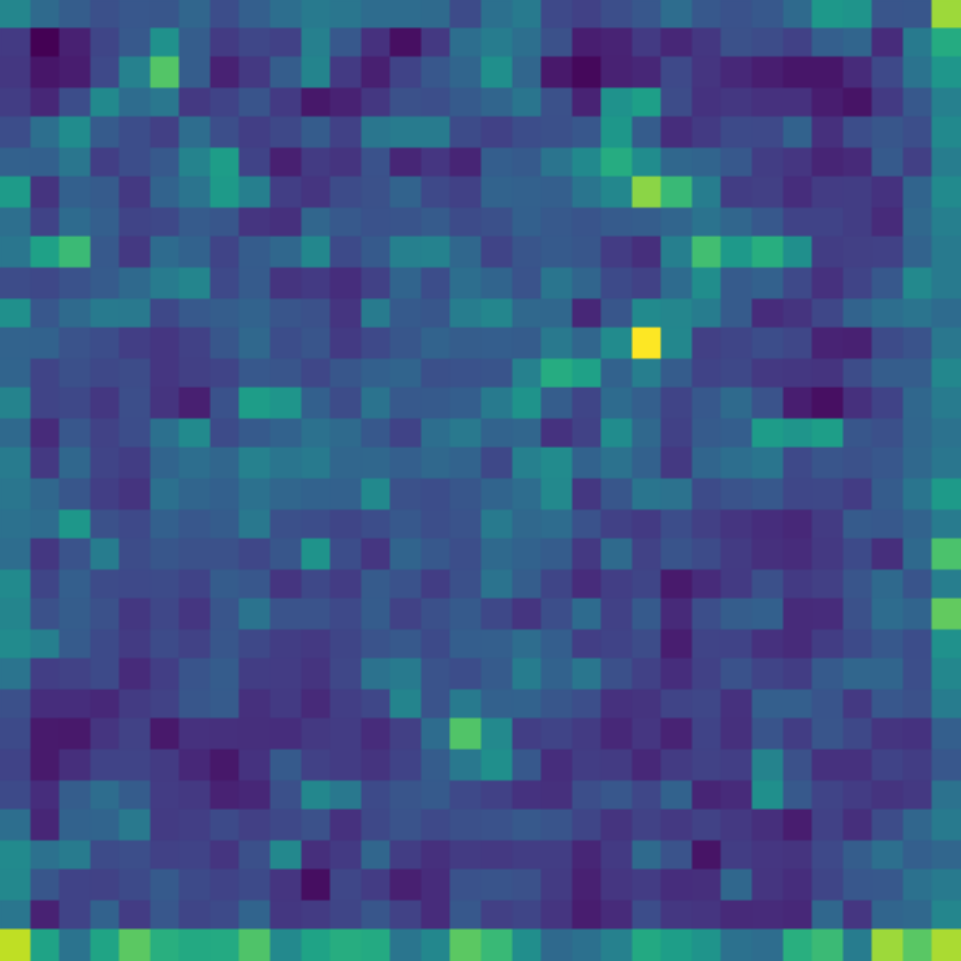}
\end{subfigure}\vspace{10pt}
\\

\begin{subfigure}{.05\textwidth}
  \centering
  \includegraphics[width=1.0\linewidth]{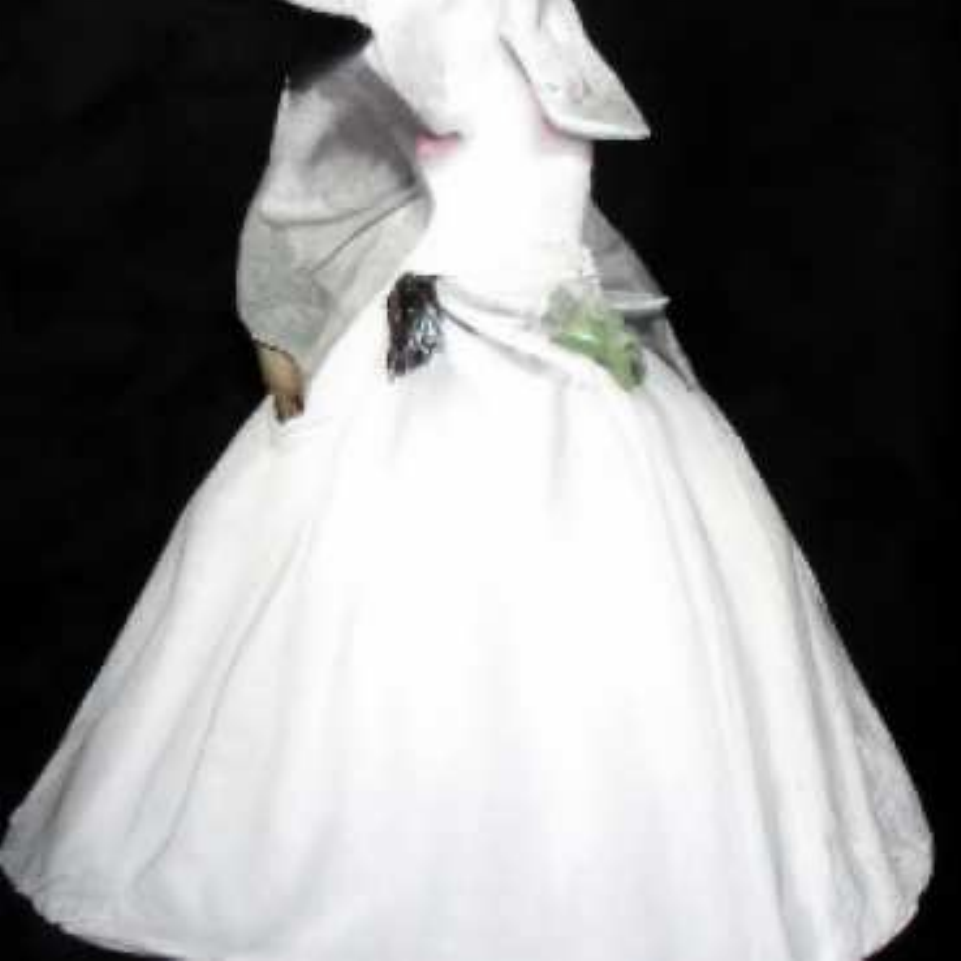}
\end{subfigure}
&
\begin{subfigure}{.05\textwidth}
  \centering
  \includegraphics[width=1.0\linewidth]{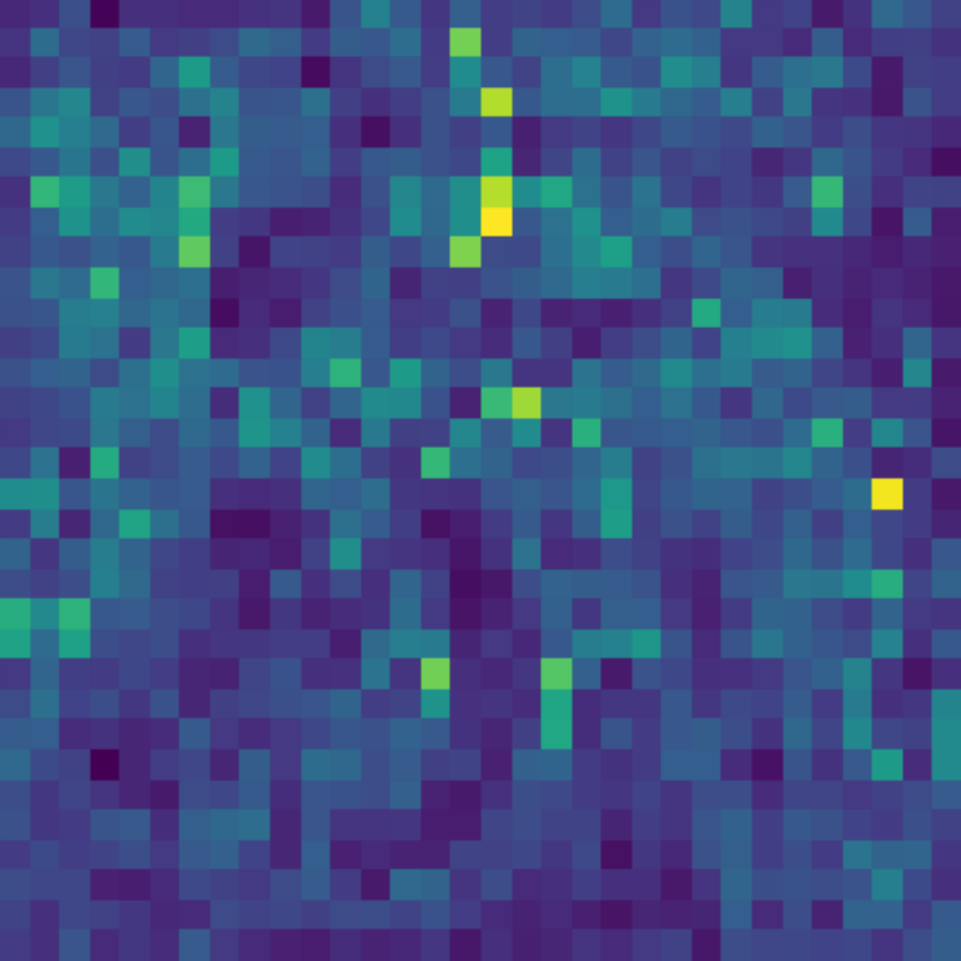}
\end{subfigure}
\begin{subfigure}{.05\textwidth}
  \centering
  \includegraphics[width=1.0\linewidth]{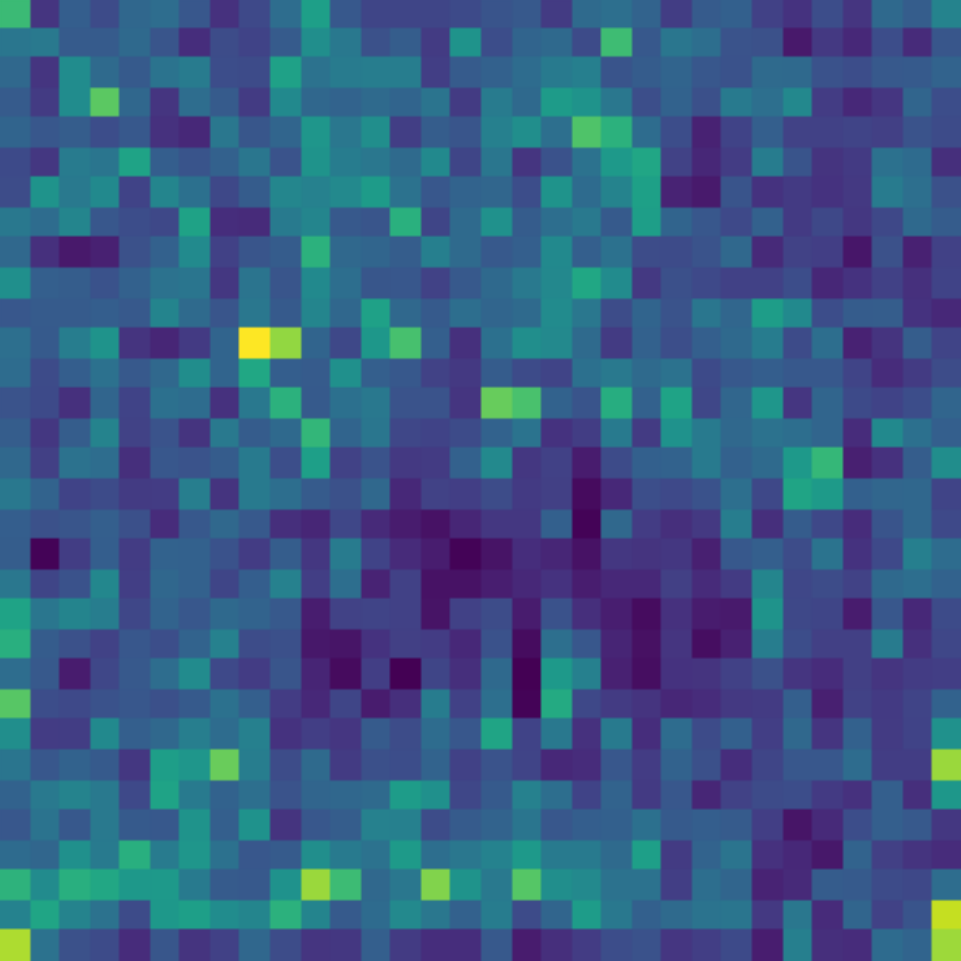}
\end{subfigure}
\begin{subfigure}{.05\textwidth}
  \centering
  \includegraphics[width=1.0\linewidth]{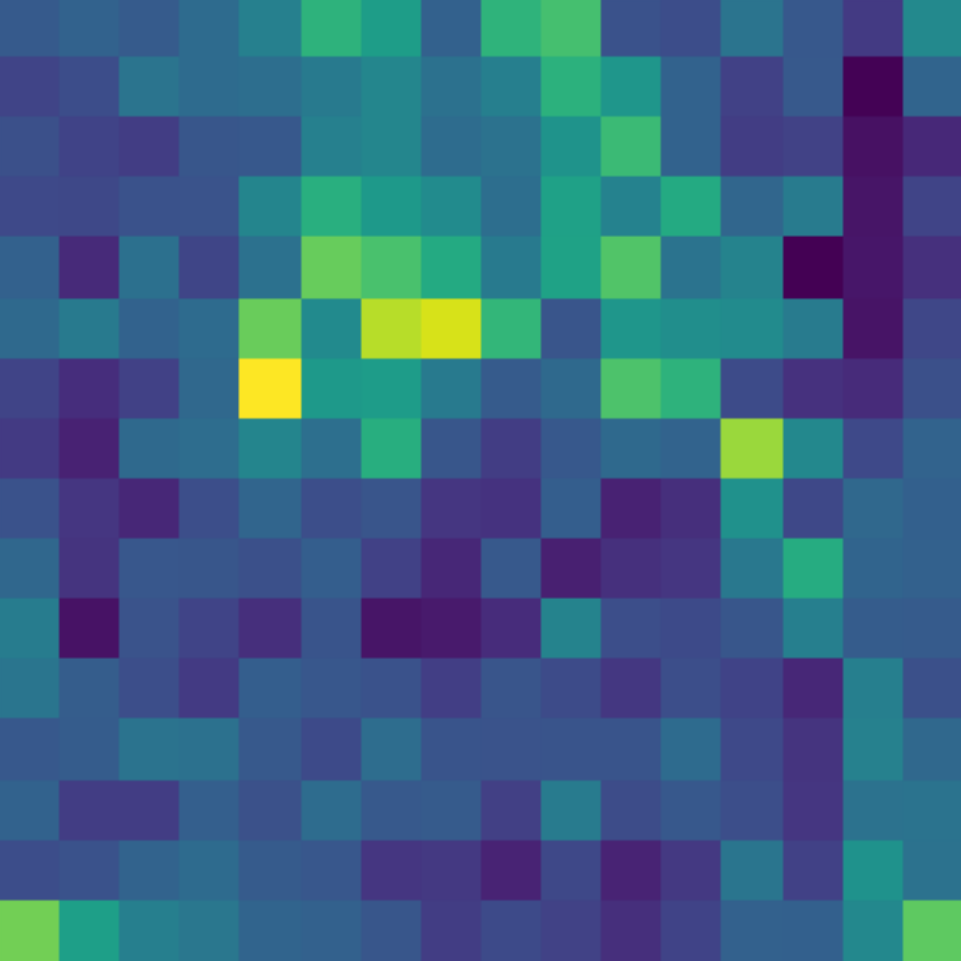}
\end{subfigure}
\begin{subfigure}{.05\textwidth}
  \centering
  \includegraphics[width=1.0\linewidth]{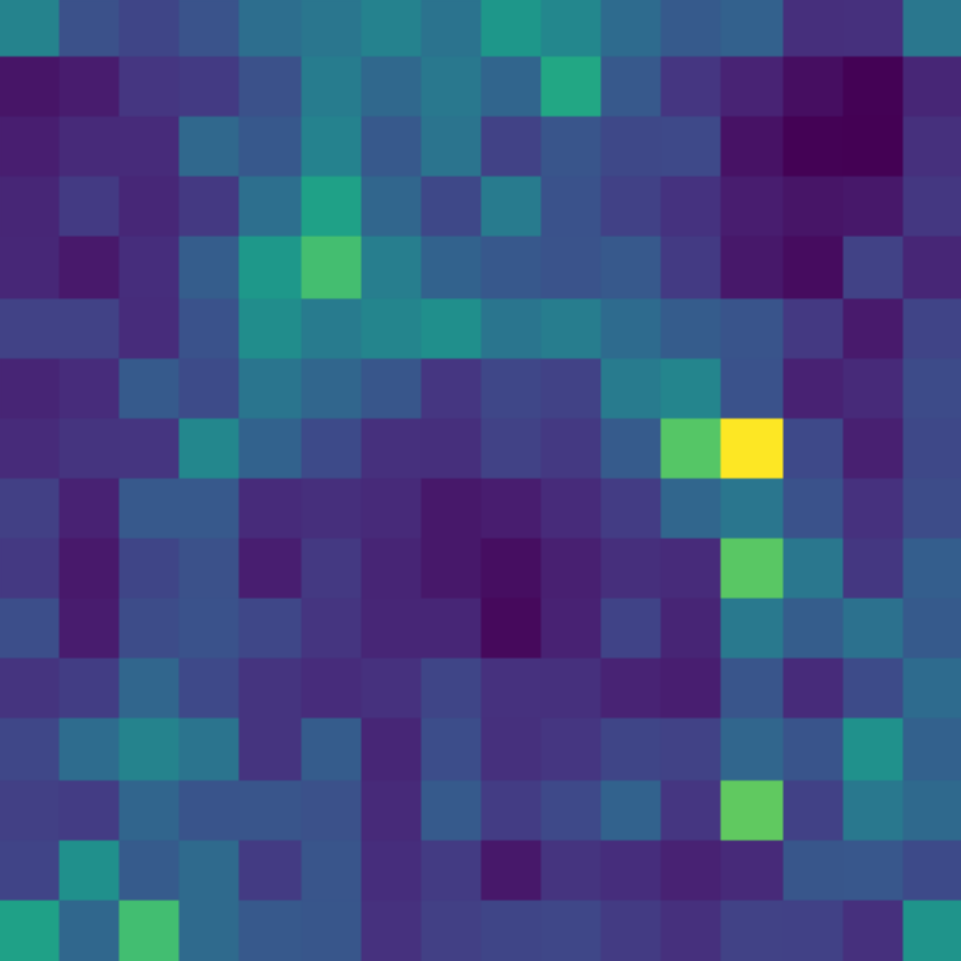}
\end{subfigure}
\begin{subfigure}{.05\textwidth}
  \centering
  \includegraphics[width=1.0\linewidth]{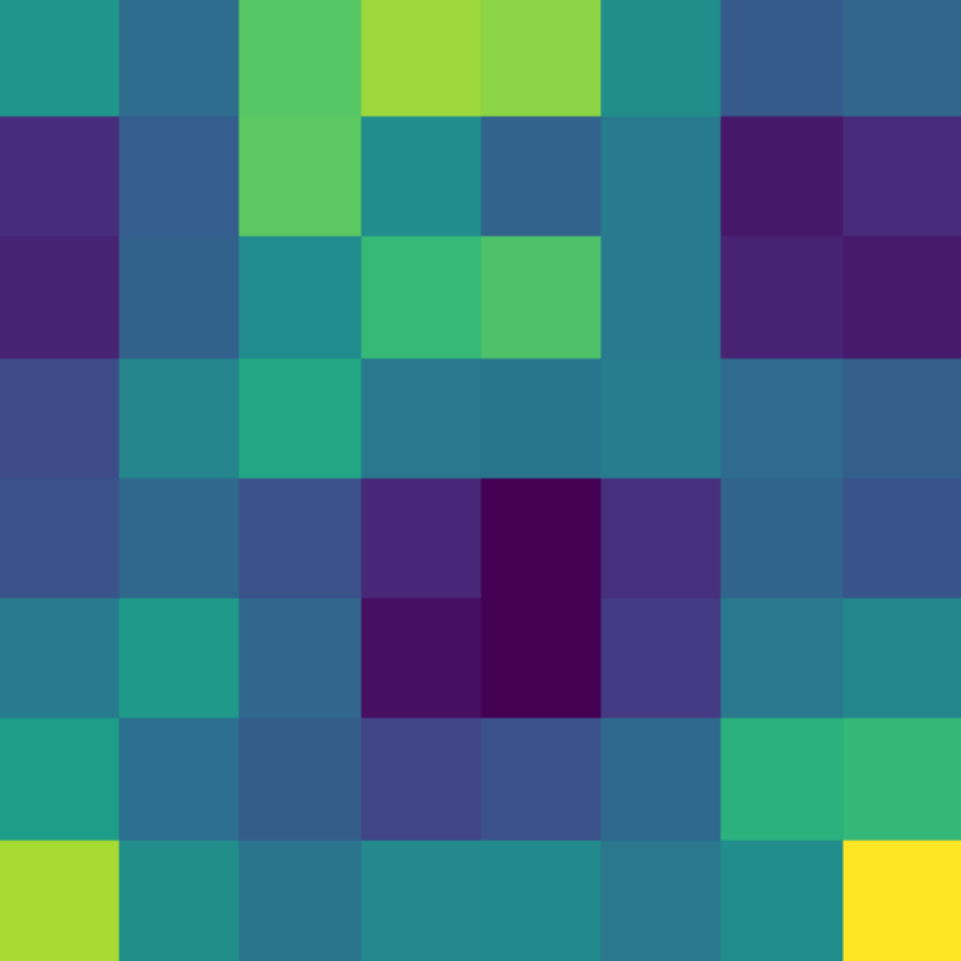}
\end{subfigure}
\begin{subfigure}{.05\textwidth}
  \centering
  \includegraphics[width=1.0\linewidth]{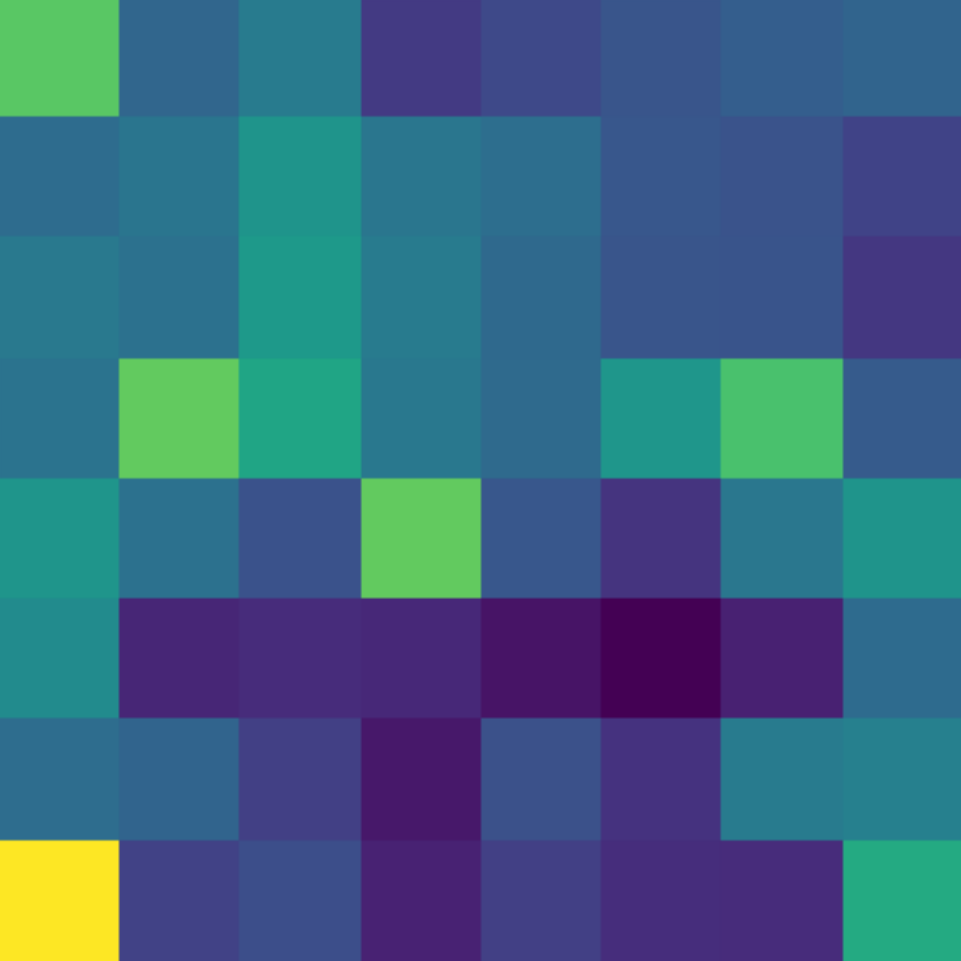}
\end{subfigure}
\begin{subfigure}{.05\textwidth}
  \centering
  \includegraphics[width=1.0\linewidth]{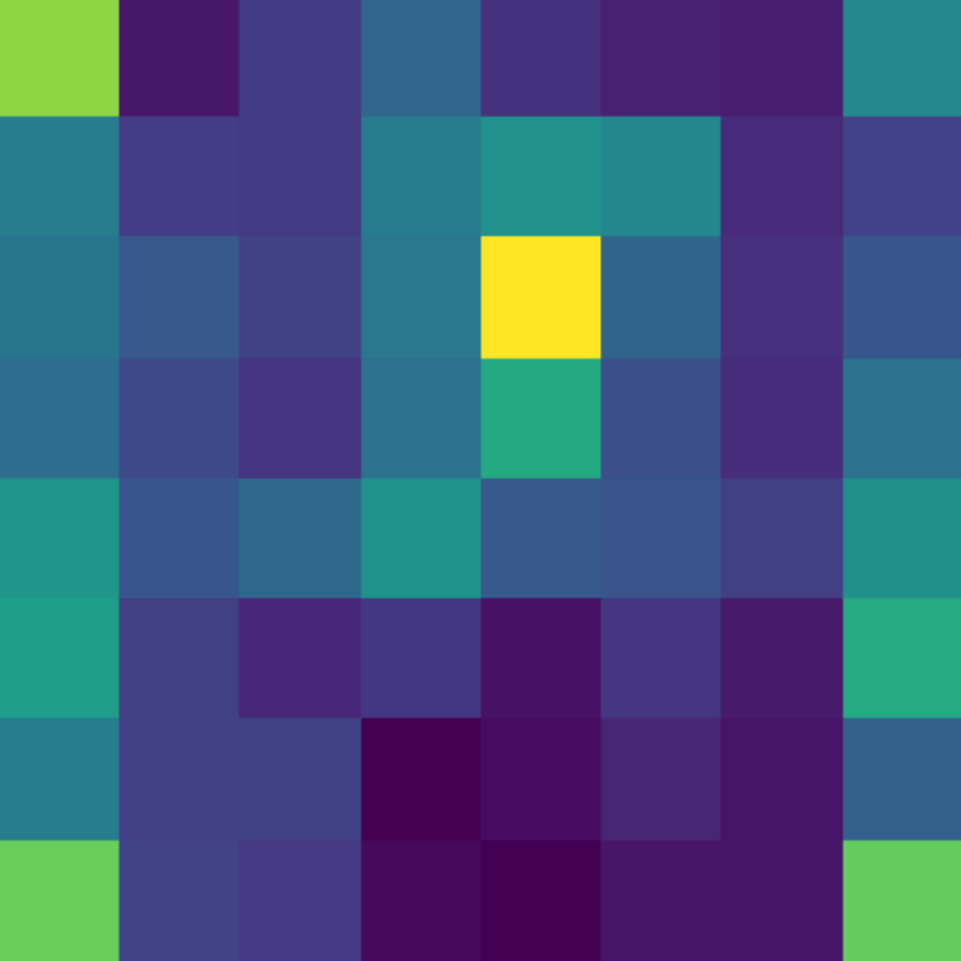}
\end{subfigure}
\begin{subfigure}{.05\textwidth}
  \centering
  \includegraphics[width=1.0\linewidth]{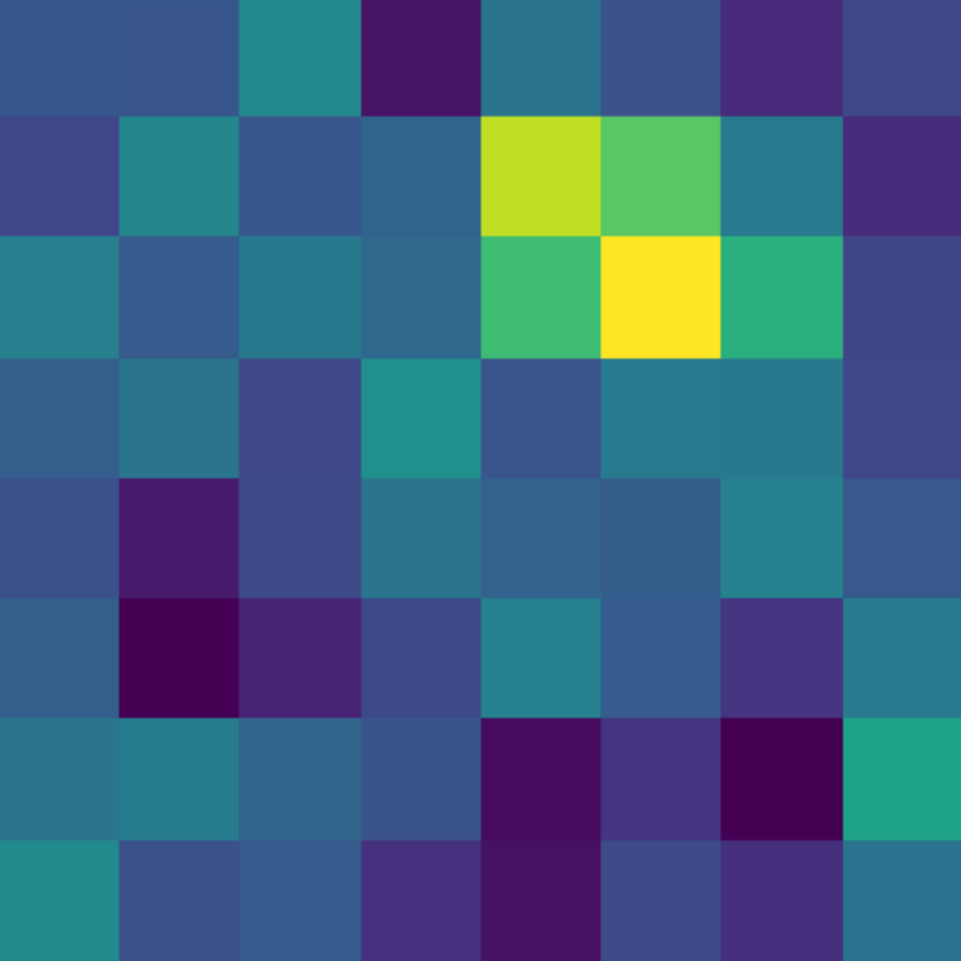}
\end{subfigure}
\begin{subfigure}{.05\textwidth}
  \centering
  \includegraphics[width=1.0\linewidth]{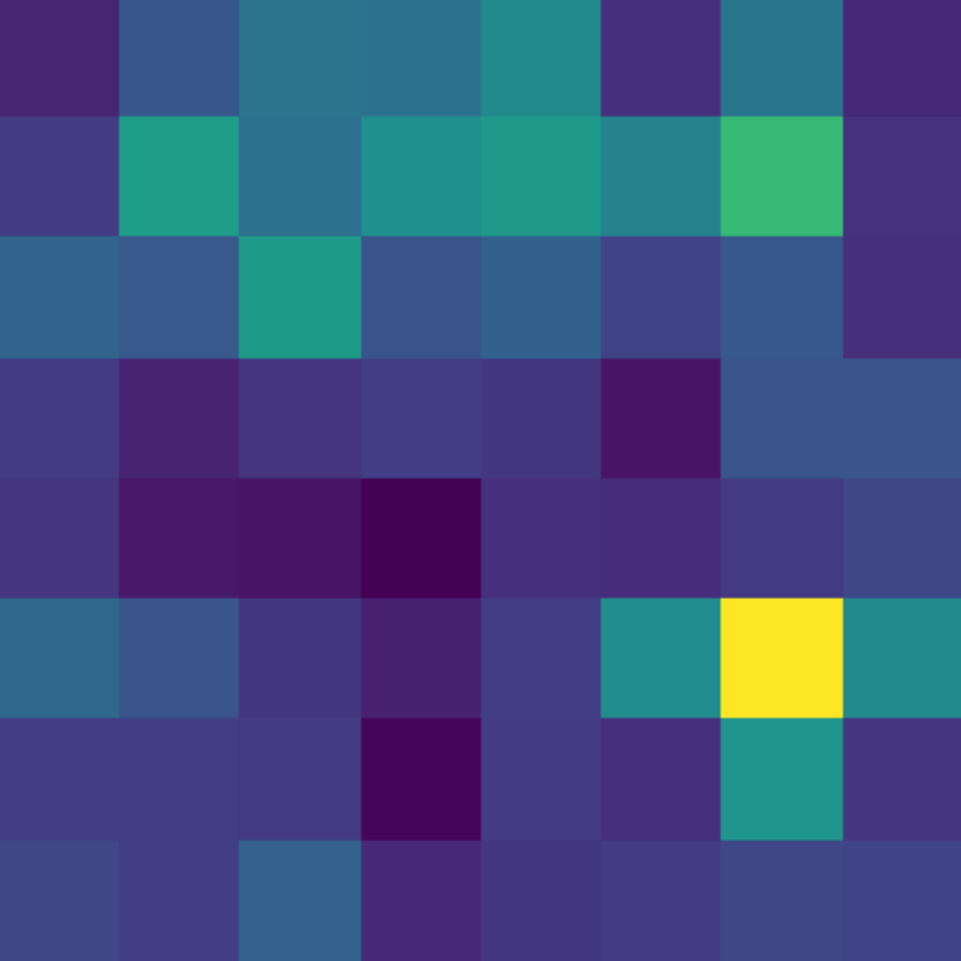}
\end{subfigure}
\begin{subfigure}{.05\textwidth}
  \centering
  \includegraphics[width=1.0\linewidth]{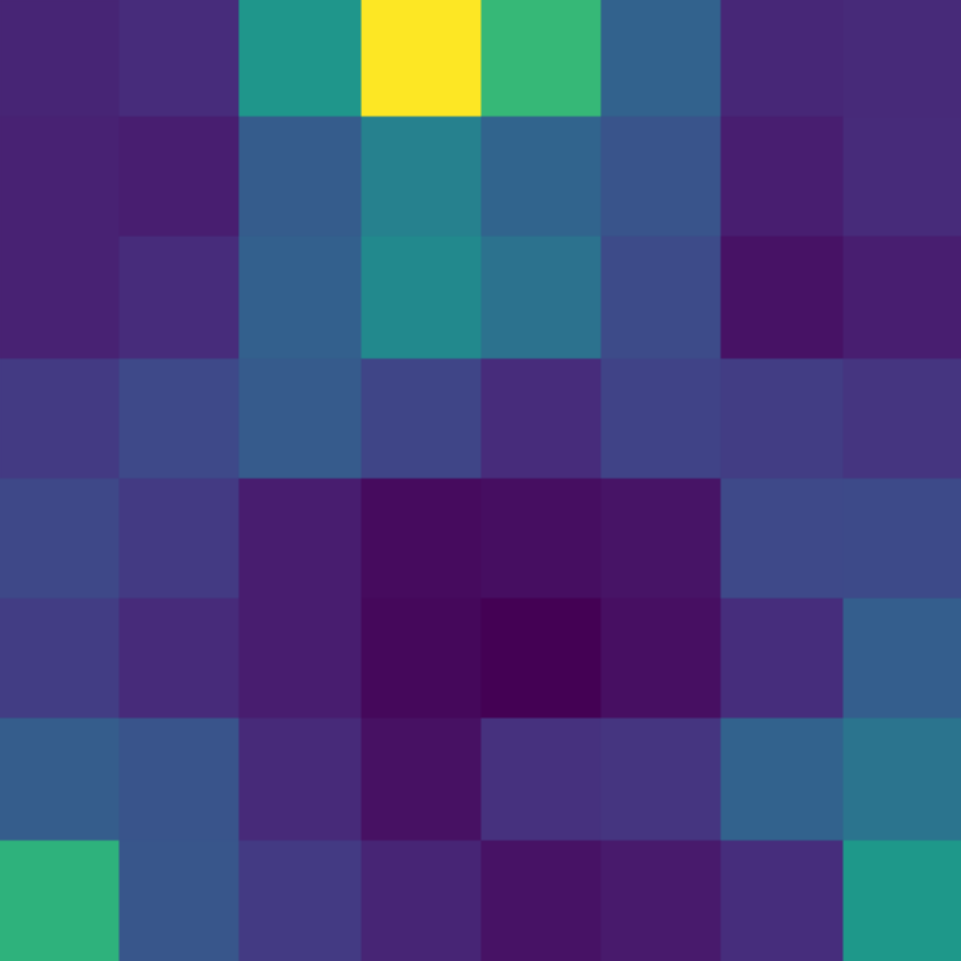}
\end{subfigure}
\begin{subfigure}{.05\textwidth}
  \centering
  \includegraphics[width=1.0\linewidth]{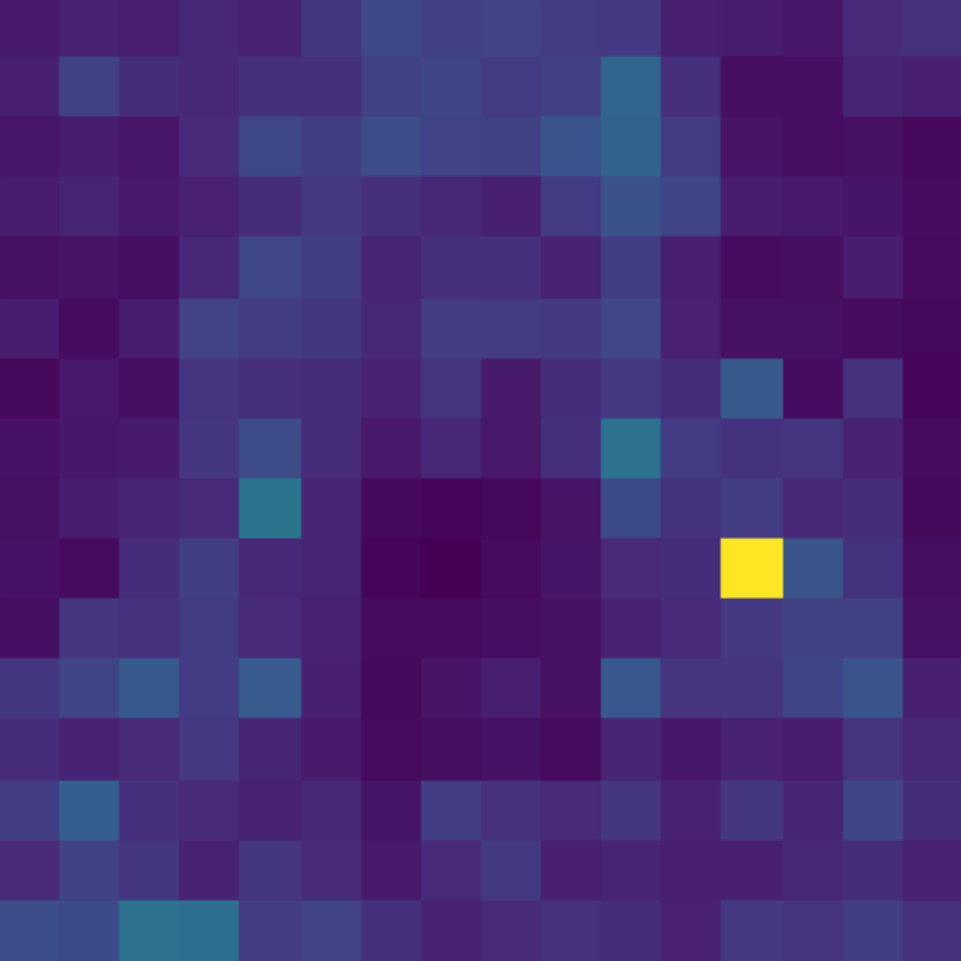}
\end{subfigure}
\begin{subfigure}{.05\textwidth}
  \centering
  \includegraphics[width=1.0\linewidth]{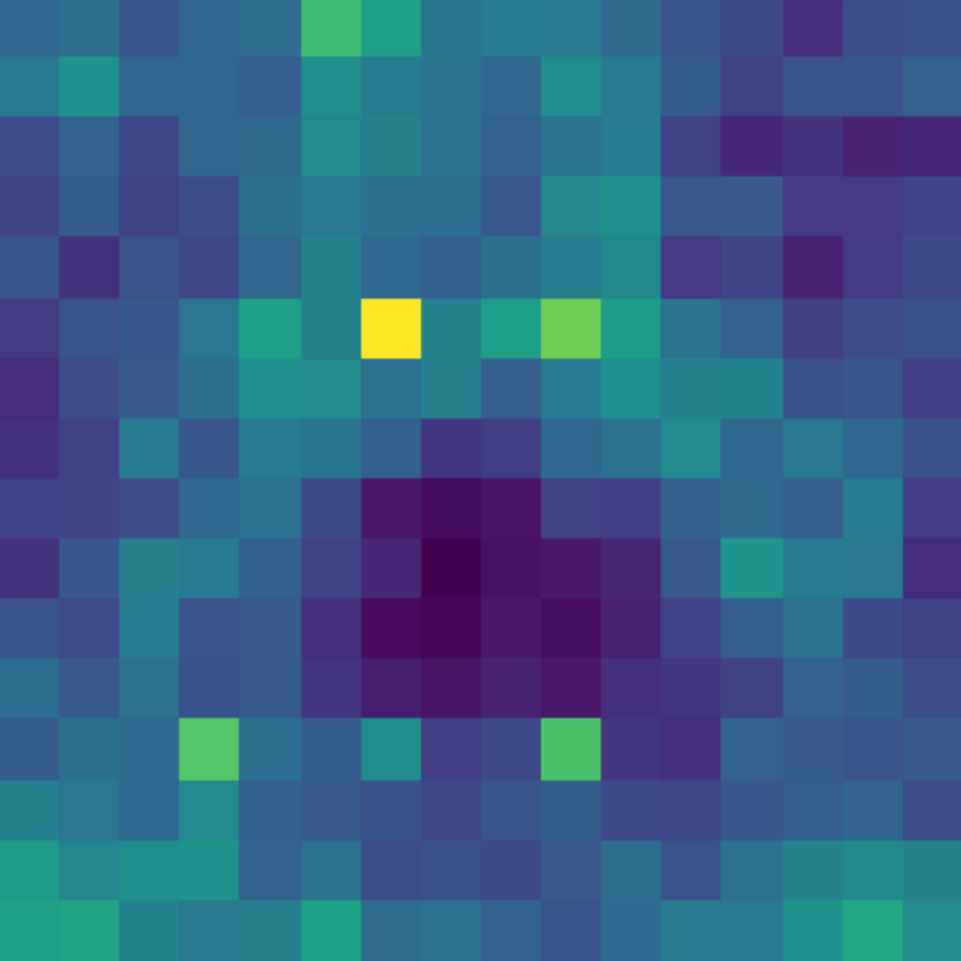}
\end{subfigure}
\begin{subfigure}{.05\textwidth}
  \centering
  \includegraphics[width=1.0\linewidth]{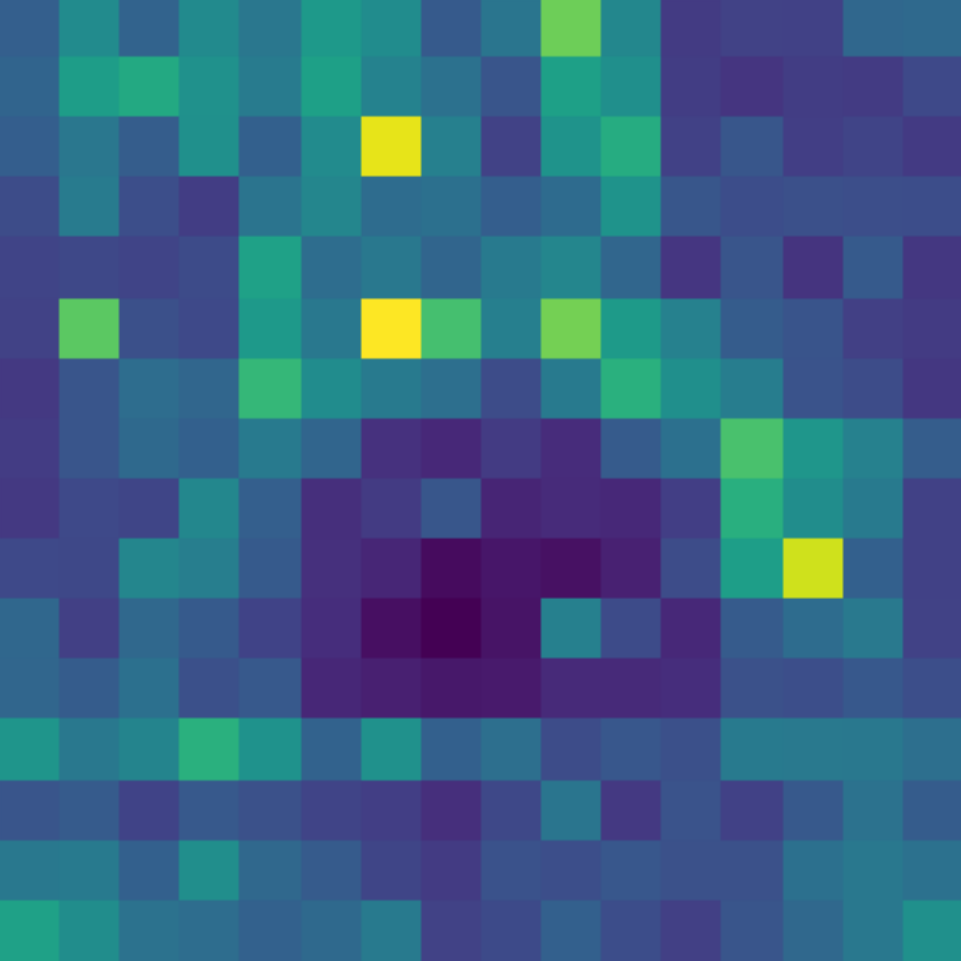}
\end{subfigure}
\begin{subfigure}{.05\textwidth}
  \centering
  \includegraphics[width=1.0\linewidth]{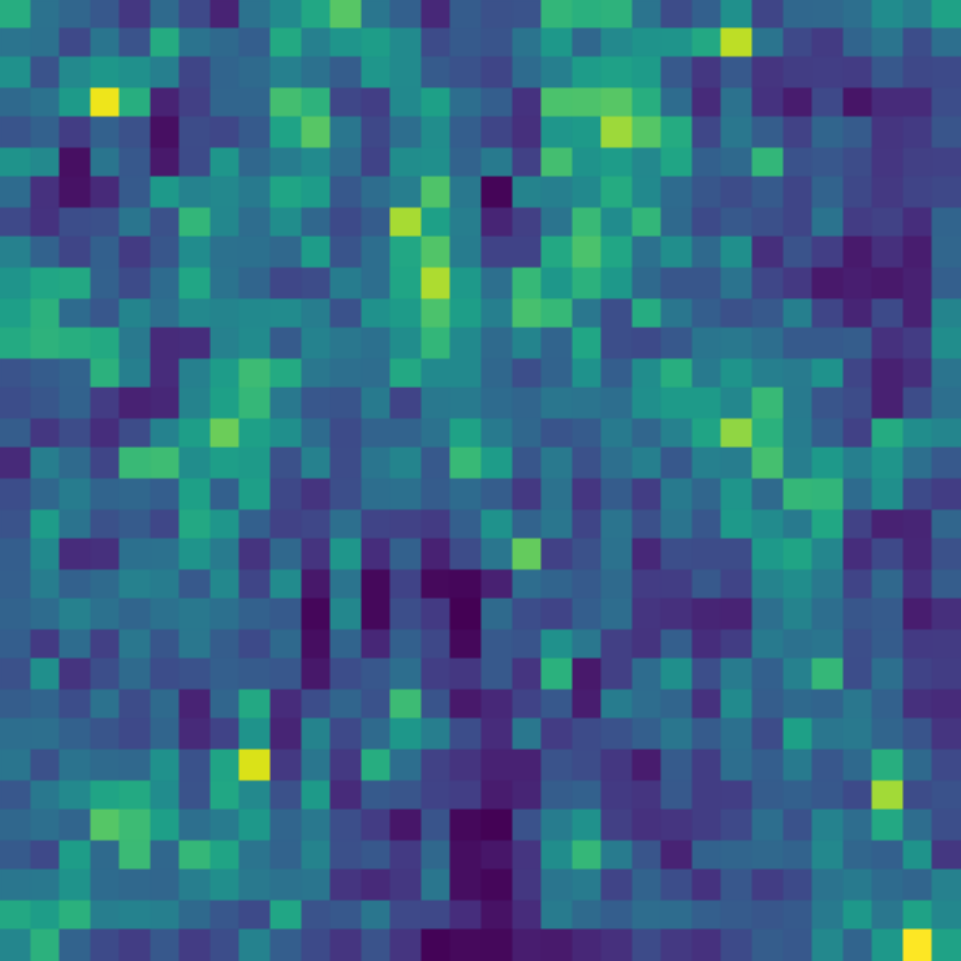}
\end{subfigure}
\begin{subfigure}{.05\textwidth}
  \centering
  \includegraphics[width=1.0\linewidth]{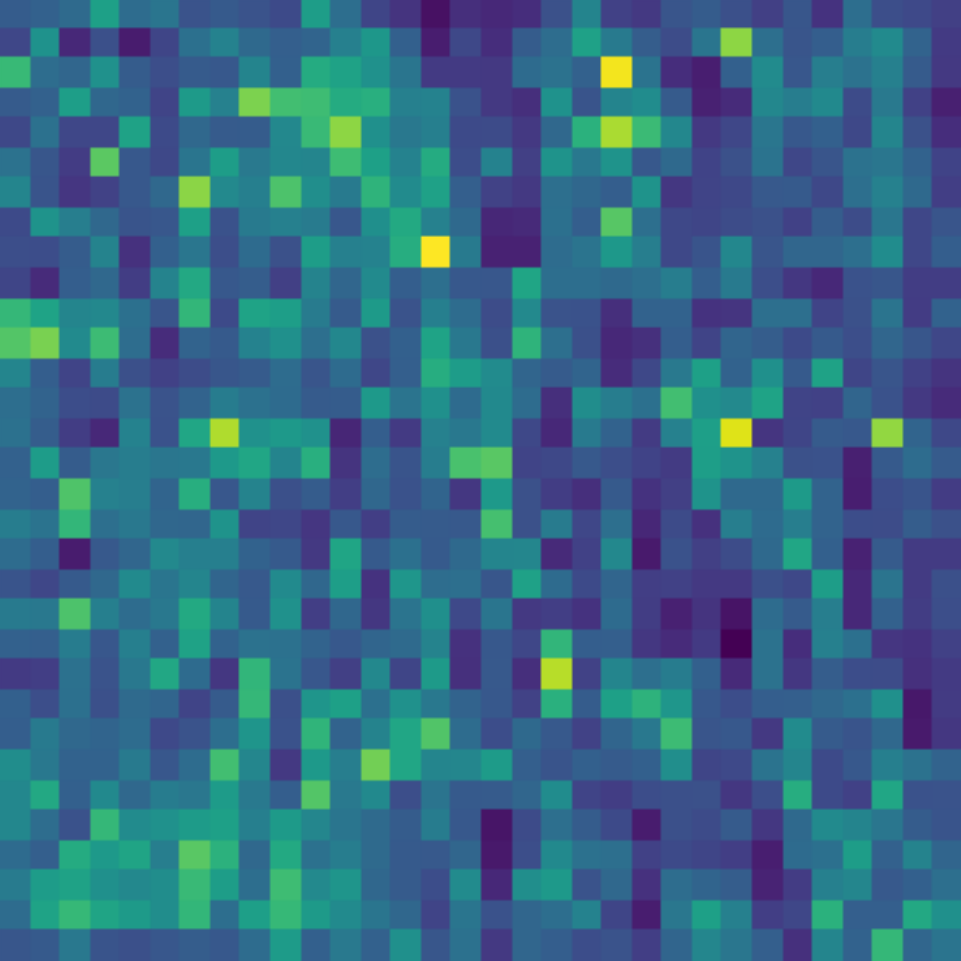}
\end{subfigure}
\begin{subfigure}{.05\textwidth}
  \centering
  \includegraphics[width=1.0\linewidth]{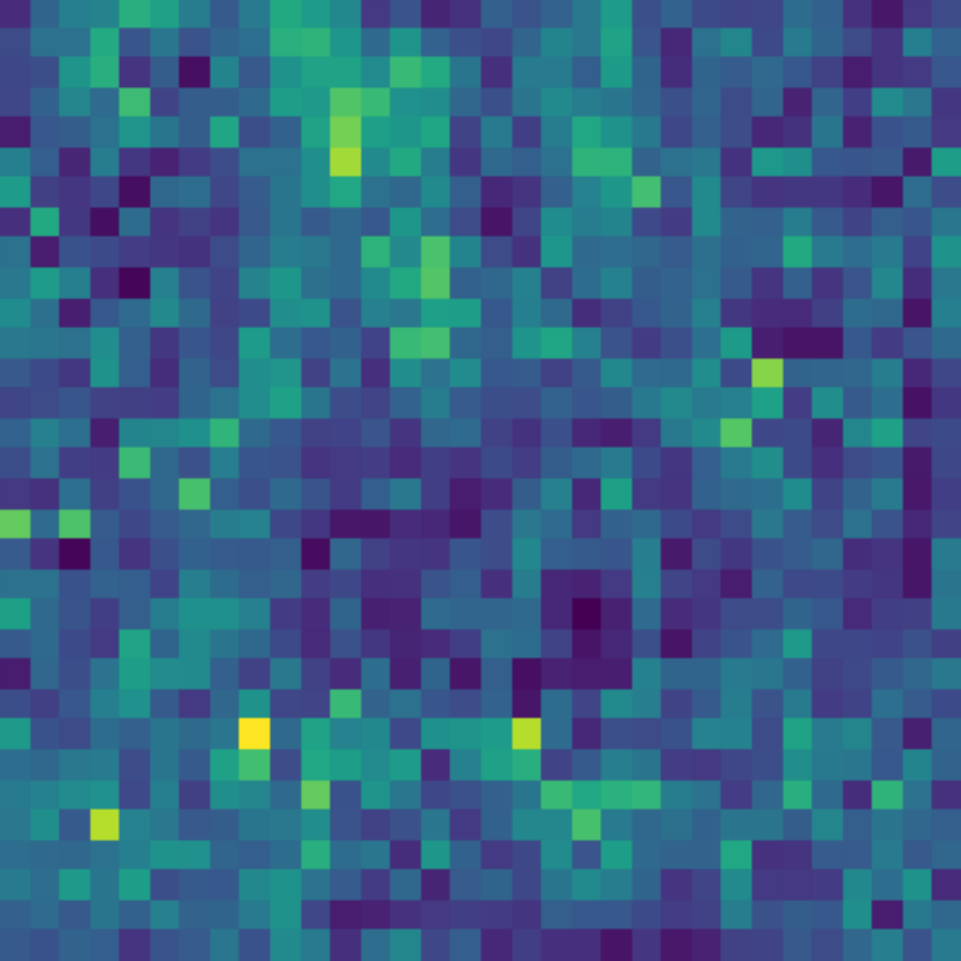}
\end{subfigure}
\\
&
\begin{subfigure}{.05\textwidth}
  \centering
  \includegraphics[width=1.0\linewidth]{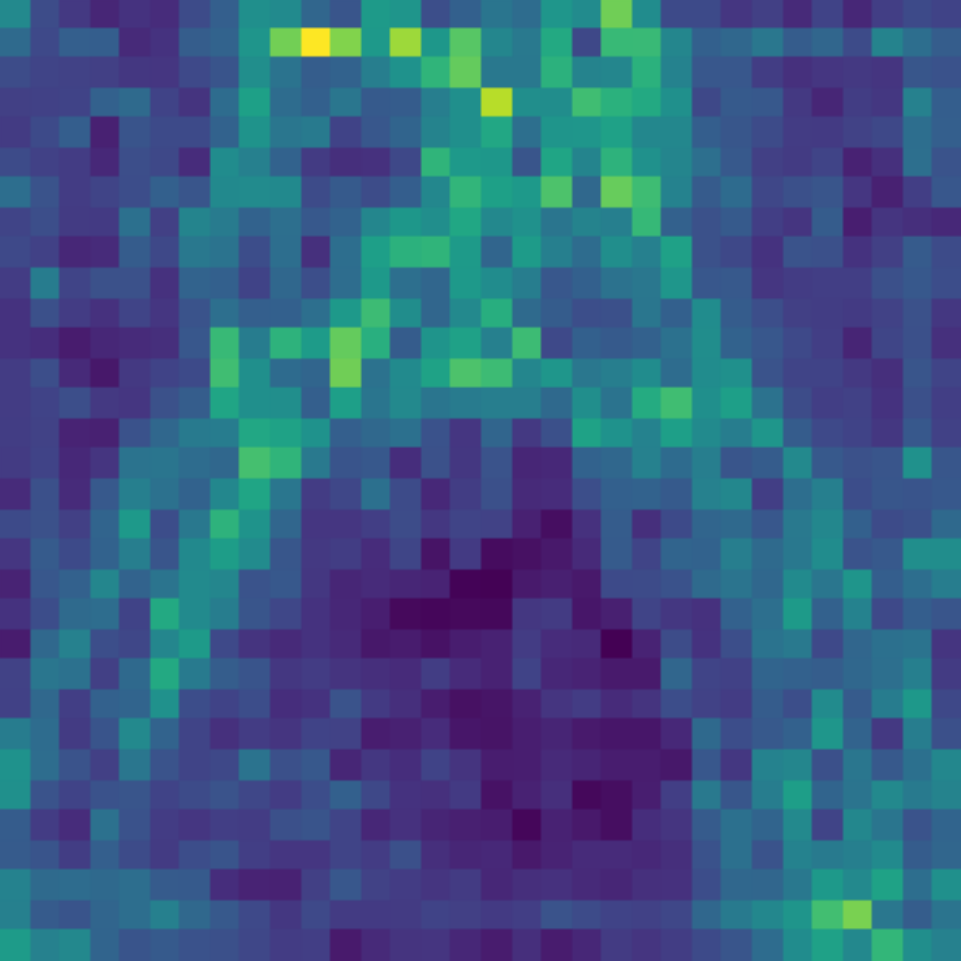}
\end{subfigure}
\begin{subfigure}{.05\textwidth}
  \centering
  \includegraphics[width=1.0\linewidth]{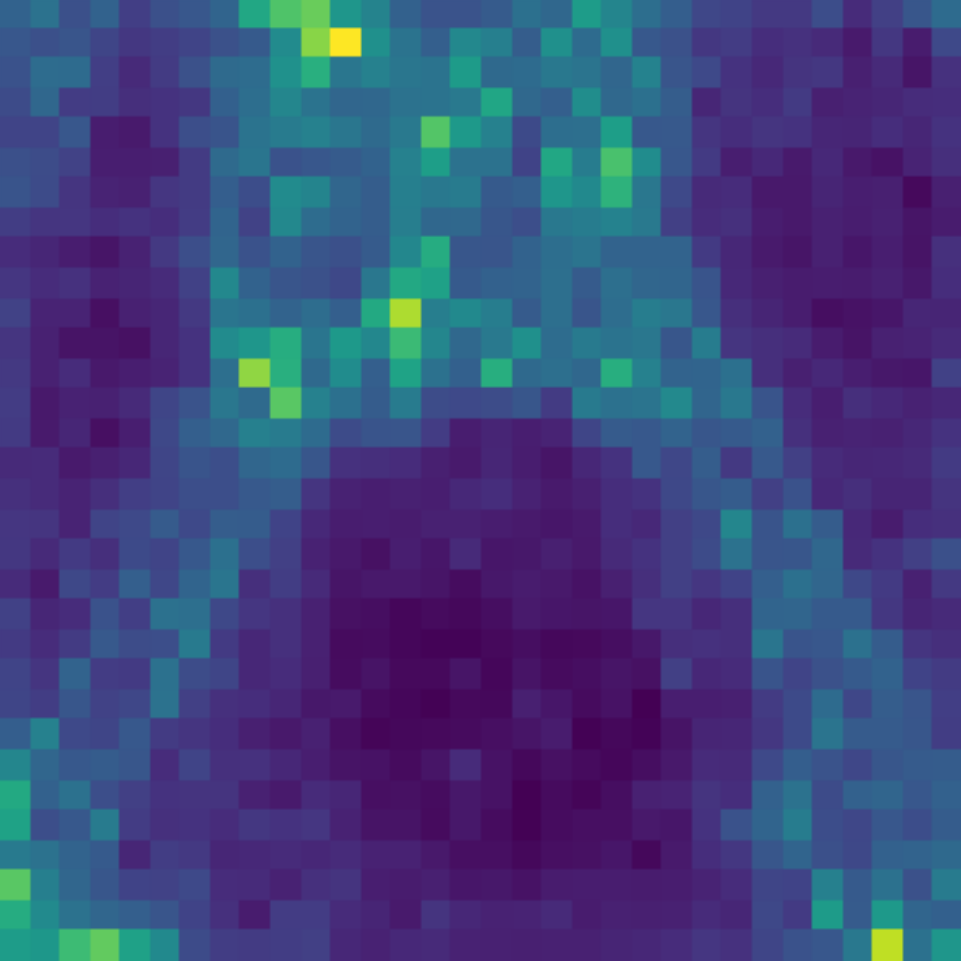}
\end{subfigure}
\begin{subfigure}{.05\textwidth}
  \centering
  \includegraphics[width=1.0\linewidth]{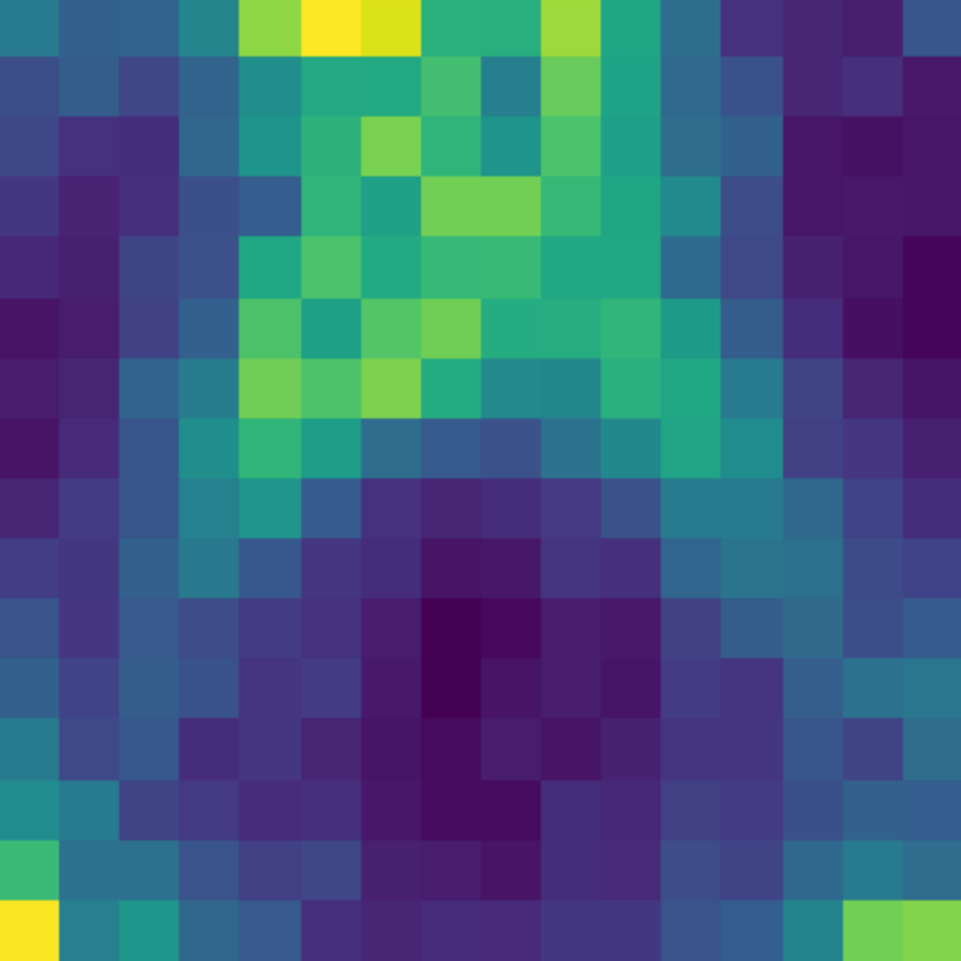}
\end{subfigure}
\begin{subfigure}{.05\textwidth}
  \centering
  \includegraphics[width=1.0\linewidth]{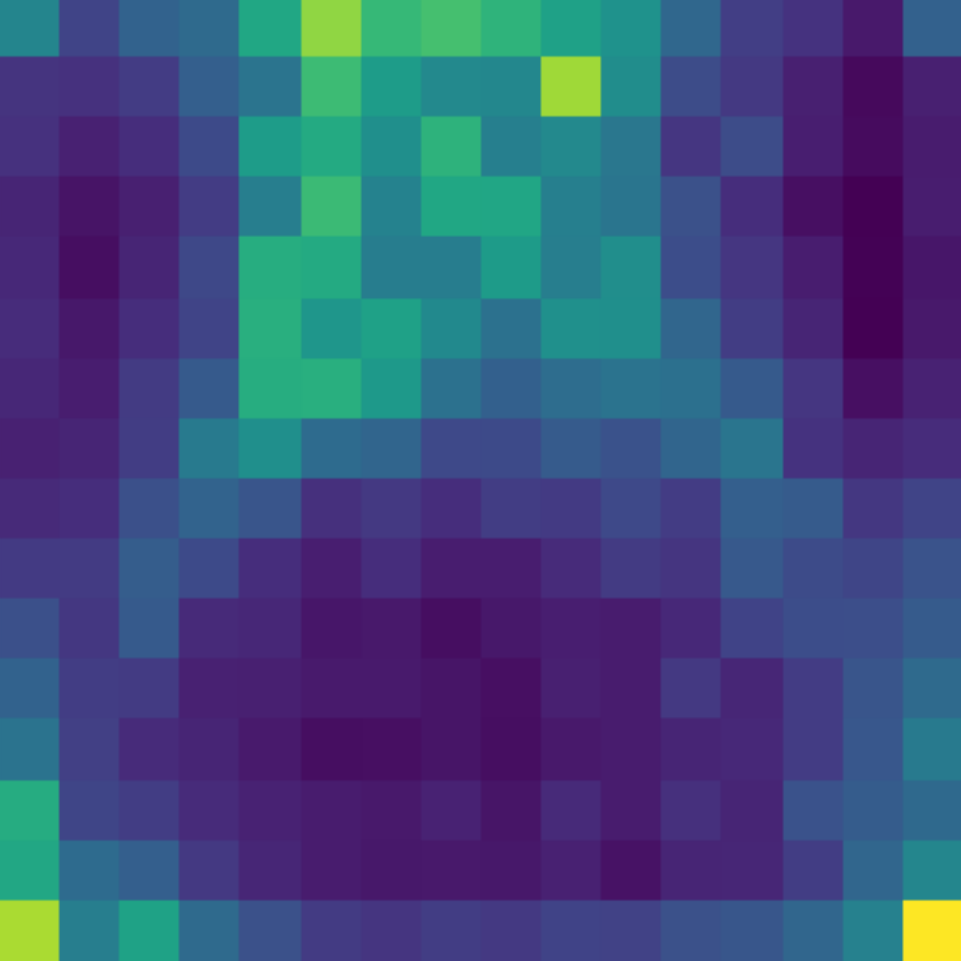}
\end{subfigure}
\begin{subfigure}{.05\textwidth}
  \centering
  \includegraphics[width=1.0\linewidth]{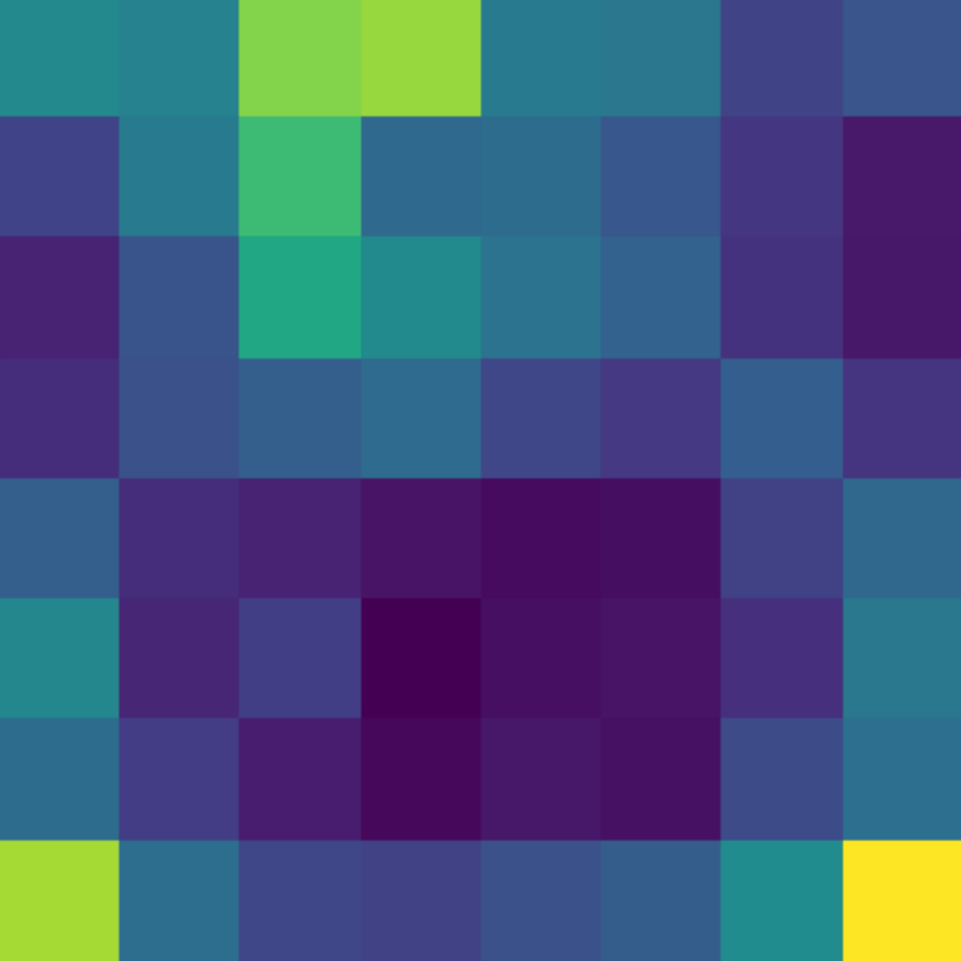}
\end{subfigure}
\begin{subfigure}{.05\textwidth}
  \centering
  \includegraphics[width=1.0\linewidth]{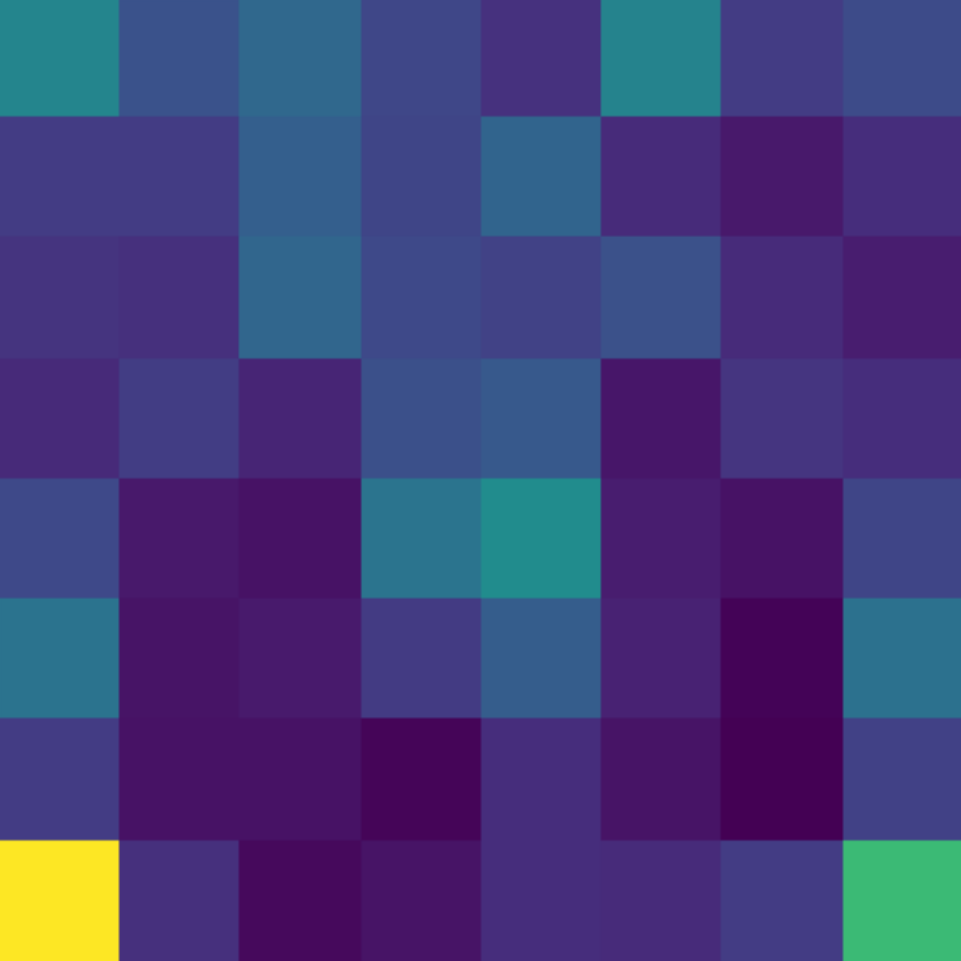}
\end{subfigure}
\begin{subfigure}{.05\textwidth}
  \centering
  \includegraphics[width=1.0\linewidth]{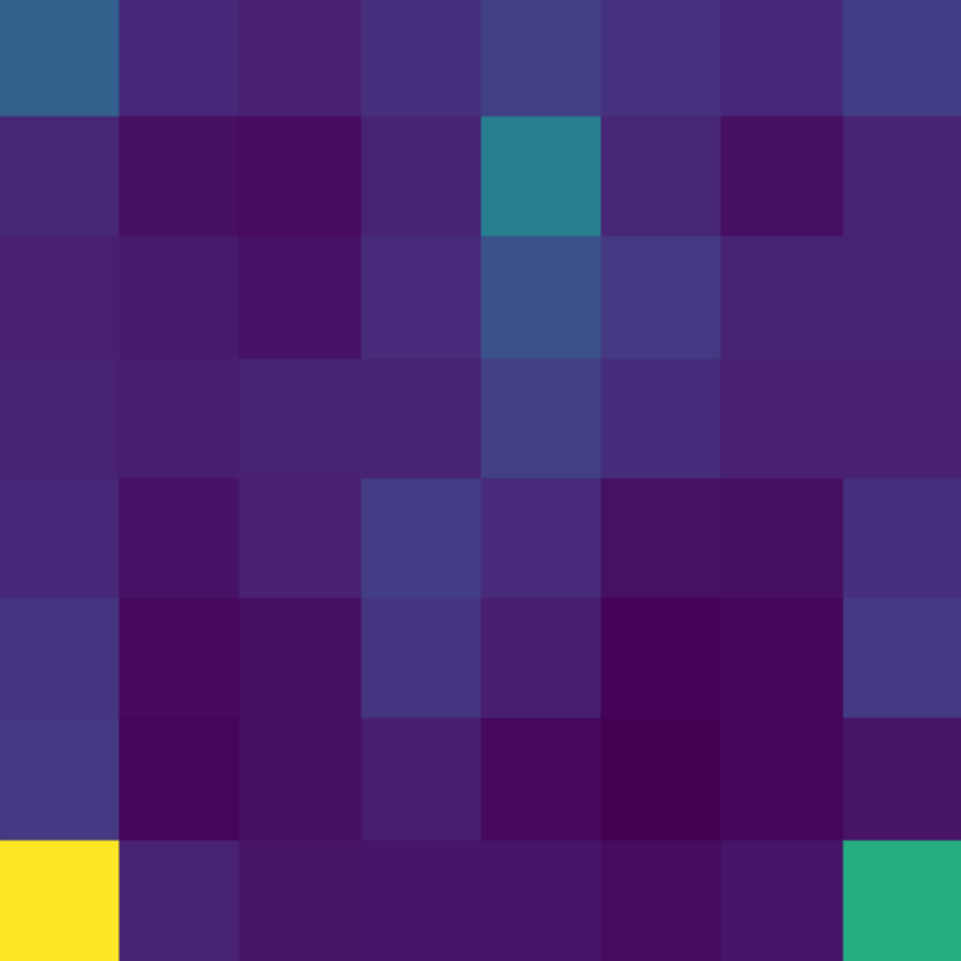}
\end{subfigure}
\begin{subfigure}{.05\textwidth}
  \centering
  \includegraphics[width=1.0\linewidth]{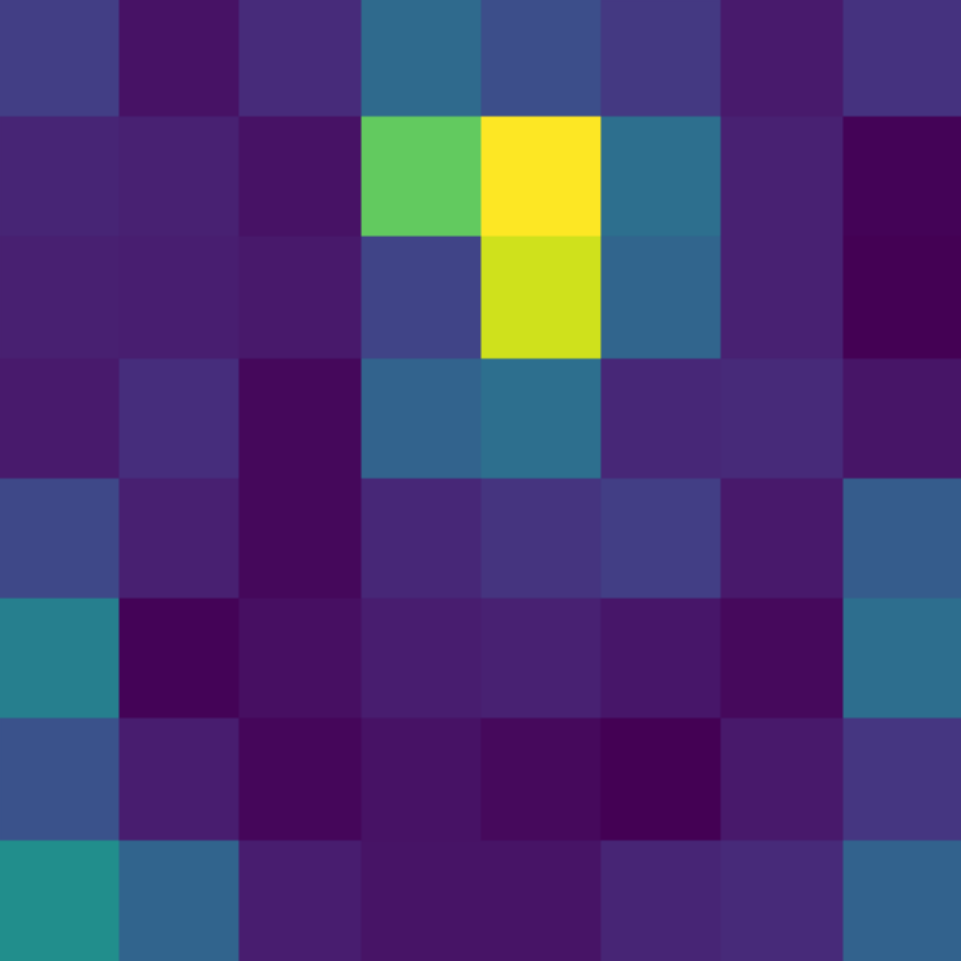}
\end{subfigure}
\begin{subfigure}{.05\textwidth}
  \centering
  \includegraphics[width=1.0\linewidth]{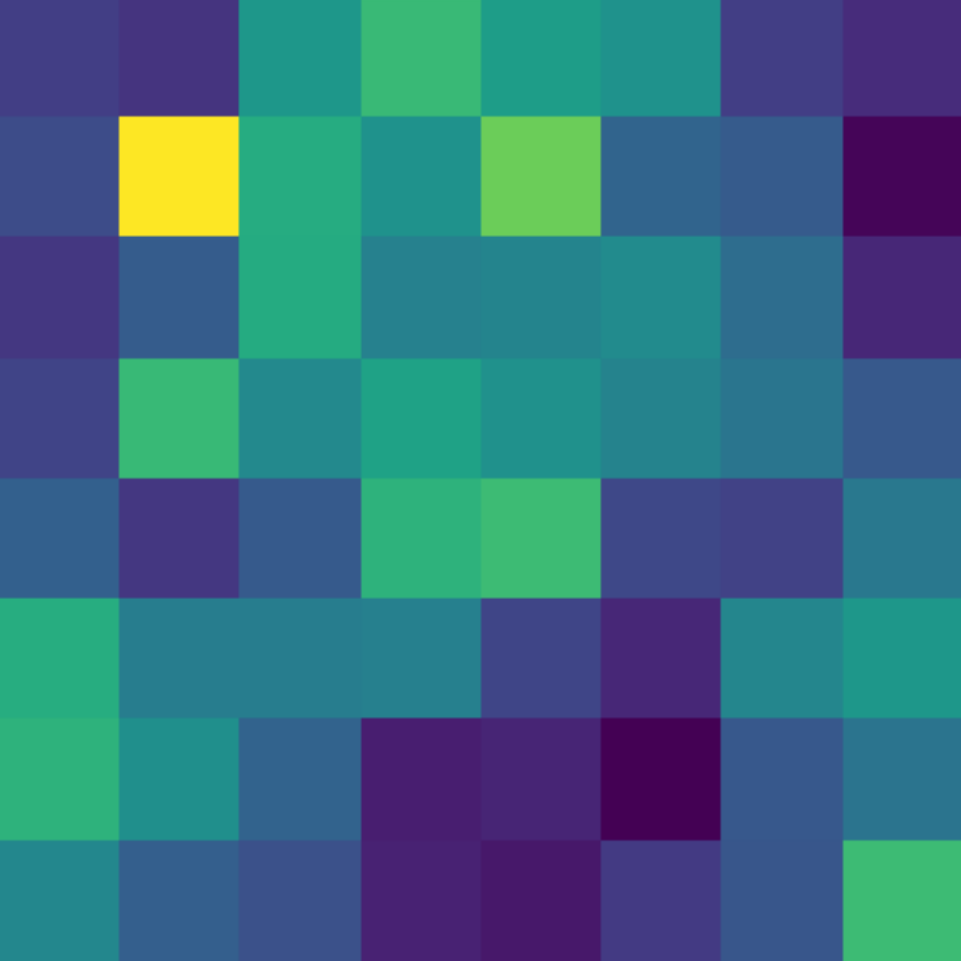}
\end{subfigure}
\begin{subfigure}{.05\textwidth}
  \centering
  \includegraphics[width=1.0\linewidth]{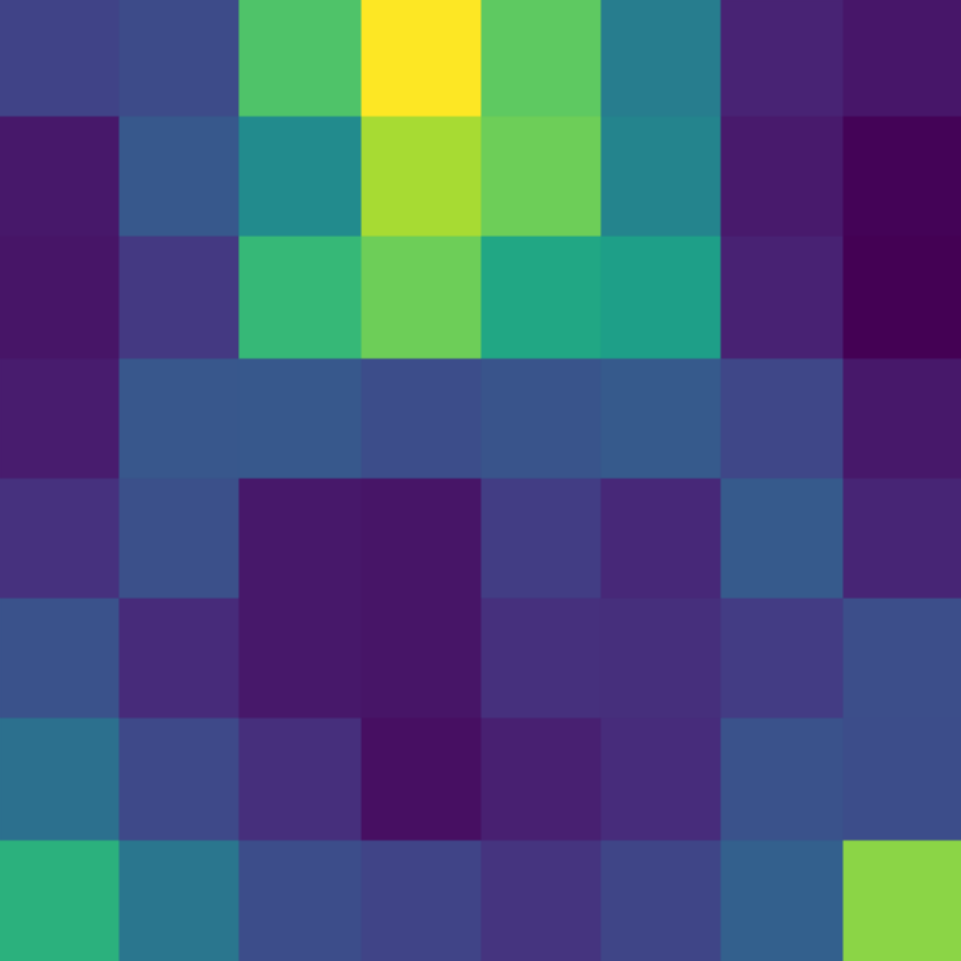}
\end{subfigure}
\begin{subfigure}{.05\textwidth}
  \centering
  \includegraphics[width=1.0\linewidth]{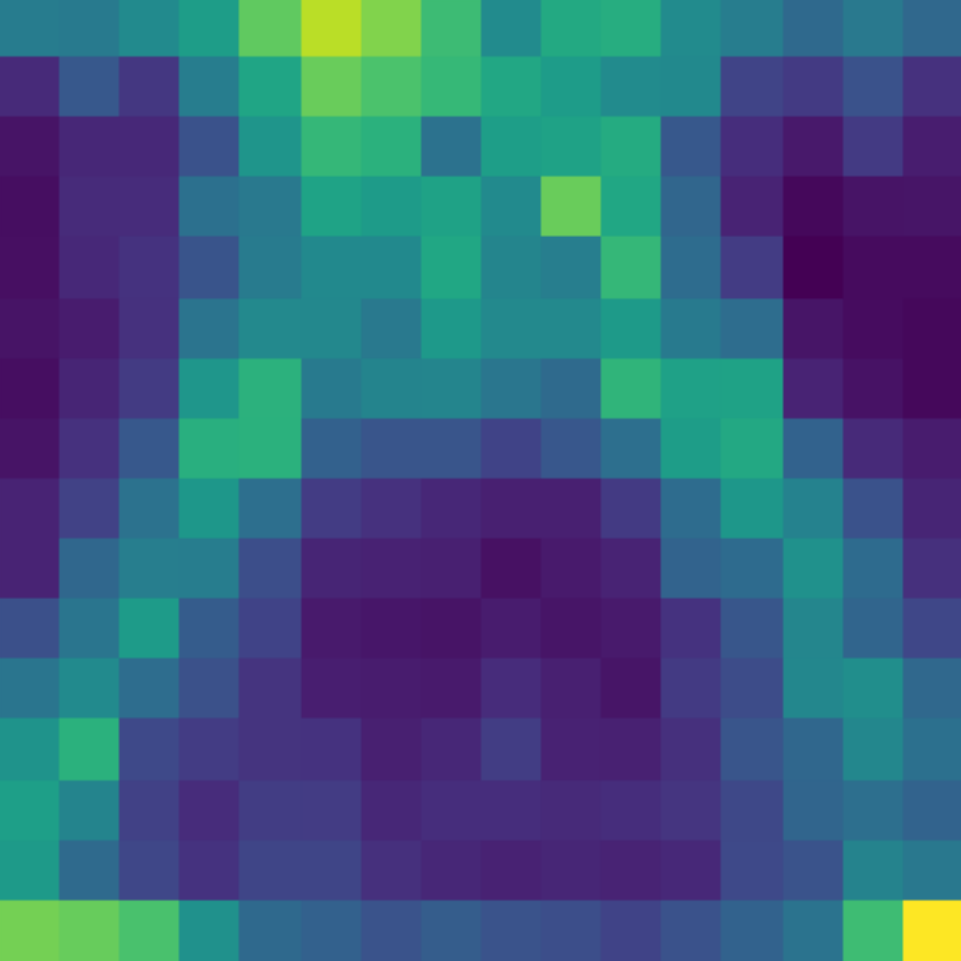}
\end{subfigure}
\begin{subfigure}{.05\textwidth}
  \centering
  \includegraphics[width=1.0\linewidth]{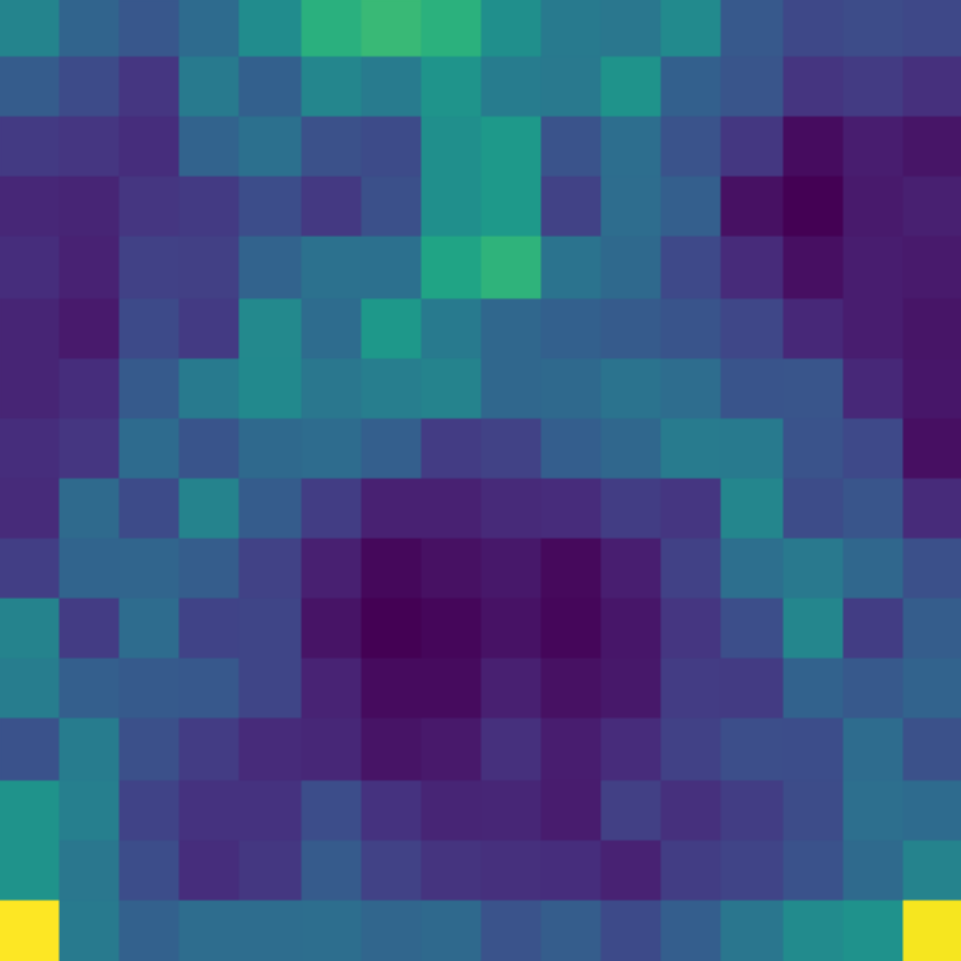}
\end{subfigure}
\begin{subfigure}{.05\textwidth}
  \centering
  \includegraphics[width=1.0\linewidth]{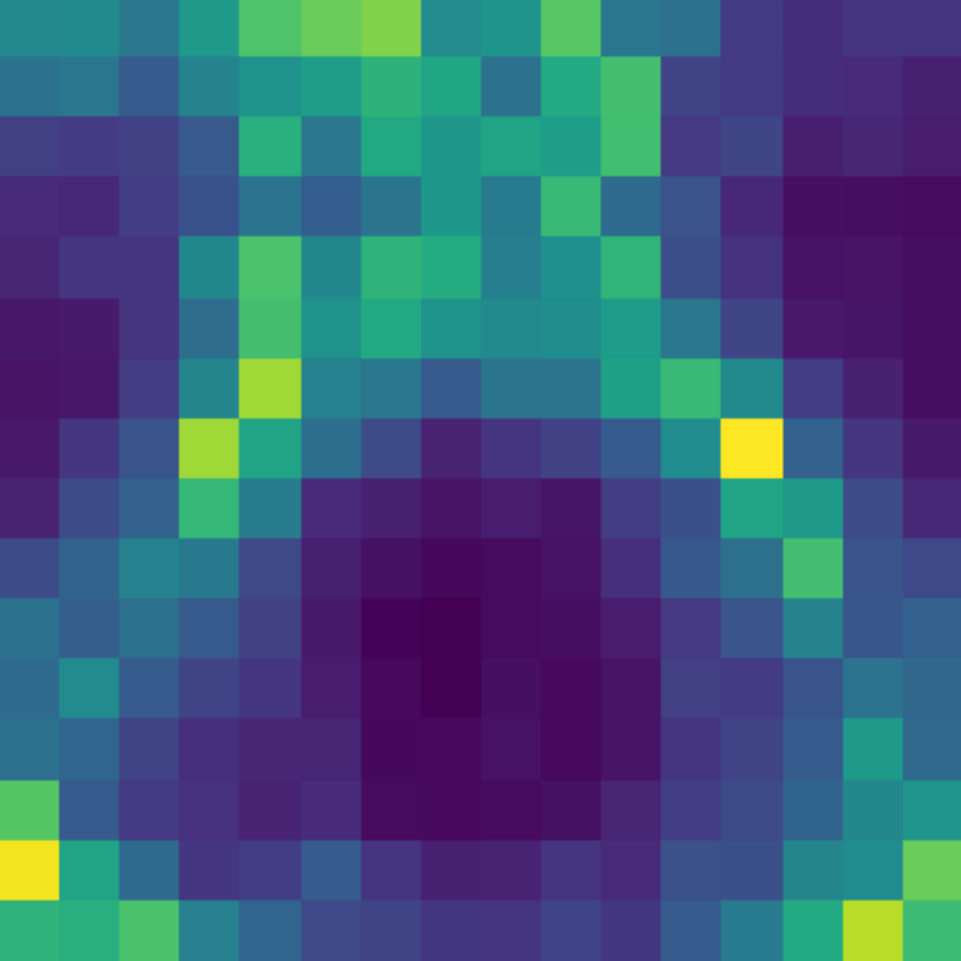}
\end{subfigure}
\begin{subfigure}{.05\textwidth}
  \centering
  \includegraphics[width=1.0\linewidth]{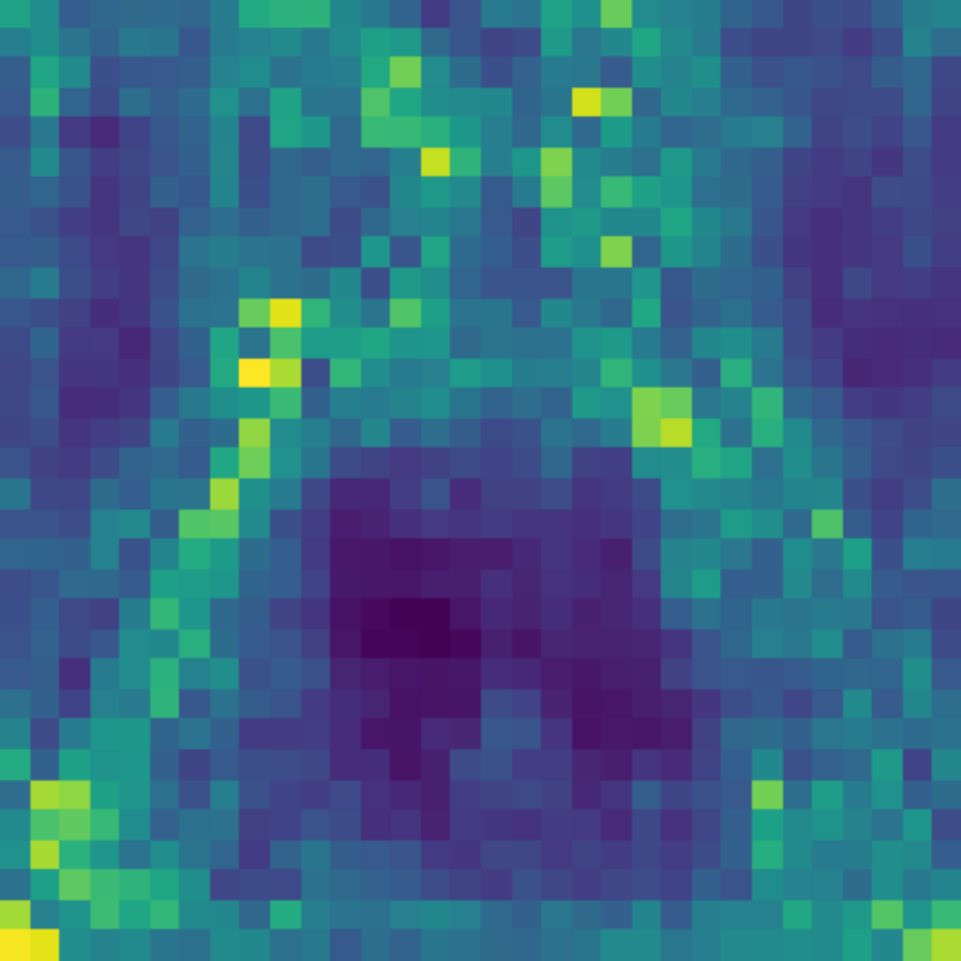}
\end{subfigure}
\begin{subfigure}{.05\textwidth}
  \centering
  \includegraphics[width=1.0\linewidth]{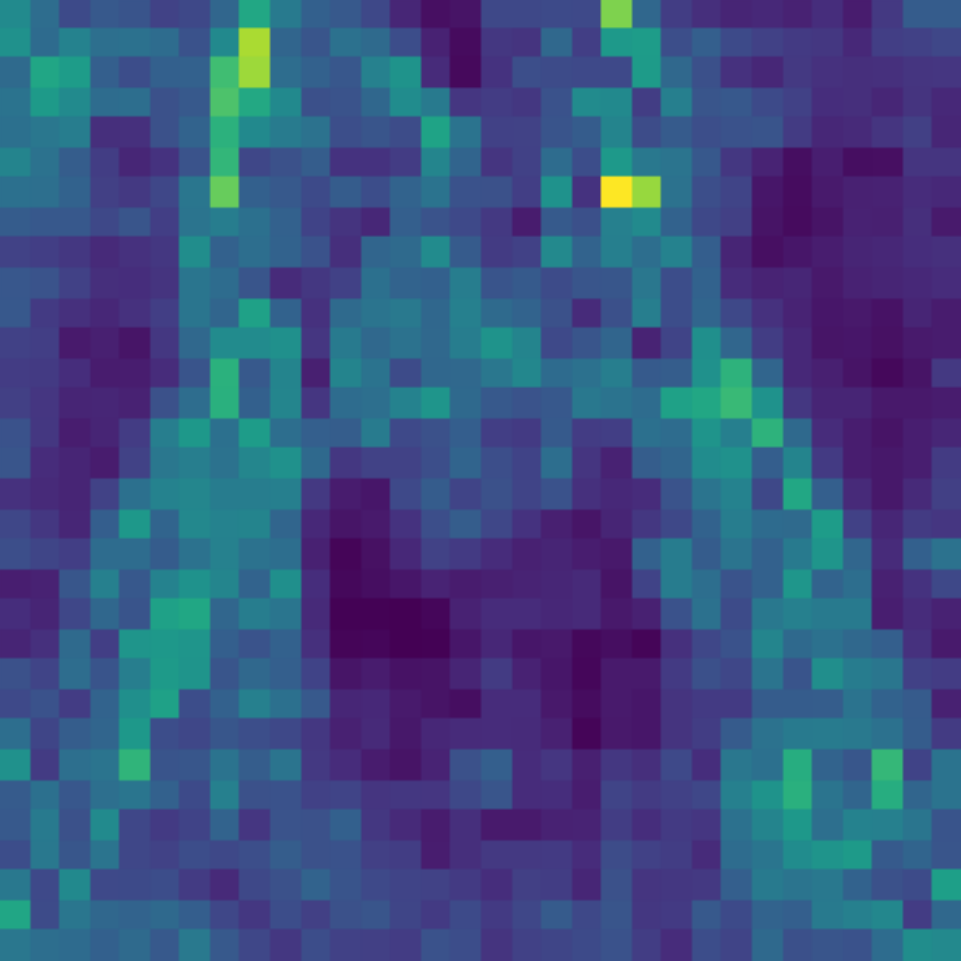}
\end{subfigure}
\begin{subfigure}{.05\textwidth}
  \centering
  \includegraphics[width=1.0\linewidth]{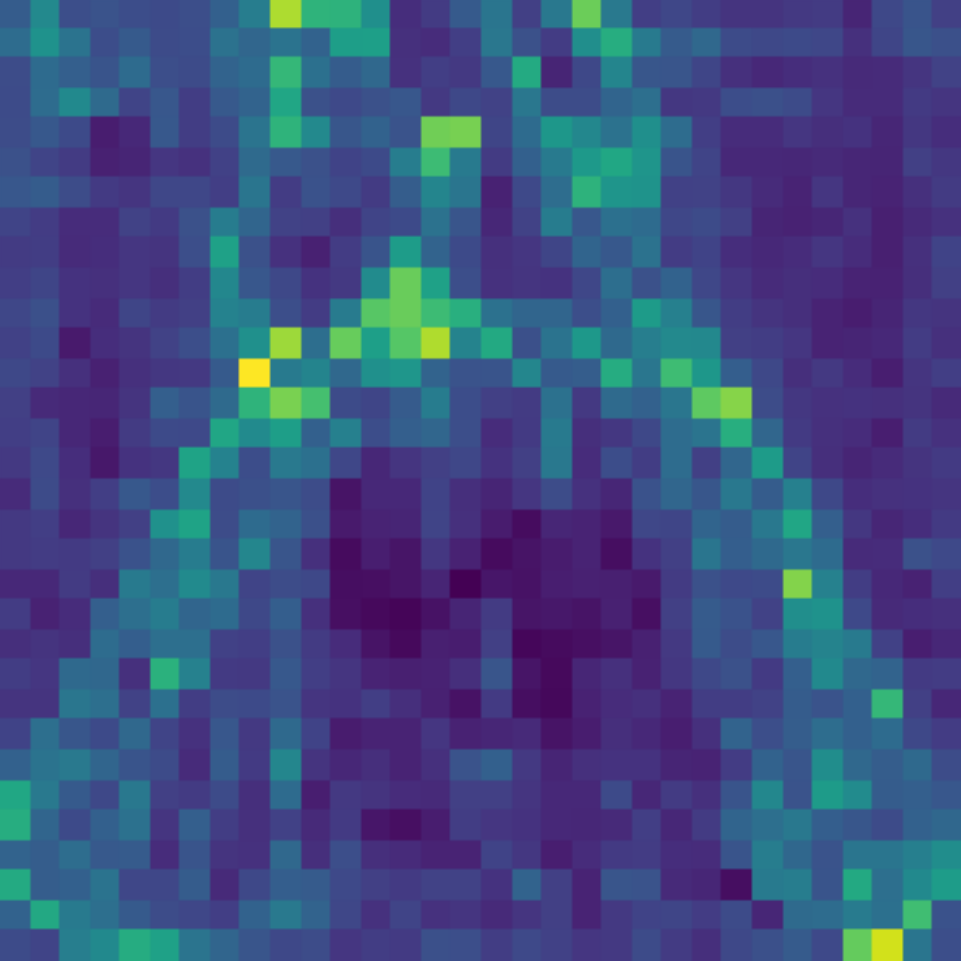}
\end{subfigure}
\\
&
\begin{subfigure}{.05\textwidth}
  \centering
  \includegraphics[width=1.0\linewidth]{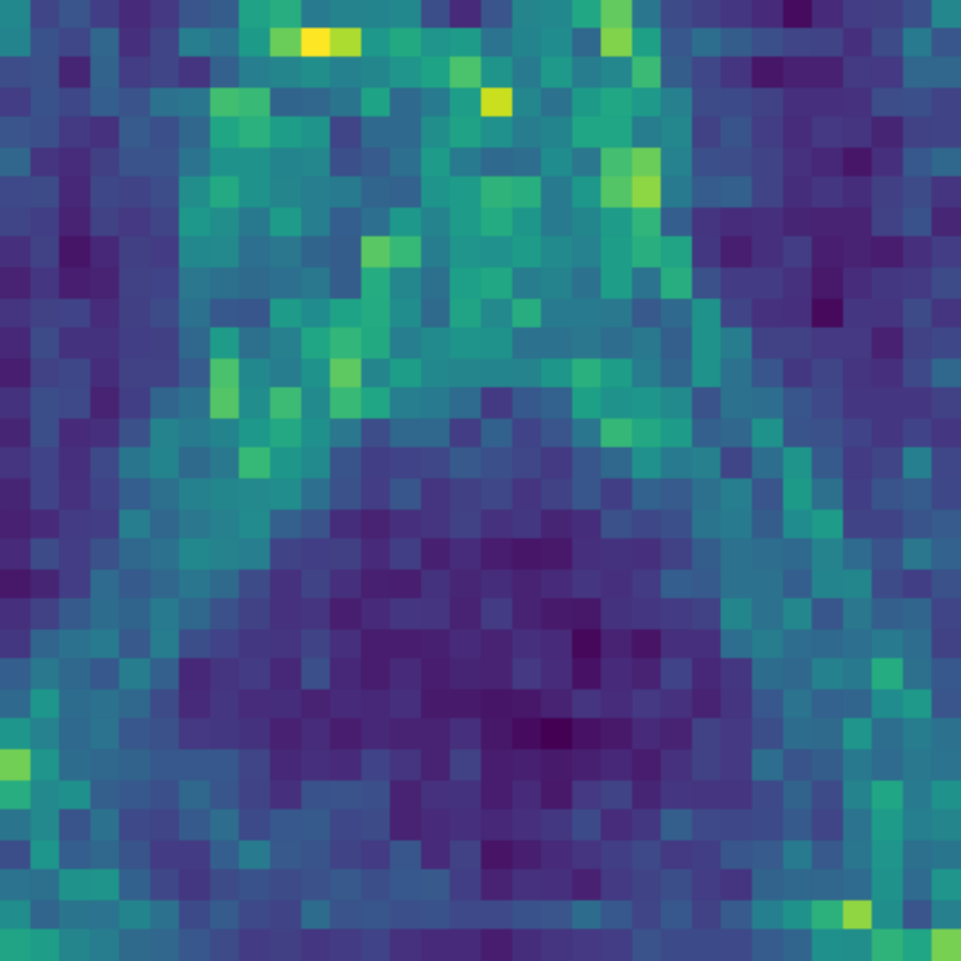}
\end{subfigure}
\begin{subfigure}{.05\textwidth}
  \centering
  \includegraphics[width=1.0\linewidth]{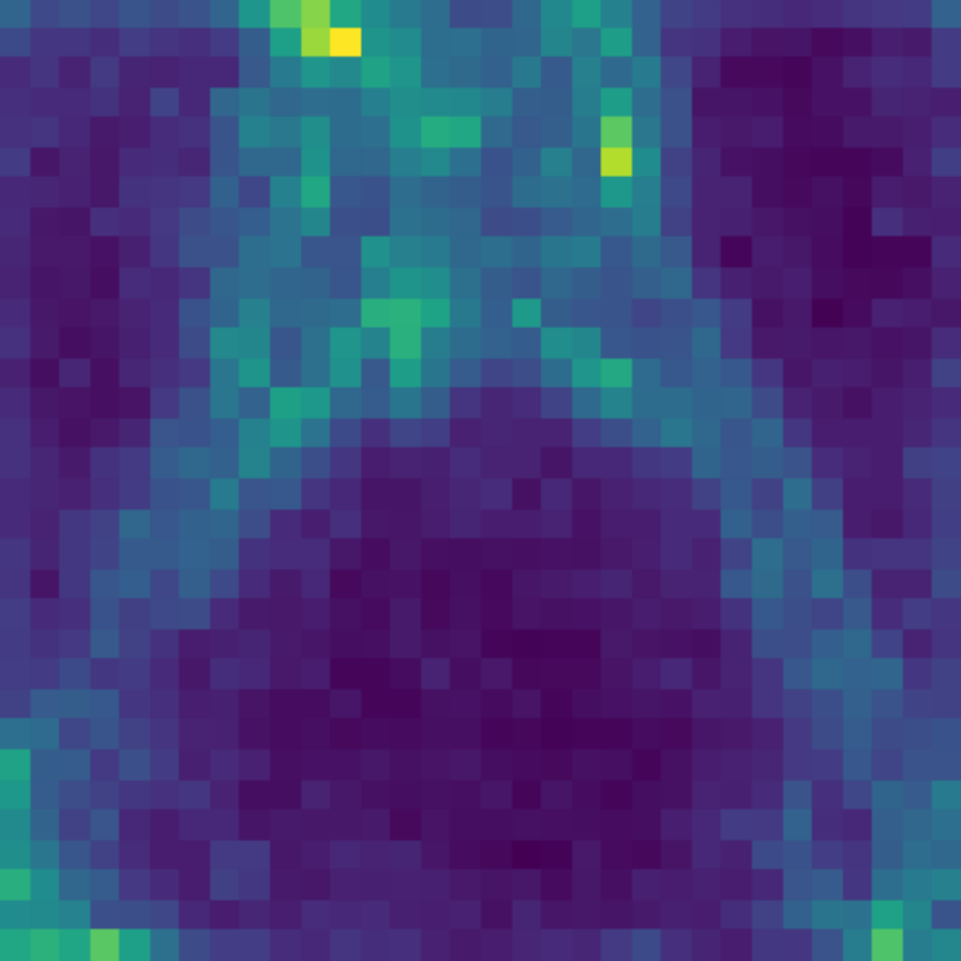}
\end{subfigure}
\begin{subfigure}{.05\textwidth}
  \centering
  \includegraphics[width=1.0\linewidth]{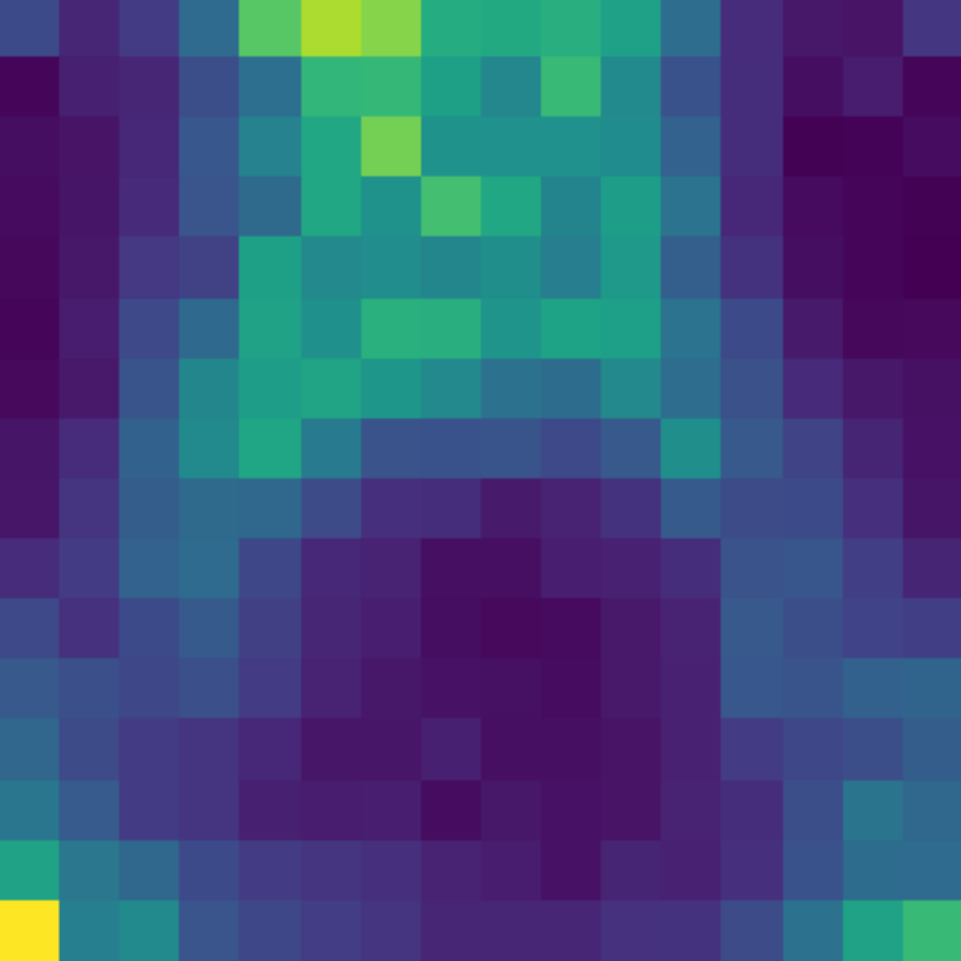}
\end{subfigure}
\begin{subfigure}{.05\textwidth}
  \centering
  \includegraphics[width=1.0\linewidth]{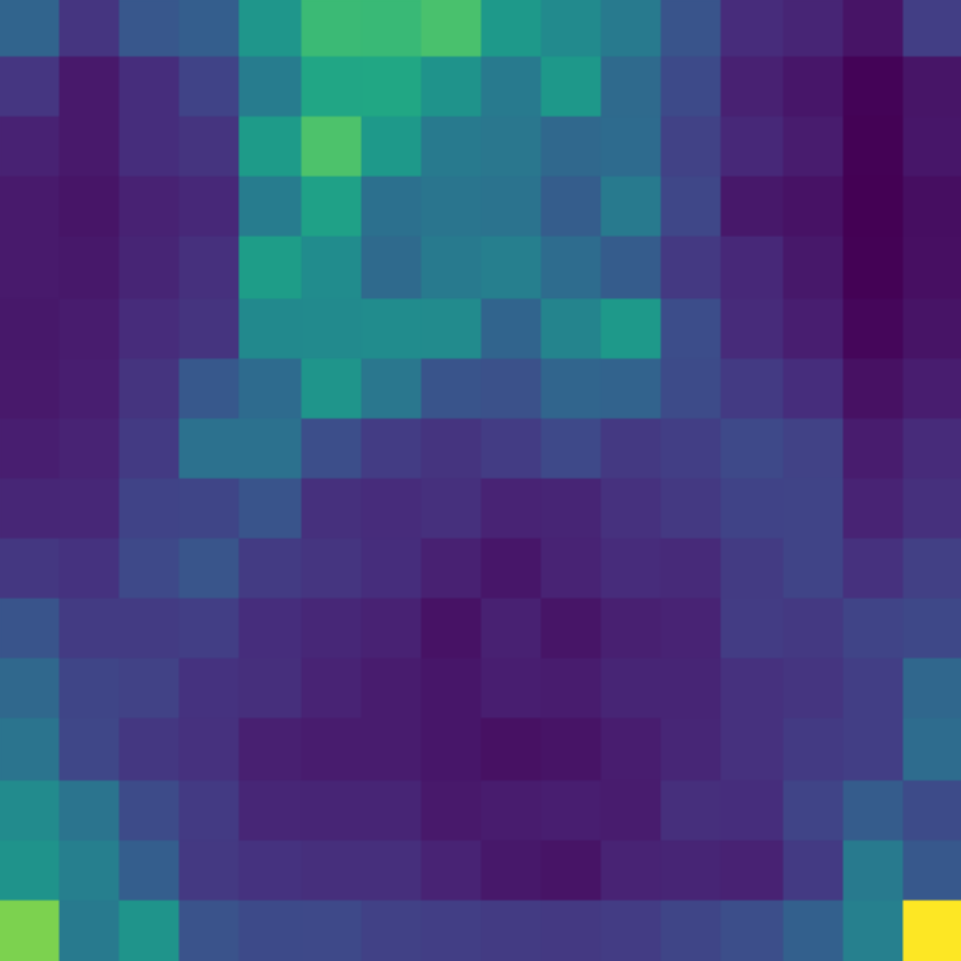}
\end{subfigure}
\begin{subfigure}{.05\textwidth}
  \centering
  \includegraphics[width=1.0\linewidth]{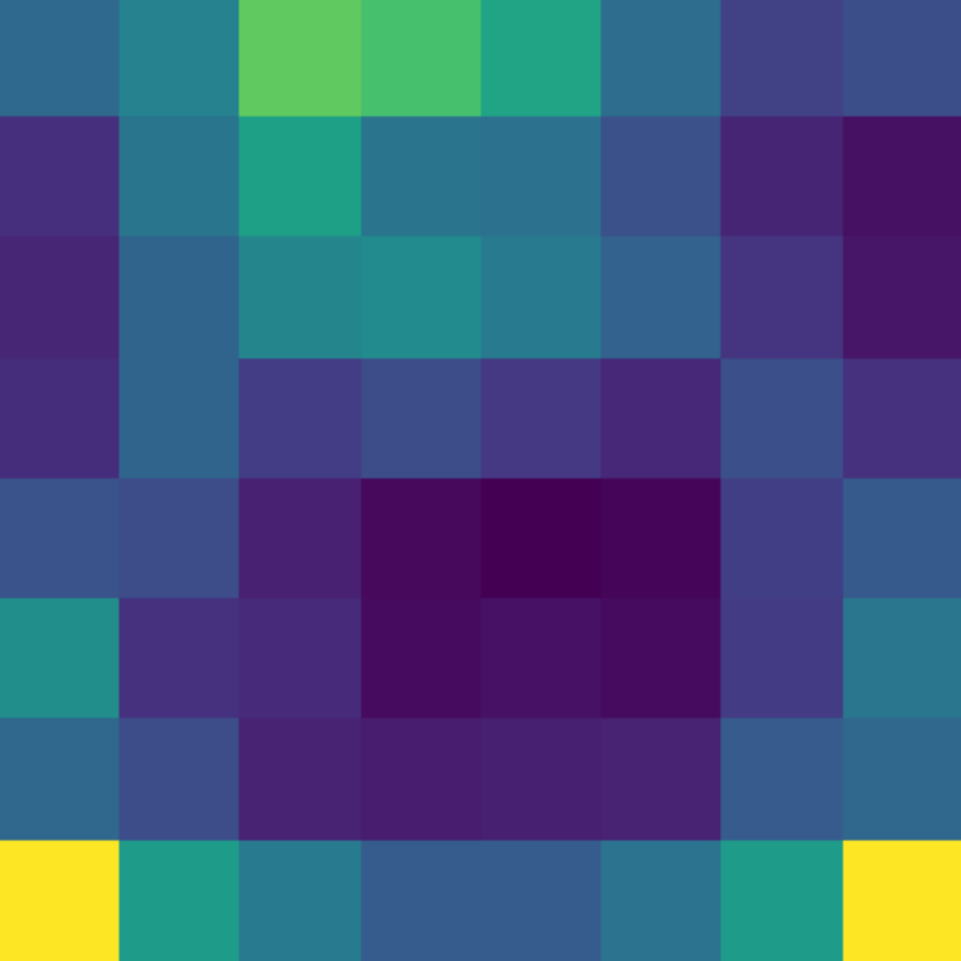}
\end{subfigure}
\begin{subfigure}{.05\textwidth}
  \centering
  \includegraphics[width=1.0\linewidth]{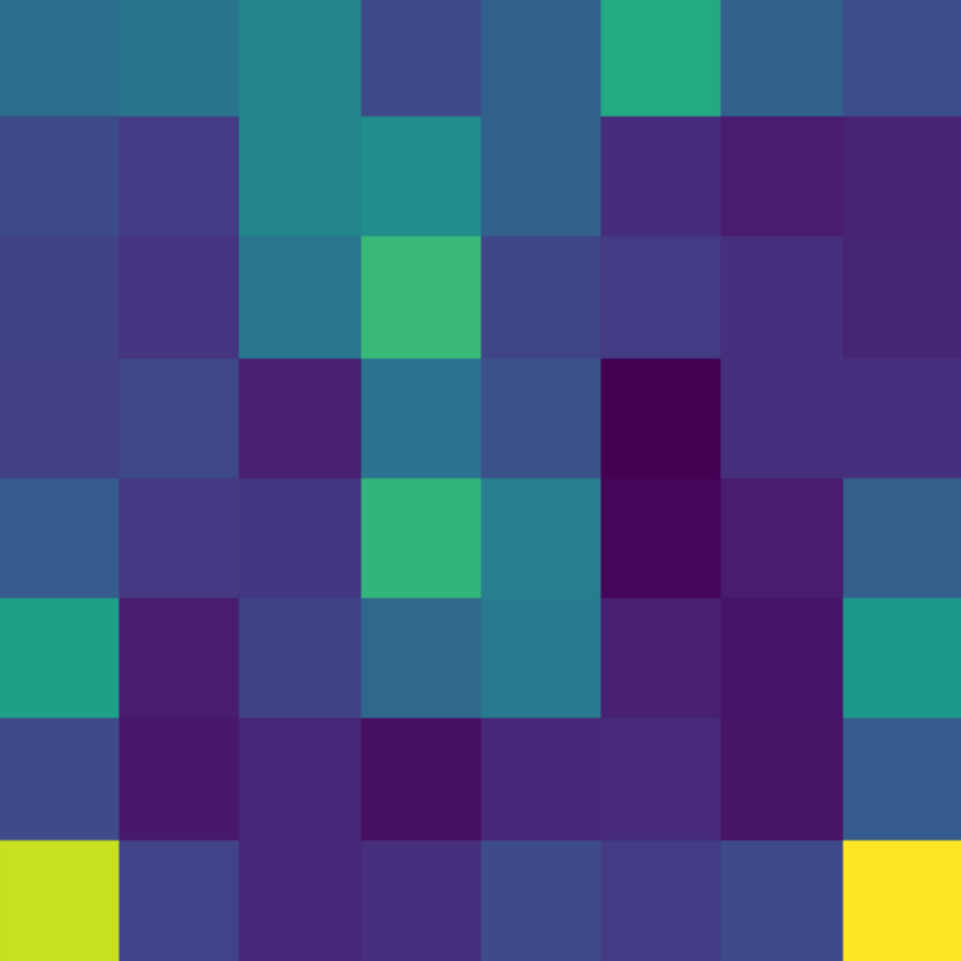}
\end{subfigure}
\begin{subfigure}{.05\textwidth}
  \centering
  \includegraphics[width=1.0\linewidth]{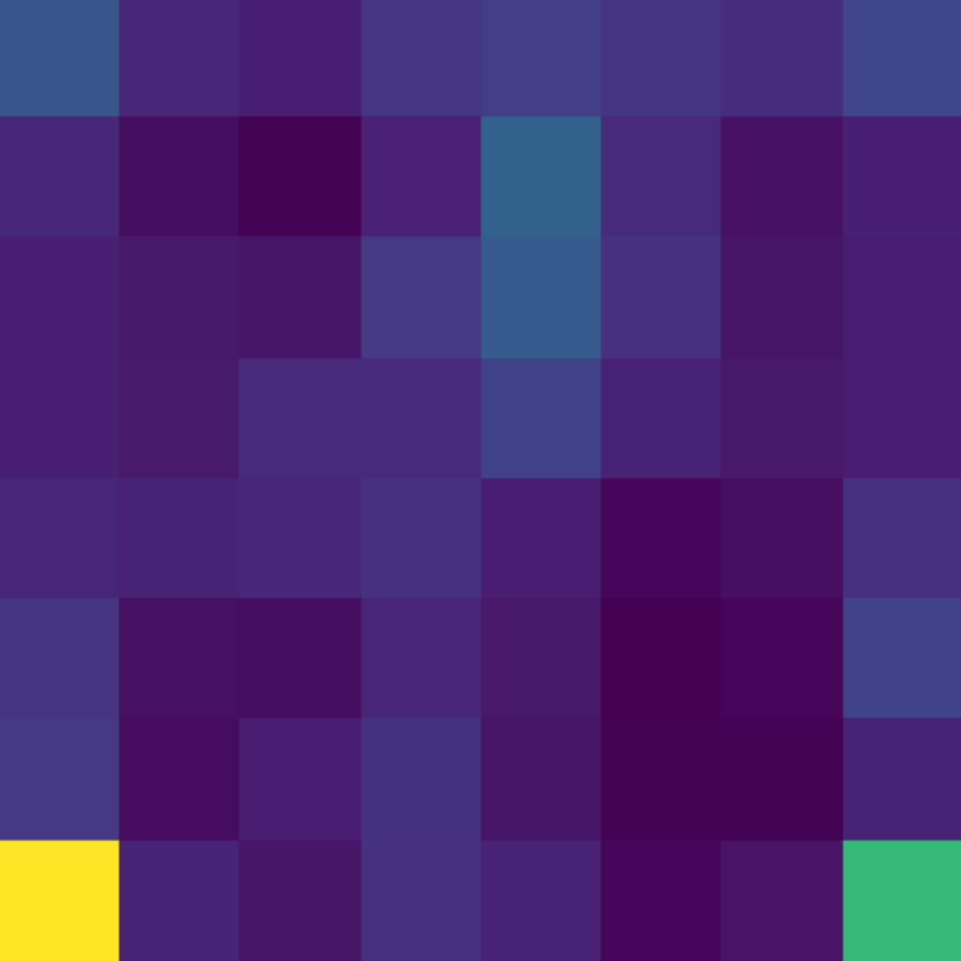}
\end{subfigure}
\begin{subfigure}{.05\textwidth}
  \centering
  \includegraphics[width=1.0\linewidth]{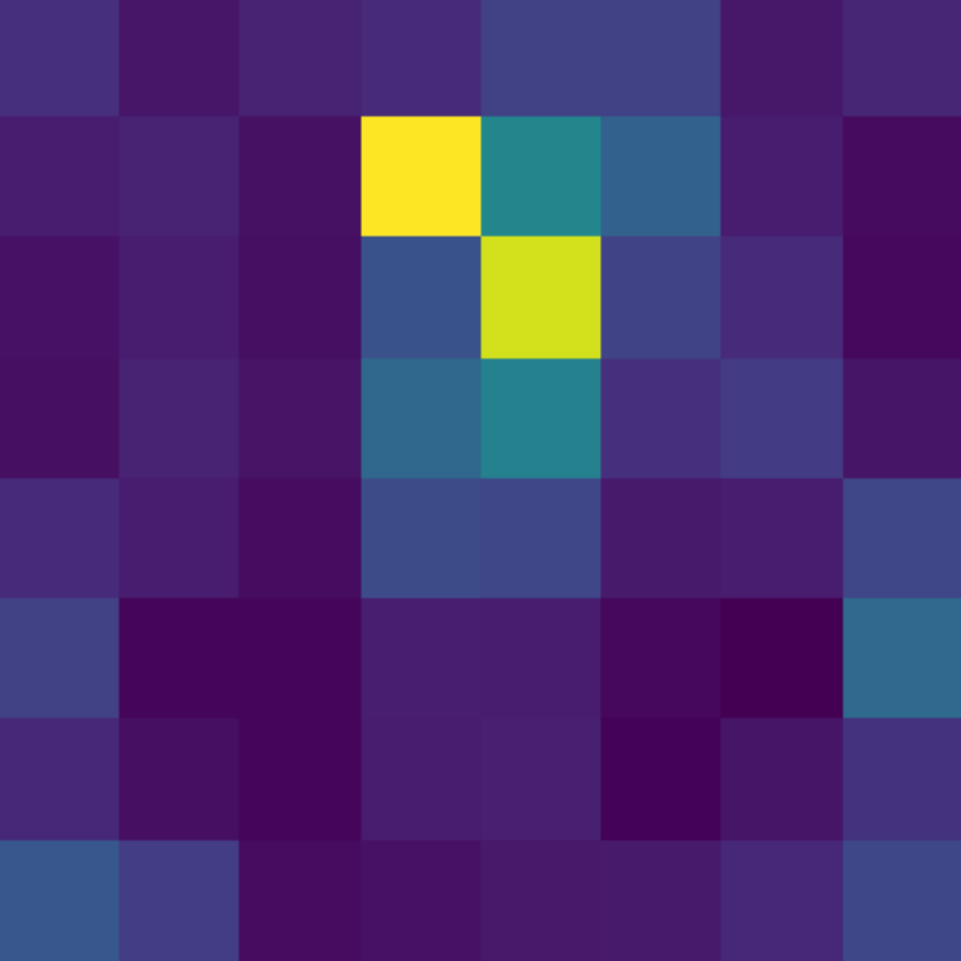}
\end{subfigure}
\begin{subfigure}{.05\textwidth}
  \centering
  \includegraphics[width=1.0\linewidth]{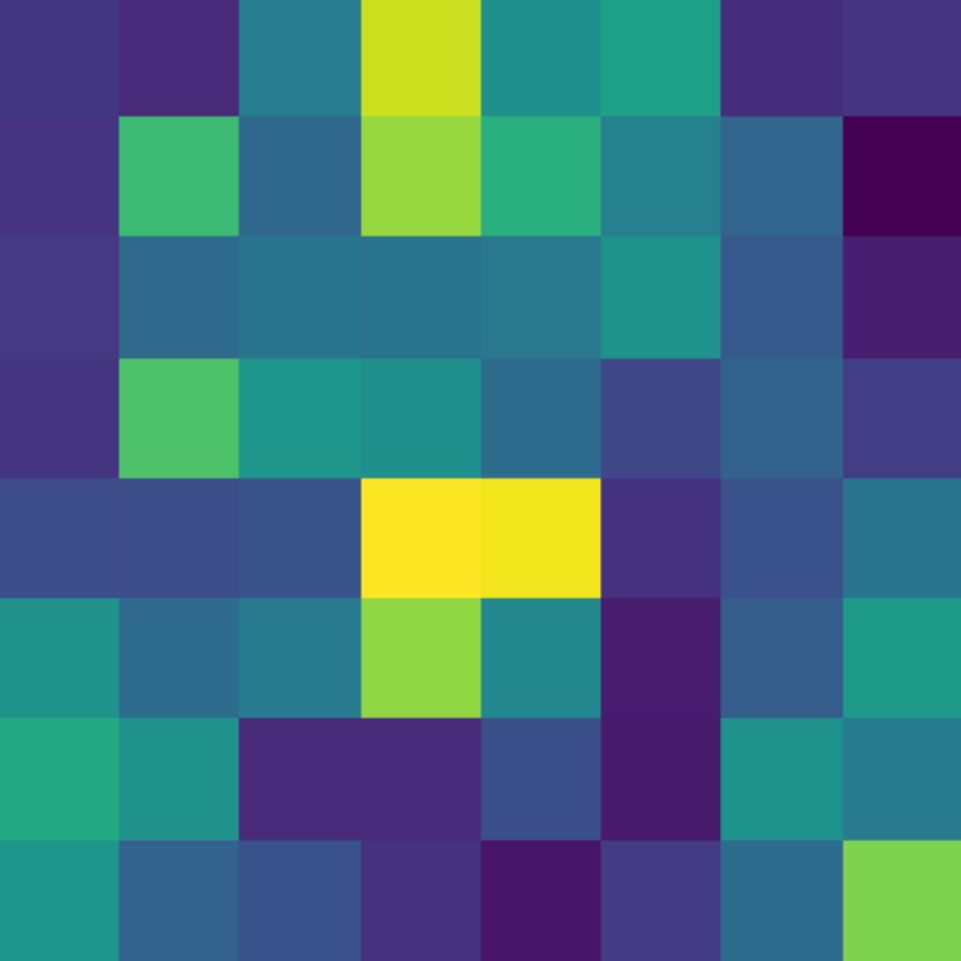}
\end{subfigure}
\begin{subfigure}{.05\textwidth}
  \centering
  \includegraphics[width=1.0\linewidth]{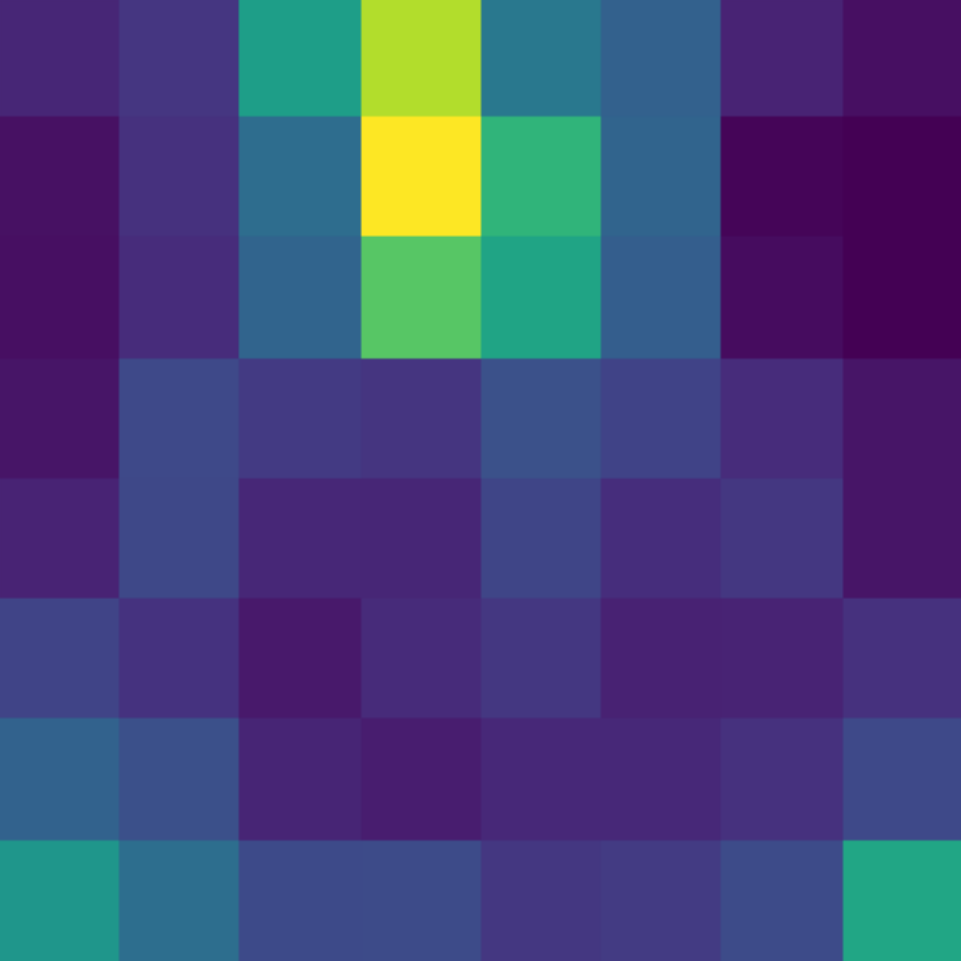}
\end{subfigure}
\begin{subfigure}{.05\textwidth}
  \centering
  \includegraphics[width=1.0\linewidth]{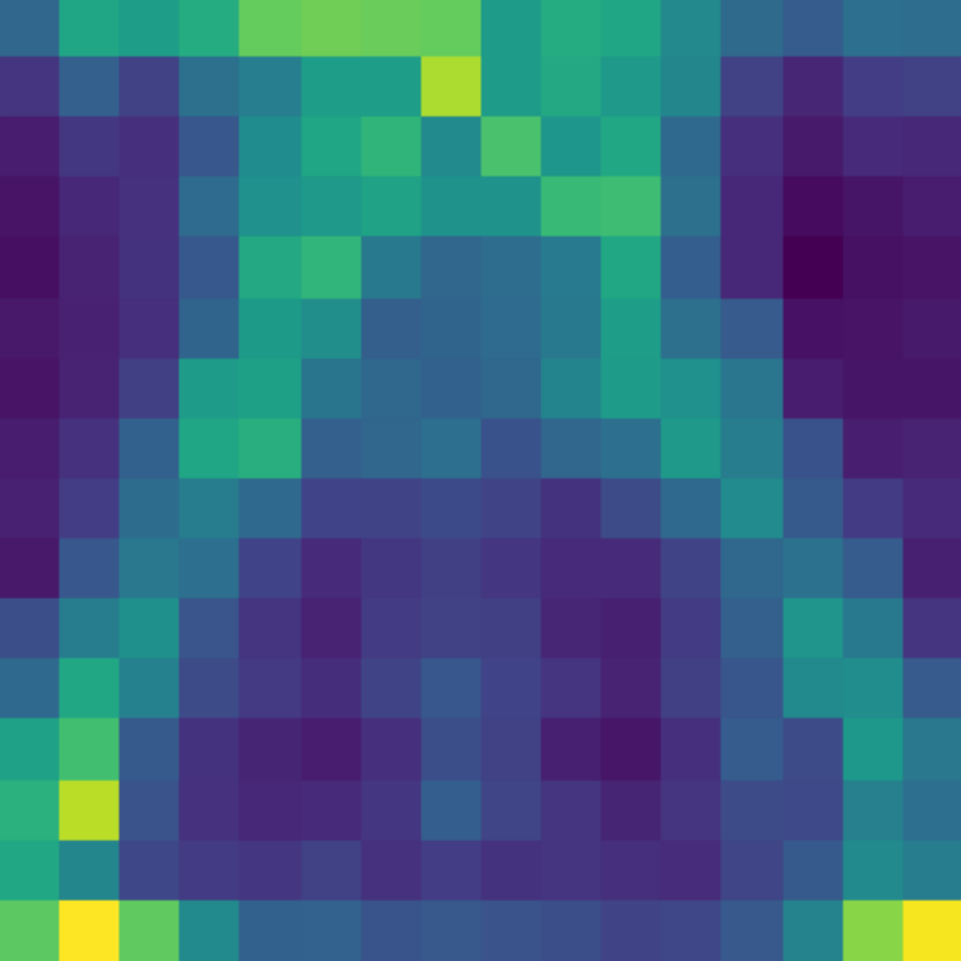}
\end{subfigure}
\begin{subfigure}{.05\textwidth}
  \centering
  \includegraphics[width=1.0\linewidth]{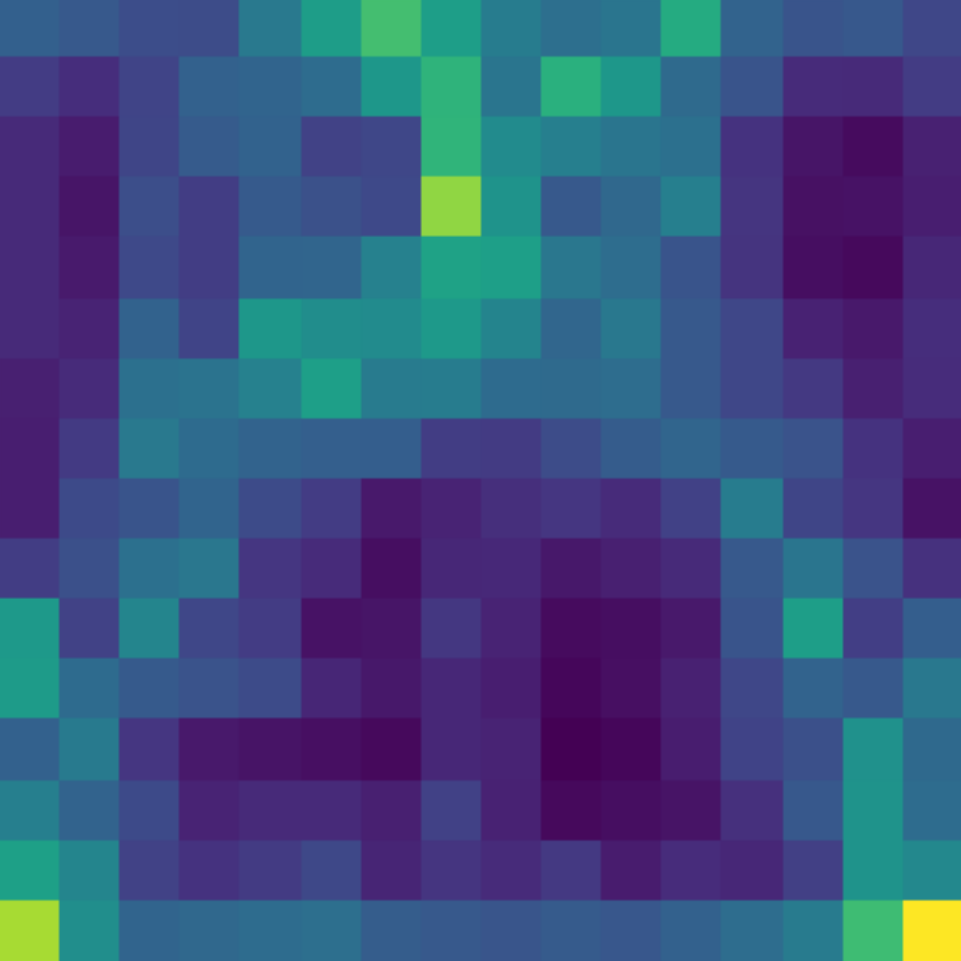}
\end{subfigure}
\begin{subfigure}{.05\textwidth}
  \centering
  \includegraphics[width=1.0\linewidth]{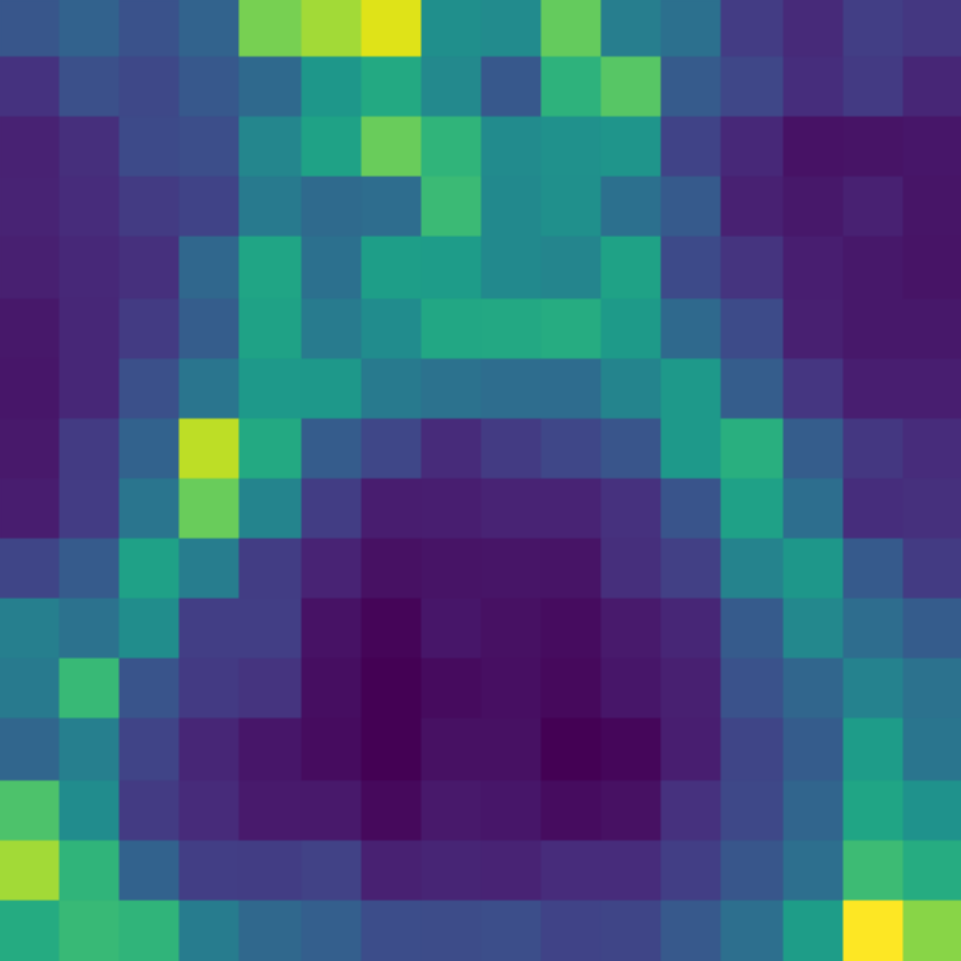}
\end{subfigure}
\begin{subfigure}{.05\textwidth}
  \centering
  \includegraphics[width=1.0\linewidth]{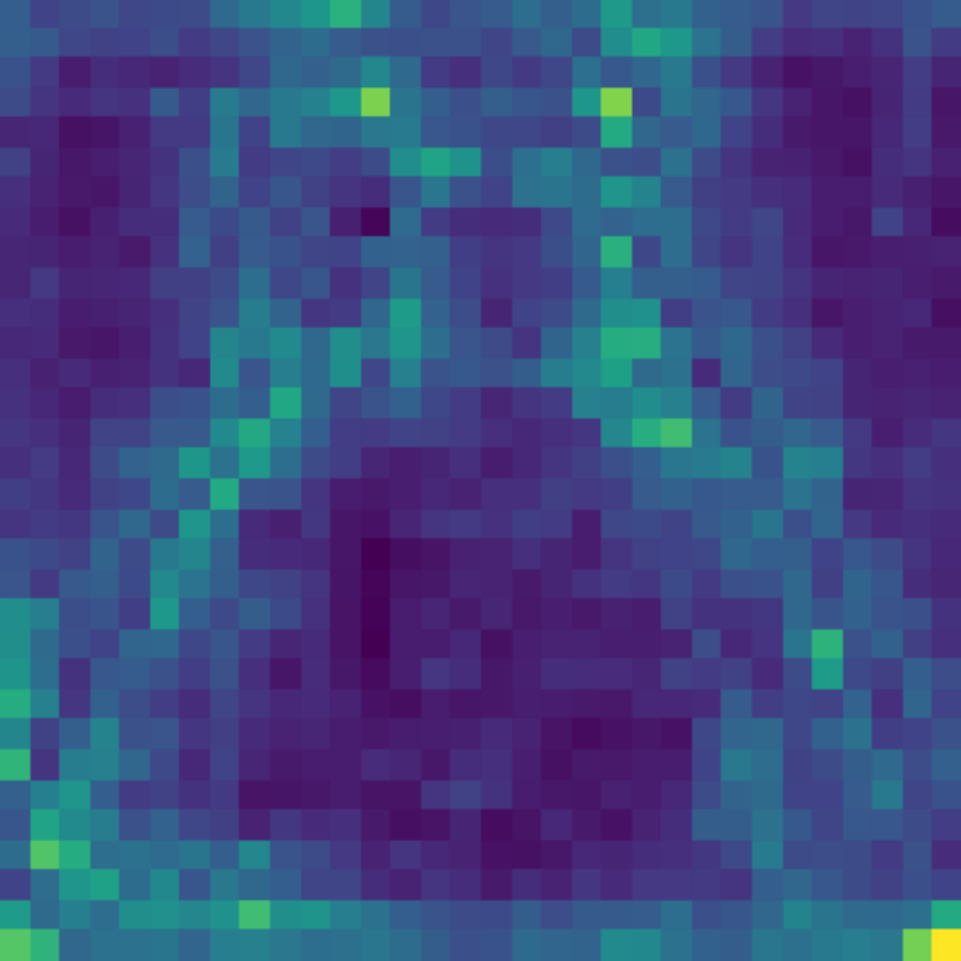}
\end{subfigure}
\begin{subfigure}{.05\textwidth}
  \centering
  \includegraphics[width=1.0\linewidth]{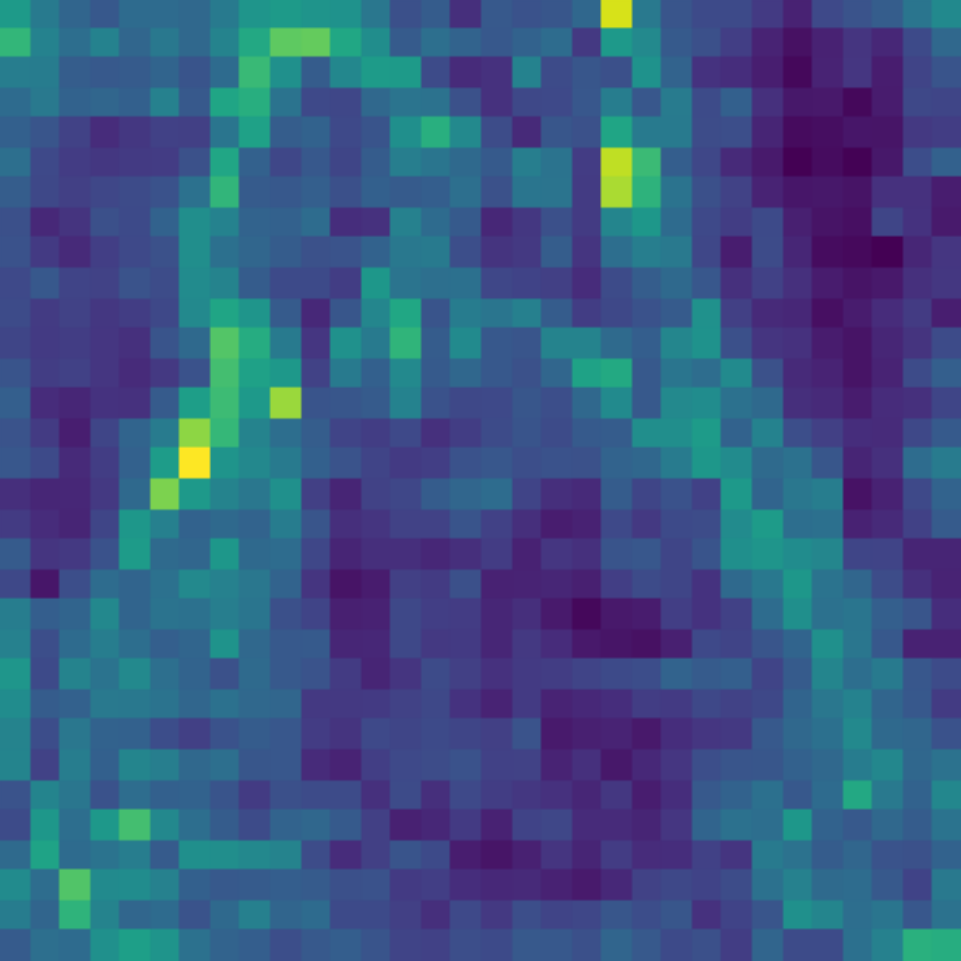}
\end{subfigure}
\begin{subfigure}{.05\textwidth}
  \centering
  \includegraphics[width=1.0\linewidth]{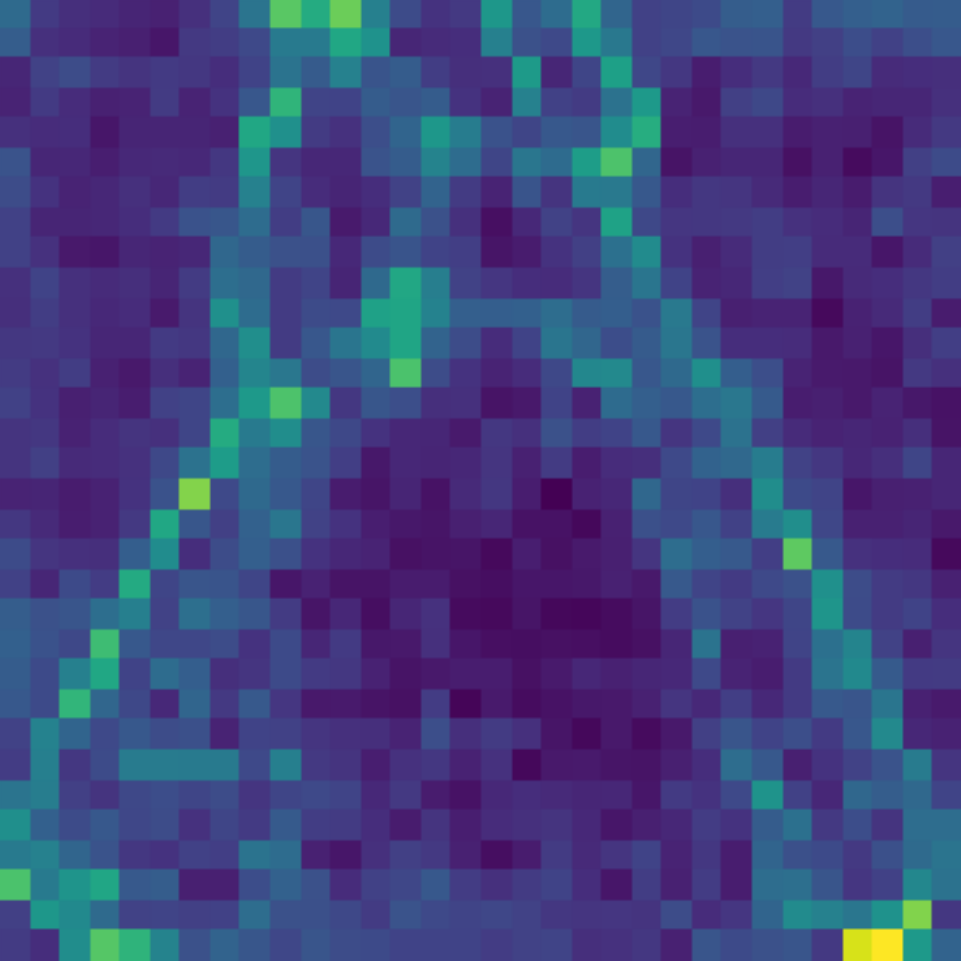}
\end{subfigure}
\\
&
\begin{subfigure}{.05\textwidth}
  \centering
  \includegraphics[width=1.0\linewidth]{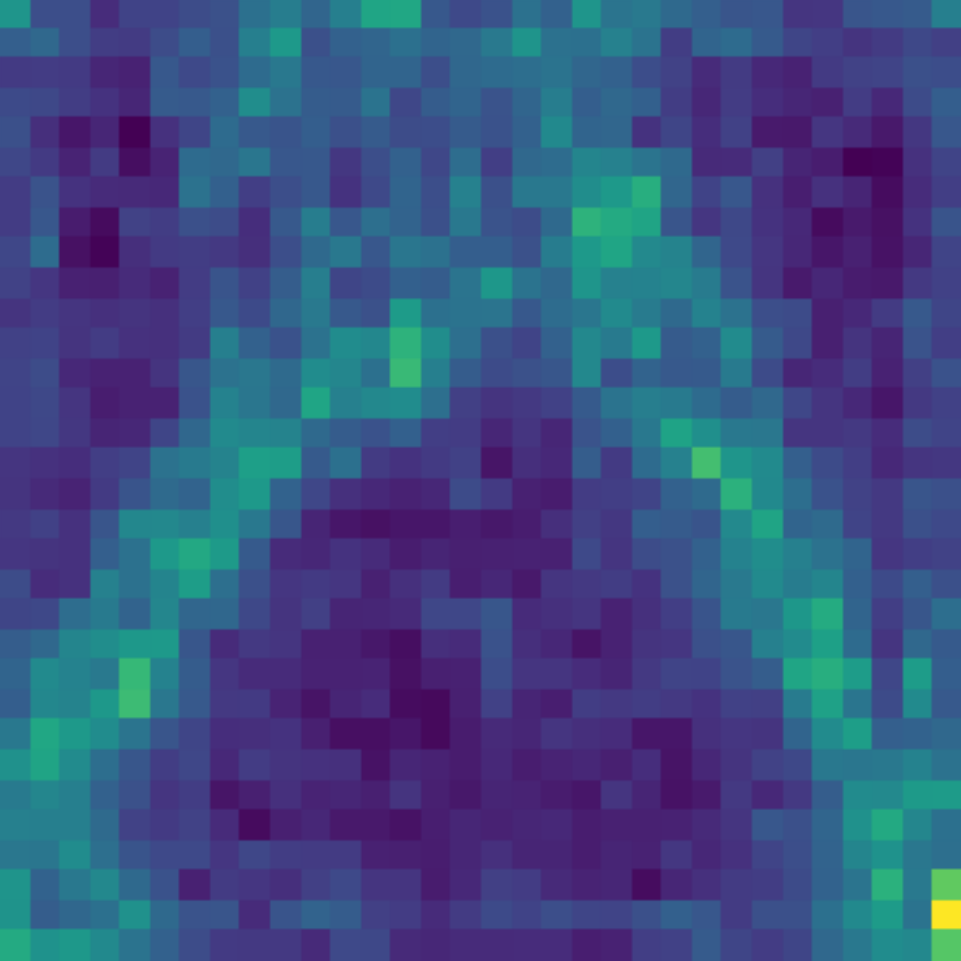}
\end{subfigure}
\begin{subfigure}{.05\textwidth}
  \centering
  \includegraphics[width=1.0\linewidth]{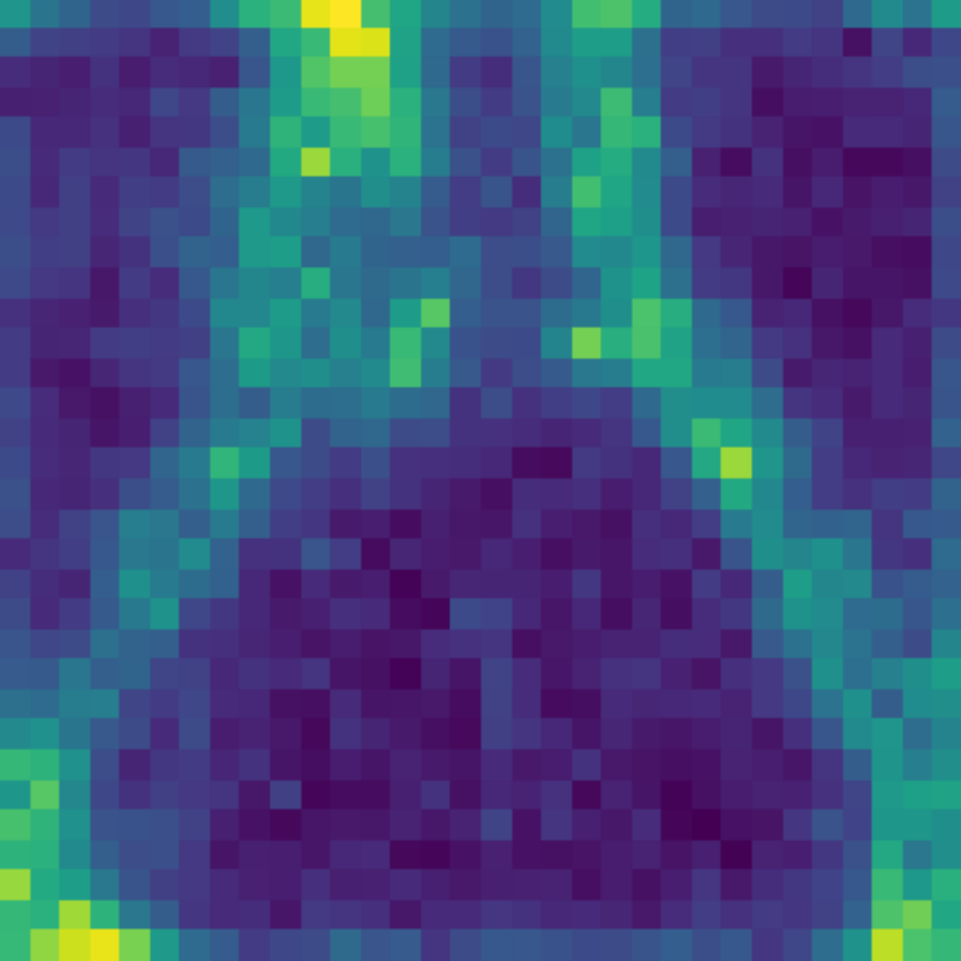}
\end{subfigure}
\begin{subfigure}{.05\textwidth}
  \centering
  \includegraphics[width=1.0\linewidth]{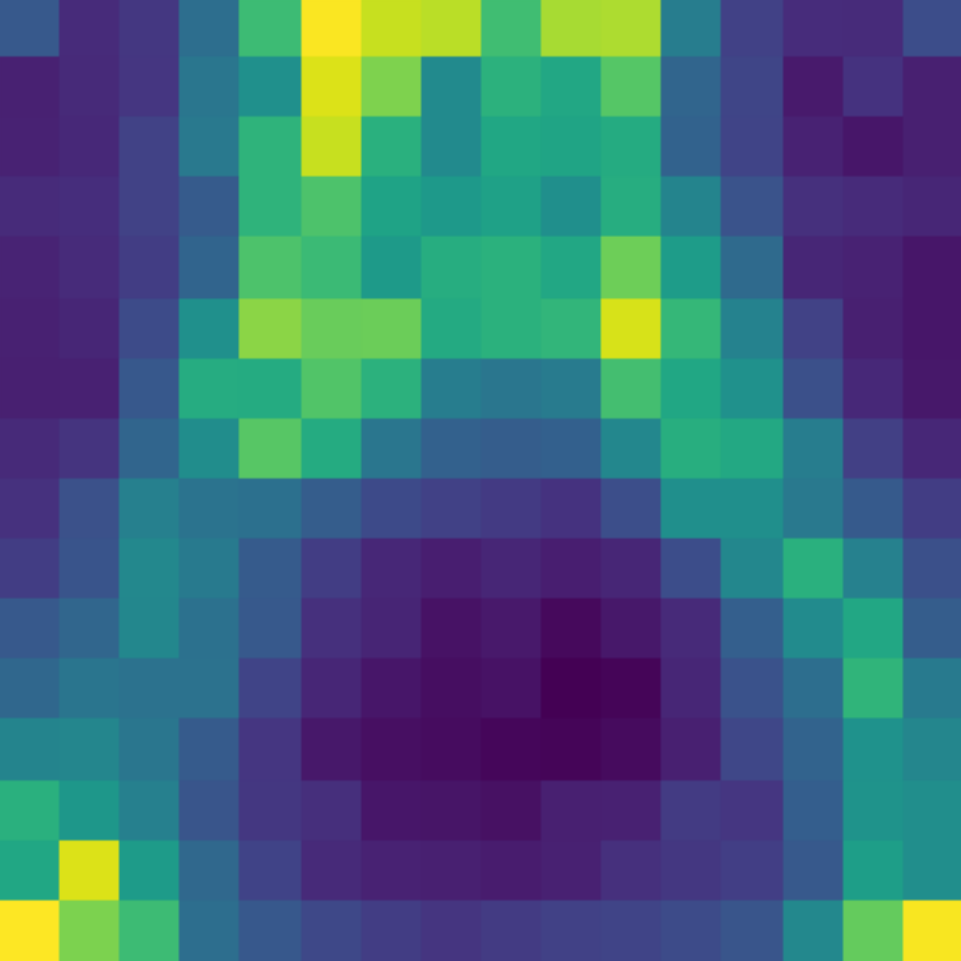}
\end{subfigure}
\begin{subfigure}{.05\textwidth}
  \centering
  \includegraphics[width=1.0\linewidth]{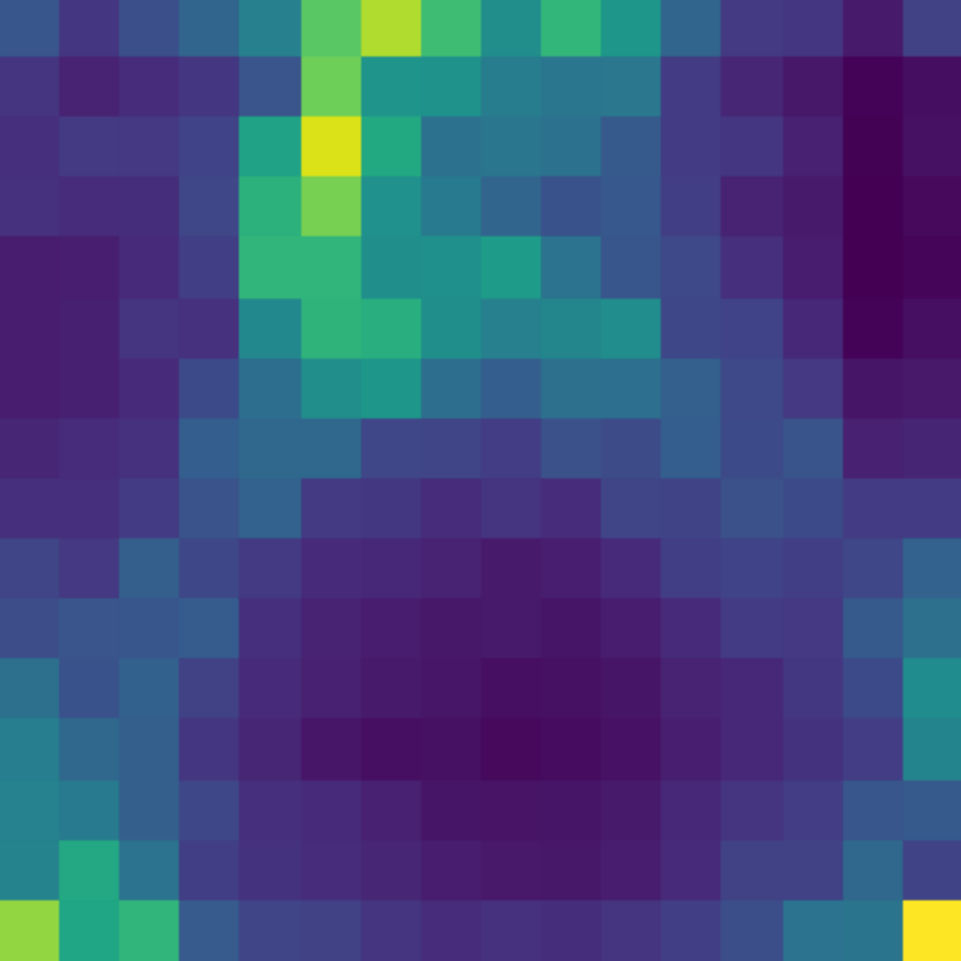}
\end{subfigure}
\begin{subfigure}{.05\textwidth}
  \centering
  \includegraphics[width=1.0\linewidth]{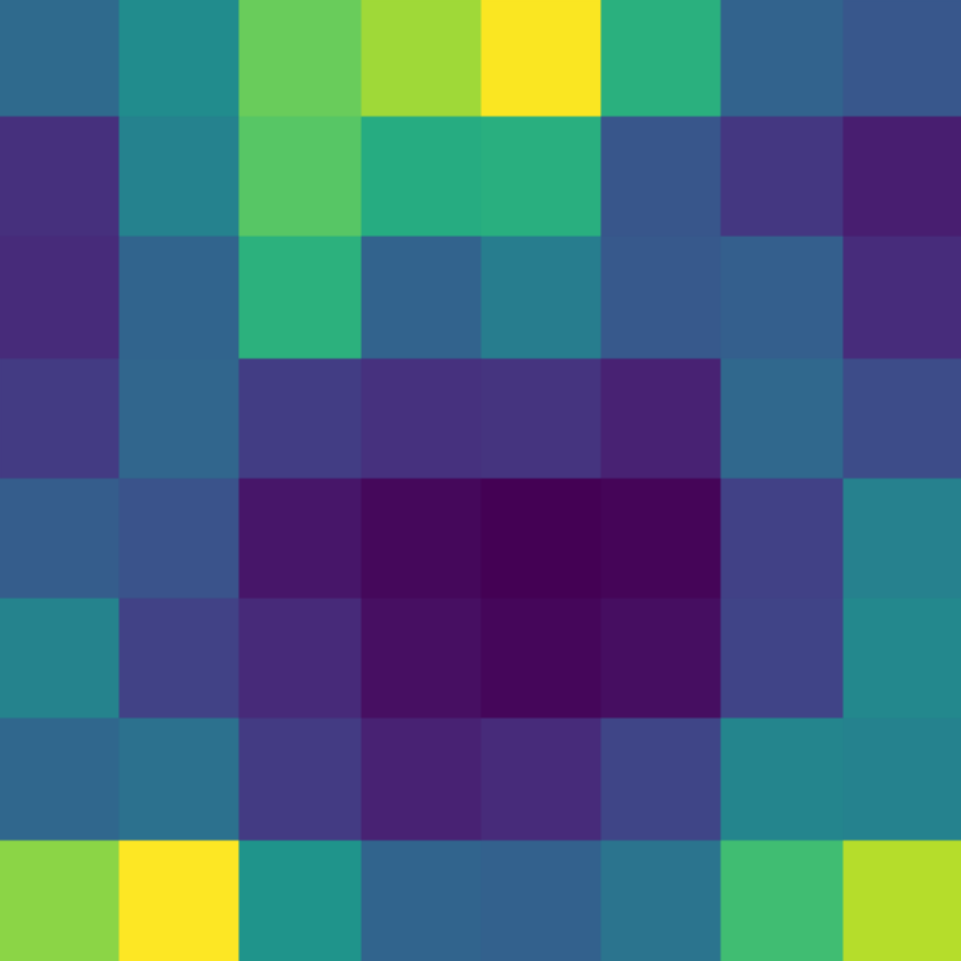}
\end{subfigure}
\begin{subfigure}{.05\textwidth}
  \centering
  \includegraphics[width=1.0\linewidth]{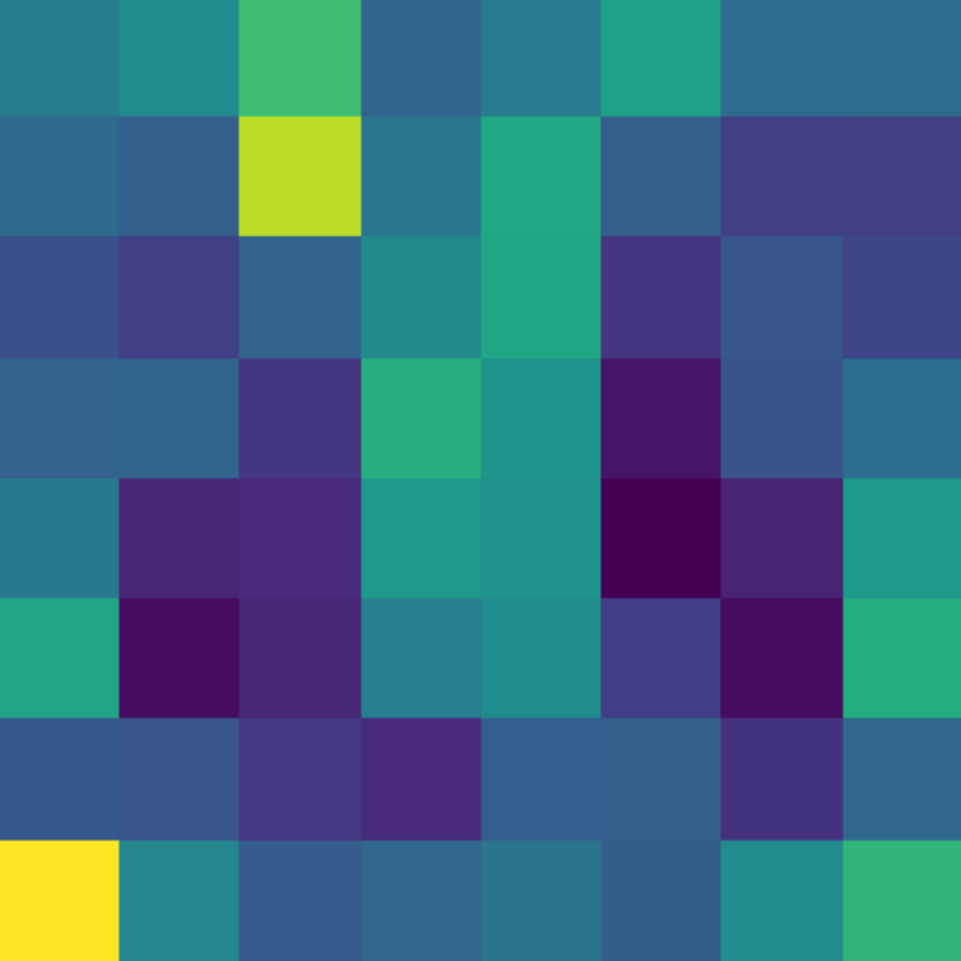}
\end{subfigure}
\begin{subfigure}{.05\textwidth}
  \centering
  \includegraphics[width=1.0\linewidth]{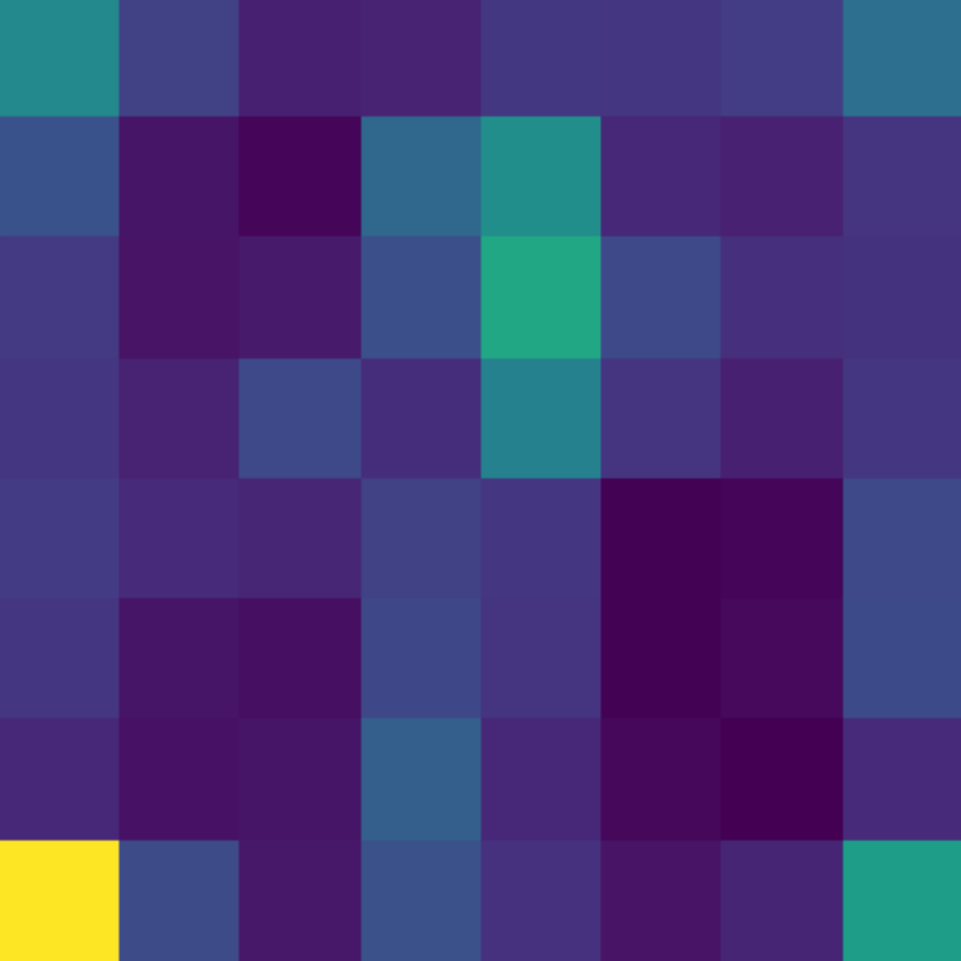}
\end{subfigure}
\begin{subfigure}{.05\textwidth}
  \centering
  \includegraphics[width=1.0\linewidth]{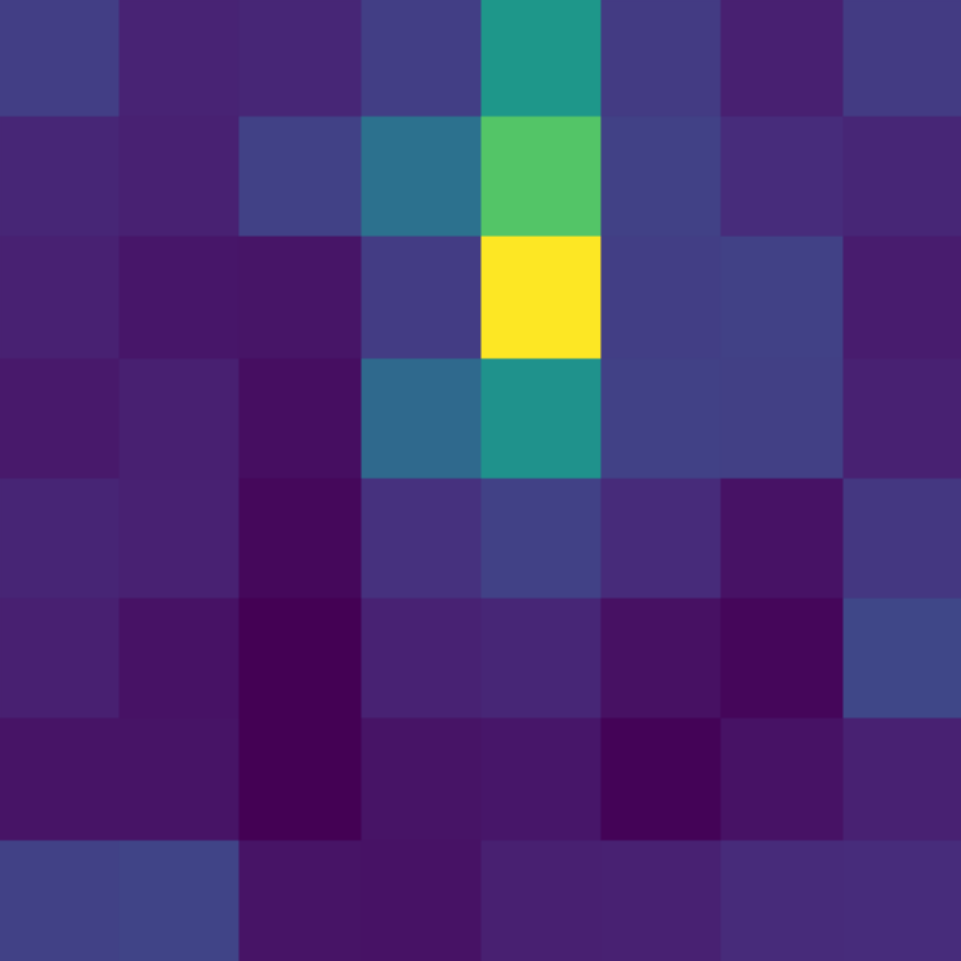}
\end{subfigure}
\begin{subfigure}{.05\textwidth}
  \centering
  \includegraphics[width=1.0\linewidth]{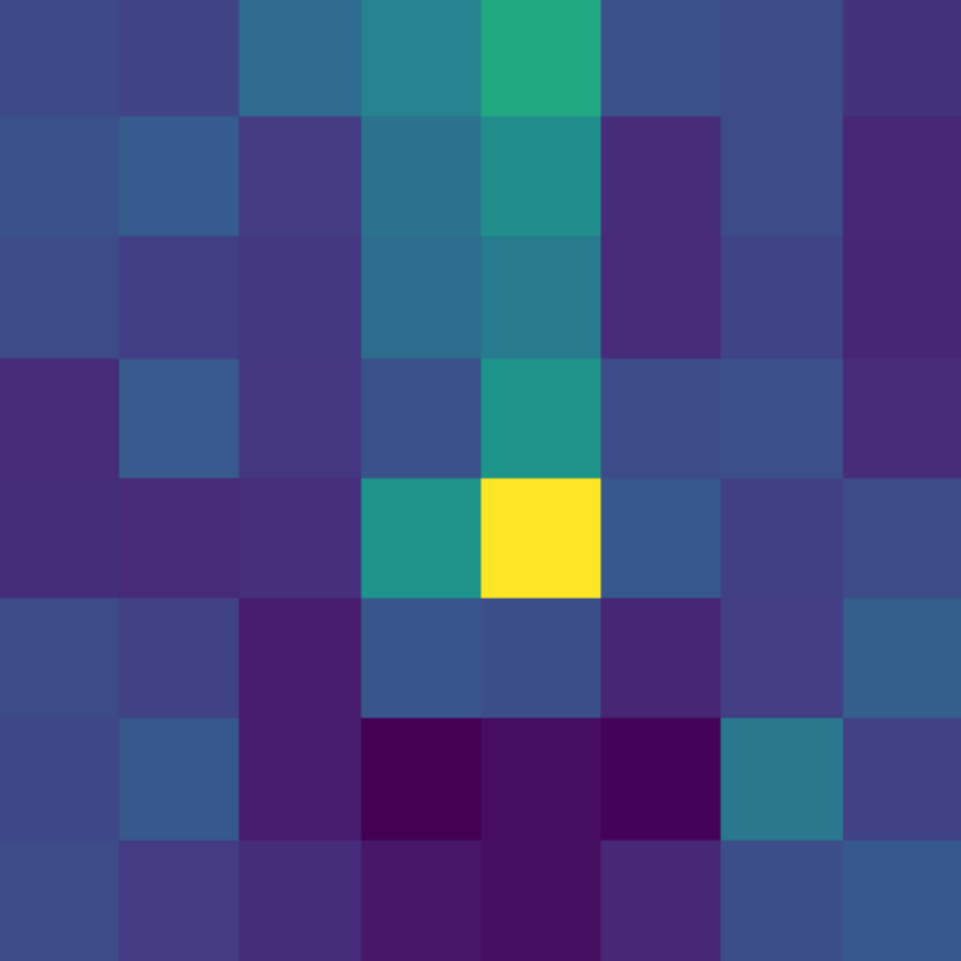}
\end{subfigure}
\begin{subfigure}{.05\textwidth}
  \centering
  \includegraphics[width=1.0\linewidth]{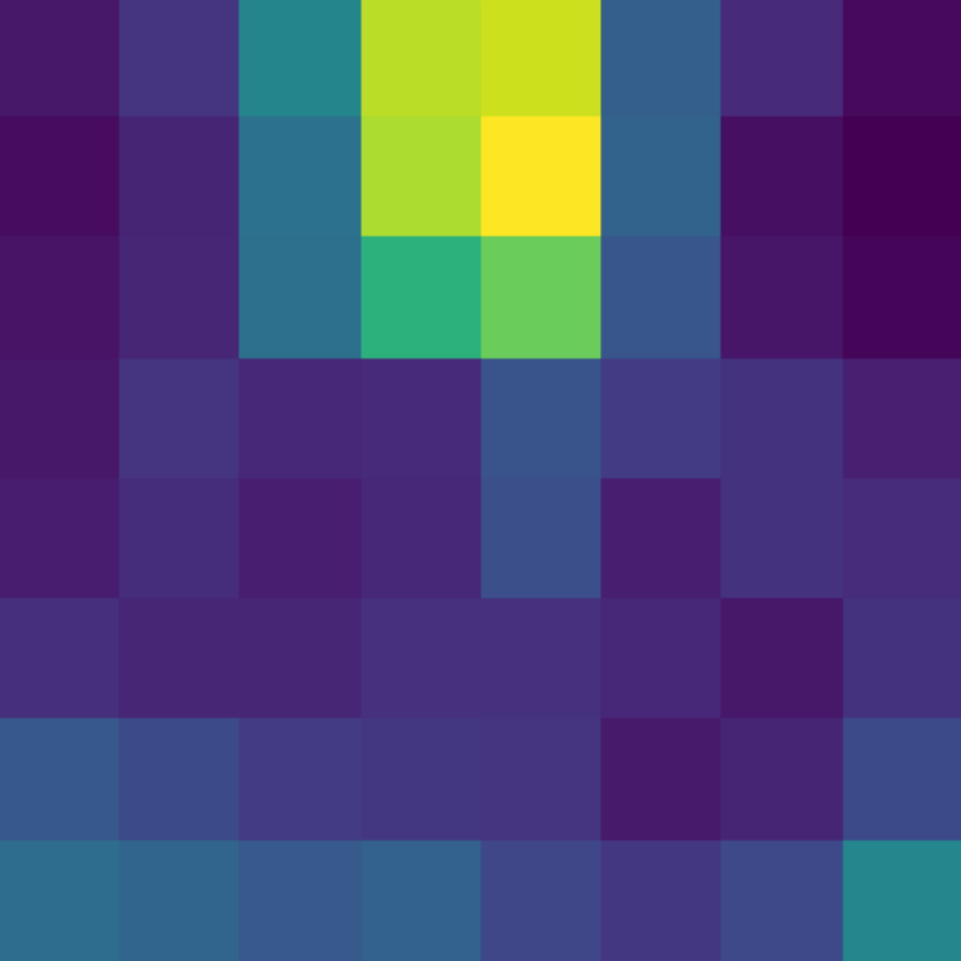}
\end{subfigure}
\begin{subfigure}{.05\textwidth}
  \centering
  \includegraphics[width=1.0\linewidth]{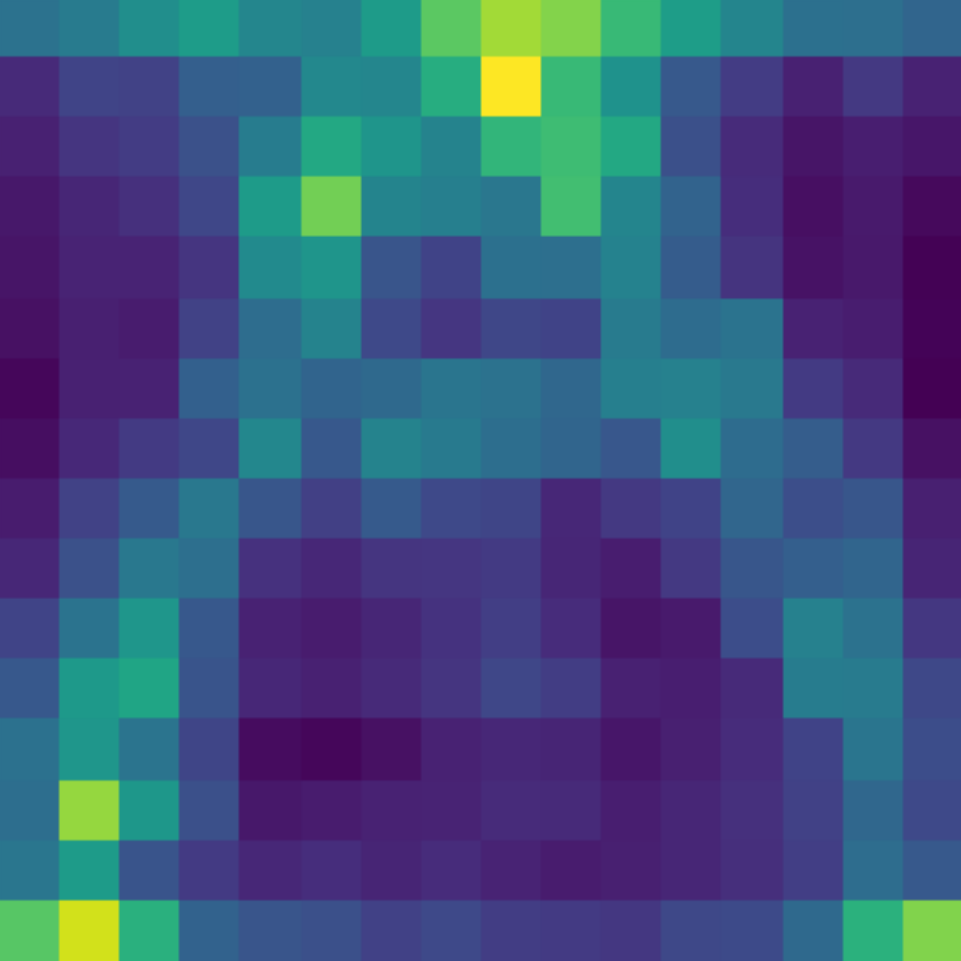}
\end{subfigure}
\begin{subfigure}{.05\textwidth}
  \centering
  \includegraphics[width=1.0\linewidth]{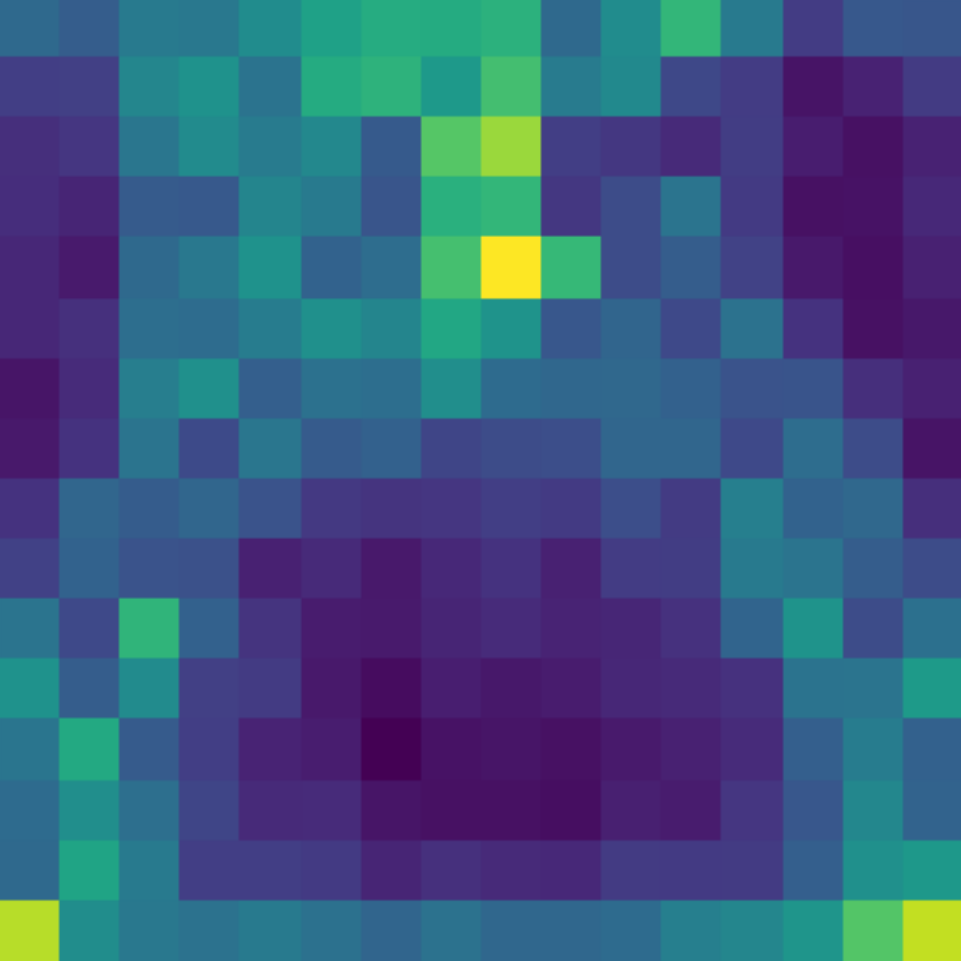}
\end{subfigure}
\begin{subfigure}{.05\textwidth}
  \centering
  \includegraphics[width=1.0\linewidth]{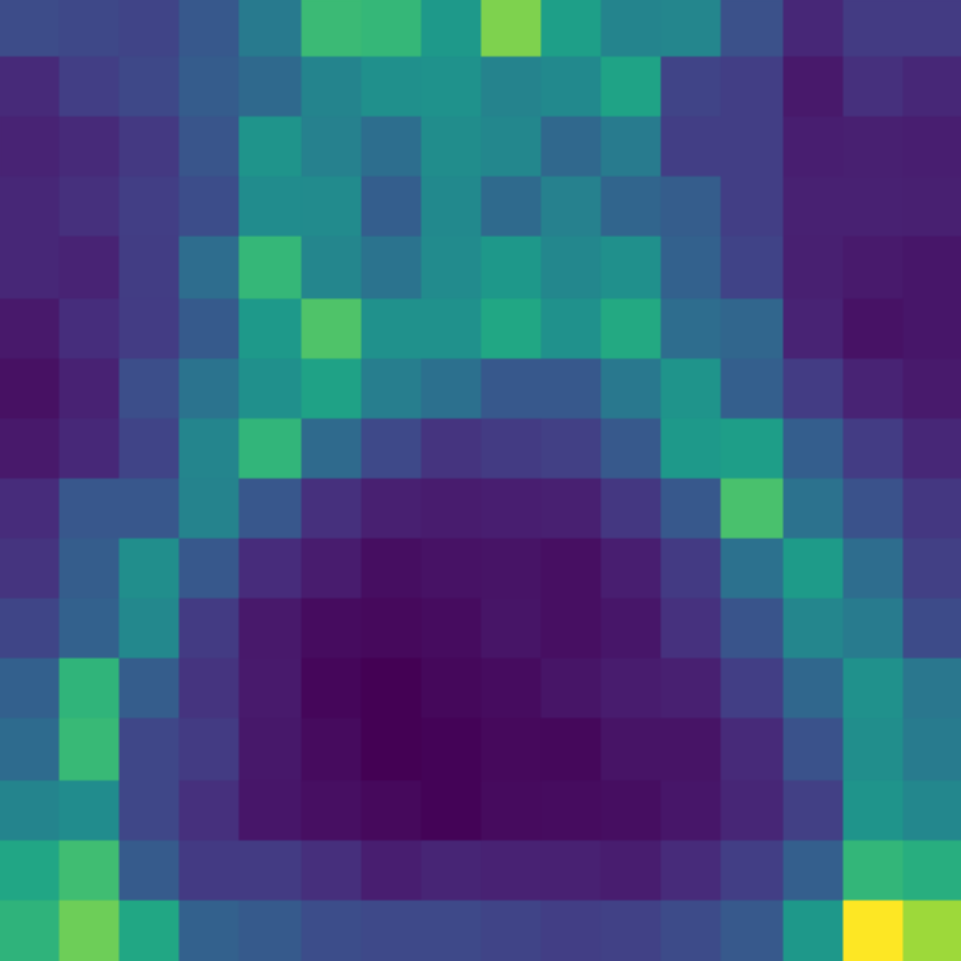}
\end{subfigure}
\begin{subfigure}{.05\textwidth}
  \centering
  \includegraphics[width=1.0\linewidth]{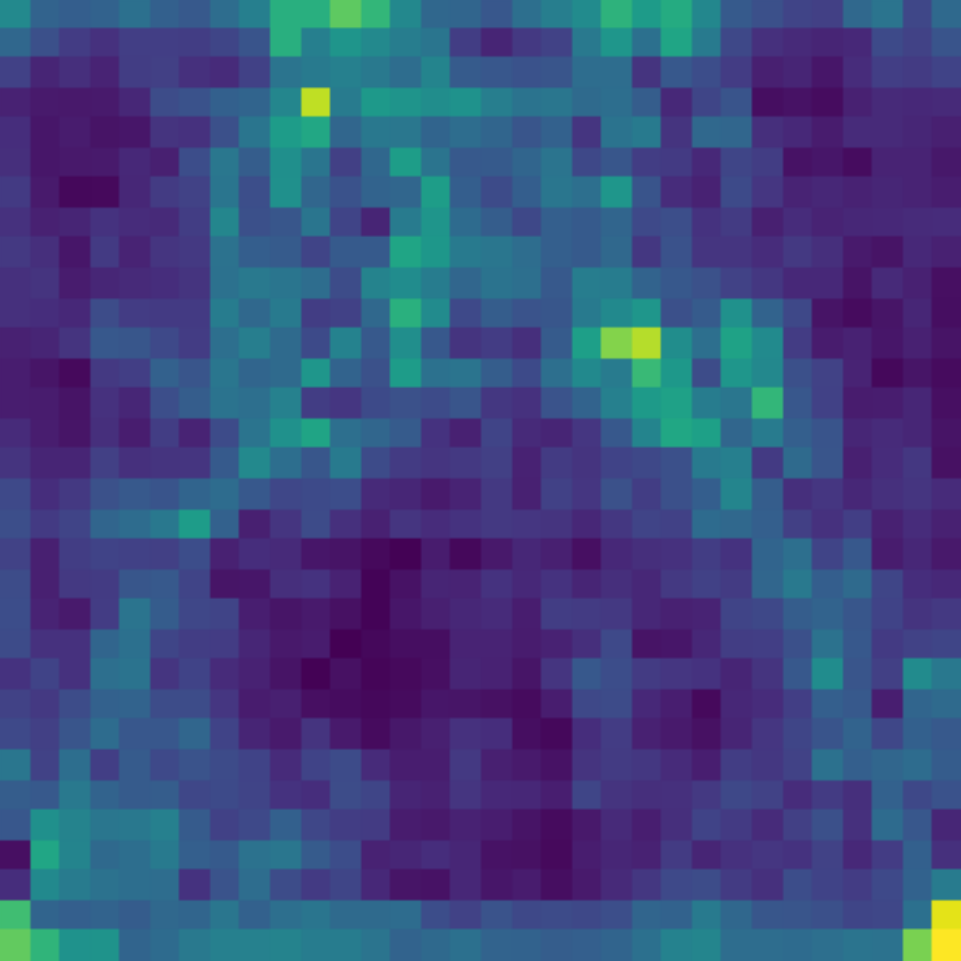}
\end{subfigure}
\begin{subfigure}{.05\textwidth}
  \centering
  \includegraphics[width=1.0\linewidth]{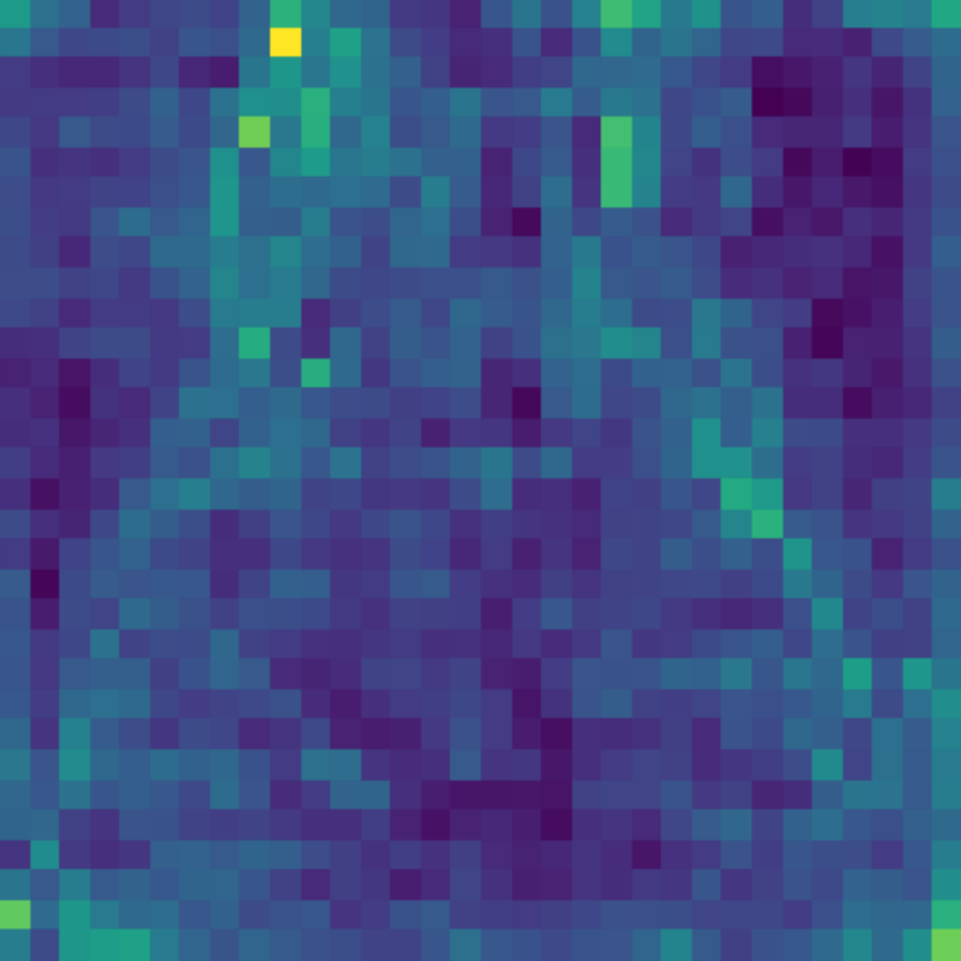}
\end{subfigure}
\begin{subfigure}{.05\textwidth}
  \centering
  \includegraphics[width=1.0\linewidth]{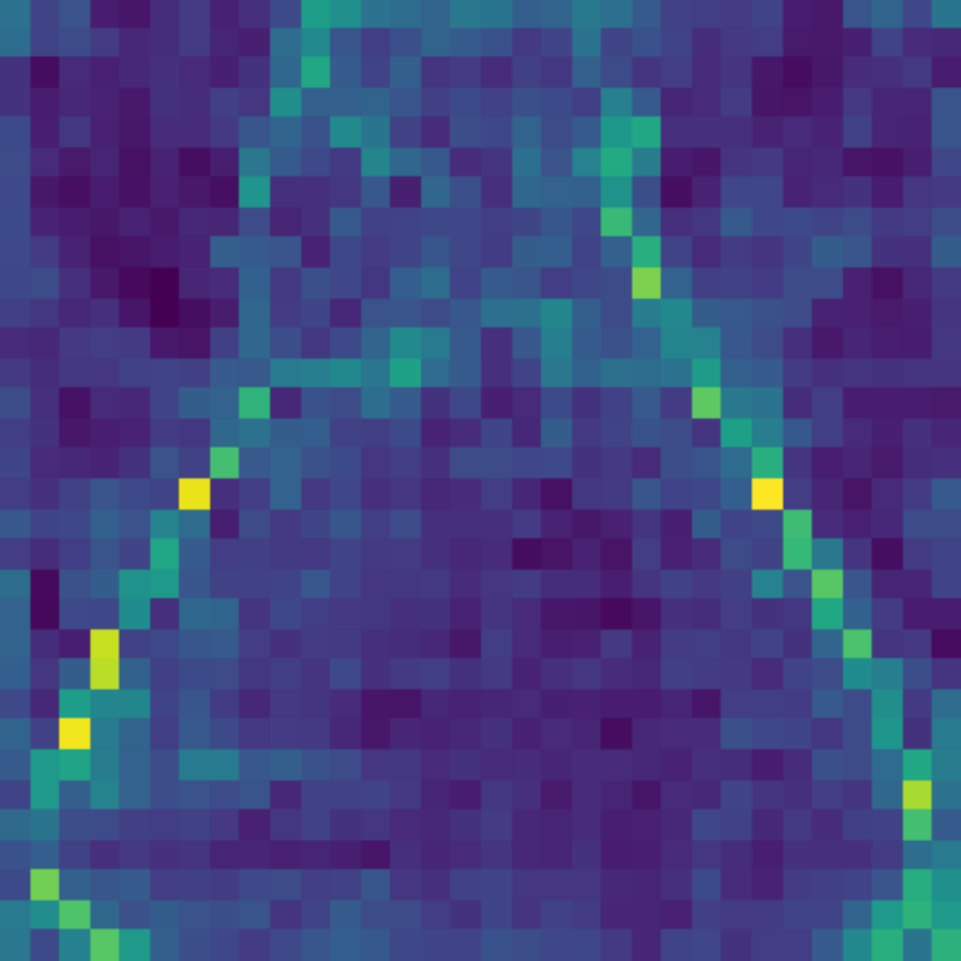}
\end{subfigure}
\\
&
\begin{subfigure}{.05\textwidth}
  \centering
  \includegraphics[width=1.0\linewidth]{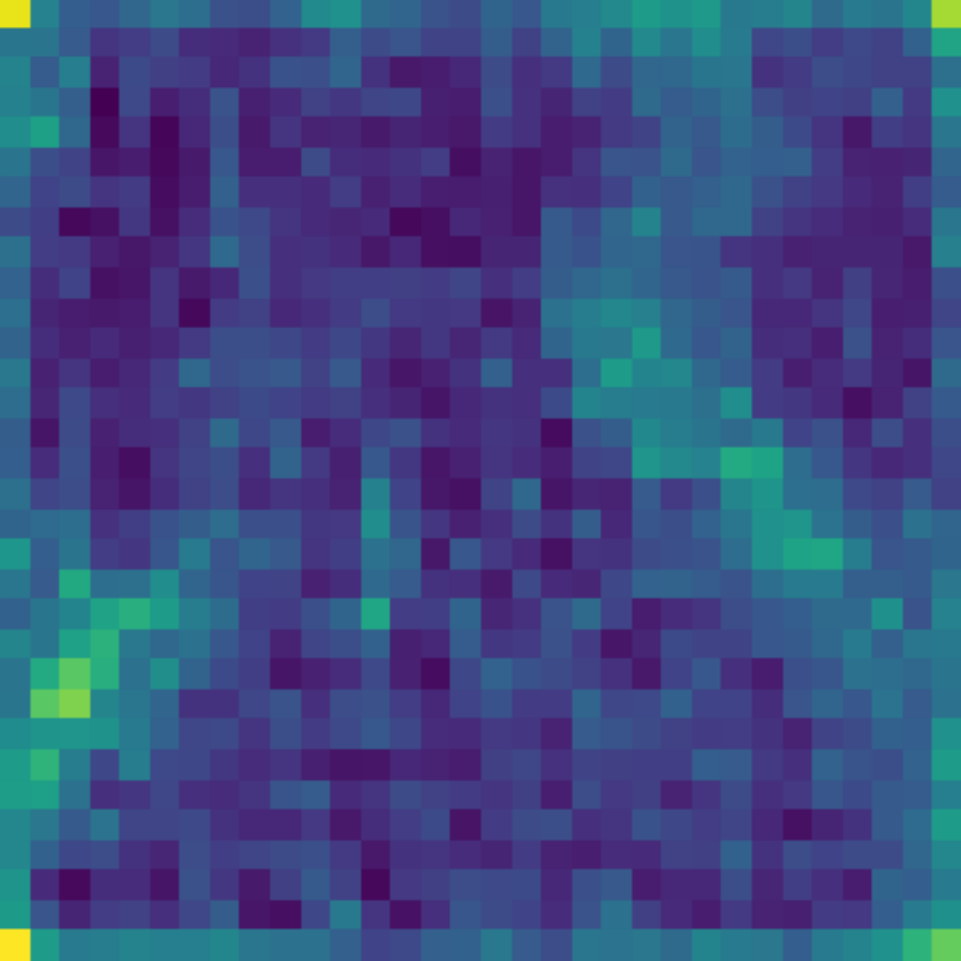}
\end{subfigure}
\begin{subfigure}{.05\textwidth}
  \centering
  \includegraphics[width=1.0\linewidth]{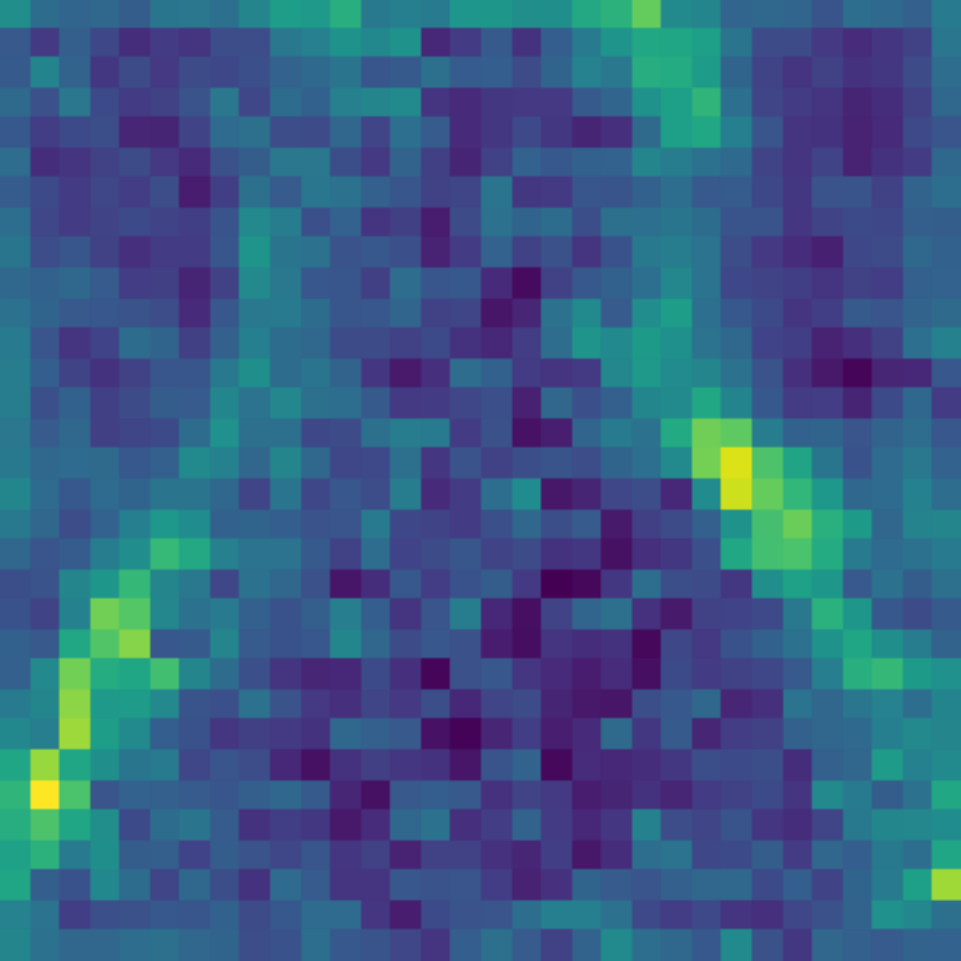}
\end{subfigure}
\begin{subfigure}{.05\textwidth}
  \centering
  \includegraphics[width=1.0\linewidth]{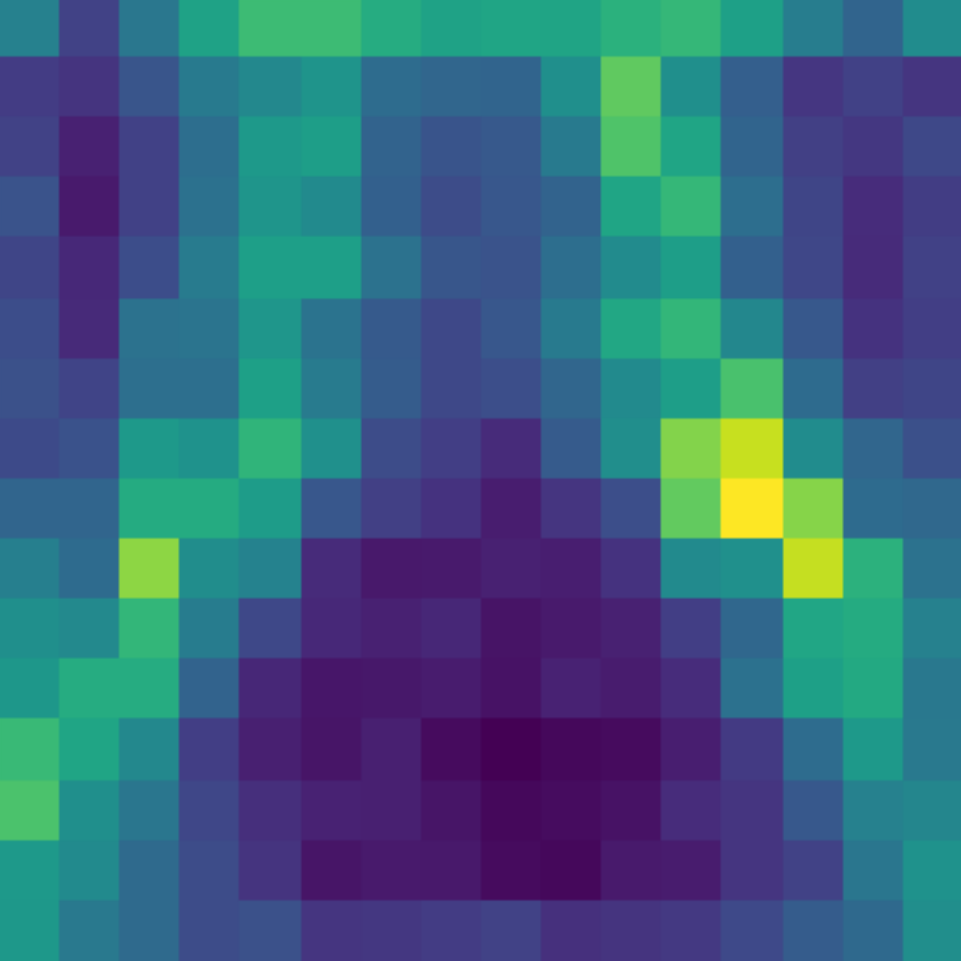}
\end{subfigure}
\begin{subfigure}{.05\textwidth}
  \centering
  \includegraphics[width=1.0\linewidth]{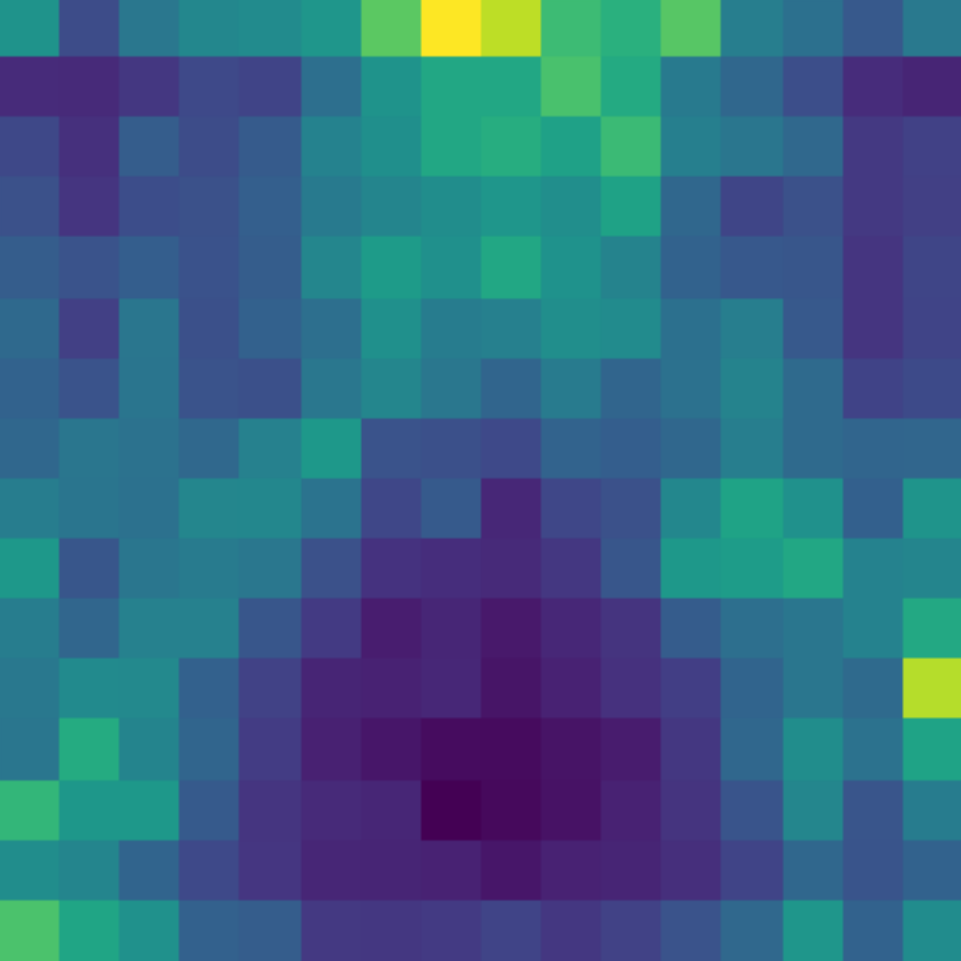}
\end{subfigure}
\begin{subfigure}{.05\textwidth}
  \centering
  \includegraphics[width=1.0\linewidth]{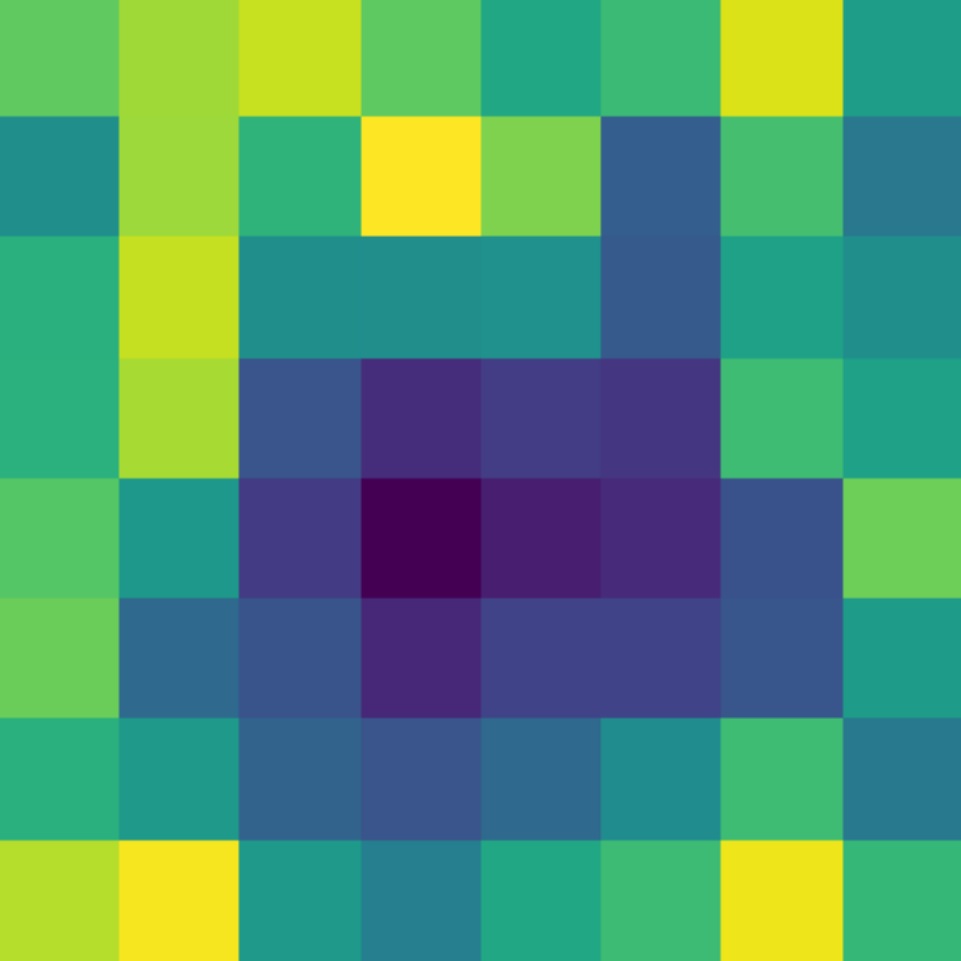}
\end{subfigure}
\begin{subfigure}{.05\textwidth}
  \centering
  \includegraphics[width=1.0\linewidth]{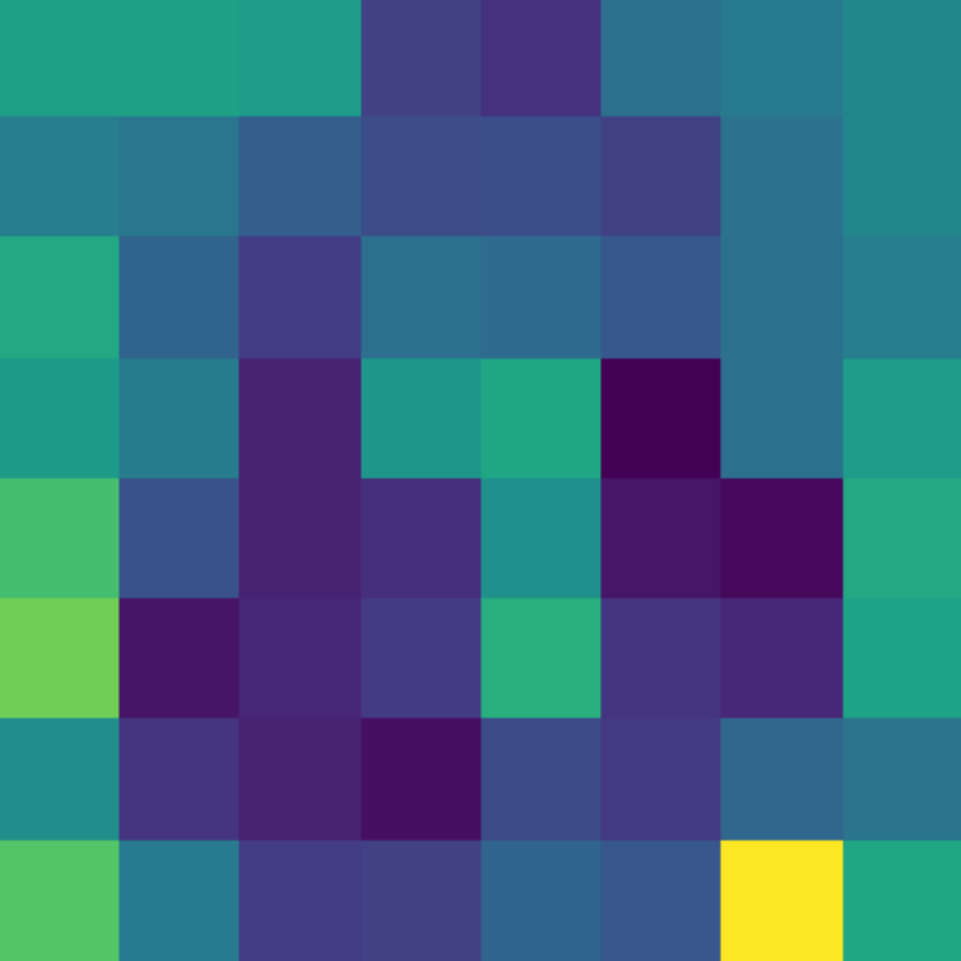}
\end{subfigure}
\begin{subfigure}{.05\textwidth}
  \centering
  \includegraphics[width=1.0\linewidth]{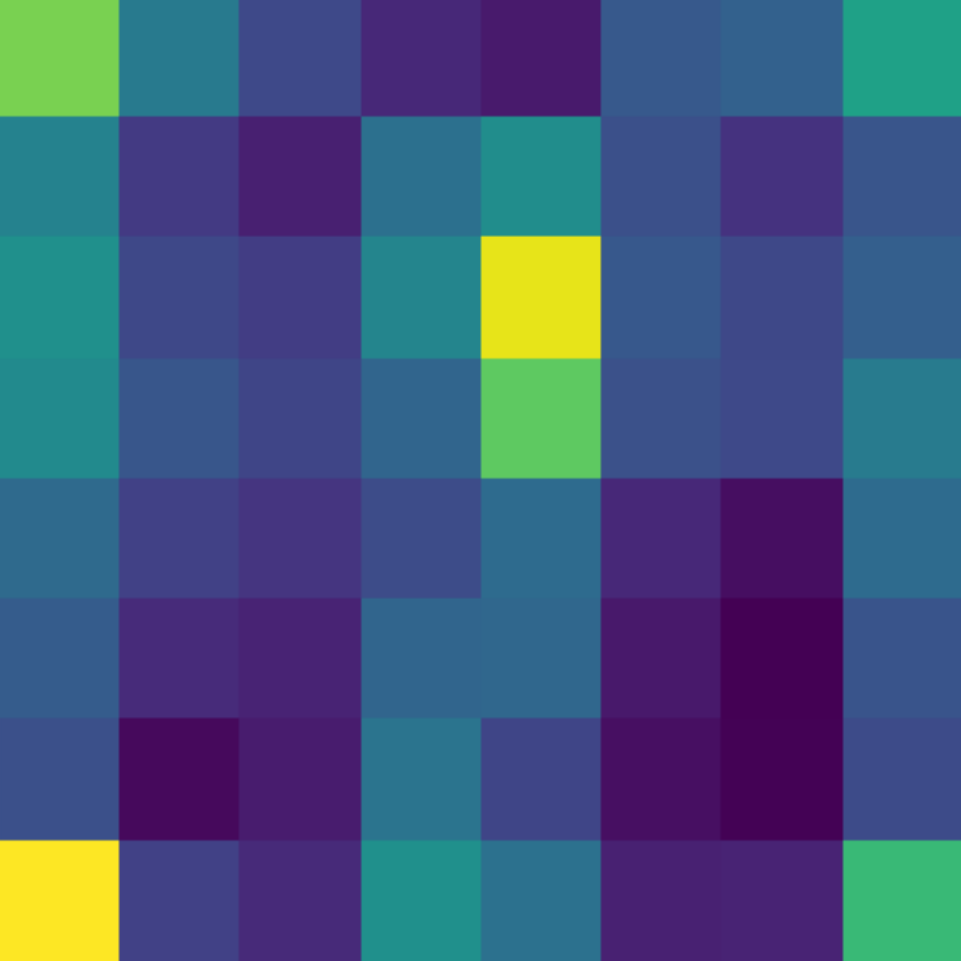}
\end{subfigure}
\begin{subfigure}{.05\textwidth}
  \centering
  \includegraphics[width=1.0\linewidth]{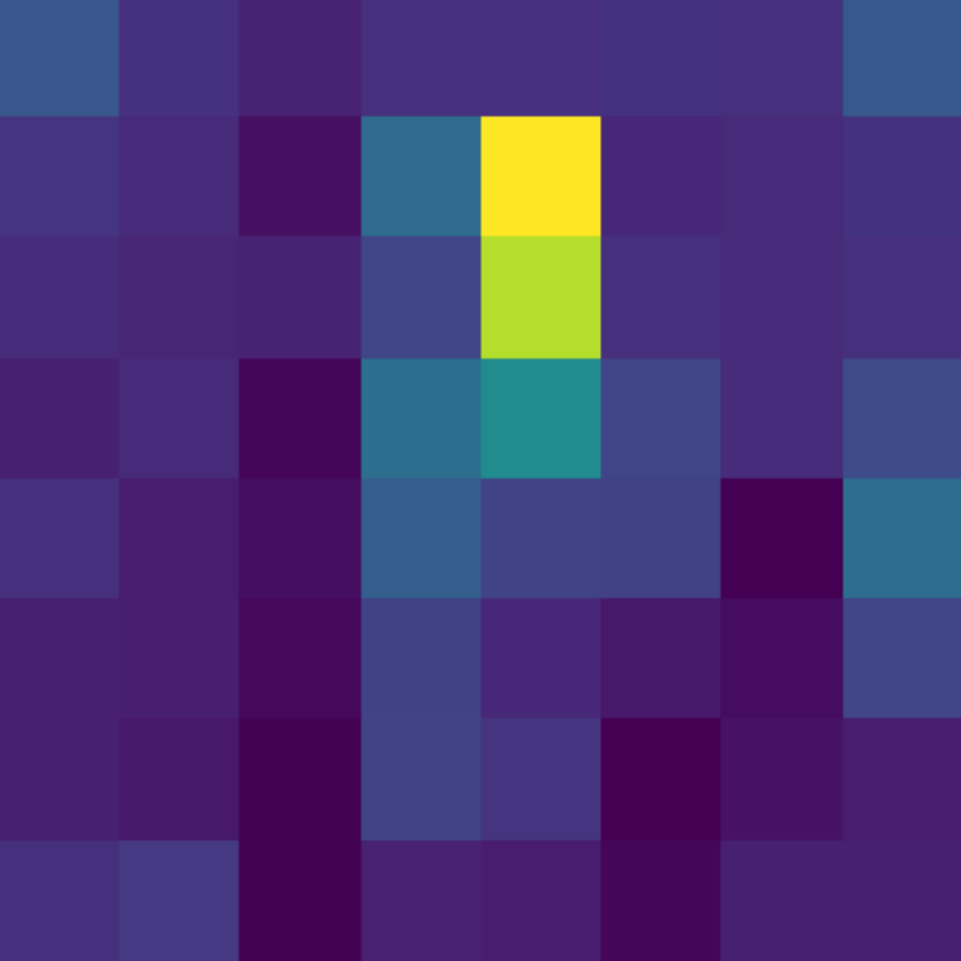}
\end{subfigure}
\begin{subfigure}{.05\textwidth}
  \centering
  \includegraphics[width=1.0\linewidth]{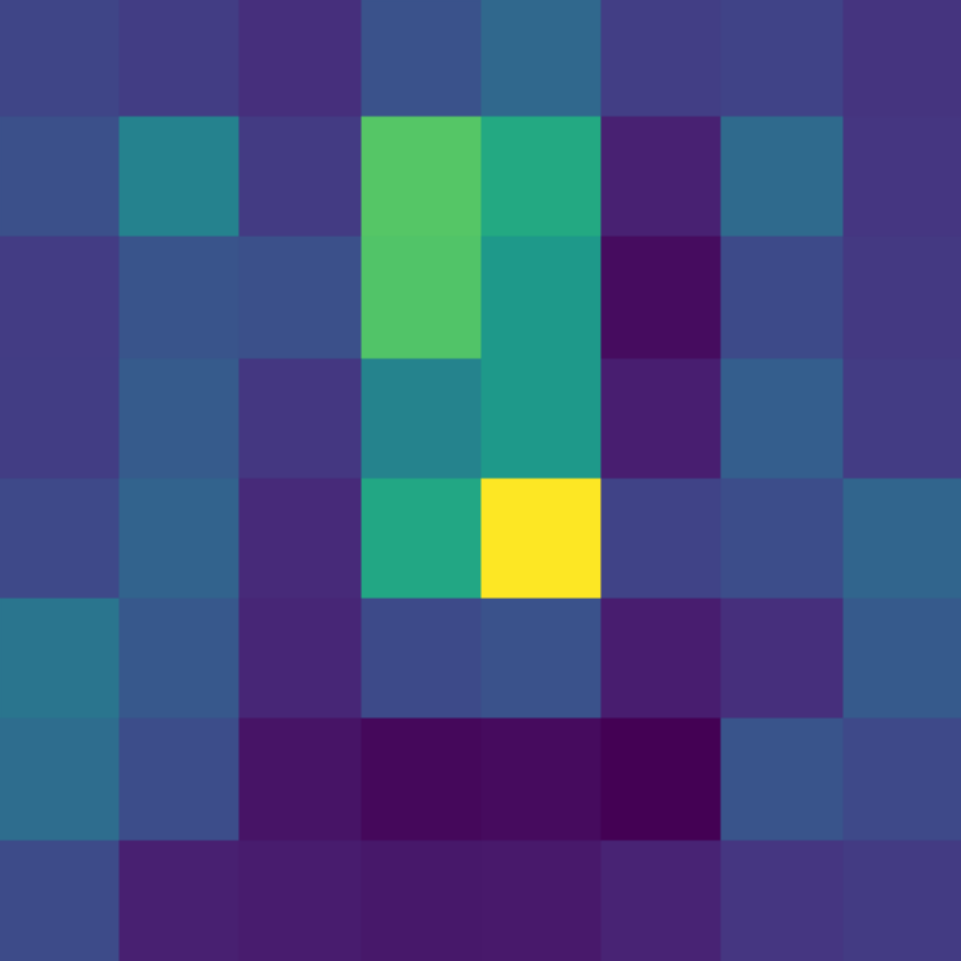}
\end{subfigure}
\begin{subfigure}{.05\textwidth}
  \centering
  \includegraphics[width=1.0\linewidth]{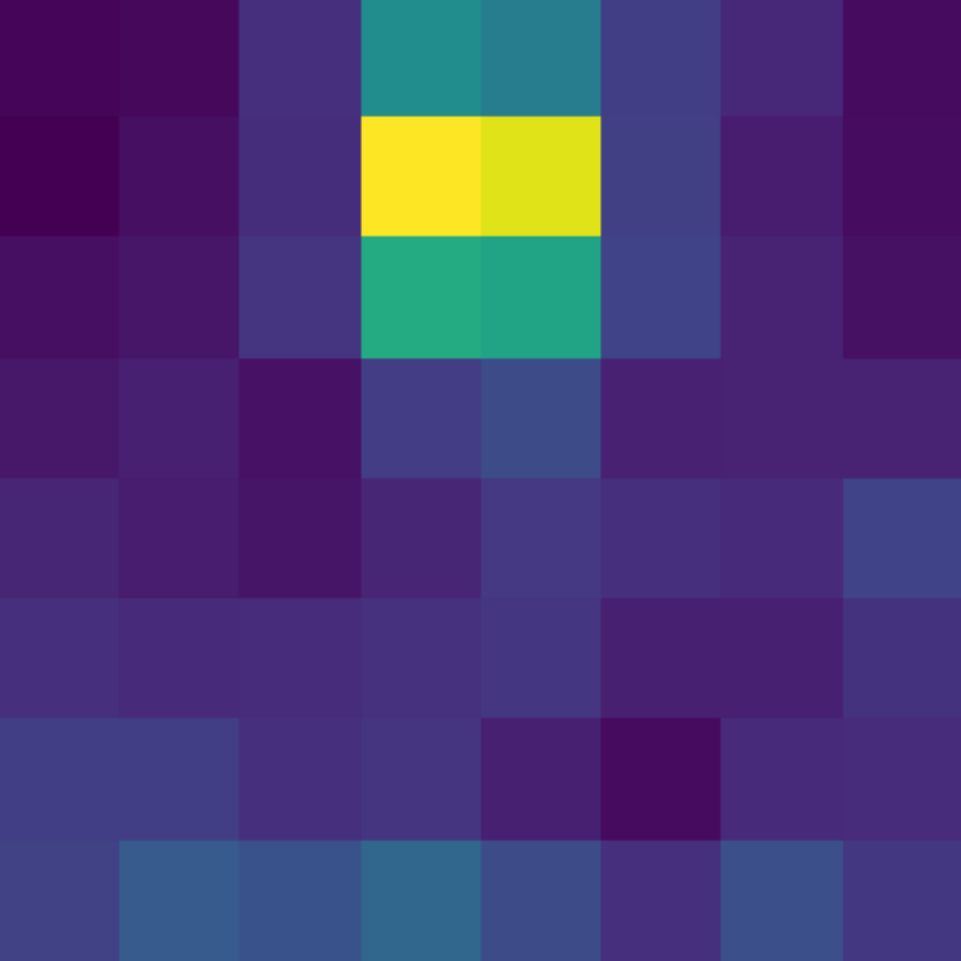}
\end{subfigure}
\begin{subfigure}{.05\textwidth}
  \centering
  \includegraphics[width=1.0\linewidth]{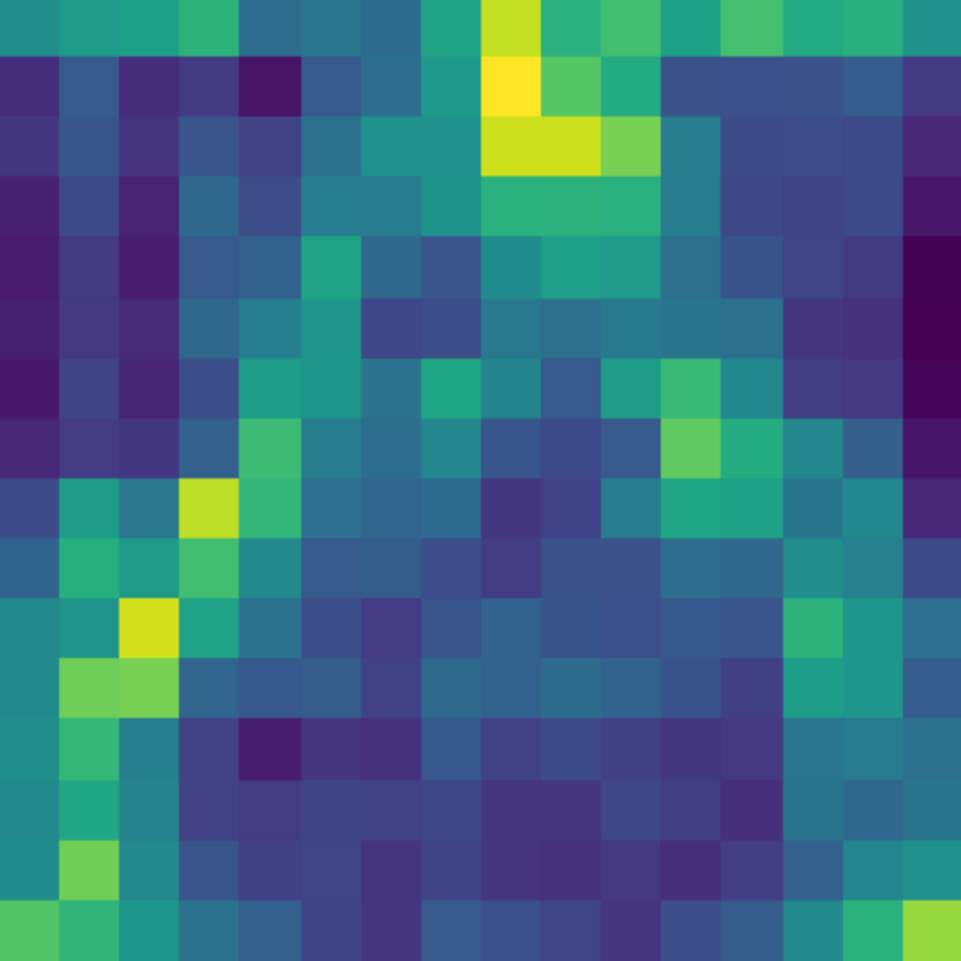}
\end{subfigure}
\begin{subfigure}{.05\textwidth}
  \centering
  \includegraphics[width=1.0\linewidth]{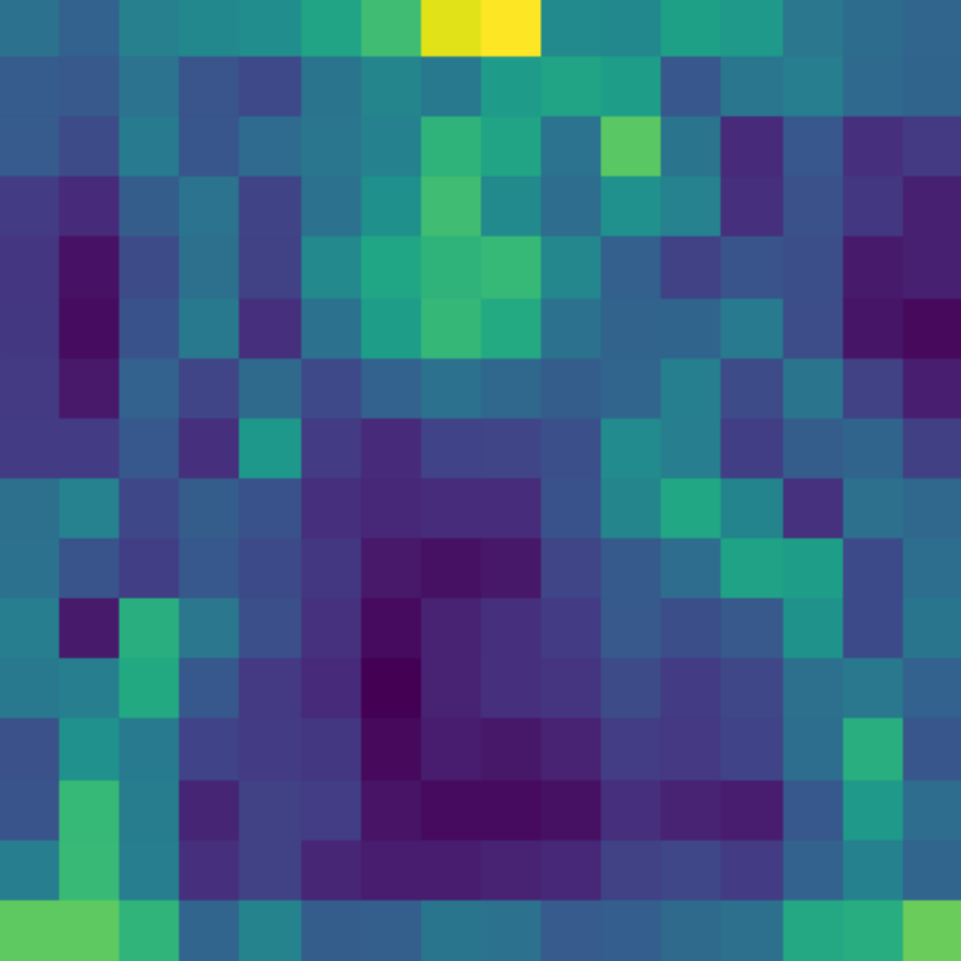}
\end{subfigure}
\begin{subfigure}{.05\textwidth}
  \centering
  \includegraphics[width=1.0\linewidth]{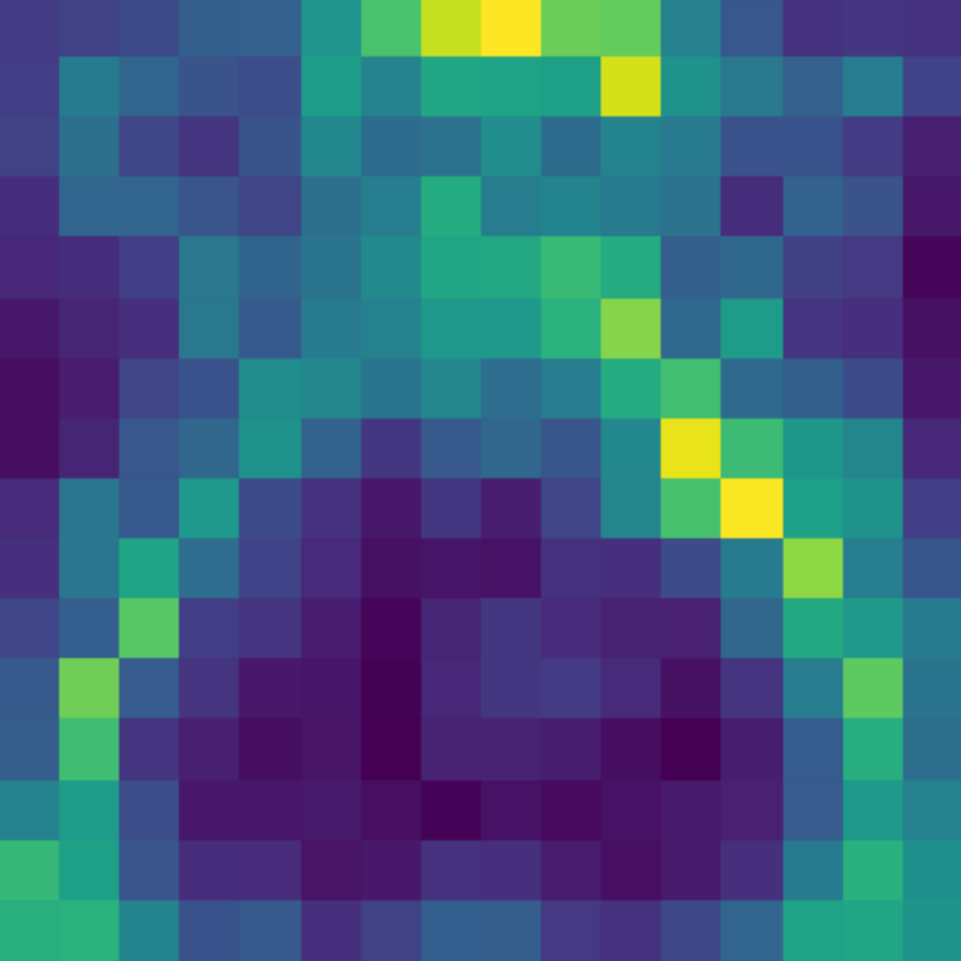}
\end{subfigure}
\begin{subfigure}{.05\textwidth}
  \centering
  \includegraphics[width=1.0\linewidth]{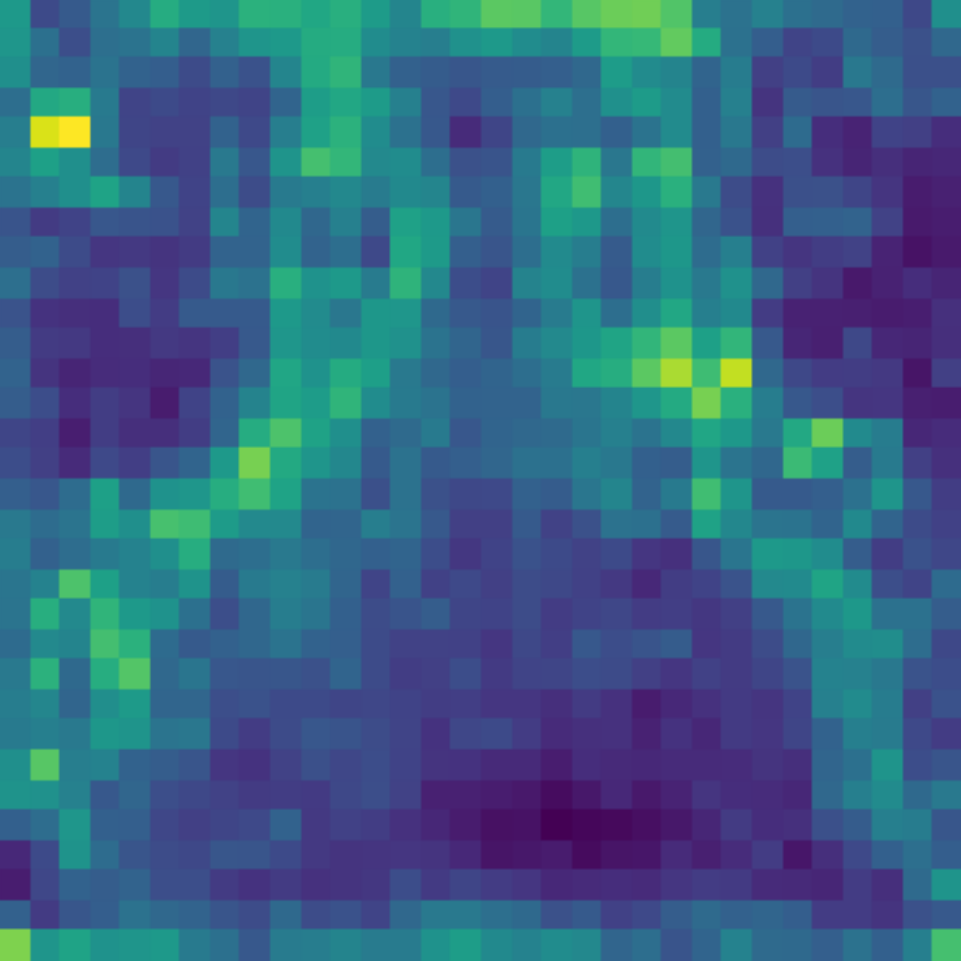}
\end{subfigure}
\begin{subfigure}{.05\textwidth}
  \centering
  \includegraphics[width=1.0\linewidth]{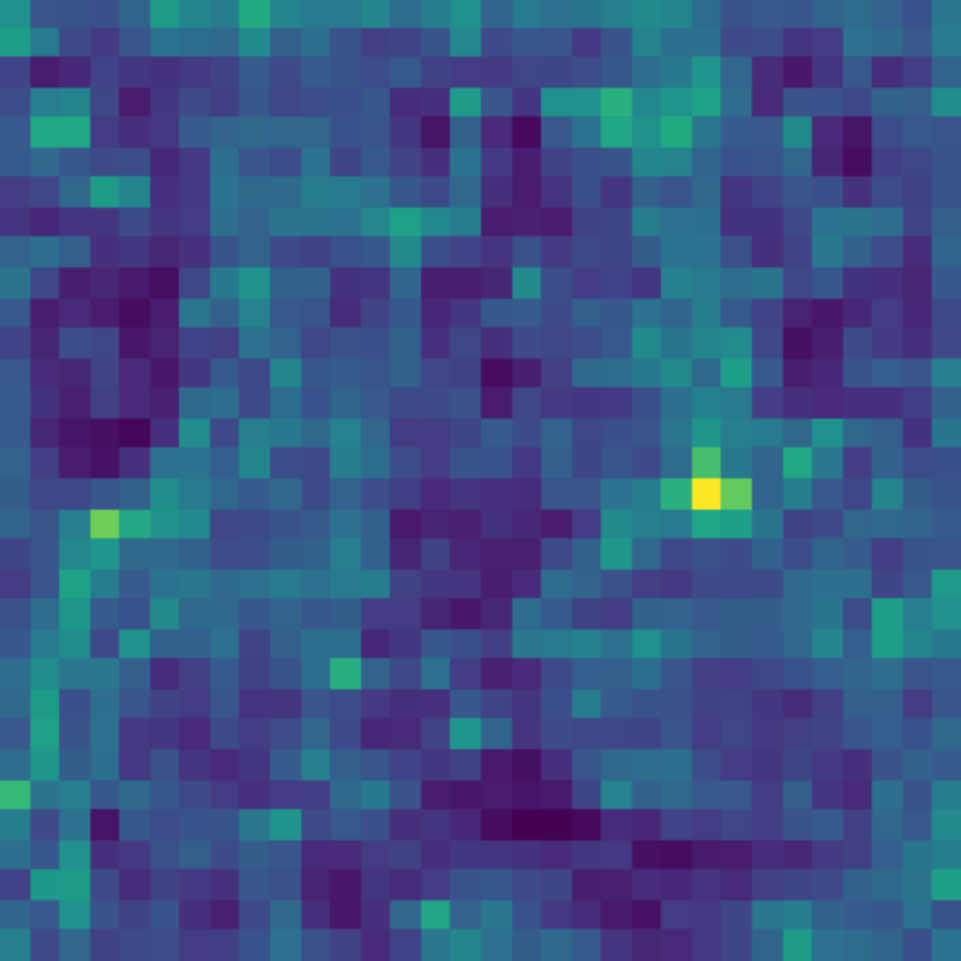}
\end{subfigure}
\begin{subfigure}{.05\textwidth}
  \centering
  \includegraphics[width=1.0\linewidth]{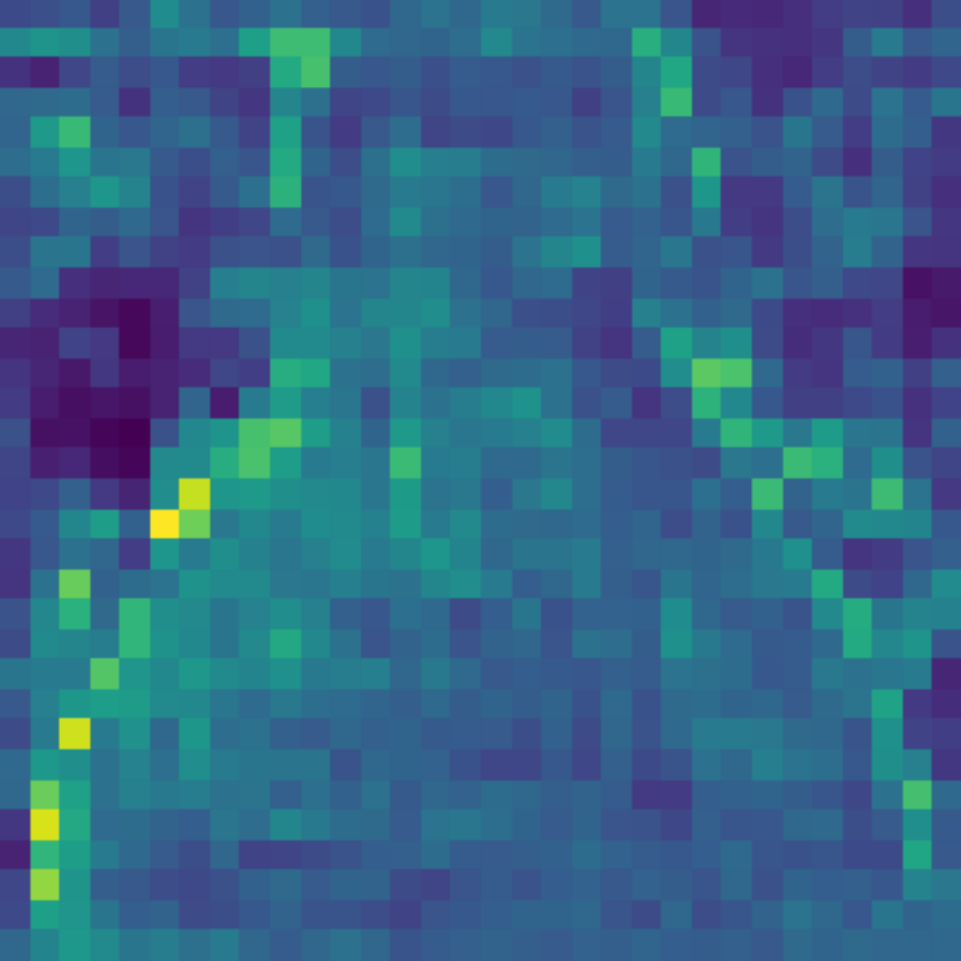}
\end{subfigure}
\\
&
\begin{subfigure}{.05\textwidth}
  \centering
  \includegraphics[width=1.0\linewidth]{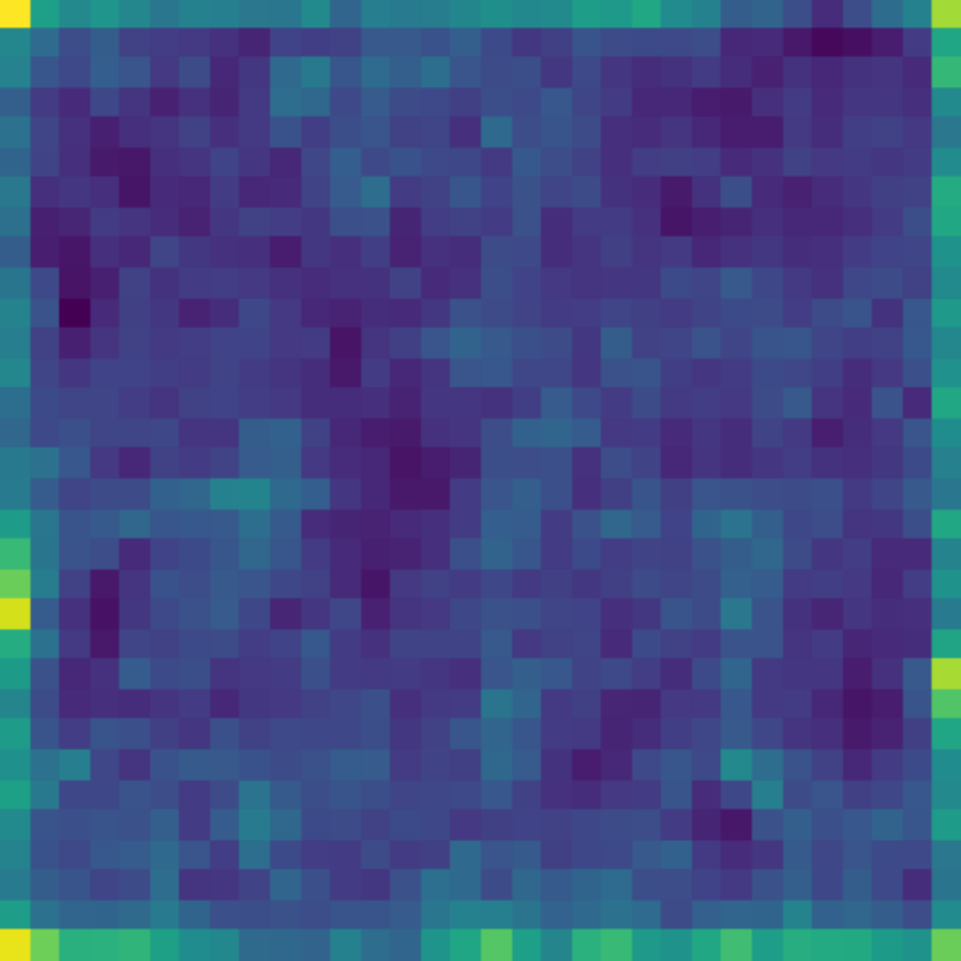}
  \caption*{In.~7}
\end{subfigure}
\begin{subfigure}{.05\textwidth}
  \centering
  \includegraphics[width=1.0\linewidth]{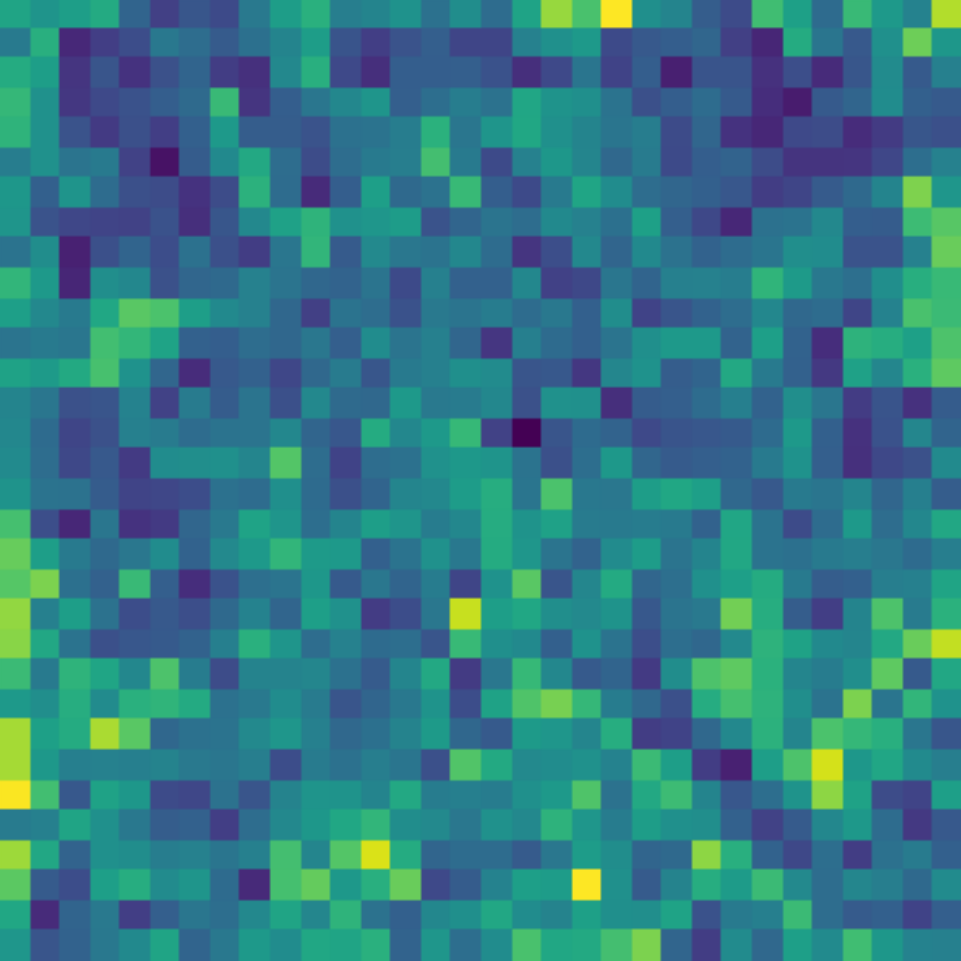}
  \caption*{In.~8}
\end{subfigure}
\begin{subfigure}{.05\textwidth}
  \centering
  \includegraphics[width=1.0\linewidth]{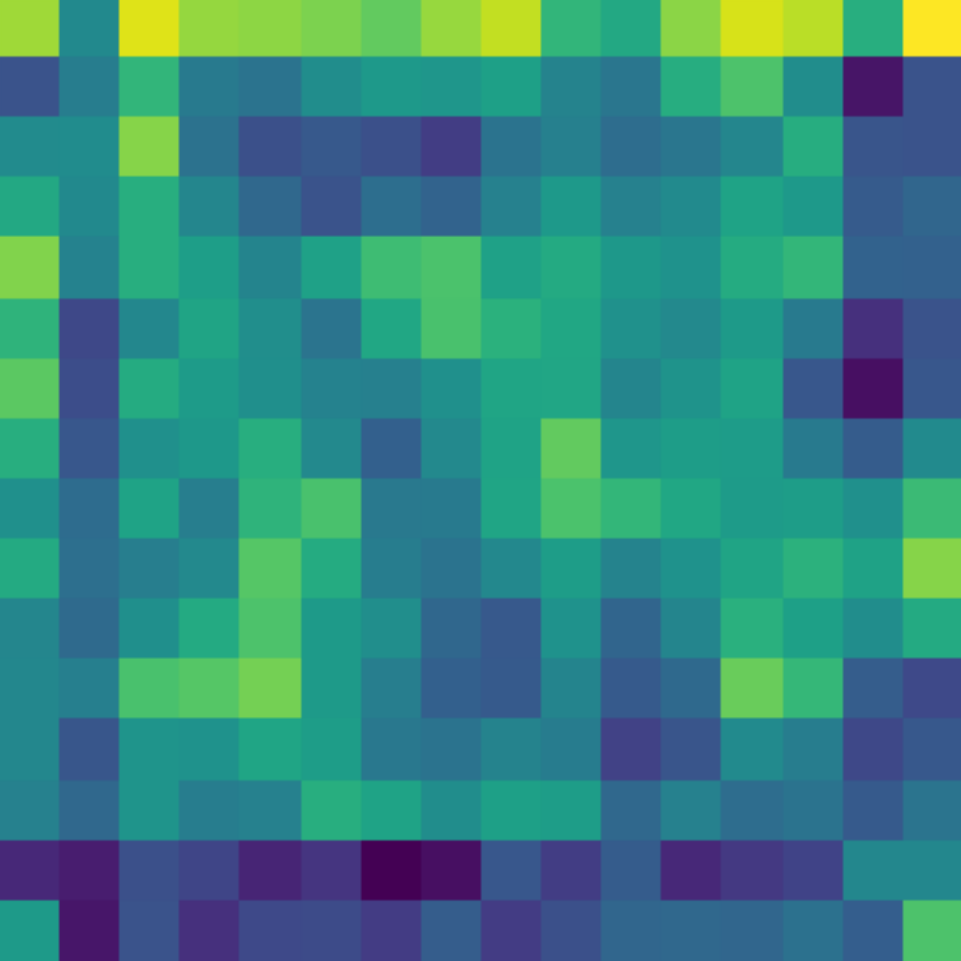}
  \caption*{In.~10}
\end{subfigure}
\begin{subfigure}{.05\textwidth}
  \centering
  \includegraphics[width=1.0\linewidth]{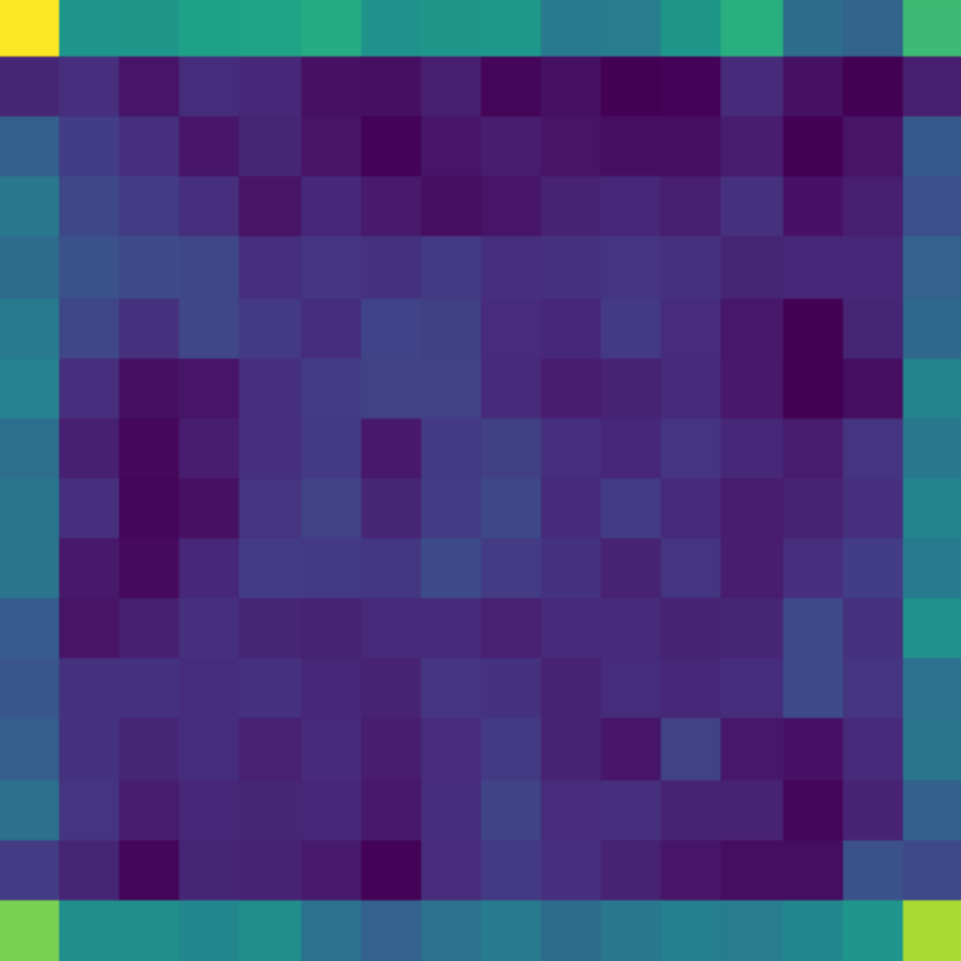}
  \caption*{In.~11}
\end{subfigure}
\begin{subfigure}{.05\textwidth}
  \centering
  \includegraphics[width=1.0\linewidth]{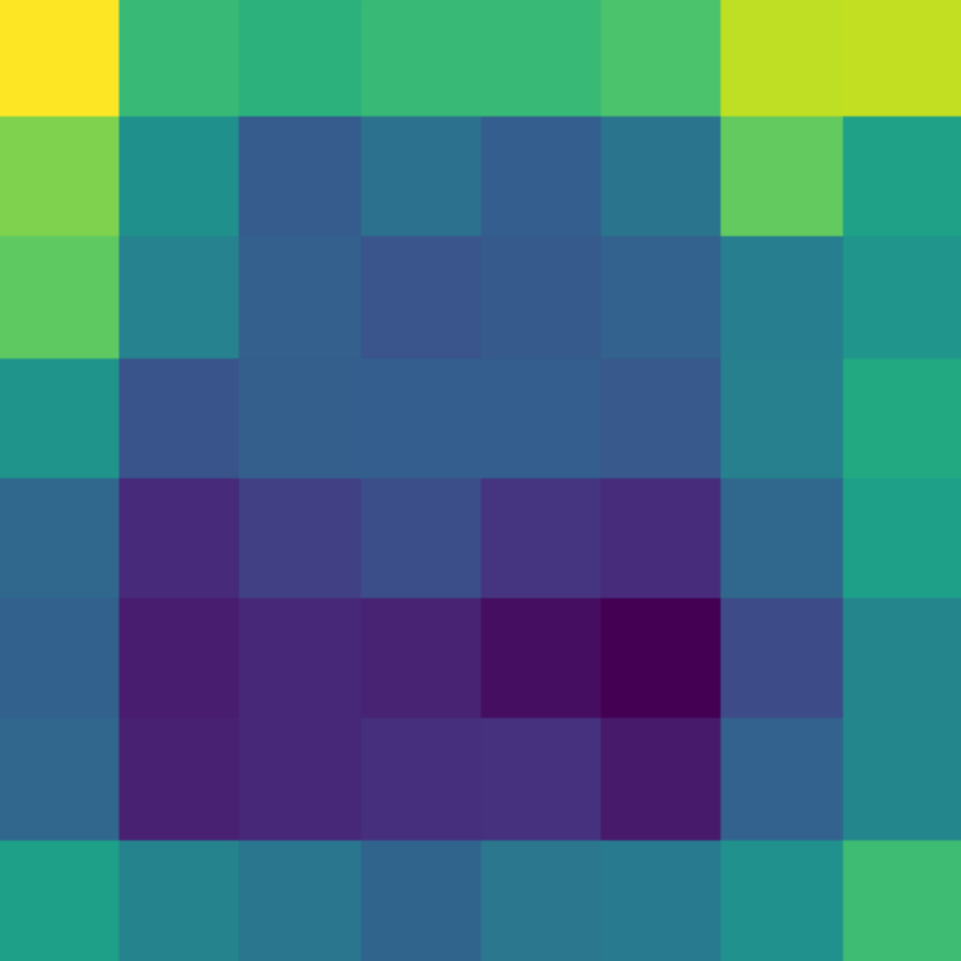}
  \caption*{In.~13}
\end{subfigure}
\begin{subfigure}{.05\textwidth}
  \centering
  \includegraphics[width=1.0\linewidth]{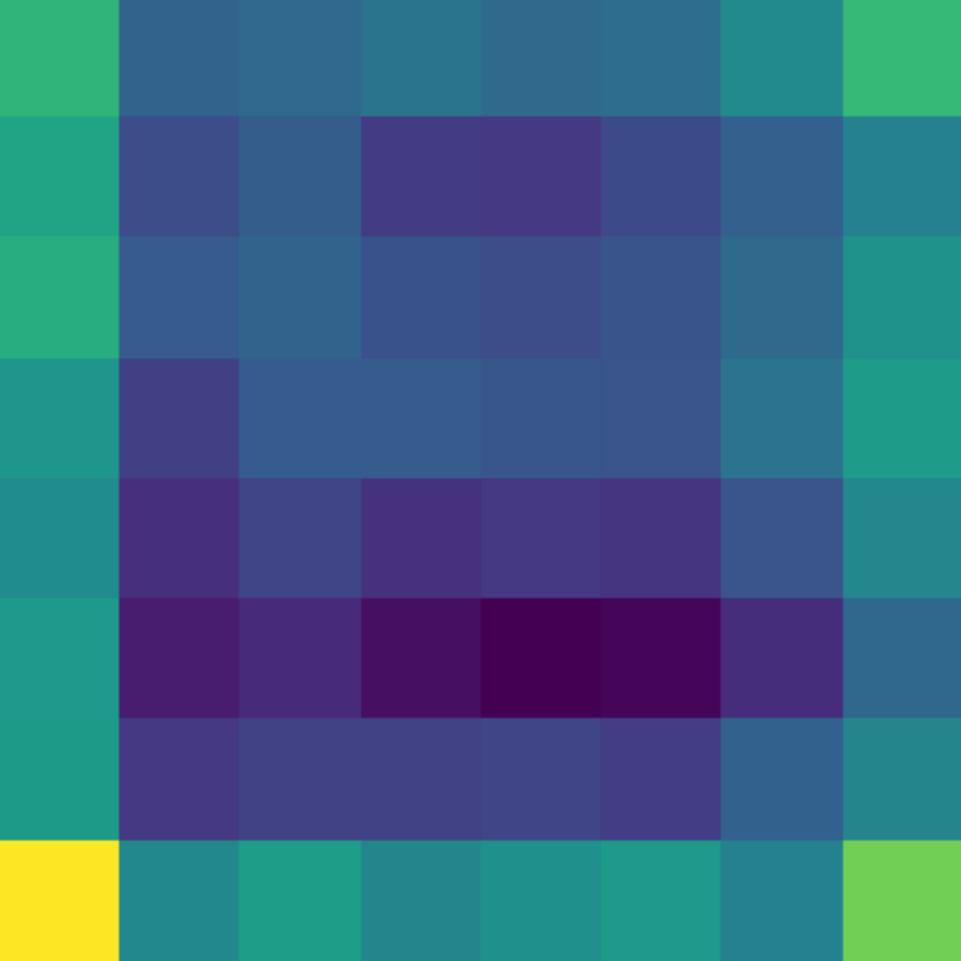}
  \caption*{In.~14}
\end{subfigure}
\begin{subfigure}{.05\textwidth}
  \centering
  \includegraphics[width=1.0\linewidth]{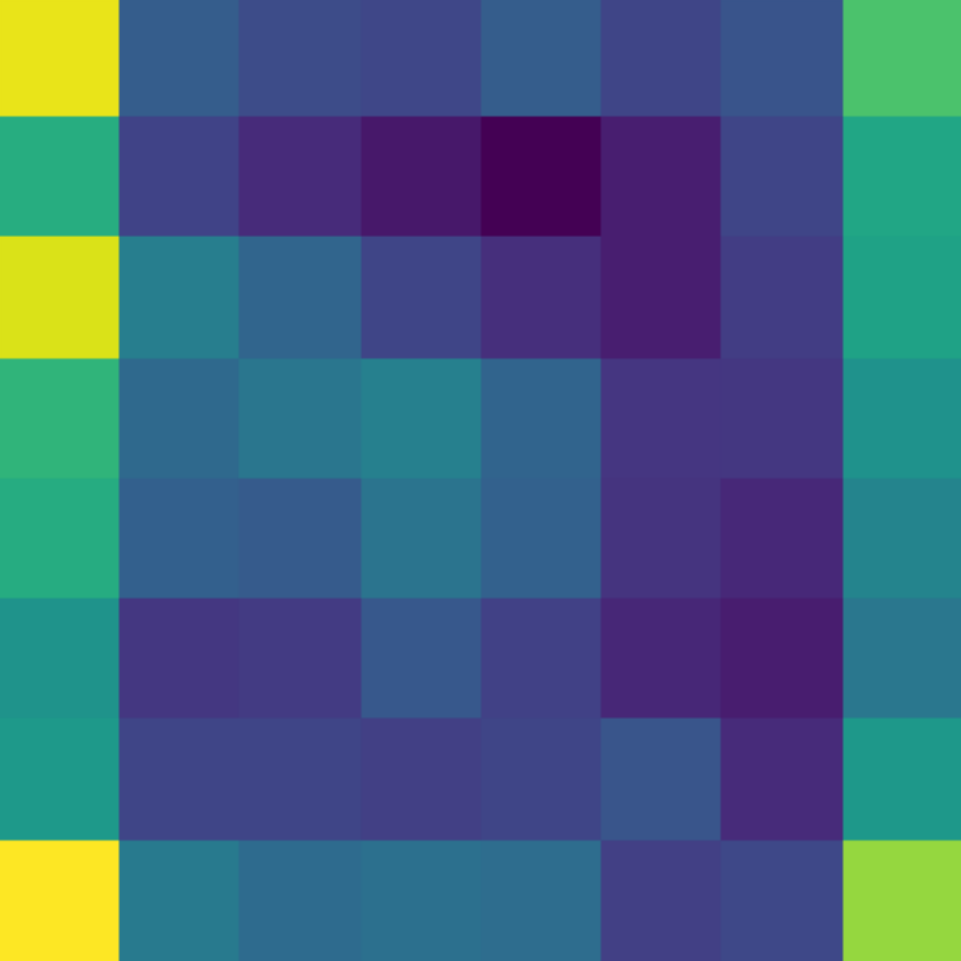}
  \caption*{Mid.}
\end{subfigure}
\begin{subfigure}{.05\textwidth}
  \centering
  \includegraphics[width=1.0\linewidth]{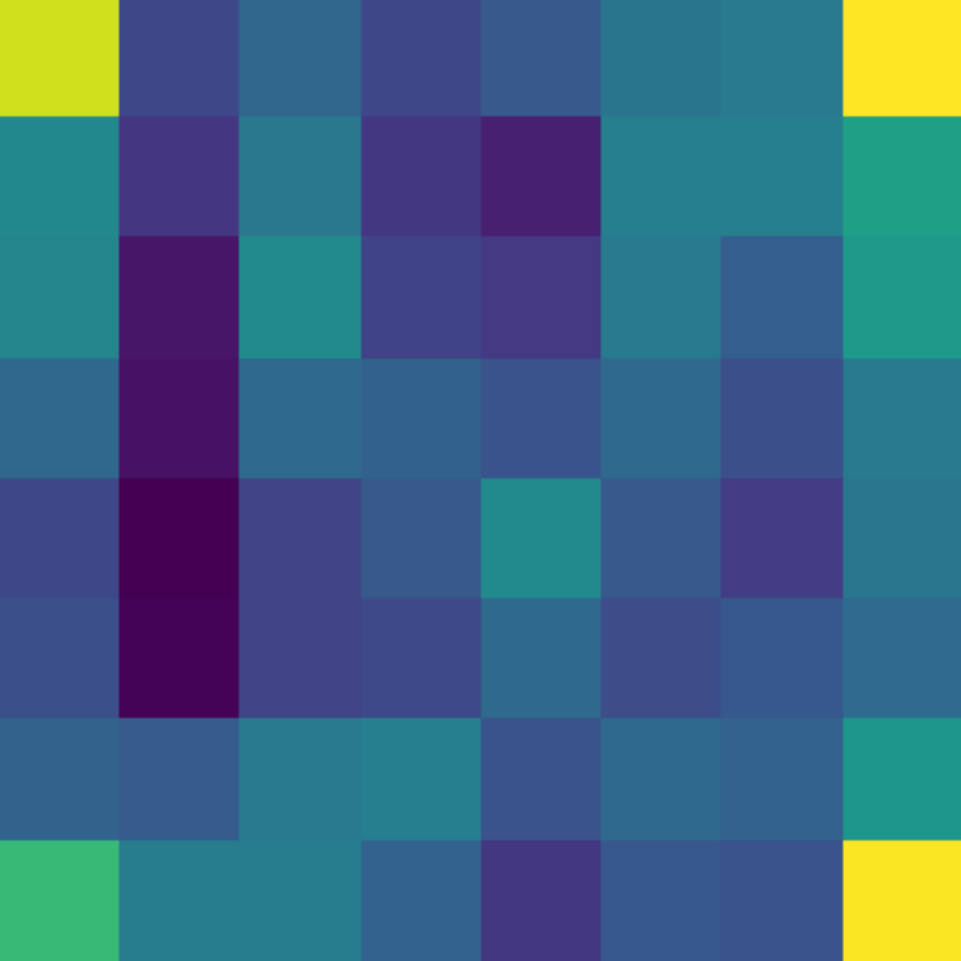}
  \caption*{Out.~0}
\end{subfigure}
\begin{subfigure}{.05\textwidth}
  \centering
  \includegraphics[width=1.0\linewidth]{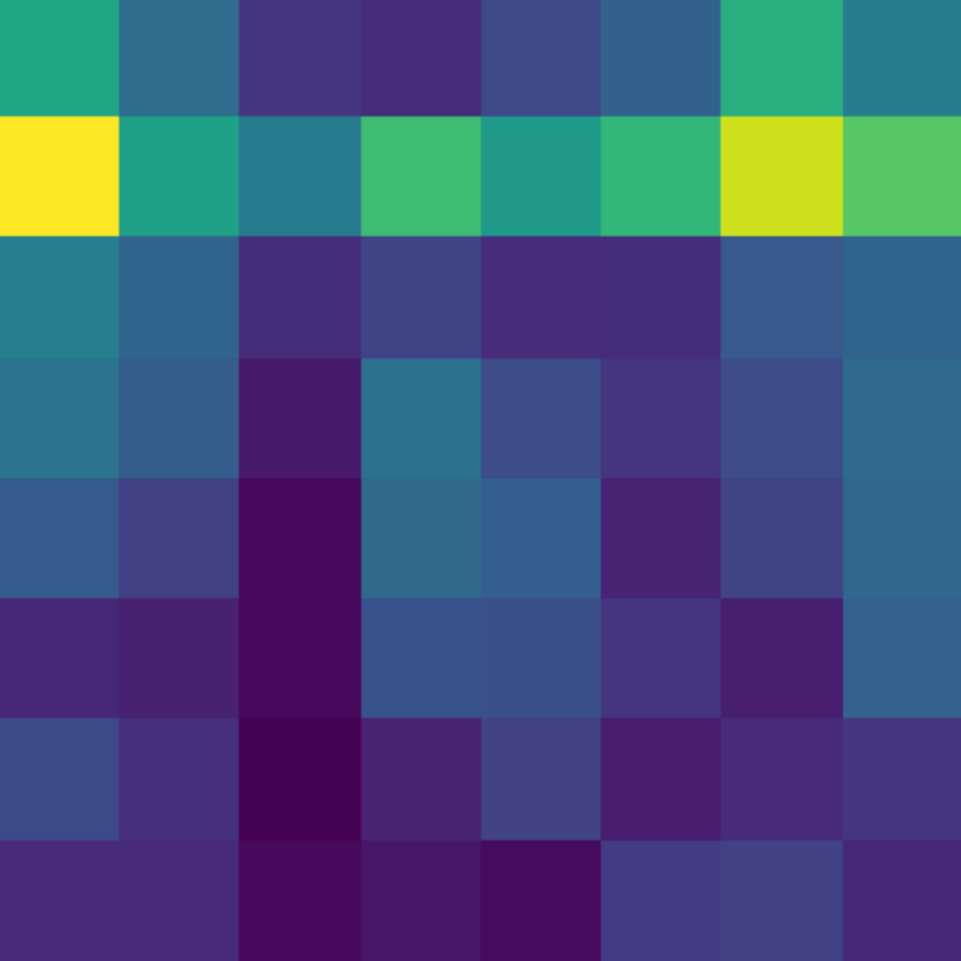}
  \caption*{Out.~1}
\end{subfigure}
\begin{subfigure}{.05\textwidth}
  \centering
  \includegraphics[width=1.0\linewidth]{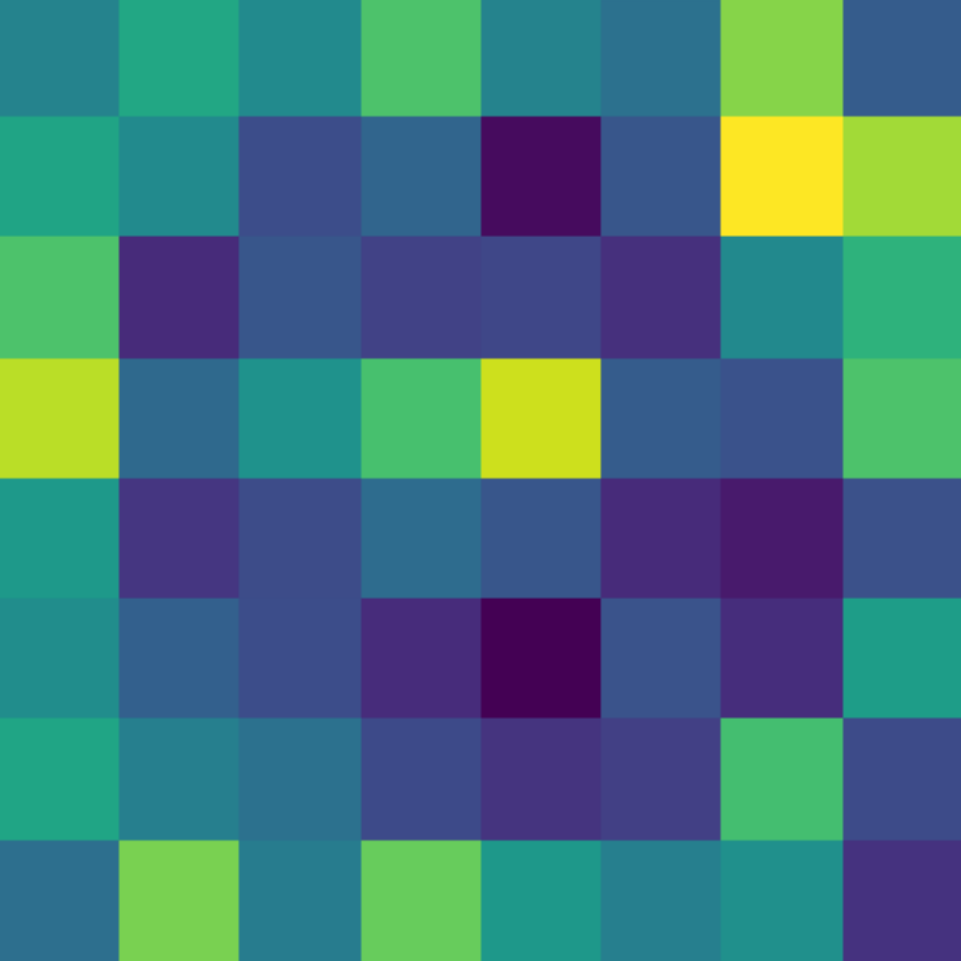}
  \caption*{Out.~2}
\end{subfigure}
\begin{subfigure}{.05\textwidth}
  \centering
  \includegraphics[width=1.0\linewidth]{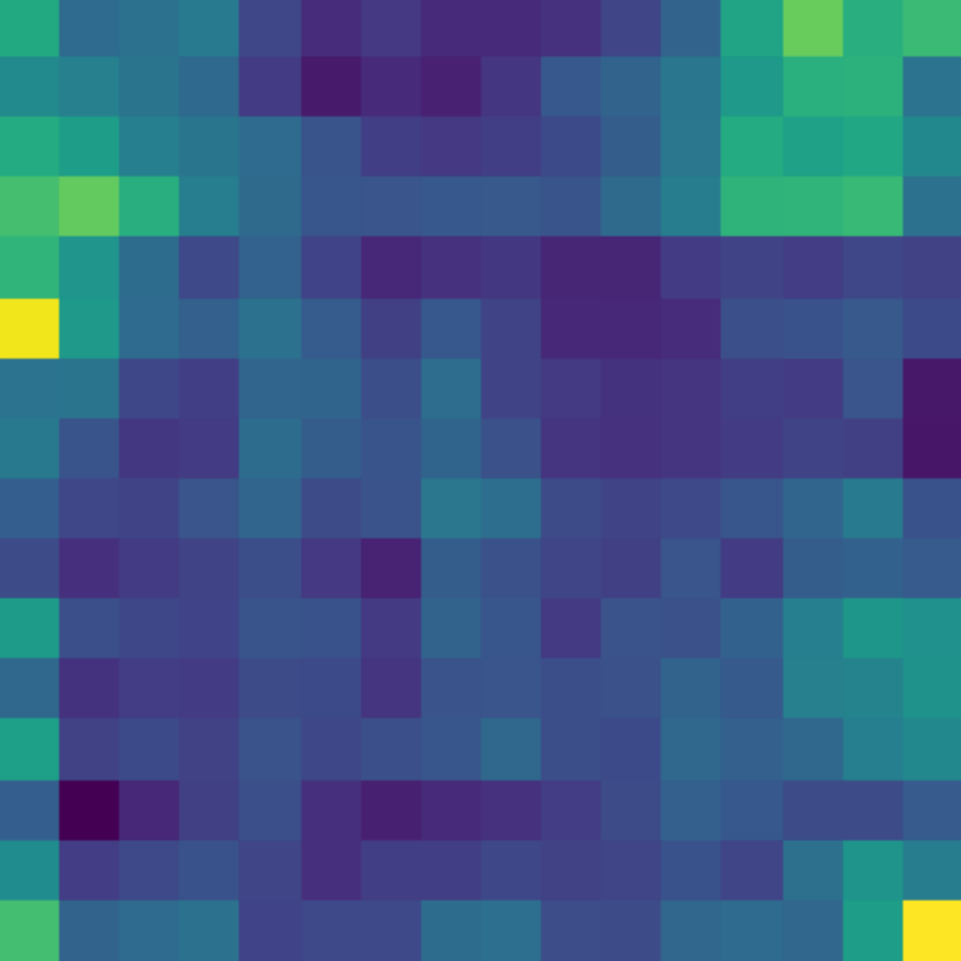}
  \caption*{Out.~3}
\end{subfigure}
\begin{subfigure}{.05\textwidth}
  \centering
  \includegraphics[width=1.0\linewidth]{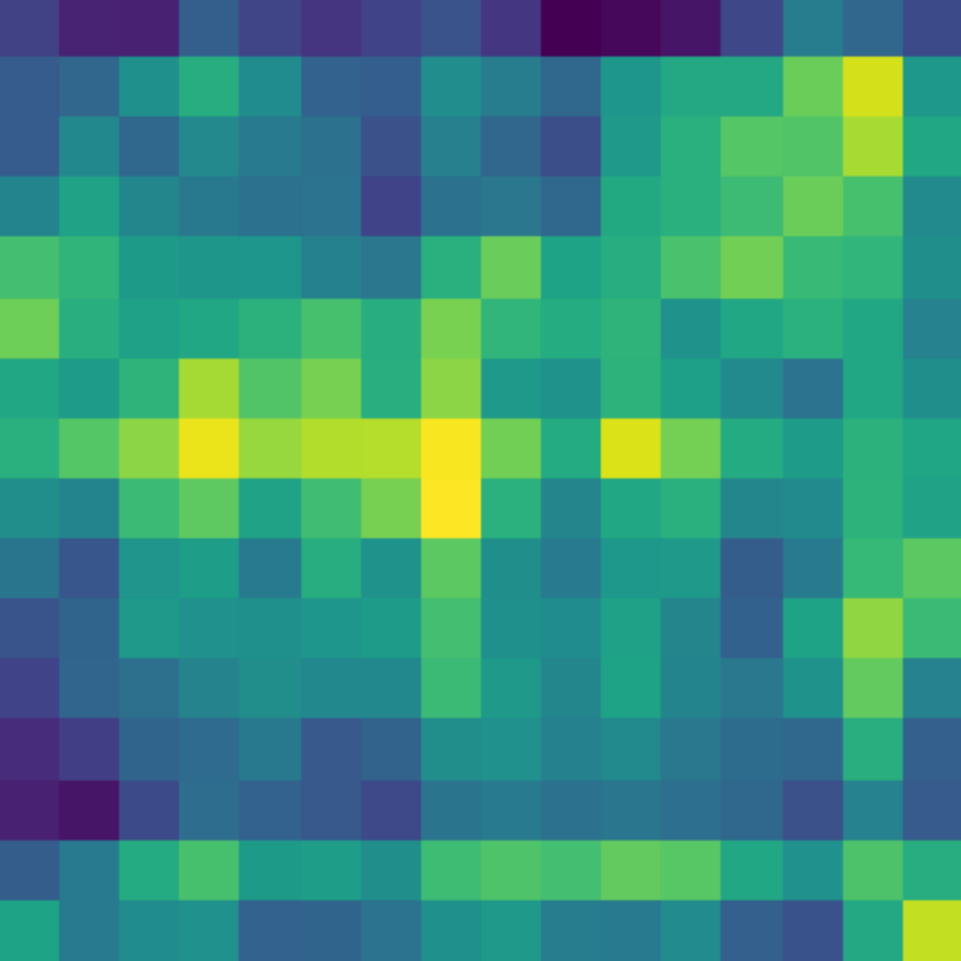}
  \caption*{Out.~4}
\end{subfigure}
\begin{subfigure}{.05\textwidth}
  \centering
  \includegraphics[width=1.0\linewidth]{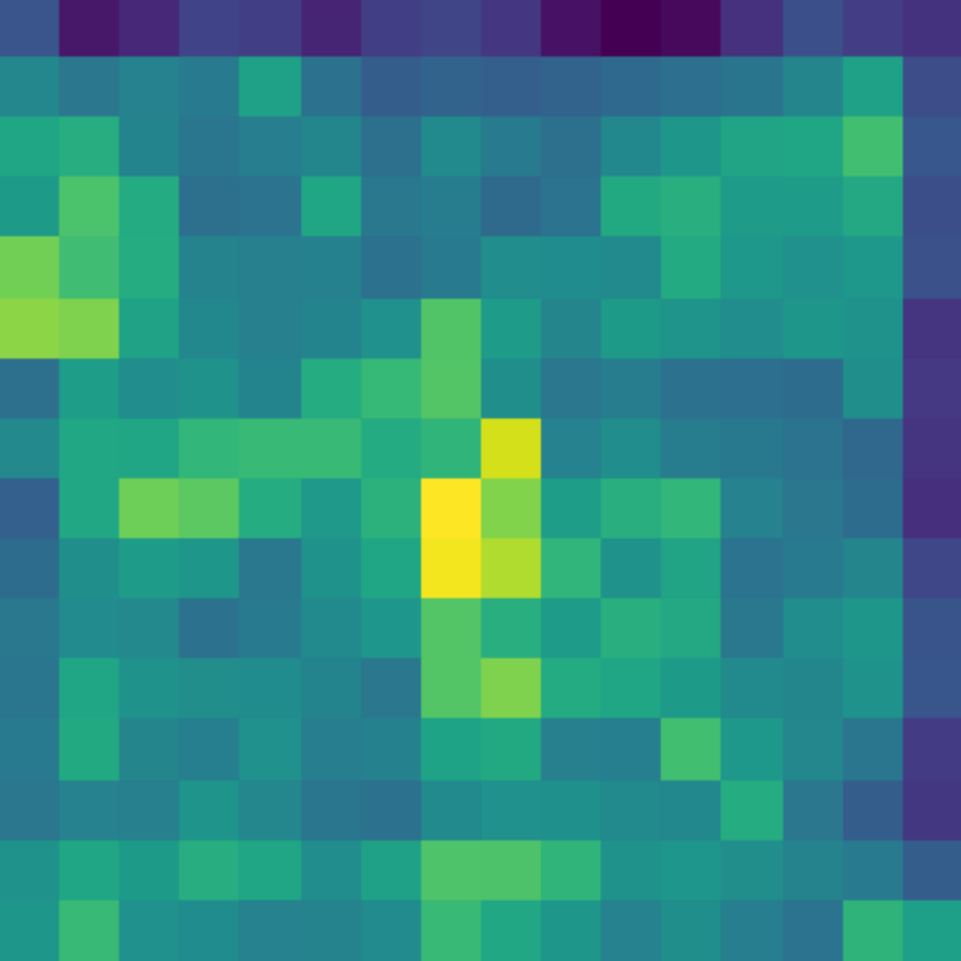}
  \caption*{Out.~5}
\end{subfigure}
\begin{subfigure}{.05\textwidth}
  \centering
  \includegraphics[width=1.0\linewidth]{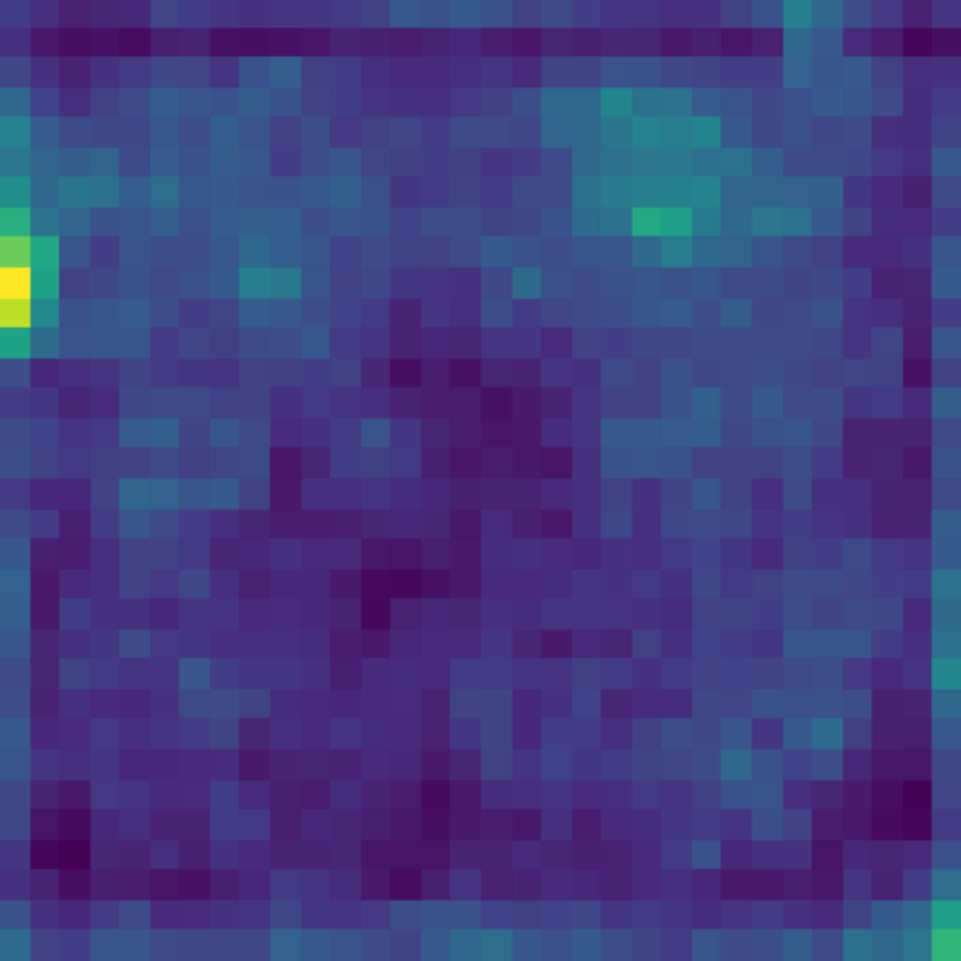}
  \caption*{Out.~6}
\end{subfigure}
\begin{subfigure}{.05\textwidth}
  \centering
  \includegraphics[width=1.0\linewidth]{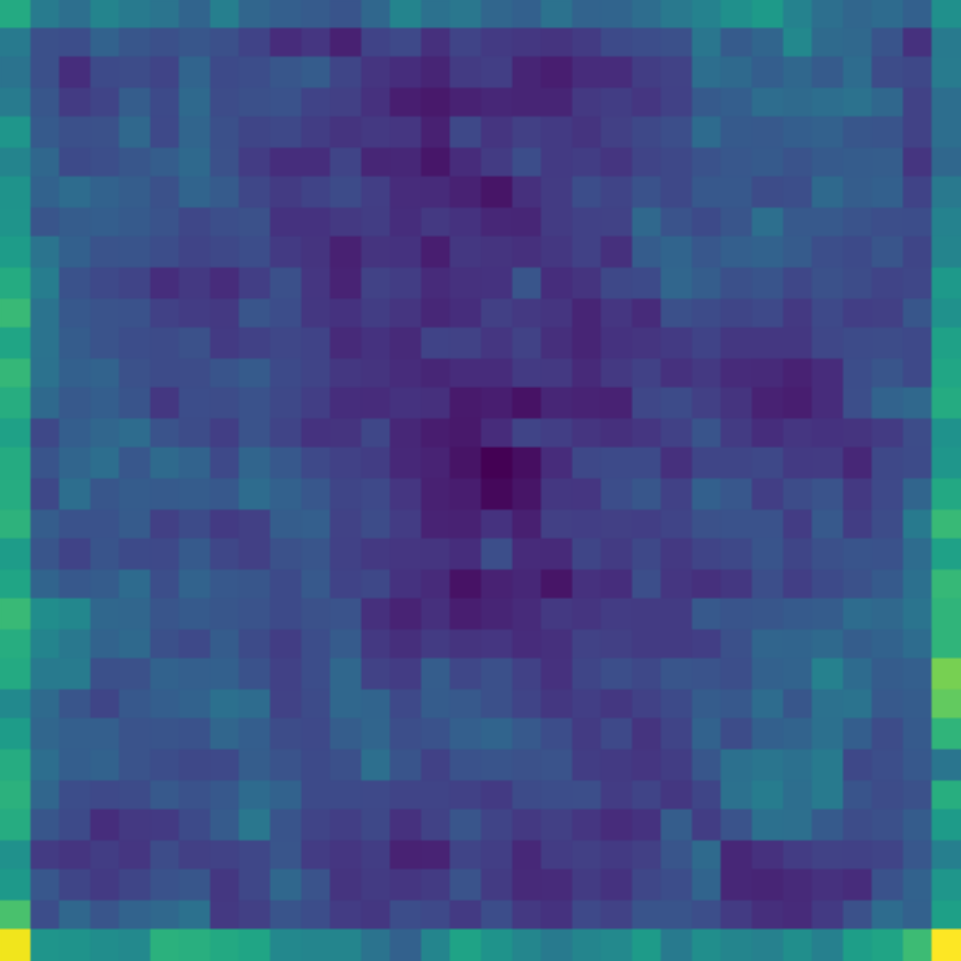}
  \caption*{Out.~7}
\end{subfigure}
\begin{subfigure}{.05\textwidth}
  \centering
  \includegraphics[width=1.0\linewidth]{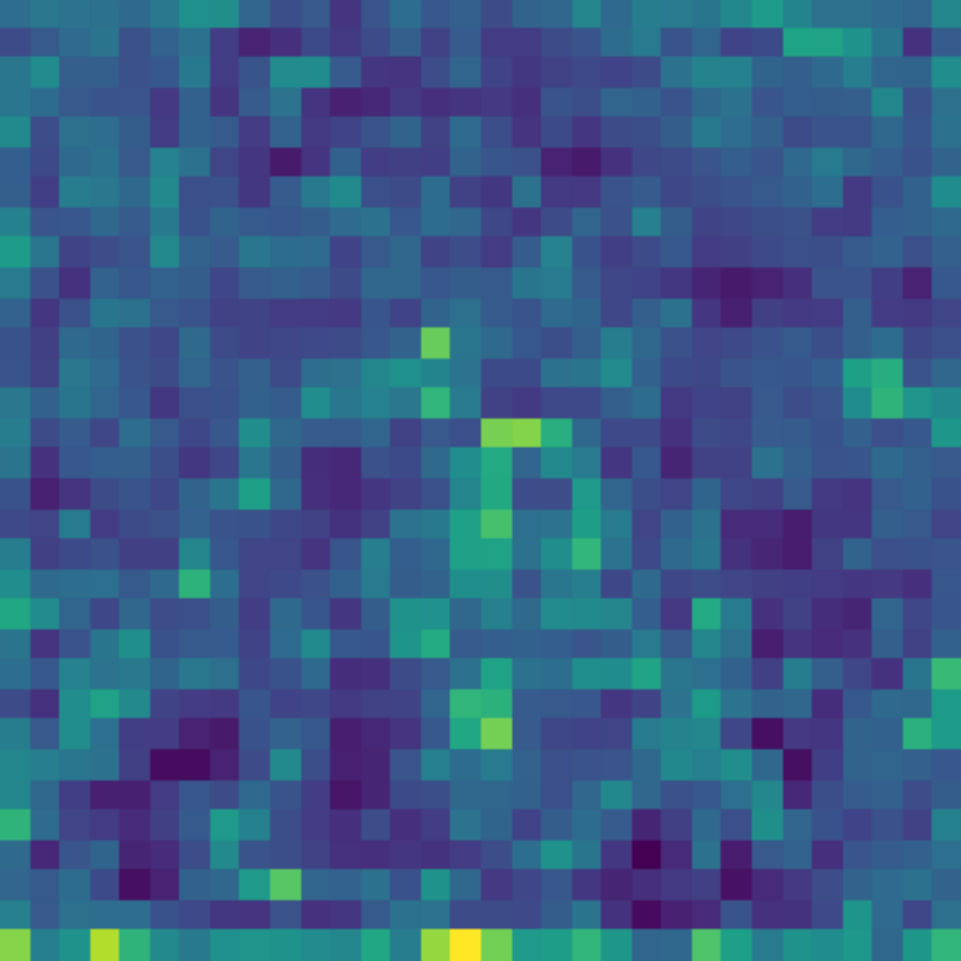}
  \caption*{Out.~8}
\end{subfigure}
\\
\end{tabular}
\vspace{-5pt}
\caption{\textbf{Attention maps at all the self-attention layers of ADM~\cite{dhariwal2021diffusion}.} In.~$n$, Mid., and Out.~$n$ denote the attention map of the $n$th block of the input blocks, the middle block, and the $n$th block of the output blocks, respectively.}
\label{fig:timestep}
\end{figure*}

\begin{figure*}[p]
\centering
\begin{tabular}{lc}
\centering
\begin{subfigure}{0.15\textwidth}
  \centering
  \includegraphics[width=1\textwidth]{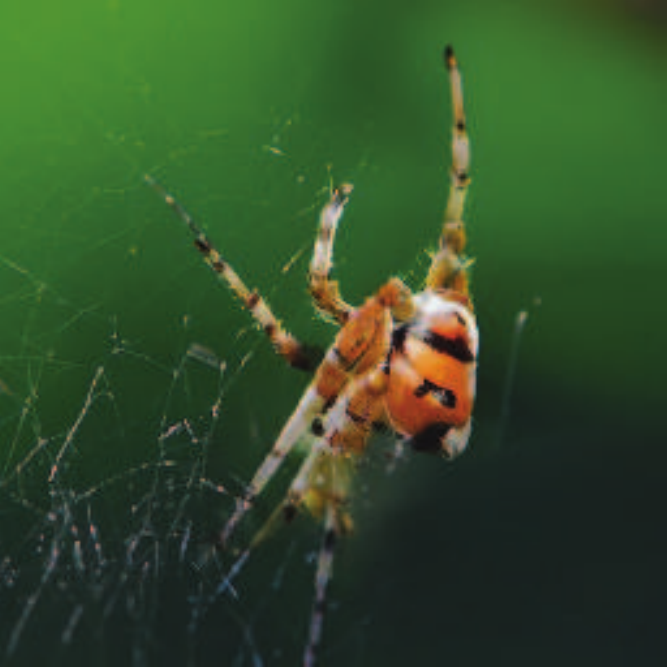}
\end{subfigure}
&
\begin{subfigure}{0.15\textwidth}
  \centering
  \includegraphics[width=1\linewidth]{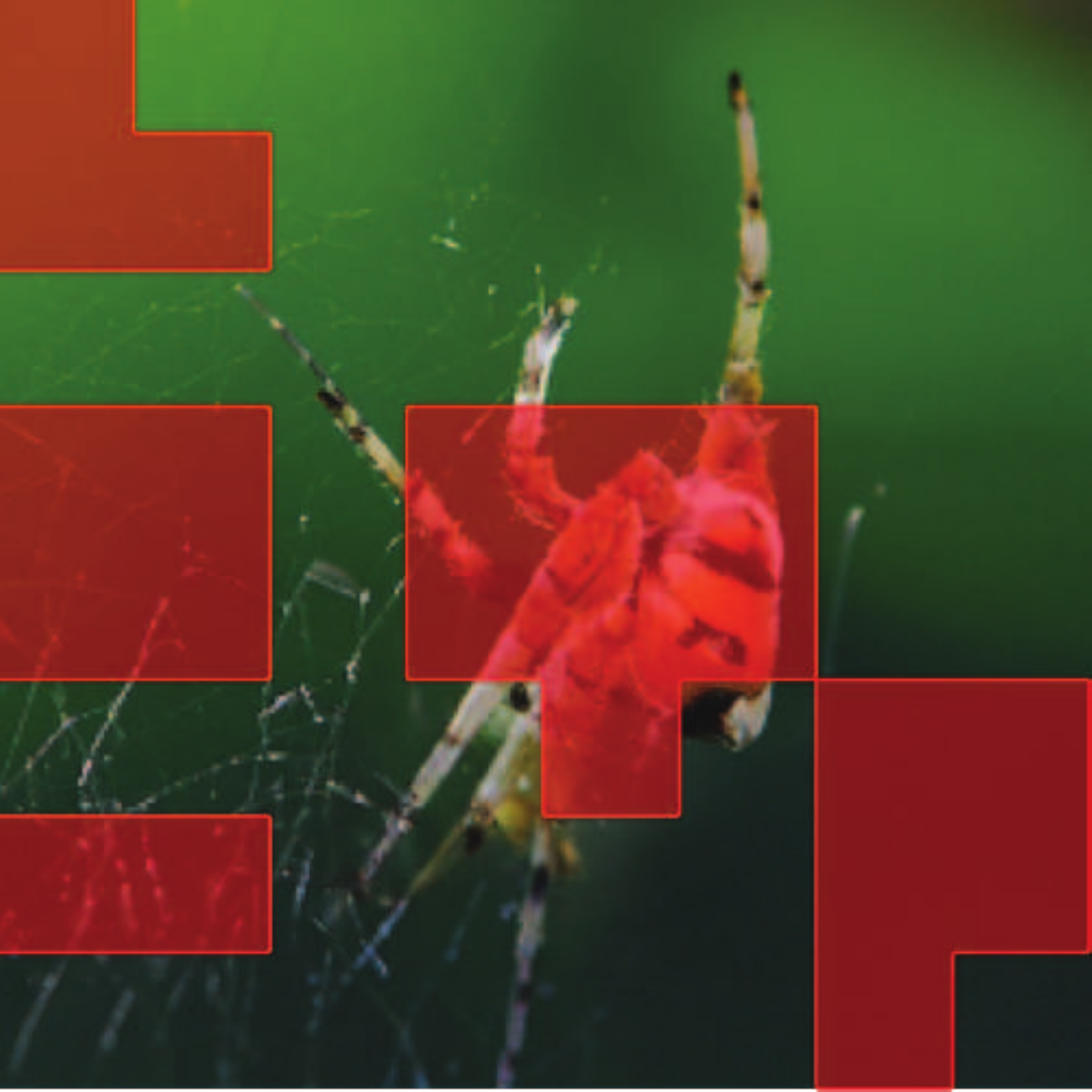}
\end{subfigure}
\begin{subfigure}{0.15\textwidth}
  \centering
  \includegraphics[width=1\linewidth]{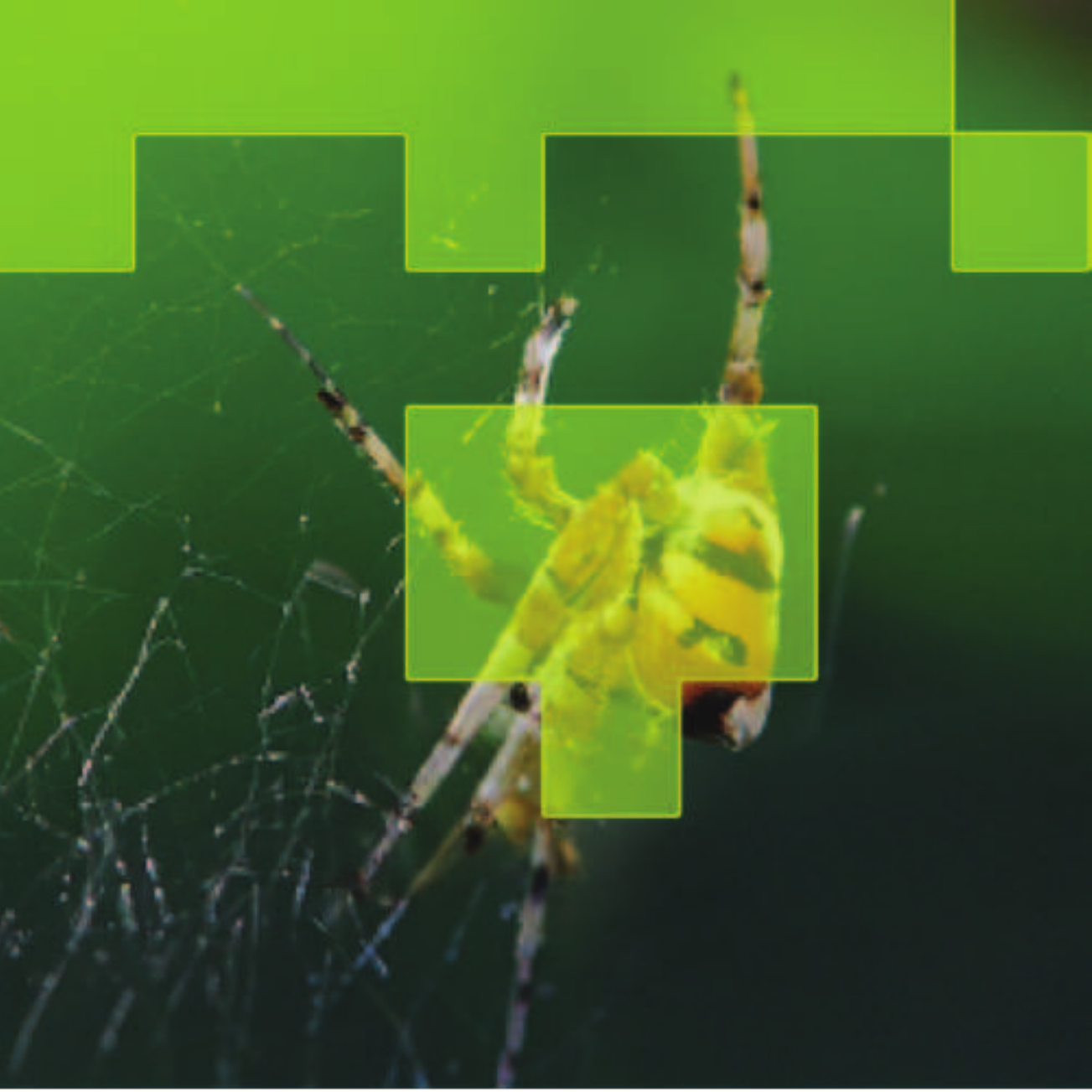}
\end{subfigure}
\begin{subfigure}{0.15\textwidth}
  \centering
  \includegraphics[width=1\linewidth]{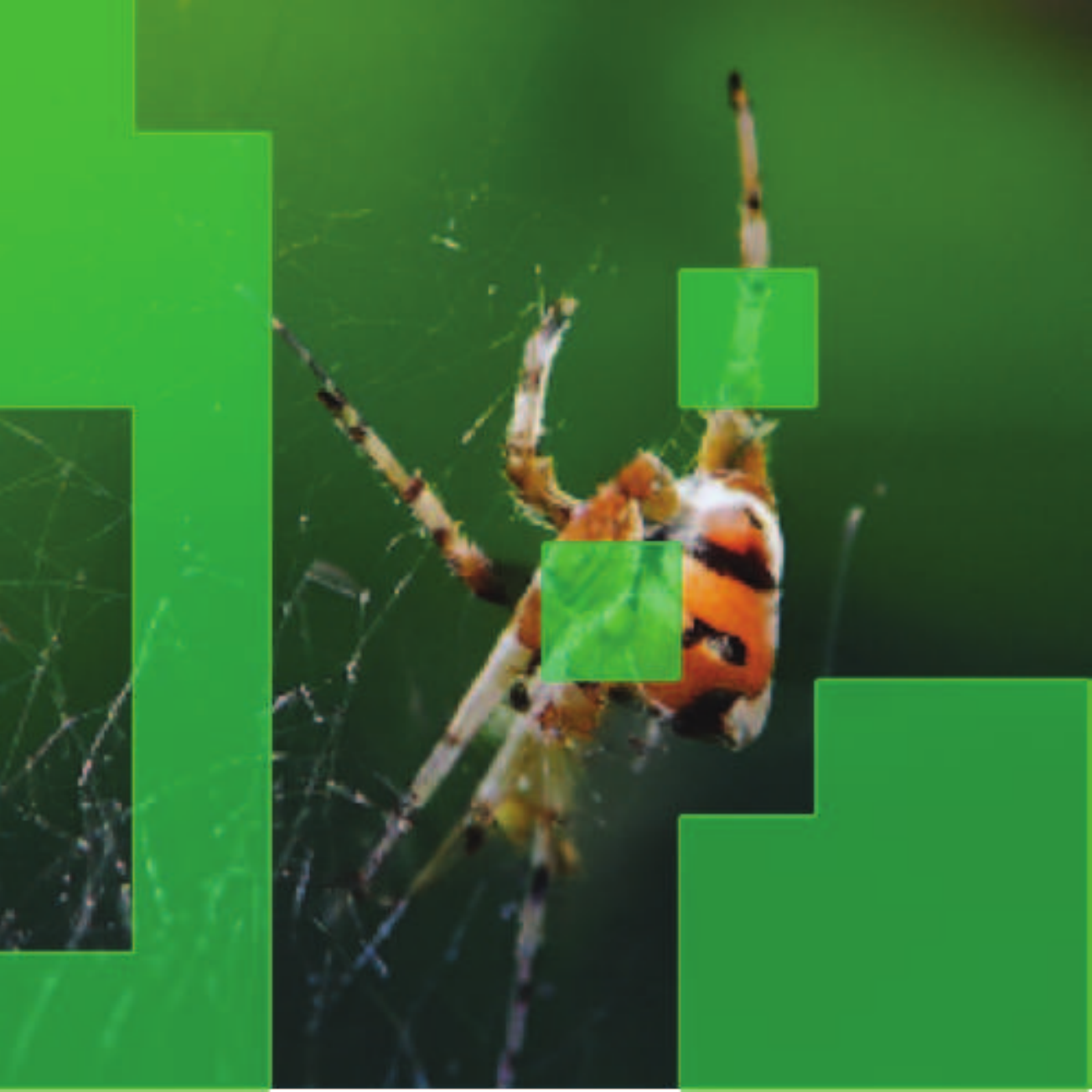}
\end{subfigure}
\begin{subfigure}{0.15\textwidth}
  \centering
  \includegraphics[width=1\linewidth]{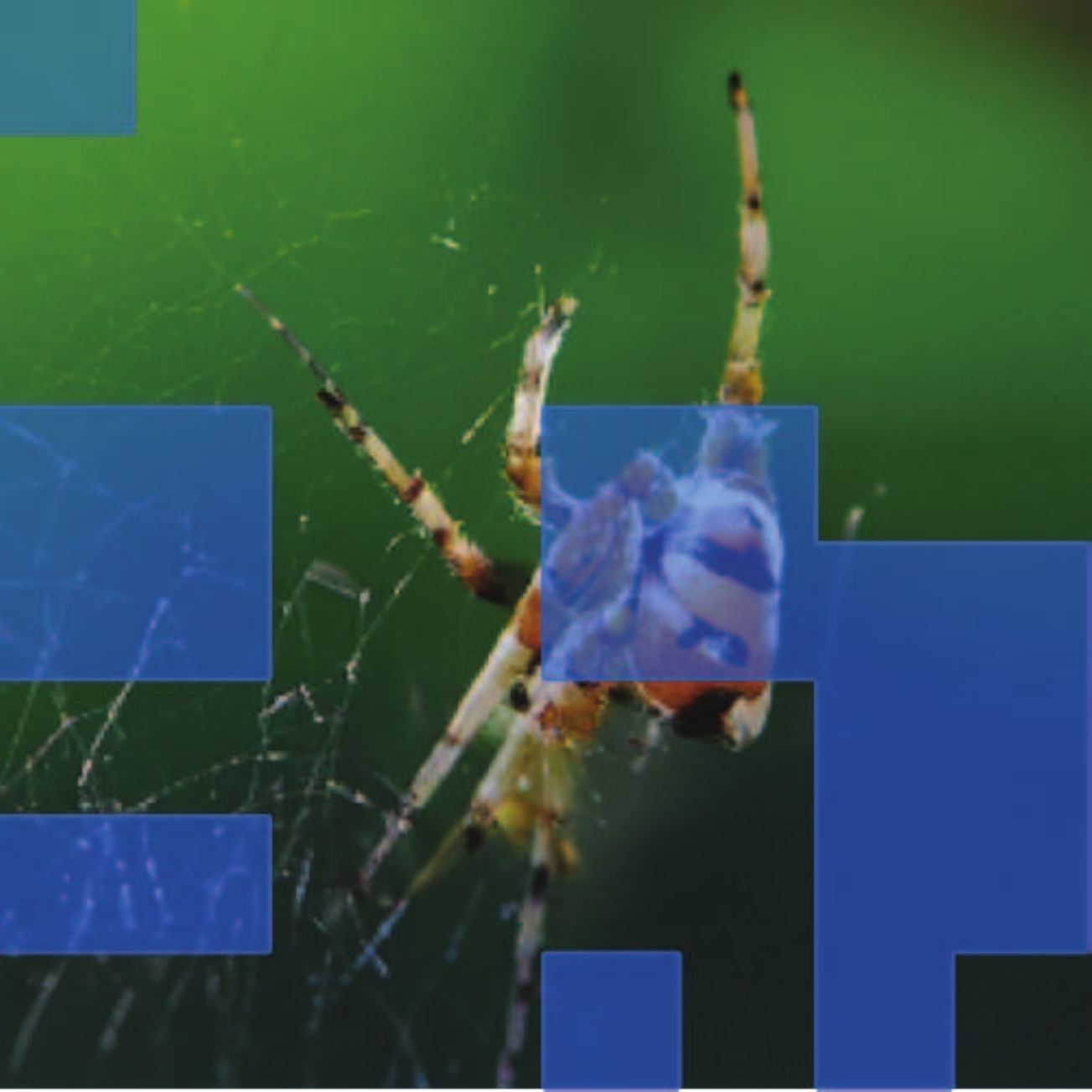}
\end{subfigure}
\begin{subfigure}{0.15\textwidth}
  \centering
  \includegraphics[width=1\linewidth]{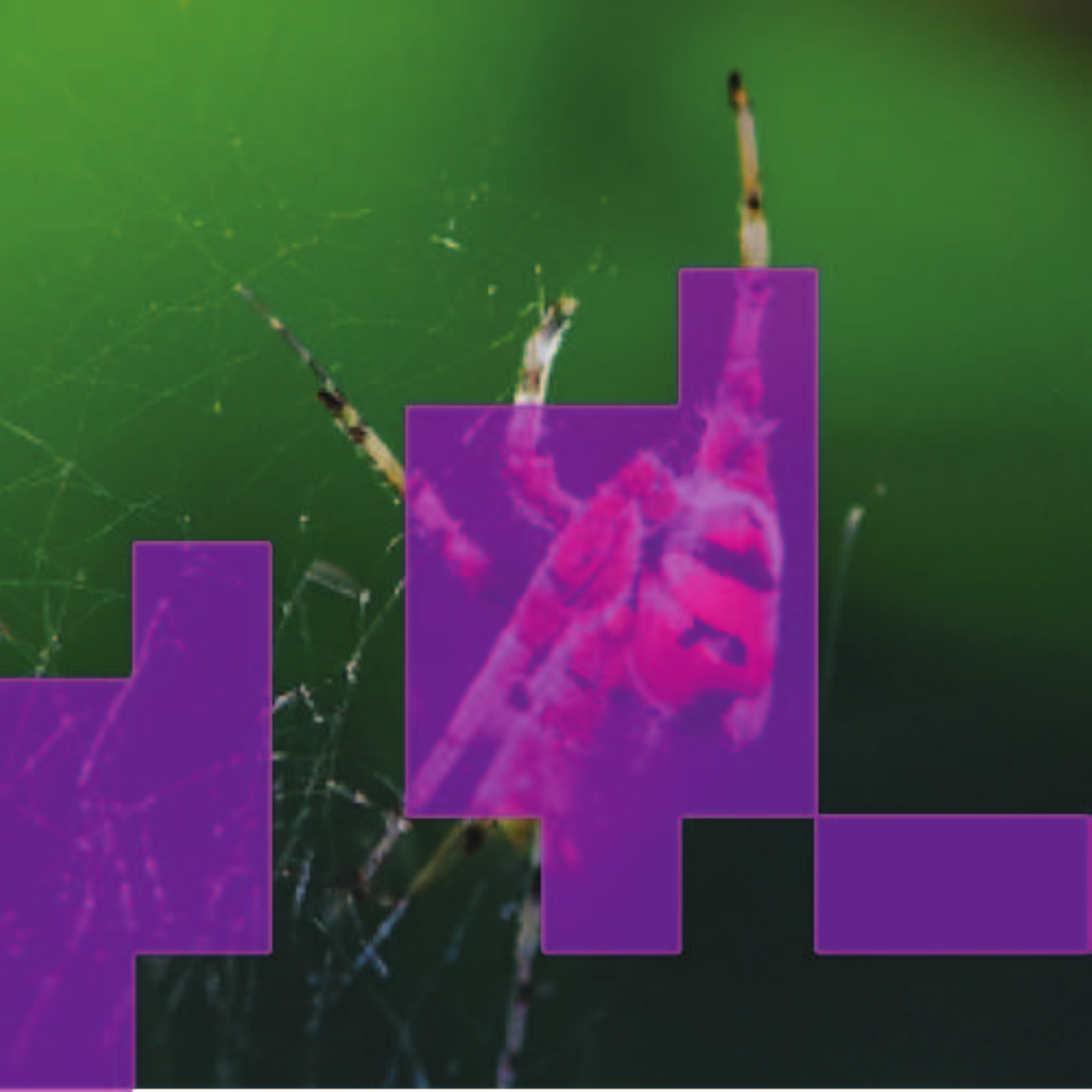}
\end{subfigure}\\
&
\begin{subfigure}{0.15\textwidth}
  \centering
  \includegraphics[width=1\linewidth]{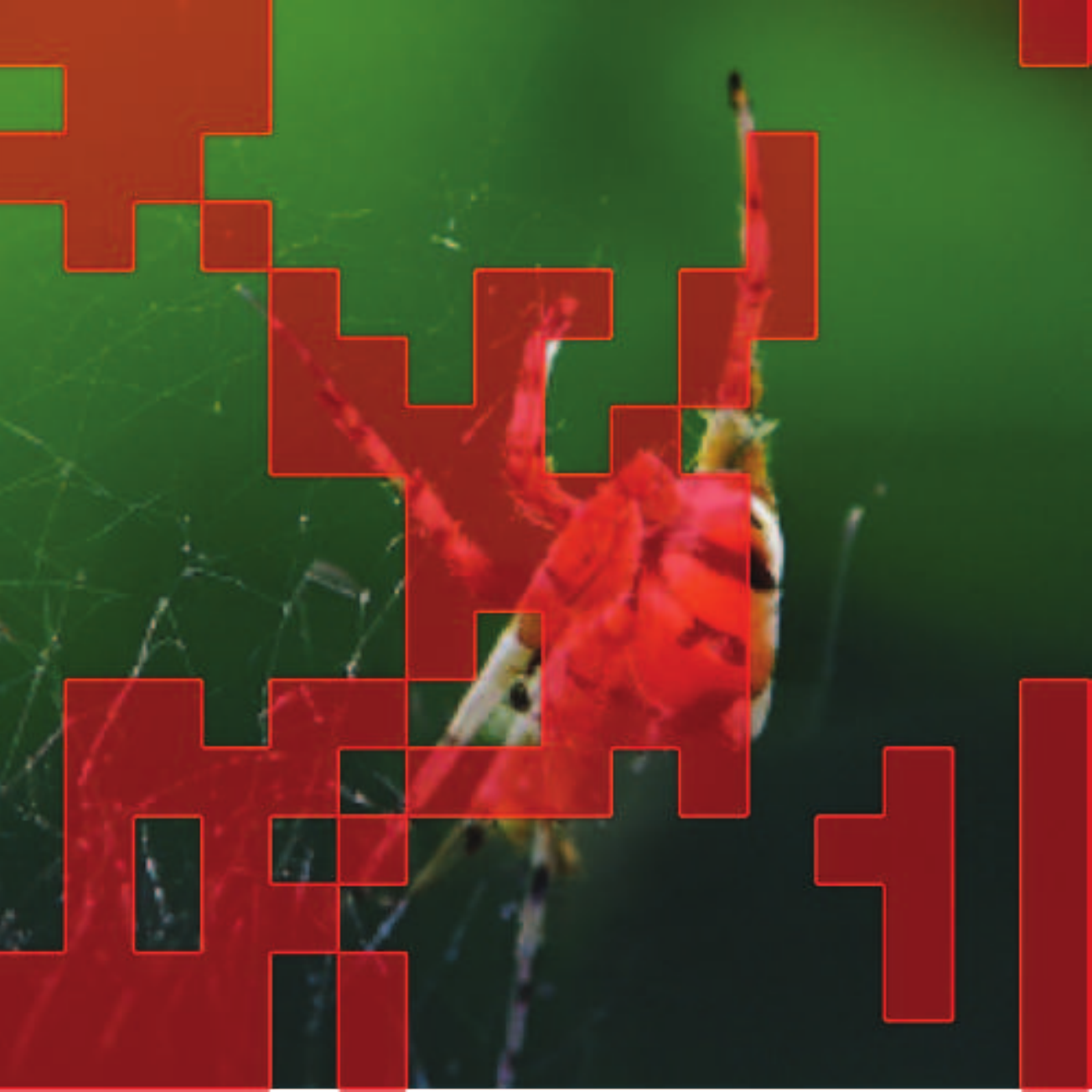}
\end{subfigure}
\begin{subfigure}{0.15\textwidth}
  \centering
  \includegraphics[width=1\linewidth]{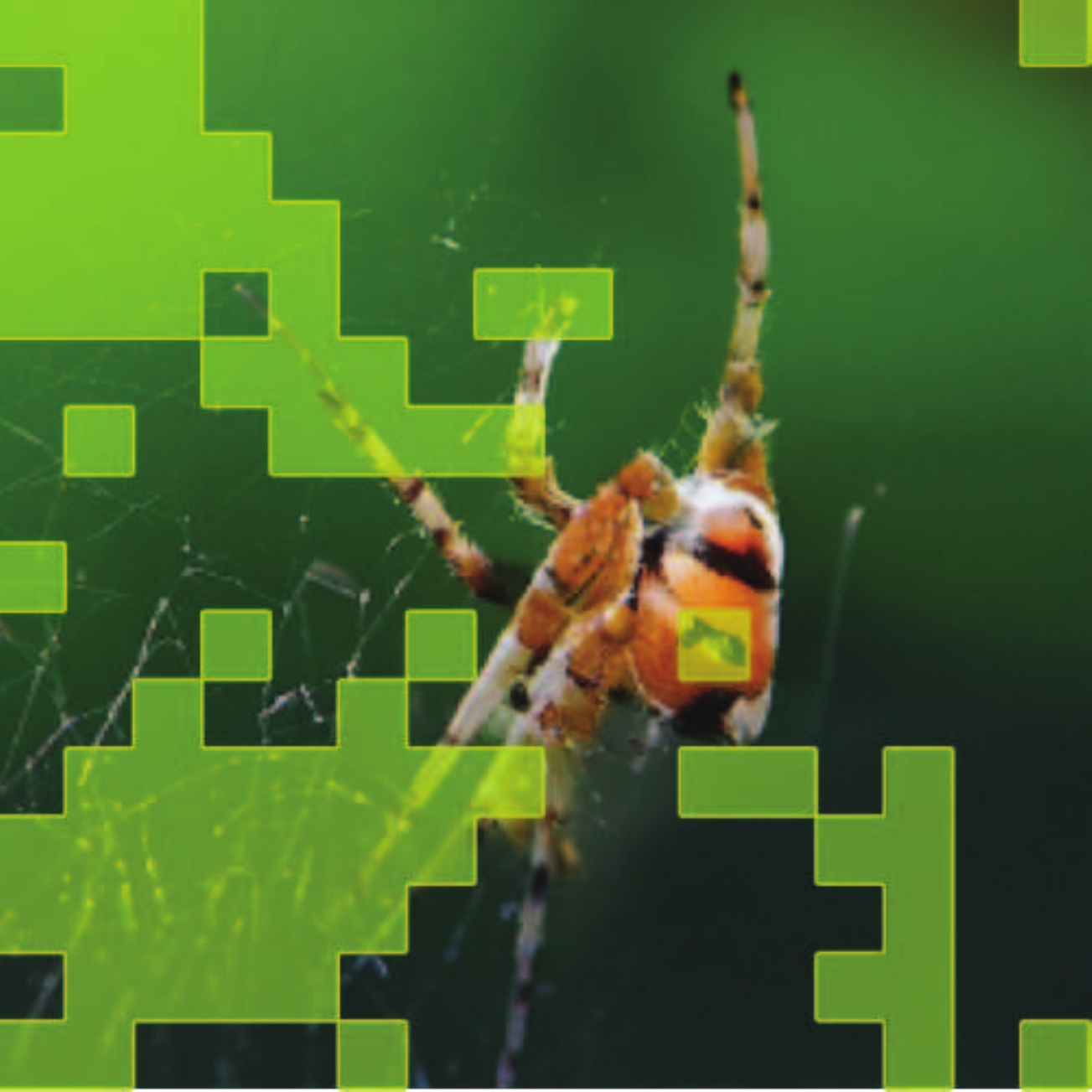}
\end{subfigure}
\begin{subfigure}{0.15\textwidth}
  \centering
  \includegraphics[width=1\linewidth]{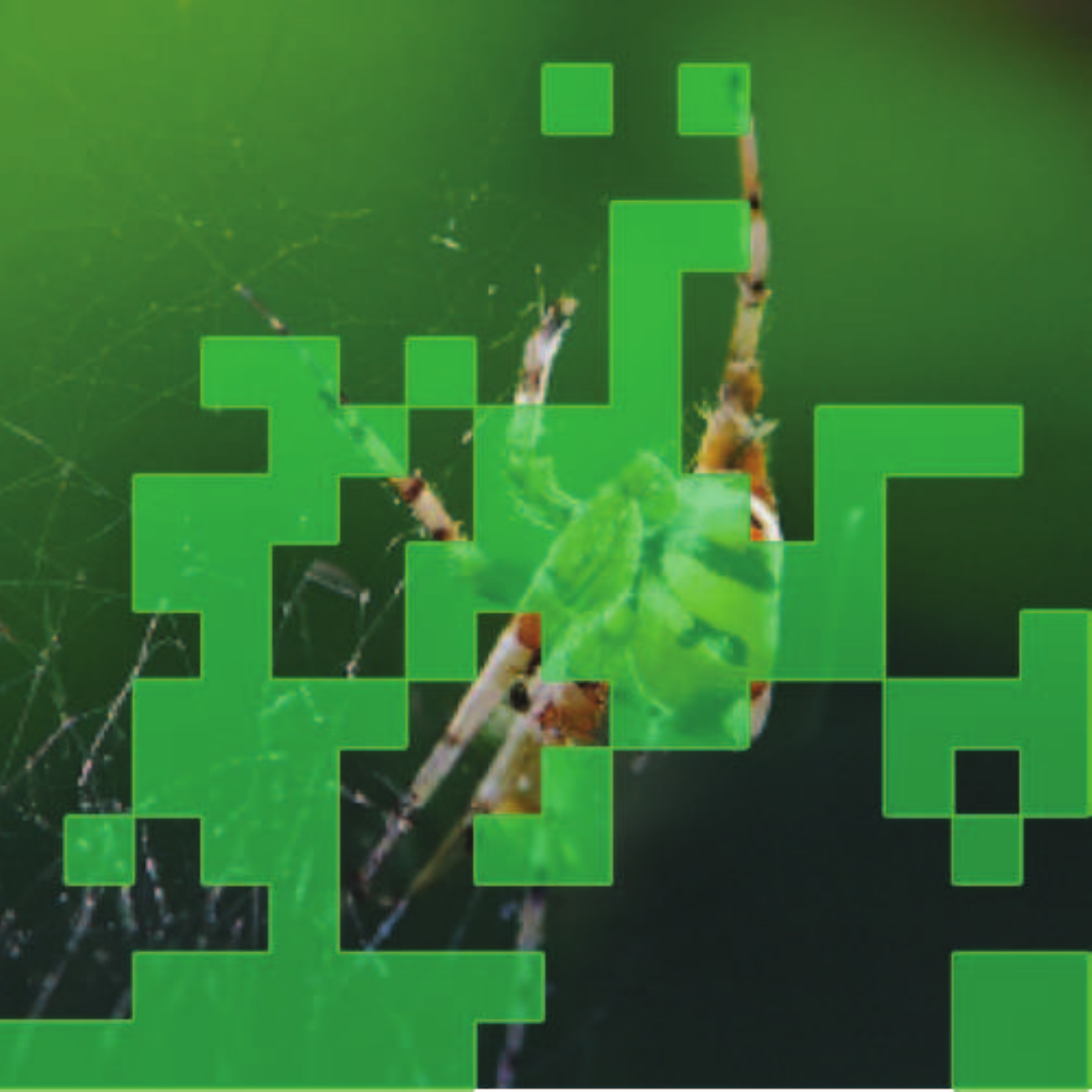}
\end{subfigure}
\begin{subfigure}{0.15\textwidth}
  \centering
  \includegraphics[width=1\linewidth]{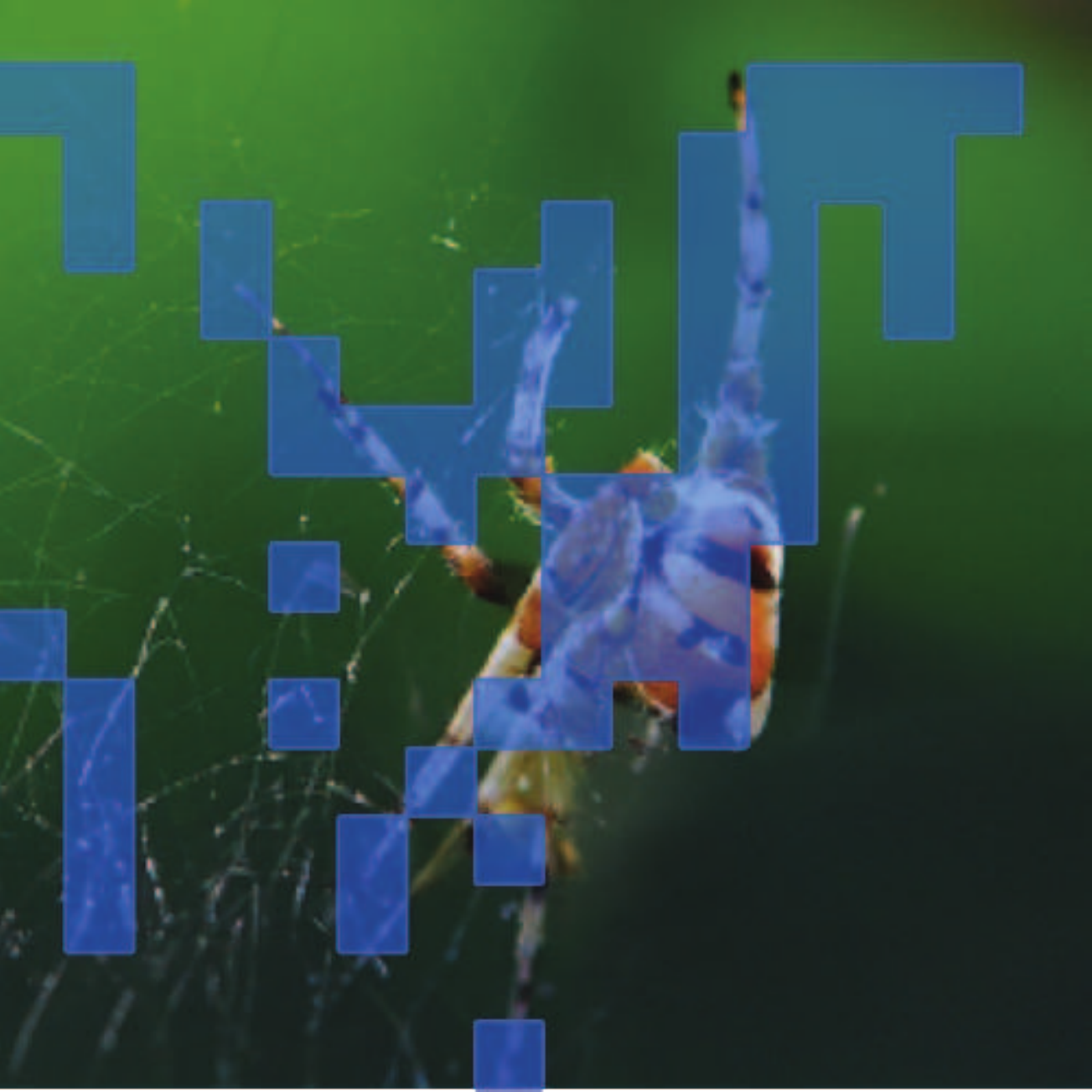}
\end{subfigure}
\begin{subfigure}{0.15\textwidth}
  \centering
  \includegraphics[width=1\linewidth]{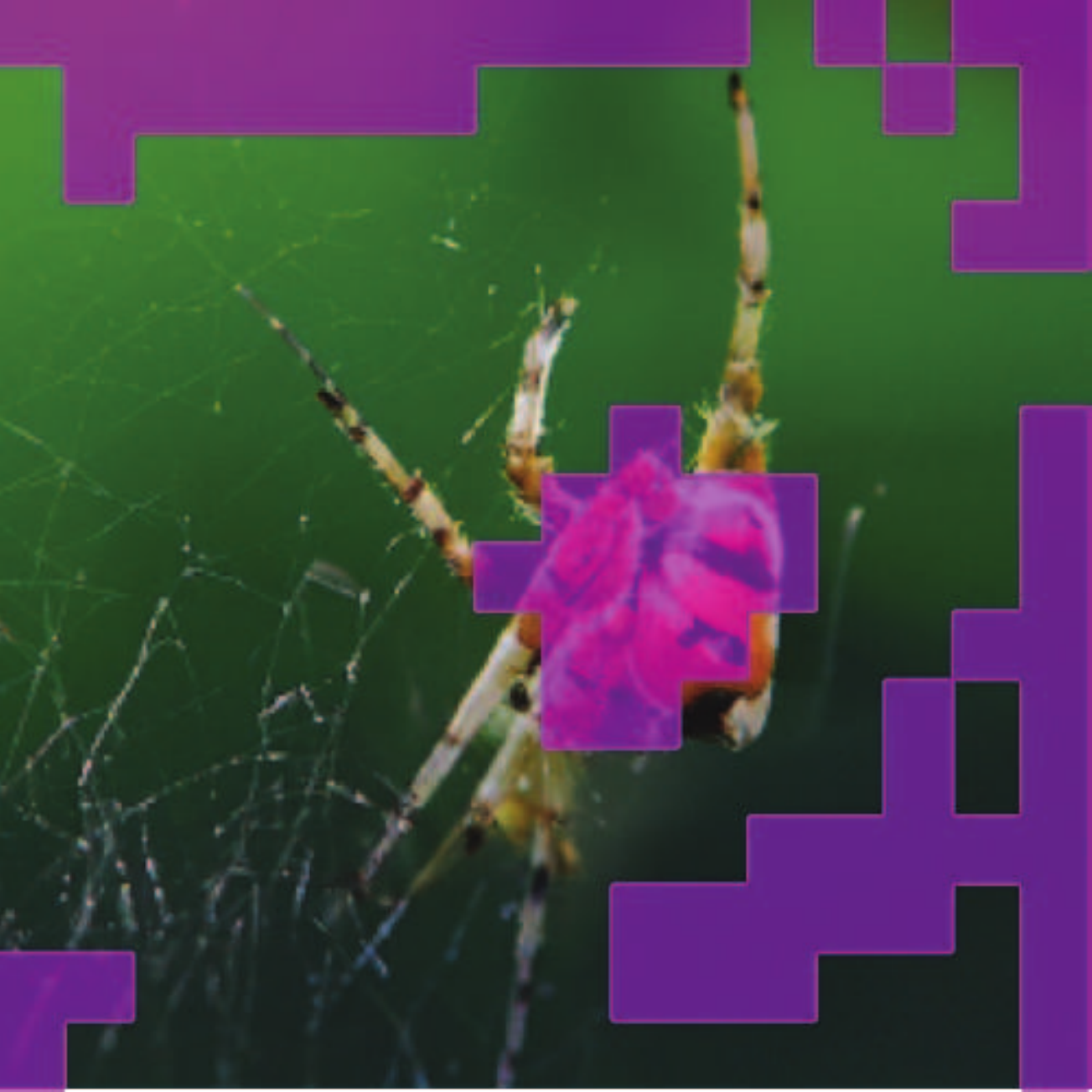}
\end{subfigure}\\
&
\begin{subfigure}{0.15\textwidth}
  \centering
  \includegraphics[width=1\linewidth]{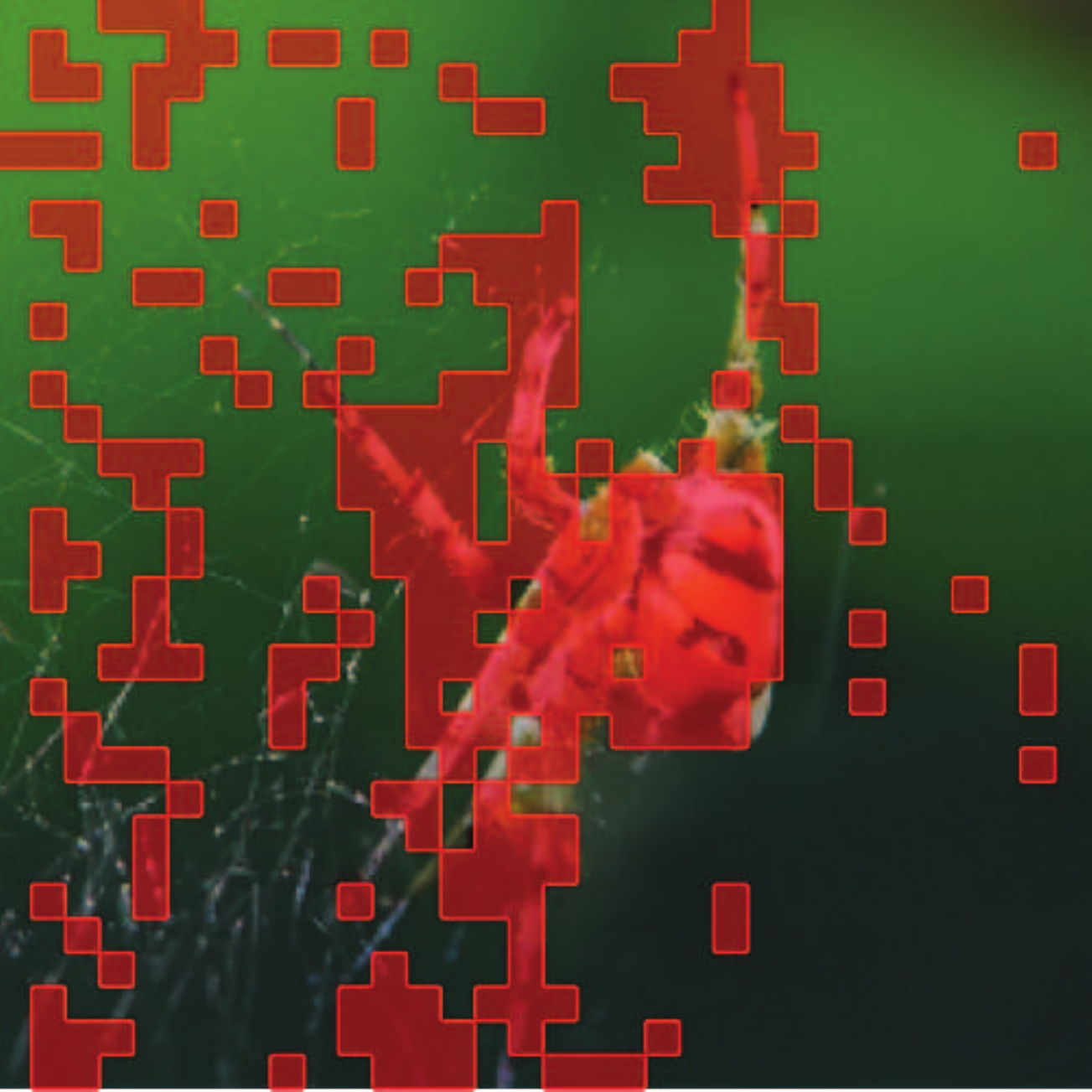}
  \caption*{Averaged}
\end{subfigure}
\begin{subfigure}{0.15\textwidth}
  \centering
  \includegraphics[width=1\linewidth]{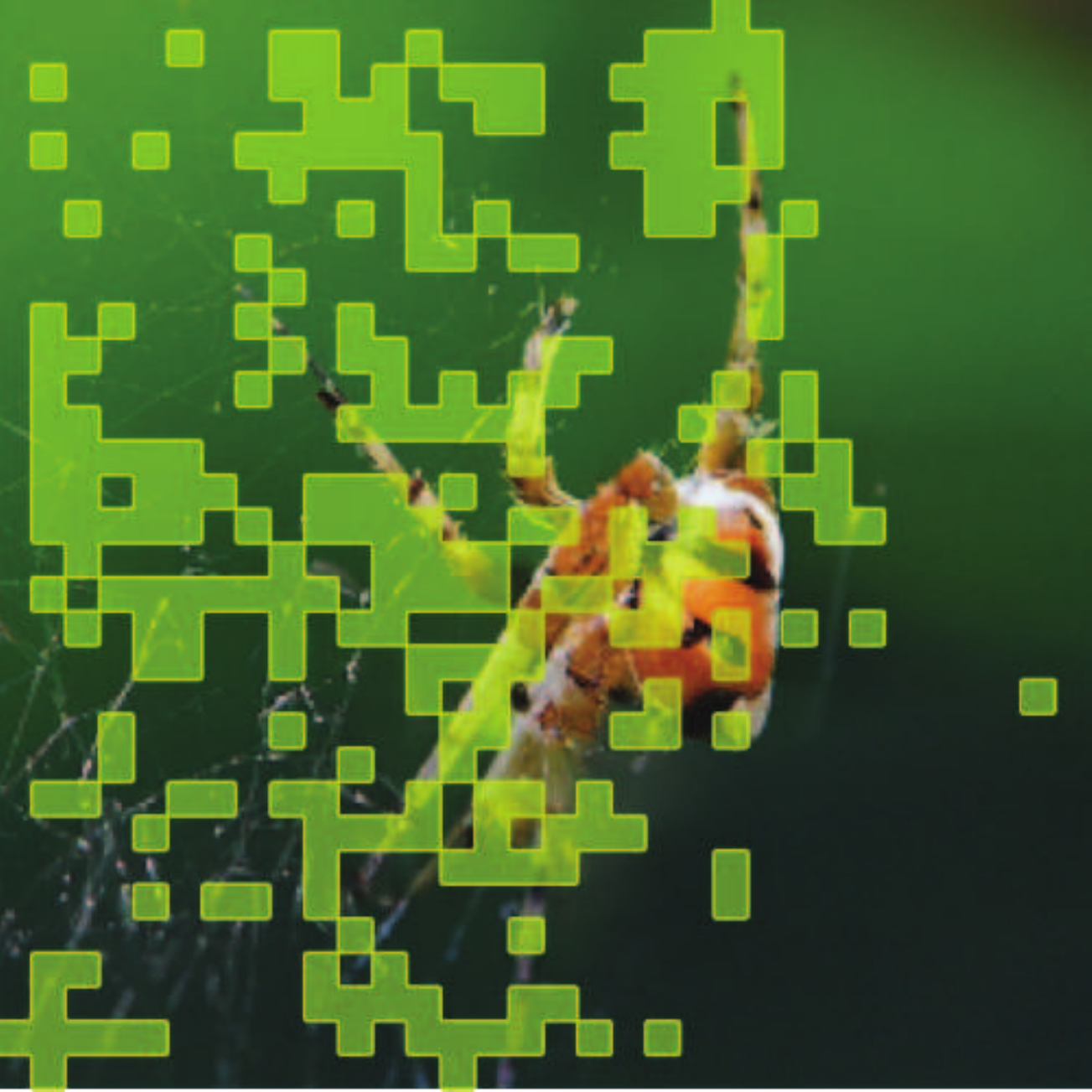}
  \caption*{Head 1}
\end{subfigure}
\begin{subfigure}{0.15\textwidth}
  \centering
  \includegraphics[width=1\linewidth]{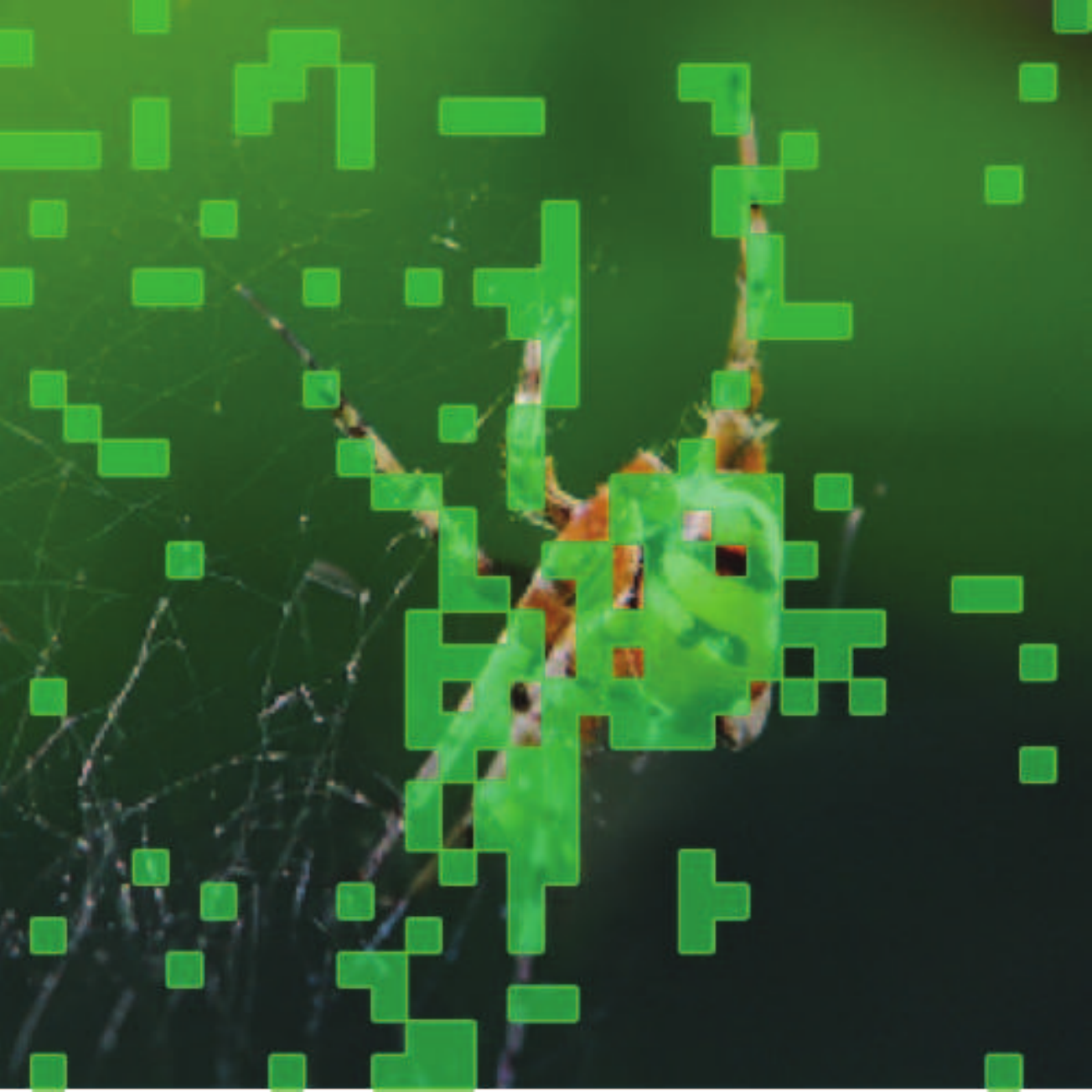}
  \caption*{Head 2}
\end{subfigure}
\begin{subfigure}{0.15\textwidth}
  \centering
  \includegraphics[width=1\linewidth]{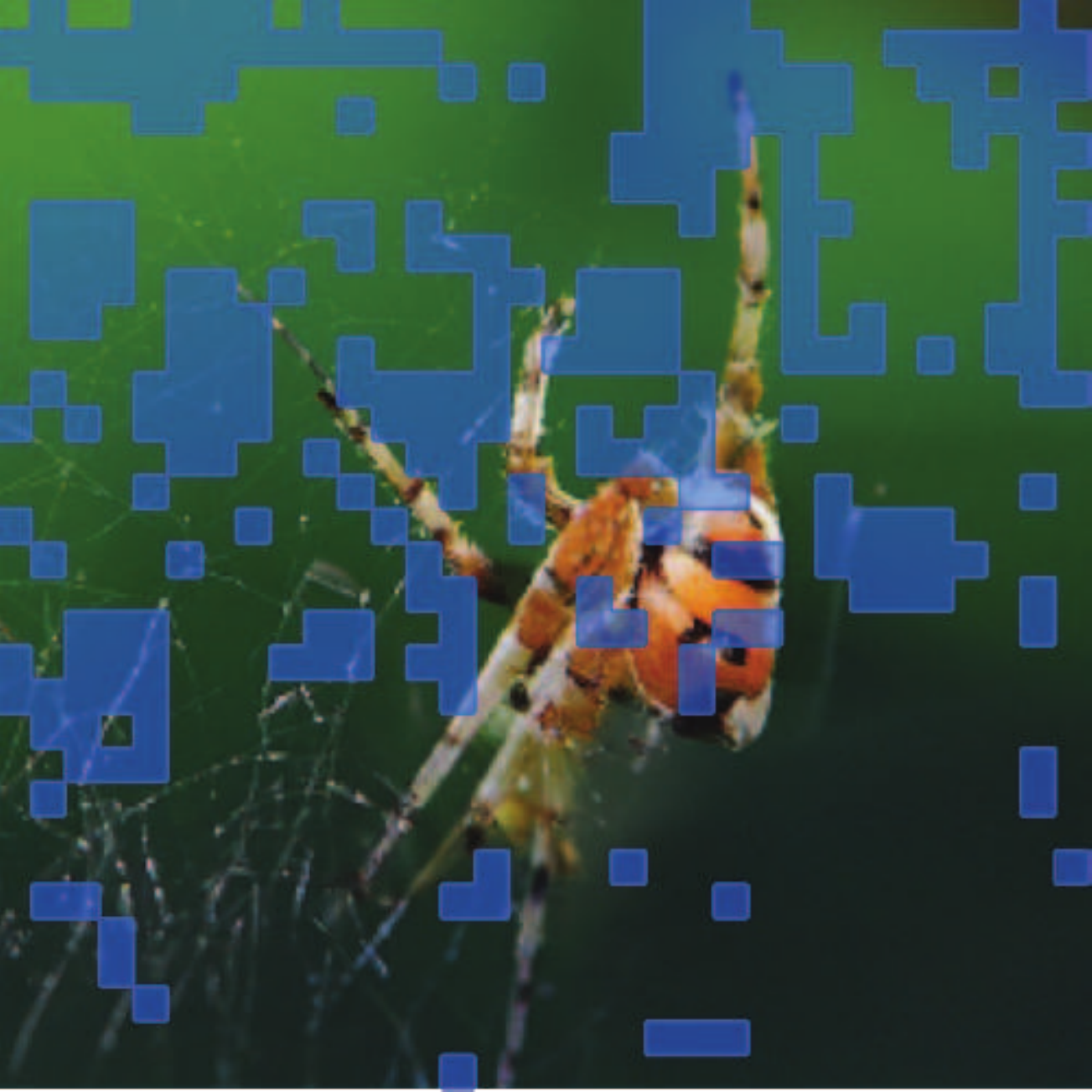}
  \caption*{Head 3}
\end{subfigure}
\begin{subfigure}{0.15\textwidth}
  \centering
  \includegraphics[width=1\linewidth]{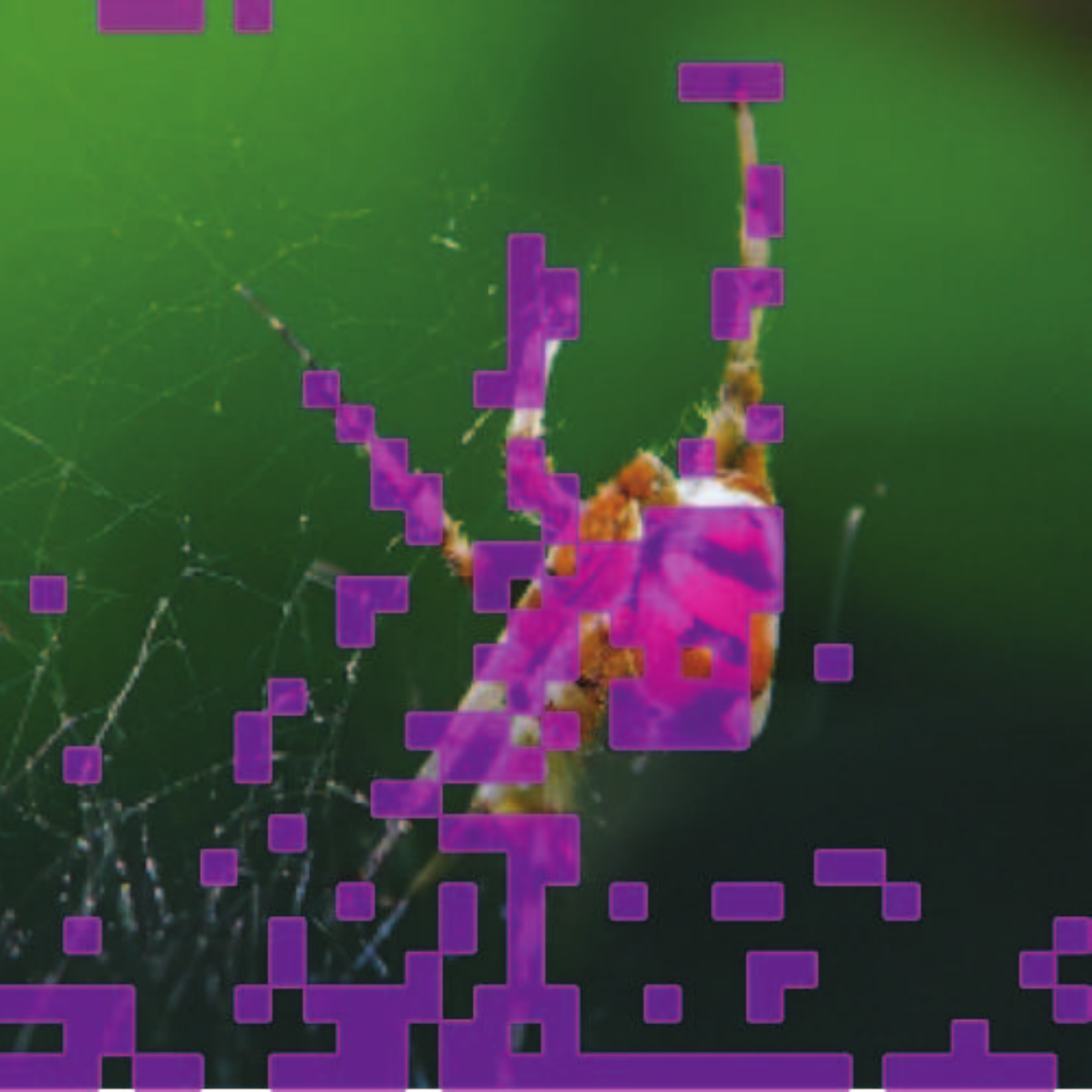}
  \caption*{Head 4}
\end{subfigure}
\\
\\
\begin{subfigure}{0.15\textwidth}
  \centering
  \includegraphics[width=1\textwidth]{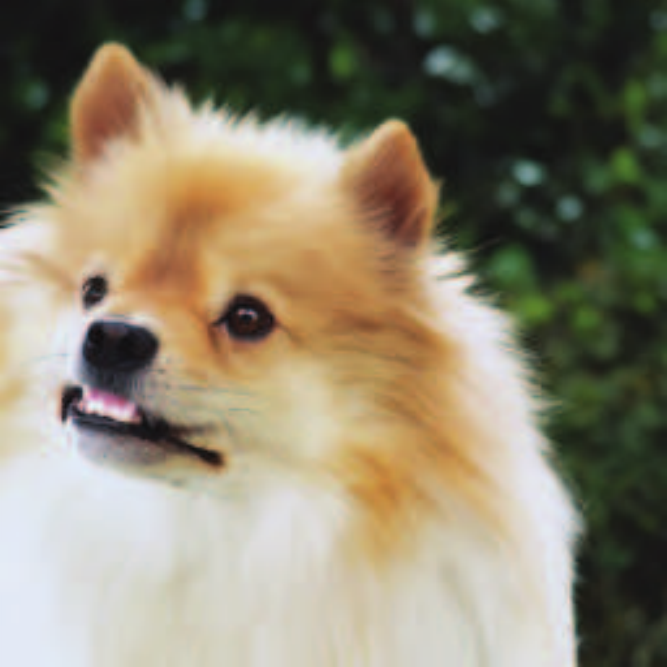}
\end{subfigure}
&
\begin{subfigure}{0.15\textwidth}
  \centering
  \includegraphics[width=1\linewidth]{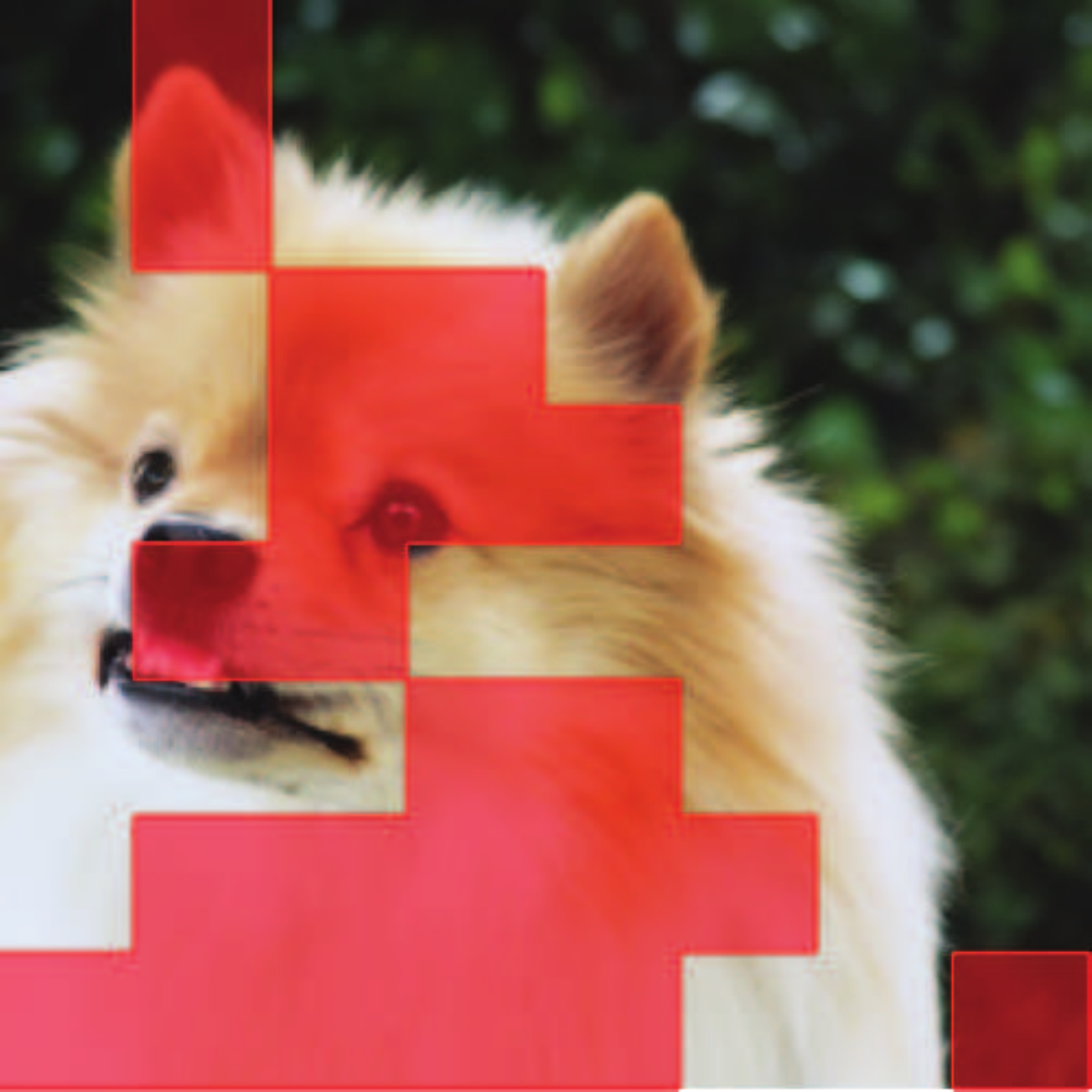}
\end{subfigure}
\begin{subfigure}{0.15\textwidth}
  \centering
  \includegraphics[width=1\linewidth]{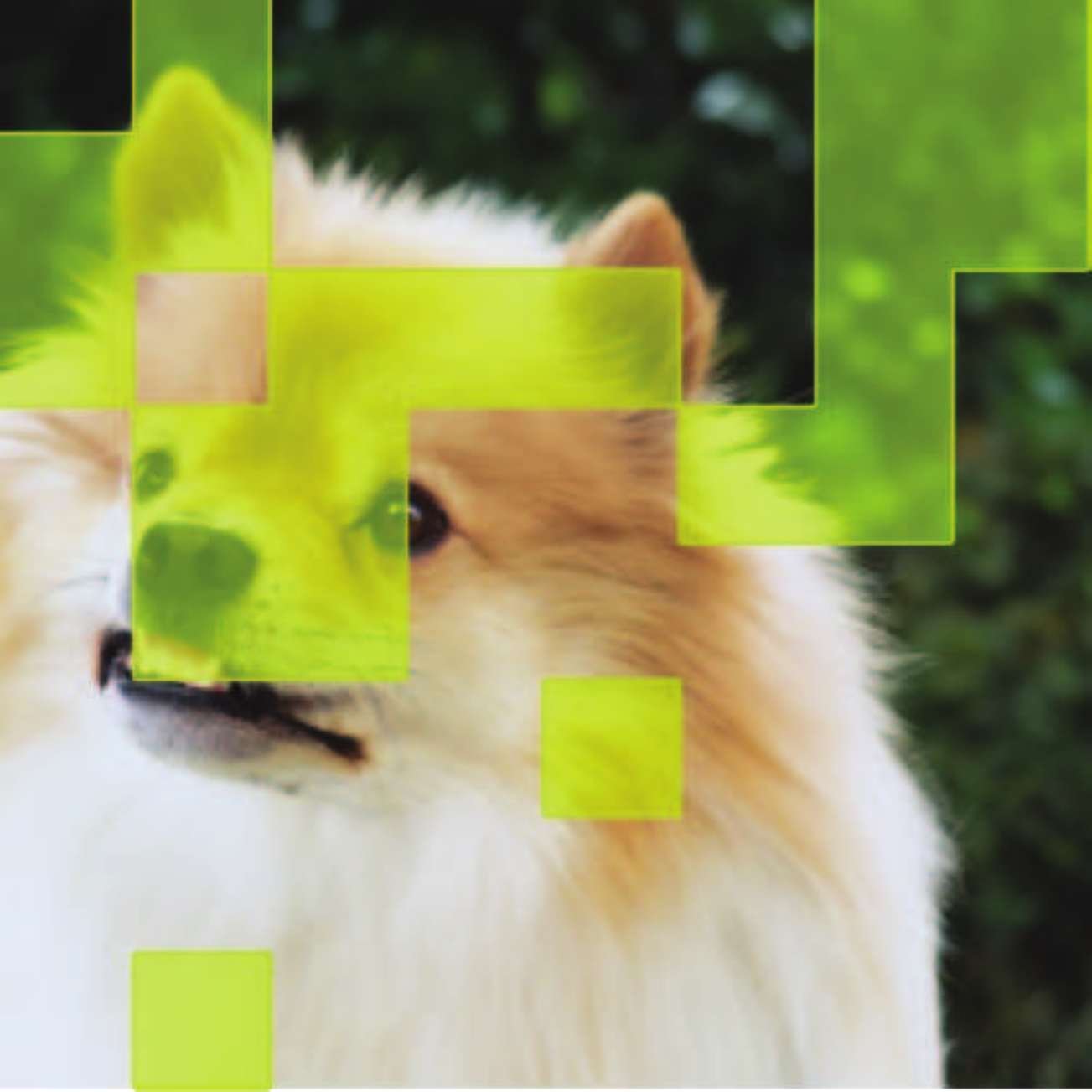}
\end{subfigure}
\begin{subfigure}{0.15\textwidth}
  \centering
  \includegraphics[width=1\linewidth]{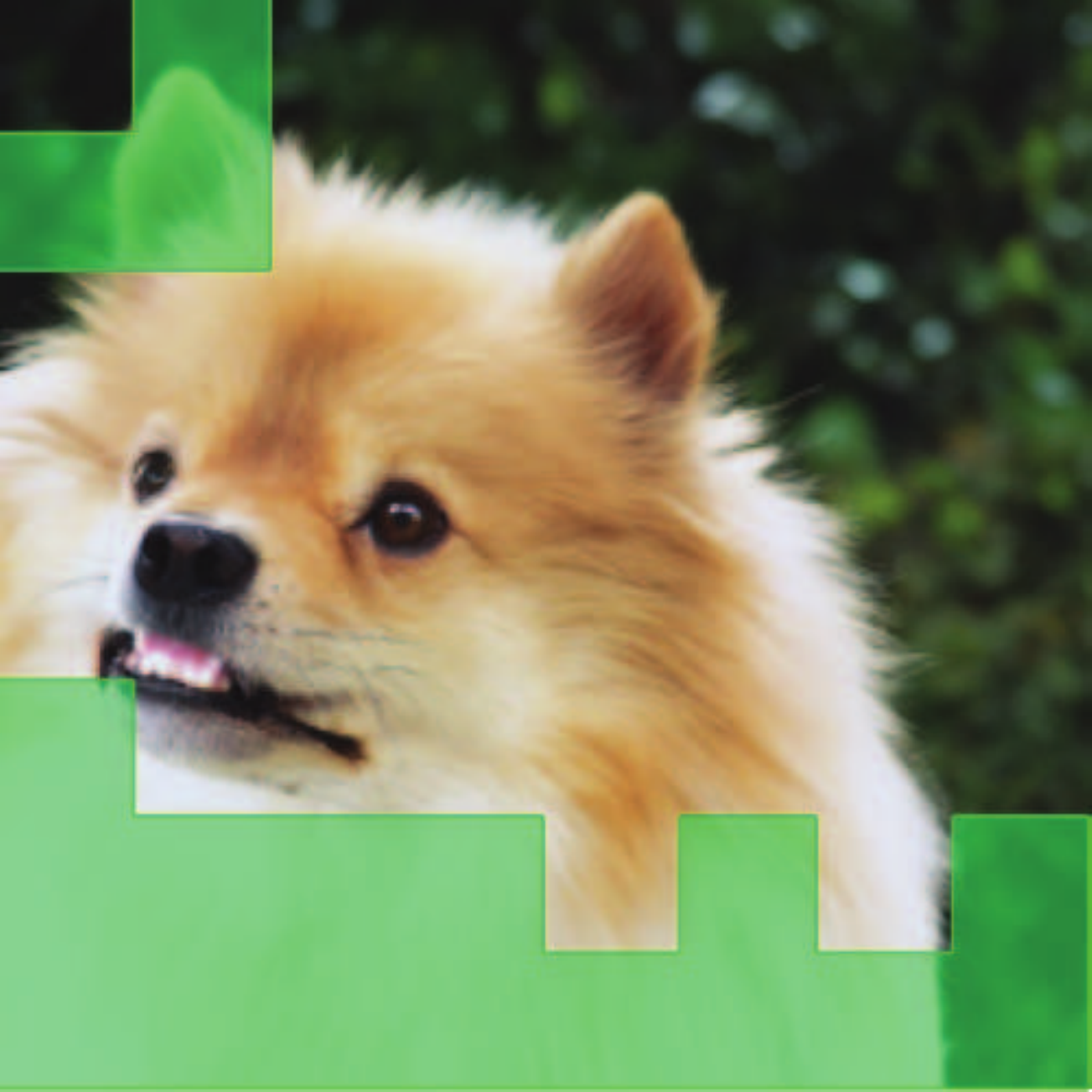}
\end{subfigure}
\begin{subfigure}{0.15\textwidth}
  \centering
  \includegraphics[width=1\linewidth]{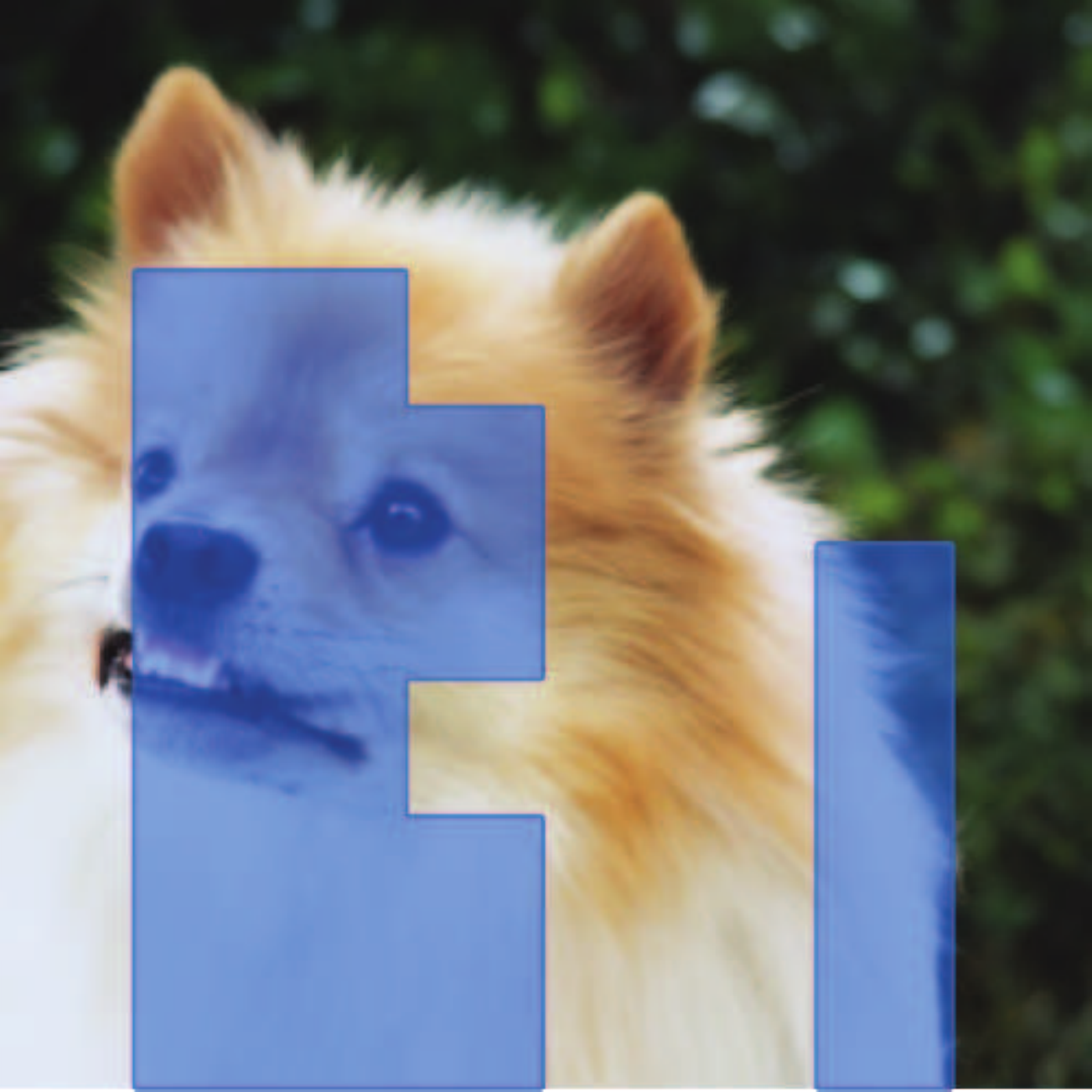}
\end{subfigure}
\begin{subfigure}{0.15\textwidth}
  \centering
  \includegraphics[width=1\linewidth]{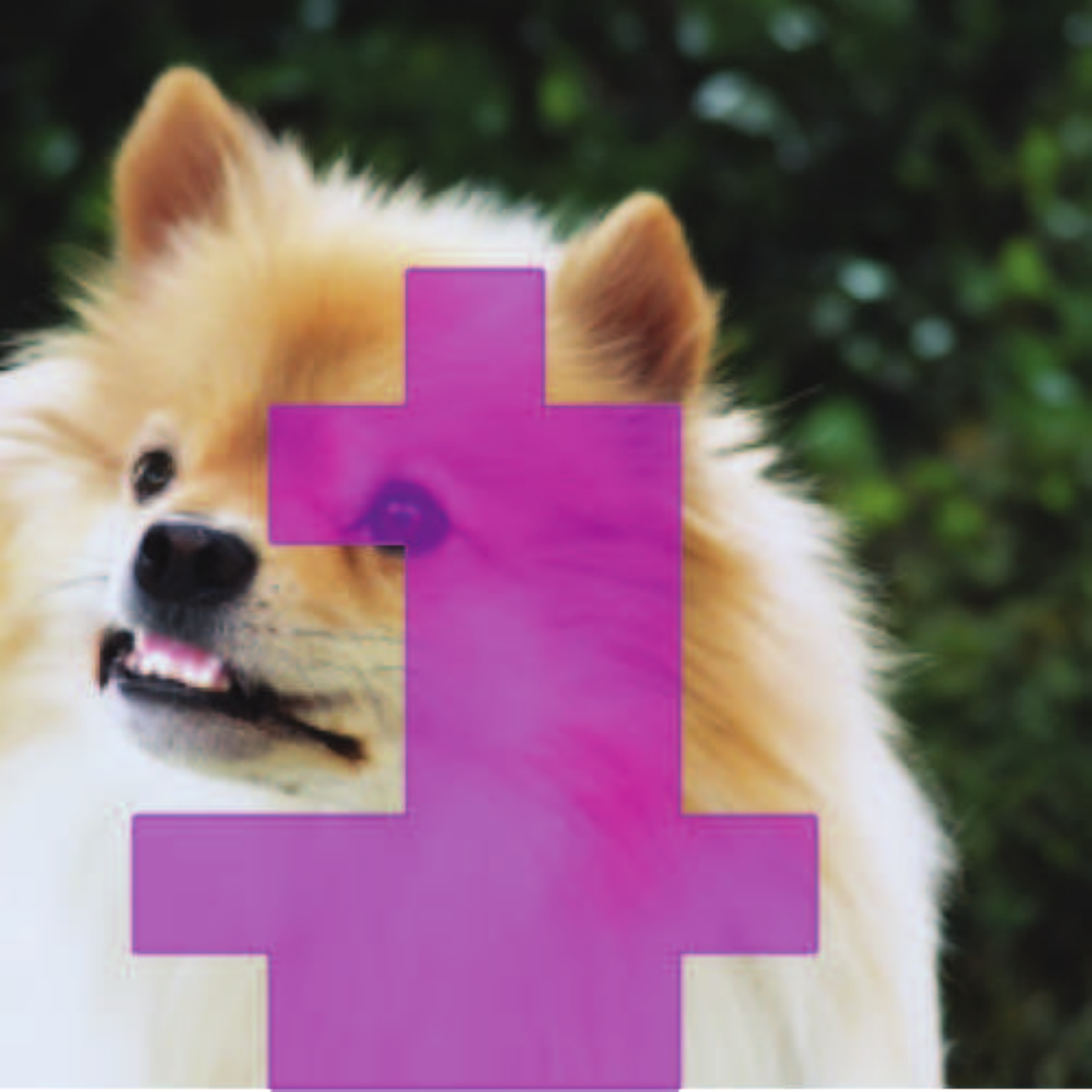}
\end{subfigure}\\
&
\begin{subfigure}{0.15\textwidth}
  \centering
  \includegraphics[width=1\linewidth]{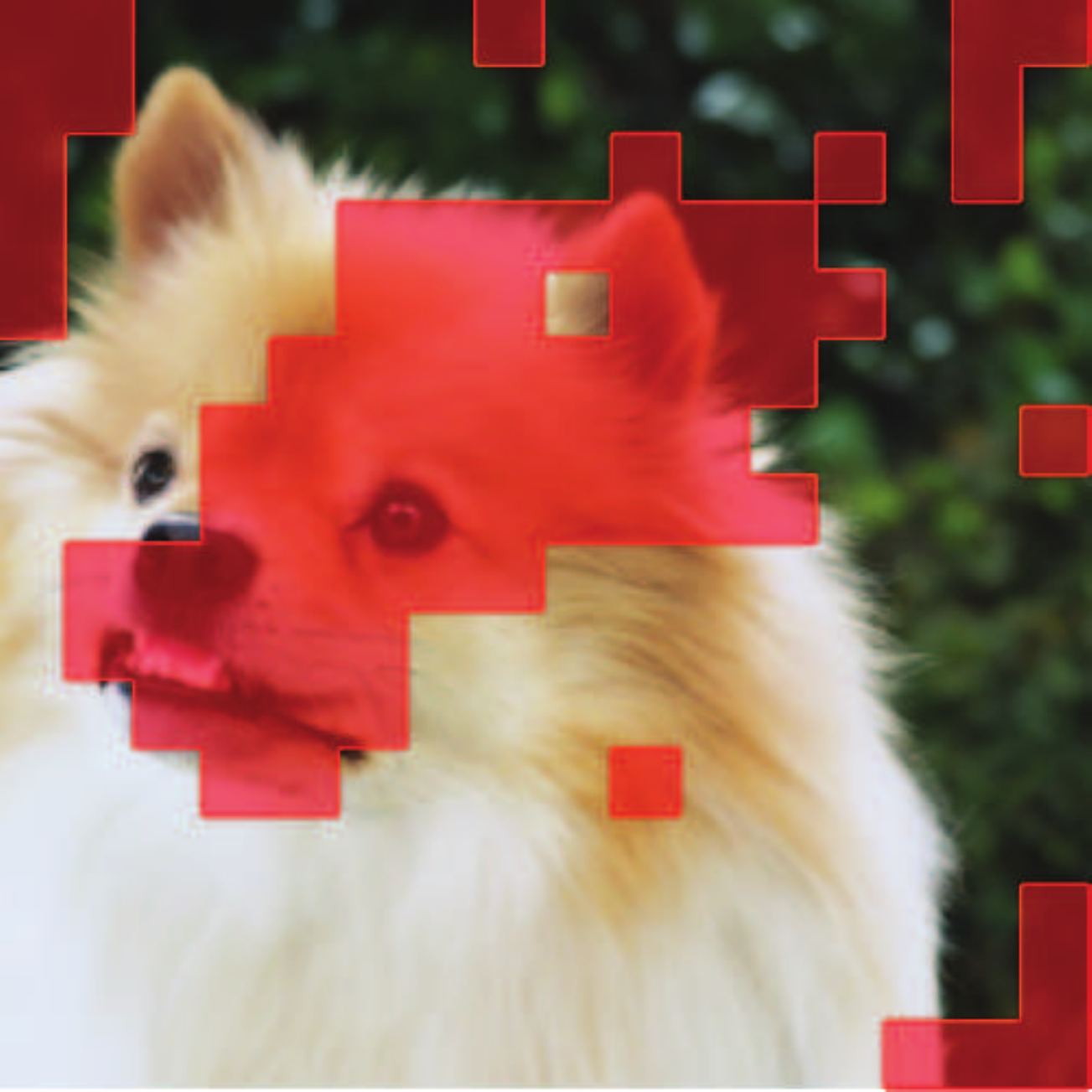}
\end{subfigure}
\begin{subfigure}{0.15\textwidth}
  \centering
  \includegraphics[width=1\linewidth]{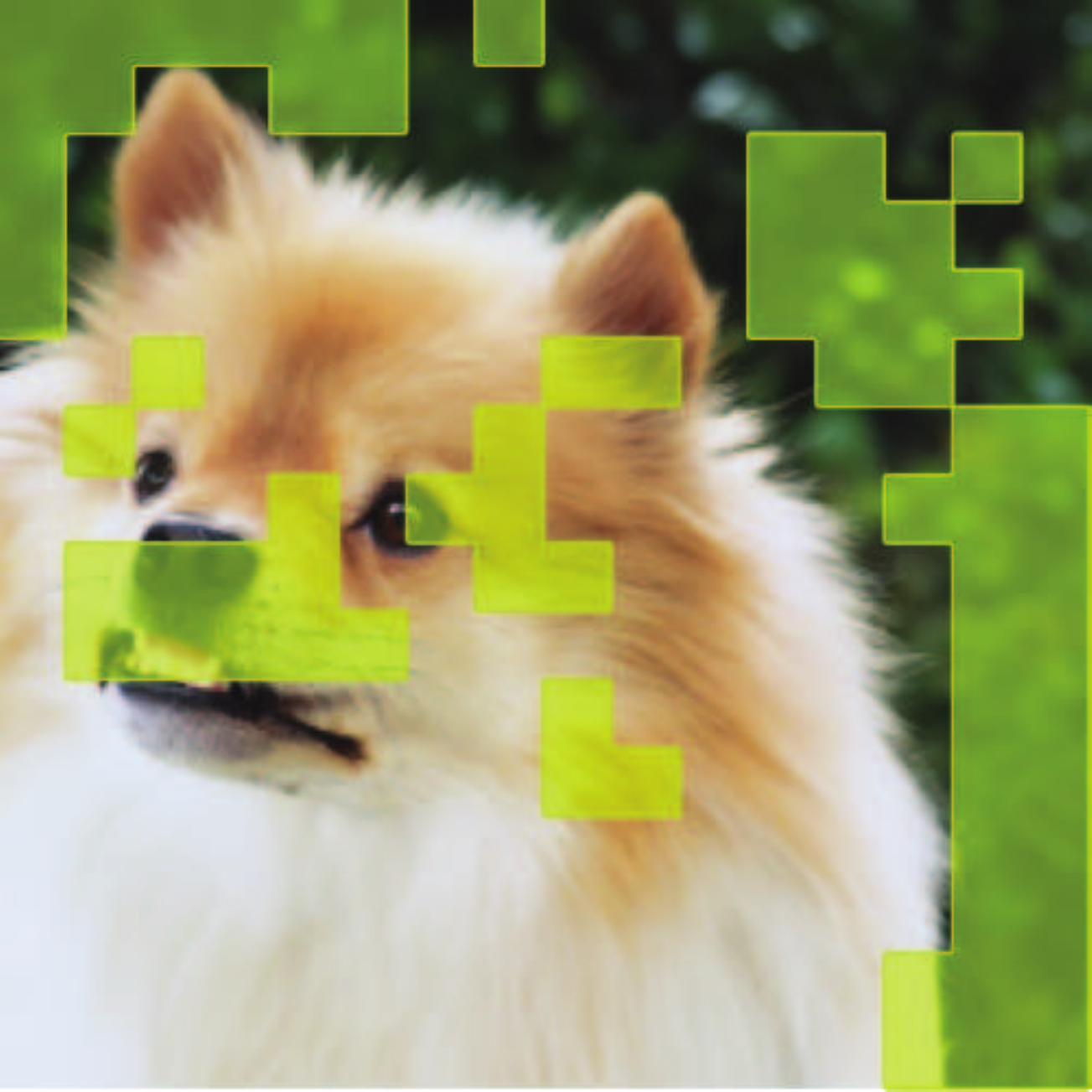}
\end{subfigure}
\begin{subfigure}{0.15\textwidth}
  \centering
  \includegraphics[width=1\linewidth]{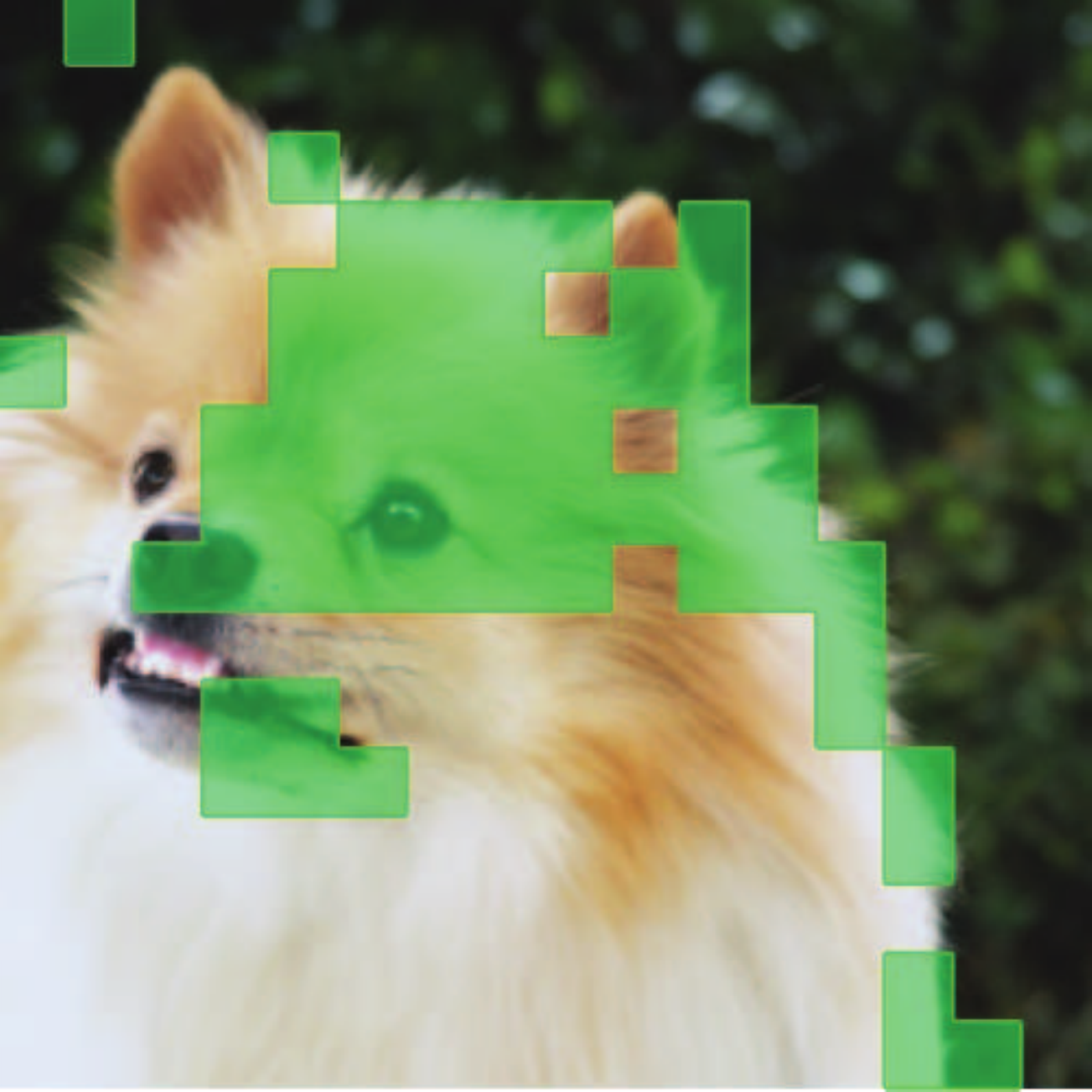}
\end{subfigure}
\begin{subfigure}{0.15\textwidth}
  \centering
  \includegraphics[width=1\linewidth]{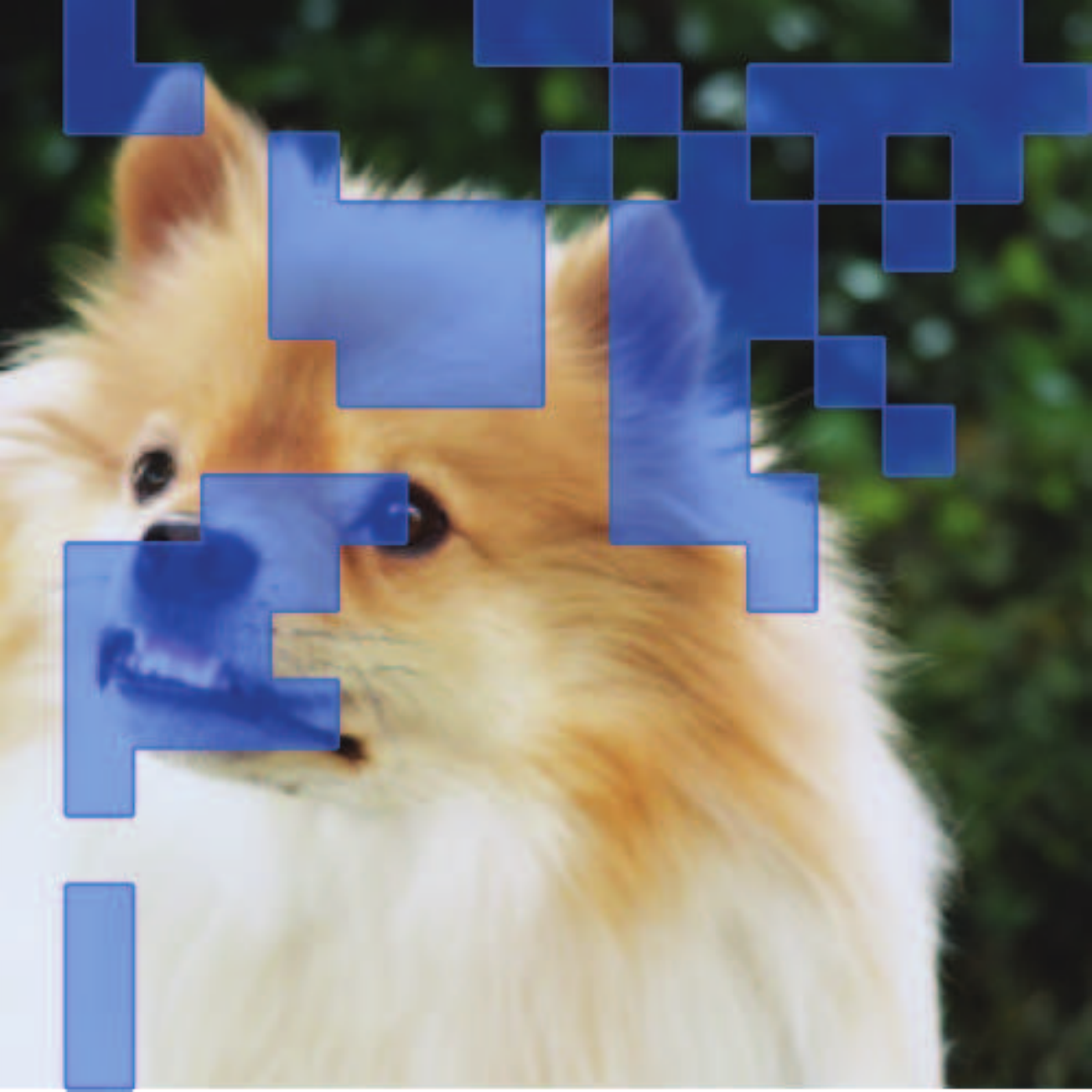}
\end{subfigure}
\begin{subfigure}{0.15\textwidth}
  \centering
  \includegraphics[width=1\linewidth]{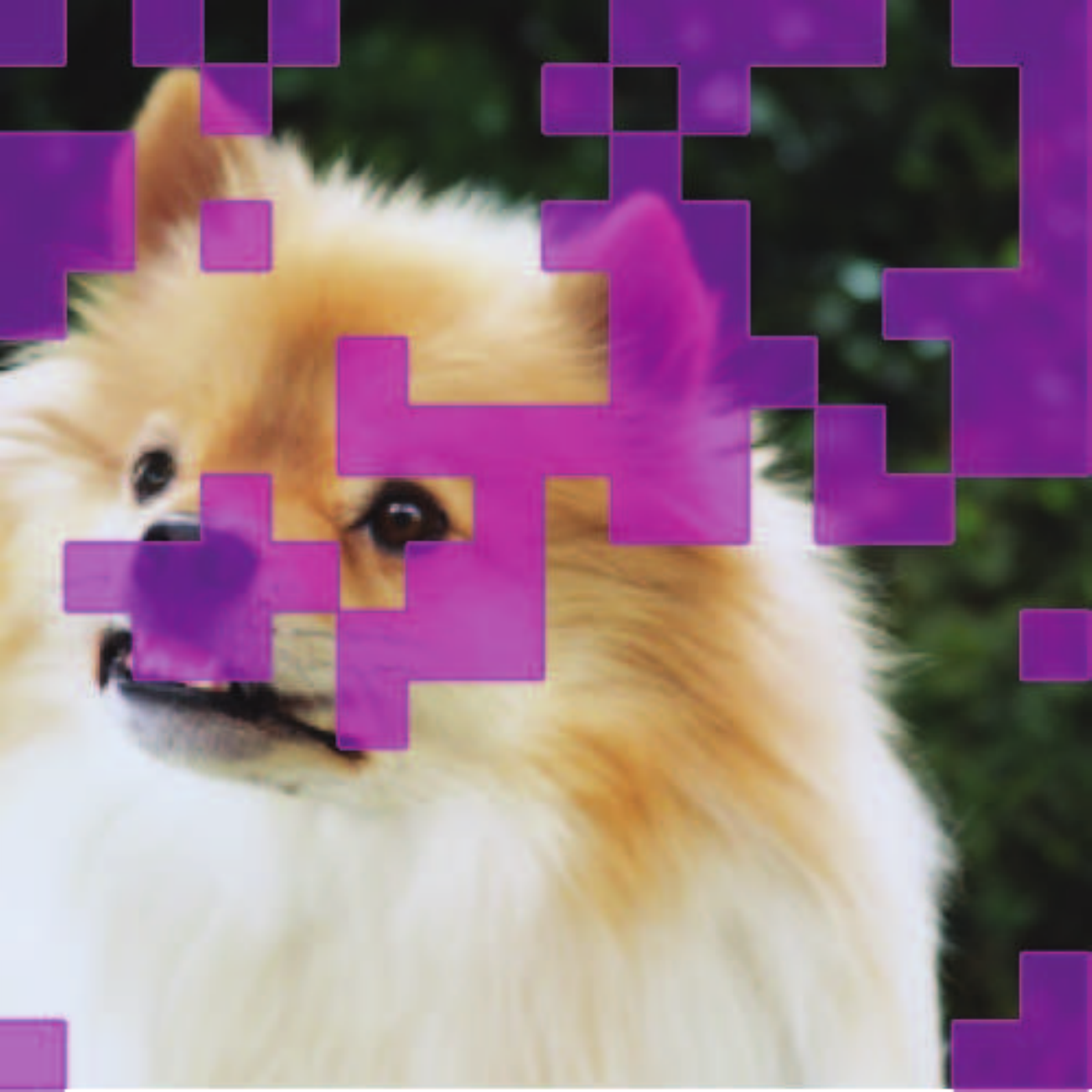}
\end{subfigure}
\\
&
\begin{subfigure}{0.15\textwidth}
  \centering
  \includegraphics[width=1\linewidth]{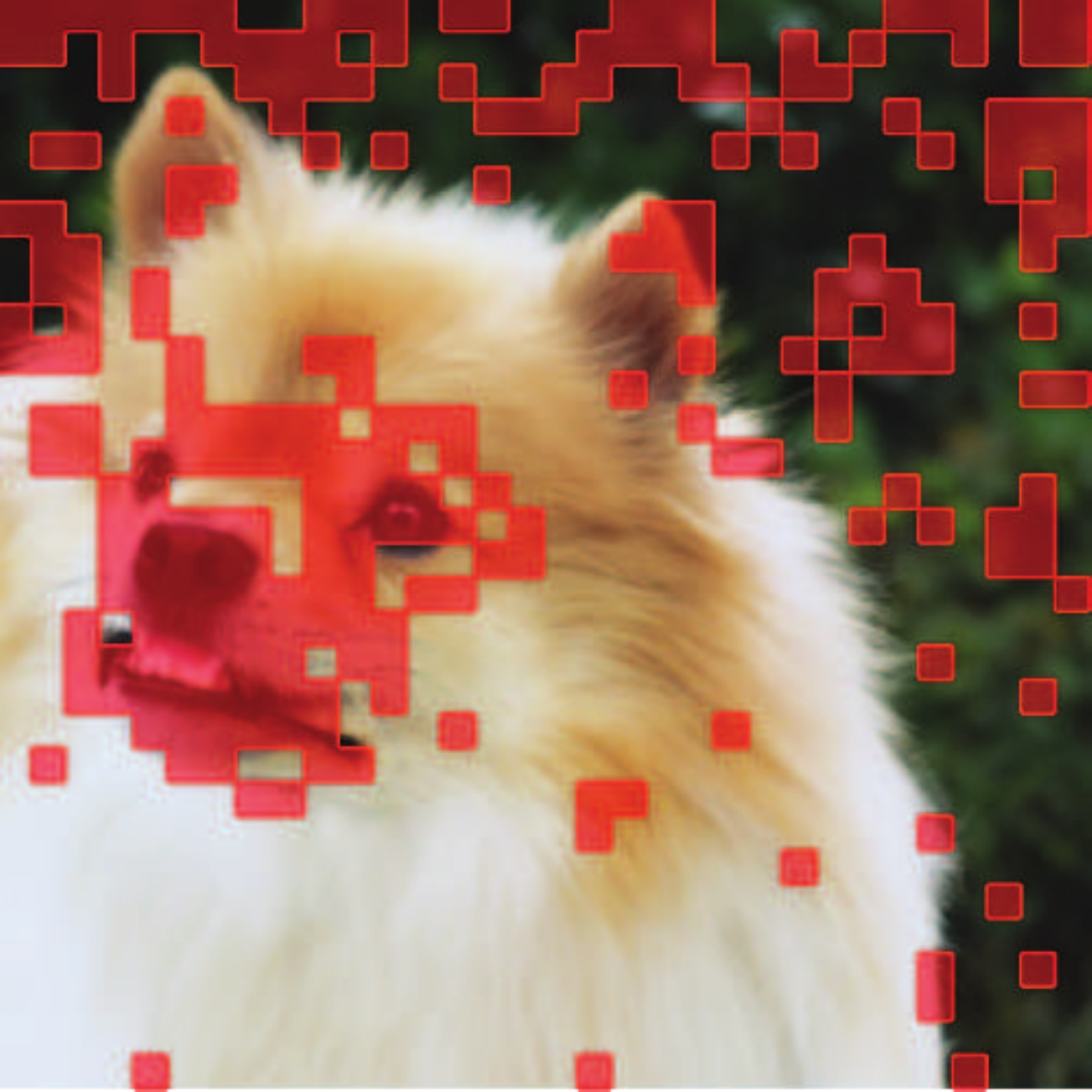}
  \caption*{Averaged}
\end{subfigure}
\begin{subfigure}{0.15\textwidth}
  \centering
  \includegraphics[width=1\linewidth]{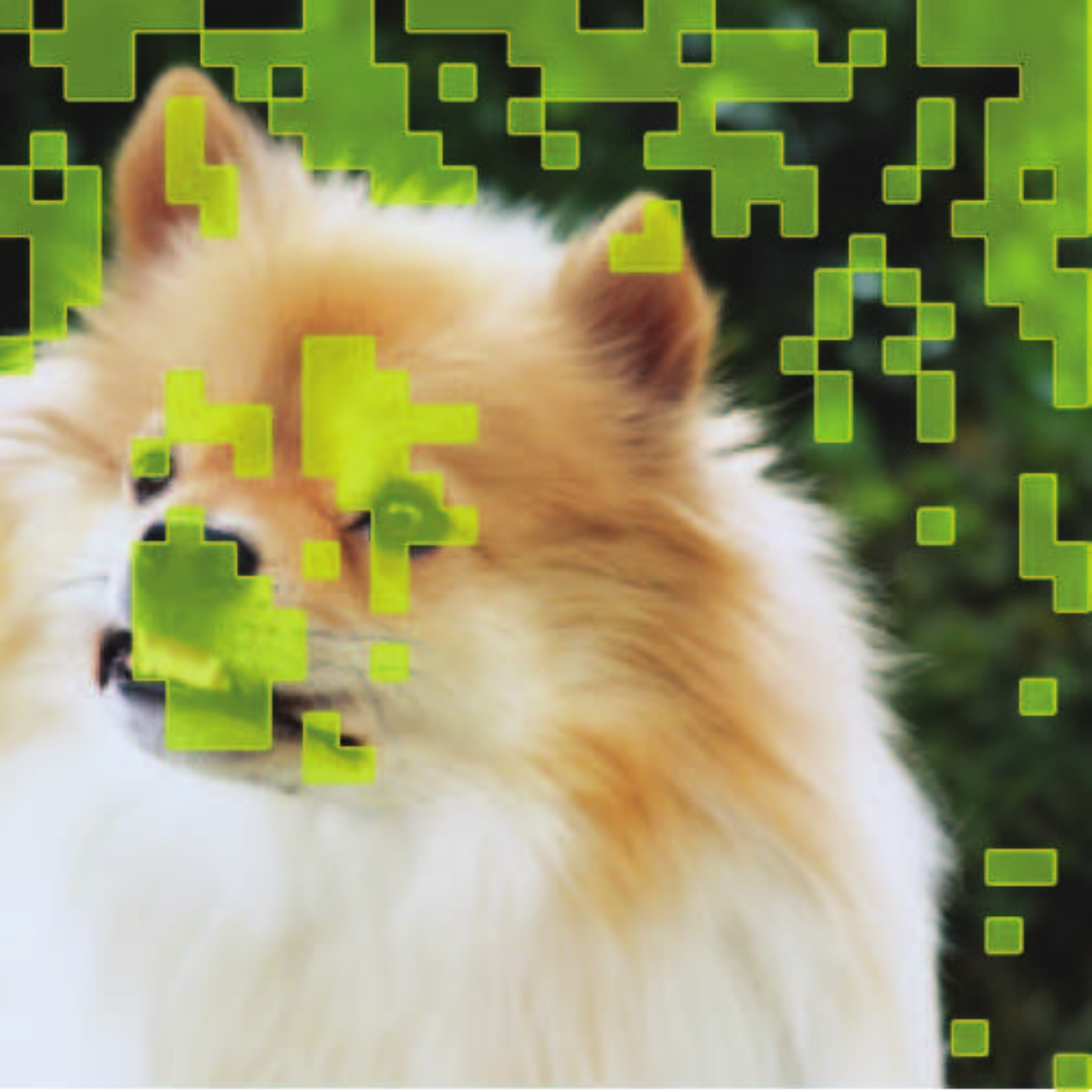}
  \caption*{Head 1}
\end{subfigure}
\begin{subfigure}{0.15\textwidth}
  \centering
  \includegraphics[width=1\linewidth]{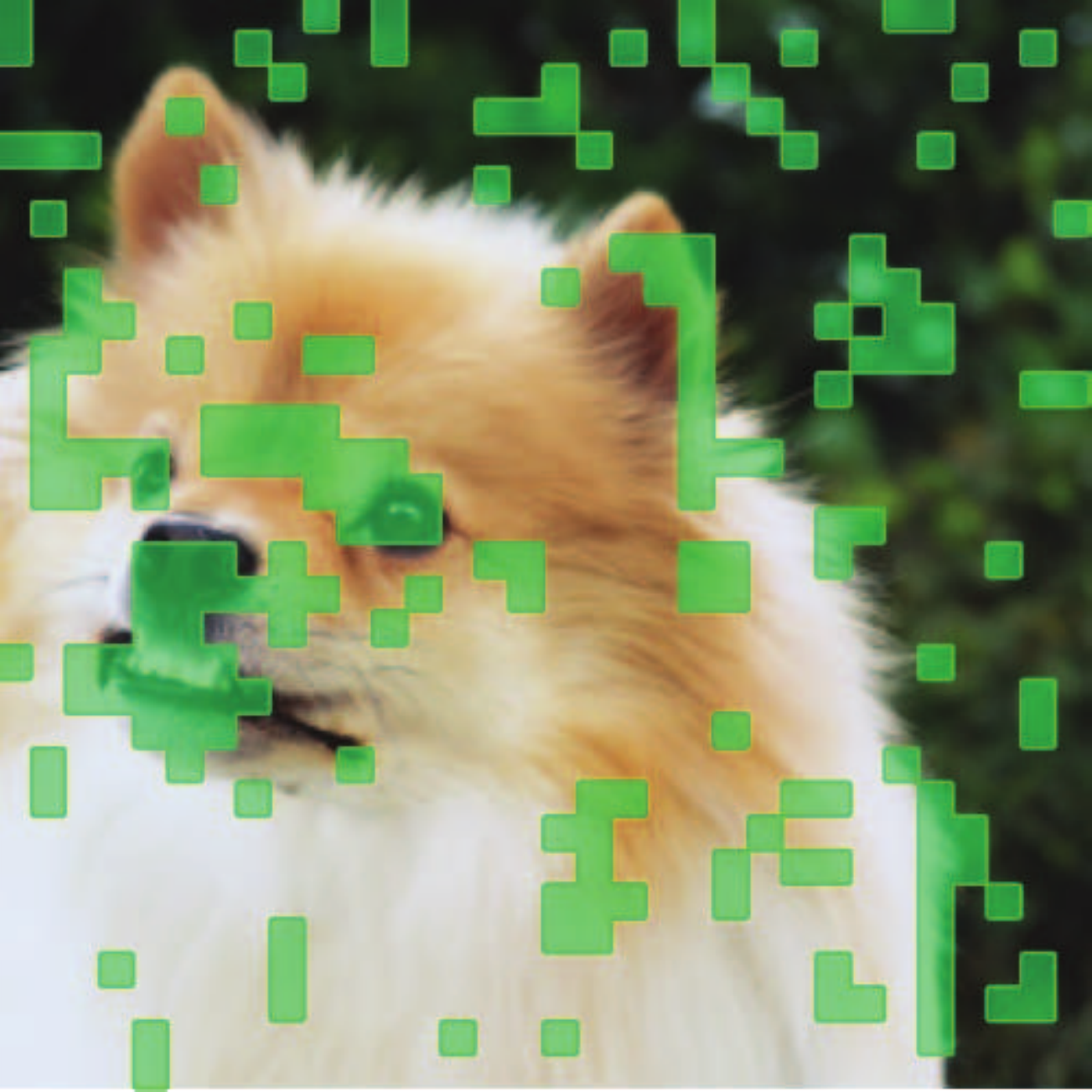}
  \caption*{Head 2}
\end{subfigure}
\begin{subfigure}{0.15\textwidth}
  \centering
  \includegraphics[width=1\linewidth]{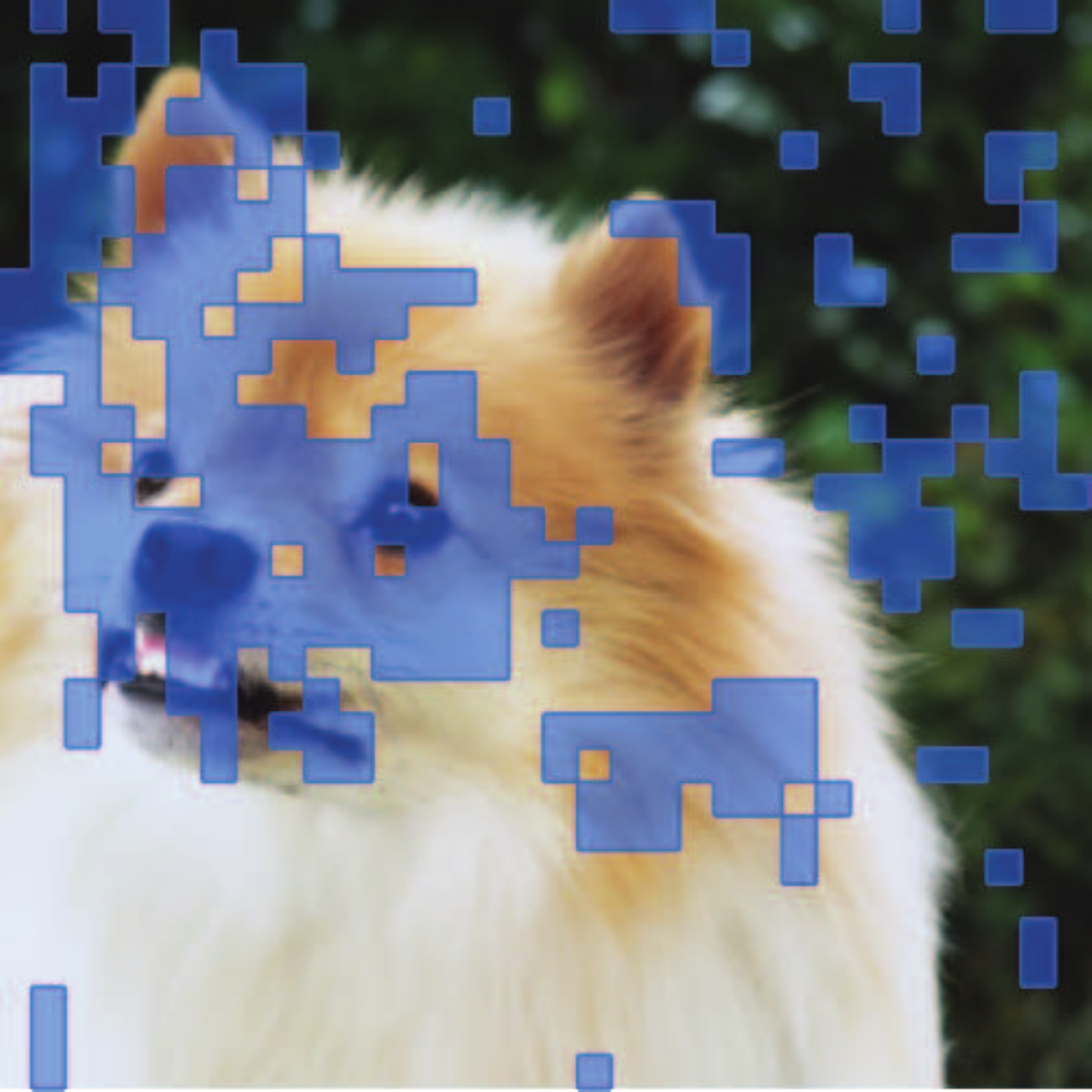}
  \caption*{Head 3}
\end{subfigure}
\begin{subfigure}{0.15\textwidth}
  \centering
  \includegraphics[width=1\linewidth]{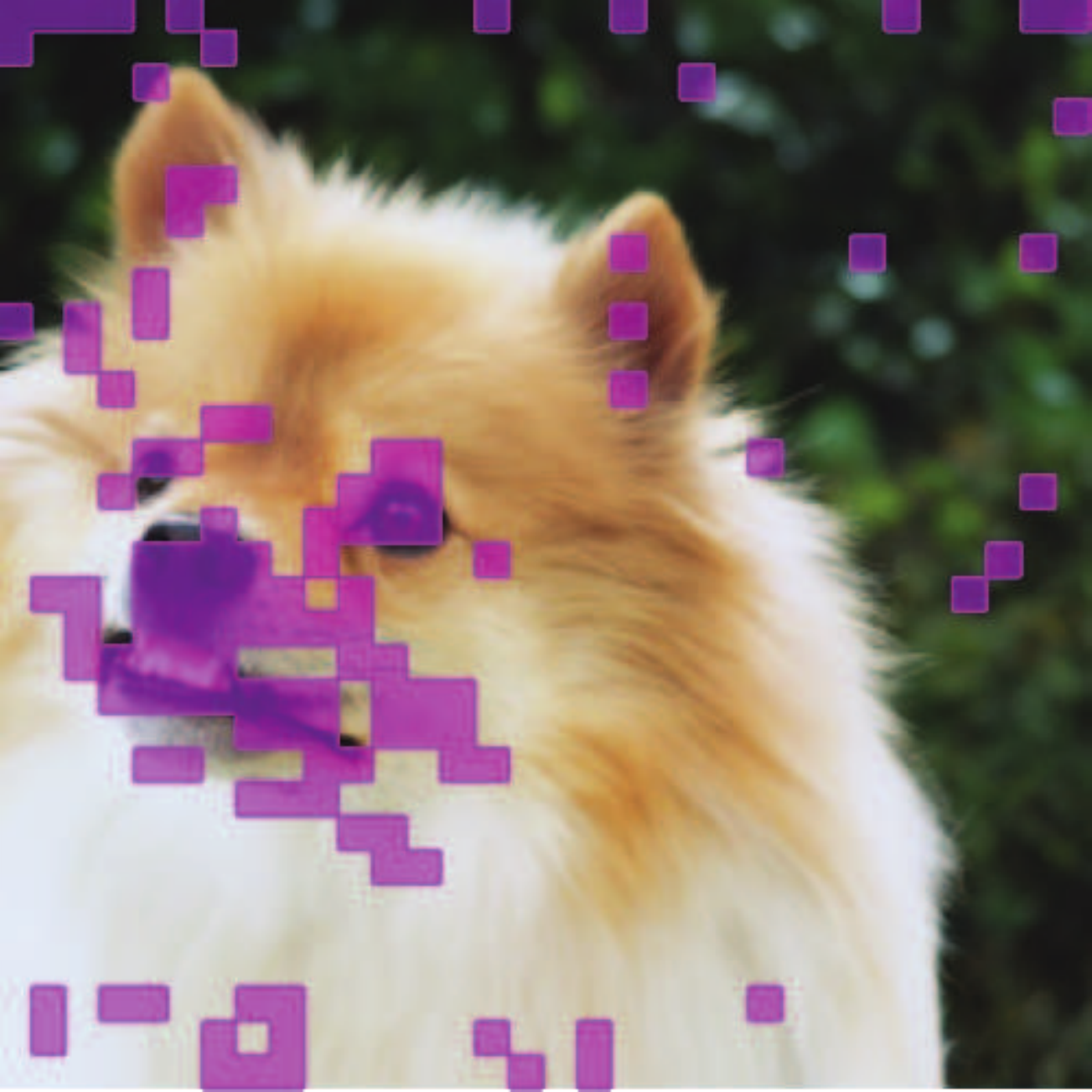}
  \caption*{Head 4}
\end{subfigure}
\end{tabular}
\vspace{-5pt}
\caption{\textbf{Visualization of self-attention masks from different layers and heads}. Each row, top to bottom, corresponds to $8\times8$, $16\times16$ and $32\times32$ self-attention layers, respectively.}
\label{fig:head1}
\end{figure*}

\begin{figure*}[p]
\centering
\begin{tabular}{lc}
\centering
\begin{subfigure}{0.15\textwidth}
  \centering
  \includegraphics[width=1\textwidth]{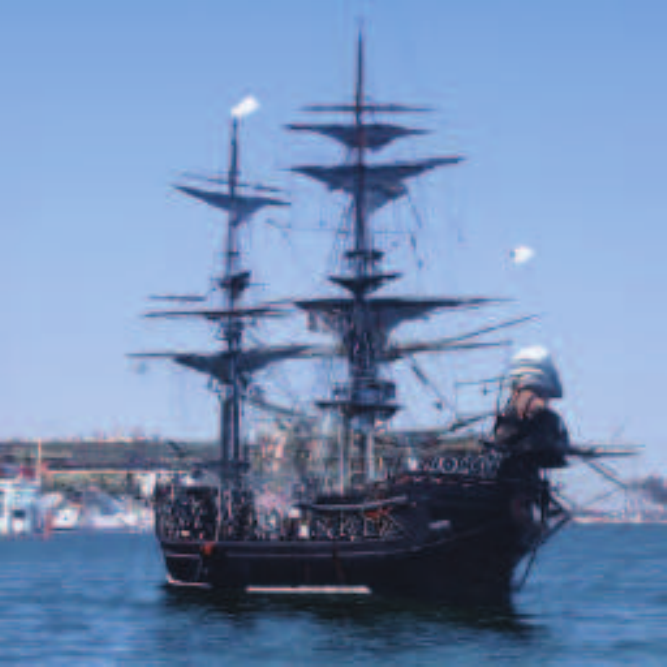}
\end{subfigure}
&
\begin{subfigure}{0.15\textwidth}
  \centering
  \includegraphics[width=1\linewidth]{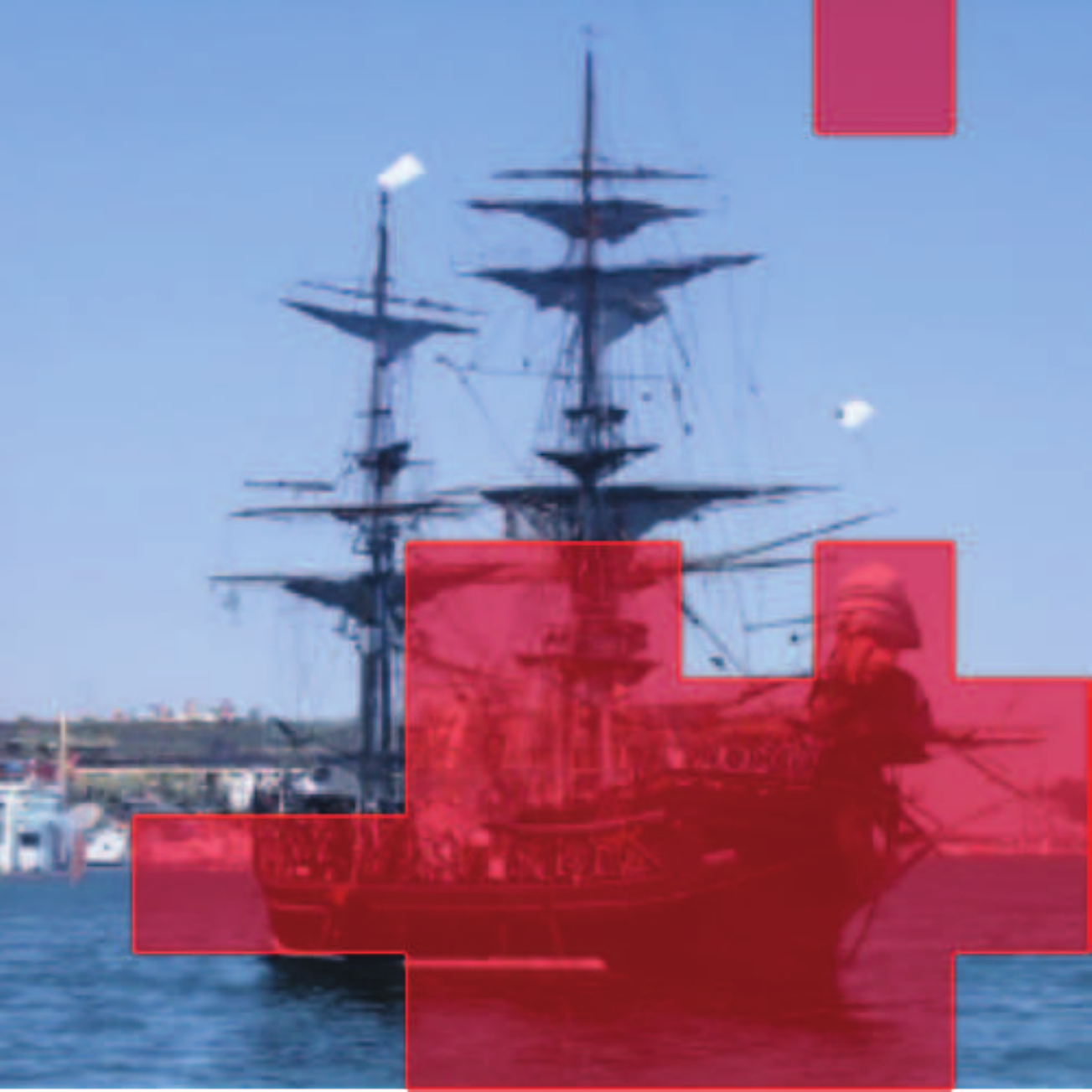}
\end{subfigure}
\begin{subfigure}{0.15\textwidth}
  \centering
  \includegraphics[width=1\linewidth]{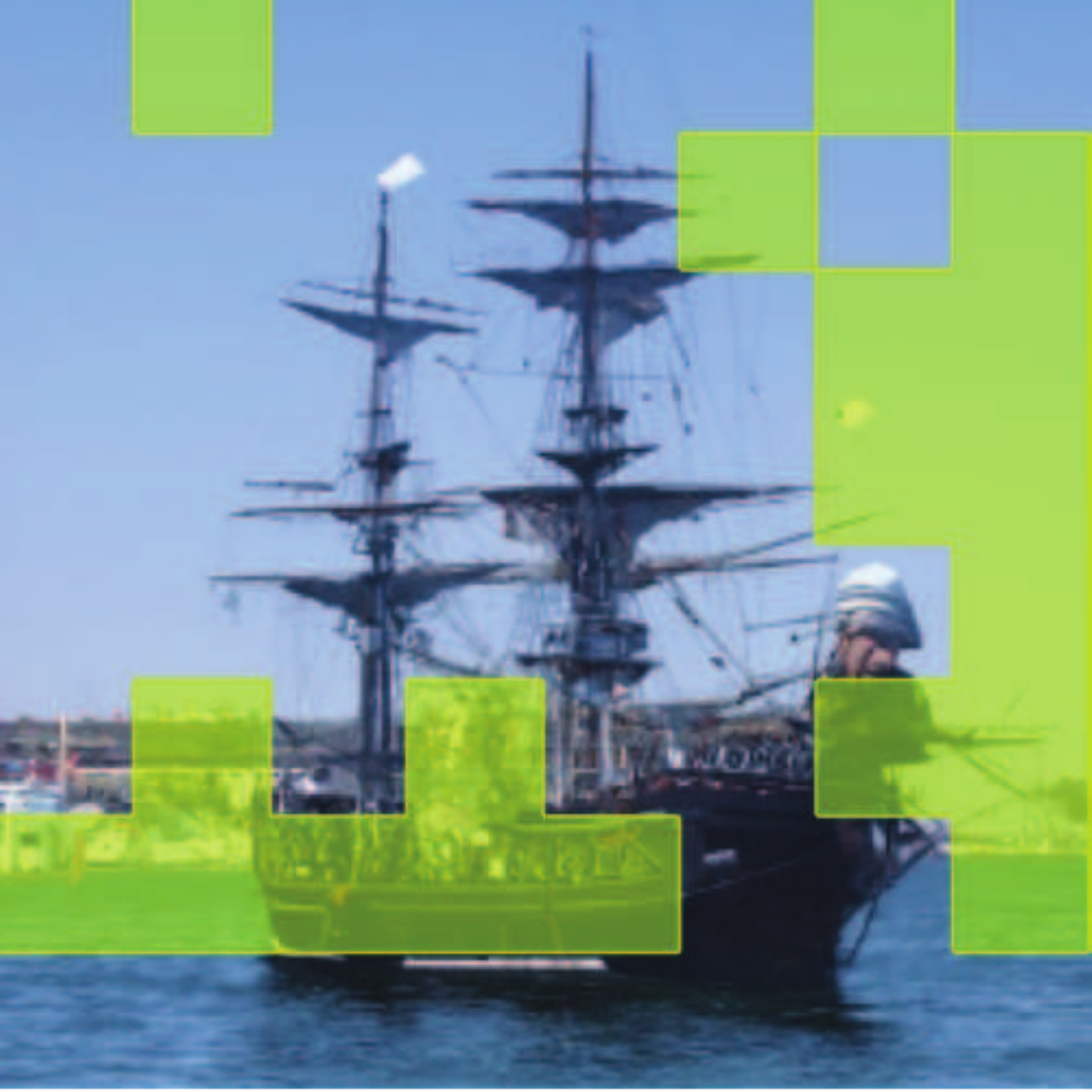}
\end{subfigure}
\begin{subfigure}{0.15\textwidth}
  \centering
  \includegraphics[width=1\linewidth]{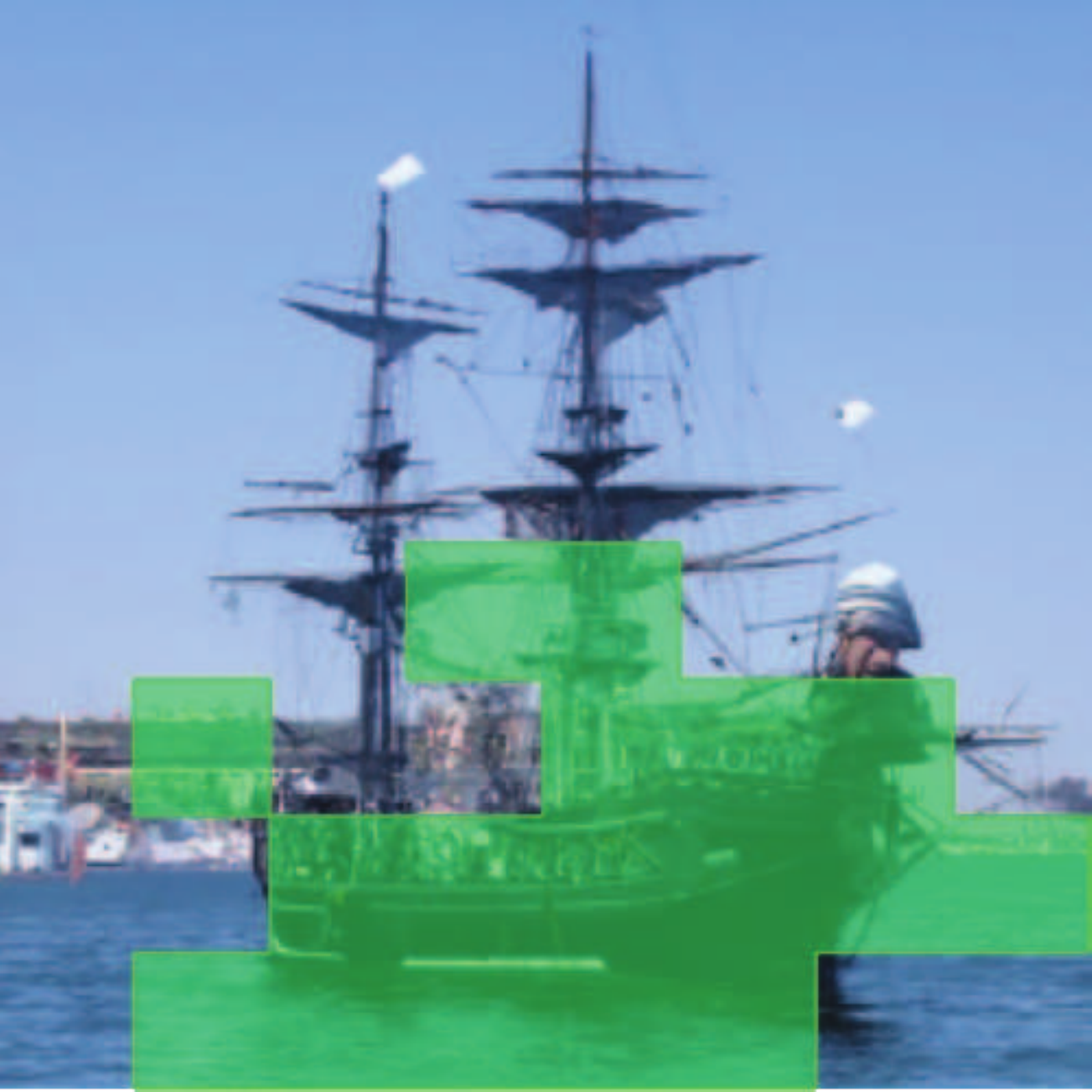}
\end{subfigure}
\begin{subfigure}{0.15\textwidth}
  \centering
  \includegraphics[width=1\linewidth]{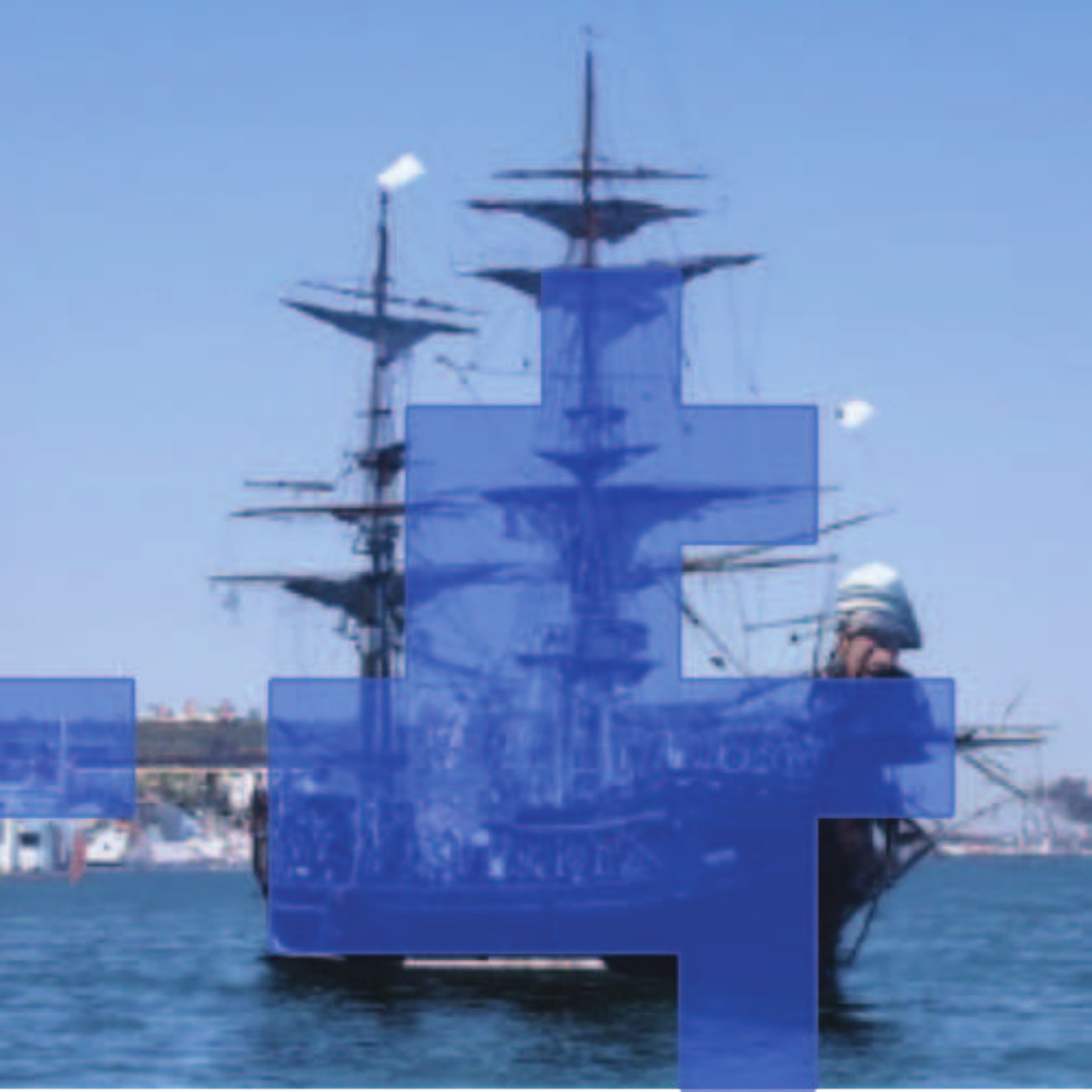}
\end{subfigure}
\begin{subfigure}{0.15\textwidth}
  \centering
  \includegraphics[width=1\linewidth]{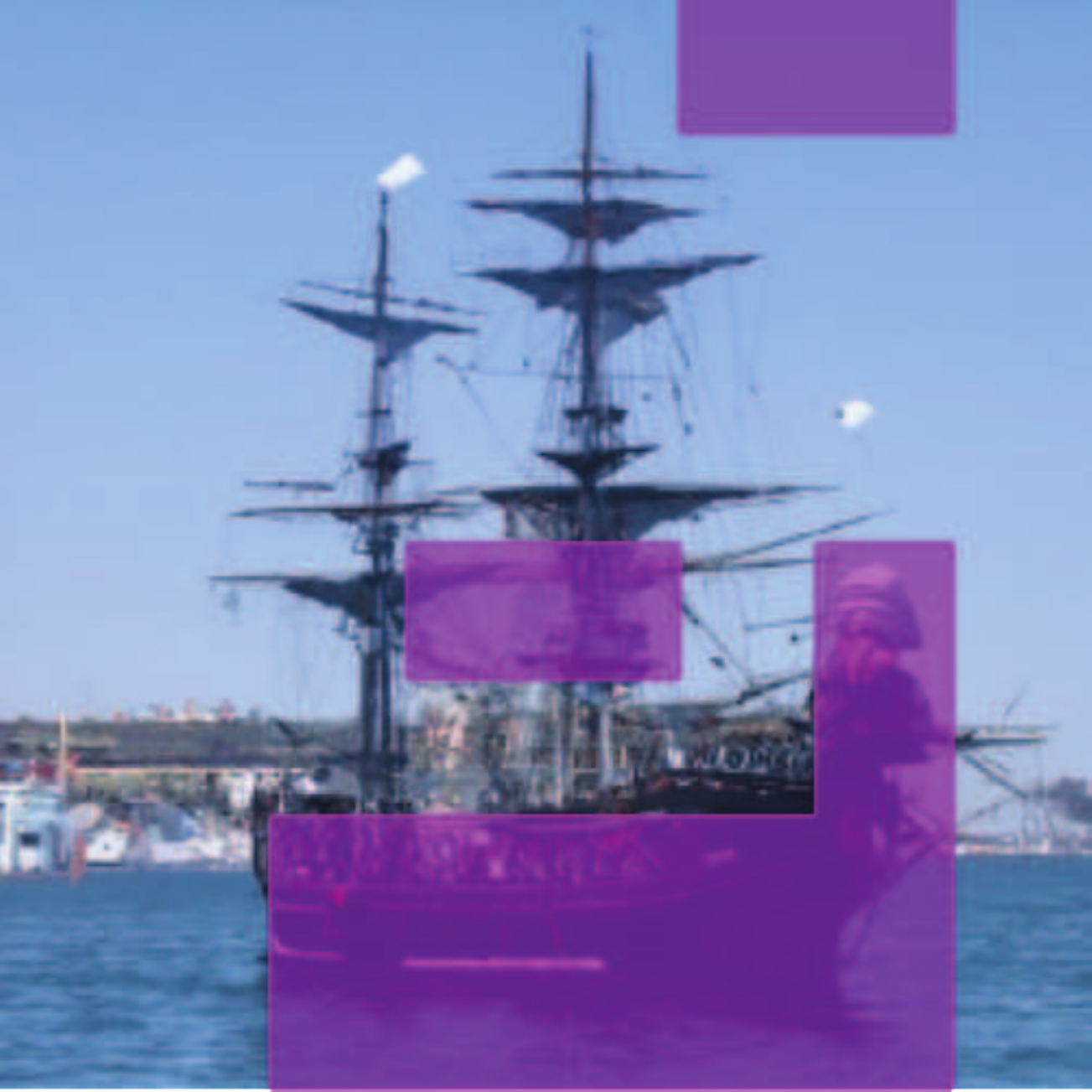}

\end{subfigure}\\
&
\begin{subfigure}{0.15\textwidth}
  \centering
  \includegraphics[width=1\linewidth]{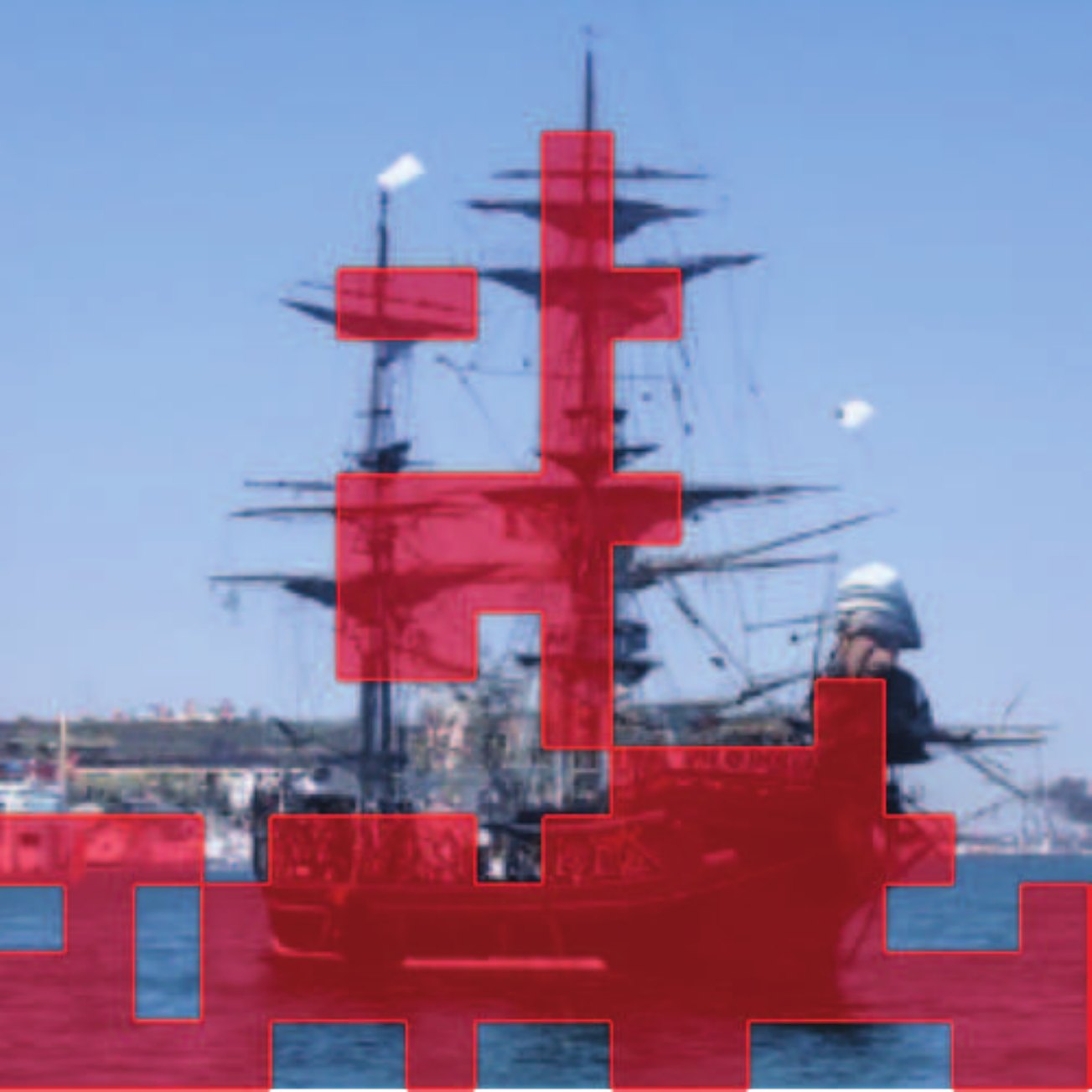}
\end{subfigure}
\begin{subfigure}{0.15\textwidth}
  \centering
  \includegraphics[width=1\linewidth]{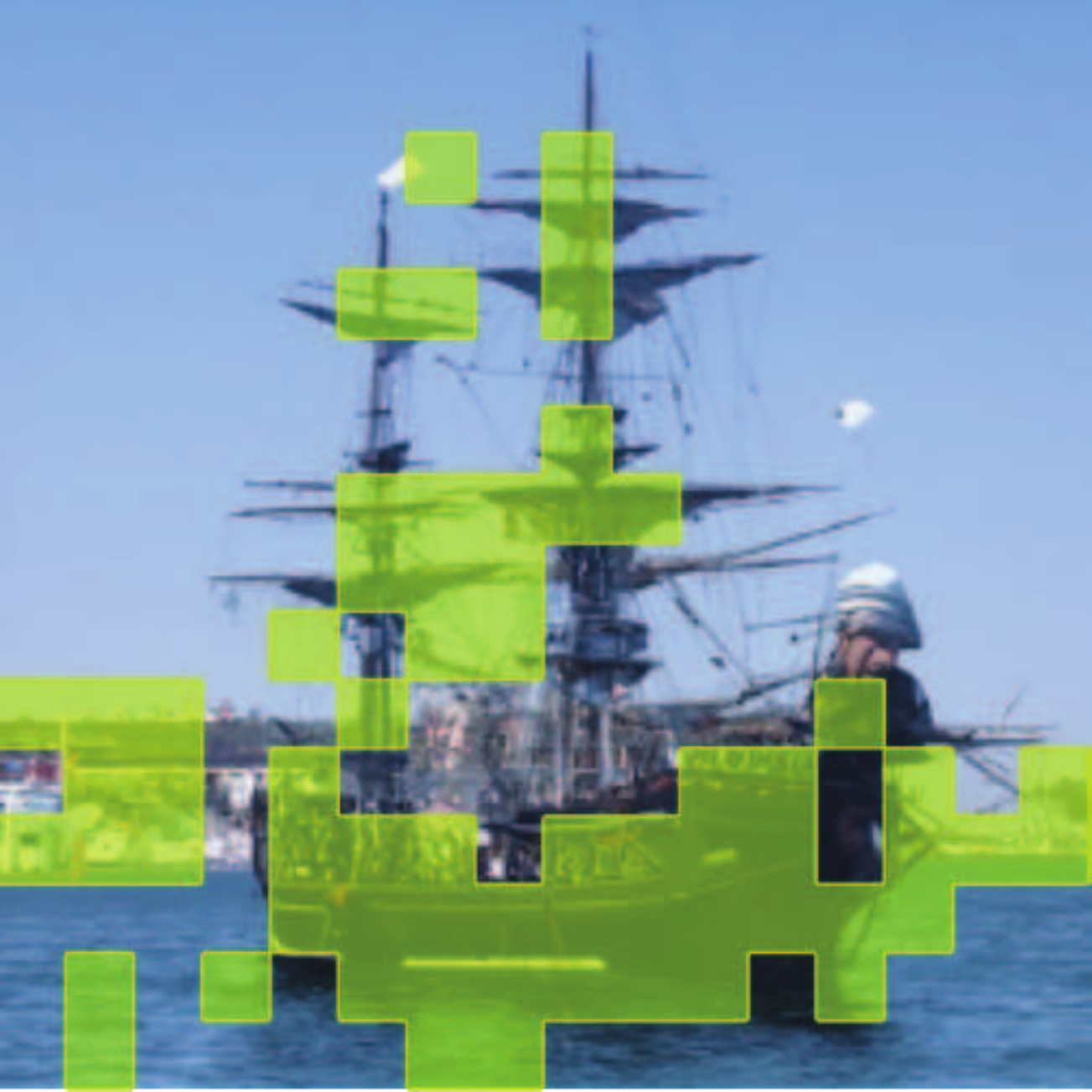}
\end{subfigure}
\begin{subfigure}{0.15\textwidth}
  \centering
  \includegraphics[width=1\linewidth]{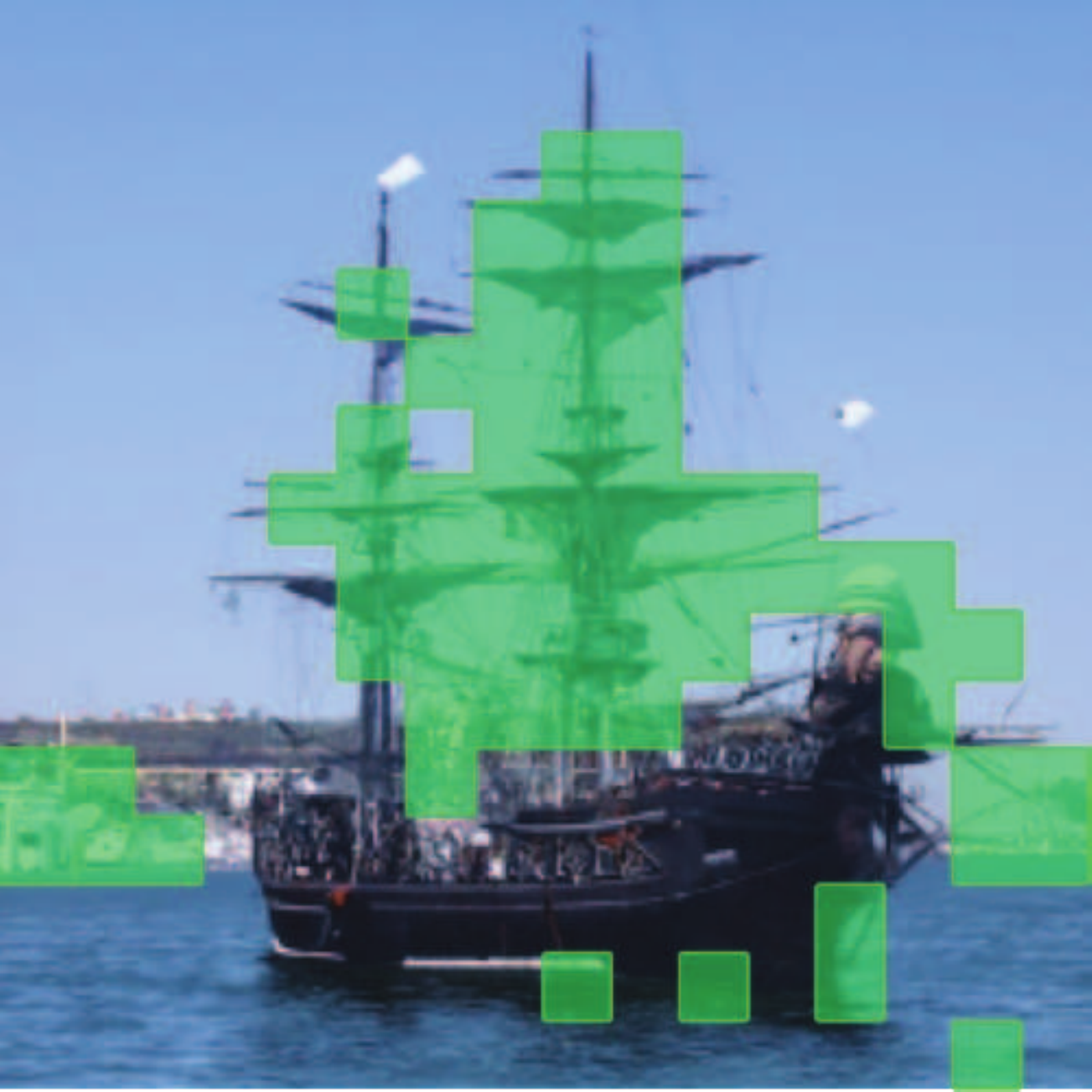}
\end{subfigure}
\begin{subfigure}{0.15\textwidth}
  \centering
  \includegraphics[width=1\linewidth]{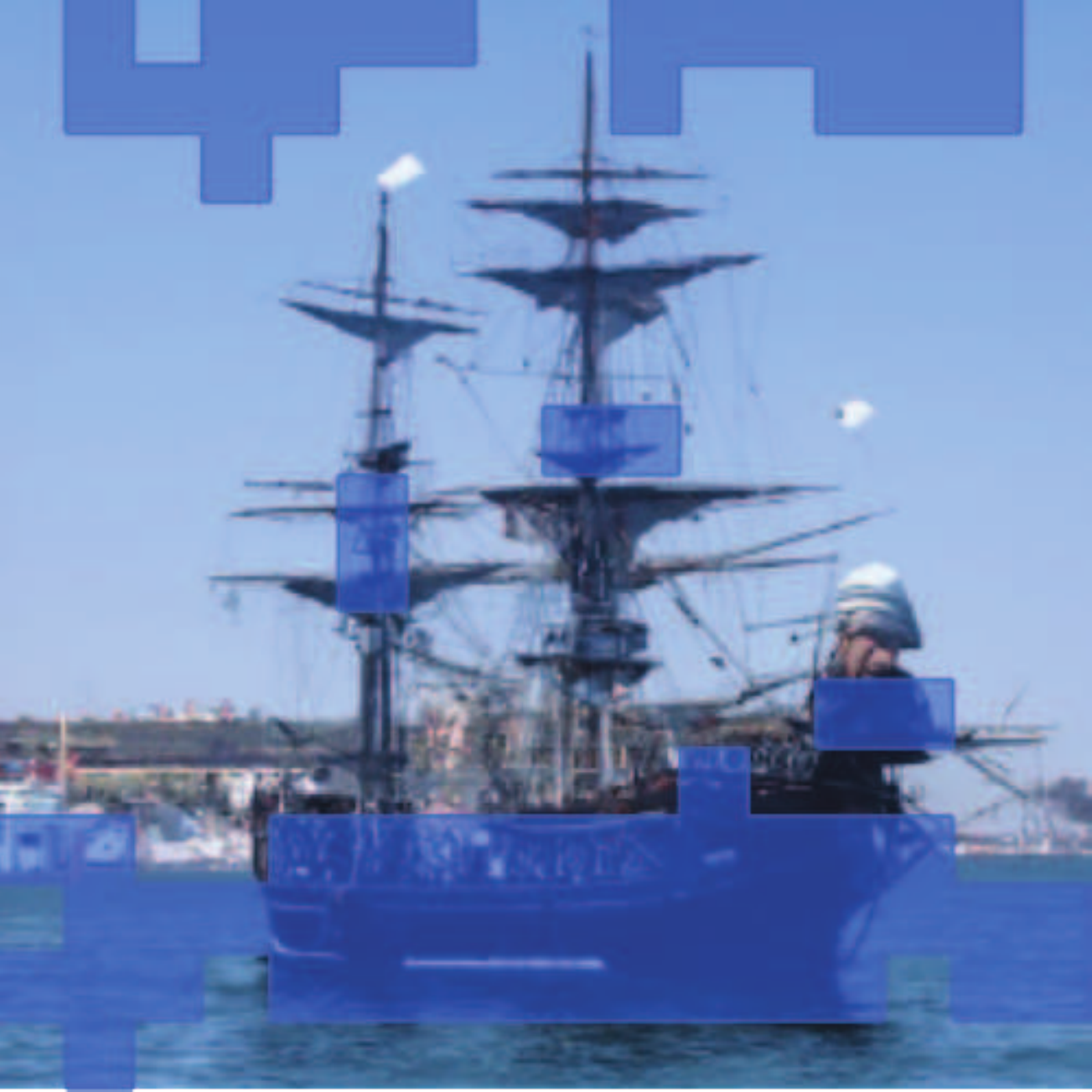}
\end{subfigure}
\begin{subfigure}{0.15\textwidth}
  \centering
  \includegraphics[width=1\linewidth]{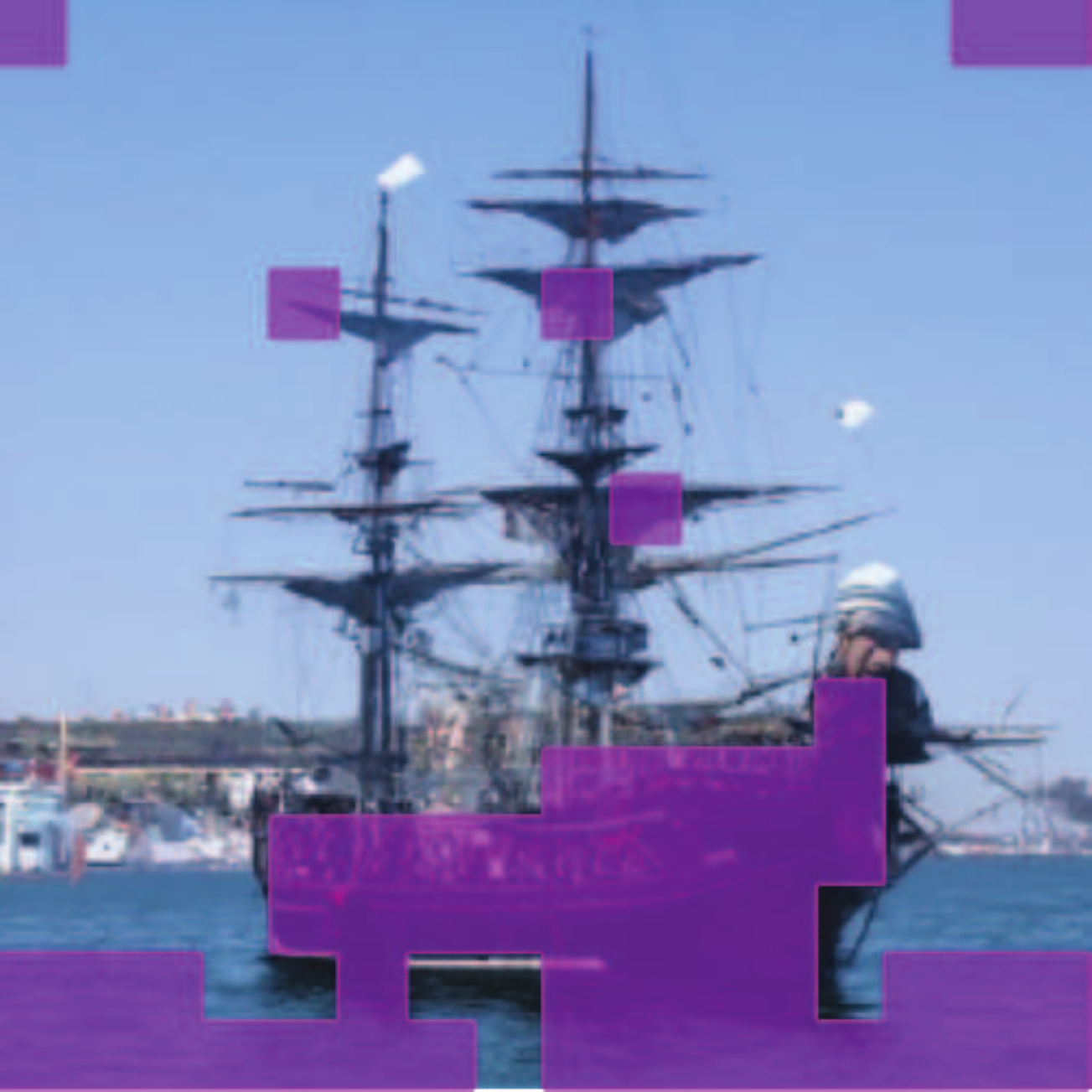}
\end{subfigure}\\
&
\begin{subfigure}{0.15\textwidth}
  \centering
  \includegraphics[width=1\linewidth]{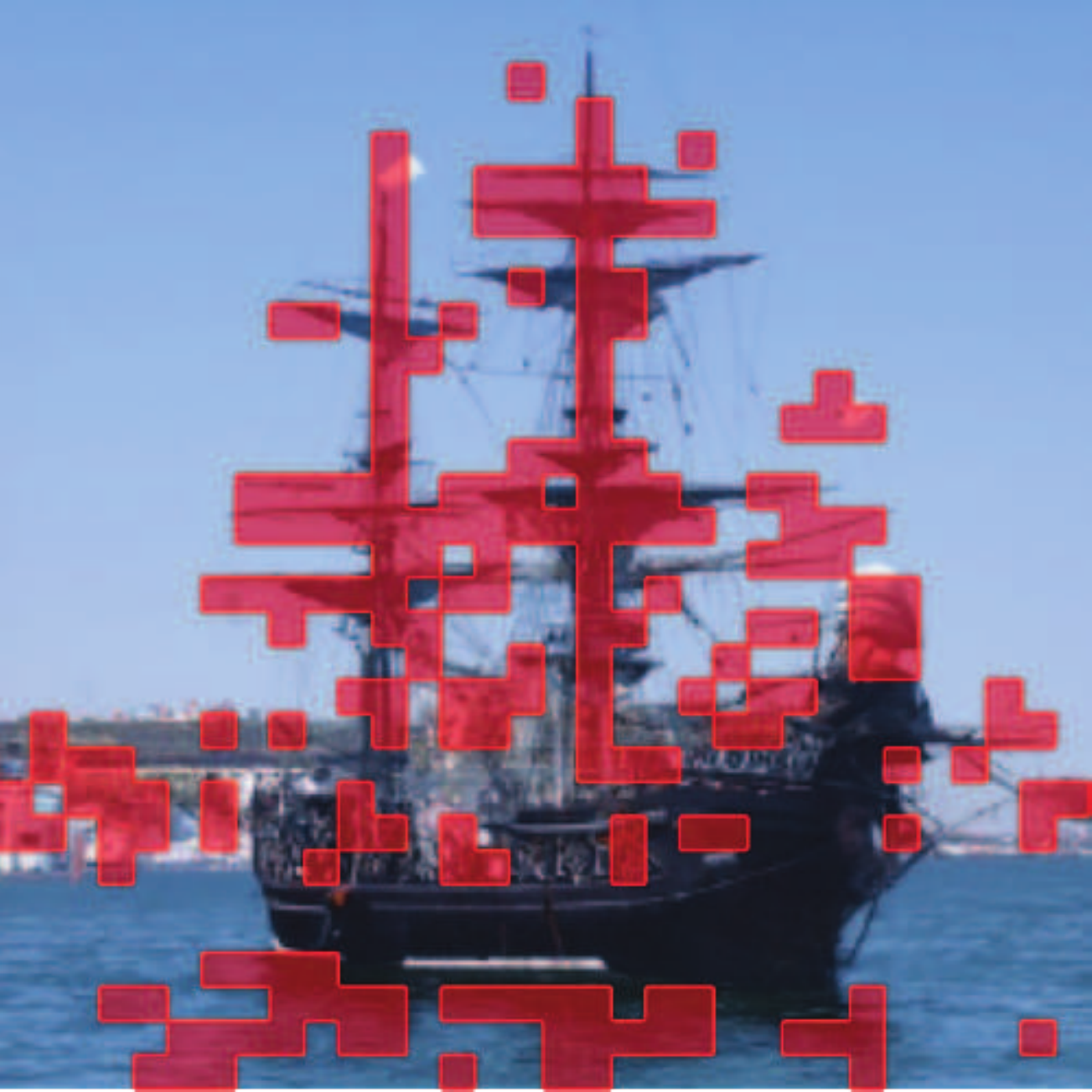}
  \caption*{Averaged}
\end{subfigure}
\begin{subfigure}{0.15\textwidth}
  \centering
  \includegraphics[width=1\linewidth]{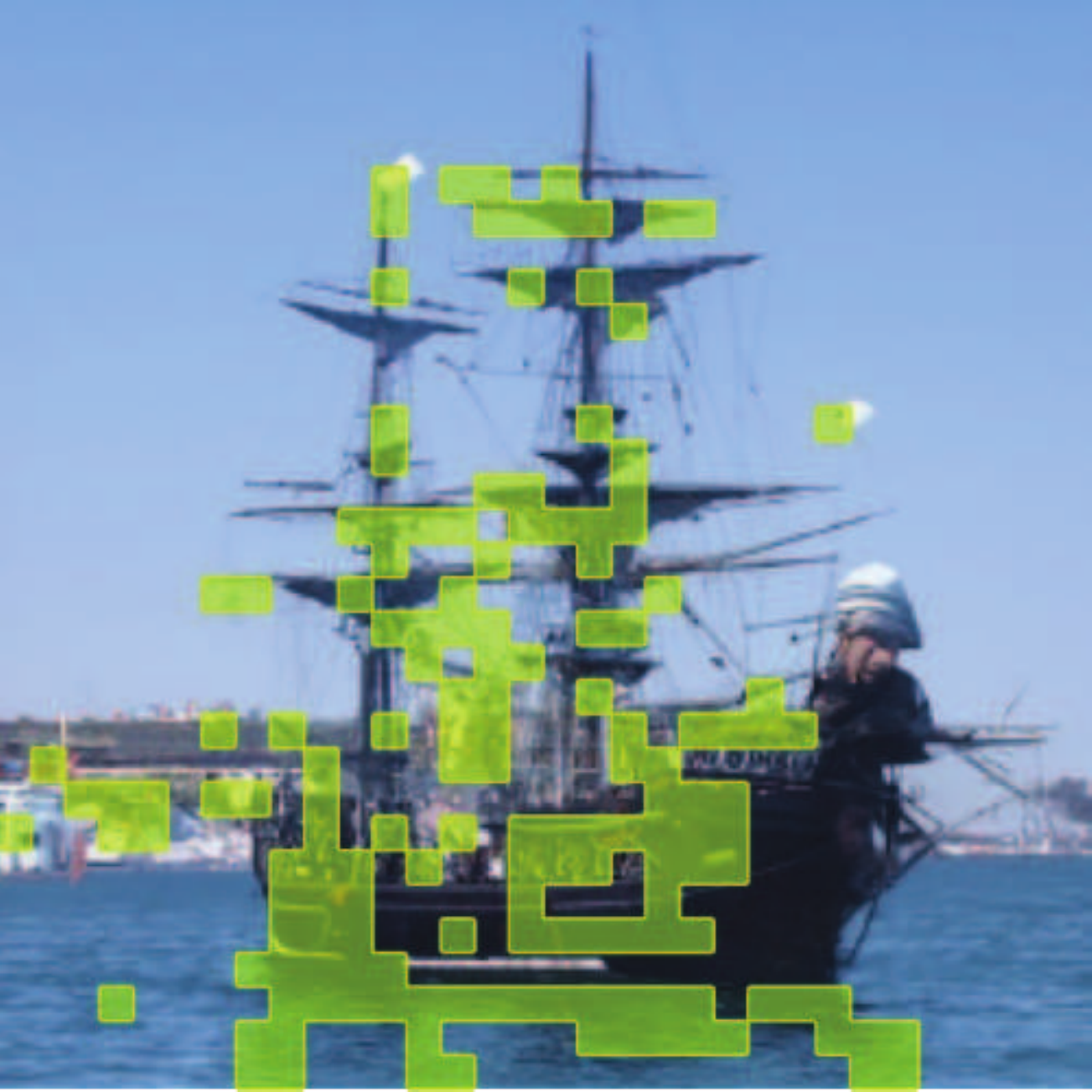}
  \caption*{Head 1}
\end{subfigure}
\begin{subfigure}{0.15\textwidth}
  \centering
  \includegraphics[width=1\linewidth]{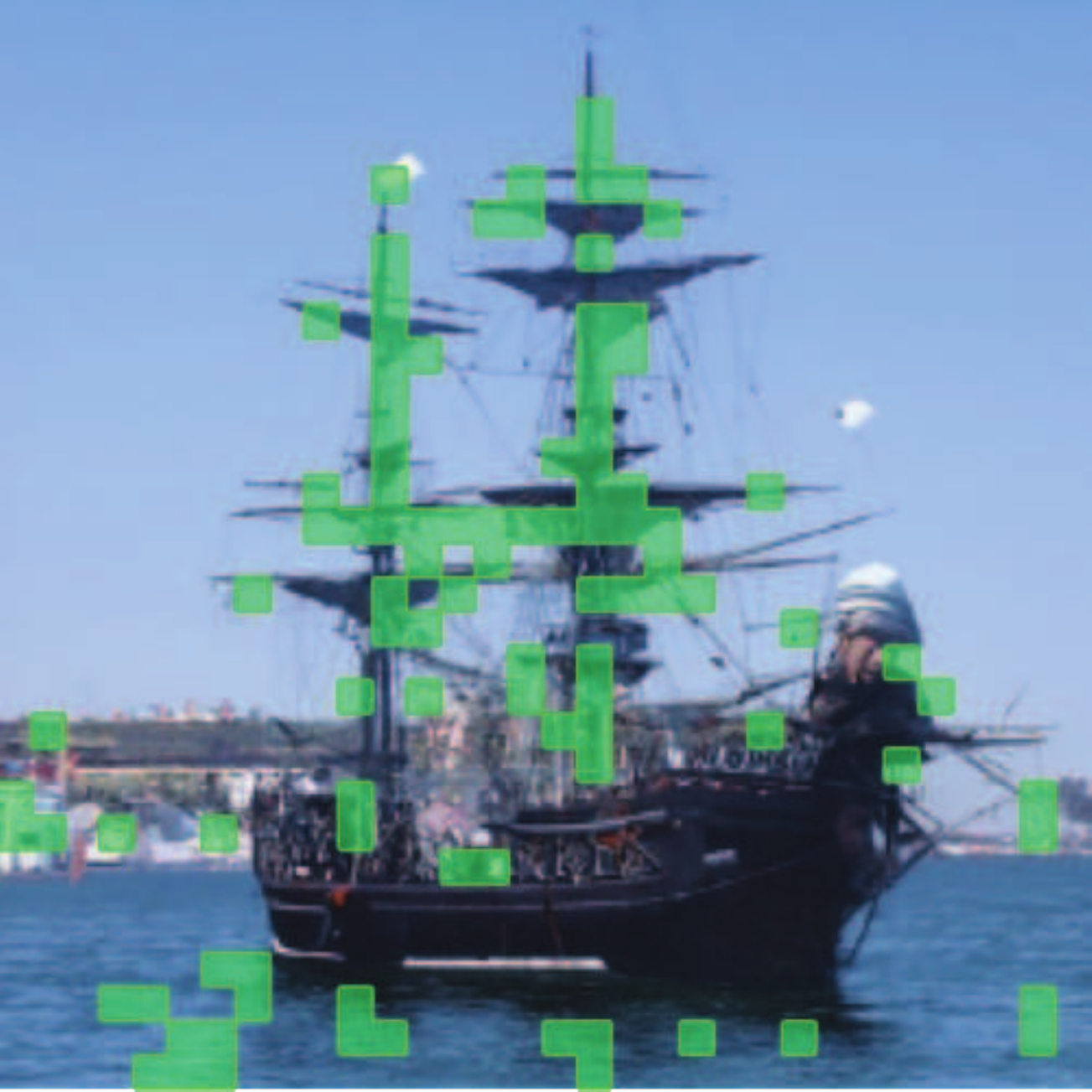}
  \caption*{Head 2}
\end{subfigure}
\begin{subfigure}{0.15\textwidth}
  \centering
  \includegraphics[width=1\linewidth]{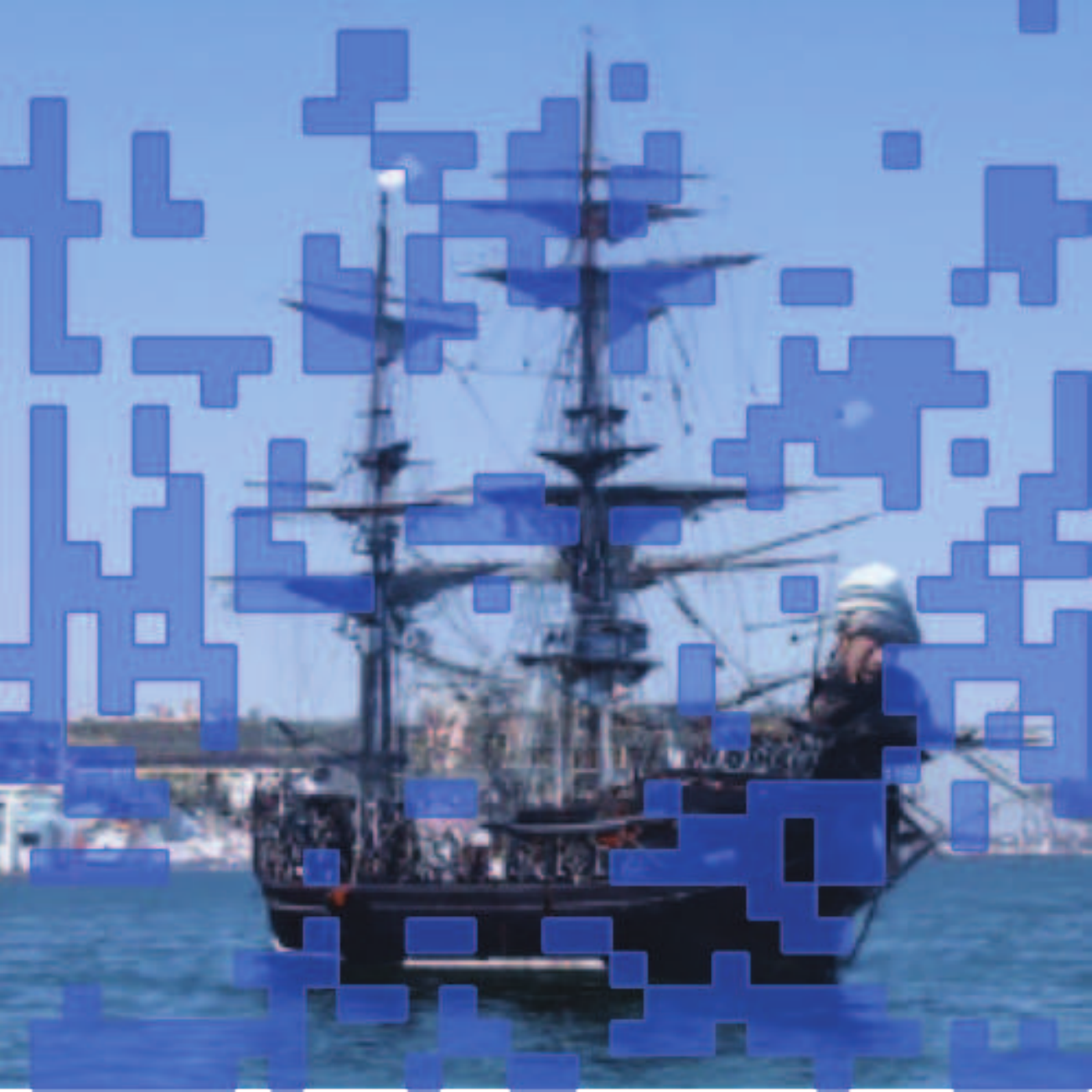}
  \caption*{Head 3}
\end{subfigure}
\begin{subfigure}{0.15\textwidth}
  \centering
  \includegraphics[width=1\linewidth]{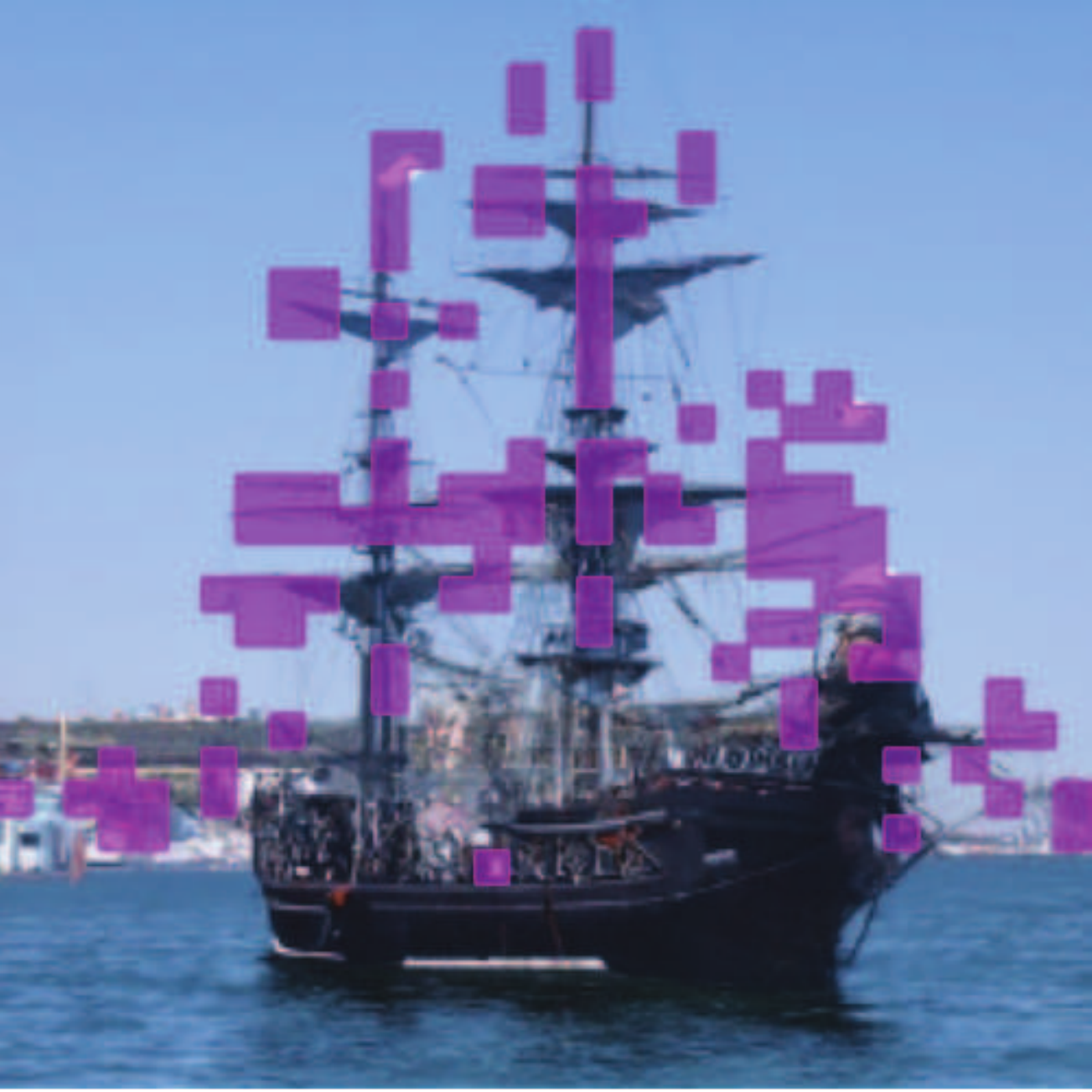}
  \caption*{Head 4}
\end{subfigure}
\\
\\
\begin{subfigure}{0.15\textwidth}
  \centering
  \includegraphics[width=1\textwidth]{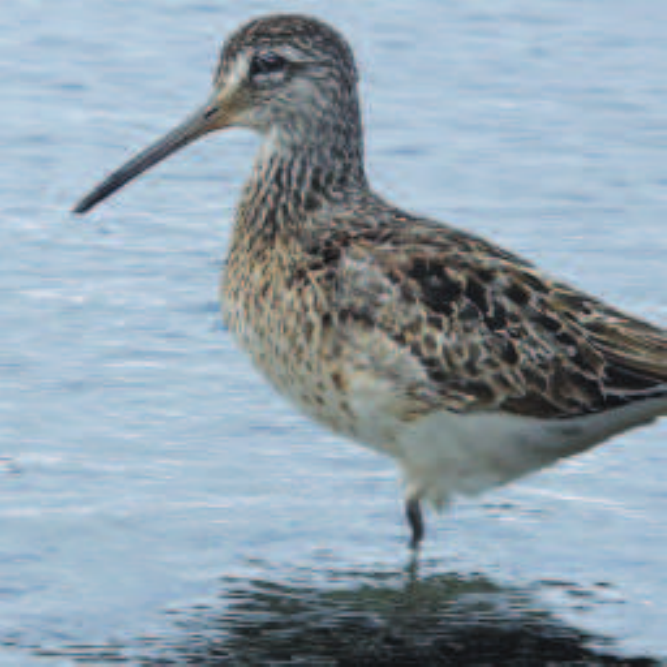}
\end{subfigure}
&
\begin{subfigure}{0.15\textwidth}
  \centering
  \includegraphics[width=1\linewidth]{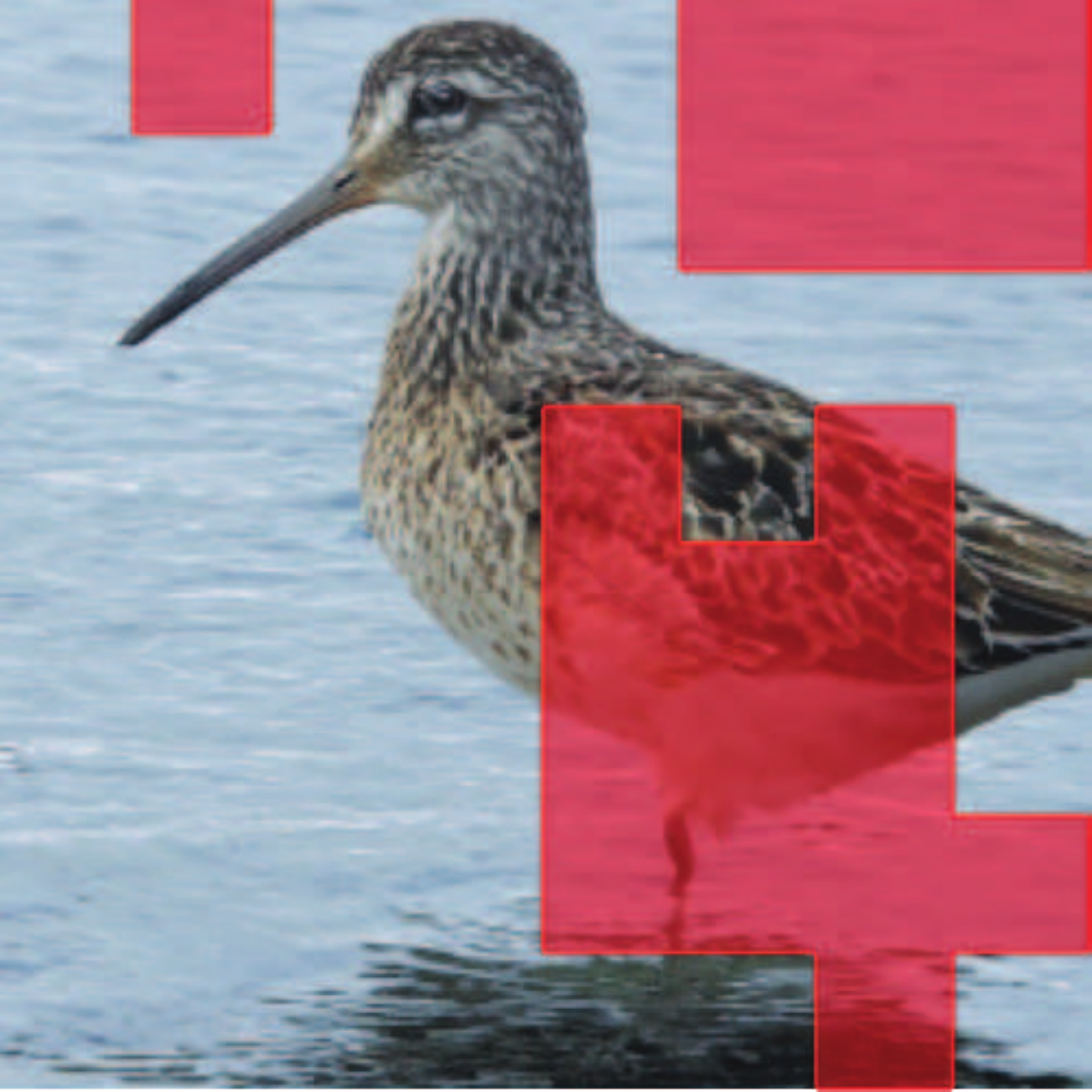}
\end{subfigure}
\begin{subfigure}{0.15\textwidth}
  \centering
  \includegraphics[width=1\linewidth]{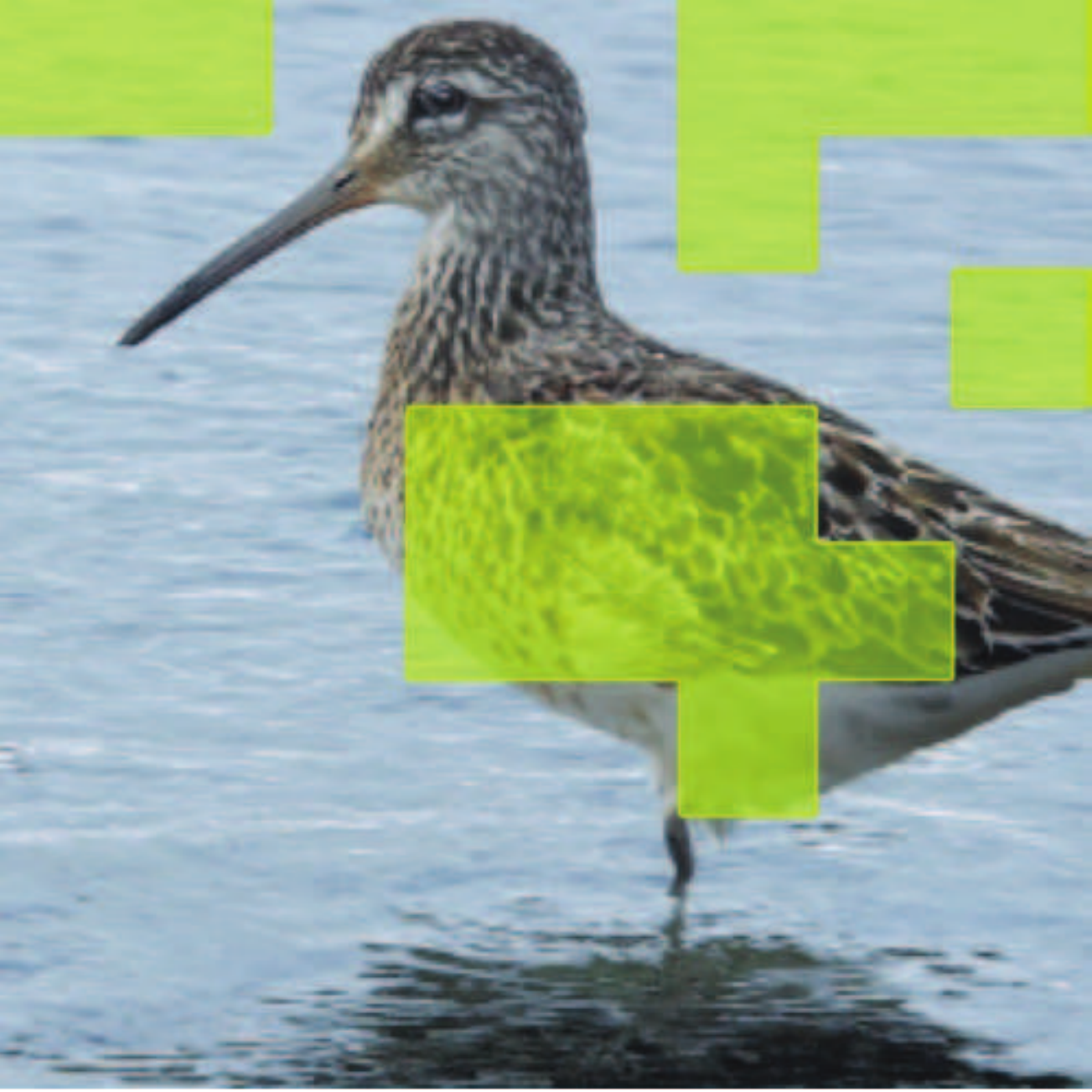}
\end{subfigure}
\begin{subfigure}{0.15\textwidth}
  \centering
  \includegraphics[width=1\linewidth]{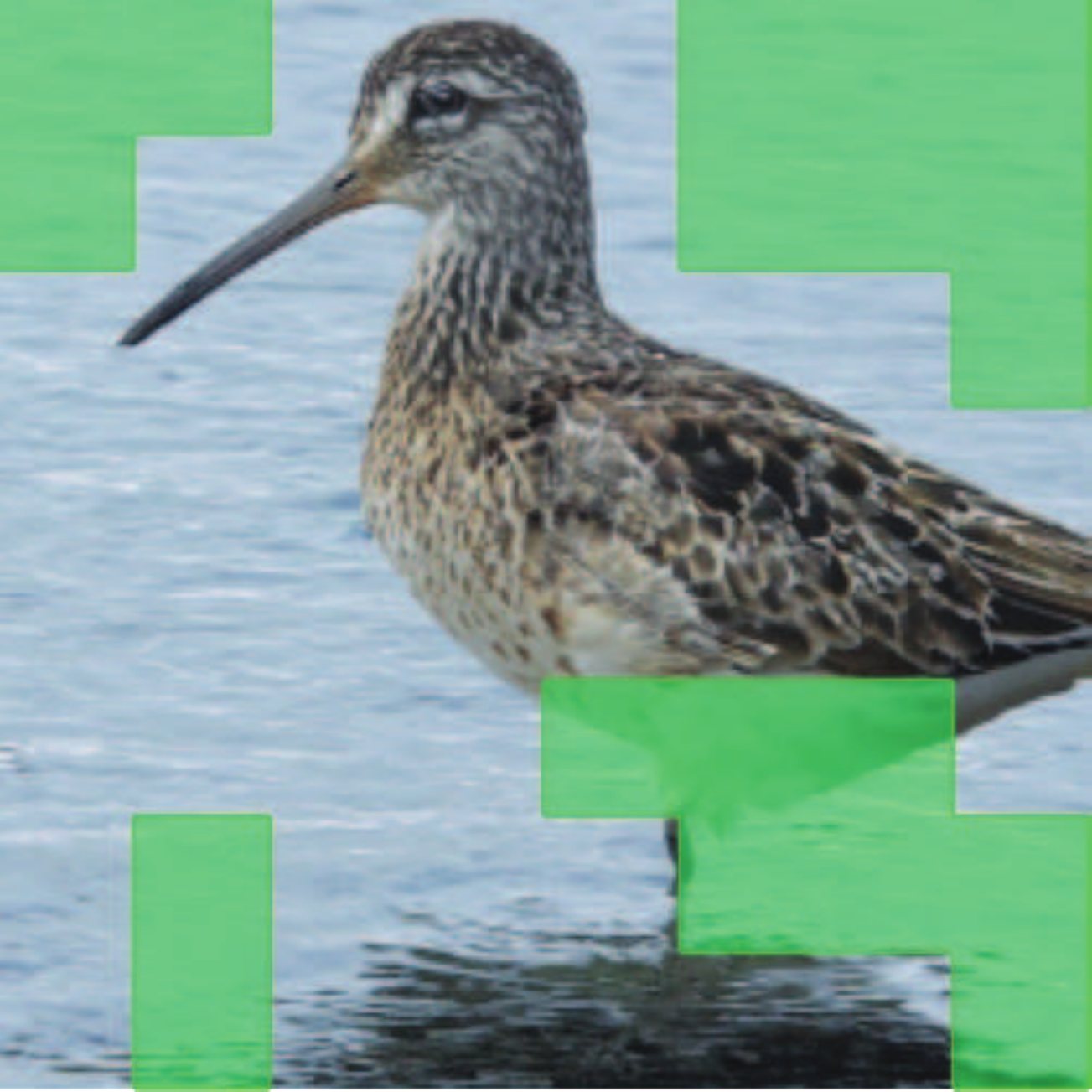}
\end{subfigure}
\begin{subfigure}{0.15\textwidth}
  \centering
  \includegraphics[width=1\linewidth]{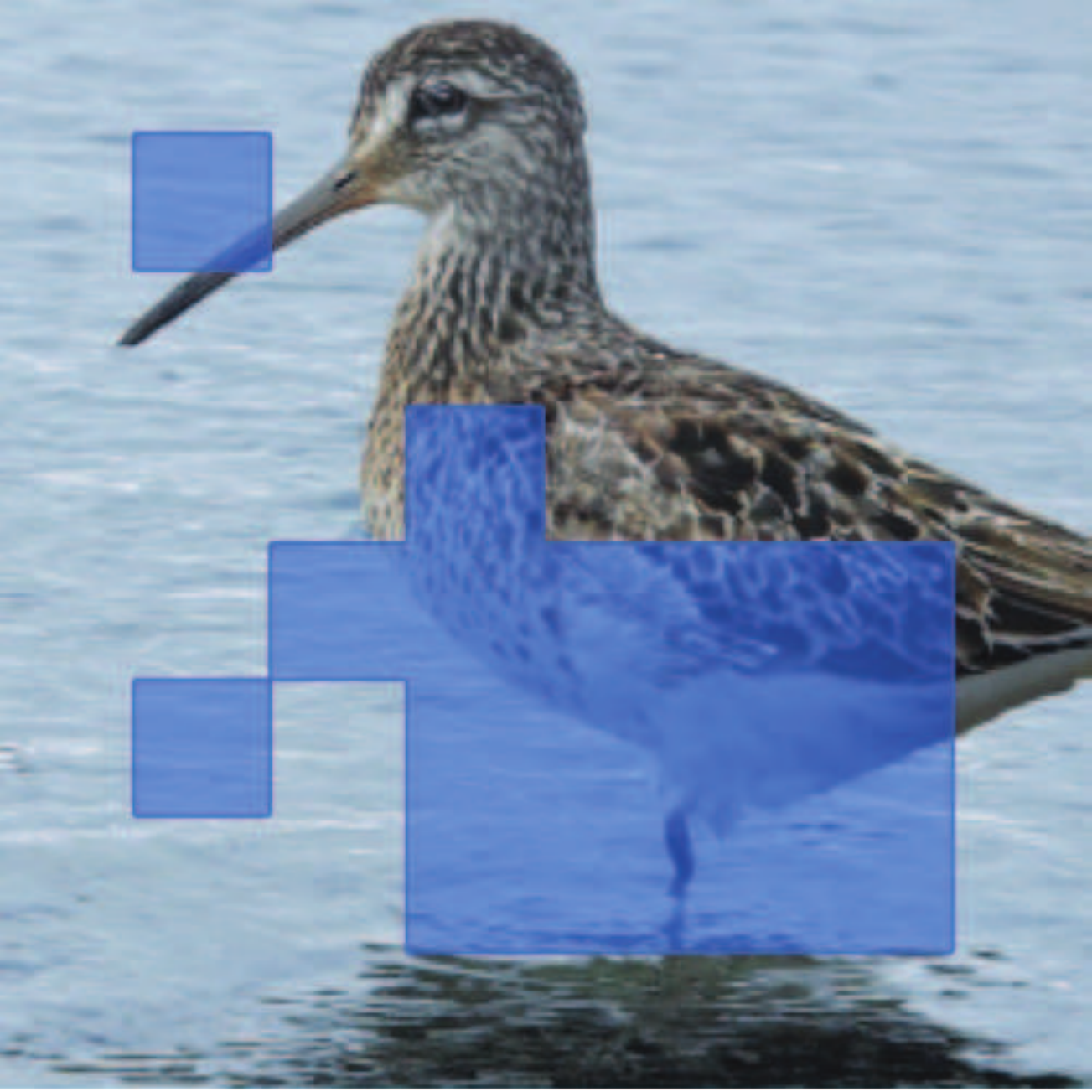}
\end{subfigure}
\begin{subfigure}{0.15\textwidth}
  \centering
  \includegraphics[width=1\linewidth]{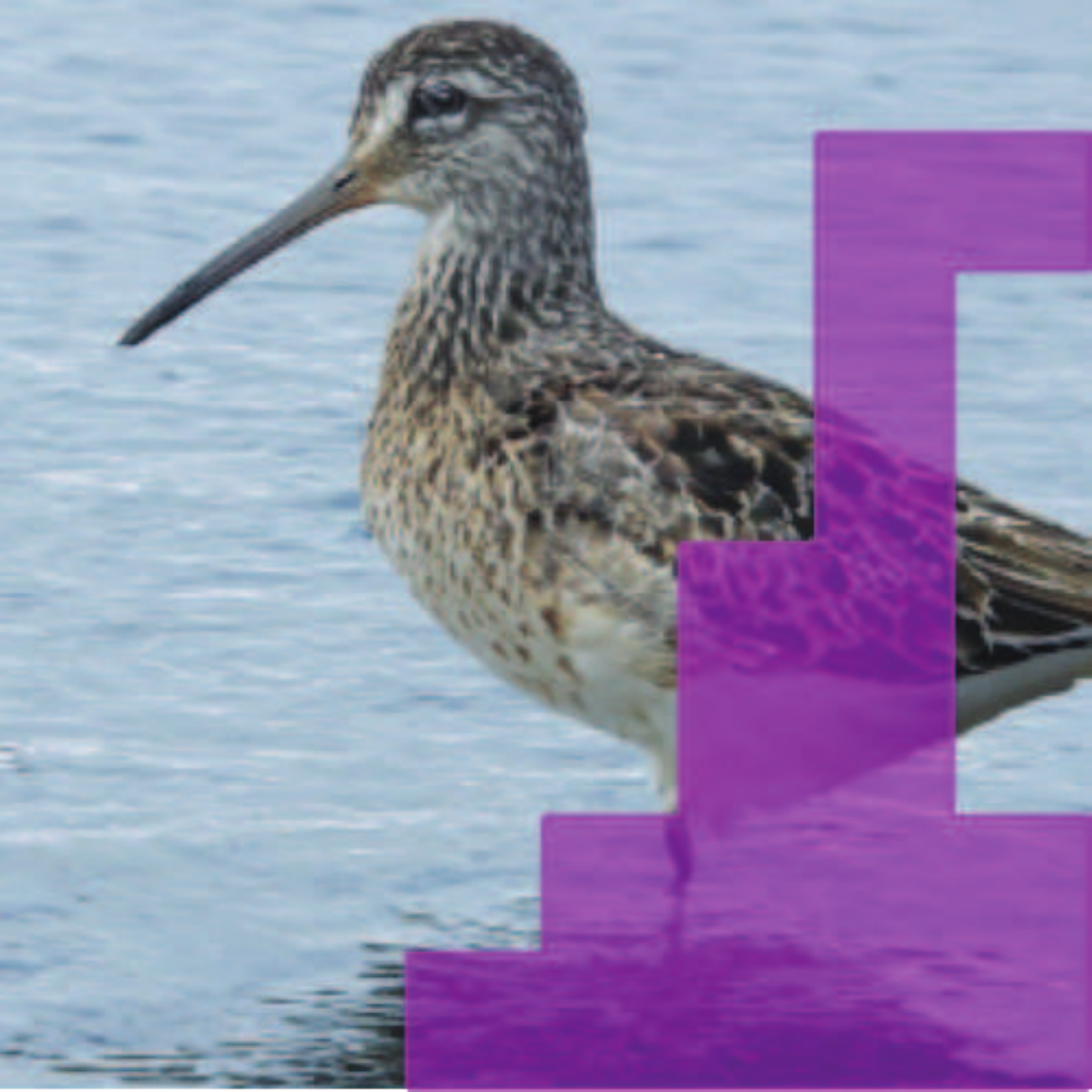}
\end{subfigure}\\
&
\begin{subfigure}{0.15\textwidth}
  \centering
  \includegraphics[width=1\linewidth]{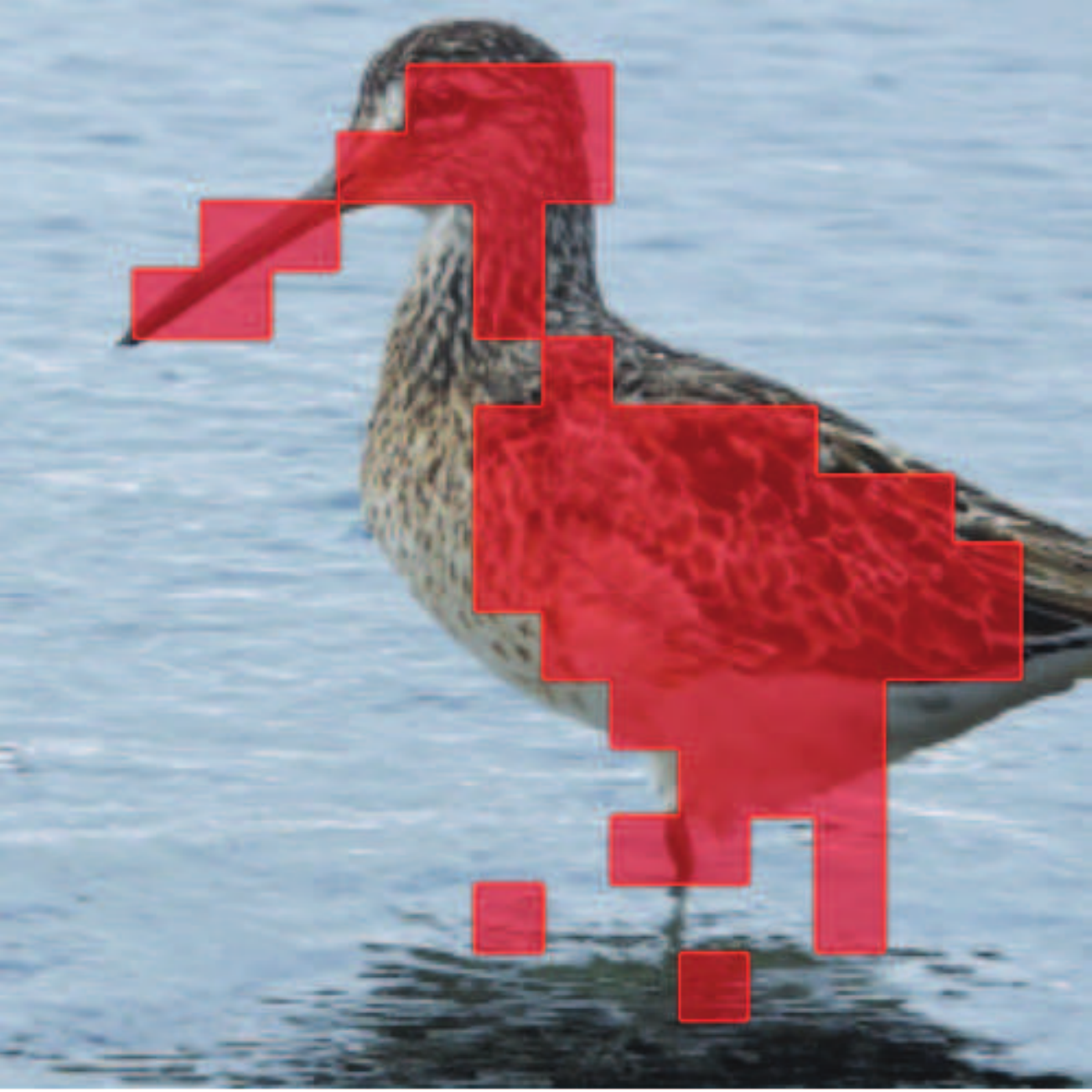}
\end{subfigure}
\begin{subfigure}{0.15\textwidth}
  \centering
  \includegraphics[width=1\linewidth]{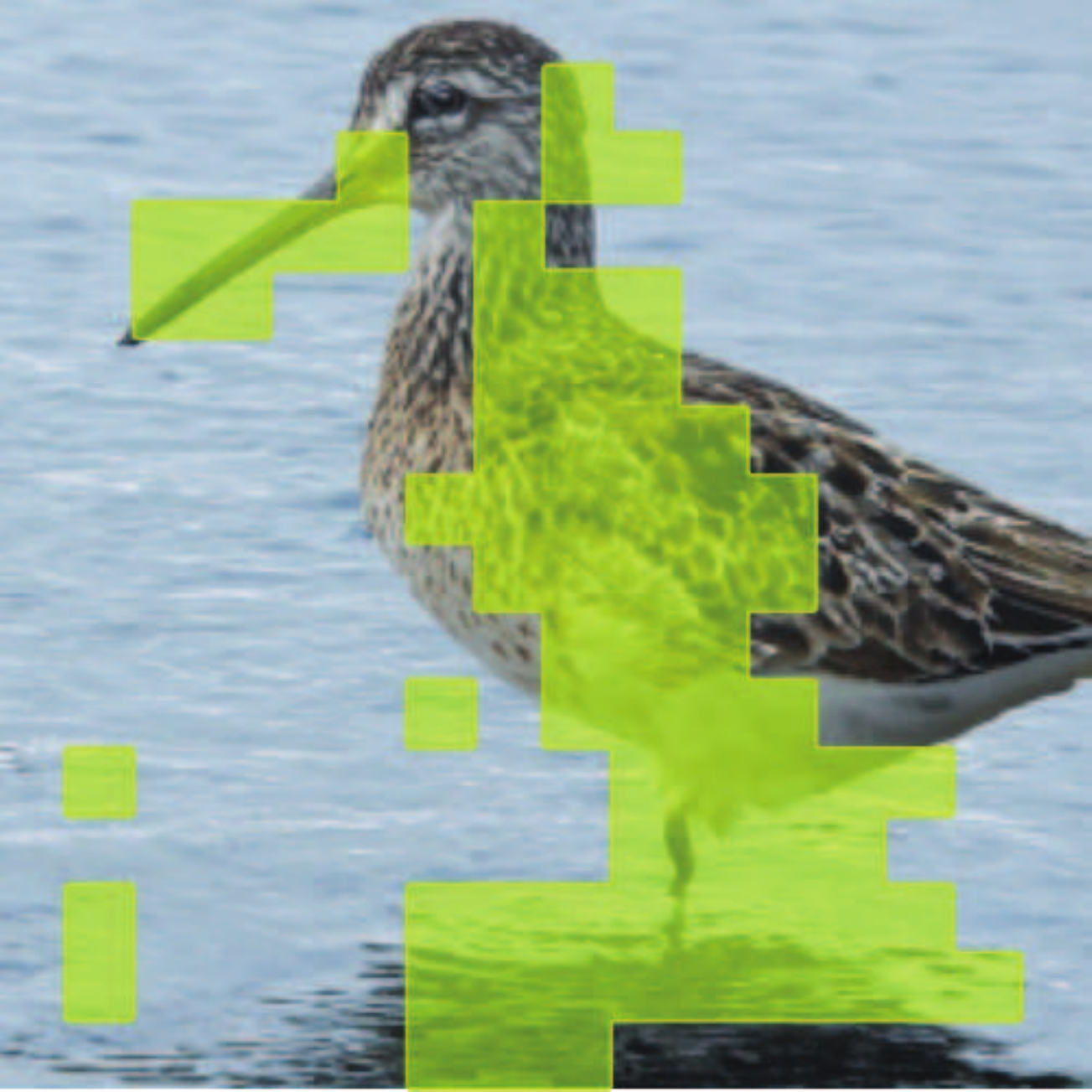}
\end{subfigure}
\begin{subfigure}{0.15\textwidth}
  \centering
  \includegraphics[width=1\linewidth]{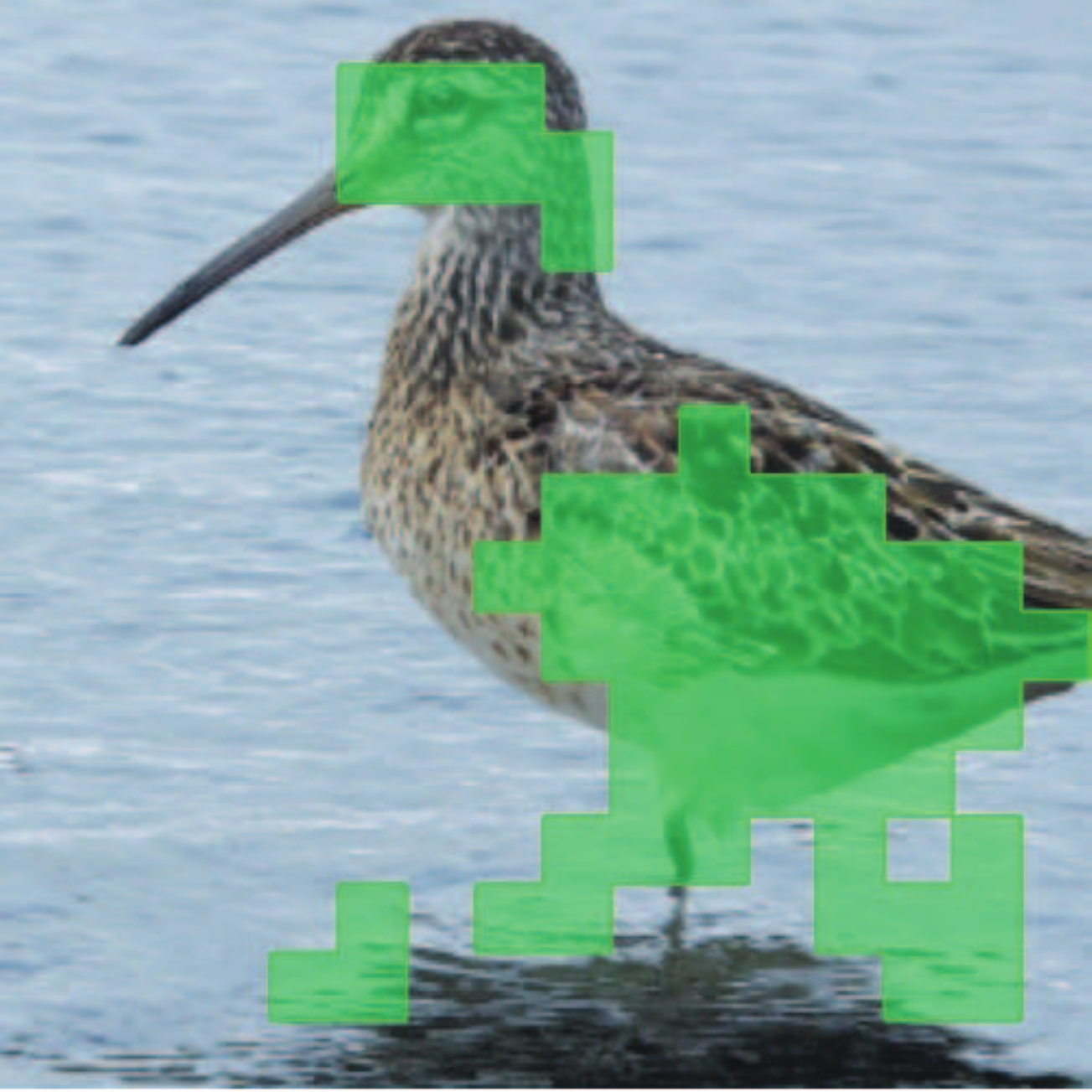}
\end{subfigure}
\begin{subfigure}{0.15\textwidth}
  \centering
  \includegraphics[width=1\linewidth]{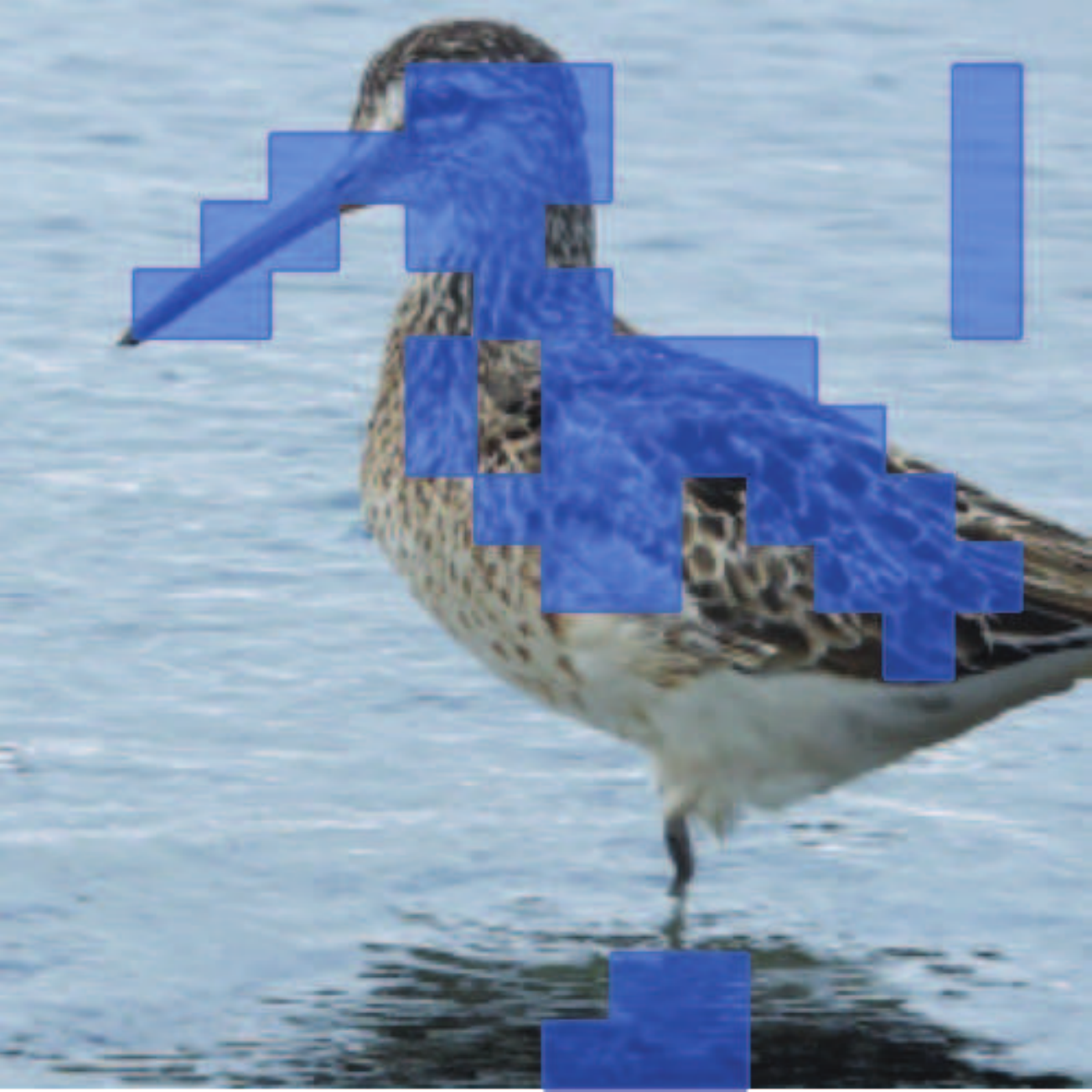}
\end{subfigure}
\begin{subfigure}{0.15\textwidth}
  \centering
  \includegraphics[width=1\linewidth]{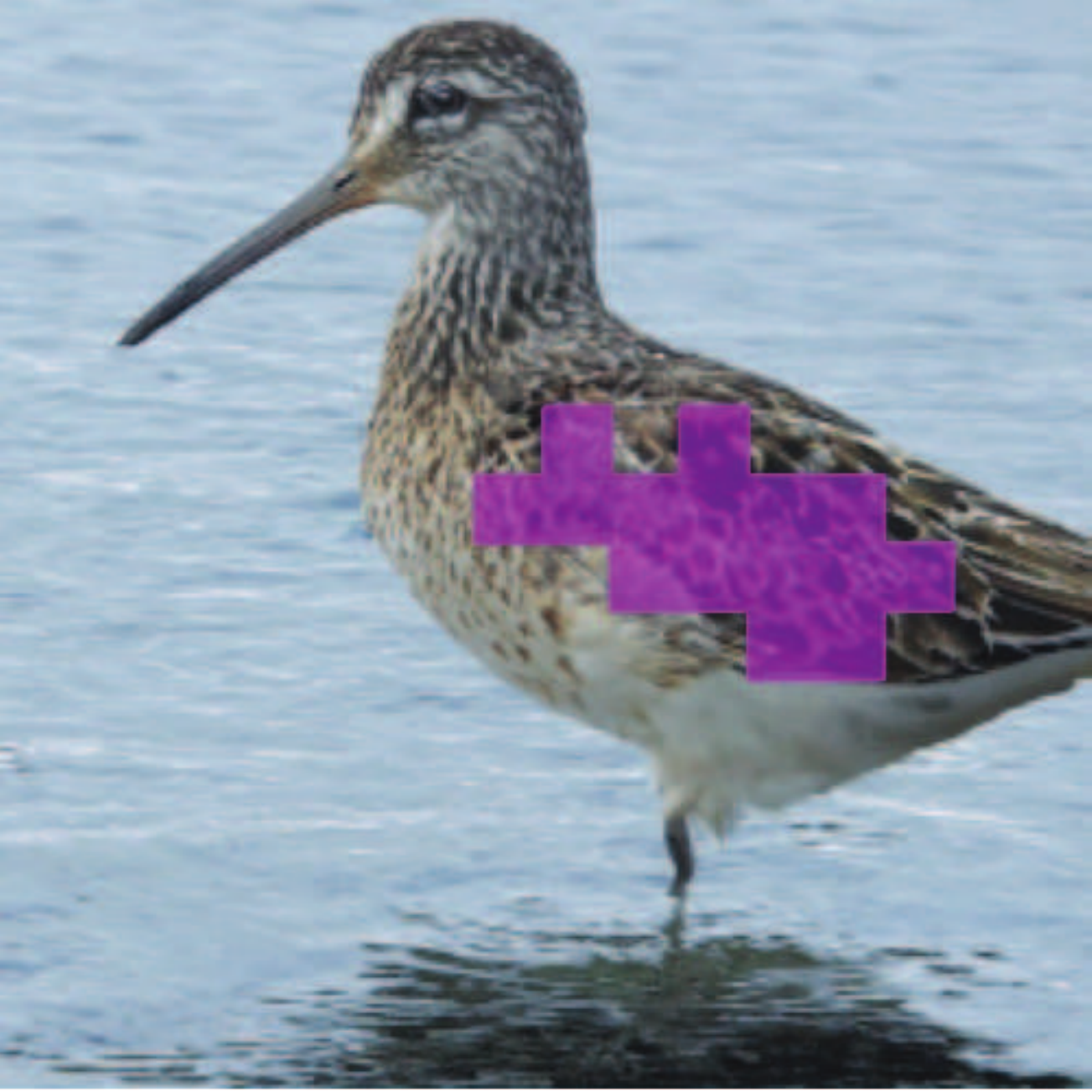}
\end{subfigure}\\
&
\begin{subfigure}{0.15\textwidth}
  \centering
  \includegraphics[width=1\linewidth]{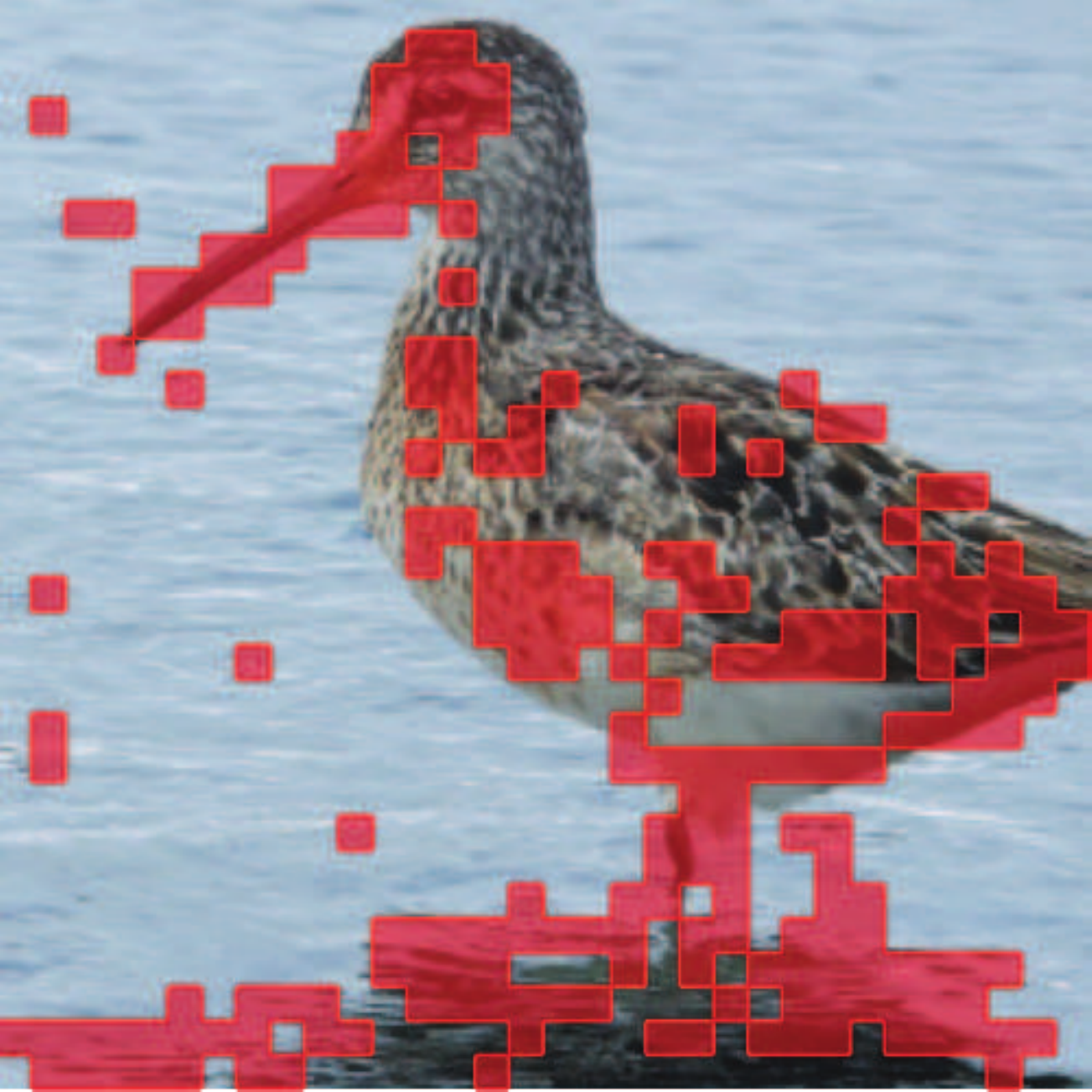}
  \caption*{Averaged}
\end{subfigure}
\begin{subfigure}{0.15\textwidth}
  \centering
  \includegraphics[width=1\linewidth]{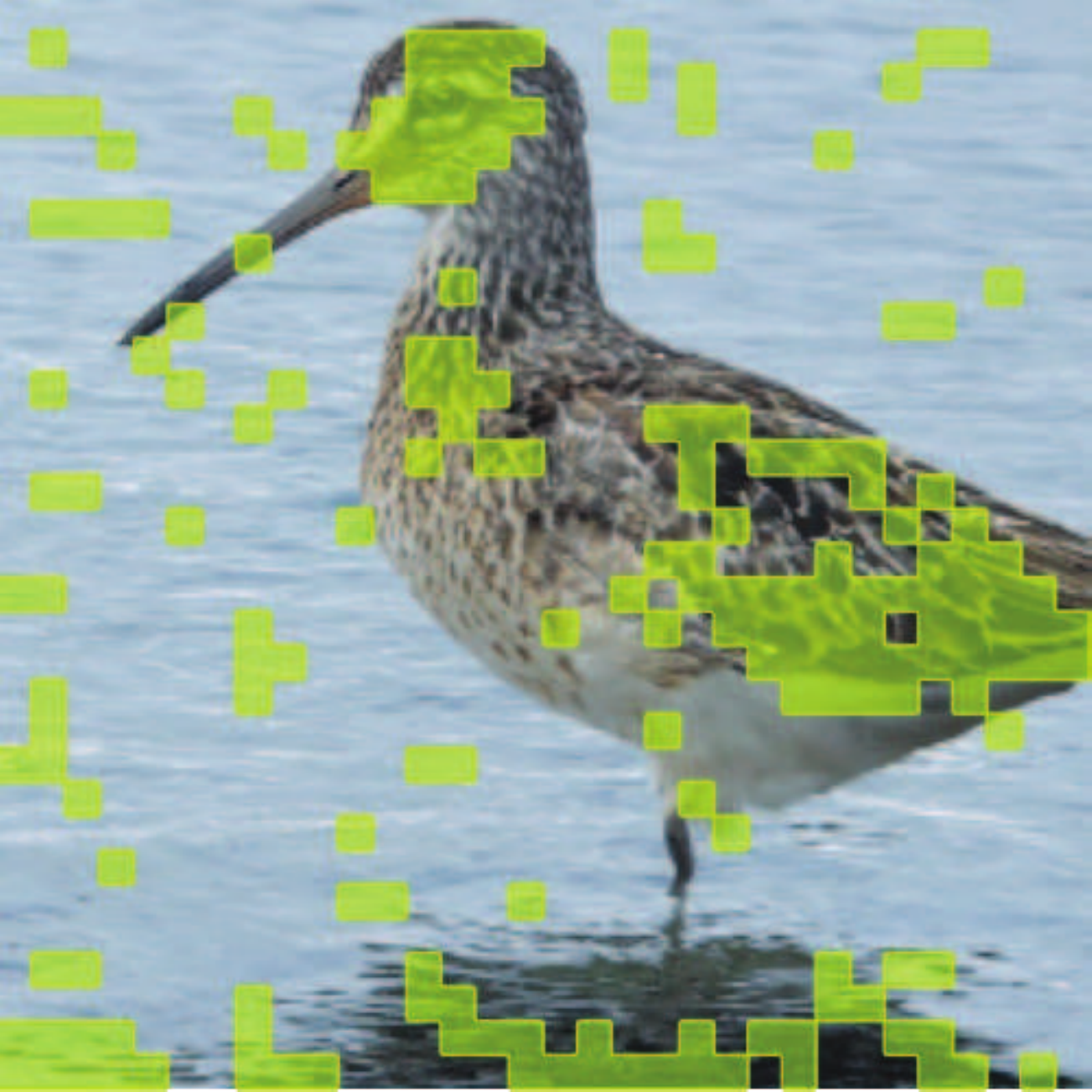}
  \caption*{Head 1}
\end{subfigure}
\begin{subfigure}{0.15\textwidth}
  \centering
  \includegraphics[width=1\linewidth]{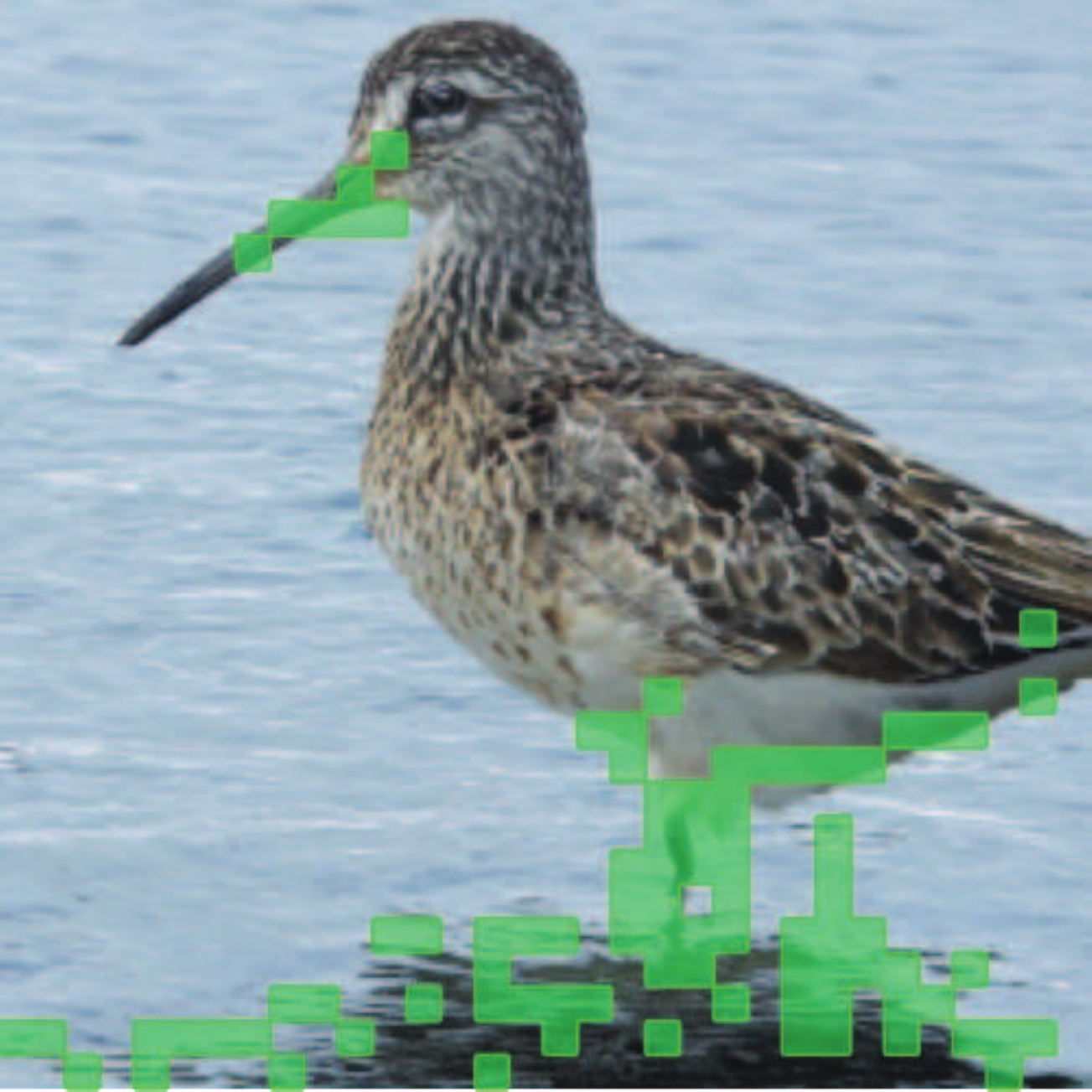}
  \caption*{Head 2}
\end{subfigure}
\begin{subfigure}{0.15\textwidth}
  \centering
  \includegraphics[width=1\linewidth]{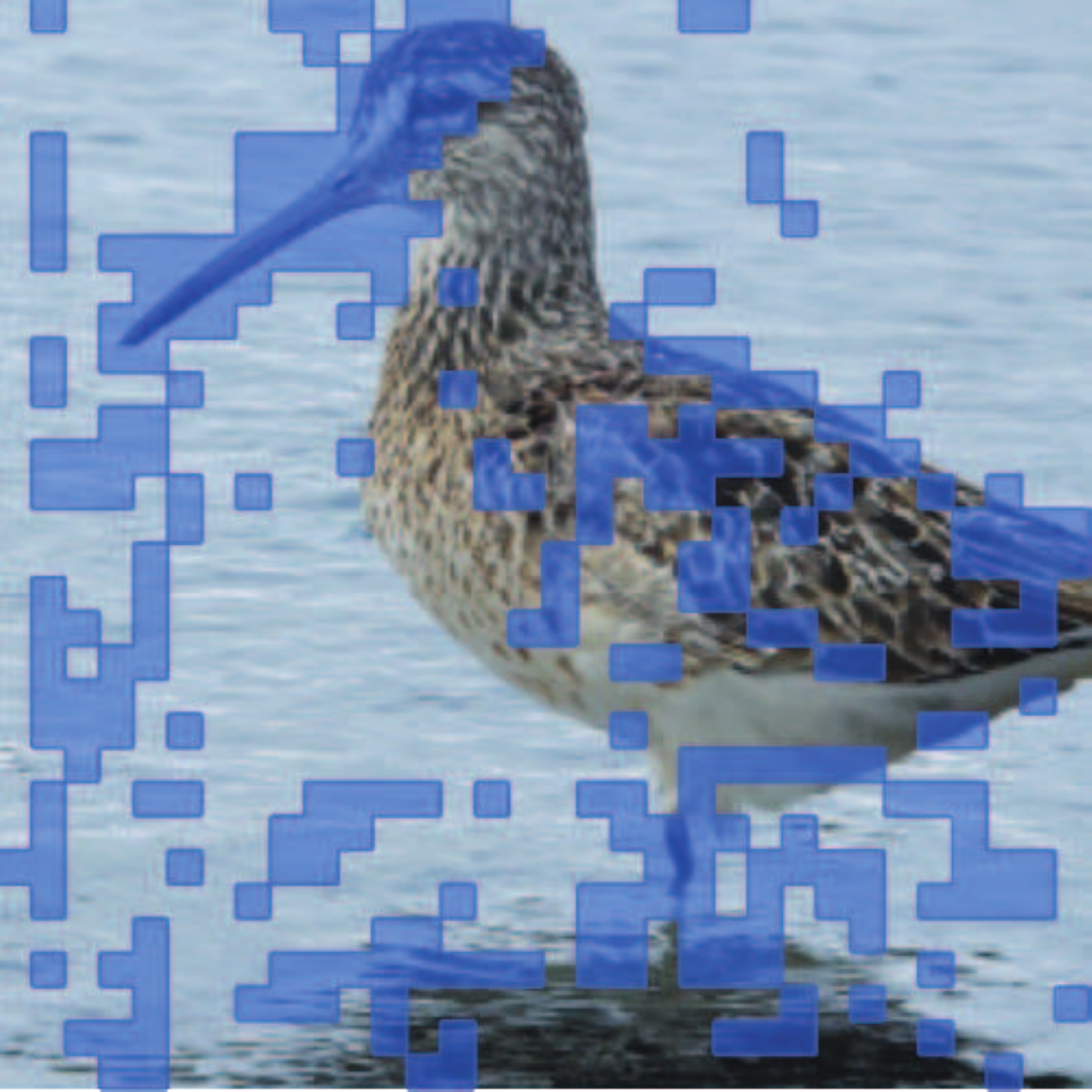}
  \caption*{Head 3}
\end{subfigure}
\begin{subfigure}{0.15\textwidth}
  \centering
  \includegraphics[width=1\linewidth]{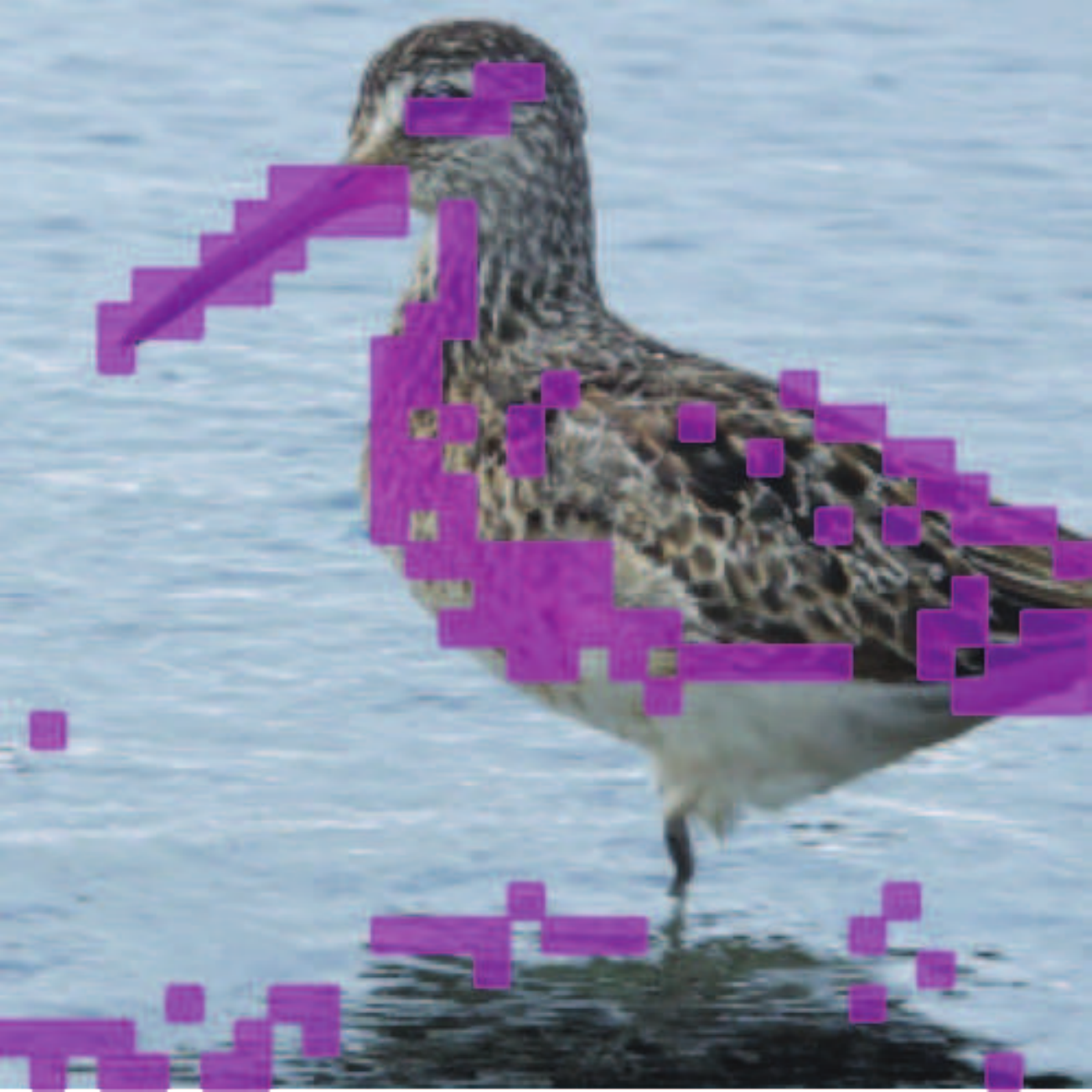}
  \caption*{Head 4}
\end{subfigure}\\
\end{tabular}
\vspace{-5pt}
\caption{\textbf{Visualization of self-attention masks from different layers and heads}. Each row, top to bottom, corresponds to the $8\times8$, $16\times16$ and $32\times32$ self-attention layers, respectively.}
\label{fig:head2}
\end{figure*}

We show the visualizations of self-attention maps in the 8$\times$8, 16$\times$16, and 32$\times$32 resolutions of the U-Net~\cite{ronneberger2015u} of ADM~\cite{dhariwal2021diffusion} in Fig.~\ref{fig:timestep}. The attention maps at $t=0, 49, 99, 149, 199, 249$ are visualized at each row in order, and the layers are aligned left to right. In this visualization, can see that the attention maps at the intermediate timesteps capture the structure of generated images. Also, we extract the self-attention masks from the different heads and layers from the U-Net and visualize them in Fig.~\ref{fig:head1} and Fig.~\ref{fig:head2}. \textit{Average} in this figure means the obtained masks after averaging attention maps of the four heads. Moreover, we compare the self-attention masks of ADM with those of DINO~\cite{caron2021emerging} in Fig.~\ref{fig:dino-vis}. Compared to the attention masks of DINO, those of ADM are more attending to multiple objects and high-frequency details of the generated images where diffusion models have to elaborate.

\begin{figure}[t]
\centering
\begin{subfigure}{.09\textwidth}
  \centering
  \includegraphics[width=1.0\linewidth]{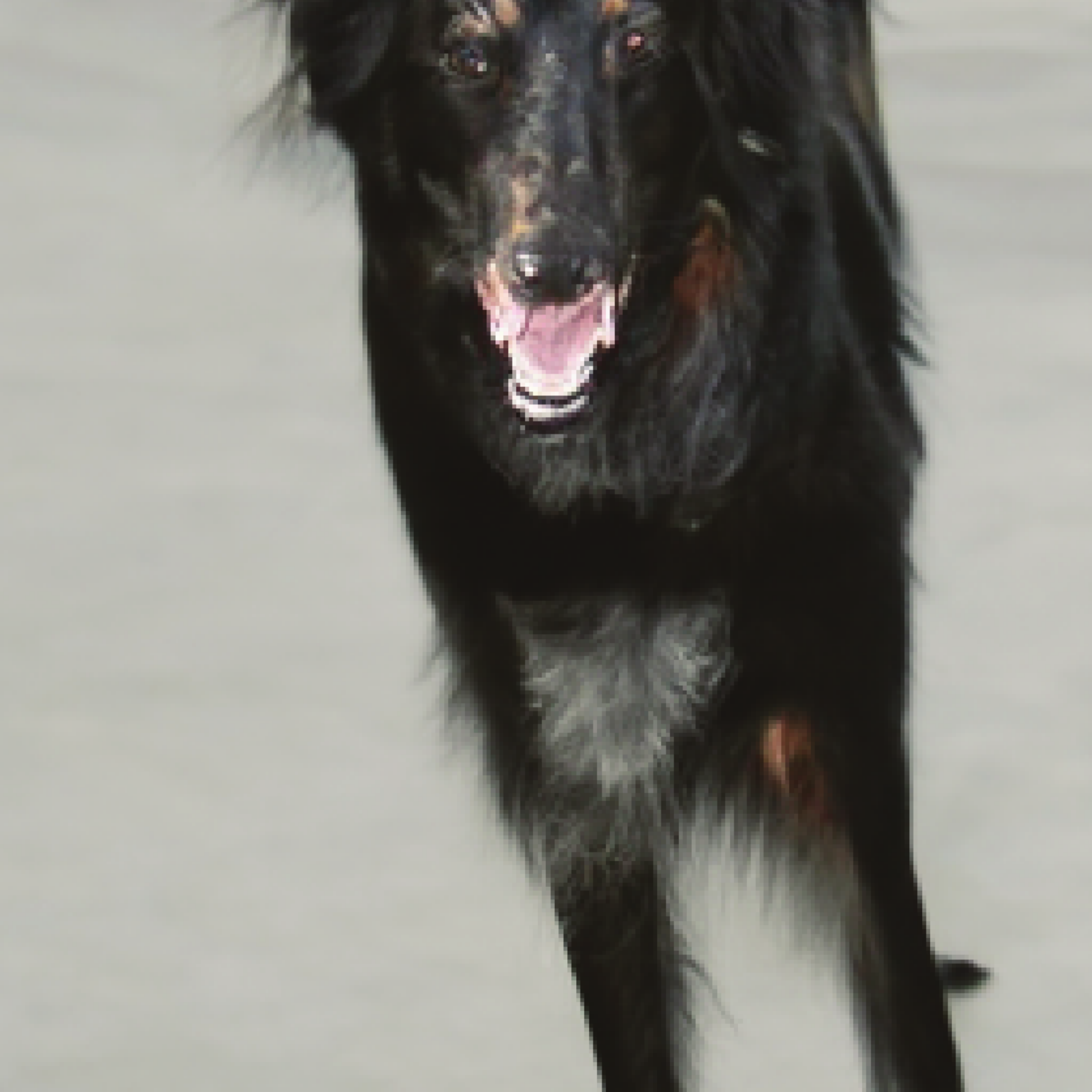}
\end{subfigure}
\begin{subfigure}{.09\textwidth}
  \centering
  \includegraphics[width=1.0\linewidth]{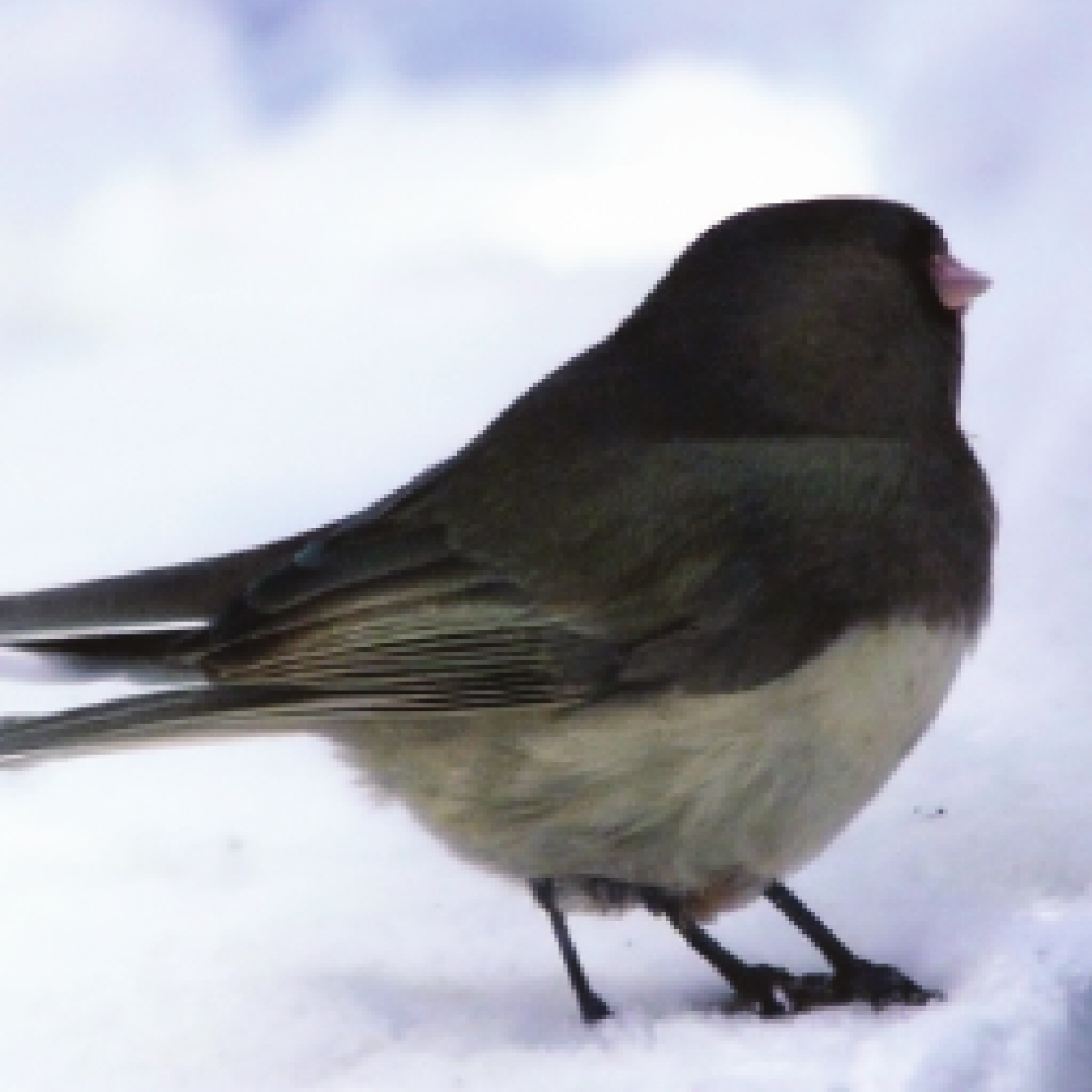}
\end{subfigure}
\begin{subfigure}{.09\textwidth}
  \centering
  \includegraphics[width=1.0\linewidth]{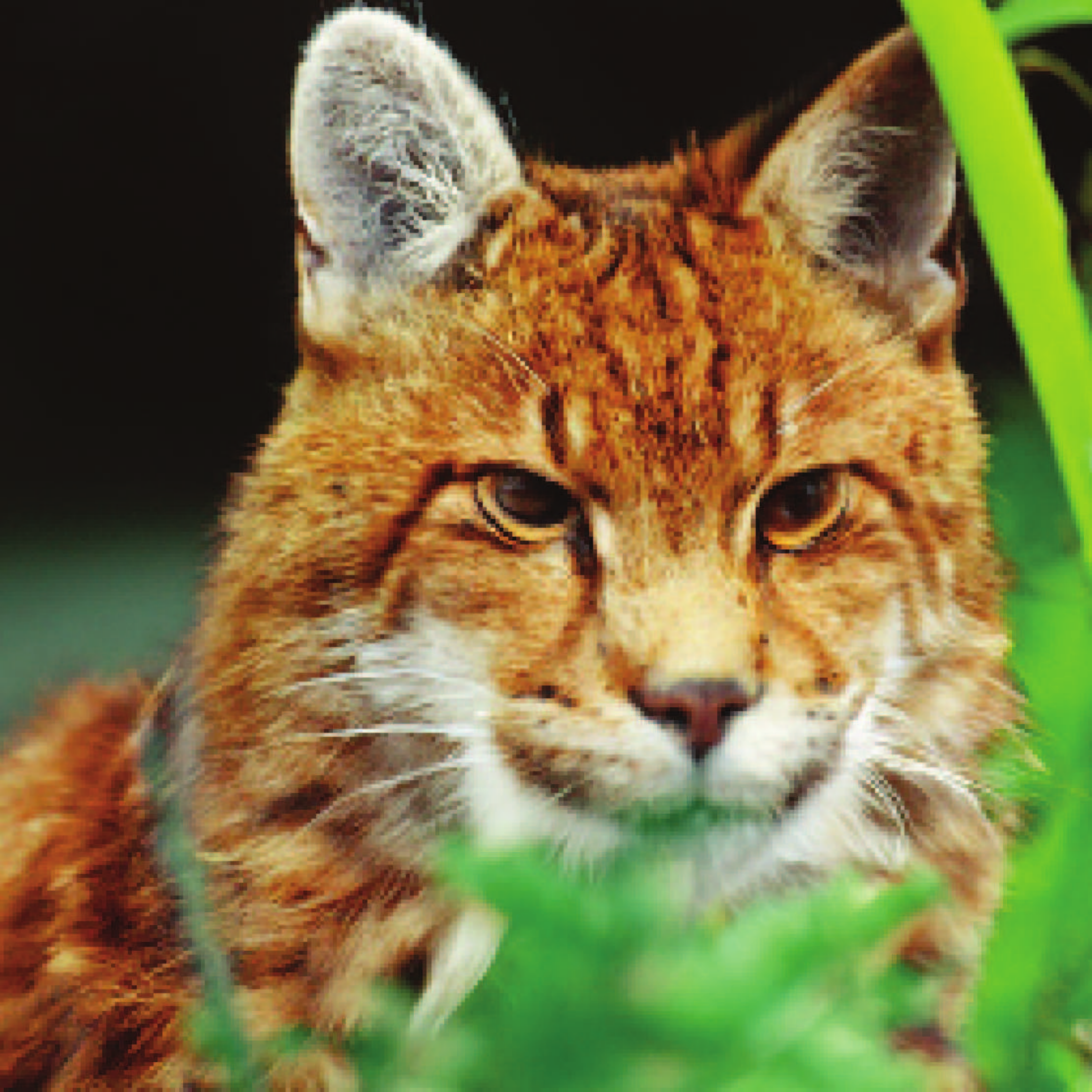}
\end{subfigure}
\begin{subfigure}{.09\textwidth}
  \centering
  \includegraphics[width=1.0\linewidth]{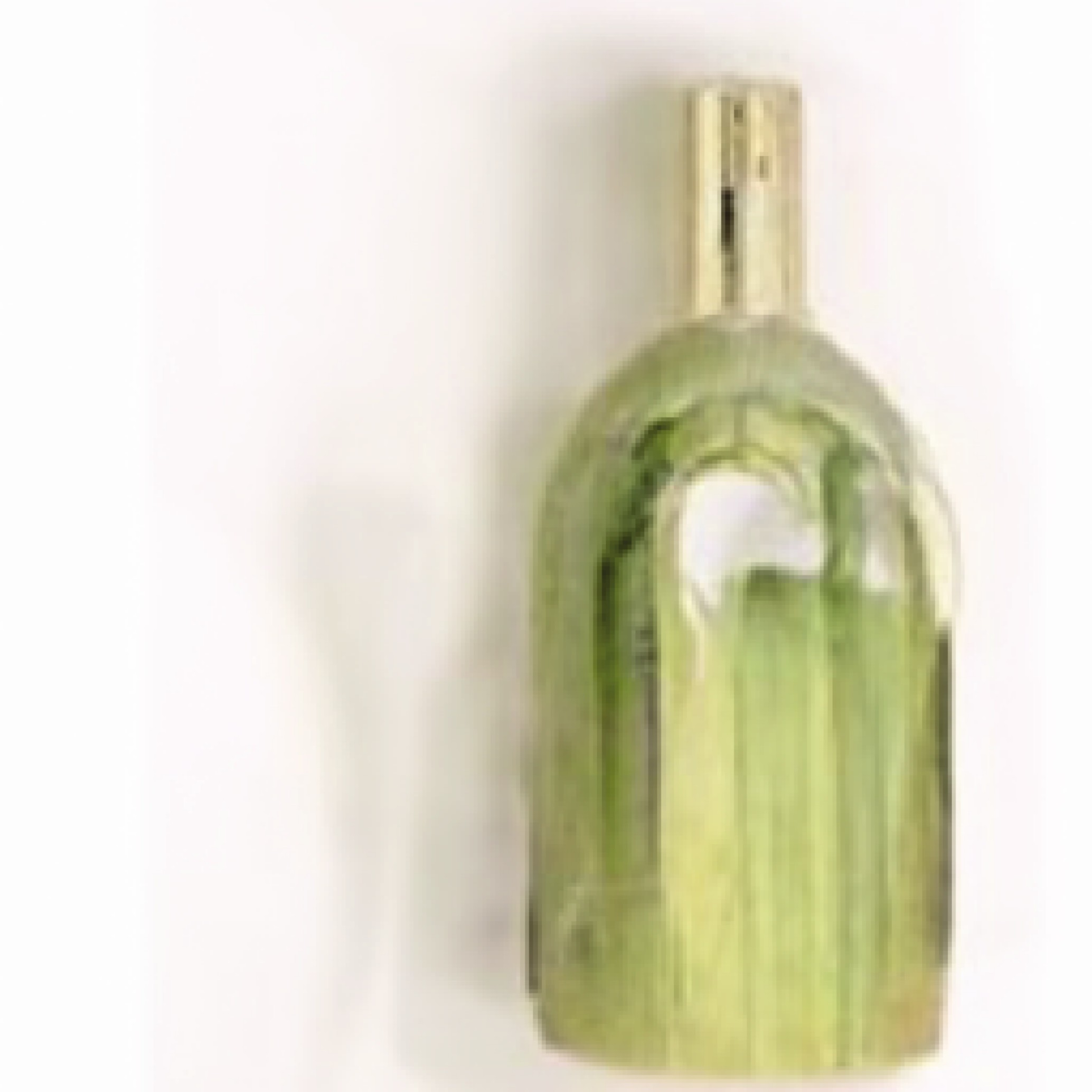}
\end{subfigure}
\begin{subfigure}{.09\textwidth}
  \centering
  \includegraphics[width=1.0\linewidth]{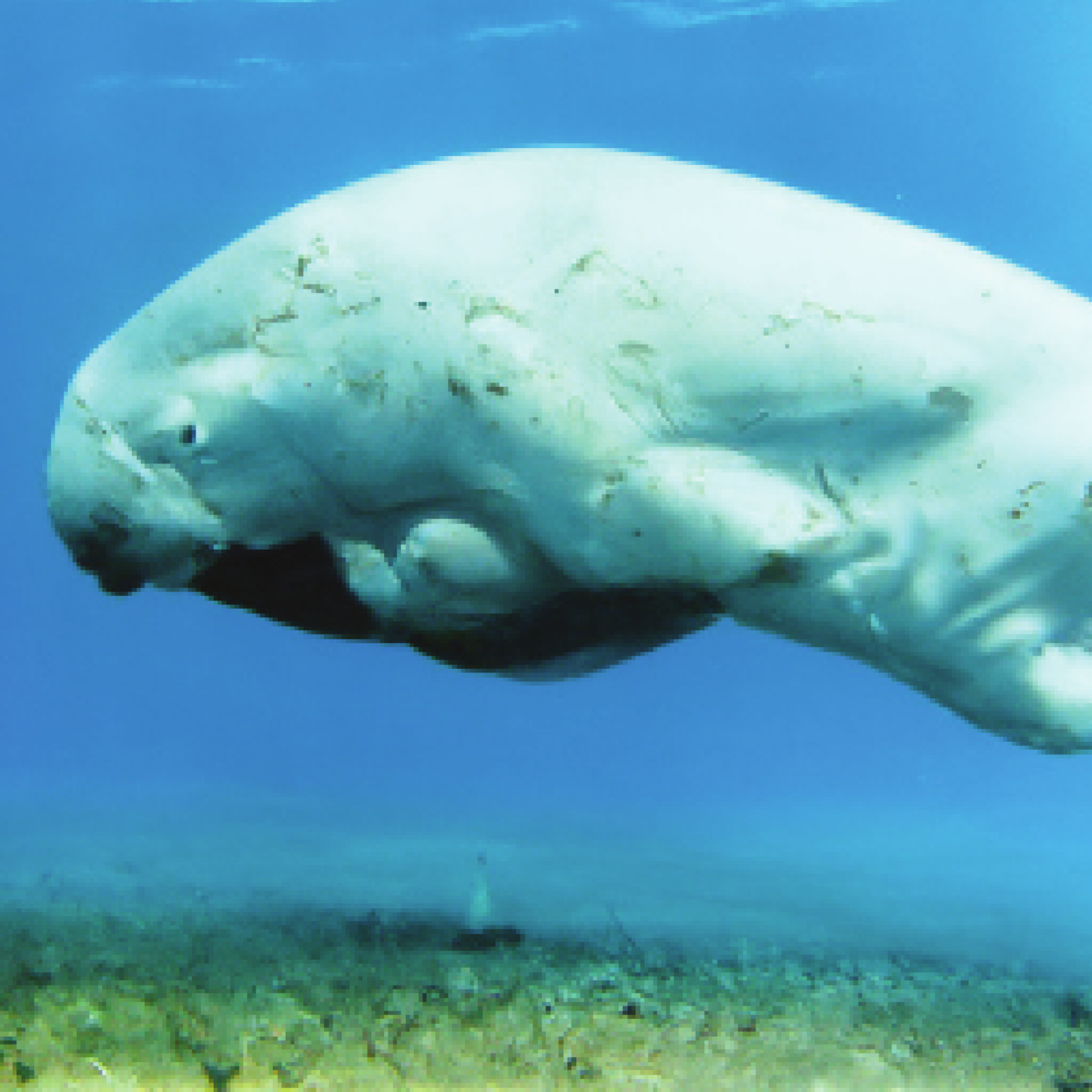}
\end{subfigure} \\
\begin{subfigure}{.09\textwidth}
  \centering
  \includegraphics[width=1.0\linewidth]{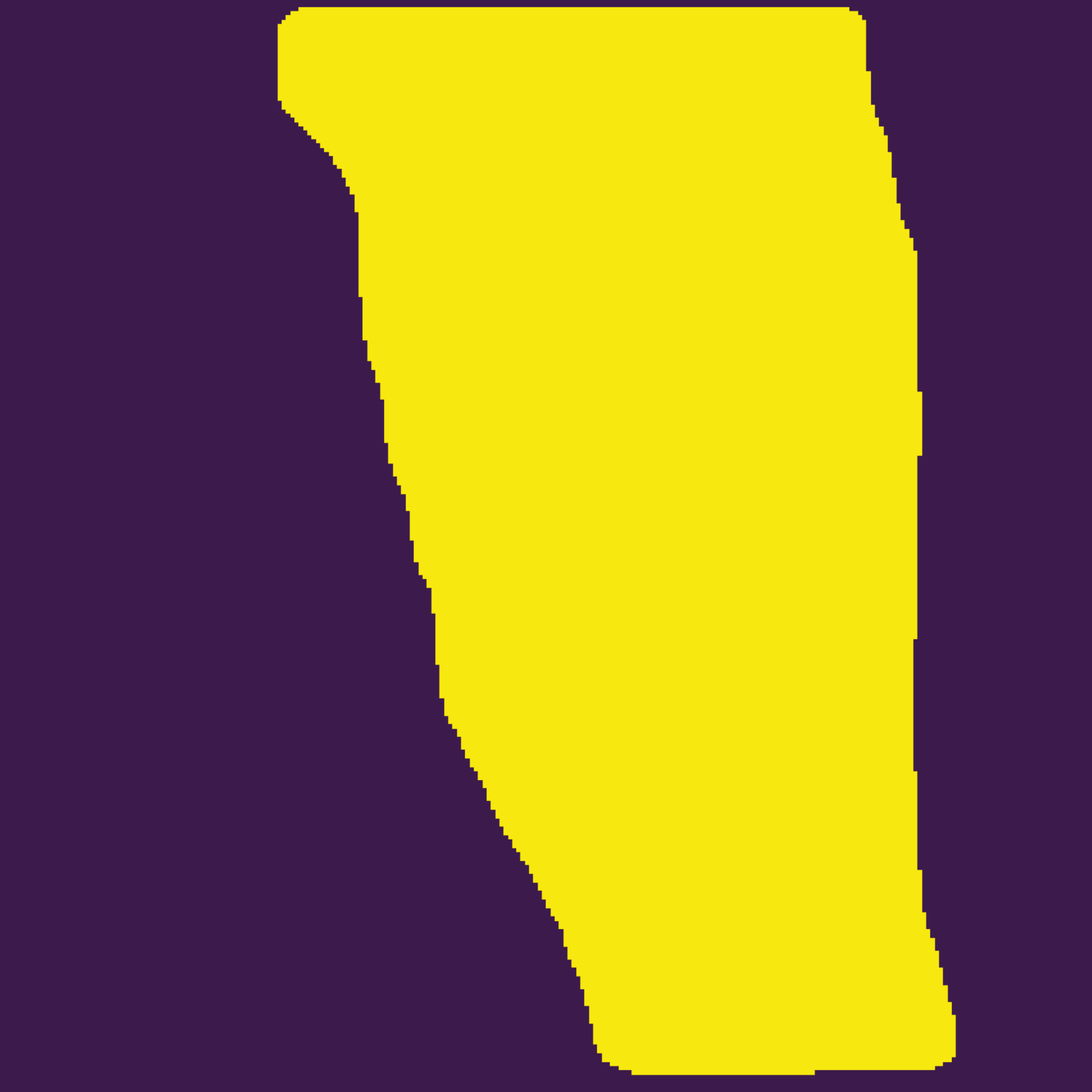}
\end{subfigure}
\begin{subfigure}{.09\textwidth}
  \centering
  \includegraphics[width=1.0\linewidth]{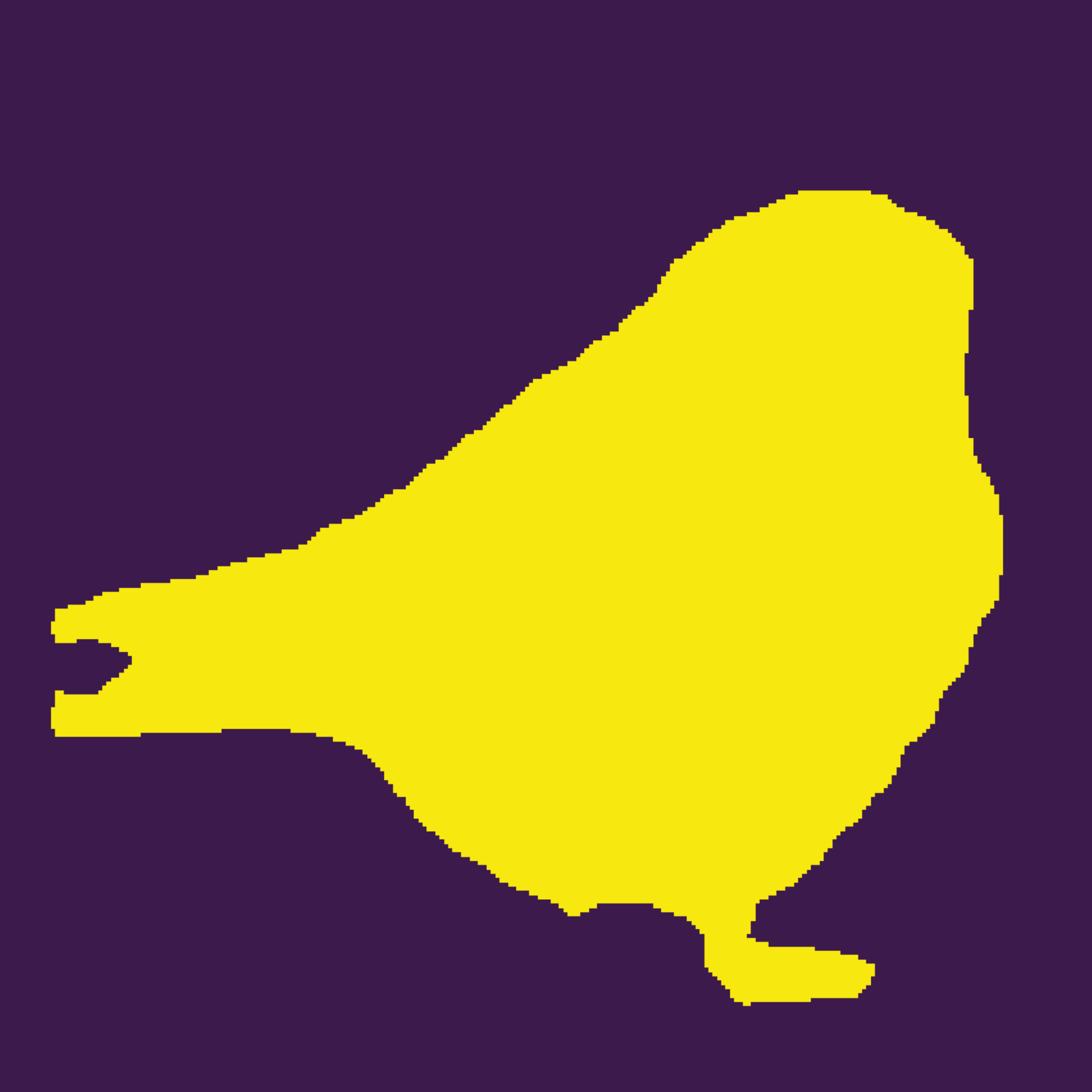}
\end{subfigure}
\begin{subfigure}{.09\textwidth}
  \centering
  \includegraphics[width=1.0\linewidth]{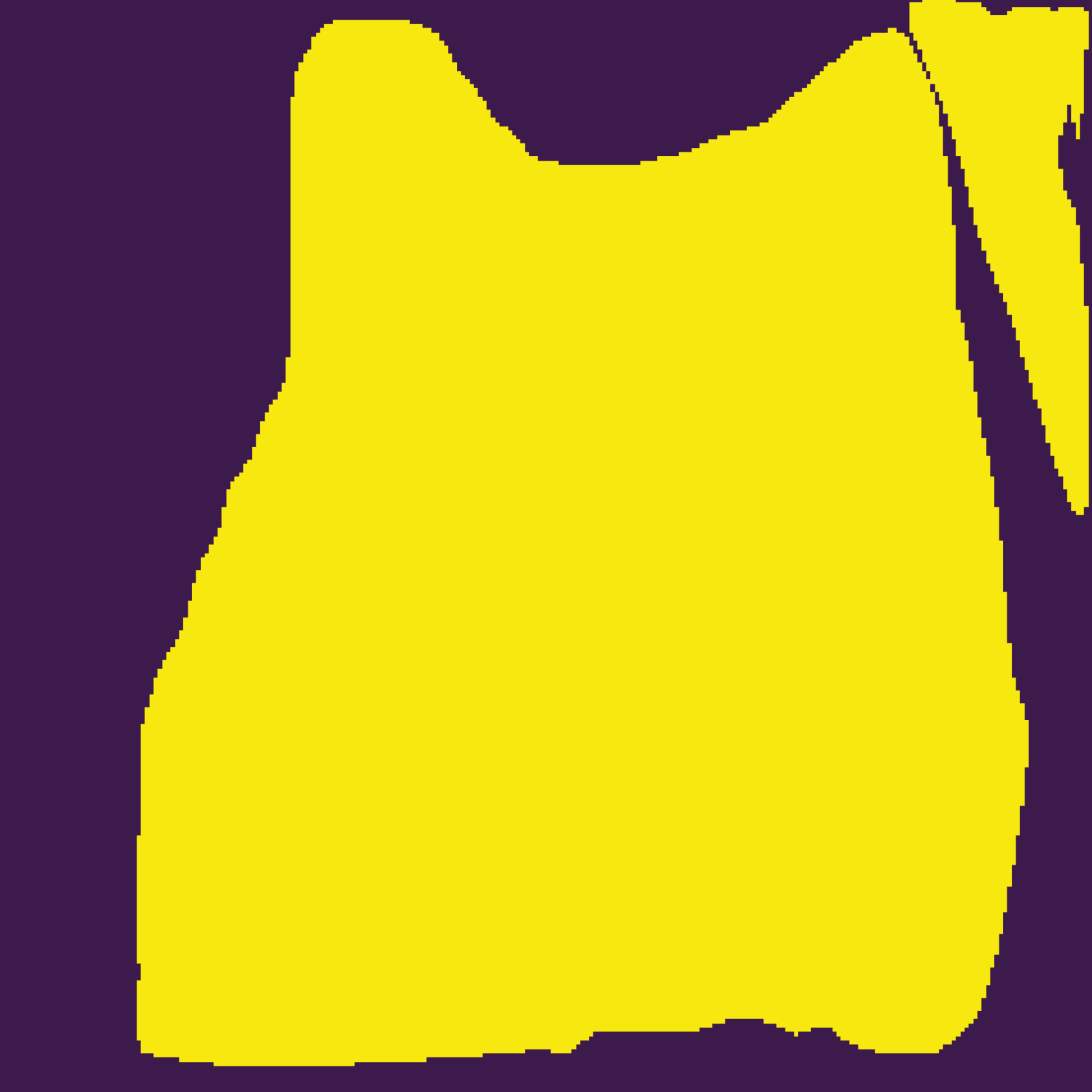}
\end{subfigure}
\begin{subfigure}{.09\textwidth}
  \centering
  \includegraphics[width=1.0\linewidth]{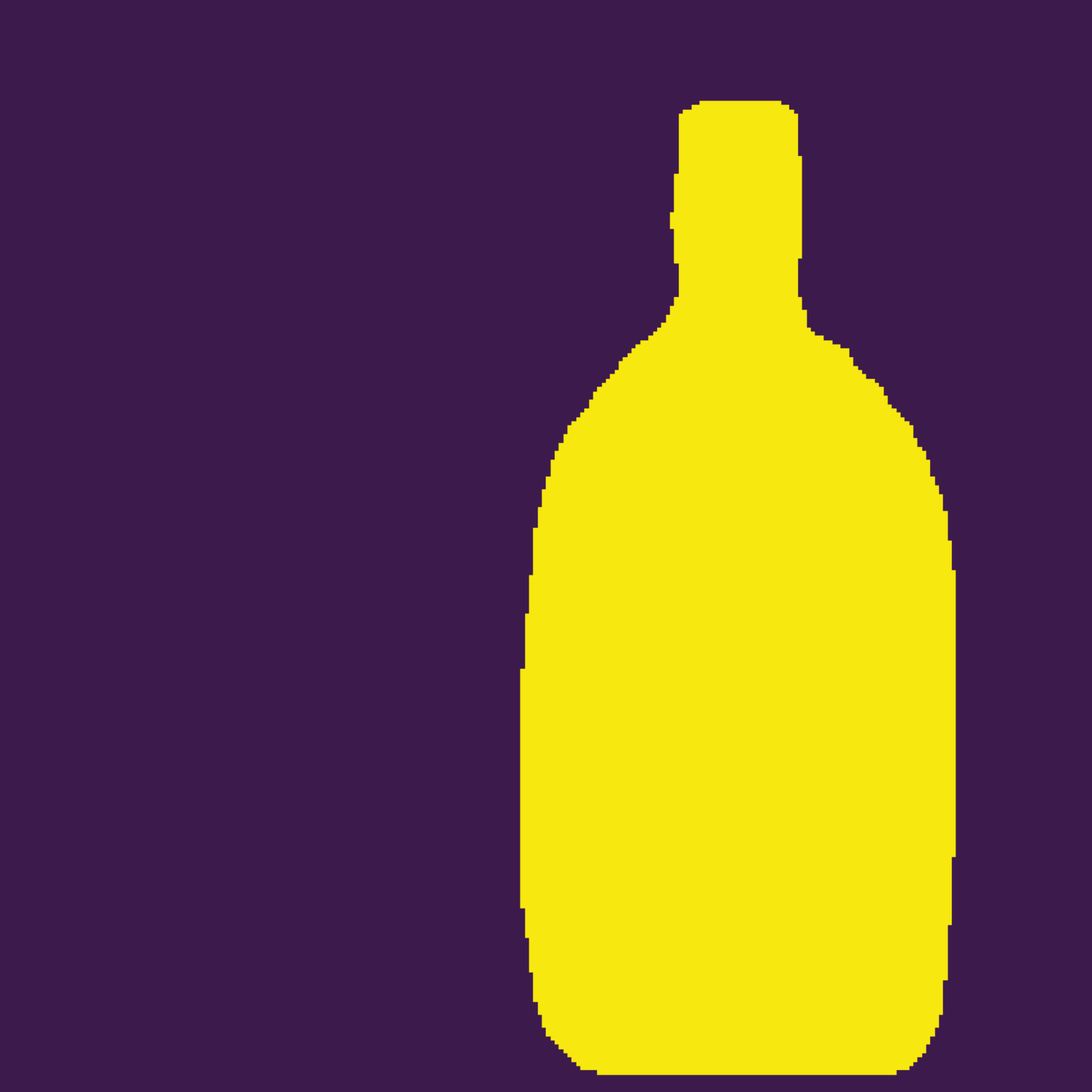}
\end{subfigure}
\begin{subfigure}{.09\textwidth}
  \centering
  \includegraphics[width=1.0\linewidth]{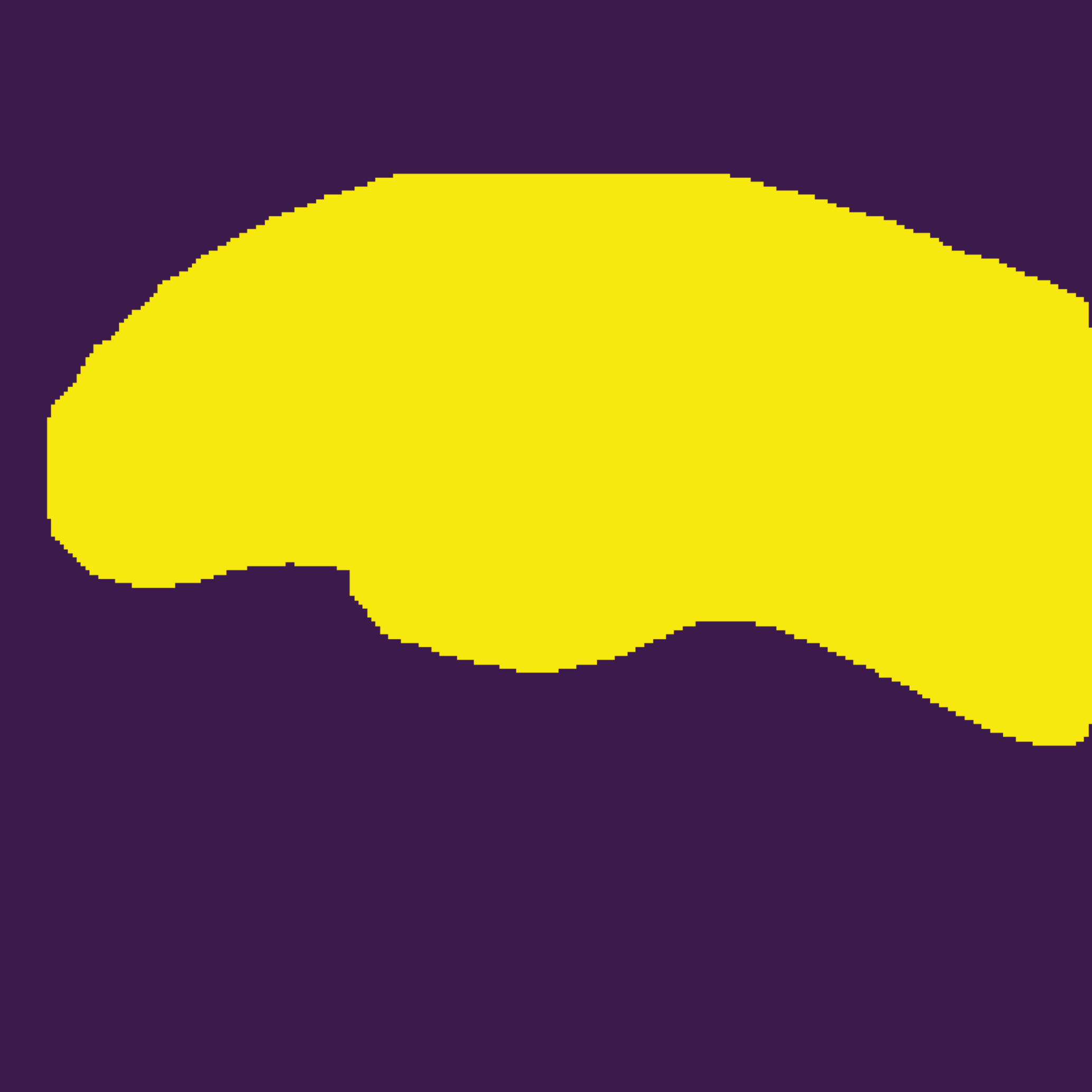}
\end{subfigure} \\
\begin{subfigure}{.09\textwidth}
  \centering
  \includegraphics[width=1.0\linewidth]{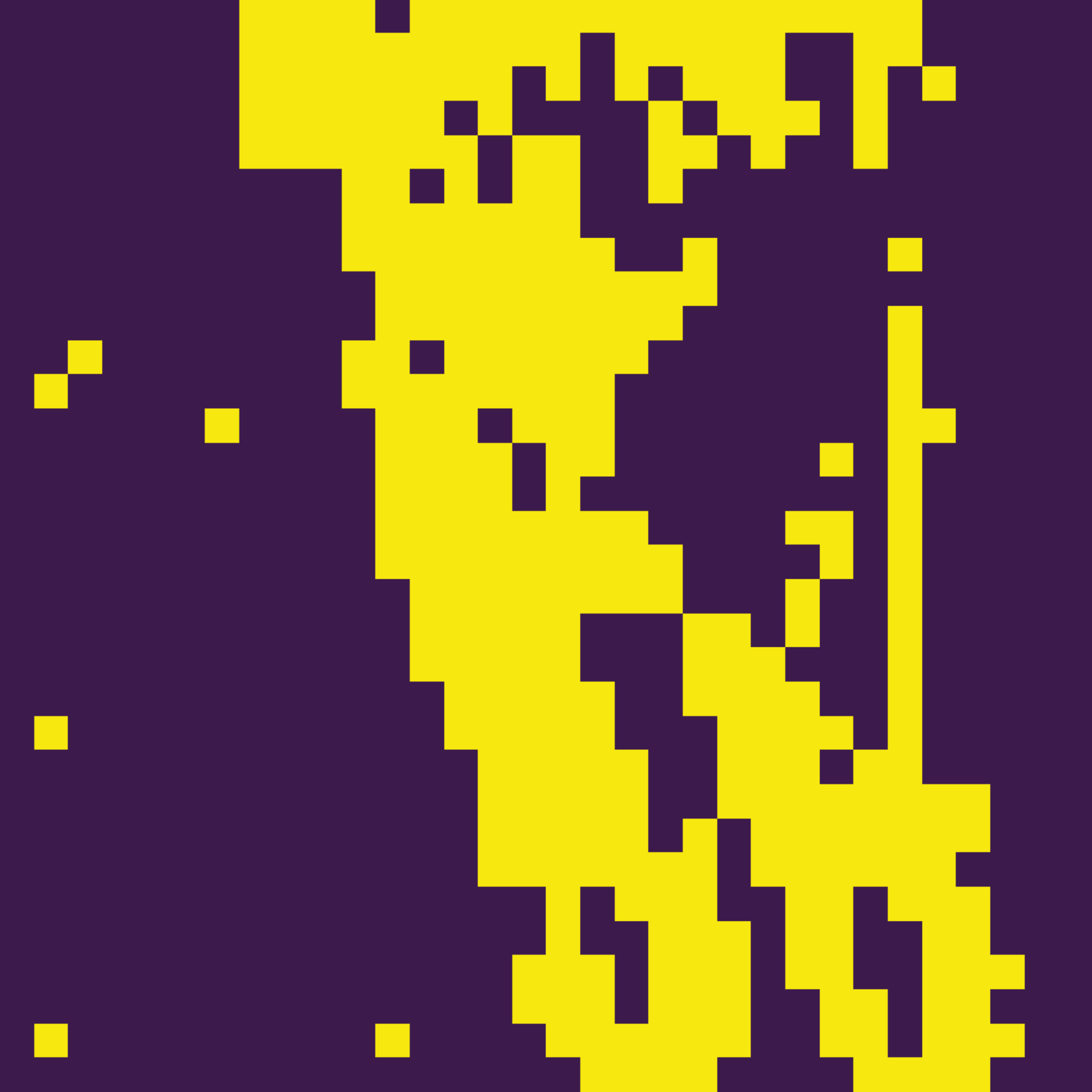}
\end{subfigure}
\begin{subfigure}{.09\textwidth}
  \centering
  \includegraphics[width=1.0\linewidth]{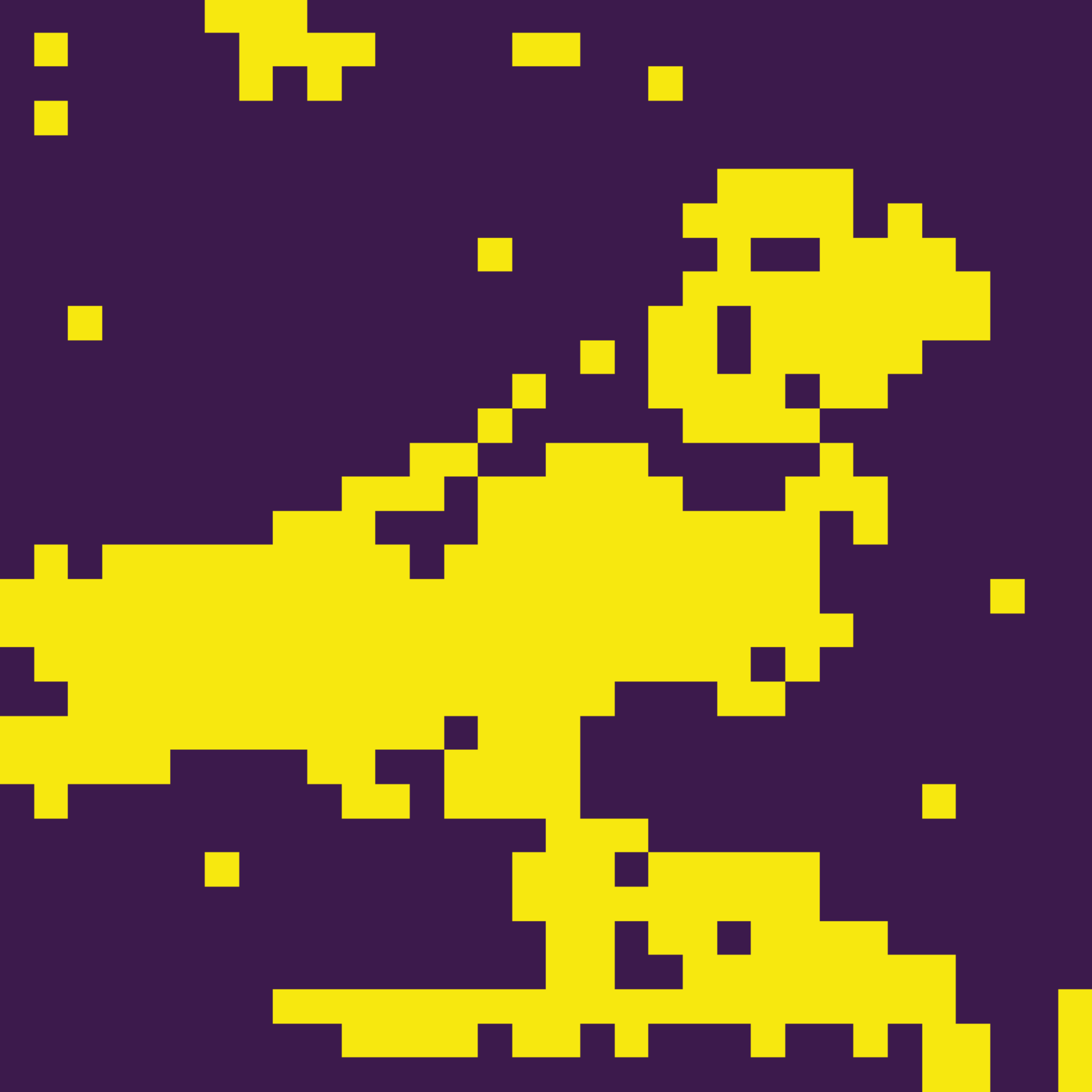}
\end{subfigure}
\begin{subfigure}{.09\textwidth}
  \centering
  \includegraphics[width=1.0\linewidth]{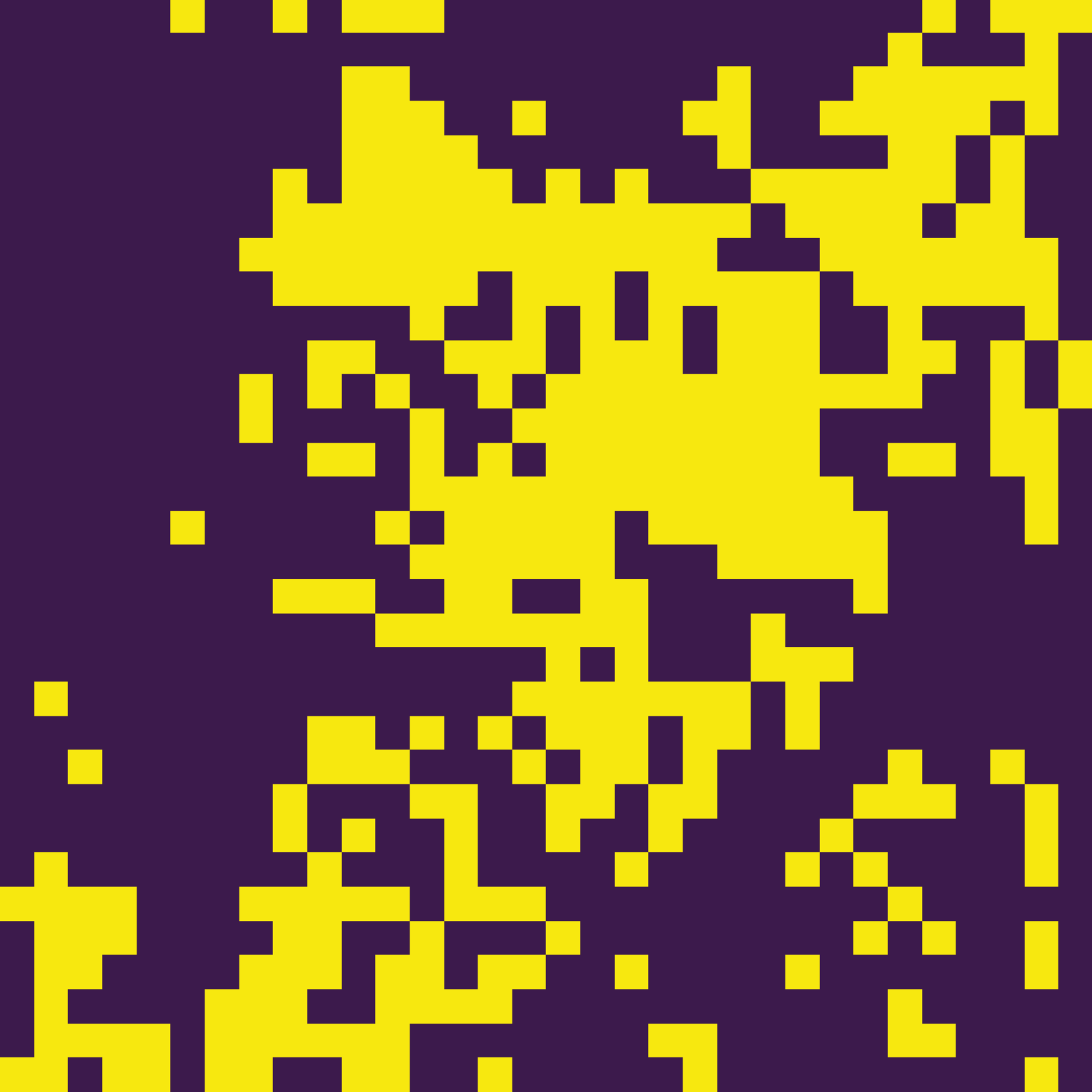}
\end{subfigure}
\begin{subfigure}{.09\textwidth}
  \centering
  \includegraphics[width=1.0\linewidth]{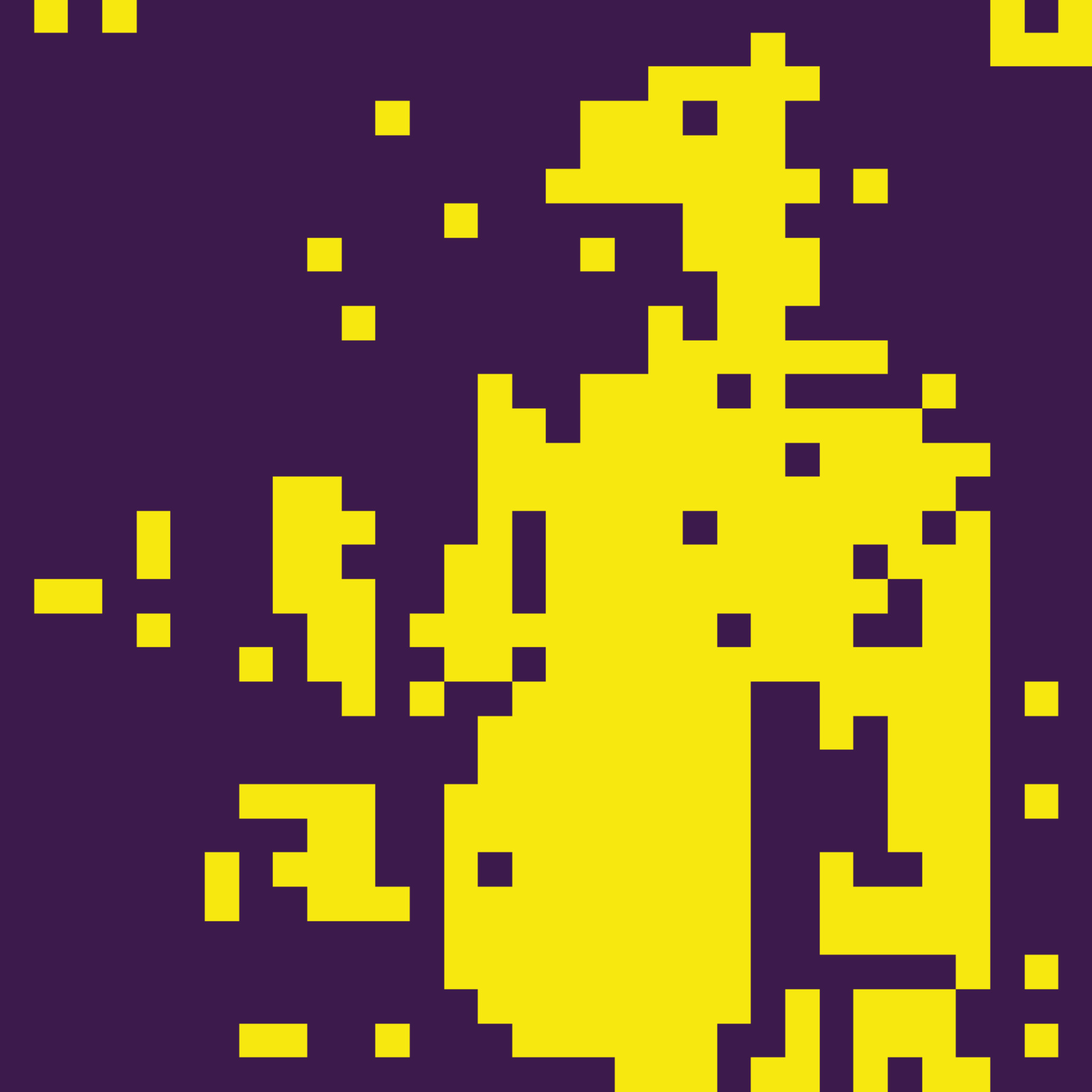}
\end{subfigure}
\begin{subfigure}{.09\textwidth}
  \centering
  \includegraphics[width=1.0\linewidth]{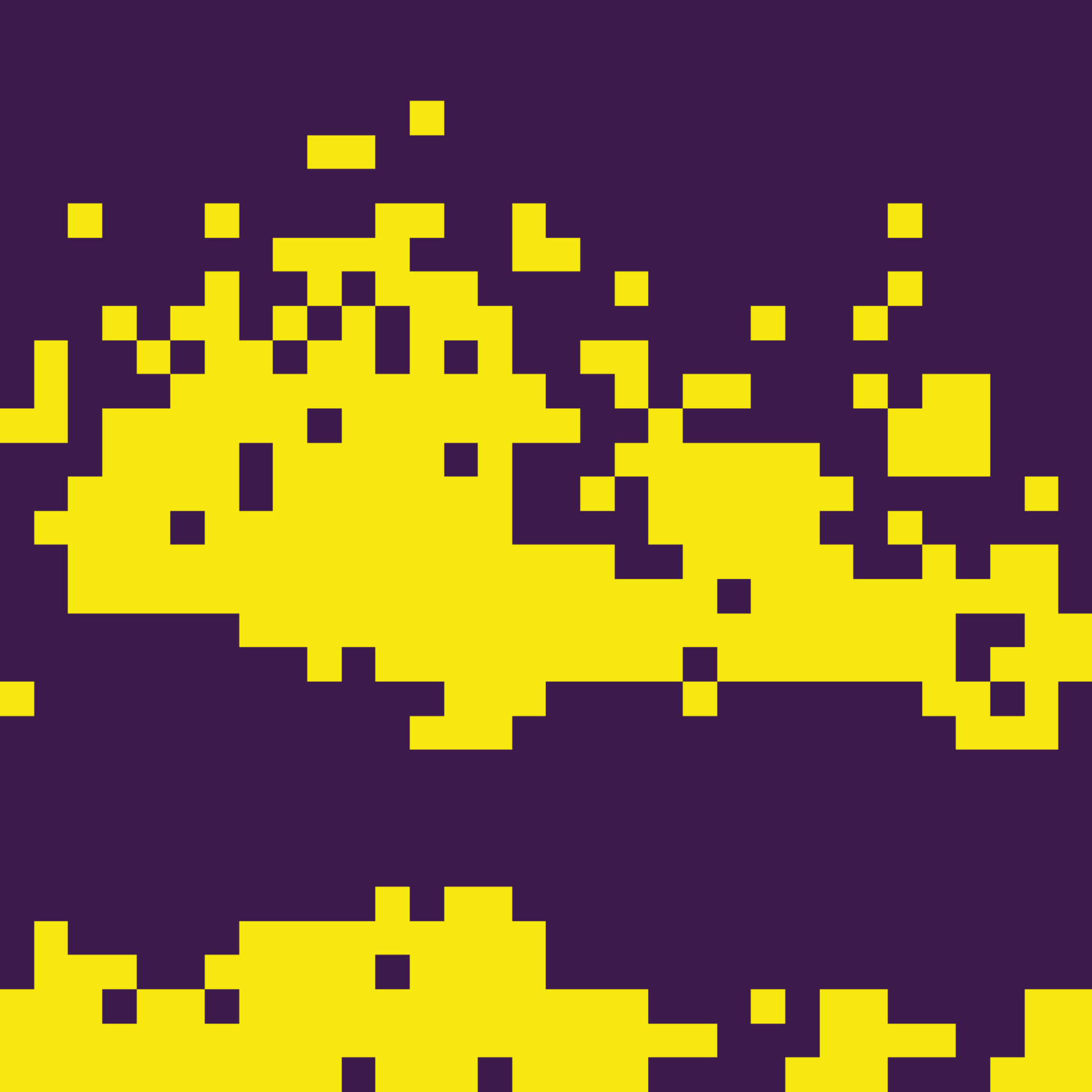}
\end{subfigure} \\
\caption{\textbf{Visualization of self-attention masks compared to object masks.} Generated images (top row), the object masks of Mask R-CNN~\cite{he2017mask} (middle row), and the self-attention masks of unconditional ADM~\cite{dhariwal2021diffusion} (bottom row).}
\label{fig:iou}
\end{figure}

Based on the observation, we are interested in two aspects that the self-attention of diffusion models attends to: the frequency and the semantics of the samples. Therefore, we first investigate how the self-attention maps correlate with frequency by comparing the frequency spectra of patches with high attention scores to those of all patches. We observe that high-attention patches contain more high-frequency details (Fig.~\ref{fig:freq_vis}). We then evaluate how the self-attention maps align with foreground objects (Table~\ref{tab:iou} and Fig.~\ref{fig:iou}) and discover that they capture some semantic information at all resolutions.

\subsection{Additional ablation studies}
We conduct experiments on the threshold of self-attention masking that affects the ratio of the blurred region with 10k samples. We test the thresholds of $0.7, 1.0$, and $1.3$. As shown in Table~\ref{tab:abl_masking_thres}, the highest metrics are obtained when the threshold value is $1.0$.

Table~\ref{tab:abl_attn_layer} shows evaluation results with respect to the attention map extraction layers, evaluated using 10k samples. We select the last self-attention layers of each resolution from the encoder and decoder, and also include the bottleneck layer that divides the encoder and decoder. Regardless of the extraction layer, performance consistently improves over the baseline, while utilizing the self-attention of the final layer yields the best FID and IS results.

\begin{table}[t]
\centering
\small
\begin{tabular}{c|c|c|c|c}
\toprule
Patch size & $\psi$ & Random & Self-attn. & \% Diff.\\
\midrule
\multirow{2}{*}{8$\times$8} & 1.0 & 0.16 & 0.23 & + 44\%\\
& 1.3 & 0.09 & 0.14 & + 56\%\\
\midrule
\multirow{2}{*}{16$\times$16} & 1.0 & 0.18 & 0.25 & + 39\%\\
& 1.3 & 0.05 & 0.11 & + 120\%\\
\midrule
\multirow{2}{*}{32$\times$32} & 1.0 & 0.18 & 0.26 & + 44\%\\
& 1.3 & 0.04 & 0.10 & + 150\%\\
\bottomrule
\end{tabular}
\caption{\textbf{Semantic analysis of the self-attention masks.} $\psi$ denotes the masking threshold, and \% Diff. denotes the percentage difference of the IoU over the random counterpart.}
\label{tab:iou}
\end{table}

\begin{table}[t]
\centering
\small
\begin{tabular}{c|c|ccc}
\noalign{\smallskip}\noalign{\smallskip}\toprule
$\psi$ & Baseline & $\psi=0.7$ & $\psi=1.0$ & $\psi=1.3$\\
\midrule
FID ($\downarrow$) & 5.98 & 5.67 & \textbf{5.47} & 5.66 \\
IS ($\uparrow$)& 141.72 & 148.60 & \textbf{151.12} & 145.58 \\
\bottomrule
\end{tabular}
\caption{\textbf{Ablation study of the masking threshold ($\psi$).} The results are derived from ADM trained on ImageNet 128$\times$128.}
\label{tab:abl_masking_thres}
\end{table}

\begin{table}[t]
\centering
\small
\begin{tabular}{c|c|cc|c|ccc}
\noalign{\smallskip}\noalign{\smallskip}\toprule
Layer & Baseline & In.~11 & In.~8 & Mid. & Out.~2 & Out.~5 & Out.~8\\
\midrule
FID ($\downarrow$) & 5.98 & 5.54 & 5.61 & 5.63 & 5.59 & 5.57 & \textbf{5.47} \\
IS  ($\uparrow$) & 141.72 & 150.07 & 148.20 & 143.44 & 150.62 & 141.73 & \textbf{151.12} \\
\bottomrule
\end{tabular}
\caption{\textbf{Ablation study of the layer where we extract the attention map.} The results are derived from ADM trained on ImageNet 128$\times$128. We denote the middle block as \textit{Mid.}, and the $n$th layer of the input and output blocks as \textit{In.~$n$} and \textit{Out.~$n$}, respectively.} 
\label{tab:abl_attn_layer}
\end{table}

\subsection{Qualitative results}

\begin{figure*}[p]
    \centering
    \includegraphics[width=0.8\linewidth]{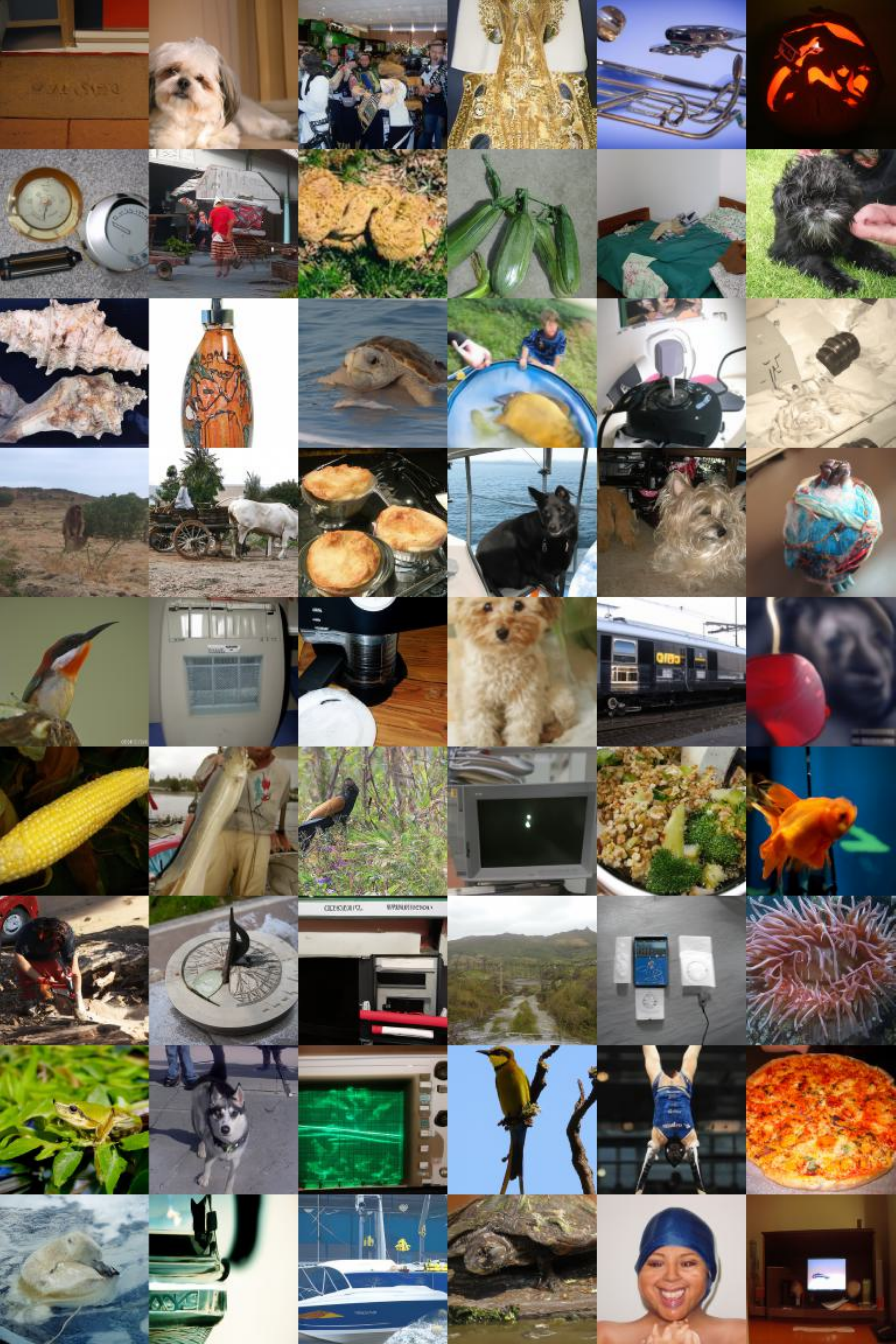}
    \caption{\textbf{Uncurated samples with our method.} The results are sampled from ADM~\cite{dhariwal2021diffusion} conditionally pre-trained in ImageNet~\cite{deng2009imagenet} 128$\times$128 with self-attention and classifier guidance in combination.}
    \label{fig:unsel_128_0}
\end{figure*}

\begin{figure*}[p]
\centering
\includegraphics[width=1.0\linewidth]{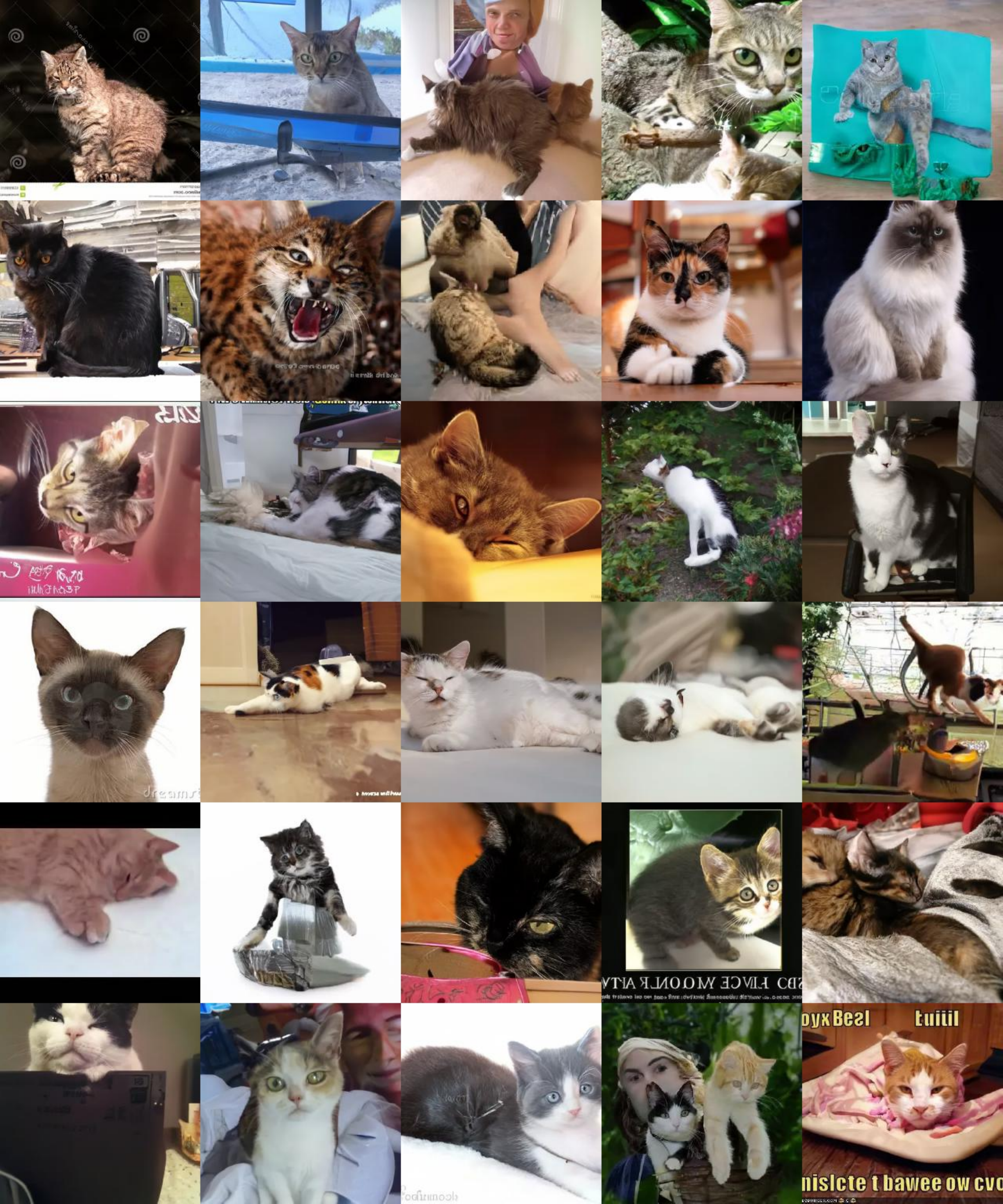}
\caption{\textbf{Uncurated samples with our method.} The results are sampled from ADM~\cite{dhariwal2021diffusion} pre-trained in LSUN Cat~\cite{yu2015lsun} with self-attention guidance.}
\label{fig:lsun_cat}
\end{figure*}

\begin{figure*}[p]
\centering
\includegraphics[width=1.0\linewidth]{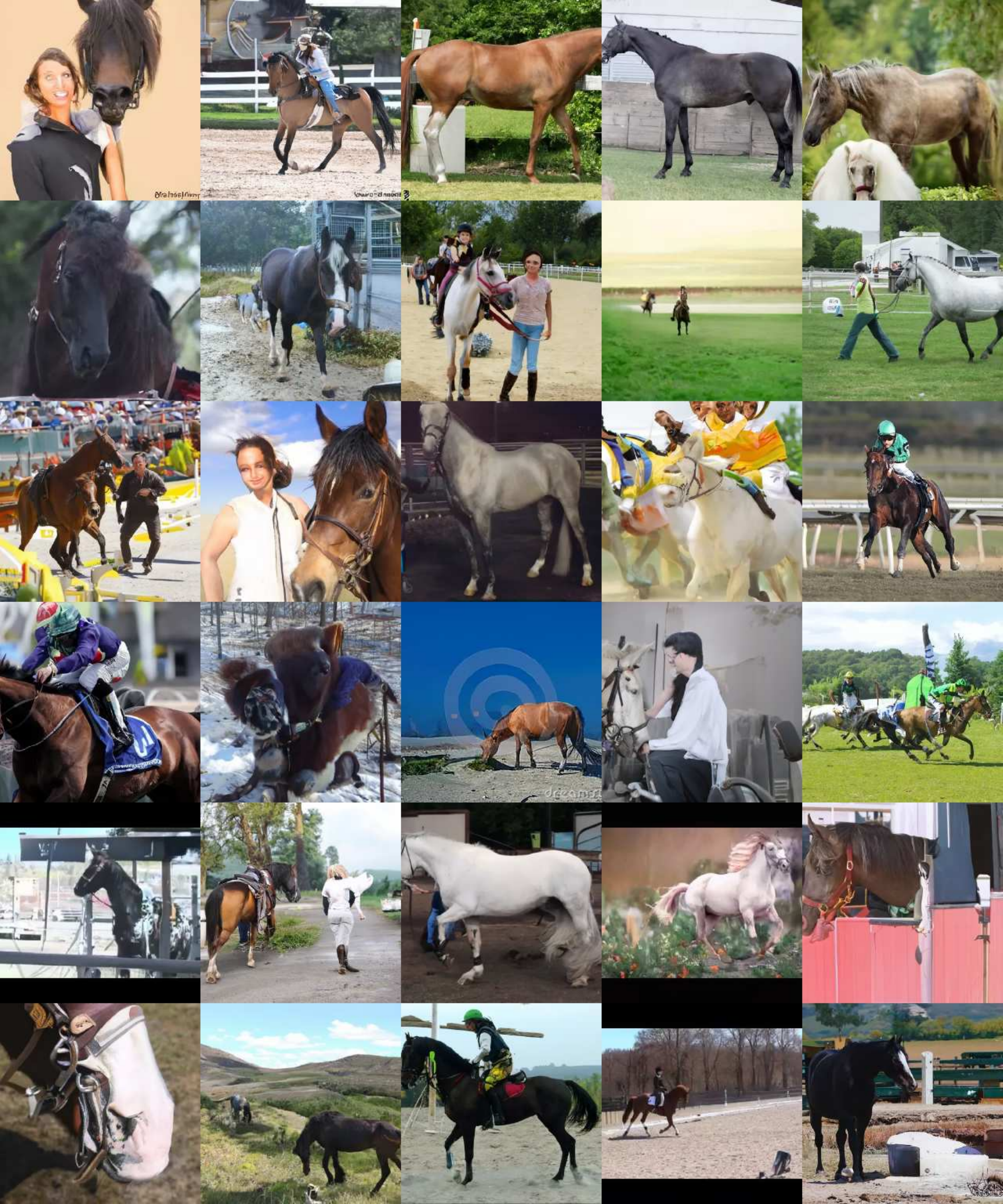}
\caption{\textbf{Uncurated samples with our method.} The results are sampled from ADM~\cite{dhariwal2021diffusion} pre-trained in LSUN Horse~\cite{yu2015lsun} with self-attention guidance.}
\label{fig:lsun_horse}
\end{figure*}

In addition to the samples in the main paper, we present random samples with SAG from ADM pre-trained with ImageNet 128$\times$128 (Fig.~\ref{fig:unsel_128_0}), LSUN Cats (Fig.~\ref{fig:lsun_cat}), and LSUN Horse (Fig.~\ref{fig:lsun_horse}).

\newpage

\section{Human Evaluation Protocol}
\begin{figure}[t]
 \centering
 \includegraphics[width=0.75\linewidth]{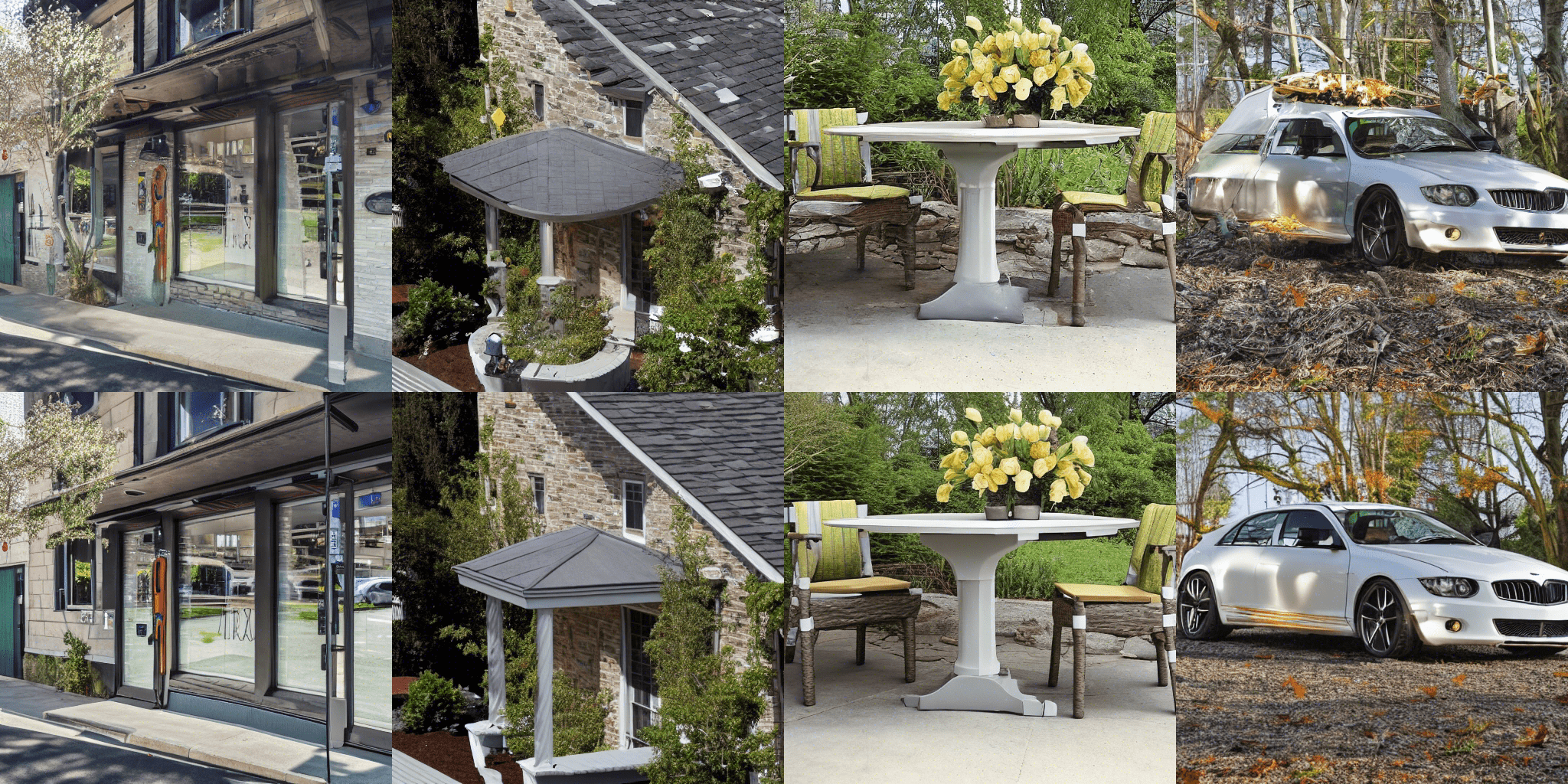}\\
 Which row do you think shows the better image quality? 1) The top row 2) The bottom row
 \caption{\textbf{An example of a question.} The participants are not told which row is sampled with our method.}
 \label{fig:user-ex}
\end{figure}

For the human evaluation of SAG with samples from Stable Diffusion~\cite{rombach2022high}, we generate 500 pairs with the empty prompt with or without SAG, and the SAG scale is $1.0$ for the samples with SAG. Each pair shares the same seed to make it comparable. We show 50 participants 2 groups of 4 samples, one with SAG and the other without SAG, and ask the participants to select a group having higher image quality. An example of a question is in Fig.~\ref{fig:user-ex}. Neither the pairs are cherry-picked nor filtered. We also do not perform any post-processing with the responses.

\section{Limitations \& Future Work}
While the increased self-conditioning typically yields results that are more visually appealing to humans, it is important to consider the perspective that the generated images may lack diversity and novelty, a topic that requires discussion. However, at the present stage, the impact of SAG can be effectively moderated by controlling its guidance scale, leading to beneficial applications. Additionally, it requires twice as many feedforward steps, a challenge that is common to CFG~\cite{ho2021classifier} and necessitates addressing. A possible solution might involve distilling guidance into diffusion models~\cite{meng2022distillation}. This could potentially lessen the computational cost associated with both SAG and CFG, without sacrificing quality.

Moreover, self-attention-based guidance may be more suitable for discrete diffusion models~\cite{tang2022improved, gu2022vector}, which directly model token probabilities instead of approximating them with continuous values. The integration of these models with our method presents an intriguing topic for future research.

{}

{}

{}

\clearpage

\end{document}